\documentclass[twoside,11pt]{article}
\usepackage{jmlr2e}

\usepackage{hyperref}
\usepackage[normalem]{ulem}
\usepackage{amsmath}

\usepackage[T1]{fontenc}    
\usepackage[utf8]{inputenc} 
\usepackage{mwe}
\usepackage{times}
\usepackage{helvet}

\usepackage{amsfonts}
\usepackage{amssymb}
\usepackage{bm}
\usepackage{thmtools}
\usepackage{thm-restate}
\usepackage{csquotes}

\usepackage{booktabs}       
\usepackage[table]{xcolor}
\usepackage{graphicx}

\usepackage{algorithm}
\usepackage{algpseudocode}

\usepackage{wrapfig}
\usepackage{mathtools}
\usepackage{multirow}
\usepackage{natbib}
\usepackage{xfrac}
\usepackage{nicefrac}
\usepackage{paralist}
\usepackage{subcaption}
\usepackage{array}
\usepackage{pgf}
\usepackage{pgfplots}
\pgfplotsset{compat=1.15}
\usepackage{tikz}
\usepackage{pifont}
\usepackage{soul}
\usepackage{url}
\usepackage{verbatim}
\usepackage{enumitem}
\usepackage{stmaryrd}
\usepackage{mathrsfs}  
\usepackage{dsfont}
\usepackage{makecell}
\usepackage[export]{adjustbox}
\usepackage[figuresright]{rotating}
\usepackage{stackengine}

\usepackage[off]{editing}

\usetikzlibrary{positioning,shapes,arrows,calc,chains,shapes.geometric}
\usetikzlibrary{arrows.meta, positioning,fit}
\usetikzlibrary{trees}
\tikzset{%
 >={Latex[width=1mm,length=1.5mm]},
  vertex/.style={draw,circle,inner sep=0mm,minimum width=5mm,font=\scriptsize},
  clique/.style={draw,circle,inner sep=1pt},
  decision/.style={draw,regular polygon,regular polygon sides=4,inner sep=0pt,font=\scriptsize},
  utility/.style={draw,diamond,inner sep=1pt,font=\scriptsize},
    bidir/.style={<->,dashed},
    dir/.style={->},
  smvertex/.style={draw,circle,inner sep=0mm,minimum width=4mm,font=\scriptsize},
  uvertex/.style={draw,circle,dashed,inner sep=0mm,minimum width=5mm,font=\scriptsize},
  action/.style={shape=circle,fill=black,inner sep=0pt,minimum size=3pt,draw},
  regime/.style={shape=rectangle,fill=black,inner sep=0pt,minimum size=3pt,draw},
  node distance=1cm
}

\tikzstyle{agent} = [rectangle, rounded corners, minimum width=1.5cm, minimum height=2cm, text centered, draw=black]
\tikzstyle{environment} = [rectangle, rounded corners, minimum width=2cm, minimum height=2cm, text centered, draw=black]
\tikzstyle{arrow} = [->,>=Latex]

\tikzstyle{SCM}=[>={Stealth},
			every node/.style={inner sep=0.25em},
			conf-path/.style={<->,dashed,
            	every edge/.append style={in=100,out=80}
            },
            removed/.style={gray!30},
            sel-node/.style={rectangle,inner sep=0.2em},
			s-node/.style={double distance=0.5mm, circle}]

\newcommand{\convexpath}[2]{
  [   
  create hullcoords/.code={
    \global\edef\namelist{#1}
    \foreach [count=\counter] \nodename in \namelist {
      \global\edef\numberofnodes{\counter}
      \coordinate (hullcoord\counter) at (\nodename);
    }
    \coordinate (hullcoord0) at (hullcoord\numberofnodes);
    \pgfmathtruncatemacro\lastnumber{\numberofnodes+1}
    \coordinate (hullcoord\lastnumber) at (hullcoord1);
  },
  create hullcoords
  ]
  ($(hullcoord1)!#2!-90:(hullcoord0)$)
  \foreach [
  evaluate=\currentnode as \previousnode using \currentnode-1,
  evaluate=\currentnode as \nextnode using \currentnode+1
  ] \currentnode in {1,...,\numberofnodes} {
    let \p1 = ($(hullcoord\currentnode) - (hullcoord\previousnode)$),
    \n1 = {atan2(\y1,\x1) + 90},
    \p2 = ($(hullcoord\nextnode) - (hullcoord\currentnode)$),
    \n2 = {atan2(\y2,\x2) + 90},
    \n{delta} = {Mod(\n2-\n1,360) - 360}
    in 
    {arc [start angle=\n1, delta angle=\n{delta}, radius=#2]}
    -- ($(hullcoord\nextnode)!#2!-90:(hullcoord\currentnode)$) 
  }
}

\pgfdeclarelayer{back}
\pgfsetlayers{back,main}

\makeatletter
\pgfkeys{%
  /tikz/on layer/.code={
    \def\tikz@path@do@at@end{\endpgfonlayer\endgroup\tikz@path@do@at@end}%
    \pgfonlayer{#1}\begingroup%
  }%
}
\makeatother

\newcolumntype{P}[1]{>{\centering\arraybackslash}p{#1}}

\newcommand{\E}{\mathbb{E}}

\newcommand\bidirectarrow{\dashleftarrow \dasharrow}

\DeclareMathOperator*{\argmax}{arg\,max}

\definecolor{betterred}{RGB}{228,26,28}
\definecolor{betterblue}{RGB}{55,126,184}
\definecolor{bettergreen}{RGB}{77,175,74}
\definecolor{betterpurple}{RGB}{152,78,163}
\definecolor{bettergray}{RGB}{211, 211, 211}
\definecolor{betteryellow}{RGB}{255,218,185}
\definecolor{LightCyan}{rgb}{0.88,1,1}

\colorlet{ebgreen}{bettergreen!25}
\colorlet{sleered}{betterred!25}
\colorlet{jzyellow}{yellow!25}

\newcommand{\cmark}{\textcolor{bettergreen}{\ding{51}}}%
\newcommand{\xmark}{\textcolor{betterred}{\ding{55}}}%

\newcommand\ci{\protect\mathpalette{\protect\independenT}{\perp}}
\newcommand\Perp{\protect\mathpalette{\protect\independenT}{\perp}}
\newcommand\independent{\protect\mathpalette{\protect\independenT}{\perp}}
\def\independenT#1#2{\mathrel{\rlap{$#1#2$}\mkern3mu{#1#2}}}

\newcommand{\mmid}{\,\middle\vert\,}
\newcommand{\braces}[2][]{#1\{#2 #1\}}
\newcommand{\Braces}[1]{\left\{ #1\right\}}
\newcommand{\brackets}[2][]{#1[#2 #1]}
\newcommand{\Brackets}[1]{\left[ #1\right]}
\newcommand{\parens}[2][]{#1(#2 #1)}
\newcommand{\Parens}[1]{\left(#1\right)}
\newcommand{\verts}[2][]{#1\lvert#2 #1\rvert}
\newcommand{\angles}[2][]{#1\langle#2 #1\rangle}
\newcommand{\Angles}[1]{\left \langle#1 \right\rangle}

\newcommand{\set}[2][]{\braces[#1]{#2}}
\newcommand{\Set}[1]{\Braces{#1}}
\newcommand{\tuple}[2][]{\angles[#1]{#2}}
\newcommand{\Tuple}[1]{\Angles{#1}}
\newcommand{\mli}[1]{\mathit{#1}}

\newcommand{\inv}[2]{P_{#2}\left ( #1\right)}
\newcommand{\invE}[2]{\E_{#2}\left [ #1\right]}
\newcommand{\invEE}[3]{\E_{#2}^{#3}\left [ #1\right]}

\newtheorem{experiment}{Experiment}

\newcommand{\I}{\mathds{1}}
\newcommand{\Ll}{\mathscr{L}}

\newcommand{\VV}{\bm{V}}

\newcommand{\YY}{\bm{Y}}

\newcommand{\D}{\mathscr{D}}

\newcommand{\G}{\mathcal{G}}

\newcommand{\M}{\mathcal{M}}

\newcommand{\PP}{\mathscr{P}}

\newcommand{\CC}{\bm{C}}

\newcommand{\Pa}{\mli{Pa}}
\newcommand{\PA}{\mli{PA}}
\newcommand{\pa}{\mli{pa}}

\newcommand{\De}{\mli{De}}

\newcommand{\An}{\mli{An}}
\newcommand{\an}{\mli{an}}
\newcommand{\de}{\mli{de}}

\newcommand{\Ch}{\mli{Ch}}
\newcommand{\ch}{\mli{ch}}

\newcommand{\doo}{\text{do}}
\newcommand{\ctf}{\text{ctf}}

\DeclarePairedDelimiterX{\infdiv}[2]{[}{]}{%
  #1\; \delimsize \|\;#2%
}

\def\*#1{\boldsymbol{#1}}
\def\1#1{\mathcal{#1}}
\def\2#1{\mathscr{#1}}
\def\3#1{\mathbb{#1}}

\let\oldfootnote\footnote
\def\footnote{\ifhmode\unskip\fi\oldfootnote}

\jmlrheading{1}{2000}{1-48}{4/00}{10/00}{bareinboim22a}{Elias Bareinboim, Junzhe Zhang, and Sanghack Lee}
\ShortHeadings{Causal Reinforcement Learning}{Bareinboim, Zhang, and Lee}
\firstpageno{1}

\begin{document} 
\title{An Introduction to Causal Reinforcement Learning}

\author{\name Elias Bareinboim \email eb@cs.columbia.edu \\
\name Junzhe Zhang \email junzhez@cs.columbia.edu \\
       \addr Department of Computer Science\\
       Columbia University\\
       New York, USA\\
\name Sanghack Lee \email sanghack@snu.ac.kr \\
    \addr Graduate School of Data Science\\
    Seoul National University\\
    Seoul, Republic of Korea }

\maketitle

\begin{abstract} \label{sec:_0_abstract}
Causal inference provides a set of principles and tools that allows one to combine data and knowledge about an environment to reason with questions of counterfactual nature -- i.e., what would have happened had reality been different -- even when no data of this unrealized reality is currently available. 
Reinforcement learning provides a collection of methods to learn a policy that optimizes a specific measure (e.g., reward, regret) when the agent is deployed in an environment and pursues an exploratory, trial-and-error approach. 
These two disciplines have evolved independently and with virtually no interaction between them. 
We note that they operate over different aspects of the same building block, i.e., counterfactual relations, which makes them umbilically connected. 
Based on these observations, we further realize that various novel learning opportunities naturally arise when this connection is explicitly acknowledged, understood, and mathematized.  
To realize this potential, we further note that any environment where the RL agent is deployed can be decomposed as a collection of autonomous mechanisms that lead to different causal invariances and which can be parsimoniously modeled as a structural causal model;  any standard RL setting today is implicitly encoding one of these models. 
This natural formalization, in turn, will allow us to put under a unifying treatment different modes of learning, including online, off-policy, and causal calculus learning, which appear seemingly unrelated in the literature.  
One may surmise that these three standard learning modalities are exhaustive in the sense that all possible counterfactual relations are learnable through their continuous implementation.  
We show that this is not the case by introducing several quite natural and pervasive classes of learning settings that do not fit these modalities but entail novel dimensions and types of analysis. Specifically, we will introduce and discuss through causal lenses the problem of generalized policy learning, where to intervene, imitation learning, and counterfactual learning. 
This new set of tasks and understanding lead to a broader view of counterfactual learning and suggests the great potential for the study of causal inference and reinforcement learning side by side, which we call \textit{causal reinforcement learning} (CRL). 
\end{abstract}

\begin{keywords}
Structural Causal Models, Interventions, Counterfactuals,  Reinforcement Learning, Identifiability, Robustness, Generalizability, Off-policy Evaluation, Imitation Learning. 
\end{keywords}


\section{Introduction}
\label{sec:_1_introduction}

AI will play an increasingly prominent role in society as significant portions of its decision-making infrastructure are being delegated to automated systems.  
The transition from human-based to AI-based decision-making systems is underway and will likely accelerate in the coming years. 
The new generation of AI systems is expected to be more efficient, robust, explainable, generalizable, and, more importantly, to lead to outcomes aligned with society's goals and expectations.
There is a growing understanding that robust decision-making relies on some knowledge of the environment's underlying causal mechanisms.
For instance, an intelligent robot needs to know the cause-and-effect relationships in its environment to plan its course of actions more robustly and communicate them with humans; 
a physician needs to understand the individual and joint effects of multiple available drugs to design an effective strategy for her patients while avoiding unethical experimentation with human subjects and potentially harmful side effects; 
an economist needs to understand the relationship between skill sets and the future job market demands so as to design new training and educational policies more efficiently. 
These examples of everyday decision-making found across society rely on some understanding of the often complex, dynamic, and almost invariably unobserved collection of causal mechanisms. 

One of the primary goals and unifying themes found across Artificial Intelligence (AI) is to develop a \emph{rational agent} that operates in an \emph{environment} capable of maximizing a performance measure, and based on the prior knowledge that the agent has about it \citep{russell2010artificial,sutton1998reinforcement}. 
Operationally, an agent perceives the environment's state through sensory input (e.g., cameras, lidar, API) and then interacts with it through (physical or virtual) \emph{actuators}. 
The agent usually follows what is known as a \emph{policy}, a sequence of decision rules that dictates the action based on the evolving history of its perception and prior decisions. When the underlying system dynamics are provided \textit{a priori} (e.g., in the form of parameters set), one could obtain an optimal policy by applying standard planning algorithms \citep{bellman57,puterman1994markov,sutton1998reinforcement,bertsekas2005dynamic}. Effective planning methods in structured environments are studied under the rubrics of influence diagrams \citep{shachter1986evaluating,lauritzen2001representing,koller2003multi}. In many practical applications, however, the parameters of the real, underlying environment are not fully known, which entails some learning processes. 

Reinforcement learning (RL) has become the \emph{de facto} framework for reasoning about optimal decision-making under uncertainty in AI and machine learning over the last decades. RL methods could generally be categorized according to the types of interactions invoked during the learning stage between the agent and the environment. First, there is the modality of \emph{off-policy learning} where the agent learns an optimal policy from offline data generated by a different behavior policy or agent. Second, there is \emph{online learning}, where the agent directly deploys policies in the actual environment and observes subsequent outcomes. Several RL algorithms have been proposed on the formalism (and the corresponding assumptions) known as Markov Decision Processes (MDPs), where a finite set of state variables is statistically sufficient to summarize the treatments and covariates history \citep{bellman57,puterman1994markov,sutton1998reinforcement,bertsekas2005dynamic,jaksch2010near}. 
There are a number of variations of this setting -- both special cases and generalizations -- including multi-armed bandits (MABs)  \citep{thompson1933likelihood,robbins52,lai1985asymptotically,Auer2002}, contextual banduits \citep{auer2002nonstochastic,langford2008epoch}, and partially-observable MDP (POMDPs)  \citep{lovejoy1991survey,singh1994learning,jaakkola1995reinforcement,littman1995learning}, 

As noted by Bellman, the  ``curse of dimensionality'' is a pervasive challenge found in the RL literature, considering the exponential growth in the state-action space when solving the corresponding dynamic programming problems  \citep{bellman57}. 
Since real-world problems often involve more than a few variables, it would appear unlikely that RL methods could be applied to practical settings of higher significance. 
Fortunately, the real world is often structured and modular, and components are usually independent of each other in most systems. Hierarchical RL leverages the independence relationships and allows the decomposition of the MDP model into a hierarchy of simpler MDPs representing subtasks \citep{dietterich2000hierarchical,guestrin2001multiagent,guestrin2003efficient,kulkarni2016hierarchical}. Consequently, by implicitly exploiting the modularity following from the causal mechanisms, the computational complexity of solving RL tasks could be significantly reduced. More recently, the use of deep learning automates the learning of lower-dimensional and hierarchical representations in RL, including deep Q-networks \citep{mnih2015human}, AlphaGo \citep{silver2016mastering}, and IBM's Watson DeepQA system \citep{ferrucci2010building}.

Given the compelling results obtained so far, one may surmise that the foundational picture is essentially complete. 
In other words, all decision-making problems would eventually be solved given additional resources -- more data, more memory, and more computation. 
We will argue in this part of the book that this is not the case. 
In fact, researchers are becoming increasingly aware that AI systems deployed in the real world are very fragile,  brittle, and are commonly sample inefficient (``data-hungry''). 
They also lack robustness and generalizability capabilities, and are opaque, being neither interpretable nor explainable, and subsequently, not trustworthy from a human perspective. 
This observation does not constitute a fundamental impediment to their evolution, of course, but is a simple realization of the current state of affairs, which is a necessary first step toward finding a solution. 
As it will become apparent throughout our exposition, underlying these challenging issues is the lack of an explicit language capable of accounting for the causal mechanisms and exogenous sources of variations that amount to the environment where the agent is deployed, and fundamentally generate the invariances, rewards, and dynamics of the system. 

Interestingly, the field of causal inference (CI) provides a set of principles and tools that allows one to combine data and structural invariances about the environment to reason about questions of counterfactual nature, i.e., what would have happened had reality been different, even when no data about this imagined reality is available, and the environment is not fully observable. 
The conditions under which the effect of an action can be computed from observational and experimental data have been extensively studied. 
Several conditions and algorithms have been proposed based on qualitative knowledge about the environment (Part II). 
Notable results exist throughout the empirical sciences \citep{cornfield:51,Flegal2005,Heckman2006science,Smoking2014}, and more recently, in machine learning \citep{kallus2018confounding,namkoong2020off,etesami2020causal,kallus2020confounding,tennenholtz2021bandits} on how to translate causal knowledge to support new policies and principled decision-making. 
However, this area is still in its infancy, requiring substantive work and a significant amount of highly skilled scientists.\footnote{For instance, economists strive to understand the \emph{root causes} of poverty, which could allow, in principle, the design of new policies (i.e., causal interventions) to improve the population's socioeconomic status (SES). 
A considerable body of evidence was accumulated for many decades, notably by the University of Chicago's Professor and Nobel Prize laureate, James Heckman, who demonstrated the effects of early childhood education on families' SES, among other indicators \citep{Heckman2006science}. 
The understanding following this causal link translated to the larger support of early childhood education and a push for new policies aligned with these findings; for example, see Obama's one billion dollar investment \citep{whitehouse2014}.   
There are many such cases throughout the empirical sciences --- e.g., evidence supports that tobacco smoking is one of the determinant factors of lung cancer \citep{cornfield:51,Smoking2014}, or obesity is responsible to shortening life expectancy \citep{Flegal2005}, which, in turn, translated into new public health policies. }
Our goal in AI is to create intelligent systems capable of reasoning and autonomous actions. This requires a transition from a heuristic grasp of the interplay between causal knowledge and decision-making to a deeper and more fundamental understanding of the principles that connect causal evidence and robust decision-making under uncertainty.

\begin{figure}[t]
	\centering
	\begin{tikzpicture}

		\node (env) [environment, fill=gray!10, dashed, minimum height=2cm, text width=2.5cm] at (0, 1) {Environment \\  \textcolor{betterred}{$\M^*$}};

		\node (agent) [agent, minimum height=2cm, text width=2.5cm] at (7, 1) {Agent \\ $\Theta$,  \textcolor{betterred}{$\G$}};

  		\draw [arrow] (agent) -- node[anchor=south] {\small Interaction} (env);
 		\draw[arrow] (env.30) to [bend left = 30] node[anchor=south] {\small State} (agent.150);
  		\draw[arrow] (env.-30) to [bend right = 30] node[anchor=south] {\small Reward} (agent.210);

		\node (scm) [text width = 3.6 cm, text centered] at (0, -1) {\footnotesize Structural Causal Model};
  		\node (diagram) [text width = 2.5cm,  text centered] at (8.5, -1.05) {\footnotesize Causal Diagram};
  		\node (param) [text width = 2.5cm, text centered ] at (5.5, -1) {\footnotesize Parametrization};

   	 	\draw [arrow] (scm) to (0, 0.5);
   	 	\draw [arrow] (param) to (6.5, 0.5);
   	 	\draw [arrow] (diagram) to (7.5, 0.5);
   	 	
	\end{tikzpicture}
	\caption{The Agent-Environment interaction from Causal Reinforcement Learning (CRL).}
	\label{fig_1_crl}
\end{figure}
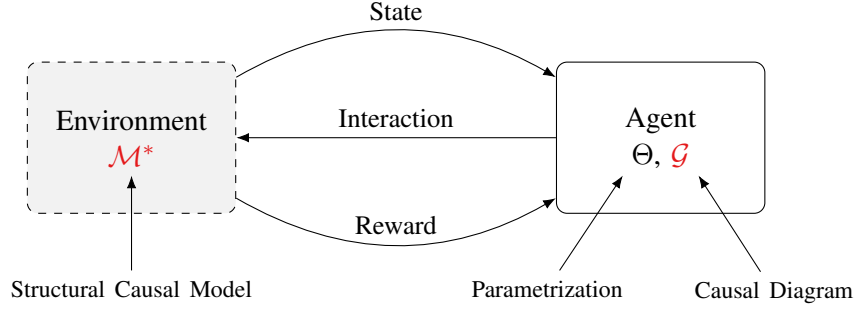
In this paper, we endeavor towards this aim, harnessing the strengths and synergies of CI and RL to devise more sample-efficient, transparent, and robust decision-making systems. 
We refer to this program as \emph{Causal Reinforcement Learning} (CRL). 
Specifically, this chapter seeks to introduce this framework and investigate the intricate and sometimes nuanced relationship between causal knowledge and decision-making.
The design of CRL agents will follow two simple and powerful observations. 
First,  the environment's underlying causal mechanisms should be accounted for explicitly in the analysis. 
In particular, this is realized through formal language and a dual view, as depicted in Fig.~\ref{fig_1_crl}: 
\begin{enumerate}[leftmargin=*]
	\item From the environment's perspective, causal mechanisms and the probability distribution over the exogenous conditions are described as a fully specified SCM $\1M^*$ (on the figure's right side). 
	\item From the agent's perspective, a parsimonious representation of the environment's invariances will be maintained in the form of a causal model  $\G$ (on the left side), such as a causal diagram. 
\end{enumerate}
In essence, while the environment and agent's perspectives differ, they are tied through the pair, an SCM $\M^*$ and its corresponding causal diagram $\G$. 
There is formally a notion of compatibility between these two objects, as discussed later in the book. 
This pairing can be articulated in different forms, including typical template-like structures such as MABs, MDPs, POMDPs. 
\footnote{The current literature evokes this relationship mostly in an implicit fashion, assuming that there exists a match between the environment and the agent knowledge. }

\begin{table}[t]
  \setlength{\tabcolsep}{4.8pt}
  \centering
  {\small
  \begin{tabular}{@{}p{0.5cm}p{2.2cm}p{1.5cm}p{3cm}p{3cm}p{3cm}@{}} \toprule
    & \textbf{Layer \newline (Symbol)} & \textbf{Typical\newline Activity} & \textbf{Quintessential\newline Question} & \textbf{Example} & \textbf{Machine Learning} \\ \midrule
   $\Ll_1$ & Associational \newline $P(y|x)$  & Seeing \newline  & What is? \newline How would seeing $X$ change belief in $Y$? & What does an observation tell us about the underlying state? & Supervised / \newline Unsupervised \newline Learning  \\ \midrule
   $\Ll_2$ & Interventional \newline $P(y|\doo(x))$ & Doing \newline & What if? \newline What if I do $X$? & What if I brake hard, will my vehicle avoid an accident? & Reinforcement\newline Learning  \\ \midrule
   $\Ll_3$ & Counterfactual \newline $P(y_{x} | x', y')$ & Imagining\newline  & Why?  \newline  What if I had acted differently? & Was it the hard brake that prevented the accident? & Explanation \newline Transparency \\ \bottomrule
  \end{tabular}}
  \caption{Summary of Pearl's Causal Hierarchy including each of its layers, the symbolic representation, typical activities and questions, and examples where it appears in ML settings.}
  \label{tab:_1_pch}
\end{table}

The second observation that ground the CRL framework follows from the understanding that every SCM $M^*$ begets a mathematical construct named the \emph{Pearl Causal Hierarchy} (PCH) \citep{pearl2018book}, which has been named in his honor and formalized in \citep{bareinboim2020pearl}. The PCH consists of three qualitatively different types of distributions that are separated into layers -- the associational, the interventional, and the counterfactual. 
The PCH will play a central role in formalizing the types of activities an agent can engage in when considering the environment it has been deployed into, including \emph{seeing}, \emph{doing}, and \emph{imagining}. 
As illustrated in Table~\ref{tab:_1_pch}, knowledge at each layer will allow the agent to reason about different classes of \emph{causal concepts}, or ``queries.'' 
In particular, Layer 1 deals with purely ``observational'', factual information when the agent passively observes the environment (or other agents interacting in the environment). Layer 2 encodes information about what \emph{would} happen, hypothetically speaking, were some interventions were performed, namely, the effects of actions. 
Interestingly, this is a typical activity in RL settings, and answering such queries may be possible from data on interventions already performed or from data collected passively under the first modality in layer 1. These will appear in the form of online and offline learning modalities, discussed later in the text. 
Finally, Layer 3 involves queries about what \emph{would have} happened, counterfactually speaking, had some interventions or actions been performed, given that something else in fact occurred, possibly conflicting with a hypothetical intervention that has not actually happened. The causal hierarchy establishes a useful classification of concepts that might be relevant for a given CRL inference task, thereby also classifying formal frameworks in terms of the questions that agents are able to represent, and ideally answer.

There is a growing literature that investigates various points in the design space of CRL agents and their policies and represents the more concrete examples currently available of this picture, and CRL tasks and inferential machinery \citep{bareinboim2015bandits,forney2017counterfactual,lee2018structural,kallus2018confounding,forney2019counterfactual,lee2019structural,de2019causal,lee2020characterizing,namkoong2020off,etesami2020causal,kallus2020confounding,bennett2021off,wang2021provably,tennenholtz2021bandits,kumor2021causal,ruan2022learning,swamy2022causal}. Still, the treatment provided in each paper represent special cases of specific problems, and were not studied in generality and in a unified manner. In fact, these problems can be seen as a basis on which a CRL agent should be built and will be studied under the same formal umbrella that motivated their very existence. 

\begin{figure}[t]
\resizebox{\linewidth}{!}{
\begin{tikzpicture}[
every node/.style = {draw=black,thick,anchor=west, minimum height=2.5em},
criteria/.style={text centered, text width=3.2cm, fill=gray!30, minimum height=2cm},
attribute/.style={%
    grow=down, xshift=-1.5cm,
    align = left, text width=4cm,
    edge from parent path={(\tikzparentnode.210) |- (\tikzchildnode.west)}},
first/.style    ={level distance=12ex},
second/.style   ={level distance=22ex},
third/.style    ={level distance=32ex},
fourth/.style   ={level distance=42ex},
fifth/.style    ={level distance=52ex},
level 1/.style={sibling distance=15em, level distance = 15ex}]
    \node{Towards Causal Reinforcement Learning}
    [edge from parent fork down]
    
    child{node (sec2) [criteria] {Section \ref{sec:_2_preliminaries} \\ Foundations of Causal Inference}
        child[attribute,first]  {node {\ref{sec:_2_1_scm} Structural Causal Models}}
        child[attribute,second] {node {\ref{sec:_2_2_obs+exp} Observational and Interventional Distributions}}
        child[attribute,third] {node {\ref{sec:_2_3_diagram} Causal Diagrams}}}
  	child{node (sec3) [criteria] {Section \ref{sec:_3} \\ Elements of CRL}
        child[attribute,first]  {node {\ref{sec:_3_1} Causal Decision Models}}
        child[attribute,second] {node {\ref{sec:_3_2} Causal Reinforcement Learning Tasks}}
        child[attribute,third] {node {\ref{sec:_3_3} Comparison with Markov Decision Process}}}
    child{node (sec4) [criteria] {Section \ref{sec:_4} \\ RL through Causal Lenses}
        child[attribute,first]  {node {\ref{sec:_4_1} Online Learning}}
        child[attribute,second] {node {\ref{sec:_4_2} Off-Policy Learning}}
        child[attribute,third]  {node {\ref{sec:_4_3} Causal Learning}}}
    child{node (sec5) [criteria] {Sections \ref{sec:_5_o2o} - \ref{sec:_8_imitation} \\ Novel CRL Tasks}
        child[attribute,first]  {node {\ref{sec:_5_o2o} Offline-to-Online Learning}}
        child[attribute,second] {node {\ref{sec:_6_wheredo} Where to Intervene}}     
        child[attribute,third]  {node {\ref{sec:_7_ctf-rand} Counterfactual Randomization}}
        child[attribute,fourth] {node {\ref{sec:_8_imitation} Causal Imitation Learning}}};
\end{tikzpicture}}
\caption{Paper's roadmap and organization.} \label{fig:_1_1_roadmap}
\end{figure}
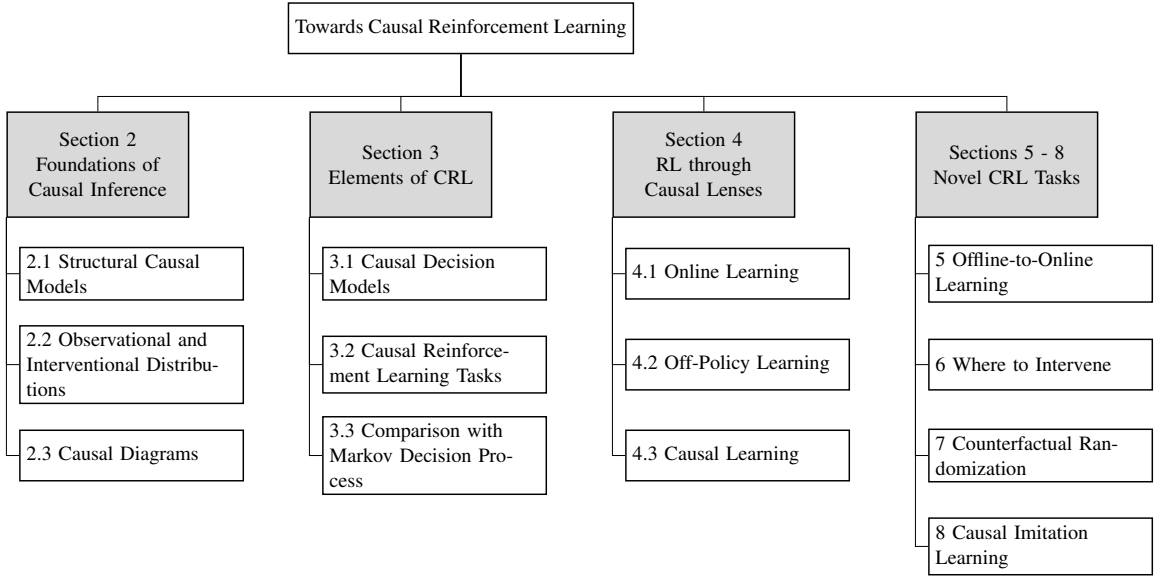

\subsection{Roadmap of the Paper} 
The remainder of the paper is organized as shown in Fig.~\ref{fig:_1_1_roadmap}. We provide in Sec.~\ref{sec:_2_preliminaries} the necessary background and a logical foundation of causal inference to understand the rest of this paper. We review the definition of structural causal models (Sec.~\ref{sec:_2_1_scm}), evaluation of observational and interventional distributions (Sec.~\ref{sec:_2_2_obs+exp}), and the construction of causal diagrams representing qualitative knowledge in SCMs (Sec.~\ref{sec:_2_3_diagram}). Extensive examples are provided to illustrate these concepts.

Sec.~\ref{sec:_3} is a foundational chapter that connects the different learning modalities found in RL to the causal language introduced in this chapter. 
In particular, Sec.~\ref{sec:_3} formalizes the policy learning problem using the semantic language of SCMs, termed causal decision models (Sec.~\ref{sec:_3_1}). 
Based on this framework, we introduce \textit{causal reinforcement learning tasks} that consider the interaction capabilities of the learning agent and the prior knowledge of the environment accessible to the agent (Sec.~\ref{sec:_3_2}). 
We compare the CRL formalisms with reinforcement learning under the standard model assumptions of Markov decision processes, emphasizing that there exists no discretion here, and causal knowledge is indispensable for solving CRL tasks.

Sec.~\ref{sec:_4} studies classic learning tasks of reinforcement learning and causal inference through the CRL framework, including off-policy learning (Sec.~\ref{sec:_4_1}), online learning (Sec.~\ref{sec:_4_2}), and causal identification (Sec.~\ref{sec:_4_3}). In particular, we discuss several conditions and algorithmic procedures for policy learning for each of these tasks. In the last section, we introduce a graphical criterion that extends off-policy learning methods to the language of structural causality where unobserved confounding is not ruled out a priori.

Sec.~\ref{sec:_5_o2o} considers the problem of causal offline-to-online learning (COOL), where the agent attempts to first pre-train informative representations of optimal policies from offline data and then fine-tune policy estimates by conducting online experimentations. Sec.~\ref{sec:_5_1} introduces a confounding robust procedure for transferring observational data in bandit models. Secs.~\ref{sec:_5_2} and \ref{sec:_5_3} extend this transfer strategy to sequential decision-making settings where the agent has to determine a series of actions in order to maximize the primary outcome (e.g., dynamic treatment regimes).

Sec.~\ref{sec:_6_wheredo} introduces a new task called \textit{mixed policy learning}. This task is concerned with whether the agent should intervene in the system, and, if so, where the intervention should be targeted. 
Sec.~\ref{sec:mixed-policy-with-no-context} investigates the structural properties inherent in a mixed policy space with atomic intervention, where the properties can help the agent to explore the space more efficiently and effectively. Sec.~\ref{sec:mixed-policy-with-context} further investigates a scenario where the agent can conduct soft intervention, selecting which variables to observe for performing soft intervention.

Sec.~\ref{sec:_7_ctf-rand} broadens the scope of policies and introduces a novel \emph{counterfactual decision criterion} that is applicable when agents have their own biases and operate in adversarial environments. Sec.~\ref{sec:_7_1} formalizes the concept of counterfactual policies, enabling agents to perform counterfactual reasoning by accounting for their initial intended actions. Sec.~\ref{sec:_7_2} presents a new type of counterfactual randomization strategy that supports the realization of the counterfactual decision criterion and facilitates the learning of an optimal counterfactual policy. In the last section, we formalize the trade-off between optimality and autonomy under the counterfactual decision criterion and provide a practical planning algorithm to address this trade-off.

Sec.~\ref{sec:_8_imitation} studies the problem of policy learning from the observational data without complete knowledge about the reward function measuring the performance of the agent - called imitation learning. Sec.~\ref{sec:_8_1} develops a complete graphical condition for learning an imitating policy, achieving the expert's performance using behavioral cloning. Sec.~\ref{sec:_8_2} extends this condition to produce a policy that could consistently dominate the expert by exploiting parametric knowledge about the unknown reward function via inverse RL. We also develop an algorithmic approach to apply inverse RL in a more generalized family of SCMs provided with a causal diagram of the environment. 

Finally, Sec.~\ref{sec:_9_conclusions} concludes by summarizing the work and algorithms studied in previous sections and giving final remarks. We also discuss other essential CRL tasks, including transportability, generalizability and model induction, and outlines future challenges in designing CRL agents.

\subsection{Notations}
We introduce the basic notations used throughout this paper. Capital letters represent variables ($X$), and small letters represent their values ($x$). Let $\D(X)$ represent the domain of $X$ and $\PP_X$ the space of probability distributions over $\D(X)$. Boldfaced capital letters $\*X$ denote a collection of variables, $|\*X|$ its dimension, $\D(\*X)$ their joint domains, and boldfaced smaller letters $\*x$ a particular joint realization in the domain $\D(\*X)$. We will consistently use $P(\*X)$ to represent the joint distribution over $\*X$ and $P(\*x)$ represent probabilities $P(\*X = \*x)$; similarly, notation $P(\*Y \mid \*X)$ represents a set of conditional distributions $P(\*Y \mid \*X = \*x)$, $\forall \*x$. Finally, indicator function $\I\{\*Z = \*z \}$ returns $1$ if $\*Z = \*z$ holds true; otherwise $\I\{\*Z = \*z \} = 0$. 

For a directed acyclic graph (DAG) $\1G$, we denote by $\*V(\G)$ the set of vertices in $\1G$; similarly, $\*E(\G)$ is the set of arrows in $\G$. A vertex-induced subgraph is denoted by brackets, e.g., $\G[\*W]$ which includes vertices $\*W \subseteq \*V(\G)$ and edges among its elements. 
For convenience, we define $\G \setminus \*X \equiv \G\left [\*V(\G) \setminus \*X \right ]$. 
For arbitrary subsets $\*W, \*Z \subseteq \*V(\G)$, $\G_{\overline{\*W}\underline{\*Z}}$ is a subgraph obtained from $\G$ by removing edges pointing into any node $W \in \*W$ and edges coming out of any node $Z\in\*Z$. 

Also, we will use standard graph-theoretic abbreviations to represent relationships among nodes: $\an(X)_{\G}$, $\de(X)_\G$, $\pa(X)_\G$ and $\ch(X)_\G$ stand for the set of ancestors, descendants, parents, and children of a node $X$ in a DAG $\G$, not including $X$; subscript $\G$ is omitted when it is obvious. The parent set of a node set $\*X$ is all parents of any node in $\*X$, i.e., $pa(\*X)_\G=\bigcup_{X\in\*X} \pa(X)_\G$; $\de(\*X)_G$, $\an(\*X)_G$, and $\ch(\*X)_G$ are similarly defined. Finally, $\Pa(\*X)_\G$ is the set union $\pa(\*X)_\G \cup \*X$, and so do $\De(\*X)_\G$, $\An(\*X)_\G$, and $\Ch(\*X)_\G$.

\section{Foundations of Causal Inference}
\label{sec:_2_preliminaries}

\begin{figure}[t]
	\centering
	\begin{tikzpicture}
		\filldraw [fill=gray!10, dashed,line width=0.1mm] (-2.5,2.6) -- (-2.5,-0.95) .. controls (2.5,-1) and (2.5,-1) .. (2.5,2.6);
		\draw [color=white, thick]  (-2.5,-1) -- (-2.5,2.6);

		\node (t1) at (0,2.2) {SCM $\M^*$};
		\node (t1) at (0,1.6) {\small{(Unobserved Environment)}};

		\node (Pu) at (0,-0.4) {\footnotesize $U_x, U_y \sim P(U_x, U_y)$};
		\node [inner sep=0.5em] (L1F) at (0,0.7) {\footnotesize $\2F = \begin{cases} X &\gets f_X(U_x)\\ Y &\gets f_Y(X, U_y) \end{cases}$};

		\draw [densely dotted,line width=0.1mm] (-1, -1.6) rectangle (8, -2.4);
		\node [draw, line width=0.08mm, align=center, label=below:{\footnotesize Observational}] (L1Pv) at (0.6,-2) {($\Ll_1$) $P(X,Y)\;$};
		\node [draw, line width=0.08mm, align=center, label=below:{\footnotesize Interventional}] (L2Pv) at (3.3,-2) {($\Ll_2$) $\inv{Y}{x}$};
		\node [draw, line width=0.08mm, align=center, label=below:{\footnotesize Counterfactual}] (L3Pv) at (6.2,-2) {($\Ll_3$) $P(Y_x | x',y')$};

		\path [-Latex] (0, -0.9) edge (L1Pv.north);
		\path [-Latex] (1, -0.85) edge (L2Pv.90);
		\path [-Latex] (2, -0.45) edge (L3Pv.155);

		\node [draw, text height =1.5cm, text width=3.5cm, align=center, label=above:{\small (c) Sec.~\ref{sec:_2_3_diagram}}] (diagram) at (5, 0.7) {};
		\node[text width=3.5cm, align=center] at (5, 1.2) {Causal Diagram $\1G$};
		\node[vertex, minimum width = 5mm] (X) at (4, 0.3) {X};
		\node[vertex, minimum width = 5mm] (Y) at (6, 0.3) {Y};
		\draw[bidir] (X) to [bend left = 45] (Y);
		\draw[dir] (X) -- (Y);

		\path [-Latex] (L1F) edge (diagram);
		\node [draw, text height=0.4cm, text width=4.2cm, align=center] (c1) at (9.6, 1.7) {};
		\node [draw, text height=0.4cm, text width=4.2cm, align=center] (c2) at (9.6, 0.7) {};
		\node [draw, text height=0.4cm, text width=4.2cm, align=center] (c3) at (9.6, -0.3) {};
		\node [align=center, inner sep=0.3em, anchor=west] at (7.5,1.7) {\small I. Templates (MAB, MDP,...)};
		\node [align=center, inner sep=0.3em, anchor=west] at (7.5,0.7) {\small  II. Prior Knowledge };
		\node [align=center, inner sep=0.3em, anchor=west] at (7.5,-0.3) { \small III. Structural Learning };

		\path [-] (diagram) edge (c1.180);
		\path [-] (diagram) edge (c2);
		\path [-] (diagram) edge (c3.180);

		\node at (-0.1,2.9) {\small (a) Sec.~\ref{sec:_2_1_scm}};
		\node [align=center] at (-1.8,-2.3) {PCH \\ \small (b) Sec.~\ref{sec:_2_2_obs+exp}};

	\end{tikzpicture}
 \caption{Building blocks of Causal RL analysis. (a)  Unobserved model of the environment; (b) the PCH (and not necessarily fully observed); (c) The structural constraints over the SCMs that may be elicited through different methods, including prior knowledge and structural learning.  }
	\label{fig:_2_1_ci}
\end{figure}
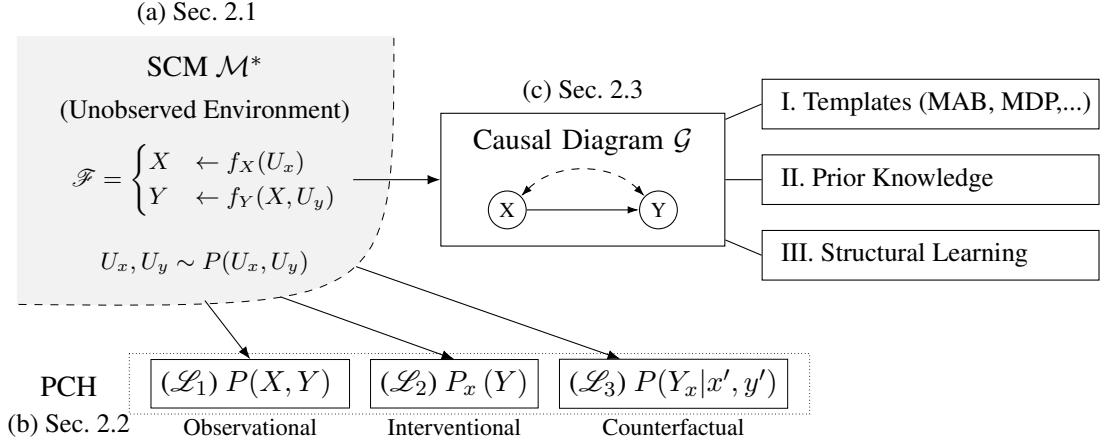

We now provide a brief outline of this section and will lay out the basic CRL building blocks, following the schema shown in Fig.~\ref{fig:_2_1_ci}. 
We will start in Sec.~\ref{sec:_2_1_scm} with the underlying environment, which will be described through formal causal semantics and the definition of SCMs. 
We will further provide some specific examples, or instantiations of some SCMs that follow from canonical examples found in the literature (e.g., MABs, MDPs). 
Each SCM induces the PCH, shown in the bottom of the figure, 
 modeling various interactions an agent can undertake in the environment, including observations, interventions, and counterfactual reasoning. 
In Sec.~\ref{sec:_2_2_obs+exp}, we will formalize the dynamics when the agent is only passively observing the environment -- 
i.e., collecting the PCH's $\Ll_1$-type of data -- which will give rise to what is known as the \textit{observational distribution}. 
We will then move to a more active mode of interaction whenever the agent can perform interventions in the environment, which constitute $\Ll_2$-type of interactions, giving rise to what is known as the \textit{interventional distribution}. In Sec.~\ref{sec:_2_3_diagram}, we discuss specifying structural assumptions about the environment in a non-parametric and parsimonious manner using causal diagrams. 

\subsection{Structural Causal Models and the Environment}\label{sec:_2_1_scm}
 
We build on the language of  \textit{structural causal models}, which is one of the most general and flexible data-generating models known to date \citep{pearl:2k,bareinboim2020pearl}.
The first element of any CRL system is the environment where the agent will be deployed, which will be instantiated as an SCM defined next.

\begin{definition}[Structural Causal Model \citep{pearl:2k}]\label{def:_2_scm}
A structural causal model (SCM, for short) $\M$ is a $4$-tuple $\tuple{\*U, \*V, \2F, P}$ where: 
\begin{itemize}
  \item  A set of background or exogenous variables $\*U = \{U_1, U_2, ..., U_k\}$,
  representing factors outside the model, which
  nevertheless, affect relationships within the model; 
  \item A set of endogenous
  variables $\*V = \{V_1, V_2, ..., V_n\}$ representing variables inside the model. 
  \item  A set $\2F$ of structural functions $\{f_i: V_i \in \*V \}$ s.t. each
  $f_i$ determines the value of $V_i \in \*V$, 
  \begin{eqnarray}
 v_i \gets f_i(\*{\pa}_i, \*u_i), 
  \end{eqnarray} where $\*{\PA}_i\subseteq \*V \setminus \set{V}$ and $\*U_i\subseteq \*U$.
  \item A probability distribution over the exogenous variables, $P(\*U)$.  \hfill $\blacksquare$
\end{itemize} 
\end{definition}
Some observations follow from this definition. 
First, note that a typical SCM $\M$ partitions the variables into two sets -- exogenous (unobserved) $\*U$  and endogenous (observed) $\*V$.
The values of exogenous variables $\*U$ are decided outside the environment where the agent is deployed, following a \emph{probability distribution} $P(\*U)$ over all possible configurations of its states.
In practical settings, these variables represent unaccounted factors of the units and of the situations that affect the environment under consideration; for instance, possibly the patients' DNA, customers' sensitive demographics, robot's physical constraints, and other features and states that affect the system but are unobserved from the perspective of the CRL agent.
For concreteness, consider a  physician who may not have access to the patient's DNA before prescribing a certain treatment. At the same time, such an attribute influences how well the patient will respond to the drug and whether they will recover from their condition. Naturally, other physicians who may have access to this information will use it when making their decisions.
Alternatively, a robot may not be able to determine its precise location in the building, while others with more accurate sensors may have an almost perfect reading of their positions. 
Of course, there is no ``right'' view of reality, and those are alternative perspectives based on the contingencies and capabilities of each agent and the environment where it is deployed. 

More generally, the exogenous set $\*U$ allows for the existence of \emph{unobserved confounding}, which is inherent in any real-world, complex setting in which not every bit of information can be measured or assessed by the agent. 
This is a common phenomenon in environments where humans are present since their existence precludes full observability. 
\footnote{Some research from Neuroscience suggests that decision-making may be a process handled largely by subconscious mental activity \citep{libet1993time}.  Even several seconds before we consciously decide, its outcome can be largely influenced by subconscious activity in the brain. 
In other words, this suggests that humans generally have a lack of understanding of our decision-making process and have a hard time measuring all factors influencing our behaviors; therefore, full observability is rarely realizable in practice.}
Unless otherwise specified, our referential frame would always be from the CRL agent's perspective, which will come with their specific perceptual and interventional ($\Ll_2$) capabilities. 

Second, the value of each endogenous variable $V_i \in \*V$ in $\M$ is determined by a causal mechanism $f_i$, which takes other endogenous and exogenous variables in the system as input. The randomness of exogenous variables $\*U$ induces variations in the endogenous variables $\*V$, which is formalized in the next section. These causal mechanisms, together with the background factors encoded in the distribution $P(\*U)$, represent the data-generating process according to which the environment decides observed states and rewards in each possible configuration. 

One feature of the representation of SCMs is its flexibility: it could contain an arbitrary collection of causal mechanisms. This naturally includes and allows one to model any standard RL environment using the SCM framework, including MABs and MDPs, as illustrated below. One important distinction is that standard RL models leave action variables $\*X$ in the environment uninstantiated. On the contrary, SCMs explicitly model some secondary controlled process associated with actions, such as a human operator or a different source agent demonstrating the task, or even the effect of nature itself (e.g., gravity). Formally, we denote by $f_{\*X} =  \left \{f_X \mid \forall X \in \*X \right \}$ the collection of functional relationships in the system determining values of $\*X$. 
By default, the system is said to be under a \textit{behavior policy}, in RL terminology \citep{sutton1998reinforcement}, or under the natural regime, in CI language  \citep{pearl:95,pearl:2k,dawid:02}.
For example, a physician makes decisions about the patient's treatment in the natural world, regardless of what the AI agent wants to do. Also, the position and momentum of a particle may change with gravity, even though no deliberate (or human) agent interferes in the system at each instant in time. 
The main observation relevant to our context is that, regardless of how $\*X$ attains its value, this does not happen naturally under the control of the CRL agent, which is the one we care about here.  

To start understanding Def.~\ref{def:_2_scm}, we consider some examples showing how some classic, increasingly refined RL environments should be modeled through the semantics of SCMs. We will take a natural regime perspective, which considers the CRL agent (learner) passively observing a different agent (e.g., a teacher, a physical law) making decisions in the environment; these other agents have a possibly different set of perceptual and interventional capabilities. \footnote{Since our goal here is not to discuss the subtleties of multi-agent systems, we will abstract away the identity of these other agents and simply use the notion of an environment. }

\begin{example}[Multi-Armed Bandit \citep{robbins52}]\label{exp:_2_1_mab} 
In a clinic where patients with a chronic disease are treated, physicians must choose how to select their treatment, one at a time. 
There are two treatments or `arms', $X = 0$ and $X = 1$, and the overall goal is to find the optimal treatment to maximize the chance of patients' recovery ($Y$). This can be seen as an example of a multi-armed bandit (MAB) model. Its roots can be traced back to work produced by \citet{thompson1933likelihood}, which was further developed in \citep{robbins52, gittins1979bandit,lai-robbins85,Auer2002}. 
Consider a MAB model described by an SCM 
\begin{eqnarray}
\M^*_{\textsc{mab}} = \langle \*U = \{U\}, \*V =  \{X, Y\}, \2F, P(U)\rangle,
\end{eqnarray}
 where $U$ represents the patient's age (e.g., normalized in a real interval $[0, 1]$). The causal mechanisms are the following:
\begin{align} 
     \2F = \begin{cases}
       X \gets \I\{U < 0.8\}, \\
       Y \gets \I\{U < 0.4 - \Delta X\}
     \end{cases} \label{eq:_2_1_mab}
\end{align}
where coefficient $\Delta$ is a real number bounded in $(0, 0.4)$; and $P(U)$ is such that values of $U$ are drawn from a uniform distribution $\texttt{Unif}(0, 1)$.

In words, the physician prescribes treatment $X = 0$ for senior patients ($U \geq 0.8$); otherwise, an alternative treatment, $X = 1$, is prescribed. Meanwhile, the recovery $Y$ of the disease depends on the treatment ($X$) and the patient's age ($U$). Since the coefficient $\Delta > 0$, the chance of recovery when prescribing $X = 0$ is higher than when prescribing $X = 1$. That is, $X = 0$ is the preferable treatment in this context. \hfill $\blacksquare$
\end{example}
Interestingly, note that the SCM given by Eq.~\ref{eq:_2_1_mab} contains two qualitatively different types of mechanisms -- $f_X$ is controlled by the physician, and $f_Y$ is controlled by Nature, representing in this case, the patient's biology. From the CRL's agent point of view, both mechanisms are external and then deemed as the environment. 

\begin{example}[Markov Decision Process \citep{puterman1994markov}] \label{exp:_2_1_mdp}
\emph{Markov Decision Process} (MDP) has emerged as the \emph{de facto} framework for reasoning about the sequential decision-making in AI \citep{sutton1998reinforcement, russell2016artificial}. An example of MDP is the day-to-day management of an inventory with a fixed maximum size \citep{2010Szepesvari}. 

On day $i = 1, 2, \dots$, the inventory manager observes the current size of the inventory $S_i$, decides whether to purchase new items to fill up the inventory $X_i$, and receives a subsequent profit $Y_i$ by the end of day $i$. 
More specifically, consider an MDP model described by the SCM:
\begin{align}
    \M^*_{\textsc{mdp}} = \left \langle \*U = \{U_{i,1}, U_{i,2}, U_{i,3}\}, \*V =  \{X_i, Y_i, S_i\}, \2F = \{\2F_i\}, P(\*U) \right \rangle_{i = 1, 2, \dots}, 
\end{align}
where $U_{i,1}, U_{i,2}, U_{i,3}$ represent, respectively, human errors when stocking the inventory, and uncertainties in demand, and monetary values of the goods. The causal mechanisms $\2F_i$ representing the system dynamics transitioning from day $i-1$ to $i$ are defined as:
 \begin{align}
    \2F_i = \begin{cases}
      S_i \gets \left (S_{i-1} \vee X_{i-1} \right) \oplus U_{i-1, 1} \oplus U_{i-1,2},\\
      X_i \gets S_i \oplus U_{i, 1} \\
      Y_i \gets S_i \oplus X_i \oplus U_{i, 1} \oplus U_{i, 3}
    \end{cases} \label{eq:_2_1_mdp}
  \end{align}
and $P(\*U)$ is such that $U_{i, 1}, U_{i, 2}, U_{i, 3} \in \{0, 1\}$ are independent variables drawn from distributions $P(U_{i, 1} = 1) = P(U_{i, 2} = 0) = P(U_{i, 3} = 0) =  0.9$. 

In words, the size of current inventory $S_i = 1$ is full if it is also full $S_{i-1} = 1$ on the previous day or gets refilled $X_{i-1} = 1$; the operation error $U_{i-1, 1}$ and customers' demand $U_{i-1, 2}$ could also affect the inventory size. The manager's decision $X_i$ depend on the inventory size $S_i$ on the day and potential operation errors $U_{i, 1}$. The store only makes a profit $Y_i = 1$ if the inventory is well managed: it is refilled when empty, or no new goods is purchased when full. Similarly, operation errors and price fluctuation could also affect the net profit on day $i$.
\hfill $\blacksquare$
\end{example}

A couple of observations follow. First, SCMs provide a flexible language that allows for the natural encoding of standard RL environments. 
Second, the specific instantiation of each SCM will, in general, not be visible to the agent, and the goal is just to represent the environment's generative processes and its implied PCH.  
Third, the SCM is different than the PCH's corresponding datasets (Sec.~\ref{sec:_2_2_obs+exp}) and the assumptions the agent may make about them (Sec~\ref{sec:_2_3_diagram}), as discussed  next.

\subsection{Learning through Observational \& Interventional Regimes (PCH's Layers 1 and 2)}\label{sec:_2_2_obs+exp}

\begin{figure}[t]
    \centering
   \begin{tikzpicture}
  \newcommand*{\smallgap}{0.25}
  \newcommand*{\biggap}{1}

        \node (env) [environment, fill=gray!10, dashed, minimum height=2cm, text width=2.5cm] at (0, 1) {SCM $\M^*$};
    
        \node (agent) [agent, minimum height=2cm, text width=2.5cm] at (7, 1) {CRL Agent $\3C$};
        \node (b-agent) [agent, minimum height=2.2cm, text width=2.5cm] at (-5, 1) {Behavior Regime\\\footnotesize{(other agents,\\ biology, physics)}};

        \node [draw, dashed, text width = 0.5cm, text centered] (obs) at (3.5, 1.7) {$\1L_1$};
        \node [draw, dashed, text width = 0.5cm, text centered] (exp) at (3.5, 1) {$\1L_2$};
        \node [draw, dashed, text width = 0.5cm, text centered] (ctf) at (3.5, 0.3) {$\1L_3$};
        
        \node [draw, text width = 1cm, text centered, minimum height = 2.3 cm, label=above:{PCH}] (int) at (3.5, 1) {};
		\node [text centered, minimum height = 0.1 cm, label=above:{Interactions}] (int) at (3.5, -1) {};
	
        \path [-Latex, betterblue, ultra thick] ($(agent.west)+(0, 2.5*\smallgap)$) edge (obs);
        \path [-Latex, bettergreen,ultra thick] (agent.west) edge (exp);
        \path [-Latex] ($(agent.west)-(0, 2.5*\smallgap)$) edge (ctf);
        
        \path [-Latex, betterblue, ultra thick]  ($(env.east) + (0, 2.5*\smallgap)$) edge  (obs);
	   	\path [-Latex, bettergreen, ultra thick] (env.east) edge  (exp);
	   	\path [-Latex] ($(env.east) - (0, 2.5*\smallgap)$) edge (ctf);
        \path [Latex-Latex, betterred, ultra thick] (b-agent.east) edge node[anchor=south]{$\*X \gets \pi^b$} (env.west);

    \end{tikzpicture}
\caption{Representation of the CRL agent (right side) interacting with the SCM (middle)  through natural (marked in blue) and interventional (green) regimes. Other behavior agents and their interactions are shown in the left side (red). }
  \label{fig:_2_2_scm-natural}
\end{figure}
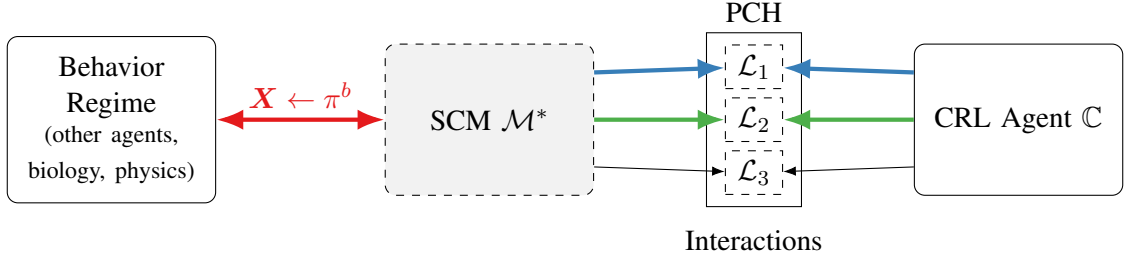

One important feature of the CRL architecture described so far is the explicit division between the agent and the environment where it lives,  which induces a natural line between the set of endogenous ($\*V$) and exogenous ($\*U$) variables. 
Interestingly, this implies that the partition between these variables is not referential-independent but agent-specific. Each agent has its way of partitioning (seeing) the environment. 
Also, the environment can be thought of as a mediator of many types of interactions, abstracting away potentially many other (behavior) agents who are also interacting with and have their own perspectives compared to the CRL agent, which is the one we care about. 
Fig.~\ref{fig:_2_2_scm-natural} illustrates this situation where the l.h.s. includes other agents interacting in the environment following their own behavior policies or natural variations, such as the laws of nature, including the wind, gravity, and natural selection. On the r.h.s., the CRL agent is depicted with its own views and capabilities. 

\subsubsection{Observational Distributions}
The CRL agent has two primary ways of interacting with the system: by perceiving the world through its sensors (by ``seeing'') or by performing intervention through its actuators (by ``doing''). 
This can be formalized through layers 1 and 2 of the PCH, which are shown in the middle of Fig.~\ref{fig:_2_2_scm-natural}. 
When the CRL agent passively observes how events unfold in time (marked in blue), the variations and dynamics in the system come from the other behavior agents (in the left), which can be natural or artificial. 
Since the CRL agent does not know the other agents' perspectives (their views, goals, and policies), these other interactions (marked in red) are of unknown nature and considered passive observations. In this case, the CRL agent does not deliberately interfere with the underlying SCM at any point in time. Actions $\*X$ attain their values under the control of the natural/behavior's regime. 

For any SCM $\M$, the collection of structural functions $\2F$ defines a mapping from the system's exogenous (unobserved) variables $\*U$ to the endogenous (observed) variables $\*V$. The distribution over the exogenous variables $P(\*U)$ induces a joint distribution over endogenous variables $\*V$, $P(\*V)$, which is called the \emph{observational distribution}.
\begin{definition}[Observational Distribution \citep{bareinboim2020pearl}]\label{def:_2_2_obs_dist}
  A SCM $\M = \langle \*U, \*V, \2F,$ $P(\*U)\rangle$ defines a joint probability distribution $P(\*Y)$ such that for any $\*Y \subseteq \*V$,
  \begin{align}
    P(\*y) = \sum_{\*u} \I \left\{\*Y(\*u) = \*y \right \} P(\*u), \label{eq:_2_2_obs_dist}
  \end{align}
  where $\*Y(\*u)$ is the solution of $\*Y$ after evaluating functions in $\2F$ given $\*U = \*u$. \hfill $\blacksquare$
\end{definition}
We will consistently use $P(\*Y; \1M)$ to represent an observational distribution $P(\*Y)$ evaluated with restriction in an SCM $\1M$. The input $\1M$ is omitted when the SCM is obvious from the context. The observational distribution in Eq.~\ref{eq:_2_2_obs_dist} could be evaluated following the procedure: 
\begin{enumerate}
  \item For each situation $\*U = \*u$, \footnote{Recall that each instantiation $\*U = \*u$ represents unobserved factors that generate variations within the environment of interest. In our context, this may represent an individual, a situation, or a state.} the environment evaluates the mechanisms in $\2F$ following a topological order (i.e., any variable in the l.h.s of function $f_i$ is evaluated after the ones in the r.h.s.), and
  \item The probability mass $P(\*u)$ is accumulated for each realization $\*U = \*u$ consistent with the event $\*Y = \*y$.
\end{enumerate}
We provide examples of this evaluation process below with the canonical RL models discussed earlier, where the CRL agent passively observes the unfolding of other agents interacting with $\M$. 

\begin{example}[MAB, Observational Distribution]\label{exp:_2_2_mab}
  Consider the MAB model $\M^*_{\textsc{mab}}$  defined earlier in Eq.~\ref{eq:_2_1_mab}. 
  By passively observing the behaviors of the physicians and collecting data, the CRL agent can estimate that the average recovery rate of each patient as
  \begin{align}
      P(Y = 1) &= P(X = 0, Y = 1) + P(X = 1, Y = 1)\\
      &= P(U \geq 0.8, U < 0.4) + P(U < 0.8, U < 0.4 - \Delta)\\
      &= P(U < 0.4 - \Delta)\\
      &= 0.4 - \Delta. \label{eq:_2_2_mab1}
  \end{align}
The agent has access only to the value in the r.h.s. of Eq.~\ref{eq:_2_2_mab1}, and Nature does the evaluation process itself. 
In particular, patients are evaluated
following the two 2-step procedure described in Def.~\ref{def:_2_2_obs_dist}, following the population's proportions and the corresponding structural mechanisms. 

To further consider the effectiveness of the treatments, the CRL agent may collect data so that it can also compute the recovery rate for the specific treatment $X = 0$. That is,
  \begin{align}
      P(Y = 1 \mid X = 0) &= P(U < 0.4 \mid X = 0)\\
      &= P(U < 0.4 \mid U \geq 0.8)\\
      &= 0
  \end{align}
  Similarly, the recovery rate conditioning on event $X = 1$ is given by
  \begin{align}
      P(Y = 1 \mid X = 1) &= P(U > 0.4 - \alpha \mid X = 1)\\
      &= P(U > 0.4 - \Delta \mid U < 0.8)\\
      &= 0.5 - 1.25 \Delta.
  \end{align}
Again, the CRL agent has access only to the r.h.s.\ of the equation through data coming from sampling. Since the coefficient $\Delta \in (0, 0.4)$, then $0.5 - 1.25\Delta > 0$, which implies that 
  \begin{align}\label{eq:_2_2_mab2}
P(Y = 1 \mid X = 0) <  P(Y = 1 \mid X = 1).
  \end{align} 
This seems to suggest that treatment $X = 1$ achieves a better recovery rate than treatment $X = 0$. 

The conclusion that follows from Eq.~\ref{eq:_2_2_mab2} seems to be at odds with the earlier analysis based on the full knowledge of the SCM and the specific mechanisms of how $Y$ comes about (as shown in Eq.~\ref{eq:_2_1_mab}), which concluded that $X = 0$ is the optimal treatment. 

Note that this evaluation is based on passively collected data (from layer the PCH's $\Ll_1$), which means that a proper causal interpretation is generally not well-advised. The behavior agent may not be efficient in allocation or have different goals in mind. 
\hfill $\blacksquare$
\end{example}

\begin{example}[MDP, Observational Distribution]\label{exp:_2_2_mdp}
Consider the MDP $\M^*_{\textsc{mdp}}$ described in Eq.~\ref{eq:_2_1_mdp}. We are interested in evaluating the inventory manager's cumulative profit $\E\left [ \sum_{i = 1}^{\infty} \gamma^{i-1} Y_i \right]$, where $\gamma$ is a discount factor in the real interval $(0, 1)$. Evaluating profit $Y_i$ on day $i = 1, 2, \dots$ in $\M^*_{\textsc{mdp}}$ gives:
\begin{align}
    Y_i &= S_i \oplus X_i \oplus U_{i, 1} \oplus U_{i, 2} \notag \\ 
    &= S_i \oplus S_i \oplus U_{i, 1} \oplus U_{i, 1} \oplus U_{i, 2} \notag \\ 
    &= U_{i, 3}
\end{align}
That is, the manager's inventory control policy generates an expected profit $\E[Y_i] = P(U_{i, 2} = 1) = 0.1$ every day. For concreteness, we consider $\gamma = 0.9$. The store's expected cumulative reward is:
\begin{align}
    \E \left [ \sum_{i = 1}^{\infty} \gamma^{i-1} Y_i\right] &=\sum_{i = 1}^{\infty} \gamma^{i-1} \E[Y_i] \notag \\
    &= \frac{0.1}{1 - \gamma} = 1. \label{eq:_2_2_mdp1}
\end{align}

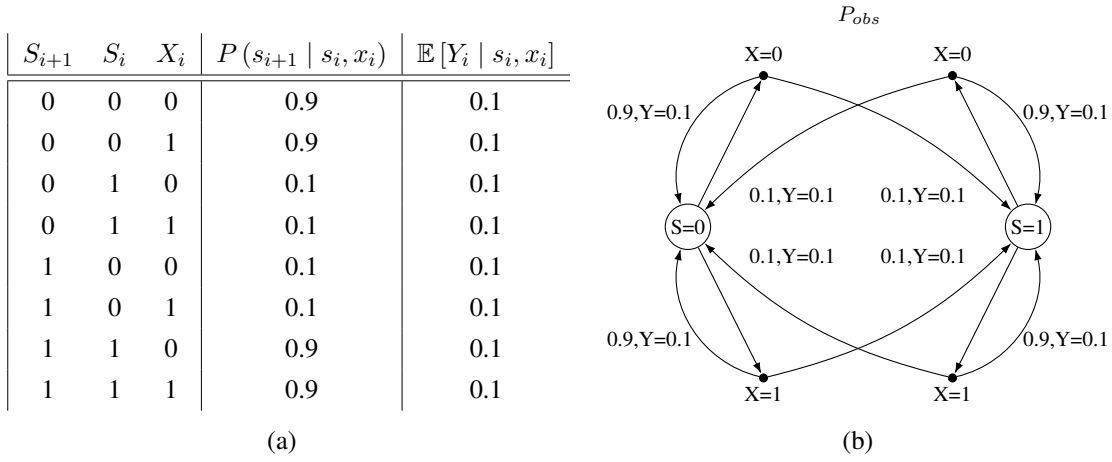
\begin{figure}[t]
\begin{subfigure}{0.5\linewidth}\centering
  \centering
  \resizebox{\linewidth}{!}{
  \renewcommand{\arraystretch}{1.25}
  \begin{tabular}{|ccc|c|c}
          $S_{i+1}$ & $S_i$ & $X_i$ & $P\left(s_{i+1} \mid s_{i}, x_i \right)$ & $\E\left[Y_{i} \mid s_{i}, x_i \right]$ \\
    \hline
    \hline
    0     & 0     & 0     & 0.9 & 0.1 \\
    0     & 0     & 1     & 0.9  &0.1\\
    0     & 1     & 0     & 0.1  &0.1\\
    0     & 1     & 1      & 0.1 &0.1\\
    1     & 0     & 0      & 0.1 &0.1\\
    1     & 0     & 1     & 0.1 &0.1\\
    1     & 1     & 0     & 0.9 &0.1\\
    1     & 1     & 1     & 0.9 &0.1
    \end{tabular}%
    }
    \caption{}
  \label{tab:_2_2_mdp_obs}%
  \end{subfigure}\hfill
  \begin{subfigure}{0.48\linewidth}\centering
  \begin{tikzpicture}
      \def\outerr{3.5}
      \def\innerr{3}

      \node[vertex, minimum width=6mm] (S0) at (0, 0) {S=0};
      \node[vertex, minimum width=6mm] (S1) at (4.5, 0) {S=1};
      \node[action, label={[shift={(0,0)}]\scriptsize X=0}] (X00) at (1, 2) {};
      \node[action, label={[shift={(0,-0.5)}]\scriptsize X=1}] (X10) at (1, -2) {};
      \node[action, label={[shift={(0,0)}]\scriptsize X=0}] (X01) at (3.5, 2) {};
      \node[action, label={[shift={(0,-0.5)}]\scriptsize X=1}] (X11) at (3.5, -2) {};
      
	  \node[text width = 1.3cm, right] at (1.85, +2.8) {\scriptsize $P_{obs}$};

      \draw[dir] (S0) to (X00);
      \draw[dir] (S0) to (X10);
      \draw[dir] (S1) to (X01);
      \draw[dir] (S1) to (X11);
      
      \node[text width = 1.3cm, align = right, left] at (0.2, 1.5) {\scriptsize 0.9,Y=0.1};
      \node[text width = 1.3cm, align = right, left] at (0.2, -1.5) {\scriptsize 0.9,Y=0.1};
      
      \node[text width = 1.3cm, align = right] at (1.3, 0.4) {\scriptsize 0.1,Y=0.1};
      \node[text width = 1.3cm, align = right] at (1.3, -0.4) {\scriptsize 0.1,Y=0.1};
      
      \node[text width = 1.3cm, align = left] at (3.2, 0.4) {\scriptsize 0.1,Y=0.1};
      \node[text width = 1.3cm, align = left] at (3.2, -0.4) {\scriptsize 0.1,Y=0.1};
      
      \node[text width = 1.3cm, right] at (4.3, 1.5) {\scriptsize 0.9,Y=0.1};
      \node[text width = 1.3cm, right] at (4.3, -1.5) {\scriptsize 0.9,Y=0.1};
      
      \draw[dir] (X00) to [bend right = 45] (S0);
      \draw[dir] (X10) to [bend left = 45] (S0);
      \draw[dir] (X00) to [bend left = 15] (S1);
      \draw[dir] (X10) to [bend right = 15] (S1);
      
      \draw[dir] (X01) to [bend left = 45] (S1);
      \draw[dir] (X11) to [bend right = 45] (S1);
      \draw[dir] (X01) to [bend right = 15] (S0);
      \draw[dir] (X11) to [bend left = 15] (S0);
  \end{tikzpicture}
  \caption{}
  \label{fig:_2_2_mdp_obs}
\end{subfigure}\hfill\null
\caption{Observational distributions of the MDP model $\M^*_{\textsc{mdp}}$ described in Eq.~\ref{eq:_2_1_mdp}}
\label{fig:_2_2_obs_mdp}
\end{figure}%

\noindent We now consider the transition distribution $P(S_{i+1} \mid S_i, X_i)$ and expected reward $\E[Y_i \mid S_i, X_i]$ conditional on decision $X_i$ and inventory size $S_i$ on day $i$. First, evaluating the inventory size $S_{i+1}$ on the next day in $\M^*_{\textsc{mdp}}$ gives:
\begin{align}
    P\left(S_{i+1} \mid S_i = s_{i}, X_i = x_i \right) &= P\left( (s_{i} \vee x_{i}) \oplus U_{i, 1} \oplus U_{i, 2} \mid S_i = s_{i}, X_i = x_i \right)
\end{align}
For $\M^*_{\textsc{mdp}}$ defined in Eq.~\ref{eq:_2_1_mdp}, observing events $S_i = 0, X_i = 0$ implies that $U_{i, 1} = 0$, so 
\begin{align}
    P\left(S_{i+1} = 1 \mid S_i = 0, X_i = 0 \right) &= P\left((0 \vee 0) \oplus 0 \oplus U_{i, 2} = 1 \mid S_i = 0, X_i = 0\right) \\
    &= P(U_{i, 2} = 1) \\
    &= 0.1.
\end{align}
The second step holds since the stochastic demand $U_{i, 2}$ is an independent factor only affecting $S_{i+1}$. Similarly, evaluating the expected reward $Y_i$ in $\M^*_{\textsc{mdp}}$ gives:
\begin{align}
    \E\left[Y_i \mid S_i = s_{i}, X_i = x_i \right] &= \E\left[s_i \oplus x_i \oplus U_{i, 1} \oplus U_{i, 3} \mid S_i = s_i, X_i = x_i\right]
\end{align}
For $S_i = 0, X_i = 0$, the above equation could be further written as
\begin{align}
    \E\left[Y_i \mid S_i = 0, X_i = 0 \right] &= \E\left[0 \oplus 0 \oplus 0 \oplus U_{i, 3} \right] \\
    &= P(U_{i, 3} = 1) \\
    &= 0.1
\end{align}
The detailed parametrizations of the $\Ll_1$-distribution, $P(S_{i+1} \mid S_i, X_i)$ and $\E\left[Y_i \mid S_i, X_i \right]$, are described in Fig.~\ref{tab:_2_2_mdp_obs}. Following the convention in reinforcement learning, this parametrization can be represented through a probabilistic finite-state machine as shown in Fig.~\ref{fig:_2_2_mdp_obs}, which is read as follows. Assuming the current state is $S = 0$, there are two outgoing transitions to $X = 0$ and $X = 1$, which represent two possible actions the inventory manager could be observed to take. 
If the manager takes action $X = 0$, the transition of returning back to the state $S = 0$ has a probability of $0.9$, and is associated with an average reward $Y = 0.1$, as indicated in the arrow. That is, the transition probability $P(S_{i+1} = 0 \mid S_i = 0, X_i = 0) = 0.9$, and the reward function $\E\left[Y_i \mid S_i = 0, X_i = 0 \right] = 0.1$. On the other hand, it may also transition to state $S = 1$ with probability $0.1$, resulting in a subsequent reward of $Y = 0.1$. That is, the transition function $P(S_{i+1} = 0 \mid S_i = 1, X_i = 0) = 0.1$, and the reward function $\E\left[Y_i \mid S_i = 1, X_i = 0 \right]= 0.1$. \hfill $\blacksquare$
\end{example}

So far, we have defined a model for typical, template-like RL environments and some implications of passively observing such environments, including collecting data from the corresponding observational ($\Ll_1$) distributions.  

\subsubsection{Interventional Distributions}\label{sec:interventions}
In this section, we discuss how to model the agent's interventions in the world whenever they aim to bring some state of affairs about.
In particular, we consider interventions that build on previous states' history and actions. 
These types of interventions are usually called \emph{soft} or \emph{policy interventions}, and we will follow the treatment developed in  \citep{correa2020calculus,correa2020general}. \footnote{There is a growing literature interested in different features of these interventions, refer to \citep[Ch.~4]{pearl:2k} for some historical discussion, and also \citep{dawid:02,didelez2006direct,tian2008dsp}. }  

Formally, we denote by $\pi_{\*X}$ a sequence of decision rules $\left \{ \pi_X \mid \forall X \in \*X \right \}$ over an arbitrary set of action variables $\*X$. Each $\pi_X$ is a function that determines values of $X$ taking some other variables $\*S_X$ as input; that is, $X \gets \pi_X(\*S_X)$. With a slight abuse of notation, we denote it by $\pi_{X}(X \mid \*S_X)$, the stochastic policy mapping from values of $\*S_X$ to the probabilities space over domains of every $X \in \*X$.\footnote{Formally, this means that the original pair, natural/behavior function $f_X$ and exogenous term $U_X$ is replaced with another pair, a new function $f_X^*$ and an independent noise $U_X^*$, generating the purported conditional distribution $\pi_{X}$ behavior. For simplicity, we ignore $U_X^*$ and write a stochastic policy $X \sim \pi(x | \*s_X)$.} 
Such policies $\pi_{\*X}$ are also referred to as adaptive treatment strategies or treatment policies in the healthcare literature \citep{murphy2001dtr,chakraborty2014dynamic}. These decision rules provide an effective vehicle for personalized medicine for chronic conditions, in which treatment is repeatedly tailored to a patient's dynamic state. 

A policy dictates the actions that an agent (e.g., a physician, an ad-placement engine) could take based on the values of the states that it observes in the underlying SCM. A policy intervention $\doo(\*X \gets \pi_{\*X})$ following a policy $\pi_{\*X}$ (for short, $\doo(\pi_{\*X})$) is an operation that replaces the original behavior policy $f_{X}$ associated with every variable $X \in \*X$ with the corresponding decision rule $\pi_{X}$. 
\footnote{This overwrites the natural/behavior policy and is, therefore, oblivious to how other agents, artificial or natural, were operating in the environment before the intervention.} 
We formally define the new world that emerges when the current  one is submitted to an intervention through the notion of a submodel:
\begin{definition}[Submodel]\label{def:_2_2_submodel}
  Let $\M = \tuple{\*U, \*V, \2F, P}$ be an SCM and let $\pi_{\*X}$ be a policy over actions $\*X \subseteq \*V$. A submodel $\M_{\pi_{\*X}}$ of $\M$ is an SCM $\langle\*U, \*V, \2F_{\pi_{\*X}}, P(\*U) \rangle$ where 
  \begin{align}
    \2F_{\pi_{\*X}} = \set{f_V: \forall V \in \*V \setminus \*X} \cup \set{\pi_{X}: \forall X \in \*X}. \label{eq:_2_2_submodel} 
  \end{align} \hfill $\blacksquare$
\end{definition}
In words, whenever the causal system $\M$ is submitted to an intervention $\pi_{\*X}$, the equations relative to the original mechanisms (whatever these were) are replaced with the ones corresponding to the intervention, and all other equations remain the same; the resultant model is called $M_{\pi_{\*X}}$. 
A significant special class of interventions is called atomic, $\doo(\*X \gets \*x)$ (for short, $\doo(\*x)$), which sets the values of variables $\*X$ to some constants $\*x$. That is, decision rules are defined as $X \gets x$ for every variable $X \in \*X$. 
\footnote{This basic primitive has appeared at different times and contexts through causality's history. It was introduced in econometrics by \cite{haavelmo:43,strotz:wol60}.  In statistics, potential outcomes were introduced in the context of randomized experiments by \cite{neyman:23} and then connected with observational
studies by \cite{rubin:74}. In mathematical logic, counterfactuals
were discussed by \cite{lewis:73a} with possible worlds semantics.
Pearl developed a general and algorithmic treatment through
graphical models in AI \citep{pearl:93c,pearl:95} .}
The submodel induced by an atomic intervention $\doo(\*x)$ in an SCM $\M$ is usually written as $\M_{\*x}$ \citep[p.~204]{pearl:2k}.
 Further note that this is a derived model defined over the original SCM $\M$. In other words, the SCM is generative of multiple worlds, while $\M_{\*x}$ represents one of these worlds. 
 \footnote{The importance of this notion comes from the fact that it will allow us to represent the idea of causal effect, which is critical in evaluating the effect of actions. Of course, this is a semantical definition and operationalizing it in practice, whenever $\M$ is unknown, will be part of the inferential challenge faced by the CRL agent.   }

With the context of the sequential decision-making setting in mind, we provide a few concrete examples below demonstrating the evaluation of policies in canonical RL environments.
\begin{example}[MAB, Submodel]\label{exp:_2_2_mab2}
Let us consider again the MAB model $\M^*_{\textsc{mab}}$ in Eq. \ref{eq:_2_1_mab}. Due to new HIPAA privacy rules, the age of patients ($U$) is protected and cannot be disclosed, meaning they are unobserved from the CRL agent's perspective. Consider a policy $\pi \triangleq X \gets x$ that sets treatment $X$ to a constant $x \in \{0, 1\}$. The submodel $\M^*_{\textsc{mab}_x}$ induced by atomic intervention $\doo(X \gets x)$ is a tuple 
\begin{eqnarray}
\M^*_{\textsc{mab}_x} =  \langle  \*U = \{U\}, \*V = \{X, Y\}, \2F_{x}, P(U)\rangle, 
\end{eqnarray}
where 
  \begin{align}
     \2F_x = \begin{cases}
       X \gets x, \\ 
       Y \gets \I\{U < 0.4 - \Delta X\}
     \end{cases} \label{eq:_2_2_mab3}
  \end{align}

More broadly, a stochastic policy $\pi(X)$ is a probability distribution over domains of arm choice $X \in \{0, 1\}$. The submodel entailed by intervention $\doo(X \sim \pi(X))$ is a tuple 
\begin{eqnarray}
\M^*_{\textsc{mab}_{\pi}} = \langle \*U = \{U\}, \*V = \{X, Y\}, \2F_{\pi}, P(U)\rangle, 
\end{eqnarray}
where 
  \begin{align}
     \2F_{\pi} = \begin{cases}
       X \sim \texttt{Bernoulli}(\pi(X = 1)), \\
       Y \gets \I\{U < 0.4 - \Delta X\}
     \end{cases} \label{eq:_2_2_mab4}
  \end{align}
  In words, the CRL agent following a stochastic policy $\pi(X)$ prescribes treatment $X = x$ with probability $\pi(X = x)$ where $x \in \{0, 1\}$. The patient's age ($U$) is not disclosed to the CRL agent and thus is not considered when choosing the treatment at the decision-making time. \hfill $\blacksquare$
\end{example}

\begin{example}[Markov Decision Process's Submodel] \label{exp:_2_2_mdp2}
Consider the MDP $\M^*_{\textsc{mdp}}$ in Eq.~\ref{eq:_2_1_mdp} and an atomic intervention $\doo(X_1 \gets x_1, \dots, X_i \gets x_i)$, where the stocking decision $X_i$ for every day $i$ is fixed at constant $x_i = 0, 1$. The induced submodel is described by the tuple 
\begin{align}
\M_{\textsc{mdp}_{\*x}}  = \left \langle \*U = \{U_{i,1}, U_{i,2}, U_{i,3}\}, \*V =  \{X_i, Y_i, S_i\}, \2F_{\*x} = \{\2F_{x_{i}}\}, P(\*U) \right \rangle_{i=1, 2, \dots}
\end{align}
and $\2F_{x_{i}}$ is the post-interventional system dynamics from day $i-1$ to day $i$ given by
 \begin{align}
    \2F_{x_i} = \begin{cases}
      S_i \gets \left (S_{i-1} \vee X_{i-1} \right) \oplus U_{i-1, 1} \oplus U_{i-1,2},\\
      X_i \gets x_i \\
      Y_i \gets S_i \oplus X_i \oplus U_{i, 1} \oplus U_{i, 3}
    \end{cases} \label{eq:_2_2_mdp2}
  \end{align}
To improve the long-term profit, the store decides to automate the day-to-day inventory control using the CRL agent. Since operation errors $U_{i, 1}$ and fluctuations in demands $U_{i, 2}$ and pricing $U_{i,3}$ are not recorded in the store's system, they are unobserved from the CRL agent's perspective. 

A policy $\pi$ in the MDP model $\1M$ is a sequence of decision rules $\pi = (\pi_1, \pi_2, \dots )$, one for every time step $i$. Every decision rule $\pi_i(X_i \mid S_i)$ is a conditional distribution mapping from the domain of state $S_i$ to action $X_i$, for every step $i = 1, 2, \dots$. The submodel entailed by intervention $\doo(X_1 \sim \pi_1(X_1 \mid S_1), X_2 \sim \pi_2(X_2 \mid S_2), \dots)$ is described by the tuple 
  \begin{align}
\M^*_{\textsc{mdp}_{\pi}} =  \left \langle \*U = \{U_{i,1}, U_{i,2}, U_{i,3}\}, \*V =  \{X_i, Y_i, S_i\}, \2F_{\pi} = \{\2F_{\pi_{i}}\}, P(\*U) \right \rangle_{i=1, 2, \dots}, 
 \end{align}
where the causal mechanisms $\2F_{\pi_{i}}$ transitioning from day $i-1$ to $i$ is given by
  \begin{align}
    \2F_{\pi_i} = \begin{cases}
      S_i \gets \left (S_{i-1} \vee X_{i-1} \right) \oplus U_{i-1, 1} \oplus U_{i-1,2},\\
      X_i \sim  \texttt{Bernoulli}(\pi_i(X_i = 1 \mid S_i))\\
      Y_i \gets S_i \oplus X_i \oplus U_{i, 1} \oplus U_{i, 3}
    \end{cases} \label{eq:_2_2_mdp3}
  \end{align}
  In words, the CRL agent following a stochastic policy $\pi = (\pi_1, \pi_2, \dots )$ decides whether to restock $X_i = 1$ with probability $\pi_i(X_i = 1 \mid S_i)$ for every time step $i = 1, 2, \dots$. The agent's decision is free from operation errors and thus is not affected by exogenous variables $U_{i, 1}, U_{i, 2}, U_{i, 3}$. 
  \hfill $\blacksquare$
\end{example}
The usefulness of a submodel is that it gives semantics to how reality will behave when submitted to a new interventional condition. 
Similar to the observational distribution (Def.~\ref{def:_2_2_obs_dist}), an SCM also gives a natural valuation of consequences induced by interventions $\doo(\pi_{\*X})$ following a policy $\pi_{\*X}$.  The impact of the intervention on reward signals $\*Y$ is called a potential response. 
The definition of potential response of atomic intervention $\doo(\*x)$ was provided in \citep[Def.~7.1.4]{pearl:2k}, and next, we introduce a generalization to policy interventions $\doo(\pi_{\*X})$.
\begin{definition}[Potential Response] \label{def:_2_2_po}
	An SCM $\M= \tuple{\*U$, $\*V, \2F, P}$, let $\*X$ and $\*Y$ be two sets of variables in $\*V$, $\pi_{\*X}$ be a policy over $\*X$, and $\*u$ be a unit. The potential response $\*Y_{\pi_{\*X}}(\*u)$ is defined as the solution for $\*Y$ of the set of equations $\2F_{\pi_{\*X}}$ in $\1M$. That is, $\*Y_{\pi_{\*X}}(\*u) \triangleq \*Y(\*u; \1M_{\pi_{\*X}})$. 
\end{definition}
Formally, the interventional distributions produced by $\doo(\pi_{\*X})$ in an SCM $\M$ are distributions over endogenous variables in submodel $\M_{\pi_{\*X}}$.
\begin{definition}[Interventional Distribution]\label{def:_2_2_inv_dist}
  An SCM $\M= \tuple{\*U$, $\*V, \2F, P(\*U)}$ induces a family of joint probability distributions over $\*V$, one for each intervention $\doo(\pi_{\*X})$,  where $\pi_{\*X}$ is a policy over actions $\*X \subseteq \*V$. For each endogenous set $\*Y \subseteq \*V$, 
  \begin{align}
    \inv{\*Y}{\pi_{\*X}} \equiv \sum_{\*u} \left \{ \*Y_{\pi_{\*X}}(\*u) = \*y\right\}P(\*u), \label{eq:_2_2_inv_dist}
  \end{align}
where $\*Y_{\pi_{\*X}}(\*u)$ is the potential response of intervention $\doo(\pi_{\*X})$ on variables $\*Y$ (Def.~\ref{def:_2_2_po}).
\end{definition}
Distributions entailed by policy interventions $\doo(\pi_{\*X})$, defined in Eq.~\ref{eq:_2_2_inv_dist}, could be evaluated using a procedure described as follows.
\begin{enumerate}
  \item Replace the mechanism of each $X \in \*X$ with the corresponding functions $\pi_X$ generating functions $\2F_{\pi_{\*X}}$ (Eq.~\ref{eq:_2_2_submodel}), which induces a submodel $\M_{\pi_{\*X}}$ (of $\M$);
  \item For each situation $\*U = \*u$, the environment evaluates $\2F_{\pi_{\*X}}$ following a valid order (where any variable in the l.h.s.\ is evaluated after the ones in the r.h.s.), and
  \item The probability mass $P(\*U = \*u)$ is then accumulated for each instantiation $\*U = \*u$ consistent with the event $\*Y = \*y$ in the submodel $\M_{\pi_{\*X}}$.
\end{enumerate}
As a special case, we denote by $ \inv{\*Y}{\*x}$ the interventional distribution entailed by atomic interventions $\doo(\*x)$, which is a joint distribution over variables $Y$ in submodel $\M_{\*x}$. The following examples demonstrate the evaluation of both atomic and policy interventional distributions in some canonical RL environments.

\begin{example}[MAB, Interventional Distribution, Atomic]\label{exp:_2_2_mab3}
  For the MAB model $\M^*_{\textsc{mab}}$ described in Eq.~\ref{eq:_2_1_mab}, we compute the interventional distribution $P(Y \mid \doo(X \gets x))$, $x \in \{0, 1\}$. Evaluating the recovery of the patient $Y$ in submodel $M^*_{\textsc{mab}_{X \gets 0}}$ described in Eq.~\ref{eq:_2_2_mab3} gives
  \begin{align}
      \inv{Y = 1}{X \gets 0} &= P(U < 0.4) \\
      &= 0.4 \label{eq:_2_2_mab5}
  \end{align}
  Similarly, the patient's recovery rate of treatment $X \gets 0$ is equal to
  \begin{align}
      \inv{Y = 1}{X \gets 1} &= P(U < 0.4 - \Delta) \\
      &= 0.4 - \Delta.  \label{eq:_2_2_mab6}
  \end{align}
  Since the coefficient $\Delta > 0$, 
  \begin{align}
\inv{Y = 1}{X \gets 0} > \inv{Y = 1}{X \gets 1}. 
  \end{align}
  
A few observations follow. 
First, a policy prescribing treatment $X \gets 0$ implies a higher chance of patients' recovery compared to  $X \gets 1$, i.e., $X \gets 0$ is the optimal treatment. 
Second, this matches the analysis based on the true mechanism of $Y$ as described in Eq.~\ref{eq:_2_1_mab}. \hfill $\blacksquare$
\end{example}

A significant challenge arises since the agent doesn't have access to the true description of the environment, encoded as the SCM $\M$, it will need to perform interventions physically and set action $X=x$, despite the other factors, to obtain samples from the two distributions described by Eqs.~\ref{eq:_2_2_mab5} and \ref{eq:_2_2_mab6}. 
In turn, we discuss more complex interventions from a non-atomic policy. 

\begin{example}[MAB, Interventional Distribution, Policy]\label{exp:_2_2_mab4}
For the same MAB environment $\M^*_{\textsc{mab}}$ described in the previous example (and Eq.~\ref{eq:_2_1_mab}), we have that event $X = 1$ is equivalent to $U < 0.5$. The observational distribution can be obtained by using Eq.~\ref{eq:_2_2_obs_dist} and $\M^*_{\textsc{mab}}$, which leads to $P(X = 1) = 0.5$.
In words, the physician seems to be prescribing treatment $X$ uniformly at random. 
  Recall that the physician's recovery rate based on Eq.~\ref{eq:_2_2_mab1} was $P(Y = 1) = 0.5 + \alpha$. 
  
One may be tempted to surmise that the CRL agent could achieve the same performance as the physician by 'cloning' its random policy, i.e., 
\begin{align}
\pi(X = x) = P(X = x) = 0.5. 	
\end{align}
Perhaps surprisingly, this is not the case. Specifically, evaluating the recovery $Y$ in submodel $\M^*_{\textsc{mab}_{\pi}}$ described in Eq.~\ref{eq:_2_2_mab3}, gives:
\begin{align}
  \inv{Y = 1}{\pi} &= \sum_{x = 0, 1} \pi(x) \inv{Y = 1}{X \gets x}\\
  &= 0.5 + 0.5 \alpha. \label{eq:_2_2_mab7}
\end{align}
Whenever coefficient $\alpha > 0$, the recovery rate obtained by the CRL agent given by Eq.~\ref{eq:_2_2_mab7} will be smaller than the physician's (Eq.~\ref{eq:_2_2_mab1}). \hfill $\blacksquare$
\end{example}

This example highlights the clear difference between observational (seeing) and interventional (doing) distributions from the perspective of the CRL agent. 
Also,  when the behavior and the CRL agents have different perceptual capabilities and views of the environment, naively copying (or cloning) the nominal behavior policy $P(X)$ may not lead to a successful policy (as shown by Ex.~\ref{exp:_2_2_mab3}). 
There is no formal basis to pursue such a strategy since these two distributions are different, and the empirical gap could be significant.  
\footnote{ This leads to a more refined discussion of imitation learning, which is provided in Sec.~\ref{sec:_8_imitation} and accompanied by proper causal modeling and solutions.}

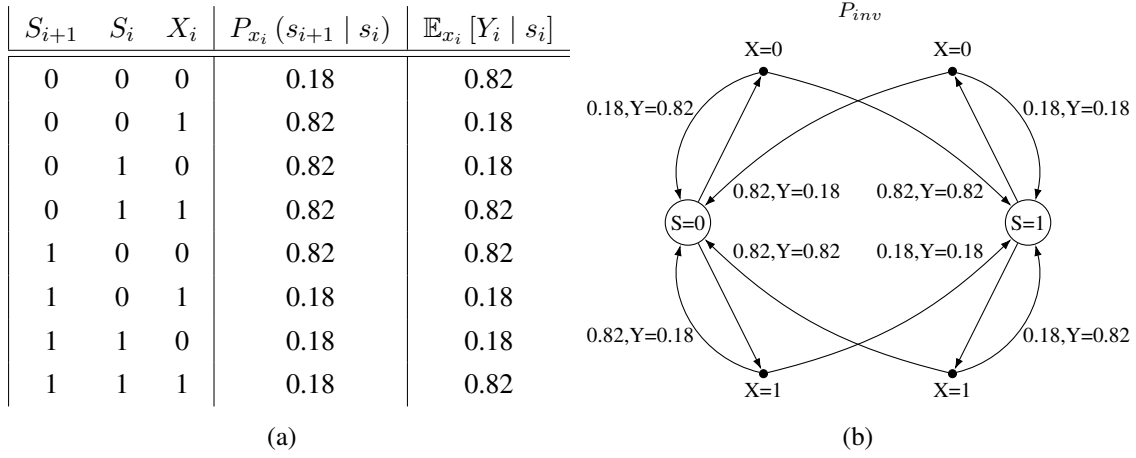
\begin{figure}[t]
\begin{subfigure}{0.5\linewidth}\centering
  \centering
  \resizebox{\linewidth}{!}{
  \renewcommand{\arraystretch}{1.25}
  \begin{tabular}{|ccc|c|c}
          $S_{i+1}$ & $S_i$ & $X_i$ & $\inv{s_{i+1} \mid s_{i}}{x_i}$ & $\invE{Y_{i} \mid s_{i}}{x_i}$ \\
    \hline
    \hline
    0     & 0     & 0     & 0.18 & 0.82 \\
    0     & 0     & 1     & 0.82  &0.18\\
    0     & 1     & 0     & 0.82  &0.18\\
    0     & 1     & 1      & 0.82 &0.82\\
    1     & 0     & 0      & 0.82 &0.82\\
    1     & 0     & 1     & 0.18 &0.18\\
    1     & 1     & 0     & 0.18 &0.18\\
    1     & 1     & 1     & 0.18 &0.82
    \end{tabular}%
    }
    \caption{}
  \label{tab:_2_2_mdp_inv}%
  \end{subfigure}\hfill
  \begin{subfigure}{0.5\linewidth}\centering
  \begin{tikzpicture}
      \def\outerr{3.5}
      \def\innerr{3}

      \node[vertex, minimum width=6mm] (S0) at (0, 0) {S=0};
      \node[vertex, minimum width=6mm] (S1) at (4.5, 0) {S=1};
      \node[action, label={[shift={(0,0)}]\scriptsize X=0}] (X00) at (1, 2) {};
      \node[action, label={[shift={(0,-0.5)}]\scriptsize X=1}] (X10) at (1, -2) {};
      \node[action, label={[shift={(0,0)}]\scriptsize X=0}] (X01) at (3.5, 2) {};
      \node[action, label={[shift={(0,-0.5)}]\scriptsize X=1}] (X11) at (3.5, -2) {};
      
	  \node[text width = 1.3cm, right] at (1.85, +2.8) {\scriptsize $P_{inv}$};
      
      \draw[dir] (S0) to (X00);
      \draw[dir] (S0) to (X10);
      \draw[dir] (S1) to (X01);
      \draw[dir] (S1) to (X11);
      
      \node[text width = 1.4cm, align = right, left] at (0.2, 1.5) {\scriptsize 0.18,Y=0.82};
      \node[text width = 1.4cm, align = right, left] at (0.2, -1.5) {\scriptsize 0.82,Y=0.18};
      
      \node[text width = 1.4cm, align = right] at (1.3, 0.4) {\scriptsize 0.82,Y=0.18};
      \node[text width = 1.4cm, align = right] at (1.3, -0.4) {\scriptsize 0.82,Y=0.82};
      
      \node[text width = 1.4cm, align = left] at (3.2, 0.4) {\scriptsize 0.82,Y=0.82};
      \node[text width = 1.4cm, align = left] at (3.2, -0.4) {\scriptsize 0.18,Y=0.18};
      
      \node[text width = 1.4cm, right] at (4.3, 1.5) {\scriptsize 0.18,Y=0.18};
      \node[text width = 1.4cm, right] at (4.3, -1.5) {\scriptsize 0.18,Y=0.82};
      
      \draw[dir] (X00) to [bend right = 45] (S0);
      \draw[dir] (X10) to [bend left = 45] (S0);
      \draw[dir] (X00) to [bend left = 15] (S1);
      \draw[dir] (X10) to [bend right = 15] (S1);
      
      \draw[dir] (X01) to [bend left = 45] (S1);
      \draw[dir] (X11) to [bend right = 45] (S1);
      \draw[dir] (X01) to [bend right = 15] (S0);
      \draw[dir] (X11) to [bend left = 15] (S0);
  \end{tikzpicture}
  \caption{}
  \label{fig:_2_2_mdp_inv}
\end{subfigure}\hfill\null
\caption{Interventional distributions of the MDP model described in Eq.~\ref{eq:_2_1_mdp}}
\label{fig:_2_2_inv_mdp}
\end{figure}%

\begin{example}[MDP, Interventional Distribution, Atomic]\label{exp:_2_2_mdp4}
Consider the MDP environment $\M^*_{\textsc{mdp}}$ described in Eq.~\ref{eq:_2_1_mdp} and the transition distribution $\inv{S_{i+1} \mid S_i}{x_i}$ and expected reward $\invE{Y_i \mid S_i}{x_i}$ induced by performing intervention $\doo(X_i \gets x_i)$ at inventory size $S_i$ on day $i$. Evaluating the inventory size $S_{i+1}$ on the next day in MDP $\M^*_{\textsc{mdp}_{\*x}}$ (Eq.~\ref{eq:_2_2_mdp2}) gives
\begin{align}
    \inv{S_{i+1} \mid S_i = s_{i}}{X_i \gets x_i} &= P\left( (s_{i} \vee x_{i}) \oplus U_{i, 1} \oplus U_{i, 2} \mid S_i = s_{i} \right)\\
    &=P\left( (s_{i} \vee x_{i}) \oplus U_{i, 1} \oplus U_{i, 2} \right) \label{eq:_2_2_mdp4}
\end{align}
The second step holds since the stochastic demand $U_{i, 2}$ is an independent variable only affecting $S_{i+1}$. For $S_i = 0, X_i \gets 0$, the above equation could be written as
\begin{align}
    \inv{S_{i+1} = 1\mid S_i = 0}{X_i \gets 0} &= P\left((0 \vee 0) \oplus U_{i, 1} \oplus U_{i, 2} = 1 \right) \\
    &= P(U_{i, 1} \oplus U_{i, 2} = 1) \\
    &= 0.82.
\end{align}
Similarly, evaluating the expected reward $Y_i$ in MDP $\M^*_{\textsc{mdp}_{\*x}}$ gives:
\begin{align}
    \invE{Y_i \mid S_i = s_{i}}{X_i \gets x_i} &= \E\left[s_i \oplus x_i \oplus U_{i, 1} \oplus U_{i, 3} \mid S_i = s_i\right]\\
    &=\E\left[s_i \oplus x_i \oplus U_{i, 1} \oplus U_{i, 3} \right] \label{eq:_2_2_mdp5}
\end{align}
For $S_i = 0, X_i \gets 0$, the company's expected profit is equal to
\begin{align}
    \invE{Y_i \mid S_i = 0}{X_i \gets 0}  &= \E\left[0 \oplus 0 \oplus U_{i, 1} \oplus U_{i, 3} \right] \\
    &= P(U_{i, 1} \oplus U_{i, 3} = 1) \\
    &= 0.82
\end{align}
The complete parametrizations of $\inv{S_{i+1} \mid S_i = s_{i}}{X_i \gets x_i}$ and $\invE{Y_i \mid S_i = s_{i}}{X_i \gets x_i}$ are shown in Fig.~\ref{tab:_2_2_mdp_inv}. The dynamic process described in Fig.~\ref{fig:_2_2_mdp_inv} shows a compact graphical representation of this parametrization where transition probabilities $\1T$ and reward function $\1R$ are given by
\begin{align}
    &\1T(s, x, s') = \inv{S_{i+1} = s' \mid S_i = s}{X_i \gets x}\\
    &\1R(s, x) = \invE{Y_i \mid S_i = s}{X_i \gets x}
\end{align} 
Note that compared with probabilities in Table~\ref{tab:_2_2_mdp_obs}, interventional distributions $\inv{S_{i+1} \mid S_i}{X_i}$, $ \hphantom{a}$ $\invE{Y_i \mid S_i}{X_i}$ and observational distributions $P\left(S_{i+1} \mid S_{i}, X_i \right)$, $\E\left[Y_i \mid S_{i}, X_i) \right]$ do not coincide in the MDP model $\M^*_{\textsc{mdp}}$ described in Eq.~\ref{eq:_2_1_mdp}. \hfill $\blacksquare$
\end{example}

Following Examples~\ref{exp:_2_2_mdp} and \ref{exp:_2_2_mdp4}, we note that the standard definition of MDP found in the literature \citep{puterman1994markov,sutton1998reinforcement}  based on a specific pair of transition probabilities $\1T$ and reward function $\1R$ only provides an abstraction for distributions in a single layer of the PCH, either observational or interventional. On the other hand, the SCM description of the MDP environment (e.g., Eq.~\ref{eq:_2_1_mdp}) provides a complete specification that allows for inferences across all layers of the PCH. 
This observation is interesting considering the distinct nature of the PCH's layers and each of the type of distributions associated with each of them. We will elaborate on this further in Sec.~\ref{sec:_3_3}.

We can also infer about the effects of non-atomic interventions following a Markov policy $\pi$ which determines values of every action $X_i$ based on the observed state $S_i$ at every decision horizon $i = 1, 2, \dots $. Our next example demonstrates such complex interventions in MDP environments.
\begin{example}[MDP, Interventional Distribution, Policy] \label{exp:_2_2_mdp5}
Consider the MDP $\M^*_{\textsc{mdp}}$ described in Eq.~\ref{eq:_2_1_mdp} and a policy $\pi = (\pi_1(X_1 \mid S_1), \pi_2(X_2 \mid S_2), \dots)$. Evaluating the transition distribution from the current state $S_i$ to next state $S_{i+1}$ in submodel $\M^*_{\textsc{mdp}_{\pi}}$  (Eq.~\ref{eq:_2_2_mdp3}) gives
\begin{align}
    \inv{S_{i+1} \mid S_i = s_{i}, X_i = x_i}{\pi} &= P\left( (s_{i} \vee x_{i}) \oplus U_{i, 1} \oplus U_{i, 2} \mid S_i = s_{i}, X_i = x_i \right)\\
    &=P\left( (s_{i} \vee x_{i}) \oplus U_{i, 1} \oplus U_{i, 2} \right)
\end{align}
The second step holds since the exogenous variable $U_{i, 2}$ is an independent variable only affecting the next state $S_{i+1}$. It follows from the evaluation in Eq.~\ref{eq:_2_2_mdp4} that the transition distribution remains invariant across atomic and policy interventions. That is,
\begin{align}
	\inv{S_{i+1} \mid S_i = s_{i}, X_i = x_i}{\pi} = \inv{S_{i+1} \mid S_i = s_{i}}{ X_i = x_i} \label{eq:_2_2_mdp6}
\end{align}
Similarly, evaluating the expected reward $Y_i$ conditioning on current state $S_i$ and observed action $X_i$ in submodel $\M^*_{\textsc{mdp}_{\pi}}$ gives:
\begin{align}
    \invE{Y_i \mid S_i = s_{i}, X_i = x_i}{\pi} &= \E\left[s_i \oplus x_i \oplus U_{i, 1} \oplus U_{i, 3} \mid S_i = s_i, X_i = x_i\right]\\
    &=\E\left[s_i \oplus x_i \oplus U_{i, 1} \oplus U_{i, 3} \right]
\end{align}
Again, it follows from Eq.~\ref{eq:_2_2_mdp5} that the conditional reward function remains the same for both atomic and policy interventions: 
\begin{align}
	\invE{Y_i \mid S_i = s_{i}, X_i = x_i}{\pi} = \invE{Y_i \mid S_i = s_{i}}{X_i \gets x_i} \label{eq:_2_2_mdp7}
\end{align}
More specifically, let policy $\pi$ be defined such that $X_i \gets S_i$ for every time step $i = 1, 2, \dots$. Evaluating the profit $Y_i$ on day $i$ evokes submodel $\M^*_{\textsc{mdp}_{\pi}}$ and is given by:
\begin{align}
     \invE{Y_i}{\pi} &= \invE{S_i \oplus X_i \oplus U_{i, 1} \oplus U_{i, 3}}{X_i \gets S_i} \\
     &=\E\left [S_i \oplus S_i \oplus U_{i, 1} \oplus U_{i, 3} \right]\\
     &= 0.82
\end{align}
The cumulative profit induced by policy $\pi$ with discount factor $\gamma = 0.9$ is then equal to:
\begin{align}
    \invE{\sum_{i = 1}^{\infty} \gamma^{i-1} Y_i}{\pi} &=\sum_{i = 1}^{\infty} \gamma^{i-1} \invE{Y_i}{\pi} \\
    &= \frac{0.82}{1 - \gamma}
\end{align}
Computing the above equation gives $\invE{\sum_{i = 1}^{\infty} \gamma^{i-1} Y_i}{\pi} = 8.2$, which outperforms the inventory manager's performance (behavior agent) $\E\left[\sum_{i = 1}^{\infty} \gamma^{i-1} Y_i\right] = 1$ (Eq.~\ref{eq:_2_2_mdp1}) under the behavior policy. In other words, it is more profitable to replace the manager with the CRL agent and automate the inventory management process in this case.
\hfill $\blacksquare$
\end{example}

\subsection{Encoding Structural Assumptions through Causal Diagrams}\label{sec:_2_3_diagram}
Even though SCMs are well-defined and provide precise semantics to the various types of distributions underlying the PCH, as discussed previously, one critical observation is that, in practice, they are usually not observable by the CRL agent. 
Further assumptions are needed if one wants to reason about the underlying SCM. We will introduce an object called a \emph{causal diagram} to encode assumptions about this SCM. There are different ways a causal diagram can be specified, including (1) as a template model (e.g., MABs, MDPs), (2) through prior knowledge, or (3) through a structural learning algorithm. We first describe the semantics of this object and a construction procedure that allows one to systematically articulate this causal diagram from a coarse, qualitative understanding of the underlying SCM. 

\begin{definition}[Causal Diagram \citep{pearl:2k, bareinboim2020pearl}] \label{def:_2_3_diagram}
Consider the SCM  $\1M = \langle \*V, \*U, \2F, P(\*U) \rangle$. A graph $\G$ is said to be a \emph{causal diagram} (of $\1M$) if:
	\begin{enumerate}
		\item there is a vertex for every endogenous variable $V_i \in \*V$,
		\item there is an edge $V_i \to V_j$ if $V_i$ appears as an argument of the mechanism $f_j \in \2F$,
		\item there is a bidirected edge $V_i\dashleftarrow\dasharrow V_j$ if the corresponding $U_i, U_j \subset \*U$ are correlated or the corresponding functions $f_i, f_j$ share some $U_{ij} \in \*U$ as an argument. \hfill $\blacksquare$
	\end{enumerate} 
\end{definition}

\begin{wrapfigure}[16]{r}{0.4\textwidth}
\centering
\vspace{-0.25in}
\includegraphics[keepaspectratio,width=0.33\columnwidth]{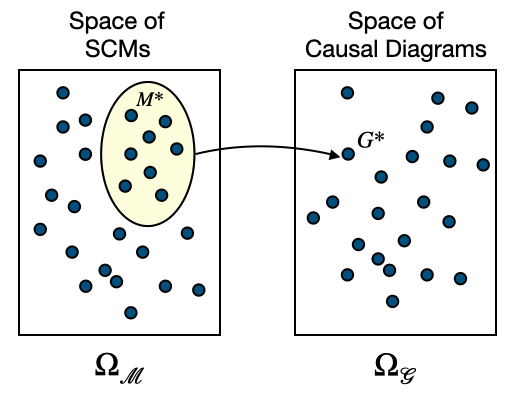} 
\caption{The space of SCMs/Causal diagrams are shown on the left/right side. The true SCM $\1M^*$ and the corresponding causal diagram $\G^*$ are explicitly shown. The yellow area represents the subspace where these other SCMs generate the same $\G^*$.  }
\label{fig:_2_3_scm-vs-dags}
\end{wrapfigure}
In words, there is an edge from endogenous variables $V_i$ to $V_j$ whenever $V_j$ ``listens to'' \footnote{This construction lies at the heart of the type of knowledge causal models represent, as suggested in \citep[pp.~129]{pearl2018book}: ``This listening metaphor encapsulates the entire knowledge that a causal network conveys; the rest can be derived, sometimes by leveraging data;'' for technical details, \cite[Sec.~1.4]{bareinboim2020pearl}. }  $V_i$ for determining its value. Similarly, a bidirected edge between $V_i$ and $V_j$ indicates shared, unobserved information affecting how both, or whether $V_i$ and $V_j$ ``listens'' to the same source of exogenous variations. Note that while the SCM contains explicit, quantitative information about all structural mechanisms ($\2F$) and exogenous probability distribution ($P(\*U)$), in contrast, the causal diagram encodes only  \textit{qualitative} information about which arguments were possibly used as inputs to the functions (from $\2F$) and how the exogenous variations are related (from $P(\*U)$). The diagram abstracts out the specifics of the mechanisms $\2F$ and distribution $P(\*U)$, retaining qualitative information about their possible arguments and independence structure, respectively. 
\footnote{
Furthermore, the existence of a directed arrow, e.g., $V_i \rightarrow V_j$, encodes the \textit{possibility} of the mechanism of $V_j$ to listen to variable $V_i$, but not its necessity. In this sense, the edges are non-committal; for instance, $f_j$ may decide not to consider $V_i$'s value. More formally, the assumptions are not encoded in the arrows in the diagram but in the missing arrows; each missing arrow ascertains that one variable is \textit{certainly} not the argument of the other or that one exogenous source of variation is not correlated to another. The same idea is true regarding the bidirected arrows and the possibility of covariation of some unobserved factors. }

\paragraph{I. Template models (knowledge-based).} 
In practice, the CRL agent will not have access to the fully specified SCM and will operate based on the assumptions encoded in the given causal diagram. This will represent a major inferential challenge since the diagram is much weaker than the SCM, and there are various SCMs (marked in yellow in Fig.~\ref{fig:_2_3_scm-vs-dags}) equally compatible with the same causal diagram. 
Still, whenever these inferences are allowed, this will translate into different gains in decision-making precision and efficiency. 

To ground this particular construction, we will consider the first way of specifying knowledge of the underlying SCM through pre-specified template models, as illustrated in the following example.

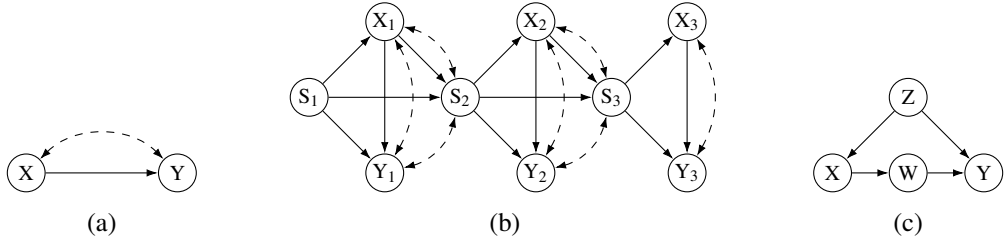
\begin{figure}[t]
	\hfill%
	\begin{subfigure}{0.2\linewidth}\centering
		\begin{tikzpicture}

			\node[vertex] (X) at (0, 0) {X};
			\node[vertex, opacity=0] (U) at (1, 2) {};
			\node[vertex] (Y) at (2, 0) {Y};
			\draw[bidir] (X) to [bend left = 45] (Y);

			\draw[dir] (X) -- (Y);
		\end{tikzpicture}
		\caption{}
		\label{fig:_2_3_1a}
	\end{subfigure}\hfill
	\begin{subfigure}{0.4\linewidth}\centering
		\begin{tikzpicture}
			\node[vertex] (S1) at (-1, -1) {S\textsubscript{1}};
			\node[vertex] (X1) at (0, 0) {X\textsubscript{1}};
			\node[vertex] (Y1) at (0, -2) {Y\textsubscript{1}};
			\node[vertex] (S2) at (1, -1) {S\textsubscript{2}};
			\node[vertex] (X2) at (2, 0) {X\textsubscript{2}};
			\node[vertex] (Y2) at (2, -2) {Y\textsubscript{2}};
			\node[vertex] (S3) at (3, -1) {S\textsubscript{3}};
			\node[vertex] (X3) at (4, 0) {X\textsubscript{3}};
			\node[vertex] (Y3) at (4, -2) {Y\textsubscript{3}};

			\draw[dir] (S1) to (S2);
			\draw[dir] (S1) to (X1);
			\draw[dir] (S1) to (Y1);
			\draw[dir] (X1) to (Y1);
			\draw[dir] (X1) to (S2);

			\draw[bidir] (X1) to [bend left = 30] (S2);
			\draw[bidir] (X1) to [bend left = 30] (Y1);
			\draw[bidir] (Y1) to [bend right = 30] (S2);

			\draw[dir] (S2) to (S3);
			\draw[dir] (S2) to (X2);
			\draw[dir] (S2) to (Y2);
			\draw[dir] (X2) to (Y2);
			\draw[dir] (X2) to (S3);

			\draw[bidir] (X2) to [bend left = 30] (S3);
			\draw[bidir] (X2) to  [bend left = 30] (Y2);
			\draw[bidir] (Y2) to [bend right = 30] (S3);

			\draw[dir] (S3) to (X3);
			\draw[dir] (S3) to (Y3);
			\draw[dir] (X3) to (Y3);

			\draw[bidir] (X3) to [bend left = 30] (Y3);
		\end{tikzpicture}
		\caption{}
		\label{fig:_2_3_1b}
	\end{subfigure}\hfill
	\begin{subfigure}{0.2\linewidth}\centering
		\begin{tikzpicture}

			\node[vertex] (X) at (0, 0) {X};
			\node[vertex] (Z) at (1, 1) {Z};
			\node[vertex] (W) at (1, 0) {W};
			\node[vertex] (Y) at (2, 0) {Y};
			\node[vertex, opacity=0] (U) at (1, 2) {};

			\draw[dir] (X) -- (W);
			\draw[dir] (W) -- (Y);
			\draw[dir] (Z) -- (Y);
			\draw[dir] (Z) -- (X);

		\end{tikzpicture}
		\caption{}
		\label{fig:_2_3_1c}
	\end{subfigure}\hfill\null
	\caption{Causal diagrams for (\subref{fig:_2_3_1a}) a multi-armed bandit (MAB); (\subref{fig:_2_3_1b}) a Markov decision process (MDP); and (\subref{fig:_2_3_1c}) an SCM representing a refinement of the MAB environment.}
	\label{fig:_2_3_1}
\end{figure}

\begin{example}[MAB's Causal Diagram]\label{exp:_2_3_mab}
Consider a template family of MAB models $\3M_{\textsc{mab}}$ which consists of SCMs $\mathcal{M}_{\textsc{mab}}$ described by a tuple 
\begin{align}
\M_{\textsc{mab}} = \langle \*U = \{U\}, \*V =  \{X, Y\}, \2F, P(U)\rangle,
\end{align}
The causal mechanisms $\2F$ are structural functions of the form:
\begin{align} 
     \2F = \begin{cases}
       X \gets f_X(U), \\
       Y \gets f_Y(X, U)
     \end{cases} \label{eq:_2_3_mab}
\end{align}
To apply the graphical construction dictated by Def~\ref{def:_2_3_diagram}, the AI engineer starts the modeling process by examining each of the endogenous variables $\*V = \{X, Y\}$, and adding them as nodes in the causal diagram. The corresponding diagram is illustrated in Fig.~\ref{fig:_2_3_1a} and will be called $\G_{\textsc{mab}}$. 

They then consider the second and third conditions of Def.~\ref{def:_2_3_diagram}. 
The mechanism underlying the context variable can be written as, 
\begin{align} \label{eq:_2_3_mab1}
X \gets f_X(U),
\end{align}
which suggests that the action is determined by an exogenous variable $U$ (in the natural regime). This is regardless of the specific form, $f_X$, of how these variables are realized in reality. The engineer may, in turn, think about the reward function, namely, 
\begin{align} \label{eq:_2_3_mab2}
Y \gets f_Y(X, U). 
\end{align}
Eq.~\ref{eq:_2_3_mab2} suggests how, in reality, the rewards that come about may be influenced by the action $X$. Graphically, this is represented through the arrow $X \rightarrow Y$. Furthermore, since the mechanisms $f_X$ and $f_Y$ share the exogenous variable $U$, a bidirected arrow $X \dashleftarrow \dashrightarrow Y$ is added to $\G_{\textsc{mab}}$.

Consider now a detailed MAB environment $\M^*_{\textsc{mab}}$ defined in Eq.~\ref{eq:_2_1_mab}. Since $\M^*_{\textsc{mab}}$ belongs to the template family $\3M_{\textsc{mab}}$, we could conclude that $\G_{\textsc{mab}}$ is a causal diagram associated with $\M^*_{\textsc{mab}}$. Note that this construction contrasts sharply with how detailed knowledge is encoded in the true SCM $\M^*_{\textsc{mab}}$, as delineated in Def.~\ref{def:_2_3_diagram}. Interestingly enough, an entirely different functional form of the reward mechanism, say 
\begin{align}
Y \gets \I\{U < 0.4 + \Delta X\}
\end{align}
would be equally compatible with the causal diagram depicted in Fig.~\ref{fig:_2_3_1a}. Compared with the original reward in Eq.~\ref{eq:_2_1_mab}, the coefficient of $X$ is flipped to $\Delta$. This means it is preferable to pull arm $X \gets 1$, which is the opposite of the optimal choice $X \gets 0$ in the original model. \hfill $\blacksquare$
\end{example}
Similar construction and argument can be used when considering an MDP environment described in Eq.~\ref{eq:_2_1_mdp}. The causal diagram in Fig.~\ref{fig:_2_3_1b} is called $\G_{\textsc{mdp}}$ and represents causal relationships among variables shared across a template family of MDP environments. 
In the same way as the template causal diagram for MABs, this diagram $\G_{\textsc{mdp}}$ is non-committal regarding the form of the mechanisms $\2F$ and the parametrization of the exogenous distribution $P(\*U)$. 

\paragraph{II. General causal models (knowledge-based).} 
The graphical models discussed so far based on templates conveniently encapsulate structural information about the state, decision, and outcome variables.
\footnote{In reality, this class of causal diagrams was introduced under the rubric of \textit{clustered diagrams} and different properties investigated, we refer readers to \citep{anand2021cluster} for further details. }
As will become apparent in the following sections,
this will naturally incur a cost in many practical CRL tasks. 
For now, we note that in some settings, additional information may be available that could be leveraged by the CRL agent. 

For concreteness, consider the diagram in Fig.~\ref{fig:_2_3_1c}. One way of thinking about it is as a refinement of the MAB diagram shown in Fig.~\ref{fig:_2_3_1a}, where a confounder $Z$ and a mediator $W$ are now explicit.  One natural question is how to test a model 
 designed by the AI engineer. Interestingly enough, there are constraints imprinted by the SCM over the observational distribution $P(\*V)$, as well as the other PCH's distribution, that will allow the CRL agent to check whether its current working hypothesis is plausible, regardless of the idiosyncrasies of the properties of the distribution of exogenous $P(\*U)$ and the causal mechanisms $\2F$ (e.g., monotonicity, linearity, separability). 

We will discuss for now constraints known as \textit{conditional independences} accompanied with a criterion known as \emph{d}-separation \citep{pearl:2k} that allows us to read such constraints from the model. A path $p$ from a node $X$ to a node $Y$ in $\G$ is a sequence of edges that does not include a particular node more than once. It may go either along or against the direction of the edges. A path consisting of only bidirected edges is called a bidirected path. Formally, it goes as follows.
\begin{definition}[\emph{d}-separation \citep{pearl:2k}]\label{def:_2_3_dsep}
A set $\*Z \subseteq \VV$ is said to block a path $p$ in $\G$ if either 
\begin{enumerate}
\item $p$ contains at least one arrow-emitting node that is in $\*Z$, or 
\item $p$ contains at least one collision node outside $\*Z$ and has no descendant in $\bm{Z}$.
\end{enumerate} 
If $\*Z$ blocks \emph{all} paths from set $\*X$ to set $\*Y$, it is said to ``$d$-separate $\*X$ and $\*Y$.''
\footnote{See \citet{hayduk:etal03}, \citet{mulaik2009linear}, and \citet[pp.\ 335]{pearl:09} for a gentle introduction to $d$-separation.} \hfill $\blacksquare$
\label{sec3-def}
\end{definition}
Before discussing some examples, we state one of the main results that connect d-separation statements made over the diagram $\G$ with constraints observed in the distribution $P(\*V)$. 
\begin{theorem}[Probabilistic Implications of \emph{d}-Separation \citep{pearl:2k}] \label{thm:_2_3_dsep}
If $\*X, \*Y$ are \emph{d}-separated by $\*Z$ in a causal diagram $\G$, then $\*X$ is independent of $\*Y$ conditional on $\*Z$ in every distribution $P$ compatible with $\G$. Conversely, if $\*X$ and $\*Y$  are not \emph{d}-separated by $\*Z$ in a diagram $\G$, then $\*X$ and $\*Y$ are dependent conditional on $\*Z$ in at least one distribution $P$ compatible with $\G$. \hfill $\blacksquare$
\end{theorem}

To illustrate these results, consider the causal diagram $\G$ in Fig.~\ref{fig:_2_3_1c} and whether $X$ and $Y$ are d-separated. Note that there are two paths from $X$ to $Y$, 
\begin{align}
	&p_1: X \leftarrow \fbox{$Z$} \rightarrow Y,  \\ 
	&p_2: X \rightarrow \fbox{$W$} \rightarrow Y. 
\end{align}
Since $\*Z = \{\}$ in this case, both paths $p_1$ and $p_2$ are opened. 
One interpretation of this result is that there is a flow of information from $X$ that is transmitted through $Z, W$ that affects $Y$. 
Now, let's consider the separation statement when the conditioning set $\*Z = \{Z, W\}$, or the confounder and mediator are in $\*Z$. 
The first condition of the criterion is then immediately satisfied, and the paths $p_1$ and $p_2$ are ``blocked.'' 
We can see through Thm.~\ref{thm:_2_3_dsep} that the following independence holds in $P(\*V)$: 
\begin{align}
 \left ( Y \ci X \mid \{Z, W\}  \right).
\end{align}
Intuitively,  once the values of $Z$ and $W$ are known, there is no information about $X$ that will affect the likelihood of $Y$ through $p_1$ and $p_2$, respectively. Readers are invited to check that the criterion is not satisfied if any intermediate variables are removed from the conditioning set. 

\paragraph{III. General causal models (learning-based).}Based on the marks imprinted by $\1M^*$ on $P(\*V)$, the agent may test whether the hypothesized graph $\G$ is compatible with the available data. There exists a traditional literature known as \textit{causal discovery} that attempts to perform the reverse process \citep{pearl:2k,spirtes2000causation,petersen:etal06}.
 In other words, from the marks readable from $P(\*V)$, the agent should infer what the compatible $\G$ that could have left these traces is. In practice, assumptions regarding the simplicity of these models (\`{a} la Occam's razor) are used to avoid situations described above, in which a saturated model would be preferred. 
 
 A growing, more recent literature is concerned with combining both observational and experimental distributions to learn a more restrictive equivalence class of causal diagrams \citep{kocaoglu2017experimental,kocaoglu2019characterization,wang2017permutation,agrawal2019abcd,JMLR:v21:17-123,jaber2020cd}. In fact, a richer set of constraints other than conditional independences emerge when we consider multiple distributions across different regimes (observational and interventional).

\section{Elements of Causal Reinforcement Learning}\label{sec:_3}
In this section, we will introduce a unified framework that lets us view the decision-making problem through causal lenses and solve reinforcement learning tasks using causal inference tools. First, Sec.~\ref{sec:_3_1} formalizes the decision-making problem in the causal language by introducing a mathematical object called causal decision models (CDMs). 
Every CDM comprises of a structural causal model representing the underlying environment, a policy space encoding what the agent can control and observe during the intervention, and a reward function that gauges the agent's performance. 

Armed with this new formalism, we can represent many canonical decision-making settings found in the literature within the semantic framework of  SCMs.
These include multi-armed bandits (MABs), Markov decision processes (MDPs), and dynamic treatment regimes (DTRs), among others. 
In practical scenarios, a detailed parametrization of the environment isn't always fully known, giving rise to reinforcement learning in SCMs. 
Sec.~\ref{sec:_3_2} introduces the concept of \emph{causal reinforcement learning task}, and provides an initial catalog of tasks that will be studied through this section. 
Each task is delineated by the manner in which the agent interacts with the environment (regime), any prior structural assumptions about that environment, and the specific policy space and reward function the agent seeks to optimize. 
Lastly, Sec.~\ref{sec:_3_3} delves into the importance of causal knowledge by examining policy learning within the conventional decision-making model of MDPs. 
Without explicitly acknowledging the learning regime and structural assumptions, we'll demonstrate that the underlying data-generating mechanisms can yield multiple MDPs, all compatible with the observed data, but with diverging implications for the optimization of decision-making. In simpler terms, the observed data typically doesn't fully dictate an optimal policy. 

\subsection{Causal Decision Models}\label{sec:_3_1}
We now formalize the policy optimization problem in SCMs based on the causal machinery introduced earlier in this section. The underlying environment will be represented as an SCM $\M^* = \tuple{\*U, \*V, \2F, P}$. We study the problem of interacting on action variables $\*X \subseteq \*V$ to optimize some performance measures over reward signals $\*Y \subseteq \*V$  evaluated by $\M^*$. The agent determines values of actions $\*X$ by performing intervention $\doo(\pi)$ following some policies $\pi$. The collection of all candidate policies $\pi$ determining values of actions $\*X$ defines a \emph{policy space} $\Pi$. Formally,
\begin{definition}[Policy Space]\label{def:_3_space}
	For an SCM $\M^* = \tuple{\*U, \*V, \2F, P}$, a policy space $\Pi$ is a collection of policies $\pi$ over actions $\*X = \{X_1, \dots, X_H\}$. Each policy $\pi$ is a sequence of decision rules $\left(\pi_1\left(X_1 \mid \*S_1\right), \dots, \pi_H\left(X_H \mid \*S_H\right) \right)$ such that for every $i = 1, \dots, H$, \footnote{The policy space $\Pi$ is also referred to a policy scope in \citep{lee2020characterizing}, which characterizes the agent's action and state scope - what it could interaction with and what it could observe at the time of interaction.}
	\begin{itemize}
		\item Action $X_i$ is a non-descendent of $X_{i+1}, \dots, X_{H}$, i.e., $X_i \in \*V \setminus \De(X_{i+1}, \dots, X_{H})$;
		\item States $\*S_i$ are non-descendants of $X_i, \dots, X_{H}$, i.e., $\*S_i \subseteq \*V \setminus \De(X_i, \dots, X_{H})$.
	\end{itemize}
	Henceforth, we will consistently denote such a policy space by $\Pi = \left \{ \langle X_1, \*S_1\rangle, \dots, \langle X_H, \*S_H \rangle \right \}$. \hfill $\blacksquare$
\end{definition}

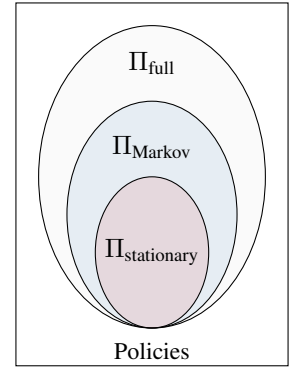
\begin{wrapfigure}[12]{r}{0.25\textwidth}
	\centering
	\vspace{-0.15in}
	\begin{tikzpicture}
		\draw[fill=none, draw=black] (-1.8, -2.5) rectangle (1.8, 2.3);
		\draw[fill=bettergray, fill opacity=0.1] (0,0) ellipse (1.5cm and 2cm);
		\draw[fill=betterblue, fill opacity=0.1, yshift = -0.5cm] (0,0) ellipse (1.125cm and 1.5cm);
		\draw[fill=betterred, fill opacity=0.1, yshift = -1cm] (0,0) ellipse (0.75cm and 1cm);

		\node at (0, -2.3) {\footnotesize Policies};
		\node at (0, 1.5) {\small $\Pi_{\text{full}}$};
		\node at (0, 0.4) {\small $\Pi_{\text{Markov}}$};
		\node at (0, -1) {\small $\Pi_{\text{stationary}}$};

	\end{tikzpicture}
	\caption{Policy spaces for a 2-stage DTR environment.}
	\label{fig:_3_1_space}
\end{wrapfigure}

Every policy $\pi \in \Pi$ is a sequence of decision rules $\left(\pi_1\left(X_1 \mid \*S_1\right), \dots, \pi_H\left(X_H \mid \*S_H\right) \right)$. An agent following policy $\pi$ selects values of actions $\*X$ following a temporal ordering $X_1, \dots, X_H$. At every step of intervention $i = 1, \dots, H$, it performs the following
\begin{enumerate}
	\item Observe some state variables $\*S_i = \*s_i$;
	\item Select a value of action $x_i \sim \pi_i(X_i \mid \*S_i = \*s_i)$ following the decision rule $\pi_i$;
	\item Perform an intervention $\doo(X_i \gets x_i)$ following the selected action $x_i$.
\end{enumerate}
In words, a policy space $\Pi$ defines the action space $\*X$ that the agent could control after being deployed in the environment and the state space $\*S_1, \dots, \*S_H$ that the agent could perceive at the time of intervention for every action $X_1, \dots, X_H$. The subsequent examples illustrate the concept of policy space in dynamic treatment regimes (for short, DTRs), which is a class of sequential decision-making environments widely applied in healthcare and personalized medicine.
\begin{example}[Dynamic Treatment Regimes \citep{RSSB:RSSB389}]\label{exp:_3_1_dtr}
	In healthcare, a typical patient is often treated at multiple stages; the physician repeatedly adapts each treatment, tailoring it to the patient's time-varying, dynamic state. Dynamic treatment regimes provide an appealing framework for managing personalized medicine in the longitudinal setting. 
	
	For instance, consider a DTR for managing alcohol-dependent patients, adapted from \citep{murphy2001dtr,chakraborty2013statistical}. The physician (i.e., the agent) has to decide the initial treatment $X_1$ and the secondary treatment $X_2$. Based on the condition of an alcohol-dependent patient ($S_1$), the physician may use behavioral therapy $(X_1 = 0$) or prescribe medication ($X_1 = 1$). The patient is then classified as a responder or a non-responder ($S_2$) based on the level of drinking within the next two months. 
 The physician must then decide whether to continue the initial treatment $(X_2 = 0)$ or switch to a more intensive plan combining medication and behavioral therapy ($X_2 = 1$). We are interested in the primary outcome $Y$ that measures the patient's days of abstinence over 12 months after the treatment.

More formally, consider a DTR environment $\1M^*_{\textsc{dtr}}$ that is a tuple given by
	\begin{align}
		\M^*_{\textsc{dtr}} = \langle \*U=\{U, U_{1}, \dots, U_4 \}, \*V=\{S_1, S_2, X_1, X_2, Y\}, \2F_{\textsc{dtr}}, P(\*U)\rangle, 
	\end{align}
where the underlying causal mechanisms $\2F_{\textsc{dtr}}$ are given by,
	\begin{align}\label{eq:_3_1_dtr1}
		\2F_{\textsc{dtr}} = \begin{cases}
			      S_1 \gets \I\{U_3 > 0\},                          \\
			      X_1 \gets \I\{3 S_1 + \alpha_1 U + U_1 > 0\},     \\
			      S_2 \gets \I\{0.1 + 0.1 S_1 + 0.1X_1 + U_4 > 0\}, \\
			      X_2 \gets \I\{3 S_2 + \alpha_2 U + U_2 > 0\},     \\
			      Y \gets \I\{ 3 U - 3S_1 - 3X_1 - 3S_1X_1 + 3 X_2 - 3S_2X_2 + 3X_1X_2> 0\},
		      \end{cases}
	\end{align}
Among quantities in the above equations, we set coefficients $\alpha_1 = \alpha_2 = 0$. However, for examples in subsequent sections, these coefficients will be non-zero.

The exogenous distribution $P(\*U)$ is defined such that for $i = 1, \dots, 4$, $U_i \sim \texttt{Logistic}(0, 1)$ is an independent variable drawn from a logistic distribution
	\begin{align}
		P \Parens{U_i < u} = \frac{1}{1 + e^{-u}}
	\end{align}
and $U \sim \texttt{Unif}(0, 1)$ is an independent variable drawn uniformly over a real interval $[0, 1]$. \hfill $\blacksquare$
\end{example}
 It is possible to define different policy spaces in the DTR environment described above, depending on the choices of input states $\*S_i$ for every action $X_i$, and parametric forms of decision rules $\pi_i$. 
\begin{example}[DTR, Policy Space]\label{exp:_3_1_dtr_space}
For the 2-stage DTR environment described in Eq.~\ref{eq:_3_1_dtr1}. We consider a policy space $\Pi_{\text{full}} = \{\langle X_1, \{S_1\} \rangle, \langle X_2, \{S_1, X_1, S_2\} \rangle\}$ satisfying the \emph{perfect recall} \citep{koller2009probabilistic}. That is, the agent determines every action based on all the past states and actions' history. More specifically, every adaptive treatment strategy (i.e., a policy) $\pi \in \Pi_{\text{full}}$ is a pair of decision rules 
 	\begin{align}
 		\pi = \Parens{\pi_1\Parens{X_1 \mid S_1}, \pi_2\Parens{X_2 \mid S_1, X_1, S_2}},
 	\end{align}
where $\pi_1$ prescribes an initial treatment $X_1$ based on the patient's initial condition $S_1$; and $\pi_2$ decides whether to continue or switch the previous plan $X_2$ based on the patient's responses $S_2$ to the previous treatment, the initial treatment, and condition $X_1, S_1$. 

One could also define a more restricted policy space $\Pi_{\text{Markov}} = \{\langle X_1, \{S_1\} \rangle, \langle X_2, \{S_2\} \rangle\}$. Every policy $\pi \in \Pi_{\text{Markov}}$ is a pair of decision rules 
\begin{align}
	\pi = \Parens{\pi_1\Parens{X_1 \mid S_1}, \pi_2\Parens{X_2 \mid S_2}},
\end{align}
where every $\pi_i$, $i = 1, 2$, prescribes a treatment $X_i$ based on the patient's condition $S_i$ at the current stage. Such policies are also referred to as Markov policies in planning literature \citep{puterman1994markov}. Note that for every Markov policy $\pi' \in \Pi_{\text{Markov}}$, one could simulate it with a general policy $\pi \in \Pi_{\text{full}}$ by setting $\pi_2(X_2 \mid S_1, X_1, S_2) = \pi'_2(X_2 \mid S_2)$. It follows that $\Pi_{\text{Markov}} \subset \Pi_{\text{full}}$, i.e., every Markov policy is contained in the general policy space $\Pi_{\text{full}}$.

Finally, we define a stationary policy space $\Pi_{\text{stationary}} \subset \Pi_{\text{Markov}}$. Every policy $\pi \in \Pi_{\text{stationary}}$ is a Markov policy satisfying an additional parametric constraint such that decision rule $\pi_i$ remains invariant across all stages $i = 1, 2$. That is, for a stationary policy $\pi = (\pi_1, \pi_2)$,
\begin{align}
	\pi_1\Parens{X_1 \mid S_1} = \pi_2\Parens{X_2 \mid S_2}
\end{align}
Since $\Pi_{\text{Markov}} \subset \Pi_{\text{full}}$, we must have $\Pi_{\text{stationary}} \subset \Pi_{\text{Markov}} \subset \Pi_{\text{full}}$. Fig.~\ref{fig:_3_1_space} shows a Venn diagram representing the relationship between $\Pi_{\text{stationary}}$, $\Pi_{\text{Markov}}$, and $\Pi_{\text{full}}$. The outer rectangle represents all possible policies over actions $X_1$ and $X_2$, including over a singleton action $X_i$, $i = 1, 2$. \hfill $\blacksquare$
 \end{example}

 The agent's performance is measured by a reward function that takes a set of reward signals in the environment as input.
\begin{definition}[Reward Function] \label{def:_3_1_reward}
	For an SCM $\M^* = \tuple{\*U, \*V, \2F, P}$, a reward function $\1R$ is a function $ \D(\*Y) \mapsto \3R$ mapping domains of a subset of endogenous variables $\*Y \subseteq \*V$ to a real value in $\3R$. Moreover, the endogenous variables $\*Y$ are called reward signals. \hfill $\blacksquare$
\end{definition}

Def.~\ref{def:_3_1_reward} covers most performance criteria in the decision-making literature. For instance, for a sequence of reward signals $\*Y = \{Y_1, \dots, Y_H\}$, the \emph{cumulative reward} $\1R_{\text{total}}(\*Y)$ is given by
\begin{align}
	\1R_{\text{total}}(\*Y) = \sum_{i = 1}^{H} Y_i. \label{eq:_3_1_reward_total}
\end{align}
When the total number of reward signals $H \to \infty$, the above cumulative reward does not necessarily converge. In this case, a reasonable criterion is to consider the \emph{average reward} given by
\begin{align}
	\1R_{\text{average}}(\*Y) = \frac{1}{H} \sum_{i = 1}^{H} Y_i. \label{eq:_3_1_reward_average}
\end{align}
The above reward function is ensured to converge as the total number of reward signals $H \to \infty$. Alternatively, let a discount factor $\gamma \in (0, 1)$ and define the \emph{discounted cumulative reward} as:
\begin{align}
	\1R_{\text{discount}}(\*Y) = \sum_{i = 1}^{H} \gamma^{i-1} Y_i. \label{eq:_3_1_reward_discount}
\end{align}
The discount factor can be interpreted in several ways; as an interest rate, the probability of living another step, or the mathematical trick for bounding the infinite sum.
\begin{definition}[Causal Decision Model]\label{def:_3_1_cdm}
	A causal decision model (CDM) is a tuple $\langle \1M^*, \Pi, \1R \rangle$ where $\1M^* = \tuple{\*U,  \*V, \2F, P}$ is an SCM, $\Pi$ is a policy space over actions $\*X \subseteq \*V$, and $\1R$ is a reward function over reward signals $\*Y \subseteq \*V$. \hfill $\blacksquare$
\end{definition}

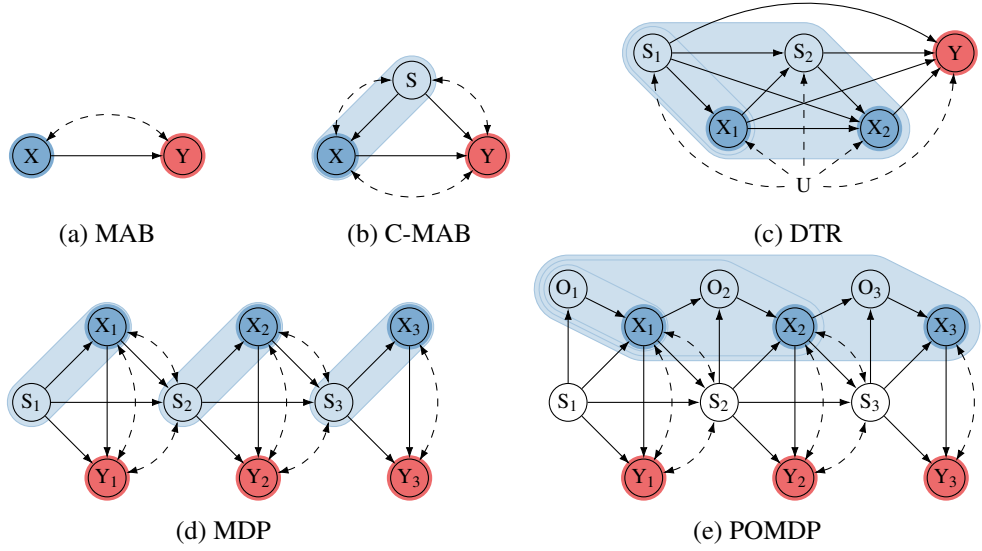
\begin{figure}[t]
	\hfill%
	\begin{subfigure}{0.2\linewidth}\centering
		\begin{tikzpicture}
			\def\outerr{3.2}
			\def\innerr{3}

			\node[vertex, opacity=0] (U) at (1, 1.7) {};
			\node[vertex] (X) at (0, 0) {X};
			\node[vertex, opacity=0] (S) at (1, 1) {};
			\node[vertex] (Y) at (2, 0) {Y};
			\draw[bidir] (X) to [bend left = 45] (Y);
			\draw[bidir, opacity=0] (X) to [bend right = 45] (Y);

			\draw[dir] (X) -- (Y);

			\begin{pgfonlayer}{back}
				\node[circle,fill=betterblue!65,draw=none,minimum size=2*\innerr mm] at (X) {};
				\node[circle,fill=betterred!65,draw=none,minimum size=2*\innerr mm] at (Y) {};
			\end{pgfonlayer}
		\end{tikzpicture}
		\caption{MAB}
		\label{fig:_3_1_mab}
	\end{subfigure}\hfill
	\begin{subfigure}{0.2\linewidth}\centering
		\begin{tikzpicture}
			\def\outerr{3.2}
			\def\innerr{3}

			\node[vertex, opacity=0] (U) at (1, 1.7) {};
			\node[vertex] (X) at (0, 0) {X};
			\node[vertex] (S) at (1, 1) {S};
			\node[vertex] (Y) at (2, 0) {Y};
			\draw[bidir] (X) to [bend right = 45] (Y);

			\draw[dir] (S) to (X);
			\draw[dir] (X) -- (Y);
			\draw[dir] (S) to (Y);
			\draw[bidir] (S) to [bend left=45] (Y);
			\draw[bidir] (S) to [bend right=45] (X);

			\begin{pgfonlayer}{back}
				\draw[fill=betterblue!25, draw = betterblue!45] \convexpath{S, X}{\outerr mm};
				\node[circle,fill=betterblue!65,draw=none,minimum size=2*\innerr mm] at (X) {};
				\node[circle,fill=betterred!65,draw=none,minimum size=2*\innerr mm] at (Y) {};
			\end{pgfonlayer}
		\end{tikzpicture}
		\caption{C-MAB}
		\label{fig:_3_1_cmab}
	\end{subfigure}\hfill
	\begin{subfigure}{0.35\linewidth}\centering
		\begin{tikzpicture}
			\def\outerr{3.5}
			\def\innerr{3}
			\node[vertex] (Z1) at (0, 1) {S\textsubscript{1}};
			\node[vertex] (X1) at (1, 0) {X\textsubscript{1}};
			\node[vertex] (Z2) at (2, 1) {S\textsubscript{2}};
			\node[vertex] (X2) at (3, 0) {X\textsubscript{2}};
			\node[vertex] (Y) at (4, 1) {Y};
			\node[vertex,draw=none] (U) at (2, -0.75) {U};

			\draw[dir] (Z1) to (Z2);
			\draw[dir] (Z1) to (X1);
			\draw[dir] (Z1) to (X2);
			\draw[dir] (Z1) to [bend left = 30] (Y);

			\draw[dir] (X1) to (Z2);
			\draw[dir] (X1) to (X2);
			\draw[dir] (X1) to (Y);

			\draw[dir] (Z2) to (Y);
			\draw[dir] (Z2) to (X2);

			\draw[dir] (X2) to (Y);

			\draw[dir, dashed] (U) to (X1);
			\draw[dir, dashed]  (U) to (X2);
			\draw[dir, dashed]  (U) to (Z2);
			\draw[dir, dashed]  (U) to [bend left = 45] (Z1);
			\draw[dir, dashed]  (U) to [bend right = 45] (Y);

			\begin{pgfonlayer}{back}
				\draw[fill=betterblue!25, draw = betterblue!45] \convexpath{X2, X1, Z1, Z2}{4 mm};
				\draw[fill=betterblue!25, draw = betterblue!45] \convexpath{X1, Z1}{\outerr mm};
				\node[circle,fill=betterblue!65,draw=none,minimum size=2*\innerr mm] at (X1) {};
				\node[circle,fill=betterblue!65,draw=none,minimum size=2*\innerr mm] at (X2) {};
				\node[circle,fill=betterred!65,draw=none,minimum size=2*\innerr mm] at (Y) {};
			\end{pgfonlayer}
		\end{tikzpicture}
		\caption{DTR}
		\label{fig:_3_1_dtr}
	\end{subfigure}\hfill\null

	\hfill
	\begin{subfigure}{0.38\linewidth}\centering
		\begin{tikzpicture}
			\def\outerr{3.2}
			\def\innerr{3}
			\node[vertex] (S1) at (-1, -1) {S\textsubscript{1}};
			\node[vertex, opacity=0] (O1) at (-1, 0.6) {};
			\node[vertex] (X1) at (0, 0) {X\textsubscript{1}};
			\node[vertex] (Y1) at (0, -2) {Y\textsubscript{1}};
			\node[vertex] (S2) at (1, -1) {S\textsubscript{2}};
			\node[vertex] (X2) at (2, 0) {X\textsubscript{2}};
			\node[vertex] (Y2) at (2, -2) {Y\textsubscript{2}};
			\node[vertex] (S3) at (3, -1) {S\textsubscript{3}};
			\node[vertex] (X3) at (4, 0) {X\textsubscript{3}};
			\node[vertex] (Y3) at (4, -2) {Y\textsubscript{3}};

			\draw[dir] (S1) to (S2);
			\draw[dir] (S1) to (X1);
			\draw[dir] (S1) to (Y1);
			\draw[dir] (X1) to (Y1);
			\draw[dir] (X1) to (S2);

			\draw[bidir] (X1) to [bend left = 30] (S2);
			\draw[bidir] (X1) to [bend left = 30] (Y1);
			\draw[bidir] (Y1) to [bend right = 30] (S2);

			\draw[dir] (S2) to (S3);
			\draw[dir] (S2) to (X2);
			\draw[dir] (S2) to (Y2);
			\draw[dir] (X2) to (Y2);
			\draw[dir] (X2) to (S3);

			\draw[bidir] (X2) to [bend left = 30] (S3);
			\draw[bidir] (X2) to  [bend left = 30] (Y2);
			\draw[bidir] (Y2) to [bend right = 30] (S3);

			\draw[dir] (S3) to (X3);
			\draw[dir] (S3) to (Y3);
			\draw[dir] (X3) to (Y3);

			\draw[bidir] (X3) to [bend left = 30] (Y3);

			\begin{pgfonlayer}{back}
				\draw[fill=betterblue!25, draw = betterblue!45] \convexpath{X1, S1}{\outerr mm};
				\draw[fill=betterblue!25, draw = betterblue!45] \convexpath{X2, S2}{\outerr mm};
				\draw[fill=betterblue!25, draw = betterblue!45] \convexpath{X3, S3}{\outerr mm};
				\node[circle,fill=betterblue!65,draw=none,minimum size=2*\innerr mm] at (X1) {};
				\node[circle,fill=betterblue!65,draw=none,minimum size=2*\innerr mm] at (X2) {};
				\node[circle,fill=betterblue!65,draw=none,minimum size=2*\innerr mm] at (X3) {};
				\node[circle,fill=betterred!65,draw=none,minimum size=2*\innerr mm] at (Y1) {};
				\node[circle,fill=betterred!65,draw=none,minimum size=2*\innerr mm] at (Y2) {};
				\node[circle,fill=betterred!65,draw=none,minimum size=2*\innerr mm] at (Y3) {};
			\end{pgfonlayer}
		\end{tikzpicture}
		\caption{MDP}
		\label{fig:_3_1_mdp}
	\end{subfigure}\hfill
	\begin{subfigure}{0.4\linewidth}\centering
		\begin{tikzpicture}
			\def\outerr{3.2}
			\def\innerr{3}
			\node[vertex] (S1) at (-1, -1) {S\textsubscript{1}};
			\node[vertex] (O1) at (-1, 0.5) {O\textsubscript{1}};
			\node[vertex] (X1) at (0, 0) {X\textsubscript{1}};
			\node[vertex] (Y1) at (0, -2) {Y\textsubscript{1}};
			\node[vertex] (O2) at (1, 0.5) {O\textsubscript{2}};
			\node[vertex] (S2) at (1, -1) {S\textsubscript{2}};
			\node[vertex] (X2) at (2, 0) {X\textsubscript{2}};
			\node[vertex] (Y2) at (2, -2) {Y\textsubscript{2}};
			\node[vertex] (O3) at (3, 0.5) {O\textsubscript{3}};
			\node[vertex] (S3) at (3, -1) {S\textsubscript{3}};
			\node[vertex] (X3) at (4, 0) {X\textsubscript{3}};
			\node[vertex] (Y3) at (4, -2) {Y\textsubscript{3}};

			\draw[dir] (S1) to (S2);
			\draw[dir] (S1) to (O1);
			\draw[dir] (O1) to (X1);
			\draw[dir] (S1) to (X1);
			\draw[dir] (S1) to (Y1);
			\draw[dir] (X1) to (Y1);
			\draw[dir] (X1) to (S2);

			\draw[bidir] (X1) to [bend left = 30] (S2);
			\draw[bidir] (X1) to [bend left = 30] (Y1);
			\draw[bidir] (Y1) to [bend right = 30] (S2);

			\draw[dir] (S2) to (S3);
			\draw[dir] (S2) to (O2);
			\draw[dir] (O2) to (X2);
			\draw[dir] (S2) to (X2);
			\draw[dir] (S2) to (Y2);
			\draw[dir] (X2) to (Y2);
			\draw[dir] (X2) to (S3);

			\draw[bidir] (X2) to [bend left = 30] (S3);
			\draw[bidir] (X2) to  [bend left = 30] (Y2);
			\draw[bidir] (Y2) to [bend right = 30] (S3);

			\draw[dir] (S3) to (O3);
			\draw[dir] (O3) to (X3);
			\draw[dir] (S3) to (X3);
			\draw[dir] (S3) to (Y3);
			\draw[dir] (X3) to (Y3);

                \draw[dir] (X1) to (O2);
                \draw[dir] (X2) to (O3);

            

			\draw[bidir] (X3) to [bend left = 30] (Y3);

			\begin{pgfonlayer}{back}
				\draw[fill=betterblue!25, draw = betterblue!45] \convexpath{X1, O1, O3, X3}{1.4*\outerr mm};
				\draw[fill=betterblue!25, draw = betterblue!45] \convexpath{X1, O1, O2, X2}{1.2*\outerr mm};
				\draw[fill=betterblue!25, draw = betterblue!45] \convexpath{X1, O1}{\outerr mm};

				\node[circle,fill=betterblue!65,draw=none,minimum size=2*\innerr mm] at (X1) {};
				\node[circle,fill=betterblue!65,draw=none,minimum size=2*\innerr mm] at (X2) {};
				\node[circle,fill=betterblue!65,draw=none,minimum size=2*\innerr mm] at (X3) {};
				\node[circle,fill=betterred!65,draw=none,minimum size=2*\innerr mm] at (Y1) {};
				\node[circle,fill=betterred!65,draw=none,minimum size=2*\innerr mm] at (Y2) {};
				\node[circle,fill=betterred!65,draw=none,minimum size=2*\innerr mm] at (Y3) {};
			\end{pgfonlayer}
		\end{tikzpicture}
		\caption{POMDP}
		\label{fig:_3_1_pomdp}
	\end{subfigure}\hfill\null
	\caption{Causal diagrams for CDMs representing canonical decision-making models.}
	\label{fig:_3_1_template}
\end{figure}

Among elements in Def.~\ref{def:_3_1_cdm}, the SCM $\1M^*$ represents the underlying environment; the policy space $\Pi$ indicates the agent's capabilities after deployed in the environment, i.e., what it could control and observe at the time of interaction; the reward function $\1R$ measures the performance of the agent, from the system designer's perspective. Formally, every CDM $\Tuple{\1M^*, \Pi, \1R}$ characterizes a planning/decision-making task \citep{bellman1966dynamic} that attempts to find an optimal strategy from the policy space $\Pi$ dictating the agent's behaviors, provided with the complete parametrization of the underlying environment $\1M^*$. An optimal policy $\pi^*$ for a CDM $\langle \1M^*, \Pi, \1R \rangle$ is a policy in space $\Pi$ that maximizes the reward function $\1R$ evaluated by the underlying SCM $\M^*$, i.e.,
\begin{align}
	 & \pi^* = \argmax_{\pi \in \Pi} \invE{ \1R \left (\*Y \right ); \1M^*}{\pi} \label{eq:_3_1_crl_opt}
\end{align}
We will graphically represent every CDM $\langle \1M^*, \Pi, \1R \rangle$ using an augmented causal diagram $\G$ constructed from the environment $\1M^*$ (Fig.~\ref{def:_2_3_diagram}); actions $\*X$ and reward signals $\*Y$ are highlighted in blue and red respectively; for every action, $X_i \in \*X$, its input states $\*S_i$ are highlighted in light blue. Fig.~\ref{fig:_3_1_template} shows the graphical representation for some canonical planning tasks.

A few observations are worth making at this point. First, the number of action variables $H = |\*X|$ represents the horizon of the decision sequence. \footnote{In episodic reinforcement learning, the agent collects data by interacting with the environment for repeated episodes $t = 1, \dots, T$. The decision horizon $H$ represents the total steps of actions $X_1, \dots, X_H$ that the agent has to decide in every episode. We will further elaborate on the episodic learning for optimizing CDMs in Sec.~\ref{sec:_3_2}.} When $H = 1$, the CDM corresponds to the single-stage decision models such as MABs \citep{robbins52}. 
On the other hand, when $H > 1$, the CDM defines a sequential decision-making problem, e.g., MDPs \citep{puterman1994markov}, when the agent has to sequentially determine values of actions $X_i$, $i = 1, \dots, H$, based on values of the observed states $\*S_i$ at the time of the intervention. 

Second, it is possible to define multiple CDMs in the same environment $\1M^*$ by changing the policy space $\Pi$ and the reward function $\1R$, resulting in different optimal policies. In other words, the optimal policy $\pi^*$ is defined with regard to the agent's capabilities to interact the environment after being deployed and how the system designer incentives its behaviors. Consequently, changing the forms of policy $\Pi$ and reward $\1R$ affects the optimal solution $\pi^*$. 
\begin{example}[CDMs in DTR] \label{exp:_3_1_cdms_dtr}
	Let $\1M^*_{\textsc{dtr}}$ be a DTR environment compatible with Fig.~\ref{fig:_3_1_dtr}. The primary outcome $Y$ is given by $Y \gets X_2 \oplus S_1$; values of $S_1$ are uniformly drawn from the binary domain $\{0, 1\}$. Let $\Pi_{\text{full}}$ and $\Pi_{\text{Markov}}$ be policy spaces defined in Example~\ref{exp:_3_1_dtr_space}. 
	
	We could define CDMs $\Tuple{\1M^*_{\textsc{dtr}}, \Pi_{\text{full}}, Y}$ and $\Tuple{\1M^*_{\textsc{dtr}}, \Pi_{\text{Markov}}, Y}$, which represent two different planning tasks. The former searches for a general policy $\pi^*_{\text{full}} \in \Pi_{\text{full}}$ determining actions $X_i$ based on the complete states and actions' history $S_1, \dots, S_i$, $X_1, \dots, X_{i-1}$; while the latter finds a Markov policy $\pi^*_{\text{Markov}} \in  \Pi_{\text{Markov}}$ selecting action $X_i$ based on the current state $S_i$. 
 
 First, for any Markov policy $\pi_{\text{Markov}} = (\pi_1, \pi_2)$ in $\Pi_{\text{Markov}}$, its expected reward is given by
	\begin{align}
		\invE{Y}{\pi_{\text{Markov}}} & = \sum_{x_2, s_1}\E[x_2 \oplus s_1] P(s_1)\sum_{s_2}\pi_2(x_2\mid s_2)P(s_2)\\
		&=0.5
	\end{align}
The last step holds since $S_1$ is uniformly drawn over the binary domain $\{0, 1\}$. Also,  solving the CDM $\Tuple{\1M^*_{\textsc{dtr}}, \Pi_{\text{full}}, Y}$ gives an optimal policy $\pi^*_{\text{full}} = (\pi^*_1, \pi^*_2)$ such that $\pi^*_2 \triangleq X_2 \gets \neg S_1$. Evaluating the expected reward of $\pi^*_{\text{full}}$ in $\1M^*_{\textsc{dtr}}$ gives
\begin{align}
	\invE{Y}{\pi^*_{\text{full}}} &= \E[S_1 \oplus \neg S_1]
\end{align}
Computing the above equation implies $\invE{Y}{\pi^*_{\text{full}}} = 1$ which outperforms the best possible Markov policy $\invE{Y}{\pi^*_{\text{Markov}}} = 0.5$. Moreover, suppose the sign of the reward function is changed to $\1R(Y) \gets - Y$. The same solution $\pi^*_{\text{full}}$ is no longer optimal in a CDM $\Tuple{\1M^*_{\textsc{dtr}}, \Pi_{\text{full}}, -Y}$ since it now minimizes the expected primary outcome $\invE{-Y}{\pi}$ instead. \hfill $\blacksquare$
\end{example}

More broadly, the formulation of CDMs permits one to represent canonical planning tasks (or equivalently, decision-making models) in the literature across disciplines, including RL and healthcare, when the detailed parametrization of the underlying environment $\1M^*$ is known. These canonical tasks are graphically represented in Fig.~\ref{fig:_3_1_template}, which we will briefly describe below.
\begin{itemize}[itemindent=-\parindent, leftmargin=*, labelwidth=*, labelindent=*]
	\item \textbf{Multi-Armed Bandit \citep{robbins52}.} Fig.~\ref{fig:_3_1_mab} is induced by a MAB model $\langle \1M^*_{\textsc{mab}}, \Pi, Y\rangle$ consisting of an arm choice $X$ and reward $Y$. Policy space $\Pi = \{\langle X, \emptyset \rangle \}$ defines a set of policies $\pi$ that selects values of action $X$ following a probability distribution $\pi(X)$. Please visit Example~\ref{exp:_3_1_mab} for a detailed instance of a MAB model.
	\item \textbf{Contextual Bandit \citep{langford-zhang08}.} Fig.~\ref{fig:_3_1_cmab} is the graphical representation of a contextual bandit (C-MAB) model $\langle \1M^*_{\textsc{cmab}}, \Pi, Y\rangle$. Compared with MABs, a context variable $S$ is now observed. The policy space $\Pi = \{\langle X, \{S\} \rangle \}$ consists of candidate policies $\pi(X|S)$ which selects values of action $X$ based on the observed context $S$.
	\item \textbf{Dynamic Treatment Regime \citep{murphy2001dtr}.} In a DTR model $\langle \1M^*_{\textsc{dtr}}, \Pi, Y \rangle$, the policy space $\Pi = \left \{\langle X_i, \{S_1, \dots, S_i, X_1, \dots, X_{i-1} \} \rangle \right \}_{i = 1}^{H}$ is a set of policies $\pi = (\pi_1, \dots, \pi_H)$ consisting of a finite sequence of decision rules. For every $i$-th stage, the decision rule $\pi_i(X_i \mid S_1, \dots, S_i, X_1, \dots, X_{i-1})$ selects values of action $X_i$ based on all the past action and states' history $S_1, \dots, S_i, X_1, \dots, X_{i-1}$. The goal is to maximize the primary outcome $Y$ after intervening on all actions $X_1, \dots, X_H$. Fig.~\ref{fig:_3_1_dtr} is the graphical representation of a $2$-stage DTR model; a detailed instance is provided in Example~\ref{exp:_3_1_dtr}.
	\item \textbf{Markov Decision Process \citep{bellman57,puterman1994markov}.} Consider a Markov decision process (MDP) model $\langle \1M^*_{\textsc{mdp}}, \Pi, \1R \rangle$. Environment $\1M^*$ consists of a set of states $\*S = \{S_i\}_{i = 1}^{\infty}$, a set of actions $\*X = \{X_i\}_{i = 1}^{\infty}$, and a set of reward signals $\*Y = \{Y_i\}_{i = 1}^{\infty}$. The policy space $\Pi = \left \{ \langle X_i, S_i \rangle \right\}_{i = 1}^{\infty}$ consists of a set of decision rules $\pi = (\pi_i(X_i \mid S_i))_{i = 1}^{\infty}$.  The reward function $\1R$ can be described as the average reward $\1R_{\text{average}}(\*Y)$ in Eq.~\ref{eq:_3_1_reward_average} or the discounted reward $\1R_{\text{discount}}(\*Y)$ in Eq.~\ref{eq:_3_1_reward_discount}. In the discounted case, rewards obtained later are discounted more than rewards obtained earlier. If the discounted factor $\gamma = 0$, the agent is said to be myopic, i.e., it is only concerned about immediate rewards. Fig.~\ref{fig:_3_1_mdp} represents the causal diagram of an MDP model spanning over steps $i = 1, 2, 3$. See  Example~\ref{exp:_3_1_mdp} for a detailed instance of a policy planning task in an MDP environment.
	\item  \textbf{Partially Observable MDP \citep{aastrom1965optimal}.} A partially observable MDP (POMDP) is a generalization of MDP in which system dynamics are determined by an MDP environment, but the agent could not directly utilize the underlying state $\*S = \{S_i\}_{i = 1}^{\infty}$ as input to determine its actions $\*X = \{X_i\}_{i = 1}^{\infty}$. Instead, it only receives a set of observation variables $\*O = \{O_i\}_{i = 1}^{\infty}$ depending on the underlying states. Fig.~\ref{fig:_3_1_pomdp} shows a POMDP model $\langle \1M^*_{\textsc{pomdp}}, \Pi, \1R \rangle$ spanning over steps  $i = 1, 2, 3$. The policy space $\Pi = \left \{ \langle X_i, \{O_1, \dots, O_i, X_1, \dots, X_{i-1}\} \rangle \right\}_{i = 1}^{\infty}$ defines a set of non-Markov policies $\pi = (\pi_i(X_i \mid O_1, \dots, O_i, X_1, \dots, X_{i-1}))_{i = 1}^{\infty}$. For every $i$-th stage of intervention, an agent following a non-Markov policy $\pi \in \Pi$ selects an action $x_i \sim \pi_i\left (X_i \mid O_1, \dots, O_i, X_1, \dots, X_{i-1} \right)$ based on all the past observations and actions.
\end{itemize}

When the policy space $\Pi$ and the reward function $\1R$ are well-specified, and detailed parameters of the underlying environment $\1M^*$ are provided, there exist efficient algorithms in the planning literature to solve for an optimal policy in a CDM $\langle \1M^*, \Pi, \1R \rangle$ \citep{bellman57,puterman1994markov,shachter1986evaluating}.
For instance, for an MDP model graphically described in Fig.~\ref{fig:_3_1_mdp}, one could obtain an optimal policy using standard dynamic programming algorithms \citep{bellman57,puterman1994markov}. The same planning procedure applies to a DTR model \citep{RSSB:RSSB389,murphy2001dtr,murphy2005generalization}, e.g., Fig.~\ref{fig:_3_1_dtr}. 
Due to the latent nature of underlying states, planning in POMDP models (e.g., \ref{fig:_3_1_pomdp}) is more computationally challenging, and requires the planning algorithm to maintain memory and possibly reason about beliefs over the states. A variety of heuristics for approximate planning in POMDPs have been proposed \citep{jaakkola1994reinforcement,hansen1998solving,hauskrecht2000value}. Optimizing policies in a general CDM has been studied under the rubrics of influence diagrams; several algorithms and approximate procedures have been proposed, including \citep{shachter1986evaluating,koller2003multi,lauritzen2001representing}.

The following examples illustrate CDMs in some canonical decision-making settings in the literature, together with the planning procedure for computing the optimal policy.
\begin{example}[MAB Planning]\label{exp:_3_1_mab}
	Consider a CDM described by the tuple
	\begin{align}
		\langle \1M^* = \1M^*_{\textsc{mab}}, \Pi = \{\langle X, \emptyset \rangle \}, \1R(Y) = Y \rangle,
	\end{align}
	where $\1M^*_{\textsc{mab}}$ is the MAB environment described in Example~\ref{exp:_2_1_mab}. Fig.~\ref{fig:_3_1_mab} shows the graphical representation of this CDM where action $X$ and reward $Y$ are highlighted in blue and red respectively.

	Every policy $\pi(X) \in \Pi$ is a probability distribution over domains of action $\D(X) = \{0, 1\}$. Evaluating the expected reward $Y$ in submodel $\1M^*_{\textsc{mab}_\pi}$ gives
	\begin{align}
		\invE{Y}{\pi} & =  \sum_{x} \sum_{u} \E \left[ Y \mid x, u \right] P(u) \pi(x)     \\
		              & = \invE{Y}{X \gets 0} \pi(X = 0) +  \invE{Y}{X \gets 1} \pi(X = 1)
	\end{align} 
	The last step follows from marginalizing over the exogenous variable $U$. Note that evaluation of expected rewards $\invE{Y}{x}$ of atomic interventions $\doo(X \gets x)$ is provided in  Example~\ref{exp:_2_2_mab3}. Replacing interventional queries $\invE{Y}{x}$ in the above equation gives
	\begin{align}
		\invE{Y}{\pi} & = 0.4 \pi(X = 0) + (0.4 - \Delta)\pi(X = 1)                        \\
		              & = 0.4 - \Delta \pi(X = 1)
	\end{align}
	The last step follows from $\sum_{x} \pi(x) = 1$. Since the coefficient $\Delta > 0$, the reward function $\1R(Y) = Y$ evaluated in $\M^*_{\textsc{mab}}$ is maximized when probability $\pi(X = 1) = 0$, That is, the optimal policy is determinsitic $\pi^*: X \gets x^*$ with the optimal arm choice $x^* = 0$. \hfill $\blacksquare$
\end{example}
\begin{example}[MDP Planning, Dynamic Programming]\label{exp:_3_1_mdp}
	Consider the MDP environment $\1M^*_{\textsc{mdp}}$ given by Eq.~\ref{eq:_2_1_mdp}. We are interested in optimizing the MDP model described by a CDM given by
	\begin{align}
		\Tuple{\1M^* = \1M^*_{\textsc{mdp}}, \Pi = \{\langle X_i, \{S_i\} \rangle \}_{i=1}^{\infty}, \1R_{\text{discount}}(\*Y) }
	\end{align}
where $\1R_{\text{discount}}(\*Y)$ is the discounted reward function given by Eq.~\ref{eq:_3_1_reward_discount} with $\gamma = 0.9$. We will focus on stationary policies $\pi = \Parens{\pi_i(X_i \mid S_i)}_{i = 1}^{\infty}$ such that decision rules $\pi_1 = \pi_2 = \dots$ remain invariant across decision horizons $i = 1, 2, \dots$. \footnote{Indeed, it has been shown that there always exists a stationary policy that could optimize the cumulative reward in an MDP model \citep{filar2012competitive}. It thus suffices to focus on stationary policies.} 

	Let $\D(S)$ and $\D(X)$ denote the domain of state $S_i$ and action $X_i$ at every stage $i$, respectively. For any stationary policy $\pi \in \Pi$, a state-action value function $Q_{\pi}: \D(S) \times \D(X) \to \3R$ (also called a Q-function) is defined as the expected cumulative reward following policy $\pi$ given the starting state $s$ and initial action $x$, i.e.,
	\begin{align}
		Q_{\pi}(s, x) = \invE{ \sum_{j = 0}^{\infty} \gamma^{j} Y_{i+j} \mid S_i = s, X_i = x}{\pi}
	\end{align}
Since the structural functions $f_{S_i}$ and $f_{Y_i}$ remains invariant across decision horizon $i = 1, 2, \dots$, for any policy $\pi$, any state $s$, and any action $x$, the above expression can be recursively defined in terms of a so-called \emph{Bellman Equation} \citep{bellman1966dynamic}:
	\begin{align}
		Q_{\pi}(s, x) &= \invE{  Y_i + \gamma Y_{i+1} + \gamma^2 Y_{i+2} + \cdots \mid S_i = s, X_i = x}{\pi}\\
		&=\invE{  Y_i + \gamma Q_{\pi}(S_{i+1}, X_{i+1})  \mid S_i = s, X_i = x}{\pi}
	\end{align}
	The last step follows from the recursive definition of the value function $Q_{\pi}(s, x)$ and the Markov property in the interventional distribution; see Example~\ref{exp:_2_2_mdp4} for details. The above equation could be further written as, by expanding on next state $S_{i+1}$ and action $X_{i+1}$,
	\begin{align}
		Q_{\pi}(s, x) &= \invE{  Y_i \mid S_i = s, X_i = x}{\pi} + \gamma \invE{Q_{\pi}(S_{i+1}, X_{i+1})  \mid S_i = s, X_i = x}{\pi}\\
		&= \invE{  Y_i \mid S_i = s, X_i = x}{\pi}  \\
		&\qquad + \gamma \sum_{s'} \inv{S_{i+1} = s' \mid S_i = s, X_i \gets x}{\pi}  \sum_{x'}\pi_{i+1}(x'\mid s') Q_{\pi}(s', x')
	\end{align}
	Since the transition distribution and the conditional reward remain invariant across atomic and policy interventions (Eqs.~\ref{eq:_2_2_mdp6} and \ref{eq:_2_2_mdp7}), the above equation could be further written as
	\begin{align}
		Q_{\pi}(s, x) &= \1R_{\text{exp}}(s, x) + \gamma \sum_{s'} \1T_{\text{exp}}(s, x, s') \sum_{x'}\pi_{i+1}(x'\mid s') Q_{\pi}(s', x')
	\end{align}
	where the transition probability $\1T_{\text{exp}}$ and the reward function $\1R_{\text{exp}}$ are given by
	\begin{align}
    	&\1T_{\text{exp}}(s, x, s') = \inv{S_{i+1} = s' \mid S_i = s}{X_i \gets x} \label{eq:_3_2_mdp_t}\\
    	&\1R_{\text{exp}}(s, x) = \invE{Y_i \mid S_i = s}{X_i \gets x} \label{eq:_3_2_mdp_r}
	\end{align}
	Detailed parametrizations of interventional quantities $\1T$ and $\1R$ are provided in Fig.~\ref{fig:_2_2_inv_mdp}. 
	
	\begin{table}[t]
			\centering
			\renewcommand{\arraystretch}{1.25}
			\begin{tabular}{|cc|c|cc |c|}
				$S$ & $X$ & $Q_*$     & $S$ & $X$ & $Q_*$     \\
				\hline
				\hline
				0     & 0     & \textbf{8.2} & 1     & 0     & 7.56 \\
				0     & 1     & 7.56         & 1     & 1     & \textbf{8.2}         \\
			\end{tabular}
		\caption{Optimal Q-function $Q_*(s, x)$ evaluated in the MDP model of Example~\ref{exp:_3_1_mdp}.}
		\label{tab:_3_1_mdp}
	\end{table}

	An optimal policy $\pi^*$ is such that $Q_{\pi^*}(s,x) \geq Q_{\pi}$ for all state-action pair $s, x$ and all policies $\pi$. Optimizing Q-function leads to an expression called the \emph{Bellman optimality equation}, i.e,
	\begin{align}
		Q_{*}(s, x) = \1R_{\text{exp}}(s, x) + \gamma \sum_{s'} \1T_{\text{exp}}(s, x, s') \max_{x'} Q_{*}(s', x') \label{eq:_3_2_mdp_optimal_bellman}
	\end{align}
	The optimal policy $\pi^*$ is given by, for every stage $i = 1,2, \dots$, 
	\begin{align}
		\pi^*_i(S_i = s) = \argmax_{x} Q_*(s, x) 
	\end{align}
	for any state $s \in \D(S)$. 
We can compute the optimal Q-function evaluated in $\1M^*_{\textsc{mdp}}$ using value iteration \citep{sutton1998reinforcement}. Detailed parametrizations are provided in Table~\ref{tab:_3_1_mdp}. Complete computations are provided in Appendix~\ref{app:_compute}.
The optimal policy $\pi^* = \Parens{\pi^*_i(X_i \mid S_i)}_{i = 1}^{\infty}$ is given by $\pi^*_i \triangleq X_i \gets S_i$, for every $i = 1, 2, \dots$. Evaluating its expected return gives $\invE{\sum_{i=1}^{\infty} \gamma^{i-1}Y_i}{\pi^*} = 8.2$; detailed derivation steps are provided in Example~\ref{exp:_2_2_mdp5}. \hfill $\blacksquare$
\end{example}
\begin{example}[DTR Planning, Dynamic Programming]\label{exp:_3_1_dtr2}
	Consider the DTR environment $\1M^*_{\textsc{dtr}}$ in Eq.~\ref{eq:_3_1_dtr1}. Our goal is to maximize the patient's days of abstinence $Y$ over 12 months after the treatment. This decision-making problem is described by a CDM given by
	\begin{align}
		\langle \1M^* = \1M^*_{\textsc{dtr}}, \Pi = \{\langle X_1, \{S_1\} \rangle, \langle X_2, \{S_1, X_1, S_2\} \rangle\}, \1R(Y) = Y \rangle 
	\end{align}
	Fig.~\ref{fig:_3_1_dtr} describes a causal diagram of $\1M^*_{\textsc{dtr}}$ where actions $X_1, X_2$ are highlighted in blue, primary outcome $Y$ in red, and input covariates $\{S_1\}$ and $\{S_1, X_1, S_2\}$ in light blue.

	\begin{table}[t]
		\begin{subtable}{\linewidth}
			\centering
			\renewcommand{\arraystretch}{1.25}
			\begin{tabular}{|cc|c|cc |c|}
				$S_1$ & $X_1$ & $Q^{(1)}_*$     & $S_1$ & $X_1$ & $Q^{(1)}_*$     \\
				\hline
				\hline
				0     & 0     & \textbf{0.8799} & 1     & 0     & \textbf{0.4716} \\
				0     & 1     & 0.8749          & 1     & 1     & 0.1003          \\
			\end{tabular}
			\caption{$Q^{(1)}_*(s_1, x_1)$}\label{tab_3_1_a}
		\end{subtable}%
		\vspace{0.1in}
		\begin{subtable}{\linewidth}
			\centering
			\renewcommand{\arraystretch}{1.25}

			\begin{tabular}{|cccc|c|cccc |c|}
				$S_1$ & $X_1$ & $S_2$ & $X_2$ & $Q^{(2)}_*$     & $S_1$ & $X_1$ & $S_2$ & $X_2$ & $Q^{(2)}_*$     \\
				\hline
				\hline
				0     & 0     & 0     & 0     & 0.7851          & 1     & 0     & 0     & 0     & 0.2149          \\
				0     & 0     & 0     & 1     & \textbf{0.9846} & 1     & 0     & 0     & 1     & \textbf{0.7851} \\
				0     & 0     & 1     & 0     & 0.7851          & 1     & 0     & 1     & 0     & 0.2149          \\
				0     & 0     & 1     & 1     & \textbf{0.7851} & 1     & 0     & 1     & 1     & \textbf{0.2149} \\
				0     & 1     & 0     & 0     & 0.2149          & 1     & 1     & 0     & 0     & 0.0008          \\
				0     & 1     & 0     & 1     & \textbf{0.9846} & 1     & 1     & 0     & 1     & \textbf{0.2149} \\
				0     & 1     & 1     & 0     & 0.2149          & 1     & 1     & 1     & 0     & 0.0008          \\
				0     & 1     & 1     & 1     & \textbf{0.7851} & 1     & 1     & 1     & 1     & \textbf{0.0154} \\
			\end{tabular}
			\caption{$Q^{(2)}_*(s_1, x_1, s_2, x_2)$}\label{tab_3_1_b}
		\end{subtable}%
		\caption{Evaluation of optimal Q-functions evaluated in the DTR system of Example~\ref{exp:_3_1_dtr}.}
		\label{tab_3_1}
	\end{table}

	For every policy $\pi \in \Pi$, evaluating recovery rate $Y$ in submodel $\1M^*_{\textsc{dtr}_\pi}$ gives
	\begin{align}
		\invE{Y}{\pi} = \sum_{s_1, x_1, s_2, x_2} \pi_2(x_2 \mid s_1, x_1,s_2) \pi_1(x_1 \mid s_1) \bigg (\sum_{u} & \E \left[Y \mid s_1, x_1, s_2, x_2, u \right]  \notag           \\
	& P\left(s_2 \mid s_1, x_1, u \right )  P(s_1 \mid u) P(u) \bigg)
	\end{align}
	Since $\pi_1, \pi_2$ are not functions of the exogenous variable $U$, summing over domain of $U$ we obtain
	\begin{align}
		\invE{Y}{\pi} = \sum_{s_1, x_1, s_2, x_2} \invE{Y\mid s_1, s_2}{x_1, x_2} \pi_2(x_2 \mid s_1, x_1,s_2) \inv{s_2 \mid s_1}{x_1} \pi_1(x_1 \mid s_1) P(s_1)
	\end{align}
	As shown in \citep{murphy2001dtr}, the optimal policy $\pi^*$ is deterministic, and satisfies the Bellman equation \citep{bellman57}
	\begin{align}
		\pi^*_1(s_1)           & = \argmax_{x_1} Q^{(1)}_*\left(s_1, x_1 \right)          \\
		\pi^*_2(s_1, x_1, s_2) & = \argmax_{x_2} Q^{(2)}_*\left(s_1, s_2, x_1, x_2\right)
	\end{align}
	The optimal Q-function is
	\begin{align}
		Q^{(1)}_*(s_1, x_1)                      & = \sum_{s_2} \max_{x_2} Q^{(2)}_*\left(s_1, s_2, x_1, x_2\right) \inv{s_2\mid s_1}{x_1} \\
		Q^{(2)}_*\left(s_1, s_2, x_1, x_2\right) & = \invE{Y\mid s_1, s_2}{x_1, x_2}
	\end{align}
We compute the parametrization of Q-functions evaluated in $\1M^*$ and provide them in Table~\ref{tab_3_1_a}. See Appendix~\ref{app:_compute} for complete computation.
The optimal actions $\pi^*_1(s_1)$, $\pi^*_2(s_1, x_1, s_2)$ given every state-action's history is highlighted. Solving for an optimal policy gives $\pi_1^* \triangleq X_1 \gets 0$ and $\pi^*_2 \triangleq X_2 \gets 1$. In words, in order to maximize the patient's days of abstinence, the physician should start with behavioral therapy and follow up with an intensive treatment combining both behavioral therapy and medication.
	The expected reward of this policy $\pi^*$ is computable as:
	\begin{align}
		\invE{Y}{\pi^*} & = \sum_{s_1} Q^{(1)}_*(s_1, X_1 = 0) P(s_1)                                                           \\
		                & = Q^{(1)}_*(S_1 = 0, X_1 = 0) P(S_1 = 0) + Q^{(1)}_*(S_1 = 1, X_1 = 0) P(S_1 = 1) \label{eq:_3_3_dtr3}
	\end{align}
	Evaluating the above equation gives the optimal expected reward $\invE{Y}{\pi^*}  = 0.6758$. \hfill $\blacksquare$
\end{example}

\subsection{Causal Reinforcement Learning Tasks}\label{sec:_3_2}
The causal decision model described so far assumes the full knowledge of the underlying environment. However, in many real-world practical applications, the detailed parametrization of the environment is very rarely known, which means that standard planning algorithms are not immediately applicable. In order for the agent to optimize the performance of the underlying system, a learning process must take place, leading to the learning paradigm of \emph{causal reinforcement learning}.

To make the argument more precise, we start the discussion of a CDM  $\langle \1M^*, \Pi, \1R \rangle$. Recall that it defines a planning task of finding an optimal policy in space $\Pi$ that maximizes the reward function $\1R$ evaluated in the environment $\1M^*$. A CRL agent $\3C$ is assumed to have access to the policy space $\Pi$ and the performance measurement $\1R$. 
\footnote{The policy space $\Pi$ and the reward function $\1R$ are provided in most of the learning tasks considered in this paper. However, there exist practical applications where the reward function $\1R$ is not fully known. In this case, the agent has to ``guess'' a surrogate reward function from a hypothesis class $\3R$ and then compute an optimal policy estimate with it. 
This setting is studied under the rubric of causal imitation learning in Sec.~\ref{sec:_8_imitation}.} 
However, the underlying environment $\1M^*$ is not fully known. Instead, the agent $\3C$ only has access to some structural assumptions $\1A$ encoding qualitative knowledge about the environment $\1M^*$, and a learning regime $\1L$ dictating how it interacts with the environment $\1M^*$ to collect data. This partial knowledge constitutes new dimensions for the task formulation of optimal decision-making under uncertainty, which we will briefly discuss below. 

\paragraph{Learning Regimes ($\1L$).} Following the discussion of the PCH, each CRL agent may be able to interact with $\M^*$ in different ways, including through passive observations (i.e., $\1L = \text{see}$) or by active interventions ($\1L = \text{do}$). These learning regimes model distinct types of interactions of the agent with the environment. The former corresponds to off-policy reinforcement learning tasks \citep{li2011unbiased,li2014minimax}; while the latter corresponds to online reinforcement learning tasks \citep{auer2002finite}. More specifically, an agent passively observing the environment does not actively determine actions. Instead, it receives observational data $\1D \sim P(\*V)$ summarizing trajectories of another agent (e.g., a human demonstrator) already operating in the environment, following a behavioral policy. On the other hand, an agent may actively control actions $\*X$ by performing interventions $\doo(\pi)$, following some policies $\pi$, and receiving experimental data $\1D \sim \inv{\*V}{\pi}$. We will investigate these reinforcement learning algorithms in depth later on in this paper.

\paragraph{Structural Assumptions ($\1A$).} Structural assumptions specify a hypothesis class of possible environmental models that the agent is operating with. One common way of specifying assumptions about the SCM $\M^*$ is through a causal diagram $\1G$ (Def.~\ref{def:_2_3_diagram}). The hypothesis class $\3M$ is thus defined as the family of SCMs $\1M$ compatible with the diagram $\G$, i.e., $\G(\1M) = \G$. For instance, the causal diagram in Fig.~\ref{fig:_3_1_mdp} specifies a family of MDP environments consisting of states $S_i$, actions $X_i$, and reward signals $Y_i$, for $i = 1, 2, \dots$; the underlying transition and reward distributions remain un-specified. Other assumptions include constraints over the features of the behavior policy \citep{pearl:rob95}, or equivalence classes of causal diagrams \citep{zhang2008causal}, to cite a few.

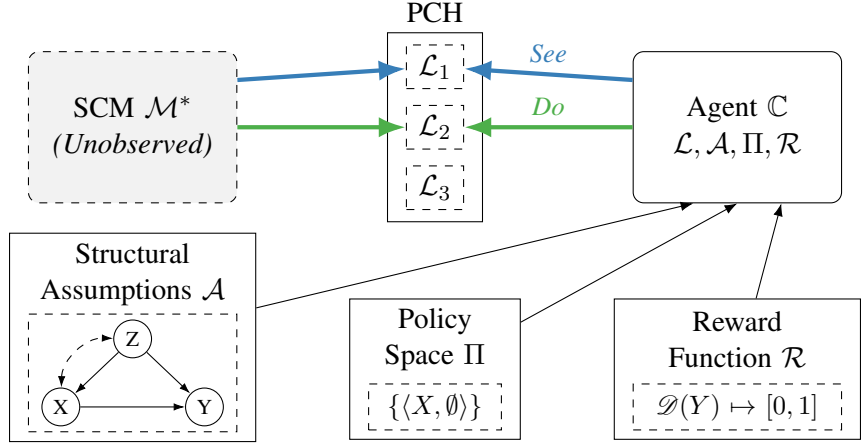
\begin{figure}[t]
	\centering
	\begin{tikzpicture}
		\newcommand*{\smallgap}{0.25}
		\newcommand*{\biggap}{1}

		\node (env1) [environment, fill=gray!10, dashed, minimum height=2cm, text width=2.5cm] at (2, 1) {SCM $\M^*$ \\ \textit{(Unobserved)}};

		\node (agent) [agent, minimum height=2cm, text width=2.5cm] at (10, 1) {Agent $\3C$ \\ $\1L, \1A, \Pi, \1R$};

		\node [draw, dashed, text width = 0.5cm, text centered] (obs1) at (6, 1.8) {$\1L_1$};
		\node [draw, dashed, text width = 0.5cm, text centered] (exp1) at (6, 1) {$\1L_2$};
		\node [draw, dashed, text width = 0.5cm, text centered] (ctf1) at (6, 0.2) {$\1L_3$};

		\node [draw, text width = 1cm, text centered, minimum height = 2.5 cm, label=above:{PCH}] (int1) at (6, 1) {};

		\node [draw, text height = 1.6cm, text width=2cm, align=center] (space) at (6,-2.2) {};
		\node [text width=2cm, align=center] at (6,-1.8) {Policy Space $\Pi$};
		\node [draw, dashed, text width=1.5cm, align=center] at (6,-2.7) {\small{$\left\{\langle X, \emptyset \rangle \right \}$}};

		\node [draw, text height = 1.6cm, text width=3cm, align=center] (reward) at (10,-2.2) {};
		\node [text width=3cm, align=center] at (10,-1.8) {Reward Function $\1R$};
		\node [draw, dashed, text width=2.5cm, align=center] at (10,-2.7) {\small{$\D(Y) \mapsto [0, 1]$}};

		\node [draw, text height = 2.5cm, text width=3cm, align=center] (assumption) at (2,-1.78) {};
		\node[text width=3cm, align=center] at (2, -0.95) {Structural Assumptions $\1A$};
		\node [draw, dashed, text height = 1.3cm, text width=2.5cm, align=center] (diagram) at (2,-2.25) {};

		\path [Latex-, betterblue, ultra thick] (obs1) edge node [anchor=south]{\small{$\textit{See}$}} ($(agent.west)+(0, 2.5*\smallgap)$);
		\path [Latex-, bettergreen,ultra thick] (exp1) edge node [anchor=south]{\small{$\textit{Do}$}} (agent.west);


		\draw[arrow] (space) to (agent.270);
		\draw[arrow] (assumption) to (agent.240);
		\draw[arrow] (reward) to (agent.300);

		\node[vertex] (X) at (1.05, -2.7) {X};
		\node[vertex] (Z) at (2, -1.8) {Z};
		\node[vertex] (Y) at (2.95, -2.7) {Y};
		\draw[bidir] (X) to [bend left = 45] (Z);
		\draw[dir] (Z) -- (X);
		\draw[dir] (X) -- (Y);
		\draw[dir] (Z) -- (Y);
		\path [-Latex, betterblue, ultra thick]  ($(env1.east) + (0, 2.5*\smallgap)$) edge  (obs1);
		\path [-Latex, bettergreen, ultra thick] (env1.east) edge  (exp1);

	\end{tikzpicture}
	\caption{Graphical representation of a causal reinforcement learning task}
	\label{fig:_3_2_crl_agent}
\end{figure}
Consider again a CDM $\Tuple{\1M^*, \Pi, \1R}$, where the model of the environment $\1M^*$ is not fully revealed to the agent. Replacing $\1M^*$ with the learning regime $\1L$ and structural assumptions $\1A$ leads to a new signature $\Tuple{\1L, \1A, \Pi, \1R}$ which characterizes a \emph{causal reinforcement learning task}. Formally,
\begin{definition}[Causal Reinforcement Learning Task]\label{def:_3_2_task}
For an SCM $\1M^* = \Tuple{\*U, \*V, \2F, P}$, a CRL task $\1T$ in the environment $\1M^*$ is a 4-tuple $\tuple{\1L, \1A, \Pi, \1R}$, where
	\begin{enumerate}
		\item $\1L$ is a learning regime of an agent's interaction with the SCM $\1M^*$, possibly \emph{see} or \emph{do};
		\item $\1A$ is a set of structural assumptions about the SCM $\M^*$;
		\item $\Pi$ is a policy space over actions $\*X$;
		\item $\1R$ is a reward function over reward signals $\*Y$. \hfill $\blacksquare$
	\end{enumerate}
\end{definition}
Formally, every CRL task $\tuple{\1L, \1A, \Pi, \1R}$ describes an optimal decision-making problem under uncertainties about the underlying environment. Provided with input $\tuple{\1L, \1A, \Pi, \1R}$, the CRL agent attempts to estimate an optimal policy $\pi^* \in \Pi$ defined in Eq.~\ref{eq:_3_1_crl_opt} maximizing the reward $\1R$ evaluated in the unknown environment $\1M^*$. Since $\1M^*$ is not fully observed, we substitute it with the learning regime $\1L$ and structural assumptions $\1A$ about the environment, depending on the specific task. 
The goal of the agent is then to find a policy $\pi^*$ such that
\begin{align}\label{eq:opt-crl-v2}
	 & \pi^* = \argmax_{\pi \in \Pi} \invEE{ \1R \left (\*Y \right ) \bigg \vert \; \1A, \1L}{\pi}{\1M^*}
\end{align}
Compared to the optimization given by  Eq.~\ref{eq:_3_1_crl_opt}, we move the unobserved SCM that evaluates the agent as a superscript and leave what the agent has access to as part of the conditioning set. 

For instance, Fig.~\ref{fig:_3_2_crl_agent} shows a graphical instance of a CRL task $\1T$ in an unknown bandit environment $\1M^*$ consisting of an arm choice $X$, reward $Y$, and a covariate $Z$. The goal is to find an optimal policy $X \gets x^*$ in the policy space $\Pi = \Braces{X, \emptyset}$ maximizing the expected reward $\invE{\1R(Y)}{x}$. The agent could passively observe the environment ($\1L = \text{See}$) and receives observational data drawn from $P(X, Y, Z)$; it could also actively intervene on the arm ($\1L = \text{Do}$) and receive interventional data drawn from $\inv{Y, Z}{x}$. The causal diagram $\1A = \G_{\text{backdoor}}$ (bottom left) encodes structural knowledge that the agent has about the environment: there is no spurious correlation between $Z$ and $Y$ and between $X$ and $Y$. 

\begin{algorithm}[t]
	\caption{Causal Reinforcement Learning Agent $\3C$}
	\label{alg:_3_2_crl_agent}
	\begin{algorithmic}[1]
		\Require CRL task $\1T = \langle \1L, \1A, \Pi, \1R \rangle$
		\Ensure a policy estimate $\hat{\pi} \in \Pi$ optimizing a CDM $\tuple{\1M^*, \Pi, \1R}$.
		\State Let data $\1D = \{\}$.
		\ForAll{every episode $t = 1, \dots, T$}
		\State Interact with the SCM $\1M^*$ following regime $\1L$ and receive samples $\*V^{(t)} \sim P_{\1L}\left(\*V; \1M^* \right)$.
		\State Update data $\1D = \1D \cup \Braces{\*V^{(t)}}$.
		\EndFor
		\State \Return an empirical estimate $\hat{\pi} \in \Pi$ of the optimal policy $\pi^*$ from data $\1D$ and assumptions $\1A$.
	\end{algorithmic}
\end{algorithm}

Alg.~\ref{alg:_3_2_crl_agent} provides pseudo-code describing the general learning strategy of a CRL agent to solve a task $\tuple{\1L, \1A, \Pi, \1R}$. We follow the episodic reinforcement learning setting \citep{sutton1998reinforcement} where the agent interacts with the environment $\1M^*$ for repeated episodes $t =1, \dots, T$. For each episode $t$, the agent interacts with the environment following the learning regime $\1L$ and receives sample $\*V^{(t)}$, consisting of realized actions $\*X^{(t)}$, observed states $\*S^{(t)}$, reward signals $\*Y^{(t)}$, and other endogenous variables. For the observational regime $\1L = \textit{see}$, the CRL agent passively observes another agent, currently deployed, to determine values of every action $X_i \in \*X$ following a behavioral policy $f_{\*X}$. For the interventional regime $\1L = \textit{do}$, the CRL agent actively intervenes on every action $X_i \in \*X$ following a policy $\pi$, and receives subsequent states $\*S$ and rewards $\*Y$. Finally, the agent $\3C$ computes an empirical estimate of an optimal policy $\pi^*$ in the policy space $\Pi$ from the combination of the collected data $\1D$ and structural assumptions $\1A$. 

The formulation of the task signature and the CRL agent summarizes existing policy learning problems and learning strategies in the reinforcement learning and causal inference literature. The following examples illustrate CRL tasks in single-stage decision-making settings.
\begin{example}[Off-Policy Learning \citep{sutton1998reinforcement}, MAB]\label{exp:_3_2_mab1}
	Consider first a MAB model $\tuple{\1M^*, \{\langle X, \emptyset \rangle\}, Y}$ graphically described in Fig.~\ref{fig:_3_1_mab}. A CRL agent $\3C$ aims to learn an arm
\begin{align}
x^* = \argmax_{x} \invE{Y}{x} 
\end{align}
	with the maximal expected reward. 
The detailed parametrization of SCM $\1M^*$ is unknown. 
Instead, the agent could only passively observe the environment and receive the observational distribution $P(X, Y)$ evaluated in $\1M^*$. 
This leads to an \emph{off-policy learning} task described through the following signature:
	\begin{align}
		\1T_{\text{off}} =\tuple{\1L =\text{see}, \1A=\text{NUC}, \Pi =  \{\langle X, \emptyset \rangle\}, \1R = Y}, \label{eq:_3_2_off}
	\end{align}
	\text{NUC} stands for the assumption of ``No Unmeasured Confounder'': there is no unobserved confounder affecting the action $X$ and the reward $Y$ simultaneously. 
	All the observed correlations between $X$ and $Y$ are fully explained by the causal relationships among them, which implies that
	\begin{align}
		\inv{Y}{x} = P(Y \mid X = x)
	\end{align}
Therefore, the agent could evaluate the expected reward of every arm $x$ from the observational data $P(X, Y)$ \footnote{We assume that the number of the observational data is sufficient, and joint distribution $P(X, Y)$ is recovered.}, and find optimal an optimal arm $x^*$ with the maximal empirical reward estimates.  \hfill $\blacksquare$
\end{example}
\begin{example}[Online Learning \citep{sutton1998reinforcement}, MAB]\label{exp:_3_2_mab2}
	We will continue with the previous example of the MAB model $\tuple{\1M^*, \{\langle X, \emptyset \rangle\}, Y}$. Suppose the CRL can now actively intervene in the underlying environment $\1M^*$. This leads to an \emph{online learning} task described by a signature
	\begin{align}
		\1T_{\text{on}} = \tuple{ \1L =\text{do}, \1A=\emptyset, \Pi =  \{\langle X, \emptyset \rangle\}, \1R = Y}, \label{eq:_3_2_on}
	\end{align}
	In order to evaluate every arm $x$, the CRL agent performs an intervention $\doo(x)$ in SCM $\1M^*$ and receives subsequent reward signals drawn from $\inv{Y}{x}$. The expected reward $\invE{Y}{x}$ is thus estimable from the experimental data by computing the empirical means.  \hfill $\blacksquare$
\end{example}
\begin{example}[Causal Identification \citep{pearl:2k}, MAB] \label{exp:_3_2_mab3}
	Consider the off-policy learning task of Eq.~\ref{eq:_3_2_off} again. 
Suppose now that the NUC assumption no longer holds. 
Instead, a previously unobserved covariate $Z$ is now revealed; the CRL agent has access to a more detailed causal diagram $\G_{\text{backdoor}}$ (Fig.~\ref{fig:_3_2_crl_agent}, bottom left) describing the underlying environment $\1M^*$. 
Replacing the NUC with structural assumptions encoded in diagram $\G_{\text{backdoor}}$ leads to a \emph{causal identification} task
	\begin{align}
		\1T_{\text{id}} =\tuple{\1L =\text{see}, \1A=\G_{\text{backdoor}}, \Pi =  \{\langle X, \emptyset \rangle\}, \1R = Y}, \label{eq:_3_2_id}
	\end{align}
	Like off-policy learning, the CRL agent observes the environment and receives the observational distribution $P(X, Y, Z)$ evaluated in $\1M^*$. Provided with the causal diagram $\G_{\text{backdoor}}$, applying the backdoor adjustment formula \cite[Ch.~3.3]{pearl:2k} implies
	\begin{align}
		\inv{y}{x} = \sum_{z} P\left( y \mid z, x \right)P \left(x \right)
	\end{align}
That is, the expected reward of every arm $x$ is computable from the observational data $P(X, Y, Z)$. Optimizing the expected reward over the action domain $X$ leads to an optimal arm.  \hfill $\blacksquare$
\end{example}

\begin{table}[t]
  \setlength{\tabcolsep}{4.2pt}
  \centering
  \begin{tabular}{@{}p{0.2cm}p{3cm}|p{2cm}p{2.5cm}p{2cm}p{2cm}|p{1.5cm}@{}}
  \multicolumn{2}{c}{} & \multicolumn{4}{|c|}{\textbf{Signature}} & \\
  \toprule
  &\textbf{Task} &  \textbf{Learning \newline Regime} \newline ($\1L$) & \textbf{Structural \newline Assumptions} ($\1A$) & \textbf{Policy \newline Space} \newline ($\Pi$) & \textbf{Reward \newline Function} \newline ($\1R$) &  \textbf{Section} \\ \midrule
  1& Off-policy  \newline Learning & See &   \cellcolor{gray!25} NUC  & $\Pi_{\textsc{exp}}$ & $\D(\*Y) \mapsto \3R$ & \ref{sec:_4_1} \\ \midrule
  2& Online  \newline Learning  & \cellcolor{gray!25} Do & -  & $\Pi_{\textsc{exp}}$ & $\D(\*Y) \mapsto \3R$ & \ref{sec:_4_2} \\ \midrule
  3& Causal \newline Identification & See& \cellcolor{gray!25} DAG $\G$  &  $\Pi_{\textsc{exp}}$ & $\D(\*Y) \mapsto \3R$  & \ref{sec:_4_3} \\ \bottomrule
  \end{tabular}
  \caption{Summary of causal reinforcement learning tasks investigated in this paper, in terms of their signatures and sections. We highlight in gray the most distinct feature introduced by the task. }
  \label{tab:_2_4_roadmap}
\end{table}

Broadly speaking, the formalization of the CRL dimensions and the corresponding tasks, semantically defined through structural causal models, allows us to describe most of the popular learning settings studied in the literature. 
It also enables us to explore and study novel CRL learning tasks beyond the current literature and that arises naturally in real-world applications. 
We first summarize the more traditional RL-CI tasks with their corresponding signatures in Table~\ref{tab:_2_4_roadmap}. 
Specifically, we will discuss in Sec.~\ref{sec:_4}, the policy learning methods for traditional CI and RL tasks through the language of CRL. 
In practice, this encompasses tasks such as off-policy and online learning and causal identification. All of these focus on optimizing policies within an experimental policy space $\Pi_{\textsc{exp}}$ (Def.~\ref{def:_3_space}). 
These tasks can be viewed as variations of the first two dimensions described in the table, namely, the interactive regime and the structural assumptions, while the other dimensions remain constant. 
Even though these tasks are prevalent in current literature, certain critical conditions weren't formally understood prior to our new formalization.
For instance, determining the validity of an off-policy method (e.g., inverse propensity weighting and dynamic programming), specifically, whether it can find a policy consistent with optimal one given by the underlying SCM.
Analyzing these classical tasks through CRL perspective will be instrumental in illuminating other foundational issues and illustrating how causal and RL formalisms intersect. 

\subsection{Comparison with Markov Decision Processes}\label{sec:_3_3}
The CRL tasks described so far (Def.~\ref{def:_3_2_task}) assume that the agent has access to either the learning regime $\1L$ under which the data are collected, or structural assumptions $\1A$ encoding causal invariances about the environment. 
In this section, we will demonstrate that such causal knowledge is generally indispensable (necessary) for learning an optimal policy in an unknown SCM. Our discussion focuses on standard MDPs, which is a class of sequential decision-making models widely used in practice.
\begin{definition}[Standard MDP \citep{puterman1994markov}]\label{def:_3_3_mdp}
\sloppy	A Markov decision process is tuple $\Tuple{\D(S), \D(X), \1T, \1R}$ where 
	\begin{enumerate}
		\item $\D(S)$ is a set of states called the state space;
		\item $\D(X)$ is a set of actions called the action space;
		\item $\1T(s, x, s') \in [0, 1]$ is a transition probability that action $X_i = x$ in state $S_i = s$ at stage $i$ will lead to state $S_{i+1} = s'$ at stage $i+1$;
		\item $\1R(s, x)$ is the immediate reward received in state $S_i = s$ due to action $X_i = x$ at stage $i$.
	\end{enumerate}
\end{definition}
A policy $\pi(x|s)$ in a standard MDP is a function mapping from the state space $\1D(S)$ to a probability distribution over the action space $\1D(X)$. For any indices $i < j \in \3N$, let $\bar{\*V}_{i:j}$ denote a sequence $\{V_i, V_{i+1}, \dots, V_j\}$. 
Given a policy $\pi$ and a distribution over the initial state $P(S_1)$, every standard MDP model defines a joint distribution over states $\bar{\*S}_{1:H}$, actions $\bar{\*X}_{1:H}$, and rewards $\bar{\*Y}_{1:H}$ up to decision horizon $H$, i.e.,
\begin{align}
	P_{\pi}(\bar{\*s}_{1:H}, \bar{\*x}_{1:H}, \bar{\*y}_{1:H}) = P(s_1) \prod_{i = 1}^H \pi(x_i\mid s_i) \1T(s_i, x_i, s_{i+1}) \I\{\1R(s_i, x_i) = y_i\}
\end{align}
The key assumption of a standard MDP is that the transition probability and reward functions depend on the past only through the current state of the system and the action selected by the decision maker in that state. This assumption is called the \emph{Markov property} \citep{puterman1994markov} and can be characterized using the following independence relationships, for every stage $i = 2, 3, \dots$,
\begin{align}
	\Parens{\bar{\*S}_{1:i-1}, \bar{\*X}_{1:i-1}, \bar{\*Y}_{1:i-1} \independent \bar{\*S}_{i+1:\infty}, \bar{\*X}_{i:\infty}, \bar{\*Y}_{i:\infty} \mid S_i} \label{eq:_3_3_markov}
\end{align}
In other words, the standard MDP could be seen as a compact representation of a family of joint distributions over observed trajectories of states $\bar{\*S}_{1:H}$, actions $\bar{\*X}_{1:H}$, and rewards $\bar{\*Y}_{1:H}$, provided that the Markov property holds. 
In the language of structural causality, the Markov property could hold in both the observational distribution $P(\bar{\*s}_{1:H}, \bar{\*x}_{1:H}, \bar{\*y}_{1:H})$ and the interventional distribution $P_{\pi}(\bar{\*s}_{1:H}, \bar{\*x}_{1:H}, \bar{\*y}_{1:H})$. 
\footnote{In this case, the decision rule $\pi(x|s)$ is set as the conditional distribution $P(X_i = x \mid S_i = s)$ defined by the behavioral policy $f_X$.} We showed in Examples \ref{exp:_2_2_mdp} and \ref{exp:_2_2_mdp4} the compression of the observational and interventional distributions of an SCM instance to standard MDP models respectively.


To make the argument more precise, consider the SCM $\1M^*$ graphically described in the MDP diagram $\1G_{\textsc{mdp}}$ of Fig.~\ref{fig:_3_1_mdp}. For every state $i = 1, 2, \dots$, conditioning on state $S_i$ and action $X_i$ blocks all paths from history $S_j, X_j, Y_j$ for $j < i$ to any future state $S_{k}$, action $X_{k}$, and reward $Y_k$ for $k > i$ (Def.~\ref{def:_2_3_dsep}). 
The observational distribution evaluated in $\1M^*$ thus satisfies the Markov property in Eq.~\ref{eq:_3_3_markov} and can be represented using a standard MDP. 
\footnote{We will consistently assume that structural functions $f_{S_i}, f_{Y_i}$ and distributions $P(\*U_{S_i}, \*U_{Y_i})$ remain invariant across decision horizons $i = 1, 2, \dots$. This is a common assumption for solving infinite-horizon MDP \citep{puterman1994markov}.}
\begin{example}[MDP, Observational]\label{exp:_3_3_mdp_obs}
Consider the MDP environment $\1M^*$ described in Eq.~\ref{eq:_2_1_mdp}. Its observational distribution $P(\bar{\*s}_{1:H}, \bar{\*x}_{1:H}, \bar{\*y}_{1:H})$ defines a standard MDP $\Tuple{\D(S), \D(X), \1T_{\text{obs}}, \1R_{\text{obs}} }$ where the transition probability and the reward function are observational quantities given by
\begin{align}
    \1T_{\text{obs}}(s, x, s') &= P\left(S_{i+1} = s' \mid S_i = s, X_i = x\right)\\
    \1R_{\text{obs}}(s, x) &= \E\left[Y_i \mid S_i = s, X_i = x \right]
\end{align}
Detailed parametrizations of system dynamics $\1T$ and $\1R$ can be compactly represented as a finite-state machine and are shown in Fig.~\ref{fig:_2_2_mdp_obs}. 

	\begin{table}[t]
			\centering
			\renewcommand{\arraystretch}{1.25}
			\begin{tabular}{|cc|c|cc |c|}
				$S$ & $X$ & $Q_*$     & $S$ & $X$ & $Q_*$     \\
				\hline
				\hline
				0     & 0     & 9 & 1     & 0     &\textbf{9} \\
				0     & 1     & \textbf{9}       & 1     & 1     & 9         \\
			\end{tabular}
		\caption{Optimal Q-function $Q_*(s, x)$ evaluated in the MDP model of Example~\ref{exp:_3_3_mdp_obs}.}
		\label{tab:_3_3_mdp_obs}
	\end{table}
Following the Bellman equation in Eq.~\ref{eq:_3_2_mdp_optimal_bellman}, we solve for the optimal Q-function $Q_*(s, x)$ in the standard MDP $\Tuple{\D(S), \D(X), \1T_{\text{obs}}, \1R_{\text{obs}} }$ and provide it in Table~\ref{tab:_3_3_mdp_obs}. Maximizing the action $x$ for every state $s$ in $Q_*(s, x)$ gives an optimal decision rule $\pi^*_{\text{obs}} \triangleq X_i \gets \neg S_i$. \hfill $\blacksquare$
\end{example}
Consider a policy space $\Pi = \Braces{\Tuple{X_i, \{S_i\}}}_{i = 1}^{\infty}$. Following a similar argument, we could show that the interventional distribution induced by any policy $\pi \in \Pi$ evaluated in SCM $\1M^*$ satisfies the Markov property, leading to an alternative standard MDP representation. 
\begin{example}[MDP, Interventional]\label{exp:_3_3_mdp_inv}
Consider the MDP environment $\1M^*$ described in Eq.~\ref{eq:_2_1_mdp}. For every policy $\pi = \Parens{\pi_i(X_i \mid S_i)}_{i = 1}^{\infty}$, its interventional distribution $P_{\pi}(\bar{\*s}_{1:H}, \bar{\*x}_{1:H}, \bar{\*y}_{1:H})$ satisfies the Markov property in Eq.~\ref{eq:_3_3_markov}. It defines a standard MDP $\Tuple{\D(S), \D(X), \1T_{\text{exp}}, \1R_{\text{exp}} }$ where the transition probability and the reward function are interventional quantities given by Eqs.~\ref{eq:_3_2_mdp_t} and \ref{eq:_3_2_mdp_r}. Their parametrizations are described in the finite-state machine of Fig.~\ref{fig:_2_2_mdp_inv}. 

We also solve for the optimal Q-function $Q_*(s, x)$ in the MDP $\Tuple{\D(S), \D(X), \1T_{\text{exp}}, \1R_{\text{exp}} }$ and obtain an optimal decision rule $\pi^*_{\text{exp}} \triangleq X_i \gets S_i$. Revisit Example~\ref{exp:_3_1_mdp} for detailed computations. \hfill $\blacksquare$
\end{example}
Some important observations follow from these two examples. First, the Markov property holds in both the observational and interventional distributions evaluated in the SCM $\1M^*$ described in Eq.~\ref{eq:_2_1_mdp}, resulting in two standard MDPs. 
Second, solving these MDPs leads to different policies $\pi^*_{\text{obs}}$ and $\pi^*_{\text{exp}}$. 
The previous discussion in Example~\ref{exp:_3_1_mdp} showed that only $\pi^*_{\text{exp}}$ is the optimal policy in the underlying environment $\1M^*$; while $\pi^*_{\text{obs}}$ is sub-optimal. 
This suggests that the model assumptions of standard MDPs are generally insufficient in determining the optimal policy in the underlying causal model, however many samples are provided.

Consider now a CRL agent that interacts with the environment $\1M^*$ and receives observed data $\1D$. Without specifying the learning regime $\1L$ (see or do) or causal knowledge $\1A$, the agent cannot determine whether data $\1D$ is drawn from the observational or interventional distribution from the Markov property. If data $\1D$ is collected from passive observations, optimizing the learned standard MDP model could lead to a sub-optimal policy, resulting in unsatisfactory performance. One may wonder if it is possible to recover interventional quantities $\1T_{\text{exp}}$ and $\1R_{\text{exp}}$ form the observational data $\1D \sim P(\*V)$ in MDP environments. Unfortunately, our next result suggests otherwise.

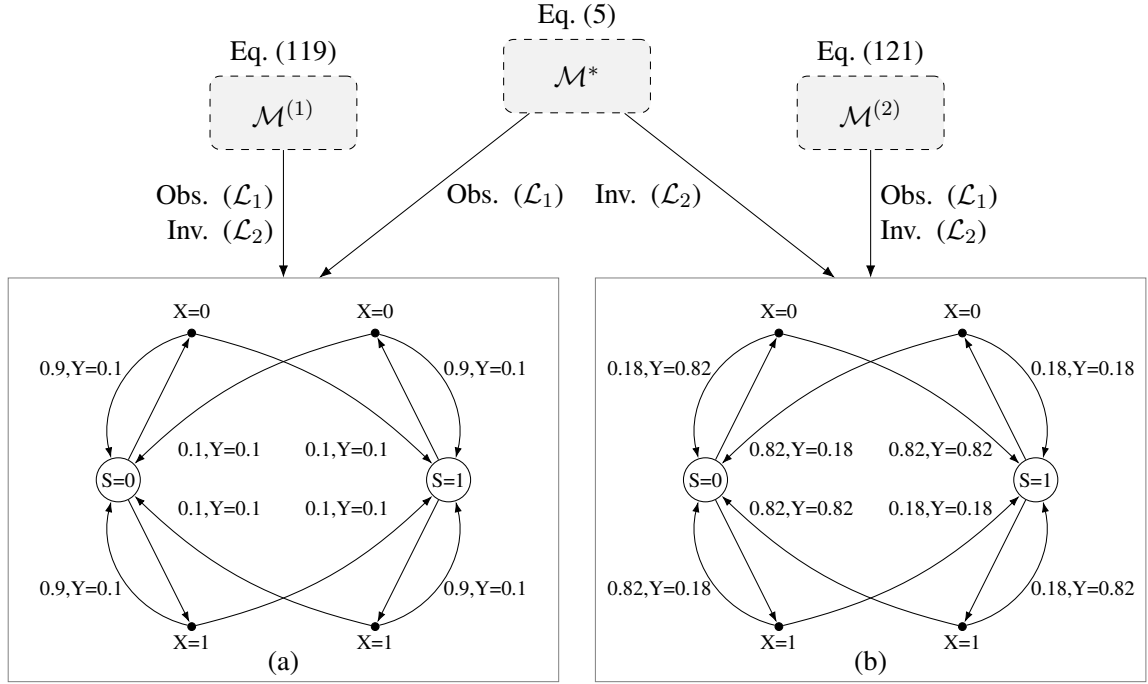
\begin{figure}[t]
\centering
  \resizebox{\linewidth}{!}{
  \begin{tikzpicture}
      \node (env) [environment, fill=gray!10, dashed, minimum height=1cm, text width=1cm, label=above:{Eq.~(5)}] at (6.25, 5.5) {$\M^*$};
      \node (env1) [environment, fill=gray!10, dashed, minimum height=1cm, text width=1cm, label=above:{Eq.~(119)}] at (2.25, 5) {$\M^{(1)}$};
      \node (env2) [environment, fill=gray!10, dashed, minimum height=1cm, text width=1cm, label=above:{Eq.~(121)}] at (10.25, 5) {$\M^{(2)}$};
        
      \node[vertex, minimum width=6mm] (S0) at (0, 0) {S=0};
      \node[vertex, minimum width=6mm] (S1) at (4.5, 0) {S=1};
      \node[action, label={[shift={(0,0)}]\scriptsize X=0}] (X00) at (1, 2) {};
      \node[action, label={[shift={(0,-0.5)}]\scriptsize X=1}] (X10) at (1, -2) {};
      \node[action, label={[shift={(0,0)}]\scriptsize X=0}] (X01) at (3.5, 2) {};
      \node[action, label={[shift={(0,-0.5)}]\scriptsize X=1}] (X11) at (3.5, -2) {};
      
      \draw[dir] (S0) to (X00);
      \draw[dir] (S0) to (X10);
      \draw[dir] (S1) to (X01);
      \draw[dir] (S1) to (X11);
      
      \node[text width = 1.3cm, align = right, left] at (0.2, 1.5) {\scriptsize 0.9,Y=0.1};
      \node[text width = 1.3cm, align = right, left] at (0.2, -1.5) {\scriptsize 0.9,Y=0.1};
      
      \node[text width = 1.3cm, align = right] at (1.3, 0.4) {\scriptsize 0.1,Y=0.1};
      \node[text width = 1.3cm, align = right] at (1.3, -0.4) {\scriptsize 0.1,Y=0.1};
      
      \node[text width = 1.3cm, align = left] at (3.2, 0.4) {\scriptsize 0.1,Y=0.1};
      \node[text width = 1.3cm, align = left] at (3.2, -0.4) {\scriptsize 0.1,Y=0.1};
      
      \node[text width = 1.3cm, right] at (4.3, 1.5) {\scriptsize 0.9,Y=0.1};
      \node[text width = 1.3cm, right] at (4.3, -1.5) {\scriptsize 0.9,Y=0.1};

      \draw[dir] (X00) to [bend right = 45] (S0);
      \draw[dir] (X10) to [bend left = 45] (S0);
      \draw[dir] (X00) to [bend left = 15] (S1);
      \draw[dir] (X10) to [bend right = 15] (S1);
      
      \draw[dir] (X01) to [bend left = 45] (S1);
      \draw[dir] (X11) to [bend right = 45] (S1);
      \draw[dir] (X01) to [bend right = 15] (S0);
      \draw[dir] (X11) to [bend left = 15] (S0);
      
      \node (mdpa) at (2.25, 0) [draw, gray, ultra thin, minimum width=7.5cm,minimum height=5.5cm] {};
      
      \node[vertex, minimum width=6mm] (S0i) at (8, 0) {S=0};
      \node[vertex, minimum width=6mm] (S1i) at (12.5, 0) {S=1};
      \node[action, label={[shift={(0,0)}]\scriptsize X=0}] (X00i) at (9, 2) {};
      \node[action, label={[shift={(0,-0.5)}]\scriptsize X=1}] (X10i) at (9, -2) {};
      \node[action, label={[shift={(0,0)}]\scriptsize X=0}] (X01i) at (11.5, 2) {};
      \node[action, label={[shift={(0,-0.5)}]\scriptsize X=1}] (X11i) at (11.5, -2) {};
      
      \draw[dir] (S0i) to (X00i);
      \draw[dir] (S0i) to (X10i);
      \draw[dir] (S1i) to (X01i);
      \draw[dir] (S1i) to (X11i);
      
      \node[text width = 1.4cm, align = right, left] at (8.2, 1.5) {\scriptsize 0.18,Y=0.82};
      \node[text width = 1.4cm, align = right, left] at (8.2, -1.5) {\scriptsize 0.82,Y=0.18};
      
      \node[text width = 1.4cm, align = right] at (9.3, 0.4) {\scriptsize 0.82,Y=0.18};
      \node[text width = 1.4cm, align = right] at (9.3, -0.4) {\scriptsize 0.82,Y=0.82};
      
      \node[text width = 1.4cm, align = left] at (11.2, 0.4) {\scriptsize 0.82,Y=0.82};
      \node[text width = 1.4cm, align = left] at (11.2, -0.4) {\scriptsize 0.18,Y=0.18};
      
      \node[text width = 1.4cm, right] at (12.3, 1.5) {\scriptsize 0.18,Y=0.18};
      \node[text width = 1.4cm, right] at (12.3, -1.5) {\scriptsize 0.18,Y=0.82};
      
      \draw[dir] (X00i) to [bend right = 45] (S0i);
      \draw[dir] (X10i) to [bend left = 45] (S0i);
      \draw[dir] (X00i) to [bend left = 15] (S1i);
      \draw[dir] (X10i) to [bend right = 15] (S1i);
      
      \draw[dir] (X01i) to [bend left = 45] (S1i);
      \draw[dir] (X11i) to [bend right = 45] (S1i);
      \draw[dir] (X01i) to [bend right = 15] (S0i);
      \draw[dir] (X11i) to [bend left = 15] (S0i);
      
       \node (mdpb) at (10.25, 0) [draw, gray, ultra thin, minimum width=7.5cm,minimum height=5.5cm] {};
       
      \node (a) at (2.25, -2.5) {(a)};
      \node (b) at (10.25, -2.5) {(b)};
      
      \path [-Latex] (env) edge node[anchor=west, text width=1.9cm] {\hphantom{a}Obs. ($\1L_1$)} (mdpa.80);
      \path [-Latex] (env) edge node[anchor=east, text width=1.7cm] {Inv. ($\1L_2$)} (mdpb.100);
      
      \path [-Latex] (env1) edge node[anchor=east, text width=1.7cm, align = right] {Obs. ($\1L_1$)\\ Inv. ($\1L_2$)} (mdpa.north);
      \path [-Latex] (env2) edge node[anchor=west, text width=1.7cm] {Obs. ($\1L_1$)\\ Inv. ($\1L_2$)} (mdpb.north);
      
  \end{tikzpicture}
  }
\caption{Causal Hierarchy Theorem (CHT) in MDP environments.}
\label{fig:_2_2_cht_mdp}
\end{figure}%

\begin{proposition}\label{prop:_3_3_mdp_obs}
	For any SCM $\1M^*$ compatible with the causal diagram $\G_{\textsc{mdp}}$ of Fig.~\ref{fig:_3_1_mdp}, there is an SCM $\1M^{(1)}$ compatible with $\G_{\textsc{mdp}}$ such that for every stage $i = 1, 2, \dots$,
	\begin{align}
		&P^{(1)}\left ( s_{i+1} \mid s_i, x_i \right ) = P^{*}\left ( s_{i+1} \mid s_i, x_i\right ), &&\E^{(1)}\left[ Y_i \mid s_i, x_i\right] = \E^{*}\left[ Y_i \mid s_i, x_i\right] \label{eq:_3_3_mdp_obs1}
	\end{align}
	while 
	\begin{align}
		&P^{(1)}_{x_i}\left ( s_{i+1} \mid s_i \right ) \neq P^{*}_{x_i}\left ( s_{i+1} \mid s_i \right ), &&\E^{(1)}_{x_i}\left[ Y_i \mid s_i\right] \neq \E^{*}_{x_i}\left[ Y_i \mid s_i\right] \label{eq:_3_3_mdp_obs2}
	\end{align} \hfill $\blacksquare$
\end{proposition}
The following example constructs an alternative SCM $\1M^{(1)}$ that generates the observational distribution as the underlying environment $\1M^*$, but differs significantly in interventional distributions. 
\begin{example}[MDP, Observational $\not \Rightarrow$ Interventional]\label{exp:_2_2_obs-mdp}
Consider the following MDP environment
\begin{align}
\M^{(1)} =
 \Tuple{\*U = \{U_{i, 1}, U_{i,2}, U_{i, 3}\}, \*V =  \{X_i, Y_i, S_i\}, \2F = \left \{\2F^{(1)}_i \right \}, P^{(1)}(\*U)}_{i = 1, 2, \dots}, 
\end{align}
where the causal mechanisms $\2F^{(1)}_t$ are defined as
 \begin{align}
    \2F^{(1)}_t = \begin{cases}
      S_i \gets S_{i-1} \oplus U_{i-1, 2},\\
      X_i \gets S_i \oplus U_{i, 1}, \\
      Y_i \gets U_{i, 3}, 
    \end{cases} \label{eq:_2_2_obs_mdp}
  \end{align}	 
and $P^{(1)}(U_{i, 1}, U_{i,2}, U_{i, 3})$ is such that $U_{i, 1}, U_{i, 2}, U_{i, 3}$ are independent variables drawn from distribution $P(U_{i, 1} = 1) = 0.9$, and $P(U_{i, 2} = 1) = P(U_{i, 3} = 1) = 0.1$. 

We compute the observational distributions $P(S_{i+1} \mid S_i, X_i)$ and $\E[Y_i \mid S_i, X_i]$ evaluated in $\1M^{(1)}$ and show their parametrization in the finite-state machine in Fig.~\ref{fig:_2_2_cht_mdp}(a). It is verifiable that MDP models $\M^{(1)}$ and $\M^*$ (defined in Eq.~\ref{eq:_2_1_mdp}) generate the same observational distributions ($\1L_1$), i.e., the equalities in Eq.~\ref{eq:_3_3_mdp_obs1} hold. On the other hand, we also derive the interventional distribution $\inv{S_{i+1} \mid S_i}{X_i}$, $\invE{Y\mid S_i}{X_i}$ evaluated in $\1M^{(1)}$. Their parametrization could also be summarized using Fig.~\ref{fig:_2_2_cht_mdp}(a). It follows from previous discussions (Examples~\ref{exp:_2_2_mdp} and \ref{exp:_2_2_mdp4}) that  $\1M^{(1)}$ and $\M^*$ (Eq.~\ref{eq:_2_1_mdp}) differ in the interventional distribution ($\1L_2$), i.e., inequalities in Eq.~\ref{eq:_3_3_mdp_obs2} hold. \hfill $\blacksquare$
\end{example} 
The above example shows that the optimal policy in an unknown MDP environment is generally underdetermined by the observational distribution and the Markov property. Conversely, we also show that one cannot recover observational quantities from the interventional data in MDP environments.
\begin{proposition}\label{prop:_3_3_mdp_inv}
	For any SCM $\1M^*$ compatible with the causal diagram $\G_{\textsc{mdp}}$ of Fig.~\ref{fig:_3_1_mdp}, there is an SCM $\1M^{(2)}$ compatible with $\G_{\textsc{mdp}}$ such that for every stage $i = 1, 2, \dots$,
	\begin{align}
		&P^{(2)}_{x_i}\left ( s_{i+1} \mid s_i \right ) = P^{*}_{x_i}\left ( s_{i+1} \mid s_i \right ), &&\E^{(2)}_{x_i}\left[ Y_i \mid s_i\right] = \E^{*}_{x_i}\left[ Y_i \mid s_i\right] \label{eq:_3_3_mdp_inv1}
	\end{align}
	while 
	\begin{align}
		&P^{(2)}\left ( s_{i+1} \mid s_i, x_i \right ) \neq P^{*}\left ( s_{i+1} \mid s_i, x_i\right ), &&\E^{(2)}\left[ Y_i \mid s_i, x_i\right] \neq \E^{*}\left[ Y_i \mid s_i, x_i\right] \label{eq:_3_3_mdp_inv2}
	\end{align} \hfill $\blacksquare$
\end{proposition}
Our next example corroborates the aforementioned proposition by constructing an alternative SCM $\1M^{(2)}$ that generates the same interventional distribution as the underlying environment $\1M^*$ but appears different from passive observations. 
\begin{example}[MDP, Interventional $\not \Rightarrow$ Observational]\label{exp:_2_2_inv-mdp}
Consider the following MDP environment
\begin{align}
\M^{(2)} =
 \Tuple{\*U = \{U_{i, 1}, U_{i,2}, U_{i, 3}\}, \*V =  \{X_i, Y_i, S_i\}, \2F = \left \{\2F^{(2)}_i \right \}, P^{(2)}(\*U) }_{i = 1, 2, \dots}, 
\end{align}
where the causal mechanisms $\2F^{(2)}_t$ are defined as:
 \begin{align}
    \2F^{(2)}_t =  \begin{cases}
      S_i \gets \left (S_{i-1} \vee X_{t-1} \right) \oplus U_{i-1,2},\\
      X_i \gets S_i \oplus U_{i, 1},\\
      Y_i \gets S_i \oplus X_i \oplus U_{i, 3}, 
    \end{cases}\label{eq:_2_2_inv-mdp}
  \end{align}	 
 and $P^{(2)}(U_{i, 1}, U_{i,2}, U_{i, 3})$ is such that $U_{i, 1}, U_{i, 2}, U_{i, 3}$ are independent variables drawn from distribution $P(U_{i, 1} = 1) = 0.9$, and $P(U_{i, 2} = 1) = 0.82 = P(U_{i, 3} = 1) = 0.82$. 
 
 We compute the interventional distribution $\inv{S_{i+1} \mid S_i}{X_i}$ and $\invE{Y\mid S_i}{X_i}$ evaluated in $\1M^{(2)}$ and summarize them in the finite-state machine described in Fig.~\ref{fig:_2_2_cht_mdp}(b). 
 It is verifiable that MDP models $\M^{(2)}$ and $\M^*$ (Eq.~\ref{eq:_2_1_mdp}) define the same interventional distributions ($\1L_2$), i.e., the equalities in Eq.~\ref{eq:_3_3_mdp_inv1} hold. 
 We also compute the observational distributions $P(S_{i+1} \mid S_i, X_i)$ and $\E[Y\mid S_i, X_i]$ of $\1M^{(2)}$ and provide their parametrizations in Fig.~\ref{fig:_2_2_cht_mdp}(b). 
 The previous discussions (Examples~\ref{exp:_2_2_mdp} and \ref{exp:_2_2_mdp4}) implied that  $\1M^{(2)}$ and $\M^*$ (Eq.~\ref{eq:_2_1_mdp}) differ significantly in the observational distribution ($\1L_1$), i.e., inequalities in Eq.~\ref{eq:_3_3_mdp_inv2} hold. 
 This complements previous examples and illustrates that interventional queries are generally under-determined by observational data in MDP environments. \hfill $\blacksquare$
\end{example}
The last two examples are summarized and the results are illustrated in Fig.~\ref{fig:_2_2_cht_mdp}. In the middle of the figure, we show the true MDP model $\M^*$ initially discussed in Example \ref{exp:_2_1_mdp} and its induced observational and interventional distributions (described in more detail in Figs.~\ref{fig:_2_2_obs_mdp} and \ref{fig:_2_2_inv_mdp} respectively). Assuming only observational data ($\1L_1$) is available, one can construct an alternative SCM $\M^{(1)}$ (left side) that matches the observational distribution but have different interventional behavior (i.e., $\1L^{*}_1 = \1L^{(1)}_1$, $\1L^{*}_2 \neq \1L^{(1)}_2$). 
Formally, the interventional distribution is underdetermined by the observational distribution. 
Practically, this means that passively observing another agent acting in the environment and collecting samples from it may not be enough to make claims about the agent's policies and their corresponding performance. 

On the other hand, the same is the case in the reverse direction;
say, whenever interventional data ($\1L_2$) is available, one can then construct an alternative SCM $\M^{(2)}$ (right side) that matches the interventional distribution but has a different observational one (i.e., $\1L^{*}_1 \neq \1L^{(2)}_1$, $\1L^{*}_2 = \1L^{(2)}_2$). Formally, the observational distribution is underdetermined by the interventional distribution. 
This may be counter-intuitive since interventions are usually believed to be more informative than just passively observing the system unfold in time.  Still, in practice, it doesn't allow the CRL agent to predict how other agents will behave when interacting in the environment. This impossibility will translate into challenges when considering the communication and exchange of experience across agents with the intent of accelerating learning. 

More generally, the Causal Hierarchy Theorem \citep[Thm.~1]{bareinboim2020pearl} states that this impossibility result is strict for almost all causal models. This means it is generically impossible to draw higher-layer inferences using only lower-layer information. 
Given that the actual underlying SCM is rarely observable in practice, and no inferences across the layers of the PCH are possible, the CRL agent will need to resort to some causal knowledge and assumptions to make claims about these underlying mechanisms, as discussed in Sec.~\ref{sec:_2_3_diagram}.  
Since the more typical language to describe standard MDPs is constrained to one particular distribution, it's somewhat limiting to consider it as a baseline to the model of the environment/agent relationship given the more general types of tasks that can be represented in terms of the PCH, as discussed earlier in this manuscript. 
\section{Reinforcement Learning through Causal Lenses}\label{sec:_4}
The formalization of causal reinforcement learning tasks (Def.~\ref{def:_3_2_task}) allows us to describe some of the most common and popular learning settings studied in the classic literature of reinforcement learning (RL) and causal inference (CI). This section will investigate learning methods for these classic RL-CI tasks, including off-policy learning (Sec.~\ref{sec:_4_1}), online learning (Sec.~\ref{sec:_4_2}), and causal identification from observational data (Sec.~\ref{sec:_4_3}). These tasks are briefly described in Table~\ref{tab:_2_4_roadmap} and can be seen as variations of the first two dimensions described in the table, i.e., interaction regime and structural assumptions, while the other dimensions are fixed. Lastly, Sec.~\ref{sec:_4_4} varies these dimensions and moves toward a catalog of novel CRL tasks.

Even though the tasks of off-policy learning, online learning, and causal identification all come from the classic literature, there are still subtle interplays between reinforcement learning and causal invariances that were only formally understood with the CRL formalization. For instance, in the language of structural causality, we will provide a formal justification for off-policy learning algorithms, e.g., inverse propensity weighting and dynamic programming. This permits one to determine when and how to apply RL algorithms to more generalized settings where unobserved confounders exist in the observational dataset. Analyzing these classic tasks through CRL lenses will shed light on other foundational issues and how CI and RL connect.

We will introduce some additional notations before studying these learning tasks in detail. Specifically, we will optimize over a policy space with a finite decision horizon $H  = |\*X| < \infty$.  Let actions $\*X$ be ordered by $X_1, X_2, \dots, X_H$, $H = |\*X|$, following a topological order in the underlying SCM $\1M^*$. For any indices $i < j$, let $\bar{\*X}_{i:j} = \{X_{i}, X_{i+1}, \dots, X_j\}$ denote a sequence of actions ranging from stage $i$ to stage $j$. Similarly, let $\bar{\*S}_{i:j} = \{\*S_{i}, \*S_{i+1}, \dots, \*S_j\}$ denote the sequence of states from stage $i$ to stage $j$. For convenience, we will consistently write $\bar{\*X}_{i} = \bar{\*X}_{1:i}$ and $\bar{\*S}_{i} = \bar{\*S}_{1:i}$. Fix a policy $\pi \in \Pi$. For any indices $i \leq j$, let $\left( \pi_i, \dots, \pi_j\right)$ denote a subsequence of decision rules constrained in $\pi$ determining values of actions $X_i, \dots, X_j$.

\subsection{Off-Policy Learning} \label{sec:_4_1}
This section investigates the off-policy learning problem where an agent attempts to learn an optimal policy from observational data generated by a different behavior policy \citep{sutton1998reinforcement}, provided that there is no unmeasured confounder (NUC) in the data (to be defined). First, we will introduce the NUC assumption and discuss how it justifies the off-policy learning approach. We then describe in Sec.~\ref{sec:_4_1_1} two primary methods in evaluating candidate policies in the off-policy setting, including inverse propensity weighting and dynamic programming \citep{bellman57}.

\begin{figure}[t]
	\centering
	\begin{tikzpicture}
		\def\outerr{3.5}
		\def\innerr{3.5}
		\def\dist{3.5}

		\draw[->, >={Latex}] (-1.5,0) -- (3*\dist+2,0) node[below] {Episode t};

		\node[vertex, minimum width=6mm] (X1) at (-1, 2) {X\textsuperscript{(1)}};
		\node[vertex, minimum width=6mm] (Y1) at (1, 2) {Y\textsuperscript{(1)}};
		\draw[bidir] (X1) to [bend left = 45] (Y1);
		\draw[dir] (X1) -- (Y1);

		\node[vertex, minimum width=6mm] (X2) at (\dist-1, 2) {X\textsuperscript{(2)}};
		\node[vertex, minimum width=6mm] (Y2) at (\dist+1, 2) {Y\textsuperscript{(2)}};
		\draw[bidir] (X2) to [bend left = 45] (Y2);
		\draw[dir] (X2) -- (Y2);

		\node[vertex, minimum width=6mm] (X3) at (2*\dist-1, 2) {X\textsuperscript{(3)}};
		\node[vertex, minimum width=6mm] (Y3) at (2*\dist+1, 2) {Y\textsuperscript{(3)}};
		\draw[bidir] (X3) to [bend left = 45] (Y3);
		\draw[dir] (X3) -- (Y3);

		\node[vertex, minimum width=6mm] (X4) at (3*\dist-1, 2) {X\textsuperscript{(4)}};
		\node[vertex, minimum width=6mm] (Y4) at (3*\dist+1, 2) {Y\textsuperscript{(4)}};
		\draw[bidir] (X4) to [bend left = 45] (Y4);
		\draw[dir] (X4) -- (Y4);

		\node (d1) at (0, 1.2) {$\*V^{(1)} \sim P\left(\*V\right)$};
		\node (d2) at (\dist, 1.2) {$\*V^{(2)} \sim P\left(\*V\right)$};
		\node (d3) at (2*\dist, 1.2) {$\*V^{(3)} \sim P\left(\*V\right)$};
		\node (d4) at (3*\dist, 1.2) {$\*V^{(4)} \sim P\left(\*V\right)$};

		\draw[very thick, betterblue, -] (0, 0) -- (0, 0.2);
		\draw[very thick, betterblue, -] (\dist, 0) -- (\dist, 0.2);
		\draw[very thick, betterblue, -] (2*\dist, 0) -- (2*\dist, 0.2);
		\draw[very thick, betterblue, -] (3*\dist, 0) -- (3*\dist, 0.2);

		\node [below] at (0, -0.05) { 0};
		\node [below] at (\dist, -0.05) { 1};
		\node [below] at (2*\dist, -0.05) { 2};
		\node [below] at (3*\dist, -0.05) { 3};

		\node [fill=betterblue!45] at (0, 0.5) {see};
		\node [fill=betterblue!45] at (\dist, 0.5) {see};
		\node [fill=betterblue!45] at (2*\dist, 0.5) {see};
		\node [fill=betterblue!45] at (3*\dist, 0.5) {see};

		\begin{pgfonlayer}{back}
			\node[circle,fill=betterblue!65,draw=none,minimum size=2*\innerr mm] at (X1) {};
			\node[circle,fill=betterred!65,draw=none,minimum size=2*\innerr mm] at (Y1) {};
			\node[circle,fill=betterblue!65,draw=none,minimum size=2*\innerr mm] at (X2) {};
			\node[circle,fill=betterred!65,draw=none,minimum size=2*\innerr mm] at (Y2) {};
			\node[circle,fill=betterblue!65,draw=none,minimum size=2*\innerr mm] at (X3) {};
			\node[circle,fill=betterred!65,draw=none,minimum size=2*\innerr mm] at (Y3) {};
			\node[circle,fill=betterblue!65,draw=none,minimum size=2*\innerr mm] at (X4) {};
			\node[circle,fill=betterred!65,draw=none,minimum size=2*\innerr mm] at (Y4) {};
		\end{pgfonlayer}
	\end{tikzpicture}
	\caption{Temporal diagram showing an off-policy learning agent interacting with the environment for repeated episodes.}
	\label{fig:_4_1_offpolicy}
\end{figure}
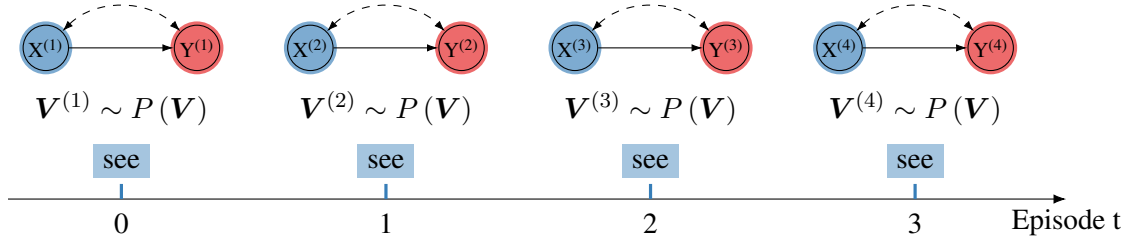

An off-policy learning agent interacts with the underlying environment (SCM) through passively observing events unfolding over time. Fig.~\ref{fig:_4_1_offpolicy} is a graphical representation of the CRL agent interacting with the environment for repeated episodes $t = 1, \dots, T$. For every episode $t$, the agent ``sees'' the SCM $\1M^*$ (Def.~\ref{def:_2_2_obs_dist}) and receives an observation $\*V^{(t)} \sim P\left( \*V \right)$.  The CRL agent currently aims to use these observations to learn a policy from candidate space $\Pi$ that maximizes a reward function $\1R(\*Y)$. This will be a useful representation to compare other learning modalities and types of interactions. The following signature characterizes this learning setting:
\begin{align}
	\1T_{\text{off}} = \left \langle \1I= \text{see}, \1A=\text{NUC}, \Pi = \{\langle X_i, \*S_i \rangle \}_{i = 1}^H, \1R = \D(\*Y) \mapsto \3R \right \rangle.
\end{align}
where the agent uses observational data combined with the critical assumption $\1A = \text{NUC}$, which means ``\emph{No Unmeasured Confounder}''. The agent's goal is to obtain an optimal policy estimate from the combination of the observational data and the NUC assumption, i.e.,
\begin{align}\label{eq:opt-crl-off}
	 & \pi^* = \argmax_{\pi \in \Pi} \invEE{ \1R \left (\*Y \right ) \big \vert \textcolor{red}{\;\1A= \text{NUC}, \;\mathcal{D}_{\text{obs}} \sim P(\*V)}}{\pi}{\1M^*}. 
\end{align}
In practice, the NUC assumption will require that at every stage of intervention on action $X_i \in \*X$, its observed correlations with the reward $Y$ given past actions and covariates' history, is entirely determined by the causal relationships between $X_i$ and $Y$. In other words, no other variables generate non-causal variations between the decision $X_i$ and the outcome $Y$. The following definition formalizes this idea.

\begin{definition}[No Unmeasured Confounder]\label{def:_4_1_nuc}
	Let $\1M^*$ be a SCM and $\Pi = \left \{\langle X_i, \*S_i \rangle \right \}_{i = 1}^H$ be a policy space (Def.~\ref{def:_3_space}). The ``no unmeasured confounder'' (for short, NUC) condition holds if for every action $X_i \in \*X$, its endogenous parents $\*\PA_{i}$ and exogenous parents $\*U_i$ satisfy the following conditions:
	\begin{enumerate}
		\item Endogenous parents $\*\PA_{i}$ are contained in the history $\bar{\*X}_{i-1} \cup \bar{\*S}_i$, i.e., $\*\PA_{i} \subseteq \bar{\*X}_{i-1} \cup \bar{\*S}_i$;
		\item Exogenous parents $\*U_{i}$ are independent from exogenous noises $\*U_j$ associated with all the other endogenous variables $V_j$ in the system, i.e., $\*U_i \ci \Braces{\*U_j \mid \forall V_j \in \*V \setminus \{X_i\}}$.
	\end{enumerate}
\end{definition}
In the above definition, Condition 1 says that all endogenous parents of every action $X_i$ are observed, contained in the past states and actions $\bar{\*S}_{i}, \bar{\*X}_{i-1}$; Condition 2 says that given the past history $\bar{\*X}_{i-1} \cup \bar{\*S}_{i}$, values of every action $X_i$ are decided by an independent noise $\*U_i$. Note that in the underlying SCM $\1M^*$, observational data are generated by a behavior policy $f_{\*X}$ which determines values of every action $X_i$ based on the endogenous $\*\PA_{i}$ and exogenous parents $\*U_i$ for all time steps $i = 1, \dots, H$. The NUC condition implies that one could simulate the behavior policy using a sequence of decision rules $\Parens{\pi_i(X_i \mid \bar{\*X}_{i-1}, \bar{\*S}_i)}_{i = 1}^{H}$ such that for every step $i = 1, \dots, H$, $\pi_i(X_i \mid \bar{\*X}_{i-1}, \bar{\*S}_i) = P(X_i \mid \*\PA_{i})$.   Allocating actions following these decision rules leads to the following independence relationships, for any sequence of actions $\bar{\*x}_H$,\footnote{The NUC assumption could also be characterized with a series of graphical conditions based on structural causality, known as \emph{sequential backdoor condition}. We will further elaborate on the graphical implication of NUC in Sec.~\ref{sec:_4_3}.}
\begin{align}
 	\left (X_i \ci S_{i+1_{\bar{\*x}_{i}}}, \dots,  S_{H_{\bar{\*x}_{H-1}}}, Y_{\bar{\*x}_H},  \mid \bar{\*X}_{i-1}, \bar{\*S}_i \right) \;\;\; \forall i = 1, \dots, H \label{eq:unc}
\end{align}
Among quantities in the above equation, the potential response $S_{i_{\bar{\*x}_{i-1}}}$, $i = 1, \dots, H$, is the observed state in submodel $\1M^*_{\bar{\*x}_{i-1}}$ induced by the atomic intervention $\doo(\bar{\*X}_{i-1} \gets \bar{\*x}_{i-1})$; similarly, $Y_{\bar{\*x}_H}$ is the future potential reward evaluated in submodel $\1M^*_{\bar{\*x}_H}$. The meaning of the NUC condition is illustrated in the next examples.
\begin{example}[DTR models where NUC holds]\label{exp:_4_1_nuc1}
	Consider a 2-stage DTR model $\Tuple{\1M^*, \Pi, Y}$ where SCM $\1M^*$ is described in Eq.~\ref{eq:_3_1_dtr1} and the policy space $\Pi = \Braces{\Tuple{X_1, \braces{S_1}}, \Tuple{X_2, \braces{S_1, X_1, S_2}}}$. We will next examine conditions of Def.~\ref{def:_4_1_nuc} and show that they hold in this CDM.

	First, note that for every treatment $X_i$, $i = 1, 2$, its endogenous parent $\*\PA_i = \{S_i\}$, which is contained in the corresponding input state $\*S_i$. This implies Condition (1) of NUC holds.

	Second, each structural function $f_{X_i}$, $i = 1, 2$ affecting treatment $X_i$, the coefficients $\alpha_i = 0$ of the exogenous variable $U$ is equal to zero. This means that there is no unobserved confounder affecting $X_i$ and other variables in the system, i.e., Condition (2) of NUC also holds. Therefore, we conclude the NUC condition holds in the DTR model $\Tuple{\1M^*, \Pi, Y}$. \hfill $\blacksquare$
\end{example}
\begin{example}[DTR models where NUC fails]\label{exp:_4_1_nuc2}
	Continuing with the DTR model $\Tuple{\1M^*, \Pi, Y}$  in the previous example, we now consider an alternative policy space $\Pi' = \Braces{\Tuple{X_1, \emptyset}, \Tuple{X_2, \emptyset }}$. Every policy $\pi \in \Pi'$ decides values of treatment $X_1, X_2$ independently, regardless of values of other variables in the system. For $i = 1, 2$, the history $\bar{\*X}_{i-1} \cup \bar{\*S}_i = \emptyset$ prior to stage $i$ is an empty set and does not contain the endogenous parent $S_i$ of treatment $X_i$. Therefore, Condition (1) of NUC fails.

	Alternatively, consider an SCM $\1M'$ where for the structural function $f_{X_i}$ of treatment $X_i$, $i = 1, 2$, the coefficient of the exogenous variable $U$ is equal to $\alpha_i = -3$. This means that there exists an unobserved confounder affecting treatments $X_1, X_2$ and the primary outcome $Y$ simultaneously. Consequently, Condition (2) of NUC does not hold. \hfill $\blacksquare$
\end{example}

\subsubsection{Off-Policy Evaluation}\label{sec:_4_1_1}
Whenever the NUC assumption holds, there exist different strategies that allow one to estimate and compare the effects of candidate policies from observational data without having to perform online experiments in the environment. The first algorithm we discuss that implements this idea is based on a technique known as ``\emph{inverse probability weighting}'' (IPW) and is widely applied in practice  \citep{rubin:74,robins2000marginal,murphy2001marginal,wang2012evaluation,swaminathan2015counterfactual,liu2018breaking}.
Formally,
\begin{restatable}[Inverse Propensity Weighting, under NUC]{theorem}{thmipw}\label{thm:_4_1_ipw}
	Let $\langle \1M^*, \Pi, \1R \rangle$ be a CDM where the policy space $\Pi = \left \{\langle X_i, \*S_i \rangle \right \}_{i = 1}^H$ and the reward function $\1R: \D(\*Y) \mapsto \3R$. If NUC holds, for any $\pi \in \Pi$, the expected reward is computable from the observational distribution $P\left (\*V \right)$ as
	\begin{align}
		\E_{\pi}\left [\1R(\*Y)\right] = \sum_{\bar{\*x}_H, \bar{\*s}_H} \underbrace{\E\left[\1R(\*Y) \mid \bar{\*x}_H, \bar{\*s}_H \right] P\left (\bar{\*x}_H, \bar{\*s}_H \right)}_{\text{observational distribution}} \underbrace{\prod_{i = 1}^H \frac{\pi_{i} \left(x_i \mid \*s_i\right)}{P\left( x_i \mid \bar{\*x}_{i-1}, \bar{\*s}_i \right)}}_{\text{ratio $\pi$ and obs. probabilities}}. \label{eq:_4_1_ipw}
	\end{align} 
\end{restatable}
Among the above quantities, $P\left( x_i \mid \bar{\*x}_{i-1}, \bar{\*s}_i \right)$ measures the natural propensity of the behavior policy for action $X_i$, $i = 1, \dots, H$, which is known as the \emph{propensity score}. The IPW estimation requires what is called the \emph{positivity} assumption, i.e., the propensity scores $P\left( x_i \mid \bar{\*x}_{i-1}, \bar{\*s}_i \right) > 0$ for every entry $\bar{\*x}_{i}, \bar{\*s}_i$.
\footnote{This quantitative assumption is called \emph{overlap} in \citep{rosenbaum1983central,imbens2004nonparametric}. There are attempts in the literature to relax it by assuming some parametric models that allow the interpolation of the unobserved areas, e.g., refer to \citep{articlerosenbaum2002observational,kallus2018confounding}. }
The following example illustrates the application of the IPW method.

\begin{example}\label{exp:_4_1_dtr3}
	Consider again the CDM $\Tuple{\1M^*, \Pi, Y}$ described in Eq.~\ref{eq:_3_1_dtr1} where coefficients $\alpha_1 = \alpha_2 = 0$. We will apply IPW estimation to evaluate the effects of the policy $\pi =(X_1 \gets 0, X_2 \gets  1)$ from the observational distribution $P(S_1, X_1, S_2, X_2, Y)$. Applying the estimation formula provided by Thm.~\ref{thm:_4_1_ipw} gives
	\begin{align}
		\E^{\textsc{ipw}}_{X_1 \gets 0, X_2 \gets 1}\left [Y \right] & = \sum_{s_1, x_1, s_2, x_2} P\left (s_1, x_1, s_2, x_2, Y = 1\right) \frac{\I\{x_1 = 0\} }{P\left(x_1\mid s_1 \right)} \frac{\I\{x_2 = 1\}}{P\left(x_2 \mid s_1, x_1, s_2\right)}
	\end{align}

	\begin{table}[t]
		\centering
		\renewcommand{\arraystretch}{1.25}
		\begin{tabular}{|ccccc|c|ccccc |c|}
			$S_1$ & $X_1$ & $S_2$ & $X_2$ & $Y$ & $P(s_1, x_1, s_2, x_2, y)$ & $S_1$ & $X_1$ & $S_2$ & $X_2$ & $Y$ & $P(s_1, x_1, s_2, x_2, y)$ \\
			\hline
			\hline
			0     & 0     & 0     & 0     & 0   & 0.0128                     & 1     & 0     & 0     & 0     & 0   & 0.0042                     \\
			0     & 0     & 0     & 0     & 1   & 0.0466                     & 1     & 0     & 0     & 0     & 1   & 0.0011                     \\
			0     & 0     & 0     & 1     & 0   & 0.0009                     & 1     & 0     & 0     & 1     & 0   & 0.0011                     \\
			0     & 0     & 0     & 1     & 1   & 0.0585                     & 1     & 0     & 0     & 1     & 1   & 0.0042                     \\
			0     & 0     & 1     & 0     & 0   & 0.0013                     & 1     & 0     & 1     & 0     & 0   & 0.0004                     \\
			0     & 0     & 1     & 0     & 1   & 0.0049                     & 1     & 0     & 1     & 0     & 1   & 0.0001                     \\
			0     & 0     & 1     & 1     & 0   & 0.0269                     & 1     & 0     & 1     & 1     & 0   & 0.0098                     \\
			0     & 0     & 1     & 1     & 1   & 0.0982                     & 1     & 0     & 1     & 1     & 1   & 0.0027                     \\
			0     & 1     & 0     & 0     & 0   & 0.0442                     & 1     & 1     & 0     & 0     & 0   & 0.1013                     \\
			0     & 1     & 0     & 0     & 1   & 0.0121                     & 1     & 1     & 0     & 0     & 1   & 0.0008                     \\
			0     & 1     & 0     & 1     & 0   & 0.0008                     & 1     & 1     & 0     & 1     & 0   & 0.0796                     \\
			0     & 1     & 0     & 1     & 1   & 0.0554                     & 1     & 1     & 0     & 1     & 1   & 0.0218                     \\
			0     & 1     & 1     & 0     & 0   & 0.0051                     & 1     & 1     & 1     & 0     & 0   & 0.0130                     \\
			0     & 1     & 1     & 0     & 1   & 0.0014                     & 1     & 1     & 1     & 0     & 1   & 0.0001                     \\
			0     & 1     & 1     & 1     & 0   & 0.0281                     & 1     & 1     & 1     & 1     & 0   & 0.2566                     \\
			0     & 1     & 1     & 1     & 1   & 0.1028                     & 1     & 1     & 1     & 1     & 1   & 0.0040                     \\
		\end{tabular}
		\caption{The observational distribution $P(X_1, X_2, S_1, S_2, Y)$ evaluated in the 2-stage DTR environment described in Example~\ref{exp:_3_1_dtr}.}
		\label{tab:_4_1_dtr1}
	\end{table}
	The detailed parametrization of the observational distribution $P(S_1, X_1, S_2, X_2, Y)$ is provided in Table \ref{tab:_4_1_dtr1}. The above equation could be further written as:
	\begin{align}
		\E^{\textsc{ipw}}_{X_1 \gets 0, X_2 \gets 1}\left [Y \right] & = \frac{P\left (S_1 = 0, X_1 = 0, S_2 = 0, X_2 = 1, Y = 1\right)}{P\left(X_1 = 0\mid S_1 = 0 \right)P\left(X_2 = 1 \mid S_1 = 0, X_1 = 0, S_2 = 0\right)} \\
		                                                             & + \frac{P\left (S_1 = 0, X_1 = 0, S_2 = 1, X_2 = 1, Y = 1\right)}{P\left(X_1 = 0\mid S_1 = 0 \right)P\left(X_2 = 1 \mid S_1 = 0, X_1 = 0, S_2 = 1\right)} \\
		                                                             & + \frac{P\left (S_1 = 1, X_1 = 0, S_2 = 0, X_2 = 1, Y = 1\right)}{P\left(X_1 = 0\mid S_1 = 1 \right)P\left(X_2 = 1 \mid S_1 = 1, X_1 = 0, S_2 = 0\right)} \\
		                                                             & + \frac{P\left (S_1 = 1, X_1 = 0, S_2 = 1, X_2 = 1, Y = 1\right)}{P\left(X_1 = 0\mid S_1 = 1 \right)P\left(X_2 = 1 \mid S_1 = 1, X_1 = 0, S_2 = 1\right)}
	\end{align}
	Evaluating the above equation gives $\E^{\textsc{ipw}}_{X_1 \gets 0, X_2 \gets 1}\left [Y \right] = 0.6757$, which matches the expected reward in Eq.~\ref{eq:_3_1_dtr1}, evaluated directly in the SCM $\1M^*$. \hfill $\blacksquare$
\end{example}

The numeric example above is one instantiation of the larger implication of the theorem showing that the agent does not have to go online and try different actions, but it can learn a policy by simply re-weighting the observational data whenever the conditions of the theorem hold.
Interestingly, we further note that  by iteratively applying Bayes' rule, the IPW formula in Eq.~\ref{eq:_4_1_ipw} can be written as
\begin{align}
	\E_{\pi}\left [\1R(\*Y)\right] = \sum_{\bar{\*x}_H, \bar{\*s}_H} \E\left[\1R(\*Y) \mid \bar{\*x}_H, \bar{\*s}_H \right]\prod_{i = 1}^H  P\left ( s_i  \mid \bar{\*x}_{i-1}, \bar{\*s}_{i-1}\right)  \pi_{i} \left(x_i \mid \*s_i\right).  \label{eq:_4_1_dp_derivation}
\end{align}
Computing the above equation following a reverse topological ordering $i = H, \dots, 1$ over actions leads to an alternative algorithm for evaluating the effects of candidate policies, based on \emph{dynamic programming} (for short, DP). DP was first introduced in \citep{bellman57} and has been widely applied in reinforcement learning \citep{puterman1994markov,sutton1998reinforcement}. The following proposition describes details for applying DP for off-policy evaluation from the observational distribution, provided that the NUC condition holds.

\begin{restatable}[Dynamic Programming]{theorem}{thmdp}\label{thm:_4_1_dp}
	Let $\langle \1M^*, \Pi, \1R \rangle$ be a CDM where $\Pi = \left \{\langle X_i, \*S_i \rangle \right \}_{i = 1}^H$ and $\1R: \D(\*Y) \mapsto \3R$. If NUC holds, for any $\pi \in \Pi$, the expected reward $\E_{\pi}\left[ \1R(\*Y) \right]$ is computable from the joint distribution $P\left (\*V \right)$ as follows:
	\begin{align}
		\E_{\pi}\left [\1R(\*Y)\right] =  \E \left[ \sum_{x_1} Q_{\pi}^{(1)}(x_1, \*S_1) \pi_1(x_1\mid \*S_1)\right], \label{eq:_4_1_dp}
	\end{align}
	where the value function $Q_{\pi}^{(i)} (\bar{\*x}_{i}, \bar{\*s}_i)$, for $i = 1, \dots, H-1$,  is given by:
	\begin{align}
		               & Q_{\pi}^{(i)} (\bar{\*x}_{i}, \bar{\*s}_i) = \E\left [\sum_{x_{i+1}}  Q_{\pi}^{(i+1)} (\bar{\*x}_{i+1}, \bar{\*s}_i, \*S_{i+1}) \pi_{i+1}\left(x_{i+1} \mid \*S_{i+1} \right) \mmid \bar{\*x}_{i}, \bar{\*s}_i \right] \\
		\text{and}\;\; & Q_{\pi}^{(H)} (\bar{\*x}_{H}, \bar{\*s}_H) = \E\left[\1R(\*Y) \mid \bar{\*x}_H, \bar{\*s}_H\right]
	\end{align} \hfill $\blacksquare$
\end{restatable}
It follows from the derivation in Eq.~\ref{eq:_4_1_dp_derivation} that IPW and DP estimation are, in principle, equivalent. That is, they return the same evaluation for $\E_{\pi}\left [Y\right]$ provided with the same candidate policy $\pi$ and observational data $P(\*V)$.
The following example illustrates this equivalence.

\begin{example}\label{exp:_4_1_dtr4}
	Consider again the CDM $\Tuple{\1M^*, \Pi, Y}$ described in Eq.~\ref{eq:_3_1_dtr1} where coefficients $\alpha_1 = \alpha_2 = 0$. We will apply the DP estimation to evaluate the effect of the policy $\pi = (X_1 \gets 0, X_2 \gets 1)$. Thm.~\ref{thm:_4_1_dp} allows us to estimate the expected reward $ \invE{Y}{\pi}$ from the observational distribution $P(S_1, X_1, S_2, X_2, Y)$ as follows:
	\begin{align}
		 & Q^{(1)}_{\pi}(s_1, x_1) = \sum_{s_2, x_1} Q^{(2)}_{\pi}(s_1, x_1, s_2, x_2) \I\{x_2 = 1\} P\left( s_2 \mid x_1, s_1 \right) \\
		 & Q^{(2)}_{\pi}(s_1, x_1, s_2, x_2) = P \left (Y = 1 | s_1, x_1, s_2, x_2 \right)
	\end{align}

	\begin{table}[t]
		\centering
		\hfill
		\begin{subtable}{\linewidth}
			\centering
			\begin{tabular}{|cc|c|cc|c|}
				$S_1$ & $X_1$ & $Q^{(1)}_{\pi}$ & $S_1$ & $X_1$ & $Q^{(1)}_{\pi}$ \\
				\hline
				\hline
				0     & 0     & 0.8799          & 1     & 0     & 0.4716          \\
				0     & 1     & 0.8749          & 1     & 1     & 0.1003          \\
			\end{tabular}
			\caption{$Q^{(1)}_{\pi}(s_1, x_1)$}\label{tab:_4_1_dtr2_a}
		\end{subtable}
		\hfill
		\vspace{0.1in}
		\begin{subtable}{\linewidth}
			\centering
			\renewcommand{\arraystretch}{1.25}
			\begin{tabular}{|cccc|c|cccc |c|}
				$S_1$ & $X_1$ & $S_2$ & $X_2$ & $Q^{(2)}_{\pi}$ & $S_1$ & $X_1$ & $S_2$ & $X_2$ & $Q^{(2)}_*$ \\
				\hline
				\hline
				0     & 0     & 0     & 0     & 0.7851          & 1     & 0     & 0     & 0     & 0.2149      \\
				0     & 0     & 0     & 1     & 0.9846          & 1     & 0     & 0     & 1     & 0.7851      \\
				0     & 0     & 1     & 0     & 0.7851          & 1     & 0     & 1     & 0     & 0.2149      \\
				0     & 0     & 1     & 1     & 0.7851          & 1     & 0     & 1     & 1     & 0.2149      \\
				0     & 1     & 0     & 0     & 0.2149          & 1     & 1     & 0     & 0     & 0.0008      \\
				0     & 1     & 0     & 1     & 0.9846          & 1     & 1     & 0     & 1     & 0.2149      \\
				0     & 1     & 1     & 0     & 0.2149          & 1     & 1     & 1     & 0     & 0.0008      \\
				0     & 1     & 1     & 1     & 0.7851          & 1     & 1     & 1     & 1     & 0.0154      \\
			\end{tabular}
			\caption{$Q^{(2)}_{\pi}(s_1, x_1, s_2, x_2)$}\label{tab:_4_1_dtr2_b}
		\end{subtable}\hfill\null

		\caption{Evaluation of value functions $Q^{(1)}_{\pi}(s_1, x_1), Q^{(2)}_{\pi}(s_1, x_1, s_2, x_2)$ for the policy $\pi = (X_1 \gets 0, X_2 \gets 1)$ in evaluated in 2-stage DTR environment described in Example~\ref{exp:_3_1_dtr}.}
		\label{tab:_4_1_dtr2}
	\end{table}

	\noindent We compute the parametrization of value functions $Q^{(1)}_{\pi}(s_1, x_1), Q^{(2)}_{\pi}(s_1, x_1, s_2, x_2)$ and provide them in Table~\ref{tab:_4_1_dtr2}. The expected reward of the policy $\pi = (X_1 \gets 0, X_2 \gets 1)$ is computable as
	\begin{align}
		\E^{\textsc{dp}}_{X_1 \gets 0, X_2 \gets 1}[Y] & = \sum_{s_1}Q^{(1)}_{\pi}(s_1, x_1) \I\{x_1 = 0\} P(s_1)
	\end{align}
	Evaluating the above equation gives $\E^{\textsc{dp}}_{X_1 \gets 0, X_2 \gets 1}\left [Y \right] = 0.6757$, which matches the expected reward in Example~\ref{exp:_3_1_dtr}, evaluated in the SCM $\1M^*$. \hfill $\blacksquare$
\end{example}

Once IPW and DP evaluation formulas are obtained, efficient methods in the literature estimate the expected rewards of candidate policies from finite samples drawn from the observational distribution $P(\*V)$. For the IPW evaluation, the agent could weigh every observed reward signal with the odds ratio between the target policy $\pi$ and the propensity score $P(x_i \mid \bar{\*x}_{i-1}, \bar{\*s}_i)$, i.e., the second term of Eq.~\ref{eq:_4_1_ipw}. The expected reward is estimable by computing the empirical mean on the weighted rewards. This IPW estimate was first developed to estimate the effects of candidate policies in the single-stage decision setting, i.e., the decision horizon $H = 1$, but later adapted to the problem of estimating the effects of policies in the sequential setting, with the decision horizon $H > 1$. See \citep{rosenbaum1983central,robins2000marginal,wang2012evaluation,nahum2012experimental} for detailed explanations of how to apply IPW estimation from finite observations provided with the NUC condition.

As for the DP evaluation, the agent first approximates the state-action value function Q in Eq.~\ref{eq:_4_1_dp} from the observational data using parametric models and then computes the expected reward of a candidate policy \citep{tsitsiklis1996analysis}. For instance, it could approximate Q-functions using a parametric family of linear functions; function parameters are obtainable using the standard least squares regression \citep{murphy2005generalization}. Other more flexible families of parametric models for the Q-functions include regression trees \citep{ernst2005tree}, kernels \citep{ormoneit2002kernel}, and neural networks \citep{mnih2013playing}. The value function approximation has been studied in the literature under the rubrics of batch reinforcement learning \citep{bertsekas1995neuro,lange2012batch}.

When the expected rewards of candidate policies are computable, the agent could then search over the policy space $\Pi$ and obtain an optimal policy estimate using policy gradient \citep{sutton1999policy}. Moreover, when the Q-function $Q^{(i)}_{\pi}(\bar{\*x}_i, \bar{\*s}_i)$ only relies on the state-action value $x_i, \*s_i$ for every stage of intervention $i = 1, \dots, H$, one could solve for an optimal policy through iteratively optimizing every decision rule $\pi_i$ following a reverse topological ordering over actions $\*X$ \citep{lauritzen2001representing,koller2003multi}. This local optimization procedure is analogous to the well-celebrated Q-learning algorithm \citep{watkins1992q}. 
We refer readers to \citep{uehara2022review} for a recent literature review on standard off-policy evaluation from finite observational data under the NUC assumption.

Despite the positive results discussed above, IPW and DP methods may fail to recover the effects of candidate policies from observational data whenever the NUC condition (Def.~\ref{def:_4_1_nuc}) does not hold. The following examples demonstrate this looming challenge.
\begin{example}\label{exp:_4_1_mab}
	Consider a MAB model $\left \langle \1M^*, \left \{\langle X, \emptyset \rangle \right\}, Y \right \rangle$ graphically described in Fig.~\ref{fig:_3_1_mab} where SCM $\1M^*$ is defined in Example~\ref{exp:_2_1_mab}. In this model, the NUC condition does not hold due to unobserved confounder $U$ affecting action $X$ and reward $Y$. This implies that IPW is not necessarily applicable to recover the expected reward $\invE{Y}{x}$ from the observational distribution $P(X, Y)$.

	We will proceed regardless and try to learn a policy $\pi: X \gets 0$. Applying Thm.~\ref{thm:_4_1_ipw} gives:
	\begin{align}
		\E^{\textsc{ipw}}_{X \gets 0}\left [Y \right] & = \sum_{x} \E \left[ Y \mid x \right] P(x) \frac{\I\{x = 0\}}{P(x)} \\
		                                              & = \frac{P\left (X = 0, Y = 1\right)}{P(X = 0)}.
	\end{align}
	Evaluating the above equation gives $\E^{\textsc{ipw}}_{X \gets 0}\left [Y \right] = 0$, which deviates significantly from the actual expected reward $\invE{Y}{X \gets 0} = 0.4$ (Eq.~\ref{eq:_2_2_mab5}) evaluated in SCM $\1M^*$.

	We also apply DP to evaluate the effect of pulling arm $X \gets 0$, and through Thm.~\ref{thm:_4_1_dp}, we have: 
	\begin{align}
		\E^{\textsc{dp}}_{X \gets 0}\left [Y \right] & =\sum_{x}  \E \left[Y \mid x \right]\I\{x = 0\} \\
		                                             & = P\left( Y = 1 \mid X = 0 \right)
	\end{align}
	Evaluating the above equation gives $\E^{\textsc{dp}}_{X \gets 0}\left [Y \right] = 0$, which, again, deviates from the actual expected reward $\invE{Y}{X \gets 0} = 0.4$ (Eq.~\ref{eq:_2_2_mab5}) evaluated directly in SCM $\1M^*$. \hfill $\blacksquare$
\end{example}
The above examples show that the validity of off-policy learning methods introduced so far hinges on the NUC assumption. Such a critical assumption could be fragile and does not necessarily hold in many practical settings. For instance, in electronic healthcare records, the physician might prescribe a new drug to patients who are more likely to access high-quality healthcare, thus making the drug appear more effective. For the remainder of this section, we will introduce alternative policy learning assumptions and methods to overcome this issue.

\subsection{Online Learning}\label{sec:_4_2}
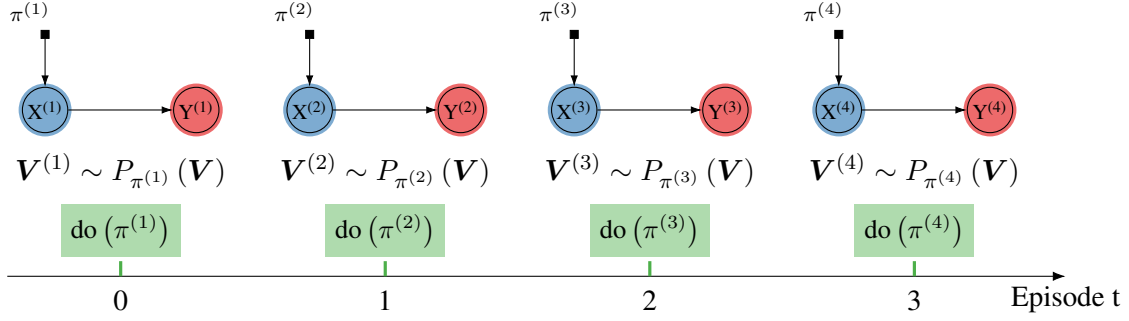
\begin{figure}[t]
	\centering
	\begin{tikzpicture}
		\def\outerr{3.5}
		\def\innerr{3.5}
		\def\dist{3.5}

		\draw[->, >={Latex}] (-1.5,0) -- (3*\dist+2,0) node[below] {Episode t};

		\node[vertex, minimum width=6mm] (X1) at (-1, 2.2) {X\textsuperscript{(1)}};
		\node[vertex, minimum width=6mm] (Y1) at (1, 2.2) {Y\textsuperscript{(1)}};
		\node[regime] (p1) at (-1, 3.2) {};
		\node[draw=none] (text) at (-1.2, 3.5) {\scriptsize $\pi^{(1)}$};
		\draw[dir] (X1) -- (Y1);
		\draw[dir] (p1) -- (X1);

		\node[vertex, minimum width=6mm] (X2) at (\dist-1, 2.2) {X\textsuperscript{(2)}};
		\node[vertex, minimum width=6mm] (Y2) at (\dist+1, 2.2) {Y\textsuperscript{(2)}};
		\node[regime] (p2) at (\dist-1, 3.2) {};
		\node[draw=none] (text) at (\dist-1.2, 3.5) {\scriptsize $\pi^{(2)}$};
		\draw[dir] (X2) -- (Y2);
		\draw[dir] (p2) -- (X2);

		\node[vertex, minimum width=6mm] (X3) at (2*\dist-1, 2.2) {X\textsuperscript{(3)}};
		\node[vertex, minimum width=6mm] (Y3) at (2*\dist+1, 2.2) {Y\textsuperscript{(3)}};
		\node[regime] (p3) at (2*\dist-1, 3.2) {};
		\node[draw=none] (text) at (2*\dist-1.2, 3.5) {\scriptsize $\pi^{(3)}$};
		\draw[dir] (X3) -- (Y3);
		\draw[dir] (p3) -- (X3);

		\node[vertex, minimum width=6mm] (X4) at (3*\dist-1, 2.2) {X\textsuperscript{(4)}};
		\node[vertex, minimum width=6mm] (Y4) at (3*\dist+1, 2.2) {Y\textsuperscript{(4)}};
		\node[regime] (p4) at (3*\dist-1, 3.2) {};
		\node[draw=none] (text) at (3*\dist-1.2, 3.5) {\scriptsize $\pi^{(4)}$};
		\draw[dir] (X4) -- (Y4);
		\draw[dir] (p4) -- (X4);

		\node (d1) at (0, 1.4) {$\*V^{(1)} \sim \inv{\*V}{\pi^{(1)}}$};
		\node (d2) at (\dist, 1.4) {$\*V^{(2)} \sim \inv{\*V}{\pi^{(2)}}$};
		\node (d3) at (2*\dist, 1.4) {$\*V^{(3)} \sim \inv{\*V}{\pi^{(3)}}$};
		\node (d4) at (3*\dist, 1.4) {$\*V^{(4)} \sim \inv{\*V}{\pi^{(4)}}$};

		\draw[very thick, bettergreen, -] (0, 0) -- (0, 0.2);
		\draw[very thick, bettergreen, -] (\dist, 0) -- (\dist, 0.2);
		\draw[very thick, bettergreen, -] (2*\dist, 0) -- (2*\dist, 0.2);
		\draw[very thick, bettergreen, -] (3*\dist, 0) -- (3*\dist, 0.2);

		\node [below] at (0, -0.05) { 0};
		\node [below] at (\dist, -0.05) { 1};
		\node [below] at (2*\dist, -0.05) { 2};
		\node [below] at (3*\dist, -0.05) { 3};

		\node [fill=bettergreen!45] at (0, 0.6) {\small $\doo\left (\pi^{(1)} \right)$};
		\node [fill=bettergreen!45] at (\dist, 0.6) {\small $\doo\left(\pi^{(2)} \right)$};
		\node [fill=bettergreen!45] at (2*\dist, 0.6) {\small $\doo\left(\pi^{(3)} \right)$};
		\node [fill=bettergreen!45] at (3*\dist, 0.6) {\small $\doo\left(\pi^{(4)} \right)$};

		\begin{pgfonlayer}{back}
			\node[circle,fill=betterblue!65,draw=none,minimum size=2*\innerr mm] at (X1) {};
			\node[circle,fill=betterred!65,draw=none,minimum size=2*\innerr mm] at (Y1) {};
			\node[circle,fill=betterblue!65,draw=none,minimum size=2*\innerr mm] at (X2) {};
			\node[circle,fill=betterred!65,draw=none,minimum size=2*\innerr mm] at (Y2) {};
			\node[circle,fill=betterblue!65,draw=none,minimum size=2*\innerr mm] at (X3) {};
			\node[circle,fill=betterred!65,draw=none,minimum size=2*\innerr mm] at (Y3) {};
			\node[circle,fill=betterblue!65,draw=none,minimum size=2*\innerr mm] at (X4) {};
			\node[circle,fill=betterred!65,draw=none,minimum size=2*\innerr mm] at (Y4) {};
		\end{pgfonlayer}
	\end{tikzpicture}
	\caption{Temporal diagram showing an online learning agent interacting with the environment for repeated episodes.}
	\label{fig:_4_2_online}
\end{figure}

An online learning agent evaluates candidate policies in space $\Pi$ by directly deploying them in the underlying environment. A temporal graph illustrating this interaction is described in Fig.~\ref{fig:_4_2_online}. In causal language, the agent intervenes in the SCM $\1M^*$ for repeated episodes $t = 1, \dots, T$. For every episode $t$, it picks a policy $\pi^{(t)} \in \Pi$, performs interventions $\doo\left(\*X \gets \pi^{(t)} \right)$ on actions $\*X$ following $\pi^{(t)}$, and receives subsequent observations $\*V^{(t)} \sim \inv{\*V}{\pi^{(t)}}$.

Formally, an online learning task is described by the following signature:
\begin{align}
	\1T_{\text{on}} = \left \langle \1R =\text{do}, \1A=\emptyset, \Pi = \{\langle X_i, \*S_i \rangle \}_{i = 1}^H, \1R = \D(\*Y) \mapsto \3R \right \rangle
\end{align}
To see the specific optimization in this task, the agent will search for a policy $\pi^*$ such that
\begin{align}\label{eq:opt-crl-on}
	 & \pi^* = \argmax_{\pi \in \Pi} \invEE{ \1R \left (\*Y \right ) \; \big \vert \; \textcolor{red}{\;\mathcal{D}_{\text{exp}} \sim \inv{\*V}{\*x}}}{\pi}{\1M^*}, 
\end{align}
Compared with the off-policy learning task ($\1T_{\text{off}}$), an online agent does not make additional structural assumptions about the underlying environment ($\1A = \emptyset$), beyond the temporal ordering over state and action variables in the policy space $\Pi$ (Def.~\ref{def:_3_space}). This means that the NUC assumption (Def.~\ref{def:_4_1_nuc}) discussed earlier does not necessarily hold.
Note that for every policy $\pi \in \Pi$, in the submodel $\1M^*_{\pi}$ induced by intervention $\doo(\pi)$, all input covariates $\*S_i$ affecting every action $X_i \in \*X$ and other variables in the system are observed and measured. This means that the NUC condition is implied in the post-interventional system. The following proposition formalizes this intuition.

\begin{restatable}[Experimental NUC]{lemma}{lemonlinenuc}\label{lem:_4_2_nuc}
	Let $\langle \1M^*, \Pi, \1R \rangle$ be a CDM where $\Pi = \left \{\langle X_i, \*S_i \rangle \right \}_{X_i \in \*X}$ and $\1R: \D(\*Y) \mapsto \3R$. For any policy $\pi \in \Pi$, the NUC condition (Def.~\ref{def:_4_1_nuc}) holds in $\Tuple{\1M^*_{\pi}, \Pi, \1R }$ induced by intervention $\doo(\pi)$. \hfill $\blacksquare$
\end{restatable}
Following the NUC condition, the result above can be seen as a formal justification for using standard off-policy learning methods, including IPW and DP, to evaluate other candidate policies $\pi' \in \Pi$ from data drawn from the interventional distribution $\inv{\*V}{\pi}$ through Thms.~\ref{thm:_4_1_ipw} and \ref{thm:_4_1_dp}.

\begin{example}\label{exp:_3_2_dtr1}
	Consider the 2-stage DTR $\Tuple{\1M^*, \Pi, Y}$ where SCM $\1M^*$ is described in Eq.~\ref{eq:_3_1_dtr1} with coefficients $\beta_i = -3$; and the policy space $\Pi = \Braces{\Tuple{X_1, \braces{S_1}}, \Tuple{X_2, \braces{S_1, X_1, S_2}}}$. It has been shown in Example~\ref{exp:_4_1_nuc2} that the NUC condition does not hold in this model. Now consider an online agent that is deployed in the environment and follows a policy $\pi = (\pi_1, \pi_2)$, where
	\begin{align}
		\pi_i \triangleq \I \Braces{3S_i + U_i > 0}
	\end{align}
	and $U_i$, $i = 1, 2$, are independent variable drawn from distribution $\texttt{Logistic}(0, 1)$. Performing interventions $\doo(\pi)$ following this policy leads to a submodel described by the following tuple
	\begin{align}
		\M^*_{\pi} = \Tuple{ \*U = \{U, U_1, \dots, U_5\}, \*V = \{S_1, X_1, S_2, X_2, Y\}, \2F_{\pi}, P(\*U)},
	\end{align}
	where the structural functions $\2F_{\pi}$ are given by
	\begin{align}\label{eq:_4_2_dtr1}
		\2F_{\pi} = \begin{cases}
			            S_1 \gets \I\{U_3 > 0\},                          \\
			            X_1 \gets \I\{3S_1  + U_1 > 0\},                  \\
			            S_2 \gets \I\{0.1 + 0.1 S_1 + 0.1X_1 + U_4 > 0\}, \\
			            X_2 \gets \I\{3S_2  + U_2 > 0\},                  \\
			            Y \gets \I\{ 3 U - 3S_1 - 3X_1 - 3S_1X_1 + 3 X_2 - 3S_2X_2 + 3X_1X_2> 0\}.
		            \end{cases}
	\end{align}
	In the above equations, the unobserved confounder $U$ no longer affects treatments $X_1, X_2$, and the NUC condition holds in the submodel $\Tuple{\1M^*_{\pi}, \Pi, Y}$. See Example~\ref{exp:_4_1_nuc1} for a detailed discussion. \hfill $\blacksquare$
\end{example}

\subsubsection{Randomized Controlled Trials}
We will discuss different algorithms that systematize the discussion above and operate over the environment in an online fashion. 
The first algorithm we consider will be called \textit{randomized controlled trials} (for short, \texttt{RCT}) and follows the idea of randomization, which dates back at least to \citep{fisher:35}. 
Fisher's very motivation for considering randomizing the treatment assignment was to eliminate the influence of unmeasured confounders in the collected data.
\footnote{Fisher's motivation at the time was to understand the effect of some pesticides on the yield of certain crops \citep{fisher:26}. 
Farmers were biased in how they used pesticides which tended to be applied in the best parts of the land. At the end of the season, the pesticides had a higher effect. 
Fisher was suspicious of this procedure since, in modern terminology, the NUC assumption did not hold. 
He then had the idea of allocating the treatment randomly to the different plots of land. 
This insight departed from a tradition led by Pearson \citep{pearson:11} and started a new and fundamental discipline of \textit{experimental design} \citep{fisher:35}.   }
It is a ``explore-then-commit'' strategy where the agent first explores the environment by determining values of actions $\*X$ uniformly at random for a fixed number of times, and then exploits by committing to a policy that appeared best during exploration.

Alg.~\ref{alg:_4_2_rct} shows the detailed experimental design of \texttt{RCT}. It interacts with the underlying environment by repeated episodes of interventions $t = 1, 2, \dots$. More specifically, for every episode $t$, \texttt{RCT} selects a policy $\pi^{(t)} \in \Pi$, performs an intervention $\doo\left (\pi^{(t)}\right)$, and receives a subsequent observation $V^{(t)} \sim \inv{\*V}{\pi^{(t)}}$. 
During the initialization, the algorithm considers a natural number $N \in \3N$, called the \emph{total number of trials}. It determines the total episodes of interventions for the algorithm to explore the environment before committing to a specific policy. For episode $t \leq N$, the algorithm selects a uniform policy $\pi_{\textsc{unif}}$ which determines values of every action $X^{(t)}_i$, $i = 1, \dots, H$, uniformly at random during the exploration phase. Note that the NUC condition holds under $\doo(\pi_{\textsc{unif}})$ intervention (Lem.~\ref{lem:_4_2_nuc}). This means that \texttt{RCT} could compute reward estimates $\hat{\E}^{(N)}_{\pi}[Y]$ for candidate policies $\pi \in \Pi$ from the experimental data $\inv{\*V}{\pi_{\textsc{unif}}}$ collected during the exploration. The evaluation procedures were previously described in Thms.~\ref{thm:_4_1_ipw} and \ref{thm:_4_1_dp}. Finally, \texttt{RCT} selects a policy with the highest empirical reward estimates and commits to it for all future episodes $t > N$.
\footnote{The \texttt{RCT} algorithm described in Alg.~\ref{alg:_4_2_rct} is also referred to as sequential multiple assignment randomized trials (for short, SMART \citep{murphy2005experimental}) where every subject (at episode $t$) is randomized multiple times, one for each stage of decision $X_1, \dots, X_H$. When the decision horizon $H = 1$, Alg.~\ref{alg:_4_2_rct} reduces to the original randomized controlled trials introduced by \citep{fisher:35}.}

\begin{algorithm}[t]
	\caption{Randomized Controlled Trails (\texttt{RCT})}
	\label{alg:_4_2_rct}
	\setlength{\textfloatsep}{0pt}
	\begin{algorithmic}[1]
		\Require the policy space $\Pi$, the total number of trials $N \in \3N$.
		\ForAll{episodes $t = 1, 2, \dots $}
		\State Choose a policy $\pi^{(t)}$ as follows.
		\If{$t \leq N$}
		\State Let $\pi^{(t)}$ be a uniform policy 
		\begin{align}
			\pi_{\textsc{unif}} = \left ( X_1 \sim \texttt{Unif}(\D(X_1)), \dots,  X_H \sim \texttt{Unif}(\D(X_H)) \right).
		\end{align}
		\Else
		\State Let $\pi^{(t)} = \argmax_{\pi \in \Pi }\hat{\E}^{(N)}_{\pi}[Y]$.
		\EndIf
		\State Perform $\doo(\pi^{(t)})$ for episode $t$ and receive observations $\*V^{(t)}$.
		\EndFor
	\end{algorithmic}
\end{algorithm}

\begin{example}\label{exp:_4_2_mab1}
	We illustrate \texttt{RCT} in an MAB model $\langle \1M^*, \Set{\Tuple{X, \emptyset}}, Y \rangle$ where $\1M^*$ is an SCM described in Eq.~\ref{eq:_2_1_mab} consisting of an arm choice $X$ and reward signal $Y$. For any policy $\pi(x)$, the expected reward $\invE{Y}{\pi}$ is given by:
	\begin{align}
		\invE{Y}{\pi} = \pi(X = 0) \invE{Y}{X\gets 0} +  \pi(X = 1) \invE{Y}{X\gets 1}
	\end{align}
	This means that for any stochastic policy $\pi(x) > 0$, $\forall x \in \D(X)$, its performance could always be improved by a deterministic policy $X \gets x^*$ where the optimal arm choice given by  
\footnote{It can be shown that in any fixed CDM $\langle \1M^*, \Pi, \1R \rangle$, the performance of a stochastic policy $\pi$ could always be improved by a deterministic one. This means that one could optimize the expected reward $\invE{\1R(\*Y)}{\pi}$ over only deterministic policies $\pi \in \Pi$ without loss of generality \cite[Lem.~2.1]{liu2012belief}.} 
\begin{align}
x^* = \argmax_{x \in \{0, 1\}} \invE{Y}{x}
	\end{align}
It is thus sufficient to estimate the expected reward $\invE{Y}{x}$ induced by atomic intervention $\doo(x)$.

Fix the total number of trials $N$ (say, $N = 1,000$). For every episode $t \leq N$, \texttt{RCT} selects an action $X^{(t)}$ uniformly at random over the binary domain $\{0, 1\}$, performs an intervention $\doo\left(X \gets X^{(t)} \right)$, and receives a reward $Y^{(t)} \sim \inv{P}{X^{(t)}}$. 
When the exploration phase is done ($t > N$), \texttt{RCT} estimates the expected reward $\invE{Y}{x}$ for every action $x \in \{0, 1\}$ from finite samples $\left \{X^{(t)}, Y^{(t)}\right\}_{t = 1, \dots, N}$. Applying the DP estimation formula (Thm.~\ref{thm:_4_1_dp}) implies
	\begin{align}
		\invE{Y}{x} = \E_{\pi_{\textsc{unif}}}\left[Y |x\right]
	\end{align}
	The empirical reward estimate for an arm $x$ is thus given by
	\begin{align}
		\hat{\E}^{(N)}_x[Y] = \frac{1}{N(x)}\sum_{t = 1}^N Y^{(t)} \I\{X^{(t)} = x\}, \label{eq:_4_2_dp}
	\end{align}
	where $N(x) = \sum_{t = 1}^{N} \I\left \{X^{(t)} = x \right\}$ is the total occurrence of event $X^{(t)} = x$ up to episode $N$. We could also apply the IPW estimation (Thm.~\ref{thm:_4_1_ipw}) and obtain:
	\begin{align}
		\invE{Y}{x} & = \sum_{x'} \invE{Y \mid x'}{\pi_{\textsc{unif}}} \frac{\I\{x' = x\}}{\pi_{\textsc{unif}}\left( x\right)} \\
		            & =\E_{\pi_{\texttt{unif}}}\left[Y \frac{\I\{X = x\}}{\pi_{\textsc{unif}}\left( X\right)}\right]
	\end{align}
	The last step follows from the definition of expected values. Given samples $\left \{X^{(i)}, Y^{(i)} \right\}_{i = 1, \dots, N}$ collected by \texttt{RCT} during exploration ($t < N$), the IPW empirical estimate for the expected reward of pulling arm $x$ is thus given by
	\begin{align}
		\hat{\E}^{(N)}_x[Y] = \frac{1}{N} \sum_{t = 1}^N Y^{(t)} \frac{\I\{X^{(t)} = x\}}{\pi_{\textsc{unif}}\left(X^{(t)} \right)} = \frac{2}{N} \sum_{t = 1}^N Y^{(t)} \I\{X^{(t)} = x\}. \label{eq:_4_2_ipw}
	\end{align}
	The last step holds since the uniform policy $\pi_{\textsc{unif}}(x) = 1/2$ for any action $x \in \{0, 1\}$. The empirical estimates of DP (defined in Eq.~\ref{eq:_4_2_dp}) and IPW (Eq.~\ref{eq:_4_2_ipw}) coincide if every arm $x \in \{0, 1\}$ is equally explored for episodes $t \leq N$, i.e., the total occurrences $N(x) = N/2$. \hfill $\blacksquare$
\end{example}

\paragraph{Cumulative Regret} We will analyze the performance of \text{RCT} algorithm to understand its properties and theoretical guarantees better.
Our analysis will focus on an MAB model $\langle  \1M^*_{\textsc{mab}}, \Pi, Y \rangle$, where $\1M^*_{\textsc{mab}}$ is an MAB environment graphically described in Fig.~\ref{fig:_3_1_mab} and the policy space $\Pi = \{ \langle X, \emptyset \rangle \}$. Recall that the optimal arm $x^* = \argmax_{x} \invE{Y}{x}$. We denote by $\Delta_x = \invE{Y}{x^*} - \invE{Y}{x}$ the gap between the expected reward of playing a suboptimal action $x \in \D(X)$ and the optimal arm $x^*$. 
For the analysis' convenience, we also assume that every arm $x \in \D(X)$ is played the same amount of times during the exploration phase, i.e., $N(x) = N / K$ where $K = |\D(X)|$.
	\footnote{For a uniform policy $\pi_{\texttt{unif}}$, every arm $x$ is expected to be played for $\E[N(x)] = N / K$ on average during exploration.}

There are several ways to measure the performance of online learning algorithms.
One popular criterion is to study the algorithm's \emph{cumulative regret} \citep{Auer2002}, which measures its cumulative loss relative to an optimal strategy that always selects the optimal arm $x^*$. Formally, the cumulative regret for an online learning algorithm in an MAB environment $\M^*$ after $T$ episodes of trials can be defined as:
\begin{align}
	R(T, \M^*) = \underbrace{T\invE{Y ; \M^*}{x^*}}_{\text{Optimal Reward}} -  \underbrace{\sum_{t = 1}^T \invE{Y; \M^*}{X^{(t)}}}_{\text{Realized}}. \label{eq:_4_2_cumulative_regret}
\end{align}
Naturally, minimizing the regret $R(T, \M^*)$ is equivalent to maximizing the total expected reward the agent obtains.
A reasonable objective is to design an online learning algorithm that could achieve a sublinear regret, i.e., $R(T, \M^*) = o(T)$. 
\footnote{Here we use $\mathcal{O}$ notation, where $f(n) = \mathcal{O}(g(n))$ if function $f$ is bounded above by function $g$ (up to constant factor) asymptotically. That is, $\exists k, \exists n_0$ such that $f(n) \leq kg(n)$ for $\forall n > n_0$. Similarly, $f(n) = o(g(n))$ if $\exists k, \exists n_0$, $f(n) < kg(n)$ for $\forall n > n_0$. 
For further details on this notation, see \citep[Ch.~1.3]{cormen2022introduction}.}
	This would imply that its average cumulative 
	regret per episode is converging to zero  \citep{lattimore2020bandit}, i.e.,
	\begin{align}
\lim_{T \to \infty} R(T, \M^*)/T = 0
	\end{align}
	The online learner will eventually close the gap between the optimal strategy that always commits to an optimal arm $x^*$.
	In other words, the learner is choosing an optimal arm almost all the time as the total number of episodes $T$ tends to be infinite, i.e.,
	\begin{align}
\lim_{t \to \infty} \invE{Y}{X^{(t)}} \to \invE{Y}{x^*}
	\end{align}

The analysis follows \citep[Theorem 6.1]{lattimore2020bandit}. Suppose that there are $\D(X) = \{1, \dots, K\}$ possible arms. Observe that the \texttt{RCT} algorithm only incurs regret in episodes $t$ where it plays a sub-optimal arm $x$ with $\Delta_x > 0$. The cumulative regret after $T > 1$ episodes of interventions could be written as:
\begin{align}
	R(T, \1M^*) = \sum_{x: \Delta_x > 0} \Delta_x \E \Brackets{\sum_{t = 1}^T \I\Braces{X^{(t)} = x}}
\end{align}
In the first $N$ episodes, every arm is played exactly $N / K$ times. Subsequently, it chooses a single action to maximize the empirical reward during exploration. This implies
\begin{align}
	\E \Brackets{\sum_{t = 1}^T \I\Braces{X^{(t)} = x}}& = \frac{N}{K} + (T - N) P\Parens{X^{(t)} = x}\\
	&\leq \frac{N}{K} + (T - N) P\Parens{\hat{\E}^{(N)}_x[Y] \neq \max_{x' \neq x} \hat{\E}^{(N)}_{x'}[Y]} \label{eq:_4_2_rct6}
\end{align}
The last step holds since for episodes $t \geq N$, a suboptimal arm $x \neq x^*$ is picked if and only if its empirical reward estimate $\hat{\E}^{(N)}_x[Y]$ is maximal (i.e., the largest). The error probability could thus be bounded by
	\begin{align}
		P\left(\hat{\E}^{(N)}_x[Y] \neq \max_{x' \neq x} \hat{\E}^{(N)}_{x'}[Y] \right) & \leq P\left(\hat{\E}^{(N)}_x[Y] > \hat{\E}^{(N)}_{x^*}[Y]\right)                                             \\
		                                                                              & \leq P\left(\hat{\E}^{(N)}_x[Y] - \hat{\E}^{(N)}_{x^*}[Y] - (\invE{Y}{x} - \invE{Y}{x^*}) > \Delta_x \right) \\
		                                                                              & \leq \exp\left( -\frac{N\Delta^2_x}{4K}\right) \label{eq:_4_2_rct5}
\end{align}
The last step follows the standard concentration inequality \citep{hoeffding1963probability}. Replacing the error probability Eq.~\ref{eq:_4_2_rct5} into Eq.~\ref{eq:_4_2_rct6} and summing over regret gives the following bound.
\begin{theorem}[Regrets of \texttt{RCT} \citep{lattimore2020bandit}]\label{thm:_4_2_rct}
		For an MAB $\langle \1M^*, \Set{\Tuple{X, \emptyset}}, Y \rangle$, let $Y \in \*V$ be the reward variable with support on $[0, 1]$ and let the domain of action $X$ be $\D(X) = \{1, \dots, K\}$. Fix the total number of trials $N \in \3N^+$. The regret of \texttt{RCT} in MAB $\1M^*$ after $T > 1$ episodes of interventions is bounded by
		\begin{align}
			R(T, \1M^*) \leq \underbrace{\frac{N}{K} \sum_{x: \Delta_x >0} \Delta_x}_{\text{exploration}} + \underbrace{(T - N) \sum_{x: \Delta_x >0} \Delta_x \exp\left( -\frac{N\Delta^2_x}{4K}\right) }_{\text{exploitation}} \label{eq:_4_2_rct}
		\end{align} \hfill $\blacksquare$
	\end{theorem}
The regret bound in Thm.~\ref{thm:_4_2_rct} illustrates a trade-off between the exploration and the exploitation stages of the agent's strategy. The first term is the regret cumulated during the exploration ($t \leq N$), and the second term is the expected regret of \texttt{RCT} for picking a suboptimal arm during the exploitation stage. 
If the total number of trials $N$ is large, the algorithm explores too long, and the regret cumulated during the exploration phase will be large. On the other hand, if $N$ is too small, then the empirical estimate $\hat{\E}_{x}^{(N)}[Y]$ is more likely to deviate from the expected reward $\invE{Y}{x}$, and the regret in the exploitation phase increases.
	
	One fundamental question is, therefore, how to choose the optimal number of trials $N$ to balance the amount of exploration versus exploitation. Assume that the number of arms $K = 2$ and the optimal arm $x^* = 1$, and write $\Delta = \Delta_2$. The bound in Eq.~\ref{eq:_4_2_rct} simplifies to
	\begin{align}
		R(T, \1M^*) \leq \frac{N}{2} \Delta+ T \Delta \exp\left( -\frac{N\Delta^2}{8}\right). \label{eq:_4_2_rct2}
	\end{align}
	For a large $T$, the right-hand side of Eq.~\ref{eq:_4_2_rct2} is minimized up to a rounding error by
	\begin{align}
		N =  \left \lceil{\frac{8}{\Delta^2}\log\left( \frac{T \Delta^2}{4}\right) }\right \rceil. \label{eq:_4_2_rct3}
	\end{align}
	For this choice and any $T > 1$, after a few simplifications, the regret of \texttt{RCT} is bounded by
	\begin{align}
		R(T, \1M^*) \leq \Delta + C \sqrt{T} \label{eq:_4_2_rct4}
	\end{align}
	where $C$ is a universal constant. That is, \texttt{RCT} is able to achieve a sublinear regret $R(T, \1M^*) = \mathcal{O}\left(T^{1/2} \right)$ by fine-tuning the total number of trials $N$.
	However, note that the choice of $N$ in Eq.~\ref{eq:_4_2_rct3} depends on the suboptimal gap $\Delta$ and the total number of episodes $T$, which are not necessarily known in advance. In the next section, we will see an online algorithm that does not depend on the prior knowledge of the model parameter $\Delta$ and the total episodes $T$.

	\begin{figure}[t]
		\centering
		\includegraphics[width=0.7\textwidth]{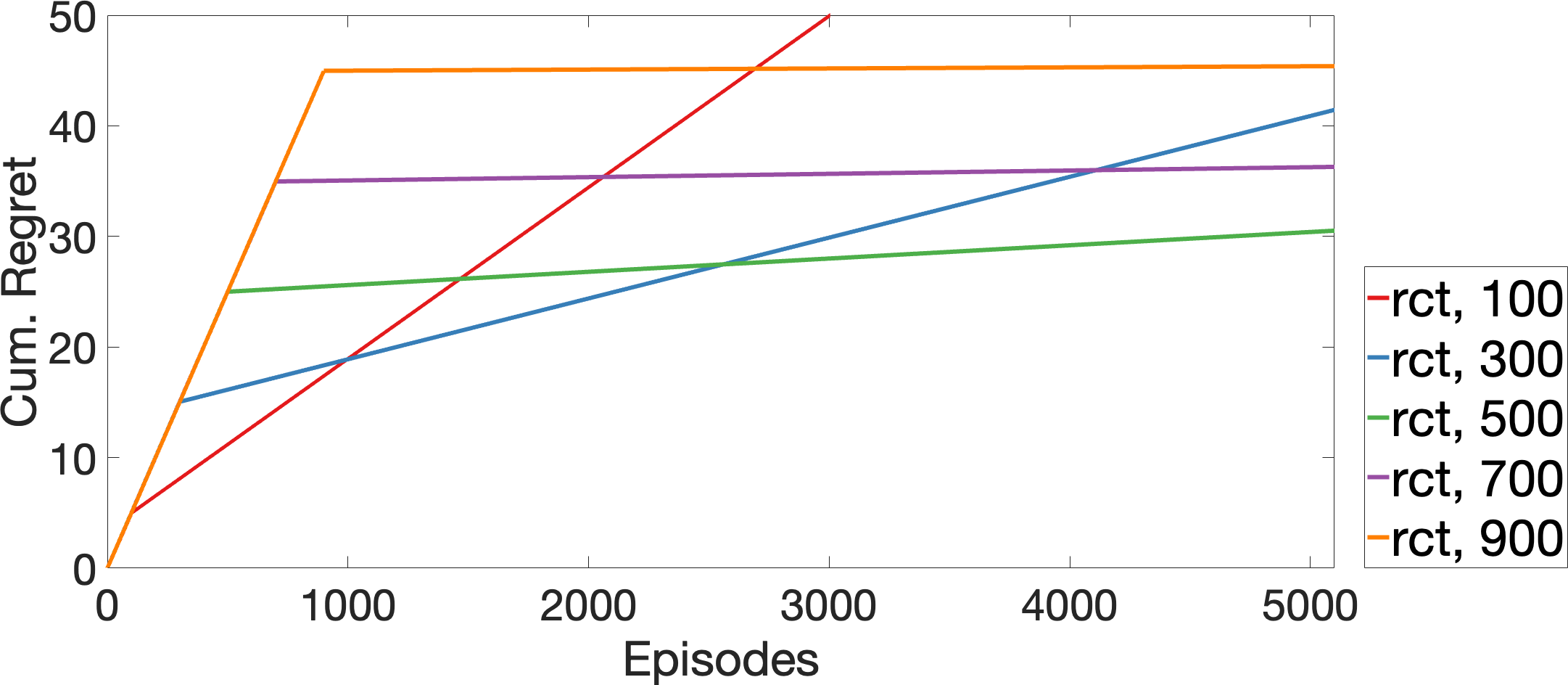}
		\caption{The regret of \texttt{RCT} with varying total number of trials.}
		\label{fig:_4_2_rct}
	\end{figure}

\begin{experiment}\label{exp:_4_2_mab2}
		Fig.~\ref{fig:_4_2_rct} shows the cumulative regrets of \texttt{RCT} when deploying in the MAB model $\langle \1M^*, \Set{\Tuple{X, \emptyset}}, Y \rangle$ described in Example~\ref{exp:_4_2_mab1} where the optimal arm choice $X \gets 0$ and the suboptimal gap $\Delta = 0.1$. The optimal arm choice is $X \gets 0$, as shown in the derivation in Example~\ref{exp:_2_2_mab3}.

Recall that the $T$ represents the total number of episodes that the \texttt{RCT} algorithm interacts with the environment, and the total number trials $N$ represents the amount of exploration it performs (out of $T$ episodes). We evaluate the \texttt{RCT} algorithm with the number of episodes set to $T = 5,000$ and the number of trials set to $N = 100, 300, 500, 700, 900$.
Each data point is the average of $1,000$ simulations, which makes the error bars invisible.
The simulations show that \text{RCT} with $N = 500$ performs the best among all strategies, which is close to the analytical result in Eq.~\ref{eq:_4_2_rct3}, setting $N \approx 878$. 
This means Eq.~\ref{eq:_4_2_rct3} provides a near-optimal choice of the total number of trials. \hfill $\blacksquare$
	\end{experiment}

\subsubsection{The Upper Confidence Bound Algorithm}
The upper confidence bound (UCB) algorithm is based on the principle of \emph{optimism in the face of uncertainty} (OFU, \citealt{auer2002finite}), which states that the agent should act as if the environment is as close to the best-case scenario as possible, given past observations. 
It offers several advantages over \texttt{RCT} introduced in the previous section, which we summarize below:
	\begin{itemize}[leftmargin=*]
		\item It does not depend on the prior knowledge of the parametrization of the underlying environment, i.e., the gap $\Delta$ in the expected rewards between an optimal and a suboptimal policy.
		\item It does not rely on prior knowledge of the total number of episodes $T$ that the online algorithm will intervene in the underlying environment.
		\item It achieves the same theoretical guarantees as \texttt{RCT} that fine-tunes the total number of trials $N$ based on prior knowledge of the suboptimal gap $\Delta$ and the total episodes $T$.
	\end{itemize}
The insight of the \texttt{UCB} algorithm is to evoke an \emph{adaptive randomization} strategy 	\citep{robbins52,lai1985asymptotically,berry1985bandit}, which means that the agent repeatedly adjusts the probability of action assignment according to the past assigned actions and observed outcomes during the experimentation. To make the argument more precise, for an MAB model $\langle \1M^*, \Set{\Tuple{X, \emptyset}}, Y \rangle$, \texttt{UCB} will utilize the prior actions and rewards' history to compute an \textbf{upper confidence bound} to each arm $x$, which is an overestimate of the unknown expected reward $\invE{Y}{x}$ with high probability.

	In order to better understand the construction of the confidence bound, we need to introduce some basic concentration results. Let $\Braces{Y^{(1)}, \dots, Y^{(n)}}$ be finite i.i.d. reward signals drawn from a discrete distribution $P(Y)$. Let empirical estimates be $\hat{\E}[Y] = \frac{1}{n} \sum_{t = 1}^n Y^{(i)}$. Applying Hoeffding's inequalities \citep{hoeffding1963probability} on the reward signal $Y$ bounded in a real interval $[0, 1]$ gives
	\begin{align}
		P \left ( \E[Y] \geq \hat{\E}[Y] + \sqrt{\frac{\log(1/\delta)}{2n}}\right) \leq \delta \;\;\; \text{for all }\delta \in (0, 1) \label{eq:_4_2_ucb1}
	\end{align}
	Fix a sequence of arm selections $x^{(1)}, \dots, x^{(t)} \in \D(X)$. Let $N_t(x) = \sum_{i = 1}^{t} \I\{X^{(t)} = x\}$ be the total occurrence of arm $x$ being played for every $x \in \D(X)$. Given finite samples $\Braces{Y^{(1)}, \dots, Y^{(t)}}$ drawn from interventional distributions $\inv{Y}{x^{(1)}}, \dots, \inv{Y}{x^{(t)}}$ respectively, it follows from Eq.~\ref{eq:_4_2_ucb1} that the upper confidence bound for the reward $\invE{Y}{x}$ is defined as \footnote{For consistency, we also define $\text{UCB}_t(x, \delta) = \infty$ if $N_t(x) = 0$.}
	\begin{align}
		\text{UCB}_t(x, \delta) = \underbrace{\hat{\E}^{(t)}_x[Y]}_{exploration} + \underbrace{\sqrt{\frac{\log(1/\delta)}{2N_t(x)}}}_{exploitation}  \label{eq:_4_2_ucb2}
	\end{align}
	Among quantities in the above equation, the second term is the confidence width computed from the concentration bound in Eq.~\ref{eq:_4_2_ucb1}. The first term is the empirical mean estimate $\hat{\E}^{(t)}_x[Y]$ for the expected reward and is given by
	\begin{align}
		\hat{\E}^{(t)}_x[Y] = \frac{1}{N_t(x)} \sum_{i = 1}^{t} Y^{(i)} \I \left \{X^{(i)} = x \right \}
	\end{align}
We summarize in Alg.~\ref{alg:_4_2_ucb} the details of the \texttt{UCB} algorithm when deployed in an unknown MAB model $\1M^*_{\textsc{mab}}$. 
For every episode $t$, it computes confidence bounds $\text{UCB}_{t-1}(x, \delta)$ for every arm $x \in \D(X)$ from prior interventional data $\Braces{Y^{(1)}, \dots, Y^{(t - 1)}}$. 
It then selects an arm $x^{(t)}$ with the maximal upper confidence bound, performs intervention $\doo\Parens{x^{(t)}}$, and receives subsequent reward $Y^{(t)}$. 
It can be shown that such an arm allocation strategy based on the upper confidence bound in Eq.~\ref{eq:_4_2_ucb2} balances the trade-off between exploration and exploitation. 
The algorithm is more likely to play an arm $x$ if it is (1) close to optimal since the empirical reward estimate $\hat{\E}^{(t)}_x[Y]$ is large, or (2) not sufficiently explored and $N_x(t)$ is small. 
At Step $3$, the error probability $\delta = t^{-4}$ decreases as the episode number $t$ increases. 
This means that the upper confidence bound estimates for every arm $x$ become increasingly accurate as the online learning process continues.

	\begin{algorithm}[t]
		\caption{Upper Confidence Bound (\texttt{UCB}) in MAB}
		\label{alg:_4_2_ucb}
		\setlength{\textfloatsep}{0pt}
		\begin{algorithmic}[1]
			\Require a policy scope $\Pi = \Braces{\Tuple{X,\emptyset}}$.
			\ForAll{episodes $t = 1, 2, \dots $}
			\State Choose an arm $x^{(t)} = \argmax_{x} \text{UCB}_{t-1}\left (x, \delta \right)$ where $\delta = t^{-4}$.
			\State Perform $\doo\Parens{x^{(t)}}$ for episode $t$ and receive reward $Y^{(t)}$.
			\EndFor
		\end{algorithmic}
	\end{algorithm}

\begin{theorem}[Regrets of \texttt{UCB} in MABs \citep{auer2002finite}]\label{thm:_4_2_ucb}
		For an MAB $\langle \1M^*, \Set{\Tuple{X, \emptyset}}, Y \rangle$, let $Y$ be the reward variable with support on $[0, 1]$, and let the domain of action $X$ be $\D(X) = \{1, \dots, K\}$. It holds the regret of \texttt{UCB} in SCM $\1M^*$ after $T > 1$ episodes is bounded by
		\begin{align}
			R(T, \1M^*) \leq  8 \sum_{x: \Delta_x > 0} \frac{\log(T)}{\Delta_x} + \left( 1 + \frac{\pi^2}{3} \right) \sum_{x: \Delta_x > 0} \Delta_x \label{eq:_4_2_ucb3}
		\end{align} \hfill $\blacksquare$
	\end{theorem}
After a few simplifications \citep{lattimore2020bandit}, the regret bound in Eq.~\ref{eq:_4_2_ucb3} could be further written as:
	\begin{align}
		R(T, \1M^*) & \leq 5\sum_{x: \Delta_x >0} \Delta_x + C \sqrt{KT\log(T)} \label{eq:_4_2_ucb4}
	\end{align}
where $C$ is a universal constant. In words, \texttt{UCB} achieves a sublinear regret $\mathcal{O}\left(T^{1/2}\log(T)^{1/2}\right)$, which is close to the regret bound obtained by \texttt{RCT} up to logarithmic terms. By employing sharper concentration inequalities for the expected reward estimates, it is possible to shave the dependence on the logarithmic term in the regret bound of \texttt{UCB} \citep{audibert2009minimax}. 
Broadly speaking, the theory supports the claim that \texttt{UCB} is able to overcome the limitation of \texttt{RCT} by removing the dependence on the prior knowledge of suboptimal gaps $\Delta$ and the total number of episodes $T$ while achieving the same asymptotic guarantees. Still, in practice, two algorithms' having similar regret bounds does not mean they will perform the same when deployed in the environment.
The reason is that the analysis might be loose for one algorithm and not to the other, or by a different margin. For this reason, we now compare the empirical performance of \texttt{UCB} and \texttt{RCT}.
\begin{experiment}\label{exp:_4_2_mab4}
Fig.~\ref{fig:_4_2_ucb} shows the cumulative regret of \texttt{UCB} when deploying in the MAB model described in Example~\ref{exp:_4_2_mab1}. The setup is the same as in Experiment~\ref{exp:_4_2_mab2}, which has $T = 10000$ and parameter $\alpha = 0.1$. As a baseline, we also include \texttt{RCT} with various choices for the total number of trials set to $N$. The simulation shows a common phenomenon -- if \texttt{RCT} is fine-tuned with the optimal choice of the trial number, it can outperform \texttt{UCB} by a small margin in the cumulative regret. However, if the trial number $N$ must be chosen without prior knowledge of the model parameter $\Delta$ and the total episodes of interventions $T$, \texttt{UCB} will generally dominate \texttt{RCT} in performance. \hfill $\blacksquare$
		\begin{figure}[t]
		\centering
		\includegraphics[width=0.7\textwidth]{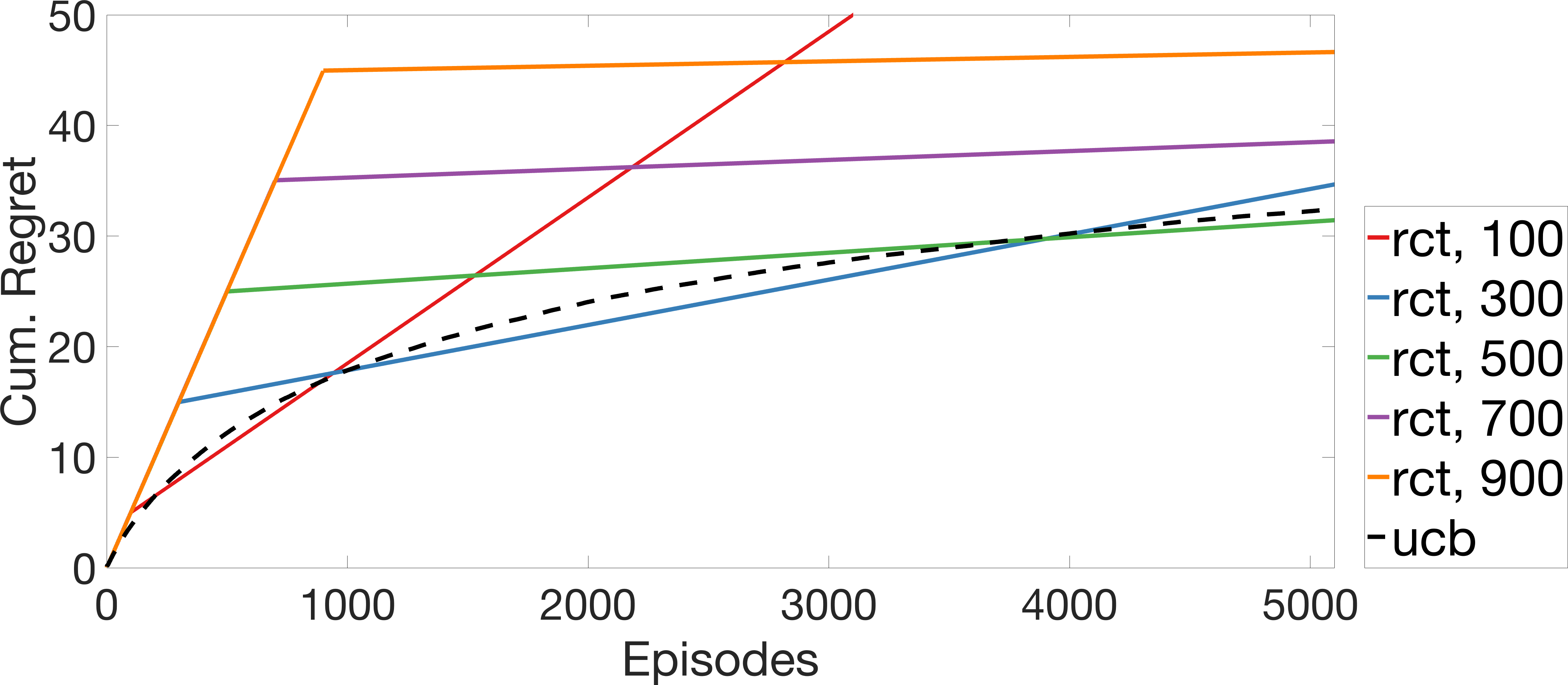}
		\caption{Simulation results comparing performance of online learning algorithms \texttt{UCB} and \texttt{RCT}.}
		\label{fig:_4_2_ucb}
	\end{figure}
\end{experiment}
The principle underpinning \texttt{UCB} has been applied to other RL environments and leads to sublinear regrets, including contextual bandit \citep{li2010contextual}, Markov decision processes \citep{auer2009near}, and factored MDPs \citep{osband2014near}, just to name a few. In Sec.~\ref{sec:_5_o2o}, we will also see a more generalized implementation of \texttt{UCB} that could find an optimal policy in an arbitrary CRL system.

\subsection{Causal Identification}\label{sec:_4_3}
Online learning algorithms ensure the NUC condition (Def.~\ref{def:_4_1_nuc}) holds by deploying candidate policies in the environment (Lem.~\ref{lem:_4_2_nuc}).
However, randomized experiments are not applicable in all settings; for example, the effect of a risk factor such as smoking cannot ethically be addressed with randomized controlled trials \citep{cornfield1959smoking}. 
Furthermore, performing randomized experiments is expensive, and may even be infeasible due to financial constraints. For instance, in a survey of phase trials of new drugs approved by the Food and Drug Administration (FDA) of the United States from 2015-2016 \citep{moore2018estimated}, the trial cost was estimated at a median of \$41,117 per patient. The increasing cost of certain trials calls for more generalized policy learning methods without direct interventions.
\footnote{The situation is more challenging in practice since even in settings where RCTs are applicable, there are still serious concerns due to issues of transportability (also known in the literature as external validity/generalizability). For further discussion, refer to \citep{bareinboim2016causal,correa2020general}.}

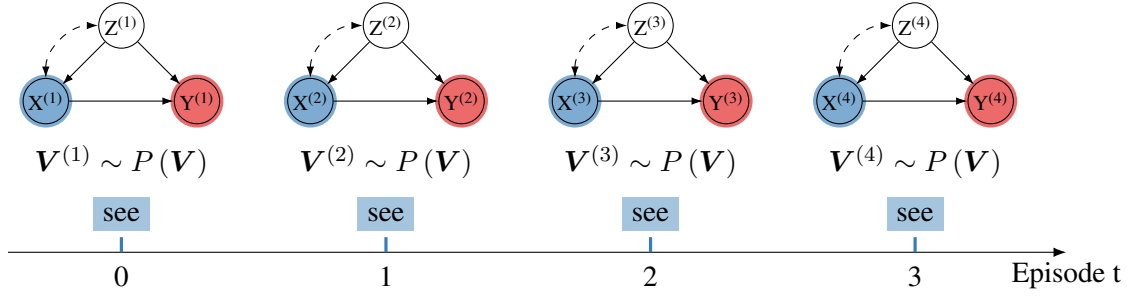
\begin{figure}[t]
	\centering
	\begin{tikzpicture}
		\def\outerr{3.5}
		\def\innerr{3.5}
		\def\dist{3.5}

		\draw[->, >={Latex}] (-1.5,0) -- (3*\dist+2,0) node[below] {Episode t};

		\node[vertex, minimum width=6mm] (X1) at (-1, 2) {X\textsuperscript{(1)}};
		\node[vertex, minimum width=6mm] (Y1) at (1, 2) {Y\textsuperscript{(1)}};
		\node[vertex, minimum width=6mm] (Z1) at (0, 3) {Z\textsuperscript{(1)}};
		\draw[bidir] (X1) to [bend left = 45] (Z1);
		\draw[dir] (X1) -- (Y1);
		\draw[dir] (Z1) -- (X1);
		\draw[dir] (Z1) -- (Y1);

		\node[vertex, minimum width=6mm] (X2) at (\dist-1, 2) {X\textsuperscript{(2)}};
		\node[vertex, minimum width=6mm] (Y2) at (\dist+1, 2) {Y\textsuperscript{(2)}};
		\node[vertex, minimum width=6mm] (Z2) at (\dist, 3) {Z\textsuperscript{(2)}};
		\draw[bidir] (X2) to [bend left = 45] (Z2);
		\draw[dir] (X2) -- (Y2);
		\draw[dir] (Z2) -- (X2);
		\draw[dir] (Z2) -- (Y2);

		\node[vertex, minimum width=6mm] (X3) at (2*\dist-1, 2) {X\textsuperscript{(3)}};
		\node[vertex, minimum width=6mm] (Y3) at (2*\dist+1, 2) {Y\textsuperscript{(3)}};
		\node[vertex, minimum width=6mm] (Z3) at (2*\dist, 3) {Z\textsuperscript{(3)}};
		\draw[bidir] (X3) to [bend left = 45] (Z3);
		\draw[dir] (X3) -- (Y3);
		\draw[dir] (Z3) -- (X3);
		\draw[dir] (Z3) -- (Y3);

		\node[vertex, minimum width=6mm] (X4) at (3*\dist-1, 2) {X\textsuperscript{(4)}};
		\node[vertex, minimum width=6mm] (Y4) at (3*\dist+1, 2) {Y\textsuperscript{(4)}};
		\node[vertex, minimum width=6mm] (Z4) at (3*\dist, 3) {Z\textsuperscript{(4)}};
		\draw[bidir] (X4) to [bend left = 45] (Z4);
		\draw[dir] (X4) -- (Y4);
		\draw[dir] (Z4) -- (X4);
		\draw[dir] (Z4) -- (Y4);

		\node (d1) at (0, 1.2) {$\*V^{(1)} \sim P\left(\*V\right)$};
		\node (d2) at (\dist, 1.2) {$\*V^{(2)} \sim P\left(\*V\right)$};
		\node (d3) at (2*\dist, 1.2) {$\*V^{(3)} \sim P\left(\*V\right)$};
		\node (d4) at (3*\dist, 1.2) {$\*V^{(4)} \sim P\left(\*V\right)$};

		\draw[very thick, betterblue, -] (0, 0) -- (0, 0.2);
		\draw[very thick, betterblue, -] (\dist, 0) -- (\dist, 0.2);
		\draw[very thick, betterblue, -] (2*\dist, 0) -- (2*\dist, 0.2);
		\draw[very thick, betterblue, -] (3*\dist, 0) -- (3*\dist, 0.2);

		\node [below] at (0, -0.05) { 0};
		\node [below] at (\dist, -0.05) { 1};
		\node [below] at (2*\dist, -0.05) { 2};
		\node [below] at (3*\dist, -0.05) { 3};

		\node [fill=betterblue!45] at (0, 0.5) {see};
		\node [fill=betterblue!45] at (\dist, 0.5) {see};
		\node [fill=betterblue!45] at (2*\dist, 0.5) {see};
		\node [fill=betterblue!45] at (3*\dist, 0.5) {see};

		\begin{pgfonlayer}{back}
			\node[circle,fill=betterblue!65,draw=none,minimum size=2*\innerr mm] at (X1) {};
			\node[circle,fill=betterred!65,draw=none,minimum size=2*\innerr mm] at (Y1) {};
			\node[circle,fill=betterblue!65,draw=none,minimum size=2*\innerr mm] at (X2) {};
			\node[circle,fill=betterred!65,draw=none,minimum size=2*\innerr mm] at (Y2) {};
			\node[circle,fill=betterblue!65,draw=none,minimum size=2*\innerr mm] at (X3) {};
			\node[circle,fill=betterred!65,draw=none,minimum size=2*\innerr mm] at (Y3) {};
			\node[circle,fill=betterblue!65,draw=none,minimum size=2*\innerr mm] at (X4) {};
			\node[circle,fill=betterred!65,draw=none,minimum size=2*\innerr mm] at (Y4) {};
		\end{pgfonlayer}
	\end{tikzpicture}
	\caption{Temporal diagram showing a causal identification agent interacting with the environment for repeated episodes.}
	\label{fig:_4_3_docalculus}
\end{figure}

An alternative approach for policy evaluation is to relax the NUC assumption and explore other more general structural assumptions in the underlying environment, represented as a causal diagram. This leads to the setting of causal identification, characterized by a signature:
\begin{align}
	\1T_{\text{id}} = \Tuple{\1R = \text{see}, \1A = \G, \Pi = \Braces{\Tuple{X_i, \*S_i}}_{i = 1}^H, \1R = \D(\*Y) \mapsto \3R}
\end{align}
The optimization in this task goes as follows: 
\begin{align}\label{eq:opt-crl-off-do-calc}
	 & \pi^* = \argmax_{\pi \in \Pi} \invEE{ \1R \left (\*Y \right ) \; \big \vert \textcolor{red}{ \1A = \G, \;\mathcal{D}_{\text{obs}} \sim P(\*V)}}{\pi}{\1M^*}, 
\end{align}
Similar to off-policy learning previously discussed in Sec.~\ref{sec:_4_1}, a causal identification agent collected data from the underlying SCM $\1M^*$ through repeated episodes of passive observations ($\1I = \text{see}$). or every episode $t$, the agent observes the SCM $\1M^*$, and receives a sample drawn from the observational distribution $P(\*V)$. The key difference is that the agent does not assume the NUC condition (Def.~\ref{def:_4_1_nuc}). Instead, it now has access to a causal diagram $\G$ associated with the underlying environment $\1M^*$ (Def.~\ref{def:_2_3_diagram}).\footnote{There exist causal discovery algorithms to learn (an equivalence class of) causal diagrams from observational  \citep{spirtes2000causation,pearl:2k} and 
experimental data \citep{kocaoglu2017experimental,kocaoglu2019characterization,jaber2020cd}. }
\footnote{This settings can be generalized to consider input distributions in which the other agent is known to have collected data under a randomized regime, albeit of another action variable different than $\*X$, say $\*Z$. This problem has been studied in the literature under the rubric of g-identification \citep{bareinboim:pea12,lee2019gid}. }
It will then incorporate experts' domain knowledge to obtain a causal diagram compatible with the underlying environment. See Sec.~\ref{sec:_2_3_diagram} for more discussion. This interaction is illustrated in the temporal diagram in Fig.~\ref{fig:_4_3_docalculus}.
 
The agent aims to learn an optimal policy $\pi^* \in \Pi$ from the combination of the observational data $P(\*V)$ and structural assumptions encoded in the causal diagram $\G$. A key challenge is to find a function of the observational distribution $P(\*V)$ that is guaranteed to be equal to the probability query of interest in the intervened submodel $\1M_{\pi}$, for any SCM $\1M$ compatible with structural assumptions encoded in the diagram $\G$. We introduce next this notion articulated more formally. 
\begin{definition}[Identifiability]\label{def:_4_3_id}
	Let subsets of variables $\*X, \*Y \subseteq \*V$ and $\pi$ be a policy over $\*X$.
	The interventional distribution $\inv{\*Y}{\pi}$ is identifiable from the structural assumptions $\1A$ if $\inv{\*Y}{\pi}$ is uniquely computable from any positive observational distribution $P(\*V)$ in any SCM $\1M$ satisfying $\1A$. That is, if for every pair of SCMs $\1M_1$ and $\1M_2$ compatible with structural assumptions $\1A$, $P_{\pi}(\*Y;\1M_1) = P_{\pi}(\*Y;\1M_2)$ whenever $P(\*V; \1M_1) = P(\*V; \1M_2) > 0$.
\end{definition}
For a causal identification task, the structural assumptions $\1A$ will be encoded as a causal diagram $\G$. We say an interventional distribution $\inv{\*Y}{\pi}$ is identifiable from a causal diagram $\G$ if for any pair of SCMs $\1M_1, \1M_2$, $P_{\pi}(\*Y;\1M_1) = P_{\pi}(\*Y;\1M_2)$ whenever $P(\*V; \1M_1) = P(\*V; \1M_2) > 0$ and $\G(\1M_1) = \G(\1M_2) = \G$. On the other hand, as for an off-policy learning task described in Sec.~\ref{sec:_4_1}, the structural assumptions $\1A$ are specified using the NUC condition (Def.~\ref{def:_4_1_nuc}), which restricts the form of the behavior policy $f_{\*X}$ determining values of actions $\*X$ in the underlying environment. It follows as a corollary from Thms.~\ref{thm:_4_1_ipw} and \ref{thm:_4_1_dp} that for any policy $\pi \in \Pi$, the expected reward $\invE{\1R(\*Y)}{\pi}$ is identifiable provided with the NUC condition (Def.~\ref{def:_4_1_nuc}).
\begin{corollary}\label{corol_4_3_nuc_id}
	For a policy space $\Pi = \Braces{\Tuple{X_i, \*S_i}}_{i = 1}^H$ and reward function $\1R: \D(\*Y) \mapsto \3R$, let a policy $\pi \in \Pi$. $\inv{\*Y}{\pi}$ is identifiable from the NUC condition (Def.~\ref{def:_4_1_nuc}) w.r.t. the policy space $\Pi$.
\end{corollary}
The following example shows the identifiability guarantee from the NUC assumption.
\begin{example}[Identification under NUC]\label{exp:_4_3_dtr1}
	We consider a 2-stage DTR $\Tuple{\1M^*, \Pi, Y}$ described in Fig.~\ref{fig:_3_1_dtr} where the policy space $\Pi = \Braces{\Tuple{X_1, \Braces{S_1}}, \Tuple{X_2, \Braces{S_1, X_1, S_2}}}$. We further assume that the NUC condition (Def.~\ref{def:_4_1_nuc}) holds, i.e., the unobserved confounder $U$ does not affect actions $X_1, X_2$. This means that for every action $X_1$ (or $X_2$), when its direct parents $S_1$ (or $S_1, X_1, S_2$) are observed and measured, its values are only affected by independent noise.

It follows from IPW estimation of Thm.~\ref{thm:_4_1_ipw} that the expected reward of any policy $\pi \in \Pi$ is computable from the observational distribution and policy's definition, and is given by:
	\begin{align}
		\invE{Y}{\pi} = \sum_{s_1, x_1, s_2, s_2} \underbrace{\frac{\E\left[Y \mid s_1, x_1, s_2, s_2 \right] P(s_1, x_1, s_2, s_2)}{P(x_1|s_1) P(x_2|s_1, x_1, s_2)}}_{\text{observational distribution}}
		\underbrace{\pi_1(x_1|s_1)\pi_1(x_2|s_1, x_1, s_2)}_{\text{policy $\pi$}}.
	\end{align}
The above equation is a function of the candidate policy $\pi$ (the second term) and the observational distribution $P(S_1, X_1, S_2, X_2, Y)$ (the first term). 
Formally, the expected reward $\invE{Y}{\pi}$ is identifiable from $P(\*V)$ provided with the NUC condition, i.e., no matter the specific form of the mechanisms $\*F$ and exogenous conditions $P(\*u)$ of the underlying, true SCM $\M^*$. We also compute the value function of policy $\pi \in \Pi$ following the DP estimation in Thm.~\ref{thm:_4_1_dp}:
	\begin{align}
		Q^{(1)}_{\pi}(x_1, s_1)           & = \sum_{s_2, x_2}  Q^{(2)}_{\pi}(x_1, x_2, s_1, s_2) P(s_2|s_1, x_1)\pi_2(x_2 \mid s_1, x_1, s_2) \\
		Q^{(2)}_{\pi}(x_1, x_2, s_1, s_2) & = \E[Y|s_1, x_1, s_2, s_2].
	\end{align}
	Finally, the expected reward is identifiable by
	\begin{align}
		\invE{Y}{\pi} = \sum_{s_1, x_1}Q^{(1)}_{\pi}(x_1, s_1)\pi_1(x_1\mid s_1) P(s_1),
	\end{align}
	and value functions $Q^{(1)}_{\pi}, Q^{(2)}_{\pi}$ are both computable from distribution $P(S_1, X_1, S_2, X_2, Y)$. \hfill $\blacksquare$
\end{example}

\subsubsection{Sequential Backdoor for Policy Evaluation}\label{sec:_4_3_1}
As previously discussed in Sec.~\ref{sec:_4_1}, both IPW (Thm.~\ref{thm:_4_1_ipw}) and DP (Thm.~\ref{thm:_4_1_dp}) are popular off-policy evaluation algorithms in causal inference and reinforcement learning literature.
Therefore, it is worth understanding the conditions under which these algorithms are applicable for identifying the expected rewards of policies in a policy space $\Pi$, provided with an arbitrary causal diagram $\G$. 
There exists a graphical condition called he \textit{sequential backdoor} \citep{pearl:rob95} that delimits whether the effect of performing a sequence of atomic interventions can be identified by covariates adjustments.
In this section, we generalize the sequential backdoor criterion to evaluate the effects of sequential policies (i.e., not necessarily atomic\footnote{For a more nuanced and detailed discussion on the difference between atomic and non-atomic interventions, please refer to \cite[Appendix~B]{correa2020calculus}.}), which select actions based on values of other observed covariates in the system.

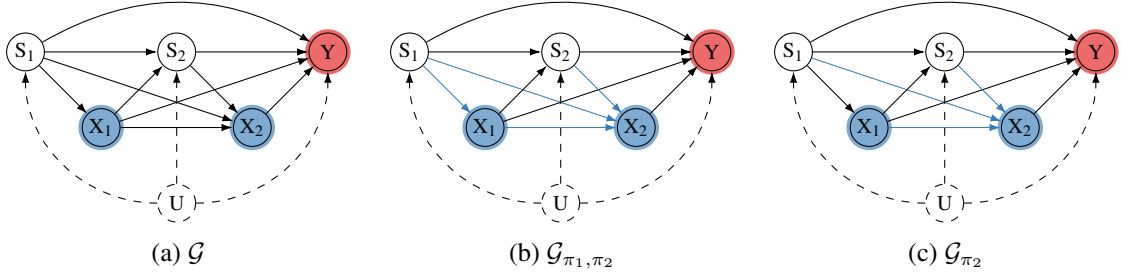
\begin{figure}[t]
	\begin{subfigure}{0.32\linewidth}\centering
		\begin{tikzpicture}
			\def\outerr{3.5}
			\def\innerr{3}
			\node[vertex] (Z1) at (0, 1) {S\textsubscript{1}};
			\node[vertex] (X1) at (1, 0) {X\textsubscript{1}};
			\node[vertex] (Z2) at (2, 1) {S\textsubscript{2}};
			\node[vertex] (X2) at (3, 0) {X\textsubscript{2}};
			\node[vertex] (Y) at (4, 1) {Y};
			\node[uvertex] (U) at (2, -1) {U};

			\draw[dir] (Z1) to (Z2);
			\draw[dir] (Z1) to (X1);
			\draw[dir] (Z1) to (X2);
			\draw[dir] (Z1) to [bend left = 30] (Y);

			\draw[dir] (X1) to (Z2);
			\draw[dir] (X1) to (X2);
			\draw[dir] (X1) to (Y);

			\draw[dir] (Z2) to (Y);
			\draw[dir] (Z2) to (X2);

			\draw[dir] (X2) to (Y);

			\draw[dir, dashed]  (U) to (Z2);
			\draw[dir, dashed]  (U) to [bend left = 45] (Z1);
			\draw[dir, dashed]  (U) to [bend right = 45] (Y);

			\begin{pgfonlayer}{back}
				\node[circle,fill=betterblue!65,draw=none,minimum size=2*\innerr mm] at (X1) {};
				\node[circle,fill=betterblue!65,draw=none,minimum size=2*\innerr mm] at (X2) {};
				\node[circle,fill=betterred!65,draw=none,minimum size=2*\innerr mm] at (Y) {};
			\end{pgfonlayer}
		\end{tikzpicture}
		\caption{$\G$}
		\label{fig:_4_3_dtr1a}
	\end{subfigure}\hfill
	\begin{subfigure}{0.32\linewidth}\centering
		\begin{tikzpicture}
			\def\outerr{3.5}
			\def\innerr{3}
			\node[vertex] (Z1) at (0, 1) {S\textsubscript{1}};
			\node[vertex] (X1) at (1, 0) {X\textsubscript{1}};
			\node[vertex] (Z2) at (2, 1) {S\textsubscript{2}};
			\node[vertex] (X2) at (3, 0) {X\textsubscript{2}};
			\node[vertex] (Y) at (4, 1) {Y};
			\node[uvertex] (U) at (2, -1) {U};

			\draw[dir] (Z1) to (Z2);
			\draw[dir, betterblue] (Z1) to (X1);
			\draw[dir, betterblue] (Z1) to (X2);
			\draw[dir] (Z1) to [bend left = 30] (Y);

			\draw[dir] (X1) to (Z2);
			\draw[dir, betterblue] (X1) to (X2);
			\draw[dir] (X1) to (Y);

			\draw[dir] (Z2) to (Y);
			\draw[dir, betterblue] (Z2) to (X2);

			\draw[dir] (X2) to (Y);

			\draw[dir, dashed]  (U) to (Z2);
			\draw[dir, dashed]  (U) to [bend left = 45] (Z1);
			\draw[dir, dashed]  (U) to [bend right = 45] (Y);
			\begin{pgfonlayer}{back}
				\node[circle,fill=betterblue!65,draw=none,minimum size=2*\innerr mm] at (X1) {};
				\node[circle,fill=betterblue!65,draw=none,minimum size=2*\innerr mm] at (X2) {};
				\node[circle,fill=betterred!65,draw=none,minimum size=2*\innerr mm] at (Y) {};
			\end{pgfonlayer}
		\end{tikzpicture}
		\caption{$\G_{\pi_1, \pi_2}$}
		\label{fig:_4_3_dtr1b}
	\end{subfigure}\hfill
	\begin{subfigure}{0.32\linewidth}\centering
		\begin{tikzpicture}
			\def\outerr{3.5}
			\def\innerr{3}
			\node[vertex] (Z1) at (0, 1) {S\textsubscript{1}};
			\node[vertex] (X1) at (1, 0) {X\textsubscript{1}};
			\node[vertex] (Z2) at (2, 1) {S\textsubscript{2}};
			\node[vertex] (X2) at (3, 0) {X\textsubscript{2}};
			\node[vertex] (Y) at (4, 1) {Y};
			\node[uvertex] (U) at (2, -1) {U};

			\draw[dir] (Z1) to (Z2);
			\draw[dir] (Z1) to (X1);
			\draw[dir, betterblue] (Z1) to (X2);
			\draw[dir] (Z1) to [bend left = 30] (Y);

			\draw[dir] (X1) to (Z2);
			\draw[dir, betterblue] (X1) to (X2);
			\draw[dir] (X1) to (Y);

			\draw[dir] (Z2) to (Y);
			\draw[dir, betterblue] (Z2) to (X2);

			\draw[dir] (X2) to (Y);

			\draw[dir, dashed]  (U) to (Z2);
			\draw[dir, dashed]  (U) to [bend left = 45] (Z1);
			\draw[dir, dashed]  (U) to [bend right = 45] (Y);
			\begin{pgfonlayer}{back}
				\node[circle,fill=betterblue!65,draw=none,minimum size=2*\innerr mm] at (X1) {};
				\node[circle,fill=betterblue!65,draw=none,minimum size=2*\innerr mm] at (X2) {};
				\node[circle,fill=betterred!65,draw=none,minimum size=2*\innerr mm] at (Y) {};
			\end{pgfonlayer}
		\end{tikzpicture}
		\caption{$\G_{\pi_2}$}
		\label{fig:_4_3_dtr1c}
	\end{subfigure}\hfill\null
	\caption{Causal diagram satisfying the NUC condition and its manipulated diagrams.}
	\label{fig:_4_3_dtr1}
\end{figure}

Before describing the details of the criterion, we first introduce some necessary notations. For every policy $\pi \in \Pi$, let $\G_{\pi}$ denote the causal diagram associated with the submodel $\1M^*_{\pi}$ induced by policy intervention $\doo(\pi)$. Operationally, the manipulated diagram $\G_{\pi}$ is obtained from $\G$ by performing the following procedures: for every $i = 1, \dots, H$,
\begin{enumerate}
	\item Remove all incoming arrows pointing into action node $X_i$;
	\item Add arrows from nodes in input states $\*S_i$ to the action node $X_i$.
\end{enumerate}
For a policy $\pi = \Parens{\pi_1, \dots, \pi_H}$, let $\Parens{\pi_i, \dots, \pi_j}$ be a sub-policy consisting of decision rules with restriction to indices $1 \leq i < j \leq H$. The manipulated graph $\G_{\pi_i, \dots, \pi_j}$ is thus a causal diagram obtained by intervention $\doo\Parens{\pi_i, \dots, \pi_j}$ following the sub-policy $\Parens{\pi_i, \dots, \pi_j}$.\footnote{Note that for $i = 1, \dots, H$, covariates $\*S_i$ are non-descendent of actions $\bar{\*X}_{i: H}$. It is verifiable that for every policy $\pi \in \Pi$, the manipulated graph $\G_{\pi_{i+1}, \dots, \pi_H}$, $i = 1 \dots, H$, is acyclic and preserves the topological ordering $\*S_1 \prec X_1 \prec \dots \prec \*S_{H} \prec X_H$.} As an example, consider the causal diagram $\G$ described Fig.~\ref{fig:_4_3_dtr1a}. For a policy $\Parens{\pi_1\Parens{X_1 \mid S_1}, \pi_2\Parens{X_2 \mid S_1, X_1, S_2}}$, Fig.~\ref{fig:_4_3_dtr1b} shows the manipulated causal diagram $\G_{\pi_1, \pi_2}$ induced by intervention $\doo(\pi_1, \pi_2)$; the added incoming arrows to actions $X_1, X_2$ are highlighted in blue. The diagram $\G_{\pi_2}$ induced by intervention $\doo(\pi_2)$ following a sub-policy with restriction on action $X_2$ is shown in Fig.~\ref{fig:_4_3_dtr1c}. Formally, the sequential backdoor condition for identifying policy interventions is defined as follows.
\begin{definition}[Sequential Backdoor Condition (Policy Intervention)]\label{def:_4_3_sbc}
	Let $\G$ be causal diagram and $\*Y \subseteq \*V$ be reward signals. A policy space $\Pi = \Braces{\Tuple{X_i, \*S_i}}_{i = 1}^H$ is said to satisfy the \emph{sequential backdoor condition} w.r.t. $\*Y$ in $\G$ (for short, $\Pi$ is backdoor admissible) if for every policy $\pi \in \Pi$, the following condition hold: for every $i = 1, \dots, H$,
	\begin{align}
		\left(X_i \ci \*Y \mid \bar{\*X}_{i-1}, \bar{\*S}_{i} \right)_{\G_{\underline{X_i},\pi_{i+1}, \dots, \pi_H}} \label{eq:_4_3_sbc}
	\end{align}
	That is, conditioning on nodes $\bar{\*X}_{i-1} \cup \bar{\*S}_{i}$ d-separates all \emph{backdoor paths} from action $X_i$ to reward signals $\*Y$ that contains an arrow pointing into $X_i$ in the manipulated graph $\G_{\pi_{i+1}, \dots, \pi_H}$.
\end{definition}
In spirit, the independence relationship in Eq.~\ref{eq:_4_3_sbc} is similar to the celebrated backdoor criterion \cite[Def.~3.3.1]{pearl:2k}. A backdoor path between nodes $\*X$ and $\*Y$ is a sequence of edges starting with an arrow pointing into a node in $\*X$. This criterion ensures that at every stage $i = 1, \dots, H$, the actions and covariates' history $\bar{\*X}_{i-1}, \bar{\*S}_{i}$ effectively summarize all the information that the behavior policy uses to determine values of action $X_i$, which are also relevant to the reward signal $Y$.
In other words, all the confounders between action $X_i$ and reward $Y$ are measured, and no other variables could generate non-causal correlations between $X_i$ and $Y$.

As a special case, let $\pi = \Parens{\pi_1, \dots, \pi_H}$ be an atomic policy such that every decision rule $\pi_i \triangleq X_i \gets x_i$ for every step $i = 1, \dots, H$. In this case, the manipulated graph $\G_{\pi_{i+1}, \dots, \pi_H}$, $i = 1, \dots, H$, is a subgraph obtained from $\G$ by removing the incoming arrows of every action node $X_{i+1}, \dots, X_H$. The independence relationship in Eq.~\ref{eq:_4_3_sbc} reduces to the sequential backdoor condition for atomic interventions $\doo(\*x)$ in \citep{pearl:rob95} and could be written as:
\begin{align}
	\left(X_i \ci \*Y \mid \bar{\*X}_{i-1}, \bar{\*S}_{i} \right)_{\G_{\underline{X_i},\overline{X_{i+1}, \dots, X_H}}}. \label{eq:_4_3_sbc_do}
\end{align}
Note that in the above equation, $\G_{\underline{X_i},\overline{X_{i+1}, \dots, X_H}}$ is a subgraph contained in the manipulated diagram $\G_{\underline{X_i},\pi_{i+1}, \dots, \pi_H}$ (w.r.t. an arbitrary policy $\pi$ over actions $\*X$). This means that the independence condition in Eq.~\ref{eq:_4_3_sbc} is stronger than the one in Eq.~\ref{eq:_4_3_sbc_do}. The atomic backdoor condition in \citep{pearl:rob95} holds whenever the policy space $\Pi$ is backdoor admissible in diagram $\G$.

Whenever the NUC condition (Def.~\ref{def:_4_1_nuc}) holds, note that the variables $\bar{\*X}_{i-1}, \bar{\*S}_{i}$ contain all direct parents of action $X_i$. There is no unobserved confounder affecting $X_i$ and any other variable in the environment.
Conditioning on the actions and states' history $\bar{\*X}_{i-1}, \bar{\*S}_{i}$ thus ``blocks'' all backdoor path from action node $X_i$ to reward nodes $\*Y$. As a corollary, it follows immediately that Def.~\ref{def:_4_3_sbc} subsumes the NUC condition (Def.~\ref{def:_4_1_nuc}).
\begin{corollary}\label{corol_4_3_nuc}
	For a CDM $\Tuple{\1M^*, \Pi, \1R}$ where $\Pi = \left \{\langle X_i, \*S_i \rangle \right \}_{X_i \in \*X}$ and $\1R: \D(\*Y) \mapsto \3R$, let $\G$ be the causal diagram associated with SCM $\1M^*$. If the NUC holds in $\Tuple{\1M^*, \Pi, \1R}$, then $\Pi$ is backdoor admissible w.r.t reward signals $\*Y$ in diagram $\G$.
\end{corollary}
Whenever the NUC condition holds in a CDM $\langle\1M^*, \Pi, \1R \rangle$, the above corollary implies that the policy space $\Pi$ must satisfy the sequential backdoor with regard to reward signals $\*Y$ in the causal diagram $\G$ of the underlying SCM $\1M^*$. The following example demonstrates this intuition.

\begin{example}[NUC $\Rightarrow$ Sequential Backdoor]\label{exp:_4_3_sbc1}
	Consider a 2-stage DTR model $\Tuple{\1M^*, \Pi, Y}$ consisting of actions $X_1, X_2$, observed states $S_1, S_2$, and a primary outcome $Y$; the policy space $\Pi = \Braces{\Tuple{X_1, \Braces{S_1}}, \Tuple{X_2, \Braces{S_1, X_1, S_2}}}$. Assume that the NUC condition holds (Def.~\ref{def:_4_1_nuc}), which means that the unobserved confounder $U$ does not affect actions $X_1, X_2$. Fig.~\ref{fig:_4_3_dtr1a} shows a more detailed causal diagram $\G$ associated with the environment $\1M^*$; the incoming arrows $U \rightarrow X_1$, $U \rightarrow X_2$ are now removed.

	We will next show that the policy space $\Pi$ satisfies sequential backdoor condition with regard to the primary outcome $Y$ in the causal diagram $\G$. Consider first the action $X_1$. For every policy $\Parens{\pi_1, \pi_2} \in \Pi$, the manipulated diagram $\G_{\pi_2}$ is shown in Fig.~\ref{fig:_4_3_dtr1c}. One could see by inspection that conditioning on covariate $S_1$ $d$-separates all backdoor paths from $X_1$ to $Y$ in $\G_{\pi_2}$, i.e.,
	\begin{align}
		\Parens{X_1 \ci Y \mid S_1}_{\G_{\underline{X_1},\pi_2}}
	\end{align}
	We also examine the independence relationship of Eq.~\ref{eq:_4_3_sbc} with regard to $X_2$, which is the last action in the decision sequence. It is thus sufficient to consider the causal diagram $\G$ of Fig.~\ref{fig:_4_3_dtr1a}. Again, conditioning input covariates $S_1, X_1, S_2$ $d$-separates backdoor paths from $X_2$ to $Y$ in $\G$, i.e.,
	\begin{align}
		\Parens{X_2 \ci Y \mid S_1, X_1, S_2}_{\G_{\underline{X_2}}}
	\end{align}
	We thus conclude the policy space $\Pi$ is backdoor-admissible w.r.t. the primary outcome $Y$ in $\G$. \hfill $\blacksquare$
\end{example}
Interestingly, Def.~\ref{def:_4_3_sbc} covers more general conditions than the NUC. There are CDMs $\langle \1M^*, \Pi, \1R \rangle$ where the NUC does not hold while the policy space $\Pi$ satisfies the sequential backdoor condition in the causal diagram $\G$ associated with the SCM $\1M^* $.
\begin{example}[Sequential Backdoor $\not \Rightarrow$ NUC] \label{exp:_4_3_sbc2}
	More formally, consider an SCM
	\begin{align}
		\M^* =
		\Tuple{\*U=\{U_i \}_{i = 1}^5, \*V=\{S_1, S_2, X_1, X_2, Y\}, \2F, P(\*U)}
	\end{align}
	consisting of actions $X_1, X_2$ and a reward signal $Y$. The causal mechanisms $\2F$ are defined as:
	\begin{align}\label{eq:_4_3_sbc1}
		\2F = \begin{cases}
			      S_1 \gets U_1,                       \\
			      X_1 \gets U_1 \oplus U_4,            \\
			      S_2 \gets S_1 \oplus X_1 \oplus U_2, \\
			      X_2 \gets U_2 \oplus U_5,            \\
			      Y \gets S_1 \oplus X_1 \oplus U_3,
		      \end{cases}
	\end{align}
	The exogenous distribution $P(\*U)$ is defined such that $U_1, U_2, U_3$ are independent binary variables following distribution $P(U_i = 0) = 0.9$, $i = 1, \dots, 3$; also, $U_4, U_5$ are independent noise uniformly drawn over $\{0, 1\}$. Fig.~\ref{fig:_4_3_sbc1a} shows the causal diagram $\G$ associated with the SCM $\1M^*$.
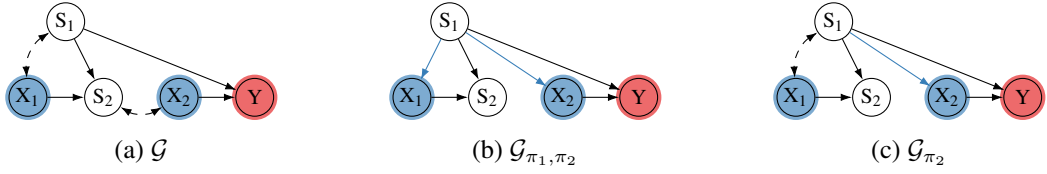
\begin{figure}
	\begin{subfigure}{0.32\linewidth}\centering
		\begin{tikzpicture}
			\def\outerr{3.5}
			\def\innerr{3}
			\node[vertex] (Z1) at (0.5, 1) {S\textsubscript{1}};
			\node[vertex] (X1) at (0, 0) {X\textsubscript{1}};
			\node[vertex] (Z2) at (1, 0) {S\textsubscript{2}};
			\node[vertex] (X2) at (2, 0) {X\textsubscript{2}};
			\node[vertex] (Y) at (3, 0) {Y};

			\draw[dir] (Z1) to (Y);

			\draw[dir] (X1) to (Z2);

			\draw[bidir] (Z2) to [bend right = 30] (X2);

			\draw[dir] (X2) to (Y);

			\draw[dir] (Z1) to (Z2);
			\draw[bidir]  (Z1) to [bend right = 30] (X1);

			\begin{pgfonlayer}{back}
				\node[circle,fill=betterblue!65,draw=none,minimum size=2*\innerr mm] at (X1) {};
				\node[circle,fill=betterblue!65,draw=none,minimum size=2*\innerr mm] at (X2) {};
				\node[circle,fill=betterred!65,draw=none,minimum size=2*\innerr mm] at (Y) {};
			\end{pgfonlayer}
		\end{tikzpicture}
		\caption{$\G$}
		\label{fig:_4_3_sbc1a}
	\end{subfigure}\hfill
	\begin{subfigure}{0.32\linewidth}\centering
		\begin{tikzpicture}
			\def\outerr{3.5}
			\def\innerr{3}
			\node[vertex] (Z1) at (0.5, 1) {S\textsubscript{1}};
			\node[vertex] (X1) at (0, 0) {X\textsubscript{1}};
			\node[vertex] (Z2) at (1, 0) {S\textsubscript{2}};
			\node[vertex] (X2) at (2, 0) {X\textsubscript{2}};
			\node[vertex] (Y) at (3, 0) {Y};

			\draw[dir] (Z1) to (Y);

			\draw[dir] (X1) to (Z2);

			\draw[dir] (X2) to (Y);

			\draw[dir] (Z1) to (Z2);
			\draw[dir, betterblue]  (Z1) to (X1);
			\draw[dir, betterblue]  (Z1) to (X2);

			\begin{pgfonlayer}{back}
				\node[circle,fill=betterblue!65,draw=none,minimum size=2*\innerr mm] at (X1) {};
				\node[circle,fill=betterblue!65,draw=none,minimum size=2*\innerr mm] at (X2) {};
				\node[circle,fill=betterred!65,draw=none,minimum size=2*\innerr mm] at (Y) {};
			\end{pgfonlayer}
		\end{tikzpicture}
		\caption{$\G_{\pi_1, \pi_2}$}
		\label{fig:_4_3_sbc1b}
	\end{subfigure}\hfill
	\begin{subfigure}{0.32\linewidth}\centering
		\begin{tikzpicture}
			\def\outerr{3.5}
			\def\innerr{3}
			\node[vertex] (Z1) at (0.5, 1) {S\textsubscript{1}};
			\node[vertex] (X1) at (0, 0) {X\textsubscript{1}};
			\node[vertex] (Z2) at (1, 0) {S\textsubscript{2}};
			\node[vertex] (X2) at (2, 0) {X\textsubscript{2}};
			\node[vertex] (Y) at (3, 0) {Y};

			\draw[dir] (Z1) to (Y);

			\draw[dir] (X1) to (Z2);

			\draw[dir, betterblue] (Z1) to (X2);

			\draw[dir] (X2) to (Y);

			\draw[dir] (Z1) to (Z2);
			\draw[bidir]  (Z1) to [bend right = 30] (X1);
			\begin{pgfonlayer}{back}
				\node[circle,fill=betterblue!65,draw=none,minimum size=2*\innerr mm] at (X1) {};
				\node[circle,fill=betterblue!65,draw=none,minimum size=2*\innerr mm] at (X2) {};
				\node[circle,fill=betterred!65,draw=none,minimum size=2*\innerr mm] at (Y) {};
			\end{pgfonlayer}
		\end{tikzpicture}
		\caption{$\G_{\pi_2}$}
		\label{fig:_4_3_sbc1c}
	\end{subfigure}\hfill\null
	\caption{Causal diagram satisfying the sequential backdoor and its manipulated diagrams.}
	\label{fig:_4_3_sbc1}
\end{figure}

	Consider a policy space $\Pi = \left \{ \langle X_1, \{S_1\}\rangle, \langle X_2, \{S_1\} \rangle \right\}$. One could see by inspection that the NUC condition does not hold in the CRL system $\langle \1M^*, \Pi, Y \rangle$ due to the presence of unobserved confounders $U_i$, $i = 1, 2$, affecting action ($X_i$) and state ($S_i$) variables, simultaneously. We next examine the scope $\Pi$ and see if it satisfies the sequential backdoor criterion in the diagram $\G$. For a policy $\Parens{\pi_1, \pi_2} \in \Pi$, the manipulated diagram $\G_{\pi_2}$ is shown in Fig.~\ref{fig:_4_3_sbc1c}. Conditioning on the covariate $S_1$ $d$-separates the backdoor path $X_1 \dashleftarrow \dashrightarrow  S_1 \rightarrow Y$ in $\G_{\pi_2}$, i.e.,
	\begin{align}
		\Parens{X_1 \ci Y \mid S_1}_{\G_{\underline{X_1},\pi_2}}
	\end{align}
	Similarly, conditioning on $S_1$ also blocks the backdoor path $X_2 \dashleftarrow \dashrightarrow S_2 \leftarrow X_1 \dashleftarrow \dashrightarrow S_1 \rightarrow Y$ between action $X_2$ and reward $Y$, i.e.,
	\begin{align}
		\Parens{X_2 \ci Y \mid S_1}_{\G_{\underline{X_2}}}
	\end{align}
	The above independence relationships imply $\Pi$ satisfies the sequential backdoor w.r.t. the reward $Y$ in the diagram $\G$ even when the NUC does not hold in the decision model $\langle \1M^*, \Pi, Y \rangle$. \hfill $\blacksquare$
\end{example}
One important observation here is that the NUC condition is hard to ascertain unless implied by the (physical) randomization procedure or assumptions following the model, such as the ones required by the sequential back-door condition.
This means that despite the NUC condition being popular throughout the literature, the same is not a primitive but a byproduct of more fundamental notions. Our next result establishes the identifiability of the effects of candidate policies in $\Pi$, provided that the policy scope $\Pi$ is backdoor-admissible in the causal diagram $\G$.

\begin{restatable}{theorem}{thmsbc}\label{thm:_4_3_sbc}
	Let $\G$ be a causal diagram, $\Pi = \Braces{\Tuple{X_i, \*S_i}}_{i = 1}^H$ be a policy space, and $\1R: \D(\*Y) \mapsto \3R$ be a reward function. If $\Pi$ is backdoor-admissible w.r.t. $\*Y$ in $\G$ (Def.~\ref{def:_4_3_sbc}), for any policy $\pi \in \Pi$, the expected reward $\invE{\1R(\*Y)}{\pi}$ is identifiable from $\G$. Moreover, $\invE{\1R(\*Y)}{\pi}$ is computable from the observational distribution $P(\*V)$ following the IPW (Thm.~\ref{thm:_4_1_ipw}) or the DP (Thm.~\ref{thm:_4_1_dp}) estimation. \hfill $\blacksquare$
\end{restatable}

In words, the sequential backdoor condition in Def.~\ref{def:_4_3_sbc} generalizes standard off-policy learning methods to settings where the NUC does not hold, and there exist unobserved confounders affecting actions and other variables in the system. As long as the sequential backdoor holds, one could apply IPW and DP algorithms to evaluate candidate policies from the observational data while ascertaining the validity of the estimation procedure in the limit.
\begin{example}\label{exp:_4_3_sbc3}
	Consider again the SCM $\1M^*$ described in Eq.~\ref{eq:_4_3_sbc1}. We are interested in evaluating a policy $\pi = (\pi_1, \pi_2)$ such that $\pi_1: X_1 \gets S_1, \pi_2: X_2 \gets  \neg S_1$. The submodel entailed by intervention $\doo(\pi_1, \pi_2)$ is described by the tuple
	\begin{align}
		\M^*_{\pi} = \Tuple{\*U = \{U_i\}_{i = 1}^5, \*V = \{S_1, X_1, S_2, X_2, Y\}, \2F_{\pi}, P(\*U)},
	\end{align}
	where the structural functions $\2F_{\pi}$ is given by
	\begin{align}\label{eq:_4_3_sbc2}
		\2F_{\pi} = \begin{cases}
			            S_1 \gets U_1,                       \\
			            X_1 \gets S_1,                       \\
			            S_2 \gets S_1 \oplus X_1 \oplus U_2, \\
			            X_2 \gets \neg S_1,                  \\
			            Y \gets S_1 \oplus X_1 \oplus U_3,
		            \end{cases}
	\end{align}
	Evaluating the expected reward $Y$ in submodel $\1M^*_{\pi}$ results in
	\begin{align}
		\invE{Y}{X_1 \gets S_1, X_2 \gets \neg S_1} & = \E \left[ S_1 \oplus \neg S_1 \oplus U_3 \right] \\
		                                            & = P(0 \oplus U_3 = 1)    \label{eq:_4_3_sbc3}
	\end{align}

	\begin{table}[t]
		\centering
		\renewcommand{\arraystretch}{1.25}
		\begin{tabular}{|cccc|c|cccc |c|}
			$S_1$ & $X_1$ & $X_2$ & $Y$ & $P(s_1, x_1,  x_2, y)$ & $S_1$ & $X_1$ & $X_2$ & $Y$ & $P(s_1, x_1,  x_2, y)$ \\
			\hline
			\hline
			0     & 0     & 0     & 0   & 0.2025                 & 1     & 0     & 0     & 0   & 0.0025                 \\
			0     & 0     & 0     & 1   & 0.0225                 & 1     & 0     & 0     & 1   & 0.0225                 \\
			0     & 0     & 1     & 0   & 0.0225                 & 1     & 0     & 1     & 0   & 0.0225                 \\
			0     & 0     & 1     & 1   & 0.2025                 & 1     & 0     & 1     & 1   & 0.0025                 \\
			0     & 1     & 0     & 0   & 0.2025                 & 1     & 1     & 0     & 0   & 0.0025                 \\
			0     & 1     & 0     & 1   & 0.0225                 & 1     & 1     & 0     & 1   & 0.0225                 \\
			0     & 1     & 1     & 0   & 0.0225                 & 1     & 1     & 1     & 0   & 0.0225                 \\
			0     & 1     & 1     & 1   & 0.2025                 & 1     & 1     & 1     & 1   & 0.0025                 \\ 
		\end{tabular}
		\caption{The observational distribution $P(S_1, X_1, X_2, Y)$ of the SCM $\1M^*$ defined in Eq.~\ref{eq:_4_3_sbc1}.}
		\label{tab_4_3_sbc1}
	\end{table}

	\noindent Evaluating the above equation gives $\invE{Y}{X_1 \gets S_1, X_2 \gets \neg S_1} = 0.9$. Since the agent does not have access to $\1M^*$, we apply next the IPW estimation procedure  (Thm.~\ref{thm:_4_1_ipw}) to evaluate the effects of policy  $\pi =(X_1 \gets S_1, X_2 \gets \neg S_1)$ from the observational distribution. Recall that scope $\Pi = \left \{ \langle X_1, \{S_1\}\rangle, \langle X_2, \{S_1\} \rangle \right \}$ is backdoor-admissible in the causal diagram $\G$ of Fig.~\ref{fig:_4_3_dtr1a}. Thm.~\ref{thm:_4_3_sbc} allows us to compute the expected reward from $P(S_1, X_1, X_2, Y)$ as follows:
	\begin{align}
		\E^{\textsc{ipw}}_{X_1 \gets S_1, X_2 \gets \neg S_1}\left [Y \right] & = \sum_{s_1, x_1, x_2} P \left (s_1, x_1,  x_2, Y = 1\right) \frac{\I\{x_1 = s_1\} }{P'\left(x_1\mid s_1 \right)} \frac{\I\{x_2 = \neg s_1\}}{P\left(x_2 \mid s_1, x_1 \right)} \notag \\
		                                                                      & = 4 \sum_{s_1, x_1, x_2} P \left (s_1, x_1,  x_2, Y = 1 \right)\I\{x_1 = s_1, x_2 = \neg s_1\}
	\end{align}
	The last step holds since $X_1 \gets U_{1} \oplus U_{4}$, $X_2 \gets U_2 \oplus U_5$; and $U_4$, $U_5$ are independent noise uniformly drawn over $\{0, 1\}$. The complete parametrization for the observational distribution $P(S_1, X_1, X_2, Y)$ is provided in Table~\ref{tab_4_3_sbc1}. The above equation could thus be further written as:
	\begin{align}
		\E^{\textsc{ipw}}_{X_1 \gets S_1, X_2 \gets \neg S_2}\left [Y \right] & = 4P(S_1 = 0, X_1 = 0, X_2 = 1, Y = 1) \notag \\
		                                                                      & +4P'(S_1 = 1, X_1 = 1, X_2 = 0, Y = 1)
	\end{align}
	The above evaluation matches the expected reward in Eq.~\ref{eq:_4_3_sbc3}, evaluated in SCM $\1M^*$. \hfill $\blacksquare$
\end{example}
\begin{example}\label{exp:_4_3_sbc4}
	We also apply the DP estimation (Thm.~\ref{thm:_4_1_dp}) to evaluate the effect of policy $\pi = (X_1 \gets S_1, X_2 \gets \neg S_1)$ in SCM $\1M^*$, described in Eq.~\ref{eq:_4_3_sbc1}. By applying Thm.~\ref{thm:_4_3_sbc}, we obtain the expected reward $ \E_{\pi}\left [Y \right]$ from the observational distribution $P(S_1, X_1, X_2, Y)$ as follows:
	\begin{align}
		Q^{(1)}_{\pi}(s_1, x_1)      & = \sum_{x_2}\I\{x_2 = \neg s_1\} Q^{(2)}_{\pi}(s_1, x_1, x_2) \\
		Q^{(2)}_{\pi}(s_1, x_1, x_2) & = P\left (Y = 1 | s_1, x_1, x_2 \right)
	\end{align}

	\begin{table}[t]
		\centering
		\hfill
		\begin{subtable}{0.38\linewidth}
			\vspace{0.65in}
			\centering
			\begin{tabular}{|cc|c|cc|c|}
				$S_1$ & $X_1$ & $Q^{(1)}_{\pi}$ & $S_1$ & $X_1$ & $Q^{(1)}_{\pi}$ \\
				\hline
				\hline
				0     & 0     & 0.9             & 1     & 0     & 0.9             \\
				0     & 1     & 0.9             & 1     & 1     & 0.9             \\
			\end{tabular}
			\caption{$Q^{(1)}_{\pi}(s_1, x_1)$}\label{tab_4_3_sbc2_a}
		\end{subtable}
		\hfill
		\begin{subtable}{0.5\linewidth}
			\centering
			\renewcommand{\arraystretch}{1.25}
			\begin{tabular}{|ccc|c|ccc|c|}
				$S_1$ & $X_1$ & $X_2$ & $Q^{(2)}_{\pi}$ & $S_1$ & $X_1$ & $X_2$ & $Q^{(2)}_{\pi}$ \\
				\hline
				\hline
				0     & 0     & 0     & 0.1             & 1     & 0     & 0     & 0.9             \\
				0     & 0     & 1     & 0.9             & 1     & 0     & 1     & 0.1             \\
				0     & 1     & 0     & 0.1             & 1     & 1     & 0     & 0.9             \\
				0     & 1     & 1     & 0.9             & 1     & 1     & 1     & 0.1             \\
			\end{tabular}
			\caption{$Q^{(2)}_{\pi}(s_1, x_1, x_2)$}\label{tab_4_3_sbc2_b}
		\end{subtable}\hfill\null
		\caption{Evaluation of value functions $Q^{(1)}_{\pi}, Q^{(2)}_{\pi}$ for policy $\pi = (X_1 \gets S_1, X_2 \gets \neg S_1)$ in SCM $\1M^*$ described in Eq.~\ref{eq:_4_3_sbc1}.}
		\label{tab_4_3_sbc2}
	\end{table}

	We compute the above value functions $Q^{(1)}_{\pi}(s_1, x_1), Q^{(2)}_{\pi}(s_1, x_1, x_2)$; their parameterizations are provided in Table~\ref{tab_4_3_sbc2}. Finally, the expected reward is identifiable by
	\begin{align}
		\E^{\textsc{dp}}_{X_1 \gets S_1, X_2 \gets \neg S_1}[Y] & = \sum_{s_1, x_1} \I\{x_1 = s_1\}Q^{(1)}_{\pi}(s_1, x_1)P(s_1) \notag \\
		                                                        & =Q^{(1)}_{\pi}(0, 0)P(S_1 = 0) + Q^{(1)}_{\pi}(1, 1)P(S_1 = 1)
	\end{align}
	The above computation matches the expected reward in Eq.~\ref{eq:_4_3_sbc3}, evaluated in SCM $\1M^*$.  \hfill $\blacksquare$
\end{example}

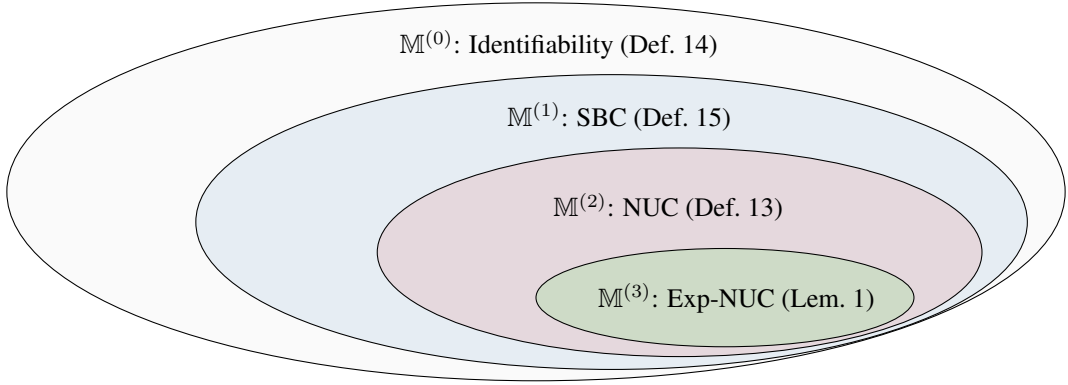
\begin{figure}[t]
	\centering
	\begin{tikzpicture}
		\draw[fill=bettergray, fill opacity=0.1] (0,0) ellipse (7cm and 2.5cm);
		\draw[fill=betterblue, fill opacity=0.1, xshift=1cm, yshift = -0.4cm] (0,0) ellipse (5.5cm and 1.95cm);
		\draw[fill=betterred, fill opacity=0.1, xshift=1.9cm, yshift = -0.8cm] (0,0) ellipse (4cm and 1.38cm);
		\draw[fill=green, fill opacity=0.1, xshift=2.5cm, yshift = -1.4cm] (0,0) ellipse (2.5cm and 0.65cm);
		\node[align=left] at (0.3, 1.95) {\small $\3M^{(0)}$: Identifiability (Def.~\ref{def:_4_3_id})};
		\node[align=left] at (1.1, 1) {\small $\3M^{(1)}$: SBC (Def.~\ref{def:_4_3_sbc})};
		\node[align=left] at (1.75, -0.2) {\small $\3M^{(2)}$: NUC (Def.~\ref{def:_4_1_nuc})};
		\node[align=left] at (2.65, -1.4) {\small $\3M^{(3)}$: Exp-NUC (Lem.~\ref{lem:_4_2_nuc})};

	\end{tikzpicture}
	\caption{Assumptions under which IPW and DP algorithms are applicable}
	\label{fig:_4_3_nuc}
\end{figure}

We provide in Fig.~\ref{fig:_4_3_nuc}  a summary of the relationships of structural assumptions about the data-generating process discussed so far that permit the evaluation of the effects of candidate policies. 
\begin{itemize}
	\item Sec.~\ref{sec:_4_1} provides a sufficient assumption called the NUC (Def.~\ref{def:_4_1_nuc}) that allows one to evaluate policy effects from the observational data via the application of IPW and DP algorithms (Thms.~\ref{thm:_4_1_ipw} and \ref{thm:_4_1_dp}). The family of SCMs satisfying the NUC condition is denoted by $\3M^{(2)}$, which is marked in red in the figure, and is the baseline of most of the current literature. However, the NUC assumption is opaque and not testable in practice, which may lead to potentially wrong inferences about the optimal policy (as shown in Examples~\ref{exp:_4_1_mab}). 
\item In Sec.~\ref{sec:_4_2}, we show that the NUC condition holds in all submodels induced by interventions $\doo(\pi)$ following candidate policies $\pi \in \Pi$, which is called Exp-NUC. 
This family of submodels is denoted by $\3M^{(3)}$ and is marked in dark green in the figure. In words, whenever an agent goes online in the environment and collects experimental data following a known policy, the same data can be used to evaluate the effect of new policies, as implied by Lemma~\ref{lem:_4_2_nuc}. 
Formally, this implies the NUC condition and constitutes a sufficient condition for the identification of the effects of policy interventions. 
\item On the other side in blue, Sec.~\ref{sec:_4_3_1} provides a more generalized graphical condition, called the sequential backdoor criterion (SBC, Def.~\ref{def:_4_3_sbc}). The family of SCMs satisfying the sequential backdoor condition is denoted by $\3M^{(1)}$. This graphical condition is defined based on the causal diagram encoding the underlying causal mechanisms, which could be more readily evaluatable from the available data and by domain experts. It is sufficient in determining whether IPW and DP are applicable to evaluate candidate policies from observational data. 
\item Finally, the outer ellipse $\3M^{(0)}$ describes the set of all environments (SCMs) where the effects of candidate policies are identifiable from the underlying data-generating mechanisms. 
\end{itemize}

After all, Fig.~\ref{fig:_4_3_nuc} shows that $\3M^{(3)} \subset \3M^{(2)} \subset \3M^{(1)}$; and this containment relationship is strict. This means that there exists a causal model that is not induced by the intervention $\doo(\pi)$ and satisfies the NUC condition (Example~\ref{exp:_4_1_nuc1}). There also exists a causal model where the NUC does not hold, but is backdoor admissible (Example~\ref{exp:_4_3_sbc2}). One may wonder if the containment $\3M^{(1)} \subset \3M^{(0)}$ is also strict. We will show next that this is the case. Particularly, there are learning settings where the sequential backdoor criterion does not hold, but the agent could still explore the structural constraints in the causal diagram to recover the effects of candidate policies from the observational data. 

\subsubsection{Do-Calculus Learning} \label{sec:_4_3_2}
Our discussion begins with an example illustrating policy evaluation from observational data under a set of non-parametric constraints known as the front-door \citep[Sec.~3.1.2]{pearl:2k}.
\begin{example}[Front-door Environment]\label{exp:_4_3_fd1}
	Consider the causal diagram $\G$ described in Fig.~\ref{fig:_4_3_fda}, which is also known as the ``front-door'' diagram. 
We are interested in evaluating the expected reward $\E_{x}[Y]$ of an atomic policy $\pi:X \gets x$ that sets the value of action $X$ to a constant $x$. 
Due to the presence of a backdoor path $X \dashleftarrow \dashrightarrow Y$, the policy space $\Pi = \{ \langle X, \emptyset \rangle \}$ does not satisfy the backdoor criterion (Def.~\ref{def:_4_3_sbc}), and standard off-policy learning algorithm do not generally apply. 

In our context, this means that the policy from which the data is coming from was implemented in the environment where the agent had access to the unobserved confounder, marked as the dashed-bidirected arrow in the graph. 
This unobserved confounder seems to suggest that the expected reward $\invE{Y}{x}$ is not identifiable from the Front-Door diagram, and the agent should, therefore, go online following the discussion in Sec.~\ref{sec:_4_2}.

However, existing results in causal inference suggest otherwise. By applying the Front-Door adjustment in \cite[Thm.~3.3.4]{pearl:2k}, the expected reward $\invE{Y}{x}$ can be computed from the observational distribution $P(X, Y, W)$ through the following mapping:
	\begin{align}
		\invE{Y}{x}  = \sum_{w}P\left (w \mid x \right) \sum_{x'}\E\left [ Y \mid w, x' \right]P(x'). \label{eq:_4_3_fd}
	\end{align}
Among the quantities in the above equation, $P\left (w \mid x \right)$, $\E\left [ Y \mid w, x' \right]$, and $P(x')$ are all functions of the observational distribution $P(X, Y, W)$. Therefore, the expected reward $\invE{Y}{x}$ is identifiable from the front-door diagram. The learner could then evaluate the expected reward of every arm $X \gets x$ from the observational data and solve for the optimal treatment $x^*$. \hfill $\blacksquare$
\end{example}

For the remainder of this section, we will introduce complete machinery to identify the expected rewards of candidate policies $\pi \in \Pi$ from the causal diagram $\G$. Our discussion begins with a procedure to reduce the original identification problem into identifying effects of atomic interventions (e.g., $\doo(\*x)$) from the same diagram $\G$. Let $\*Y \subseteq \*V$ be an arbitrary subset of endogenous variables. For any policy $\pi \in \Pi$, evaluating the joint distribution over $\*Y$ in submodel $\1M_{\pi}$ is given by:
\begin{align}
	\inv{\*y}{\pi} & = \sum_{ \*v \setminus \*y} \sum_{\*u} P(\*u) \prod_{V \in \*V \setminus \*X} P(v \mid \*\pa_V, \*u_V) \prod_{X_i \in \*X} \pi_i(x_i \mid \*s_i)
\end{align}
Recall that $\G_{\pi}$ is a manipulated graph obtained from $\G$ by replacing incoming arrows of node $X_i$ with arrows from covariates $\*S_i$ to $X_i$ for every action $X_i \in \*X$. Let $\*Z = \An(\*Y)_{\G_{\pi}}$ be ancestors of nodes $\*Y$. Then all the non-ancestor nodes can be summed out from the above equation leading to
\footnote{The decomposition in Eq.~\ref{eq:_4_3_pid1} was introduced in \citep{tian2004identifying} and extended for identifying policy effects from out-of-domain distributions in \citep{correa2020general} .}
\begin{align}
	\inv{\*y}{\pi} & = \sum_{\*z \setminus \*y} \sum_{\*u} P(\*u) \prod_{V \in \*Z \setminus \*X} P(v \mid \*\pa_V, \*u_V) \prod_{X_i \in \*X \cap \*Z} \pi_i(x_i \mid \*s_i)                                  \\
	               & =\sum_{\*x^*, \*s^*} \underbrace{P_{\*x^*}(\*y, \*s^*)}_{\text{atomic intervention}}\prod_{X_i \in \*X^*} \underbrace{\pi_i(x_i \mid \*s_i)}_{\text{new policy} \;\pi} \label{eq:_4_3_pid1}
\end{align}
where $\*X^* = \*X \cap \*Z$ are actions in $\*X$ that are ancestors of $\*Y$ in the post-interventional graph $\G_{\pi}$; $\*S^* = \*Z \setminus (\*X \cup \*Y )$ are ancestors of $\*Y$ in $\G_{\pi}$, excluding $\*Y$ and $\*X$. It follows from Eq.~\ref{eq:_4_3_pid1} that the interventional distribution $\inv{\*Y}{\pi}$ is identifiable if the distribution $P_{\*x^*}(\*Y, \*S^*)$ induced by atomic intervention $\doo(\*x^*)$ is identifiable. The following proposition implies that the reverse also holds.
\begin{proposition}[\cite{correa2019statistical,tian2004identifying}]\label{lem:_4_3_pid}
	Let $\G$ be a causal diagram, $\Pi$ be a policy space $\Braces{\Tuple{X_i, \*S}}_{i = 1}^H$, and $\*Y \subseteq \*V$ be a set of variables. For any policy $\pi \in \Pi$, $P_{\pi}(\*Y)$ is identifiable from $\1G$ if and only if $P_{\*x^*}(\*Y, \*S^*)$ for any $\*x^* \in \D(\*X^*)$ is identifiable from $\G$. \hfill $\blacksquare$
\end{proposition}
The following examples demonstrate the decomposition in Eq.~\ref{eq:_4_3_pid1} with various causal diagrams.
\begin{example}\label{exp:_4_3_pid2}
	Consider the Front-door diagram $\G$ described in Fig.~\ref{fig:_4_3_fda} and a policy scope $\Pi = \left \{ \langle X, \emptyset \rangle \right\}$. For any policy $\pi(X)$, the post-interventional graph $\G_{\pi}$ is a chain $X \rightarrow W \rightarrow Y$. For the reward $Y$ in this graph, its ancestor action $\*X^* = \An(Y) \cap \{X\} = \{X\}$ and other ancestor nodes $\*S^* = \An(Y) \setminus \{Y, X\} = \{W\}$. Following the decomposition in Eq.~\ref{eq:_4_3_pid1}, the expected reward $\invE{Y}{\pi}$ for any policy $\pi \in \Pi$ could be written as
	\begin{align}
		\invE{Y}{\pi} & = \sum_{y} y \inv{y}{\pi} \notag                                  \\
		              & = \sum_{y} y \sum_{x, w} \inv{y, w}{x} \pi(x) \label{eq:_4_3_pid3}
	\end{align}
Among quantities in the above equation, $\inv{Y, W}{x}$ is the interventional distribution induced by atomic intervention $\doo(X \gets x)$.  \hfill $\blacksquare$
\end{example}
\begin{example}\label{exp:_4_3_pid1}
	Figs.~\ref{fig:_4_3_dtr1a} and \ref{fig:_4_3_dtr1b} show a causal diagram $\G$ and the post-interventional diagram $\G_{\pi}$ associated with a policy space $\Pi = \left \{ \langle X_1, \{S_1\}\rangle, \langle X_2, \{S_1\} \rangle \right\}$. For the reward node $Y$ in graph $\G_{\pi}$, ancestor actions $\*X^* = \An(Y) \cap \{X_1, X_2\} = \{X_2\}$; and covariates $\*S^* = \An(Y) \setminus \{Y, X_1, X_2\} = \{S_1\}$. For any policy $\pi \in \Pi$, the expected reward $\invE{Y}{\pi}$ could be written as:
	\begin{align}
		\invE{Y}{\pi} & = \sum_{y} y \inv{y}{\pi} \notag                                                  \\
		              & = \sum_{y} y \sum_{x_2, s_1} \inv{y, s_1}{x_2} \pi_2(x_2|s_1) \label{eq:_4_3_pid2}
	\end{align}
	The last step follows from the decomposition of Eq.~\ref{eq:_4_3_pid1}. Among quantities in the above equation, $\inv{Y, S_1}{x_2}$ is the interventional distribution induced by atomic intervention $\doo(X_2 \gets x_2)$.  \hfill $\blacksquare$
\end{example}
Lem.~\ref{lem:_4_3_pid} implies that in order to evaluate candidate policies from the observational distribution, it is sufficient to identify the corresponding effects induced by atomic interventions. Such a problem has been studied in the literature, and several algorithms and graphical criteria have been proposed \citep{pearl:2k,spirtes2001}. First and foremost, we formally introduce do-calculus \citep{pearl:95}, which consists of three inferential rules. Each rule dictates that two interventional distributions are equivalent under a condition that can be read off from the causal diagram corresponding to the underlying, unobserved SCM.
\begin{theorem}[Rules of do-calculus \citep{pearl:2k}]\label{def:_4_3_do-calculus}
	Let $\G$ be a causal diagram compatible with a structural causal model $\M$, with endogenous variables $\*V$. For any disjoint subsets $\*X, \*Y, \*Z, \*W \subseteq \*V$, the following rules hold for interventional distributions compatible with $\G$:
	\begin{itemize}[leftmargin=40pt,topsep=0pt,itemsep=0pt,parsep=0pt,partopsep=0pt]
		\item [\textbf{Rule 1}] Insertion/deletion of observations:
		      \begin{align}
			      P_{\*x}(\*y \mid \*z, \*w) = P_{\*x}(\*y \mid \*w) \textrm{ if } (\*Y \ci \*Z \mid \*X, \*W) \textrm{ in }\G_{\overline{\*X}}
		      \end{align}
		\item [\textbf{Rule 2}] Action/observation exchange:
		      \begin{align}
			      P_{\*x, \*z}(\*y \mid \*w) = P_{\*x}(\*y \mid \*z, \*w) \textrm{ if } (\*Y \ci \*Z \mid  \*X, \*W) \textrm{ in } \G_{\overline{\*X}\underline{\*Z}}
		      \end{align}
		\item [\textbf{Rule 3}] Insertion/deletion of actions:
		      \begin{align}
			      P_{\*x, \*z}(\*y \mid \*w) = P_{\*x}(\*y \mid \*w) \textrm{ if } (\*Y \ci \*Z \mid  \*X, \*W) \textrm{ in } \G_{\overline{\*X}\overline{\*Z(\*W)}}
		      \end{align}
	\end{itemize}
	\noindent where $\*Z(\*W)$ is the subset of nodes in $\*Z$ that are not ancestors of $\*W$-nodes in $\G_{\overline{\*X}}$ \hfill $\blacksquare$
\end{theorem}
The first rule affirms that the $d$-separation criterion also holds for causal diagrams under intervention.
The second rule gives the condition for when observing and intervening on variables $\*Z$ are equivalent from the perspective of outcomes $\*Y$.
The third rule gives the conditions for when the do-operator can be removed entirely from the expression, i.e., there is no causal effect of $\
	\*Z$ on $\*Y$.

We call \textit{do-calculus learning} an algorithmic procedure to identify causal effects from the observational distribution through the applications of do-calculus together with standard mathematical rules,  or in some equivalent, perhaps more systematic form \citep{tian2002general}.
Such a procedure can be shown sufficient and necessary to identify causal effects from observational \citep{shpitser:pea06-r327,huang:val06} and interventional distributions \citep{bareinboim2012causal,lee2019gid}. This means that if $P_{\*x}(\*Y)$ cannot be expressed in terms of observational probabilities $P(\*V)$ (or $P_{\*z}(\*V)$, for some $\*Z$) by repeated applications of these three rules together with basic probability algebra, such an expression does not exist, and the effect is non-identifiable.
\begin{proposition}\label{thm:_4_3_do-id}
	Rules of Do-calculus, together with standard probability manipulations, are sound and complete for determining the identifiability of all interventional distributions of the form $P_{\*x}(\*Y)$ from a causal diagram $\G$ and the available observational and interventional distributions. \hfill $\blacksquare$
\end{proposition}
The following examples demonstrate how to apply do-calculus learning to evaluate candidate policies with an arbitrary policy space $\Pi$ from the observational distribution in different causal diagrams.

\begin{example}\label{exp:_4_3_id2}

	\begin{figure}
		\hfill\begin{subfigure}{0.3\textwidth}\centering
			\begin{tikzpicture}
				\def\outerr{3.5}
				\def\innerr{3}

				\node[vertex] (X) at (0, 0) {X};
				\node[vertex] (W) at (1.3, 0) {W};
				\node[vertex] (Y) at (2.6, 0) {Y};

				\draw[bidir] (X) to [bend left = 45] (Y);
				\draw[dir] (X) to (W);
				\draw[dir] (W) to (Y);

				\begin{pgfonlayer}{back}
					\node[circle,fill=betterblue!65,draw=none,minimum size=2*\innerr mm] at (X) {};
					\node[circle,fill=betterred!65,draw=none,minimum size=2*\innerr mm] at (Y) {};
				\end{pgfonlayer}
			\end{tikzpicture}
			\caption{$\G$}
			\label{fig:_4_3_fda}
		\end{subfigure}\hfill%
		\begin{subfigure}{0.3\textwidth}\centering
			\begin{tikzpicture}
				\def\outerr{3.5}
				\def\innerr{3}

				\node[vertex] (X) at (0, 0) {X};
				\node[vertex] (W) at (1.3, 0) {W};
				\node[vertex] (Y) at (2.6, 0) {Y};

				\draw[bidir] (X) to [bend left = 45] (Y);
				\draw[dir] (W) to (Y);

				\begin{pgfonlayer}{back}
					\node[circle,fill=betterblue!65,draw=none,minimum size=2*\innerr mm] at (X) {};
					\node[circle,fill=betterred!65,draw=none,minimum size=2*\innerr mm] at (Y) {};
				\end{pgfonlayer}
			\end{tikzpicture}
			\caption{$\G_{\underline{X}}$}
			\label{fig:_4_3_fdb}
		\end{subfigure}\hfill%
		\begin{subfigure}{0.3\textwidth}\centering
			\begin{tikzpicture}
				\def\outerr{3.5}
				\def\innerr{3}

				\node[vertex] (X) at (0, 0) {X};
				\node[vertex] (W) at (1.3, 0) {W};
				\node[vertex] (Y) at (2.6, 0) {Y};

				\draw[bidir,opacity=0] (X) to [bend left = 45] (Y);
				\draw[dir] (X) to (W);

				\begin{pgfonlayer}{back}
					\node[circle,fill=betterblue!65,draw=none,minimum size=2*\innerr mm] at (X) {};
					\node[circle,fill=betterred!65,draw=none,minimum size=2*\innerr mm] at (Y) {};
				\end{pgfonlayer}
			\end{tikzpicture}
			\caption{$\G_{\overline{X}\underline{W}}$}
			\label{fig:_4_3_fdc}
		\end{subfigure}\hfill\null
		\smallskip

		\hfill\begin{subfigure}{0.3\textwidth}\centering
			\begin{tikzpicture}
				\def\outerr{3.5}
				\def\innerr{3}

				\node[vertex] (X) at (0, 0) {X};
				\node[vertex] (W) at (1.3, 0) {W};
				\node[vertex] (Y) at (2.6, 0) {Y};

				\draw[bidir,opacity=0] (X) to [bend left = 45] (Y);
				\draw[dir] (W) to (Y);

				\begin{pgfonlayer}{back}
					\node[circle,fill=betterblue!65,draw=none,minimum size=2*\innerr mm] at (X) {};
					\node[circle,fill=betterred!65,draw=none,minimum size=2*\innerr mm] at (Y) {};
				\end{pgfonlayer}
			\end{tikzpicture}
			\caption{$\G_{\overline{X,W}}$}
			\label{fig:_4_3_fdd}
		\end{subfigure}\hfill%
		\begin{subfigure}{0.3\textwidth}\centering
			\begin{tikzpicture}
				\def\outerr{3.5}
				\def\innerr{3}

				\node[vertex] (X) at (0, 0) {X};
				\node[vertex] (W) at (1.3, 0) {W};
				\node[vertex] (Y) at (2.6, 0) {Y};

				\draw[bidir] (X) to [bend left = 45] (Y);
				\draw[dir] (X) to (W);

				\begin{pgfonlayer}{back}
					\node[circle,fill=betterblue!65,draw=none,minimum size=2*\innerr mm] at (X) {};
					\node[circle,fill=betterred!65,draw=none,minimum size=2*\innerr mm] at (Y) {};
				\end{pgfonlayer}
			\end{tikzpicture}
			\caption{$\G_{\underline{W}}$}
			\label{fig:_4_3_fde}
		\end{subfigure}\hfill%
		\begin{subfigure}{0.3\textwidth}\centering
			\begin{tikzpicture}
				\def\outerr{3.5}
				\def\innerr{3}

				\node[vertex] (X) at (0, 0) {X};
				\node[vertex] (W) at (1.3, 0) {W};
				\node[vertex] (Y) at (2.6, 0) {Y};

				\draw[bidir] (X) to [bend left = 45] (Y);
				\draw[dir] (W) to (Y);

				\begin{pgfonlayer}{back}
					\node[circle,fill=betterblue!65,draw=none,minimum size=2*\innerr mm] at (X) {};
					\node[circle,fill=betterred!65,draw=none,minimum size=2*\innerr mm] at (Y) {};
				\end{pgfonlayer}
			\end{tikzpicture}
			\caption{$\G_{\overline{W}}$}
			\label{fig:_4_3_fdf}
		\end{subfigure}\hfill\null
		\caption{A front-door graph and its manipulated representations.}
		\label{fig:_4_3_fd}
	\end{figure}
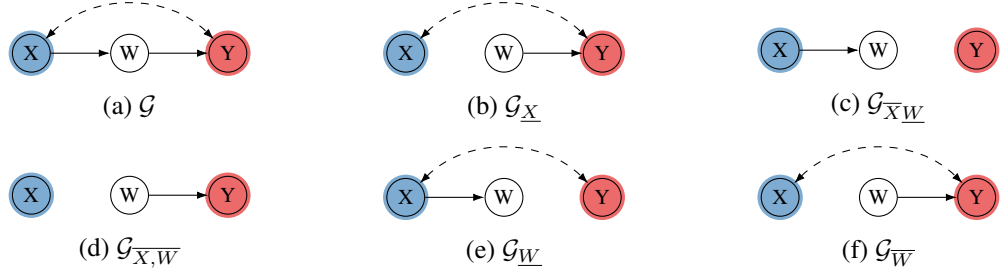

	Consider the front-door diagram $\G$ in Fig.~\ref{fig:_4_3_fda}. The decomposition in Eq.~\ref{eq:_4_3_pid3} implies that in order to evaluate the effect of a policy $\pi(X)$, it is sufficient to identify the interventional distribution $\inv{Y, W}{x}$.  We try to remove the subscript from every probability term that appears in the expression of $\inv{Y, W}{x}$ since its absence represents the fact that the causal effect is expressible in terms of the observational distribution, hence computable from the available data (independent from the underlying functions and exogenous variables). A derivation of a `subscript-free' expression for $\inv{Y, W}{x}$ is given below Eqs.~\ref{eq:_4_3_fd:deriv:1} and \ref{eq:_4_3_fd:last}. We illustrate the application of rules of do-calculus, Eqs.~\ref{eq:_4_3_fd:deriv:2}-\ref{eq:_4_3_fd:last}, in Figs.~\ref{fig:_4_3_fdb}-\ref{fig:_4_3_fdf}.
	\begingroup\allowdisplaybreaks\begin{align}
		\inv{y, w}{x} & = P_x(y |w) P_x(w)                          & \textrm{Probability Axioms} \label{eq:_4_3_fd:deriv:1}                                        \\
		              & =  P_x(y |w) P(w | x)                       & \textrm{Rule 2 } (W \ci X)_{\G_{\underline{X}}} \label{eq:_4_3_fd:deriv:2}                    \\
		              & = P_{x, w}(y) P(w | x)                      & \textrm{Rule 2 } (Y \ci W \mid X)_{\G_{\overline{X}\underline{W}}} \label{eq:_4_3_fd:deriv:3} \\
		              & =  P_w(y) P(w | x)                          & \textrm{Rule 3 } (Y \ci X\mid W)_{\G_{\overline{X,W}}} \label{eq:_4_3_fd:deriv:4}             \\
		              & =  P(w | x)  \sum_{x'} P_w(y | x') P_w(x')  & \textrm{Probability Axioms} \label{eq:_4_3_fd:deriv:5}                                        \\
		              & =  P(w | x)  \sum_{x'} P(y | w, x') P_w(x') & \textrm{Rule 2 } (Y \ci W \mid X)_{\G_{\underline{W}}} \label{eq:_4_3_fd:deriv:6}             \\
		              & =  P(w | x)  \sum_{x'} P(y | w, x') P(x')   & \textrm{Rule 3 } (X \ci W)_{\G_{\overline{W}}} \label{eq:_4_3_fd:last}                        %
	\end{align}\endgroup%
	We note that the do-operator (i.e., the subscript) does not appear in the final expression in Eq.~\ref{eq:_4_3_fd:last}, so even though we do not possess any quantitative knowledge about the unobservable variable $U$ (neither its distribution nor its dimensionality), besides the fact that it influences both $\{X, Y\}$, we are still able to compute the causal effect purely from the observational distribution $P(\*V)$ together with the assumption encoded in $\G$. This together with Eq.~\ref{eq:_4_3_pid3} allows us to evaluate the effect of any policy $\pi(X)$ from the observational data, i.e.,
	\begin{align}
		\E_{\pi}[Y] & =\sum_{y} y \sum_{x, w} P(w | x)  \sum_{x'} P(y | w, x') P(x') \pi(x) \\
		            & = \sum_{x, w} P(w | x)  \sum_{x'} \sum_{y} yP(y | w, x') P(x') \pi(x) \\
		            & = \sum_{x, w} P(w | x)  \sum_{x'} \E[Y | w, x'] P(x') \pi(x)
	\end{align}  
This recovers the front-door adjustment formula in Eq.~\ref{eq:_4_3_fd}. \hfill $\blacksquare$
\end{example}

	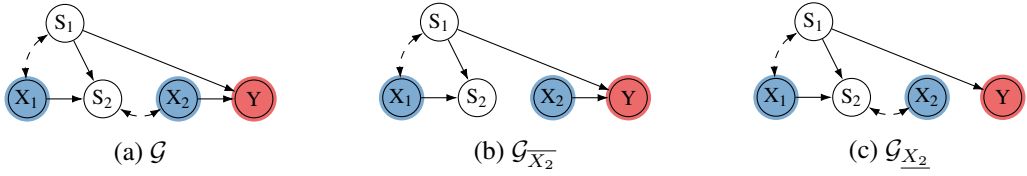
\begin{figure}
		\hfill\begin{subfigure}{0.3\textwidth}\centering
			\begin{tikzpicture}
				\def\outerr{3.5}
				\def\innerr{3}
				\node[vertex] (Z1) at (0.5, 1) {S\textsubscript{1}};
				\node[vertex] (X1) at (0, 0) {X\textsubscript{1}};
				\node[vertex] (Z2) at (1, 0) {S\textsubscript{2}};
				\node[vertex] (X2) at (2, 0) {X\textsubscript{2}};
				\node[vertex] (Y) at (3, 0) {Y};

				\draw[dir] (Z1) to (Y);

				\draw[dir] (X1) to (Z2);

				\draw[bidir] (Z2) to [bend right = 30] (X2);

				\draw[dir] (X2) to (Y);

				\draw[dir] (Z1) to (Z2);
				\draw[bidir]  (Z1) to [bend right = 30] (X1);

				\begin{pgfonlayer}{back}
					\node[circle,fill=betterblue!65,draw=none,minimum size=2*\innerr mm] at (X1) {};
					\node[circle,fill=betterblue!65,draw=none,minimum size=2*\innerr mm] at (X2) {};
					\node[circle,fill=betterred!65,draw=none,minimum size=2*\innerr mm] at (Y) {};
				\end{pgfonlayer}
			\end{tikzpicture}
			\caption{$\G$}
			\label{fig:_4_3_sbc2a}
		\end{subfigure}\hfill%
		\begin{subfigure}{0.3\textwidth}\centering
			\begin{tikzpicture}
				\def\outerr{3.5}
				\def\innerr{3}
				\node[vertex] (Z1) at (0.5, 1) {S\textsubscript{1}};
				\node[vertex] (X1) at (0, 0) {X\textsubscript{1}};
				\node[vertex] (Z2) at (1, 0) {S\textsubscript{2}};
				\node[vertex] (X2) at (2, 0) {X\textsubscript{2}};
				\node[vertex] (Y) at (3, 0) {Y};

				\draw[dir] (Z1) to (Y);

				\draw[dir] (X1) to (Z2);

				\draw[dir] (X2) to (Y);

				\draw[dir] (Z1) to (Z2);
				\draw[bidir]  (Z1) to [bend right = 30] (X1);

				\begin{pgfonlayer}{back}
					\node[circle,fill=betterblue!65,draw=none,minimum size=2*\innerr mm] at (X1) {};
					\node[circle,fill=betterblue!65,draw=none,minimum size=2*\innerr mm] at (X2) {};
					\node[circle,fill=betterred!65,draw=none,minimum size=2*\innerr mm] at (Y) {};
				\end{pgfonlayer}
			\end{tikzpicture}
			\caption{$\G_{\overline{X_2}}$}
			\label{fig:_4_3_sbc2b}
		\end{subfigure}\hfill%
		\begin{subfigure}{0.3\textwidth}\centering
			\begin{tikzpicture}
				\def\outerr{3.5}
				\def\innerr{3}
				\node[vertex] (Z1) at (0.5, 1) {S\textsubscript{1}};
				\node[vertex] (X1) at (0, 0) {X\textsubscript{1}};
				\node[vertex] (Z2) at (1, 0) {S\textsubscript{2}};
				\node[vertex] (X2) at (2, 0) {X\textsubscript{2}};
				\node[vertex] (Y) at (3, 0) {Y};

				\draw[dir] (Z1) to (Y);

				\draw[dir] (X1) to (Z2);

				\draw[bidir] (Z2) to [bend right = 30] (X2);

				\draw[dir] (Z1) to (Z2);
				\draw[bidir]  (Z1) to [bend right = 30] (X1);

				\begin{pgfonlayer}{back}
					\node[circle,fill=betterblue!65,draw=none,minimum size=2*\innerr mm] at (X1) {};
					\node[circle,fill=betterblue!65,draw=none,minimum size=2*\innerr mm] at (X2) {};
					\node[circle,fill=betterred!65,draw=none,minimum size=2*\innerr mm] at (Y) {};
				\end{pgfonlayer}
			\end{tikzpicture}
			\caption{$\G_{\underline{X_2}}$}
			\label{fig:_4_3_sbc2c}
		\end{subfigure}\hfill\null
		\caption{A causal diagram of the SCM described in Eq.~\ref{eq:_4_3_sbc1} and its manipulated diagrams.}
		\label{fig:_4_3_sbc2}
	\end{figure}
	
\begin{example}\label{exp:_4_3_id1}
	Consider the causal diagram $\G$ in Fig.~\ref{fig:_4_3_sbc2a}. We are interested in evaluating the effects of a policy of the form $\pi = \Parens{\pi_1(X_1 \mid S_1), \pi_2(X_2\mid S_1)}$. The decomposition in Eq.~\ref{eq:_4_3_pid2} implies that it is sufficient to identify the interventional distribution $\inv{Y, S_1}{x_2}$. 
	
	Our goal then will be to remove the subscript from every probability term in the expression of $\inv{Y, S_1}{x_2}$, as shown next.
	We illustrate the application of the do-calculus in the equations below and with the sub-graphs shown in Figs.~\ref{fig:_4_3_sbc2b}-\ref{fig:_4_3_sbc2c}.
	We start by writing the target expression:
	\begingroup\allowdisplaybreaks\begin{align}
		\inv{y, s_1}{x_2} & = P_{x_2}(y |s_1) P_{x_2}(s_1) & \textrm{Probability Axioms} \label{eq:_4_3_id:deriv:1}                                  \\
		                  & =  P_{x_2}(y |s_1) P(s_1)      & \textrm{Rule 3 } (S_1 \ci X)_{\G_{\overline{X_2}}} \label{eq:_4_3_id:deriv:2}           \\
		                  & =P(y |s_1, x_2)P(s_1)          & \textrm{Rule 2 } (Y \ci X_2 \mid S_1)_{\G_{\underline{X_2}}} \label{eq:_4_3_id:deriv:3}
	\end{align}\endgroup%
	The above formula, together with Eq.~\ref{eq:_4_3_pid2}, allows us to identify the expected reward of any policy of the form $\pi = \Parens{\pi_1(X_1 \mid S_1), \pi_2(X_2\mid S_1)}$ from the observational data, i.e.,
	\begin{align}
		\E_{\pi}[Y] & =\sum_{y} y \sum_{x_2, s_1} P(y |s_1, x_2)P(s_1) \pi_2(x_2|s_1)   \\
		            & = \sum_{x_2, s_1} \sum_{y} y  P(y |s_1, x_2)P(s_1) \pi_2(x_2|s_1) \\
		            & = \sum_{x_2, s_1} \E[Y |s_1, x_2]P(s_1) \pi_2(x_2|s_1)
	\end{align}
	More specifically, let policy $\pi = \Parens{X_1 \gets S_1, X_2 \gets S_1}$. The above equation leads to an evaluation of the expected reward given by:
	\begin{align}
		\E_{X_1 \gets S_1, X_2 \gets \neg S_2}\left [Y \right] & = \sum_{x_2, s_1} \E[Y |s_1, x_2]P(s_1) \I\{x_2 = \neg s_1\}
	\end{align}
	The complete parametrizations for the conditional reward $\E[Y|S_1, X_1]$ and distribution $P(S_1)$ are provided in Table~\ref{tab_4_3_sbc3}. The above equation could thus be further written as:
	\begin{align}
		\E_{X_1 \gets S_1, X_2 \gets \neg S_2}\left [Y \right] & = \E[Y |S_1 = 0, X_2 = 1]P(S_1 = 0)
	\end{align}
	The above computation matches the reward in Eq.~\ref{eq:_4_3_sbc3}, as evaluated in SCM $\1M^*$.  \hfill $\blacksquare$
	
		\begin{table}[t]
		\centering
		\hfill
		\begin{subtable}{0.2\linewidth}
			\centering
			\renewcommand{\arraystretch}{1.25}
			\begin{tabular}{|c|c|}
				$S_1$ & $P(s_1)$ \\
				\hline
				\hline
				0     & 0.9      \\
				1     & 0.1      \\
			\end{tabular}
			\caption{$P(S_1)$}\label{tab_4_3_sbc3a}
		\end{subtable}
		\hfill
		\begin{subtable}{0.53\linewidth}
			\centering
			\renewcommand{\arraystretch}{1.25}
			\begin{tabular}{|cc|c|cc |c|}
				$S_1$ & $X_1$ & $\E[Y|s_1, x_1]$ & $S_1$ & $X_1$ & $\E[Y|s_1, x_1]$ \\
				\hline
				\hline
				0     & 0     & 0.1              & 1     & 0     & 0.9              \\
				0     & 1     & 0.9              & 1     & 1     & 0.1              \\
			\end{tabular}
			\caption{$\E[Y|S_1, X_1]$}\label{tab_4_3_sbc3b}
		\end{subtable}\hfill\null
		\caption{Evaluation of $P(S_1)$ and $\E[Y|S_1, X_1]$ in SCM $\1M^*$ defined in Eq.~\ref{eq:_4_3_sbc1}.}
		\label{tab_4_3_sbc3}
	\end{table}

\end{example}
In the above examples, how we apply the rules of do-calculus in the right sequence to obtain a desirable expression is rather unclear. Fortunately, researchers have algorithmatized the procedure to obtain such a sequence, in light of identifiability problems \citep{tian2002general,shpitser2006identification,huang:identifiability,bareinboim2012causal,lee2019gid}. This means that there exist efficient algorithms to determine the identifiability of the expected rewards of candidate policies from the causal diagram, and if exits, return the identification formula for the target effects from the observational distribution. The algorithms run in a polynomial number of steps relative to the number of nodes and edges in the causal diagram.

\subsection{Novel Causal Reinforcement Learning Tasks} \label{sec:_4_4}
We recall the CRL agent is embedded in a CDM $\Tuple{\1M^*, \Pi, \1R}$ (Def.~\ref{def:_3_1_cdm}), where the SCM $\1M^*$ is not fully observed, $\Pi$ represents the policy space, and $\1R$ is the reward function. 
Even though the agent is still evaluated by  $\1M^*$, we substitute it with the learning regime $\1L$ and structural assumptions $\1A$ about the environment, which lead to a new signature $\Tuple{\1L, \1A, \Pi, \1R}$, characterizing a \emph{causal reinforcement learning task} (Def.~\ref{def:_3_2_task}).
The goal of the agent is then to find a policy $\pi^*$ such that
\begin{align}\label{eq:opt-crl-v2}
	 & \pi^* = \argmax_{\pi \in \Pi} \invEE{ \1R \left (\*Y \right ) \mid \; \1A, \1L}{\pi}{\1M^*}
\end{align}
In words, the CRL agent aims to find an optimal policy $\pi^*$ within the policy space $\Pi$ that maximizes the reward $\1R$ when evaluated in the unknown environment $\1M^*$ while having assumptions about the environment  $\1A$ and access to data collected through a learning regime $\1L$. 

A summary of the signature of the tasks studied so far in this section is shown in the upper part of Table~\ref{tab:_4_4_roadmap}, serving as a grounding tool for the discussion here. 
For instance, each of the tasks accounts for a different dimension in terms of the task signature, as was previously discussed. 
Off-policy learning considers the more traditional offline modality where the NUC assumption is assumed to hold. In this case, traditional DP or IPW methods could be applied to leverage data collected under one regime -- collected under an observational policy -- to make inferences about another -- a new interventional policy. 
We also introduced online learning where the learning regime is interventional, and data is collected in an active manner by the agent. 
Causal assumptions are minimal in this case since the data precisely matches the inferential target. 
Finally, we studied a more nuanced case of offline learning called ``causal identification,'' which relies on more explicit causal knowledge that allows one to verify the NUC condition or evaluate the optimization given in Eq.~\ref{eq:opt-crl-v2}, even when unconfoundedness doesn't hold. 
These three modalities touch on different learning regimes and structural assumptions about the underlying $\M^*$.

\begin{table}[t]
  \setlength{\tabcolsep}{4.2pt}
  \centering
  \begin{tabular}{@{}p{0.2cm}p{3cm}|p{2cm}p{2.5cm}p{2cm}p{2cm}|p{1.5cm}@{}}
  \multicolumn{2}{c}{} & \multicolumn{4}{|c|}{\textbf{Signature}} & \\
  \toprule
  &\textbf{Task} &  \textbf{Learning \newline Regime} \newline ($\1L$) & \textbf{Structural \newline Assumptions} ($\1A$) & \textbf{Policy \newline Space} \newline ($\Pi$) & \textbf{Reward \newline Function} \newline ($\1R$) &  \textbf{Section} \\ \midrule
  1& Off-policy  \newline Learning & See &   \cellcolor{gray!25} NUC  & $\Pi_{\textsc{exp}}$ & $\D(\*Y) \mapsto \3R$ & \ref{sec:_4_1} \\ \midrule
  2& Online  \newline Learning  & \cellcolor{gray!25} Do & -  & $\Pi_{\textsc{exp}}$ & $\D(\*Y) \mapsto \3R$ & \ref{sec:_4_2} \\ \midrule
  3& Causal \newline Identification & See& \cellcolor{gray!25} DAG $\G$  &  $\Pi_{\textsc{exp}}$ & $\D(\*Y) \mapsto \3R$  & \ref{sec:_4_3} \\ \midrule \midrule
  4 &Offline-to-Online \newline Learning  & \cellcolor{gray!25} See + Do  &  - & $\Pi_{\textsc{exp}}$ & $\D(\*Y) \mapsto \3R$  & \ref{sec:_5_o2o} \\ \midrule
  5 & Where to do \newline \& What to see  & Do &  DAG $\G$  & \cellcolor{gray!25} $\Pi_{\textsc{mix}}$ & $\D(\*Y) \mapsto \3R$ & \ref{sec:_6_wheredo} \\ \midrule
  6 &Counterfactual \newline randomization & \cellcolor{gray!25} Ctf-Do &  - & $\Pi_{\textsc{ctf}}$ & $\D(\*Y) \mapsto \3R$ & \ref{sec:_7_ctf-rand}\\ \midrule
  7 &Causal Imitation \newline Learning  & See &DAG $\G$  &   $\Pi_{\textsc{exp}}$& \cellcolor{gray!25} - & \ref{sec:_8_imitation} \\ \bottomrule
  \end{tabular}
  \caption{Summary of causal reinforcement learning tasks investigated in this paper, in terms of their signatures and sections. We highlight in gray the most distinct feature introduced by the task. }
  \label{tab:_4_4_roadmap}
\end{table}

For the remainder of this paper, we will study natural and pervasive classes of learning tasks that do not fit into these existing modalities but involve novel dimensions and types of analysis relevant to real-world applications. Up next, we list some of these tasks that are also shown in Table~\ref{tab:_4_4_roadmap}:

\begin{enumerate}[itemindent=-\parindent, leftmargin=2cm, labelwidth=*, label=CRL \arabic*.] 
	\item \textbf{Causal Offline-to-Online Learning (COOL).} How can we pre-train an online agent to accelerate its learning process by leveraging imperfect knowledge about the effects of candidate policies obtained from confounded observational data? 
\\
	Computing the effect of candidate policies from observational data might be infeasible, as discussed in Secs.~\ref{sec:_4_1} and \ref{sec:_4_3}. On the other hand, it is also undesirable for the AI system  to rely solely on brute force, trial-and-error-based experimentation to improve its accuracy. 
 	How can we minimize the number of interventions the AI system makes by leveraging the invariances extrapolated from the causal model?
	In terms of the prototypical CRL agent depicted in Fig.~\ref{fig:_3_2_crl_agent}, the green line represents the online learning interactions, while the blue line represents the offline regime using observational data. 
	In Sec.~\ref{sec:_5_o2o}, we explore how to combine both modalities when the conditions of offline learning are provably not attainable, yet unlimited experimentation remains undesirable. 
		
	\item \textbf{Where to do and What to look for.}  Should an agent intervene in the environment to achieve its goal of bringing about a certain state of affairs? If so, where should the intervention take place?
The agent's objective is to learn an optimal policy from a collection of candidate policies, each encompassing different actions to intervene and input states to consider when determining these actions, including the null intervention (allowing the system to evolve naturally). 

As considered earlier, the agent has a fixed action space $\Pi_{\textsc{exp}}$ and tries to identify the intervention $\doo({X=x})$ that optimizes its reward measure. 
In Sec.~\ref{sec:_6_wheredo}, we explore the structure of the action space in complex systems, $\Pi_{\textsc{mix}}$, focusing on settings where each action plays a qualitatively different role. 
Practically, this challenge could arise when evaluating the effectiveness of drug combinations given the exponential growth in the total number of possible interactions. 
	
	\item \textbf{Counterfactual Decision-Making.} The agent makes a certain decision $\do{X = x}$ and wonders: would I be better off had I taken an alternative action, $\doo({X = x'})$?
	 The agent's objective is to evaluate this counterfactual statement to account for its natural and potentially biased decision-making process. 
	 
	 The previous settings considered an experimental policy scope, where typical Fisherian randomization eliminated the agent's natural inclinations. 
	 In Sec.~\ref{sec:_7_ctf-rand}, we expand the possibilities and allow for such introspective construct, which evokes a new learning regime \textit{Ctf-Do}. 
	 This new regime based on what we call counterfactual randomization will allow the agent to navigate through the larger scope of counterfactual policies, $\Pi_{\textsc{ctf}}$. In practice, this challenge could appear when evaluating adversarial settings where the agent's natural inclinations were leveraged to trick the agent and minimize its reward in a systematic fashion. 

	\item \textbf{Causal Imitation Learning.} Does perfectly mimicking an expert always lead to high decision-making performance? If not, under what conditions does imitation learning work? 
		The goal of the agent here is to learn an effective policy from the combination of observational data and a causal diagram when the reward function is not well-specified and unobserved confounding generally exists.

	The previous settings we investigated assumed that the reward function was known, which is not always the case. For instance, consider an autonomous vehicle trained from the observed trajectories of a human driver operating the vehicle. It is non-trivial to design a universal reward function evaluating the human's driving performance. How can we program the autonomous vehicle to operate effectively from the demonstration data without knowing the driver's performance measure? 
\end{enumerate}

This expanded set with new tasks and understanding paves the way to a broader view of counterfactual learning. It underscores the potential of studying causal inference and reinforcement learning side by side, a program we call \textit{causal reinforcement learning}.

\section{Causal Offline-to-Online Learning (CRL Task 1)}\label{sec:_5_o2o}
Learning algorithms introduced in the previous section rely exclusively on one type of interaction with the underlying environment, either through passive observation (``see''/offline) or direct intervention (``do''/online), despite their strong theoretical guarantees. 
A natural question that arises is whether the agent could combine both learning regimes and achieve better performance. 
This leads to the setting of offline-to-online learning. 
Existing offline-to-online methods in reinforcement learning literature \citep{taylor2009transfer,lazaric2012transfer,lee2022offline} rely on the NUC assumption (Def.~\ref{def:_4_1_nuc}), thus are not applicable when unobserved confounders generally exist in the observed data. 
We will relax the NUC assumption and first study \emph{causal offline-to-online learning} task (for short, COOL) from confounded observational data.
\footnote{More recently, there is growing interest in causal inference to identify treatment effects by combining observational and experimental datasets   \citep{bareinboim2012causal,lee2019gid}, and further estimating these under the NUC condition \citep{colnet2020causal,rosenman2020combining,cho2022robust,lin2023many,ball2023efficient}, or more general conditions \citep{jung2023joint,jung2023combination}. 
These works are orthogonal to offline-to-online learning since they focus on the offline setting where experimental data are provided in priori; the learner does not control the experiments. On the other hand, the key challenge in COOL is to use observational data to design randomized experiments.} 
This task was studies in \citep{zhang2017transfer,zhang2019near}, where several algorithms have been proposed. This section will summarize such results under a more unified CRL framework.

\begin{figure}[t]
	\centering
	\centering
	\begin{tikzpicture}
		\def\outerr{3.5}
		\def\innerr{3.5}
		\def\dist{3.5}

		\draw[->, >={Latex}] (-1.5,0) -- (3*\dist+2,0) node[below] {Episode t};

		\node[vertex, minimum width=6mm] (X1) at (-1, 2.2) {X\textsuperscript{(1)}};
		\node[vertex, minimum width=6mm] (Y1) at (1, 2.2) {Y\textsuperscript{(1)}};
		\draw[bidir] (X1) to [bend left = 45] (Y1);
		\draw[dir] (X1) -- (Y1);

		\node[vertex, minimum width=6mm] (X2) at (\dist-1, 2.2) {X\textsuperscript{(2)}};
		\node[vertex, minimum width=6mm] (Y2) at (\dist+1, 2.2) {Y\textsuperscript{(2)}};
		\draw[bidir] (X2) to [bend left = 45] (Y2);
		\draw[dir] (X2) -- (Y2);

		\node[vertex, minimum width=6mm] (X3) at (2*\dist-1, 2.2) {X\textsuperscript{(3)}};
		\node[vertex, minimum width=6mm] (Y3) at (2*\dist+1, 2.2) {Y\textsuperscript{(3)}};
		\node[regime] (p3) at (2*\dist-1, 3.2) {};
		\node[draw=none] (text) at (2*\dist-1.2, 3.5) {\scriptsize $\pi^{(3)}$};
		\draw[dir] (X3) -- (Y3);
		\draw[dir] (p3) -- (X3);

		\node[vertex, minimum width=6mm] (X4) at (3*\dist-1, 2.2) {X\textsuperscript{(4)}};
		\node[vertex, minimum width=6mm] (Y4) at (3*\dist+1, 2.2) {Y\textsuperscript{(4)}};
		\node[regime] (p4) at (3*\dist-1, 3.2) {};
		\node[draw=none] (text) at (3*\dist-1.2, 3.5) {\scriptsize $\pi^{(4)}$};
		\draw[dir] (X4) -- (Y4);
		\draw[dir] (p4) -- (X4);

		\node (d1) at (0, 1.4) {$\*V^{(1)} \sim P\left(\*V\right)$};
		\node (d2) at (\dist, 1.4) {$\*V^{(2)} \sim P\left(\*V\right)$};
		\node (d3) at (2*\dist, 1.4) {$\*V^{(3)} \sim \inv{\*V}{\pi^{(3)}}$};
		\node (d4) at (3*\dist, 1.4) {$\*V^{(4)} \sim \inv{\*V}{\pi^{(4)}}$};

		\draw[very thick, betterblue, -] (0, 0) -- (0, 0.2);
		\draw[very thick, betterblue, -] (\dist, 0) -- (\dist, 0.2);
		\draw[very thick, bettergreen, -] (2*\dist, 0) -- (2*\dist, 0.2);
		\draw[very thick, bettergreen, -] (3*\dist, 0) -- (3*\dist, 0.2);

		\node [below] at (0, -0.05) { 0};
		\node [below] at (\dist, -0.05) { 1};
		\node [below] at (2*\dist, -0.05) { 2};
		\node [below] at (3*\dist, -0.05) { 3};

		\node [fill=betterblue!45] at (0, 0.6) {see};
		\node [fill=betterblue!45] at (\dist, 0.6) {see};
		\node [fill=bettergreen!45] at (2*\dist, 0.6) {\small $\doo\left(\pi^{(3)} \right)$};
		\node [fill=bettergreen!45] at (3*\dist, 0.6) {\small $\doo\left(\pi^{(4)} \right)$};

		\begin{pgfonlayer}{back}
			\node[circle,fill=betterblue!65,draw=none,minimum size=2*\innerr mm] at (X1) {};
			\node[circle,fill=betterred!65,draw=none,minimum size=2*\innerr mm] at (Y1) {};
			\node[circle,fill=betterblue!65,draw=none,minimum size=2*\innerr mm] at (X2) {};
			\node[circle,fill=betterred!65,draw=none,minimum size=2*\innerr mm] at (Y2) {};
			\node[circle,fill=betterblue!65,draw=none,minimum size=2*\innerr mm] at (X3) {};
			\node[circle,fill=betterred!65,draw=none,minimum size=2*\innerr mm] at (Y3) {};
			\node[circle,fill=betterblue!65,draw=none,minimum size=2*\innerr mm] at (X4) {};
			\node[circle,fill=betterred!65,draw=none,minimum size=2*\innerr mm] at (Y4) {};
		\end{pgfonlayer}
	\end{tikzpicture}
	\caption{Temporal diagram showing an offline-to-online learning agent interacting with the environment for repeated episodes.}
	\label{fig:_5_o2o}
\end{figure}
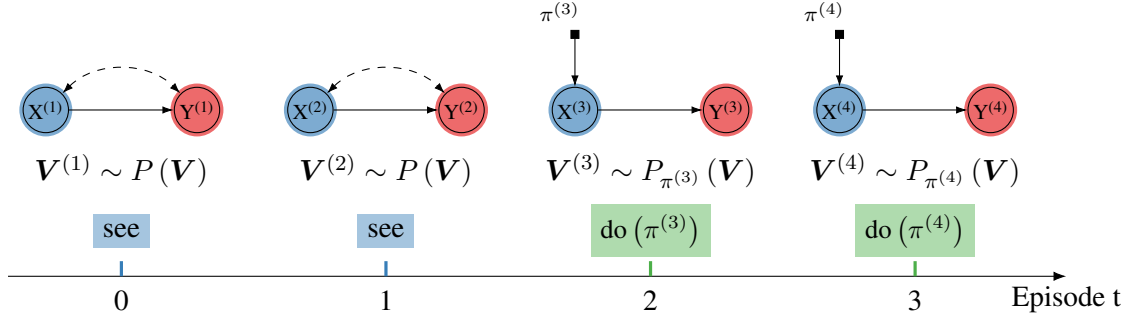

The mechanism of how the CRL agent operates in this task and switches from an offline to an online mode when interacting with the underlying environment is illustrated in  Fig.~\ref{fig:_5_o2o}. 
Specifically, the CRL agent first passively observes the environment for a number of episodes $t = 1, \dots, n$, and receive the observational samples $\*V^{(t)} \sim P(\*V)$. For episode $t = n + 1, \dots, n + T$, the agent then picks a policy $\pi^{(t)}$, directly intervenes $\doo\Parens{\pi^{(t)}}$ in the environment, and receives subsequent outcomes $\*V^{(t)} \sim \inv{\*V}{\pi^{(t)}}$. 
The agent will leverage the observational data $\Braces{\*V^{(1)}, \dots, \*V^{(n)}}$ to accelerate the future online learning process. The following task signature characterizes this offline-to-online learning setting:
\begin{align}
	\1T_{\text{off+on}} = \Tuple{\1I = \{\text{see}, \text{do}\}, \1A= \emptyset, \Pi = \Braces{\Tuple{ X_i, \*S_i } }_{X_i \in \*X}, \1R = \D(\*Y) \mapsto \3R }.
\end{align}
This means that the agent will try to find a policy $\pi^*$ such that
\begin{align}\label{eq:opt-crl-off+on}
	 & \pi^* = \argmax_{\pi \in \Pi} \invEE{ \1R \left (\*Y \right )  \;\bigg\vert  \;\textcolor{red}{\mathcal{D}_{\text{obs}} \sim P(\*V), \; \mathcal{D}_{\text{exp}} \sim \inv{\*V}{\*x} }}{\pi}{\1M^*}, 
\end{align}
where the distinct feature here is the combination of observational and interventional interactions. 

In order to make this argument more precise, we will describe an online-to-offline strategy that combines the observational data with the learning process of \texttt{UCB} algorithm (Alg.~\ref{alg:_4_2_ucb}), provided that the NUC assumption (Def.~\ref{def:_4_1_nuc}) holds. 
Consider an MAB model $\Tuple{ \1M^*, \Braces{\Tuple{X, \emptyset}}, Y }$ graphically described in Fig.~\ref{fig:_3_1_mab}. Let $\1D_{\text{obs}} = \Braces{X^{(i)}, Y^{(i)} }_{i = 1}^n$ be i.i.d. samples drawn from the observational distribution $P(X, Y)$. For \texttt{UCB} algorithm allocating an arm at episode $t$, let $\1D_{\text{exp}}^{(t)} = \Braces{X^{(n+i)}, Y^{(n+i)}}_{i = 1}^{t-1}$ be the experimental data collected by \texttt{UCB} up to episode $t$. By combining the observational data $\1D_{\text{obs}}$ and experimental data $\1D_{\text{exp}}^{(t)}$, we define the empirical reward estimate for every arm $x$ as follows:
\begin{align}
	\hat{\E}^{(n + t)}_x[Y] = \frac{1}{N_{n+t}(x)} \left (  \sum_{i = 1}^{n} Y^{(i)} \I \left \{X^{(i)} = x \right \} + \sum_{i = 1}^{t-1} Y^{(n + i)} \I \left \{X^{(n + i)} = x \right \} \right) \label{eq:_5_empirical}
\end{align}
where $N_{n+t}(x) = \sum_{i = 1}^{n+t-1} \I\{X^{(i)} = x\}$ is the total occurrence of observing arm $x$ being played in the combined dataset $\1D_{\text{obs}} \cup \1D_{\text{exp}}^{(t)}$. The augmented upper confidence bound for an arm $x$ by combining the observational and interventional data is given by
\begin{align}
	\text{UCB}_{n+t}(x, \delta) = \hat{\E}^{(n+t)}_x[Y] + \sqrt{\frac{\log(1/\delta)}{2N_{n+t}(x)}}  \label{eq:_5_ucb1}
\end{align}
The augmented \texttt{UCB} algorithm directly transfers observational data as if they were obtained from direct interventions. 
We summarize in Alg.~\ref{alg:_5_ucb_direct} details of a direct online-to-offline transfer strategy using standard off-policy learning methods, which we call \texttt{UCB}\textsuperscript{-}. 
For every episode $t$, it computes an upper confidence bound $\text{UCB}_{n+t}(x, \delta)$ for every arm $x$, plays an arm with the most significant confidence bound, and observed subsequent reward.

We will analyze the performance of \texttt{UCB}\textsuperscript{-} and show that the direct transfer strategy could accelerate the \texttt{UCB}'s performance, under the NUC assumption (Def.~\ref{def:_4_1_nuc}). 
Suppose the NUC condition holds in the MAB model $\Tuple{ \1M^*, \Braces{\Tuple{X, \emptyset}}, Y }$. Applying Thm.~\ref{thm:_4_1_dp} we compute the expected reward of every arm $x \in \D(X)$ from the observational distribution $P(X, Y)$,
\begin{align}
	\invE{Y}{x} = \E\left [Y \mid x \right] \label{eq:_5_mab1}
\end{align}
Recall that for any policy $\pi(X)$, the NUC holds in the intervened model $\Tuple{ \1M^*_{\pi}, \Braces{\Tuple{X, \emptyset}}, Y }$. The expected reward of arm $x$ is computable from the interventional distribution $\inv{X, Y}{\pi}$ as
\begin{align}
	\invE{Y}{x} = \invE{Y\mid x}{\pi} \label{eq:_5_mab2}
\end{align}
The above estimation formulas allow the agent to evaluate the effects of arms by pooling the observational and experimental data. Eq.~\ref{eq:_5_empirical} provides consistent estimates for the expected rewards $\invE{Y}{x}$ provided with the NUC assumption. 
When sufficient observations are provided, \texttt{UCB}\textsuperscript{-} will immediately identify the optimal arm; only a few episodes of online interventions are required. 
Broadly, when the NUC assumption holds, the learner could consistently evaluate candidate policies from the observational data using standard off-policy learning methods. These pre-trained estimations could then be directly transferred to ``warm-start'' the future online learning process. \footnote{Sec.~\ref{sec:_4_1} provides a more detailed discussion about the NUC condition and off-policy learning algorithms.}

\begin{algorithm}[t]
	\caption{Upper Confidence Bound in Direct Offline-to-Online Transfer (\texttt{UCB}\textsuperscript{-}) }
	\label{alg:_5_ucb_direct}
	\setlength{\textfloatsep}{0pt}
	\begin{algorithmic}[1]
		\State {\bfseries Input:} a policy space $\Pi = \Braces{\Tuple{X, \emptyset}}$, observational data $\1D_{\text{obs}} = \left \{ X^{(i)}, Y^{(i)} \right\}_{i = 1}^n$
		\ForAll{episodes $t = 1, 2, \dots $}
		\State Choose an arm
		\begin{align}
X^{(n+t)} = \argmax_{x \in \D(X)} \text{UCB}_{n+t}(x, \delta), \text{ where } \delta = t^{-4}
		\end{align}
		\State Perform $\doo(X^{(n+t)})$ for episode $t$ and receive reward $Y^{(n+t)}$.
		\EndFor
	\end{algorithmic}
\end{algorithm}

\begin{figure}[t]
	\centering
	\null
	\begin{minipage}[b]{0.32\linewidth}
		\centering
		\includegraphics[width=1.0\textwidth]{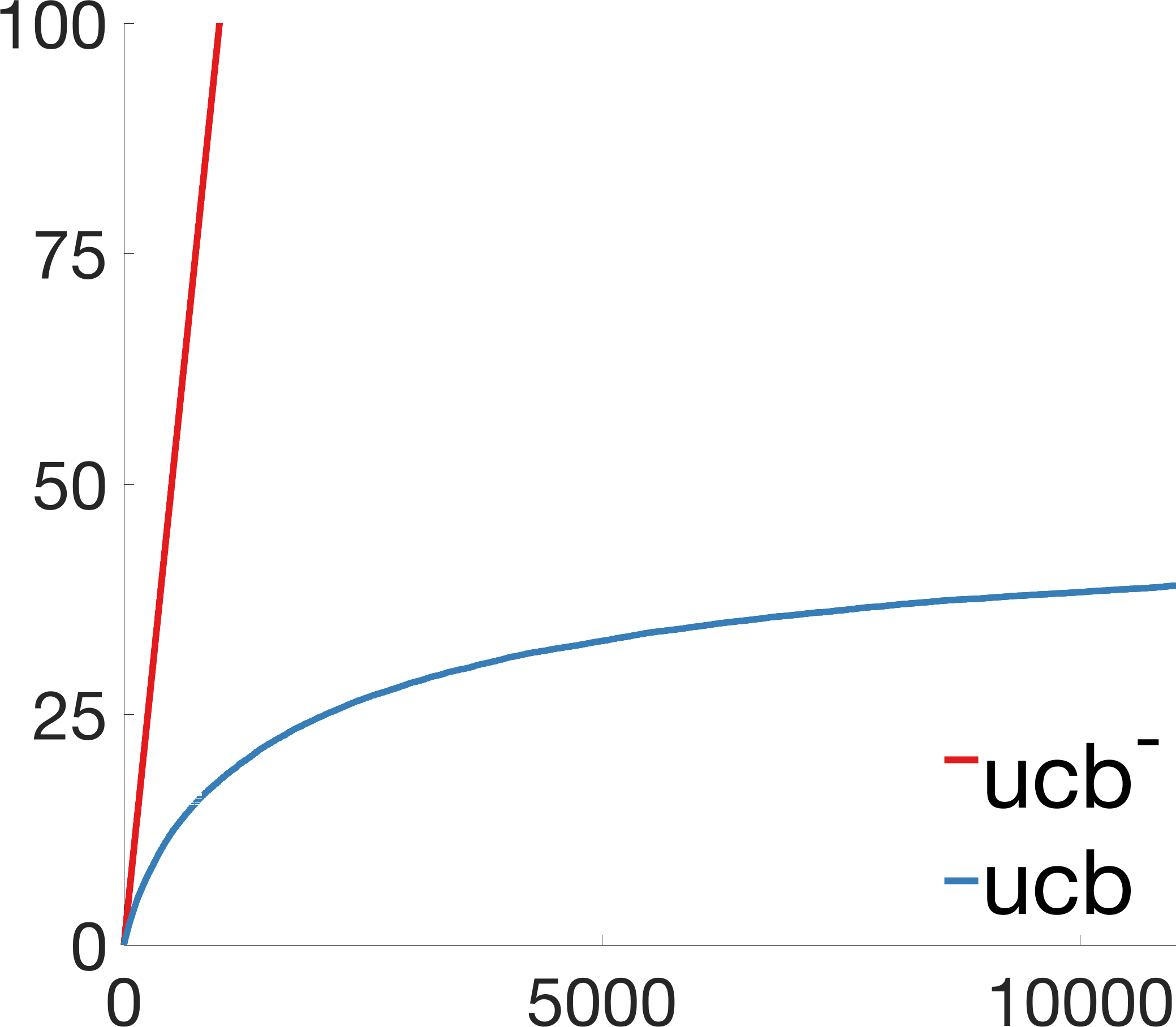}
		\subcaption{Cumulative Regret}
		\label{fig:_5_mab1_a}
	\end{minipage}\hfill
	\begin{minipage}[b]{0.31\linewidth}
		\centering
		\includegraphics[width=1.0\textwidth]{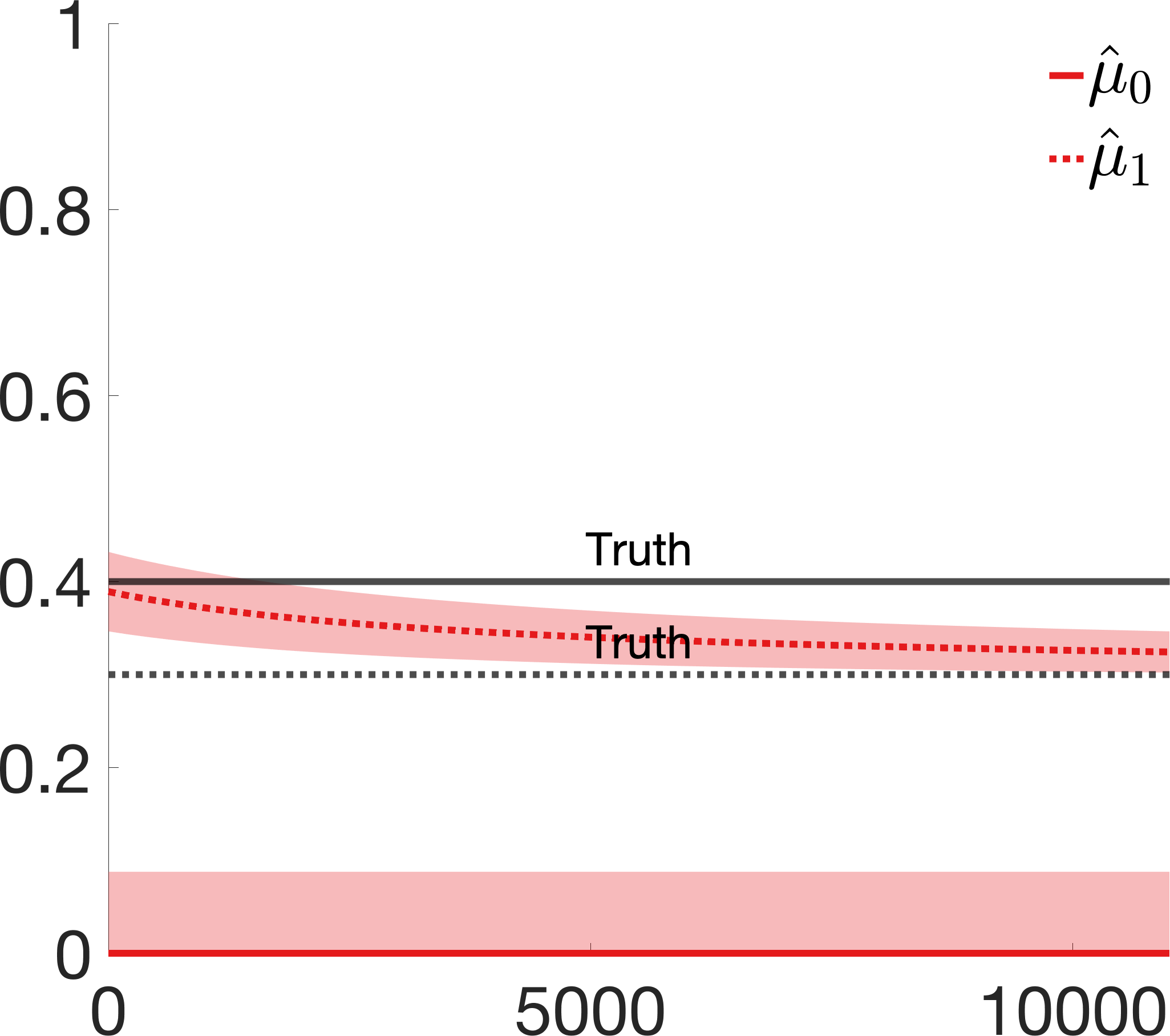}
		\subcaption{\texttt{UCB}\textsuperscript{-}}
		\label{fig:_5_mab1_b}
	\end{minipage}\hfill
	\begin{minipage}[b]{0.31\linewidth}
		\centering
		\includegraphics[width=1.0\textwidth]{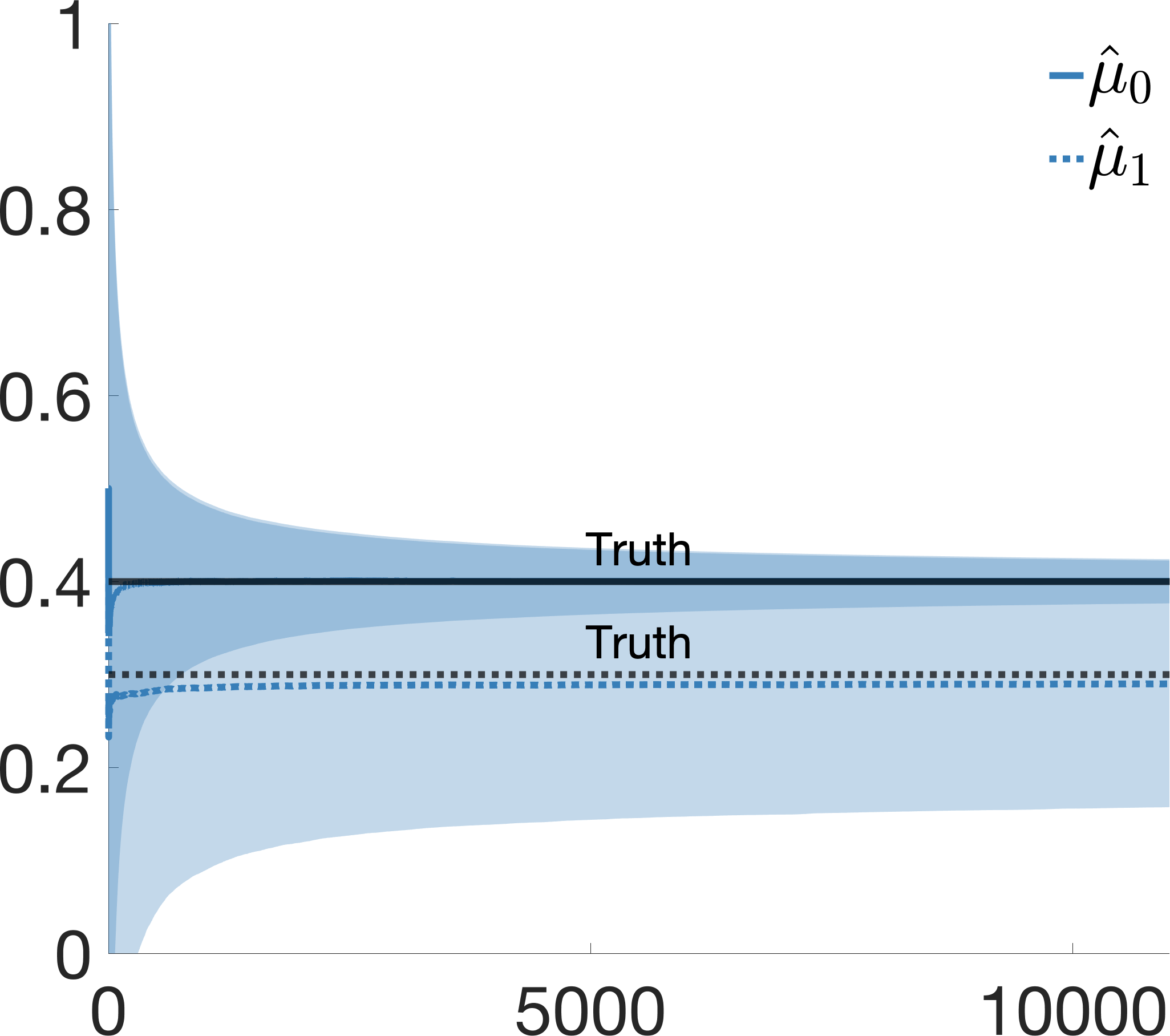}
		\subcaption{\texttt{UCB}}
		\label{fig:_5_mab1_c}
	\end{minipage}\hfill\null
	\caption{Simulation results comparing \texttt{UCB} learner with direct transfer of observational data (\texttt{UCB}\textsuperscript{-}) and standard \texttt{UCB} without any prior observations.}
	\label{fig:_5_mab1}
\end{figure}

On the other hand, the NUC assumption could be fragile and does not hold in many practical applications. For this reason, it is wise to evaluate the performance and robustness of \texttt{UCB}\textsuperscript{-} when the NUC assumption doesn't hold, which we do in the following experiment.
\begin{experiment}\label{exp:_5_mab1}
	Fig.~\ref{fig:_5_mab1_a} shows the cumulative regret of \texttt{UCB}\textsuperscript{-} in the MAB environment $\1M^*$ described in Example~\ref{exp:_2_1_mab} with the suboptimal gap $\Delta = 0.1$, taking as input $5,000$ observational samples drawn from the distribution $P(X, Y)$. The NUC assumption does not hold in this model due to the unobserved confounder $U$ affecting action $X$ and reward $Y$ simultaneously. As a baseline, we also include a vanilla \texttt{UCB} starting from scratch, which does not utilize any prior observations. One can see by inspection the significant disparity between the performance of \texttt{UCB} (blue) and \texttt{UCB}\textsuperscript{-} (red).

We show in Fig.~\ref{fig:_5_mab1_b} the empirical estimates $\hat{\mu}_x$ of the expected rewards $\invE{Y}{x}$ computed by \texttt{UCB}\textsuperscript{-}; shaded areas represent confidence intervals evaluated at $95\%$ percentile. 
For comparison, Fig.~\ref{fig:_5_mab1_c} shows the empirical reward estimates computed by the standard \texttt{UCB} without using prior observations. 
Simulation results demonstrate a significant bias in the reward estimation of \texttt{UCB}\textsuperscript{-}, favoring the suboptimal arm $x = 1$. 
This bias was not fully corrected until the end of the online learning process ($T = 10,000$). 
On the other hand, \texttt{UCB} is able to obtain accurate estimations of the expected rewards after a few episodes of interventions and identify the optimal arm $x^* = 0$. \hfill $\blacksquare$
\end{experiment}

\begin{table}[h]
	\centering
	\renewcommand{\arraystretch}{1.25}
	\begin{tabular}{|c|c|c|c|}
		$X$     & $\E[Y|X = x]$      & $\invE{Y}{X \gets x}$ & Causal Bound                   \\
		\hline
		\hline
		$x = 0$ & $0$                & $0.4$                 & $[0, 0.8]$                     \\
		$x = 1$ & $0.5 - 1.25\Delta$ & $0.4 - \Delta$        & $[0.4 - \Delta, 0.6 - \Delta]$ \\
	\end{tabular}
	\caption{Evaluations of $\E[Y|x]$ and $\invE{Y}{x}$ in MAB environment $\1M^*$ defined in Example~\ref{exp:_2_2_mab}.}
	\label{tab:_5_mab}
\end{table}

The above example suggests that in MAB models where unobserved confounders exist, the NUC condition does not hold, which implies that directly transferring observational data may introduce a significant bias into the empirical estimation of arms' expected rewards. 
This, in turn, slows down the learning process of online algorithms, and in some cases may even hinder these algorithms' convergence. 
In order to confirm this intuition and further explain the negative transfer phenomenon, we compute the expected reward $\E\left[ Y \mid x \right]$ conditioning on the event that the learner observes arm $X = x$ is played in the MAB environment $\1M^*$ defined in Example~\ref{exp:_2_2_mab}. 
We also compute the expected reward $\invE{Y}{x}$ induced by the learner playing an arm $\doo(X \gets x)$ in $\1M^*$. The analytical results are summarized in Table~\ref{tab:_5_mab}. 
One can see by inspection that the evaluations of observed expected rewards $\E\left[Y \mid x\right]$ differ significantly from the interventional expected rewards $\invE{Y}{x}$. This means that the identification formula in Eq.~\ref{eq:_5_mab1} does not apply due to unobserved confounding between the action $X$ and reward $Y$, which makes some arms $x$ appear observationally more effective than they interventionally are. When the suboptimal gap $\Delta >0$, optimizing the observed reward $\E\left[Y \mid x\right]$ leads to a suboptimal arm $x = 1$. On the other hand, the optimal arm $x^* = 0$ maximizes the interventional reward $\invE{Y}{x}$ in the underlying MAB environment.

Broadly, off-policy estimation methods may fail to recover the unknown expected rewards of candidate policies without the NUC assumption. 
Naively transferring estimated rewards introduces inaccuracies in optimal policy estimation of the online learning algorithm, resulting in a negative impact on its performance. 
Moreover, since the effects of interventions are never measured (before the online learning stage starts), the learner could not detect biases arising from the off-policy evaluation step based on observational data. 
\footnote{On the other hand, whenever the agent goes online, a sufficient test would entail evaluating whether $P_{x}(Y) = P(Y | X = x)$, 
which is known as marginal ignorability or no-confounding conditions \cite[Ch.~6]{pearl:2k}; see also  \cite[Def.16(iii)]{bareinboim2020pearl}. In practice, the finite-sample version of such test has to be evaluated. } 
This implies to significant challenges in offline-to-online learning when the NUC does not generally hold. 
One may surmise, therefore, that the learner should start the online learning process from scratch without utilizing past observations of the environment, however abundant they are. 
\footnote{This is, of course, against human experience where we learn by observing other agents interacting, even when our perceptions and models of the world do not fully match. Here, we can see that one bit difference between the input of the behavioral agent versus the agent who is using the data can lead to a catastrophic behavior.   }

This section aims to show that this is not the case and overcome the challenges outlined by the confounded situation as described above. 
We will study the problem of causal offline-to-online learning (COOL), which accelerates online reinforcement learning by leveraging offline observational data. 
We focus on the settings where the NUC condition does not hold, and the expected rewards of candidate policies are not computable from the observational data.\footnote{Of course, whenever NUC holds, this would be a trivial, special case for the approach discussed here. } 
Directly applying off-policy evaluation could lead to significant bias in the reward estimation, harming the online learning process instead. More specifically, the remainder section is divided as follows.
\begin{itemize}
	\item Sec.~\ref{sec:_5_1} introduces a novel causal offline-to-online learning strategy in MAB models and proves that it consistently dominates standard \texttt{UCB} algorithm in term of performance. It utilizes bounds to evaluate unknown expected rewards from the observational data, which are then incorporated to accelerate the online learning process.
	\item Sec.~\ref{sec:_5_2} generalizes \texttt{UCB} algorithm to general CDMs (beyond MABs) where the agent needs to decide on a sequence of actions based on values of corresponding states at the time of intervention. This algorithm achieves the near-optimal regret bound without additional observational data and structural knowledge about the underlying environment.
	\item Sec.~\ref{sec:_5_3} derives novel bounds capable of exploiting observational data to infer underlying interventional transitional probabilities and the reward functions. These bounds are then incorporated, in a systematic way, to accelerate online learning in an arbitrary CDM.
\end{itemize}

\subsection{Confounding Robust Offline-to-Online Learning}\label{sec:_5_1}
This section studies the offline-to-online learning in MAB models when the NUC assumption does not hold and standard off-policy learning algorithms, including IPW (Thm.~\ref{thm:_4_1_ipw}) and DP (Thm.~\ref{thm:_4_1_dp}) estimation, do not apply. Causal researchers may wonder if it is possible to estimate the expected rewards of arms from the observational data using causal identification algorithms, e.g., do-calculus learning (Def.~\ref{def:_4_3_do-calculus}). Indeed, it has been shown that the expected rewards are not identifiable in MAB environments without additional assumptions. The following corollary could be derived based on the formal definition of identifiability described in Def.~\ref{def:_4_3_id}.
\begin{corollary}[Non-Identifiability]\label{corol_5-non-id}
	Consider endogenous variables $\*X, \*Y \subseteq \*V$ and let $\pi$ be a policy over $\*X$. The interventional (policy) distribution $\inv{\*Y}{\pi}$ is not identifiable from structural assumptions $\1A$ and observational distribution $P(\*V$ if there exist two SCMs $\1M_1, \1M_2$ compatible with $\1A$ such that $P(\*V; \1M_1) = P(\*V; \1M_2) > 0$ while $P_{\pi}(\*Y;\1M_1) \neq P_{\pi}(\*Y;\1M_2)$. \hfill $\blacksquare$
\end{corollary}
In words, the expected rewards $\invE{Y}{x}$ of arms $x$ are not identifiable in MAB models if there exist two MAB environments that generate the same observational distribution $P(X, Y)$, but differ in the expected rewards $\invE{Y}{x}$. This means the agent could not uniquely determine the expected rewards of arms from the observational distribution alone. Our next example demonstrates this non-identifiability result in MAB models.
\begin{example}\label{exp:_5_1_mab1}
	Consider an MAB environment $\1M'$ described by an SCM
	\begin{align}
		\M' = \langle \*U = \{U_1, U_2\}, \*V =  \{X, Y\}, \2F', P(U_1, U_2)\rangle,
	\end{align}
	The causal mechanisms are the following:
	\begin{align}
		\2F' = \begin{cases}
			       X \gets \I\{U_1 < 0.8\}, \\
			       Y \gets \I\{U_2 < 0.5 - 1.25\Delta \} \times X
		       \end{cases} \label{eq:_5_1_mab1}
	\end{align}
	where coefficient $\Delta$ is a real number bounded in $(0, 0.5)$; and $P(U_1, U_2)$ is such that $U_1, U_2$ are independent variables drawn from a uniform distribution $\texttt{Unif}(0, 1)$. It is verifiable that $\1M'_{\textsc{mab}}$ defines the same observational distribution $P(X, Y)$ as the MAB environment $\1M^*$ defined in Example~\ref{exp:_2_2_mab}. First, marginal probabilities $P(X = 0) = 0.2$ and $P(X = 1) = 0.8$ in $\1M'_{\textsc{mab}}$ since $U_1$ is uniformly drawn from the real interval $[0, 1]$. Evaluating the conditional distribution $P(Y|X)$ in $\1M'$ gives
	\begin{align}
		P(Y = 1 \mid X = 0) & = P(\I\{U_2 < 0.5 - 1.25\Delta \} \times 0 = 1 \mid X = 0) \notag \\
		                    & = 0
	\end{align}
	Similarly, the recovery rate conditioning on event $X = 1$ is given by
	\begin{align}
		P(Y = 1 \mid X = 1) & = P(U_2 < 0.5 - 1.25\Delta) \notag \\
		                    & = 0.5 - 1.25\Delta.
	\end{align}
	On the other hand, $\1M'$ defines different expected rewards for interventions $\doo(X \gets x)$ from that defined by $\1M'$. More precisely, the submodel $\M_{x}$ induced by $\doo(X \gets x)$ is a tuple
	\begin{eqnarray}
		\M'_{x} =  \langle  \*U = \{U_1, U_2\}, \*V = \{X, Y\}, \2F'_{x}, P(U_1, U_2)\rangle,
	\end{eqnarray}
	where the structural functions $\2F'_x$ is defined as
	\begin{align}
		\2F'_x = \begin{cases}
			         X \gets x, \\
			         Y \gets \I\{U_2 < 0.5 - 1.25 \Delta\} \times X
		         \end{cases} \label{eq:_5_1_mab2}
	\end{align}
	Evaluating the expected reward $Y$ in submodel $\M'_{X \gets 0}$ described in Eq.~\ref{eq:_5_1_mab2} gives
	\begin{align}
		\invE{Y}{X \gets 0} & = \E \left[\I\{U_2 < 0.5 - 1.25 \Delta\} \times 0 \right] \notag \\
		                    & = 0 \label{eq:_5_1_mab3}
	\end{align}
	Similarly, the expected reward of playing an arm $\doo(X \gets 1)$ is equal to
	\begin{align}
		\invE{Y}{X \gets 1} & =  P(U_2 < 0.5 - 1.25\Delta) \notag      \\
		                    & = 0.5 - 1.25\Delta.  \label{eq:_5_1_mab4}
	\end{align}
	For detailed computations of $P(X, Y)$ and  $\invE{Y}{x}$ in MAB model $\1M^*$, revisit Examples~\ref{exp:_2_2_mab} and \ref{exp:_2_2_mab3}. To sum up, MAB models $\M'$ and $\M^*$ are both compatible with the causal diagram $\G_{\textsc{mab}}$ of Fig.~\ref{fig:_3_1_mab} such that they define the same observational distribution $P(X, Y)$ while differ significantly in the expected reward $\invE{Y}{x}$. This means the expected rewards of arms $x$ are not identifiable from the observational distribution $P(X, Y)$ in MAB models. \hfill $\blacksquare$
\end{example}
The result so far seems to suggest that when unobserved confounders exist and no additional causal knowledge is provided, no prior observations could be useful in evaluating the expected rewards of arms in MAB models. However, we will show this is not the case by deriving bounds over the unknown expected rewards from the observational data, which we call \emph{causal bounds}. This means that while it is infeasible to determine the values of the non-identifiable expected reward, the learner could still extrapolate partial knowledge from the observational data to improve the estimates of its feasible region. For MAB models with an arm choice $X$ and a reward $Y$, the seminal results in \citep{manski:90,robins:89b} allow the derivation of informative causal bounds (to be defined) containing the expected reward of an arm $x$ from the observational distribution.
\begin{theorem}[Natural Bounds \citep{manski:90}]\label{thm:_5_1_nb}
	For any SCM $\1M^*$ containing an action $X$ and a reward $Y$, let the domain of $X$ be discrete and finite, and $Y$ be bounded in the real interval $[0, 1]$. The expected reward for any arm $x$ is bounded in $\invE{Y}{x} \in \left[ l_x, r_x \right]$ where
	\begin{align}
		 & l_x = \underbrace{\E\left[Y \mid x \right ] P(x)}_{observational}, &  & r_x = \underbrace{\E\left[Y \mid x \right ] P(x) + 1 - P(x) }_{observational}\label{eq:_5_1_nb}
	\end{align} \hfill $\blacksquare$
\end{theorem}
The lower and upper bounds in Eq.~\ref{eq:_5_1_nb} are both functions of the observational distribution $P(X, Y)$ and are, therefore, estimable from the observational data. 
The above bounds are informative and strictly contained in the interval $[0, 1]$ when marginal probabilities $P(x) > 0$ are positive for any arm $x \in \D(X)$. 
The natural bounds have been proved to be optimal in MAB models \citep{zhang2017transfer,zhang2021bounding}, i.e., they cannot be improved without additional assumptions and data. 
\begin{example}\label{exp:_5_1_nb}
	Consider the MAB environments $\M^*, \M'$ described in Examples~\ref{exp:_2_1_mab} and \ref{exp:_5_1_mab1}, respectively. Applying Thm.~\ref{thm:_5_1_nb} we obtain a lower bound contained the expected reward $\invE{Y}{X \gets 0}$ computed from the observational distribution $P\left(X, Y\right)$ as
	\begin{align}
		\invE{Y}{X \gets 0} & \geq \E \left [Y \mid X = 0\right] P(X = 0) \notag \\
		                    & = 0.
	\end{align}
	The upper bound over the expected reward for arm $x = 0$ is given by:
	\begin{align}
		\invE{Y}{X \gets 0} & \leq \E \left [Y \mid X = 0\right] P(X = 0) + 1 - P(X = 0)\notag \\
		                    & \leq 0.8
	\end{align}
	Similarly, we could also obtain a natural bound for the expected reward of arm $x = 1$ from the observational distribution $P(X, Y)$ and is given by
	\begin{align}
		\invE{Y}{X \gets 1} & \geq \E \left [Y \mid X = 1\right] P(X = 1) \notag \\
		                    & = (0.5 - 1.25\Delta) \times 0.8 \notag             \\
		                    & = 0.4 - \Delta.
	\end{align}
	and the upper bound implies
	\begin{align}
		\invE{Y}{X \gets 1} & \leq \E \left [Y \mid X = 1\right] P(X = 1) + 1 - P(X = 1)\notag \\
		                    & \leq 0.4 - \Delta + P(X = 0) \notag                              \\
		                    & \leq 0.6 - \Delta
	\end{align}
	We summarize in Table~\ref{tab:_5_mab} natural bounds computed from the observational distribution $P(X, Y)$. The results support the soundness of the natural bound in Thm.~\ref{thm:_5_1_nb} in MAB models since it contains the real expected rewards $\invE{Y}{x}$ evaluated in both $\M^*$ and $\M'$. \hfill $\blacksquare$
\end{example}
The causal bounds $\invE{Y}{x} \in \left [l_x, r_x \right ]$ could be used to improve the estimate of the upper confidence bound assigned to every arm $x$ during online learning. More precisely, for \texttt{UCB} algorithm selecting an arm at episode $t$, the causally-clipped upper confidence bound for arm $x$ is defined as
\begin{align}
	\overline{\text{UCB}}_t(x, \delta) = \min \left \{ \max \left \{\text{UCB}_t(x, \delta), l_x \right\}, r_x \right\}. \label{eq:_5_1_ucb}
\end{align}
Among quantities in the above equation, $\text{UCB}_t(x, \delta)$ is the standard upper confidence bound for MAB models defined in Eq.~\ref{eq:_4_2_ucb2}. The clipping ensures the new upper bound $\overline{\text{UCB}}_t(x, \delta) \in [l_x, r_x]$ is contained in the causal bound. In words, the causal bound for an arm $x$ always takes priority when it is incompatible with the confidence bound computed from the experimental data collected from online learning to episode $t$. The incompatibilities generally arise at the beginning of the learning process ($t$ is small) where the standard upper bound $\text{UCB}_t(x, \delta)$ is loose. It eventually converges between the causal bound $[l_x, r_x]$ as more experimental data $N_t(x)$ are collected.
The algorithm consistently prefers causal bounds since the observational data is often abundant while conducting interventions is expensive; causal bounds in Thm.~\ref{thm:_5_1_nb} are valid and could be accurately estimated with sufficient observational data.

Alg.~\ref{alg:_5_1_ucb+} summarizes the augmented \texttt{UCB} algorithm incorporating causal bounds computed from the observational data, which we call \texttt{UCB}\textsuperscript{+}. It takes as input arguments causal bounds $\invE{Y}{x} \in \left[l_x, r_x\right]$ over the expected rewards for candidate arms $x$. For every episode $t$, it computes the clipped upper bound $\overline{\text{UCB}}_{n+t}(x, \delta)$ for every arm $x$ by combining the experimental data collected up to episode $t$ and the causal bound $\left [l_x, r_x \right]$. Finally, the agent plays an arm with the highest clipped confidence bound and receives a subsequent reward.

\begin{algorithm}[t]
	\caption{Upper Confidence Bound combined with Causal Bounds in MAB (\texttt{UCB}\textsuperscript{+}) }
	\label{alg:_5_1_ucb+}
	\setlength{\textfloatsep}{0pt}
	\begin{algorithmic}[1]
		\State {\bfseries Input:} a policy space $\Pi = \Braces{\Tuple{X, \emptyset}}$, causal bounds $\invE{Y}{x} \in \left[l_x, r_x\right]$.
		\ForAll{episodes $t = 1, 2, \dots $}
		\State Choose an arm
		\begin{align}
X^{(t)} = \argmax_{x \in \D(X)} \overline{\text{UCB}}_{n+t}(x, \delta), \text{ where } \delta = t^{-4}.	
		\end{align}
		\State Perform $\doo(X^{(t)})$ for episode $t$ and receive reward $Y^{(t)}$.
		\EndFor
	\end{algorithmic}
\end{algorithm}

\begin{restatable}{theorem}{thmucbandit}\label{thm:_5_1_ucb}
	For any MAB model $\langle \1M^*, \Set{\Tuple{X, \emptyset}}, Y \rangle$, let $Y$ be the reward variable with support on $[0, 1]$ and let the domain of action $X$ be $\D(X) = \{1, \dots, K\}$. It holds the regret of \texttt{UCB}\textsuperscript{+} in SCM $\1M^*$ after $T > 1$ episodes is bounded by
	\begin{align}
		R(T, \1M^*) \leq  8 \sum_{x: \substack{\Delta_x > 0 \\r_x \geq \mu_{x^*}}} \frac{\log(T)}{\Delta_x} + \left( 1 + \frac{\pi^2}{3} \right) \sum_{x: \Delta_x > 0} \Delta_x \label{eq:_5_1_ucb2}
	\end{align} \hfill $\blacksquare$
\end{restatable}
Let $\D(X)^- = \left \{x \in \D(X) \mid \Delta_x > 0 \right\}$ be the set of suboptimal arms. Let $\D(X)^*$ be the set of suboptimal arms such that the causal upper bound $r_x$ for every arm $x$ is larger than or equal to the optimal expected reward $\mu_{x^*} = \invE{Y}{x^*}$, i.e., $\D(X)^* = \left \{x \in \D(X) \mid \Delta_x > 0, r_x \geq \mu^*\right\}$. Since $\D(X)^* \subseteq \D(X)^-$, the regret bound of Thm.~\ref{thm:_5_1_ucb} consistently dominates the regret bound of the standard \texttt{UCB} in Thm.~\ref{thm:_4_2_ucb}. When there are some suboptimal arms $x$ with $r_x < \mu_{x^*}$, the augmented \texttt{UCB}\textsuperscript{+} is able to outperform \texttt{UCB} by utilizing quantitative knowledge extrapolated from the observational data.\footnote{Another line of popular bandit algorithm is called Thompson sampling (\texttt{TS}, \cite{thompson1933likelihood,chapelle2011empirical,agrawal2012analysis}). We show in Appendix~\ref{app:_ts} that causal bounds could also be utilized to accelerate the convergence of the \texttt{TS} algorithm.}

To illustrate, assume the total number of arms $K = 2$ and $x = 0$ is the optimal arm; the suboptimal gap $\Delta = \invE{Y}{X \gets 0} - \invE{Y}{X \gets 1}$. Applying \texttt{UCB} gives the regret bound
\begin{align}
	R(T, \1M^*) \leq \frac{8\log(T)}{\Delta} + \left( 1 + \frac{\pi^2}{3} \right) \Delta \label{eq:_5_1_ucb3}
\end{align}
Suppose that the causal bound $[l_1, r_1]$ of arm $x = 1$ is informative and $r_1 < \mu_0$. Thm.~\ref{thm:_5_1_ucb} implies that the regret bound of \texttt{UCB}\textsuperscript{+} taking into account this causal bound is
\begin{align}
	R(T, \1M^*) \leq \left( 1 + \frac{\pi^2}{3} \right) \Delta \label{eq:_5_1_ucb4}
\end{align}
In words, when the causal bound is informative, \texttt{UCB}\textsuperscript{+} enjoys a constant regret $\mathcal{O}(1)$ which is orders of magnitude smaller than the sublinear regret $\mathcal{O}\left ( \log(T) / \Delta\right)$ of \texttt{UCB}. On the other hand, if the causal bound is not informative and $r_1 \geq \mu_0$, the regret bound of \texttt{UCB}\textsuperscript{+} coincides with the regret of \texttt{UCB} in Eq.~\ref{eq:_5_1_ucb3}, and no negative transfer occurs.

\begin{figure}[t]
	\centering
	\hfill
	\begin{minipage}[b]{0.33\linewidth}
		\centering
		\includegraphics[width=1.0\textwidth]{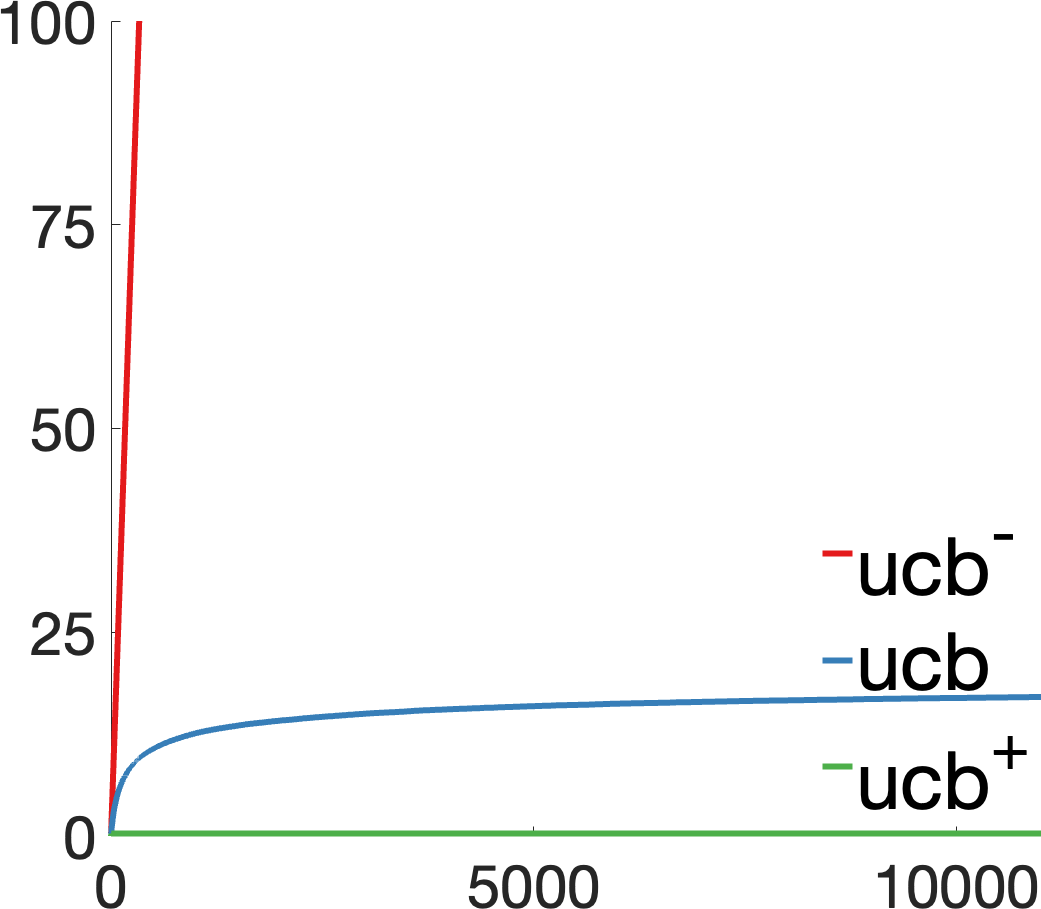}
		\subcaption{$\Delta = 0.3$}
		\label{fig:_5_1_mab1_a}
	\end{minipage}\hfill
	\begin{minipage}[b]{0.33\linewidth}
		\centering
		\includegraphics[width=1.0\textwidth]{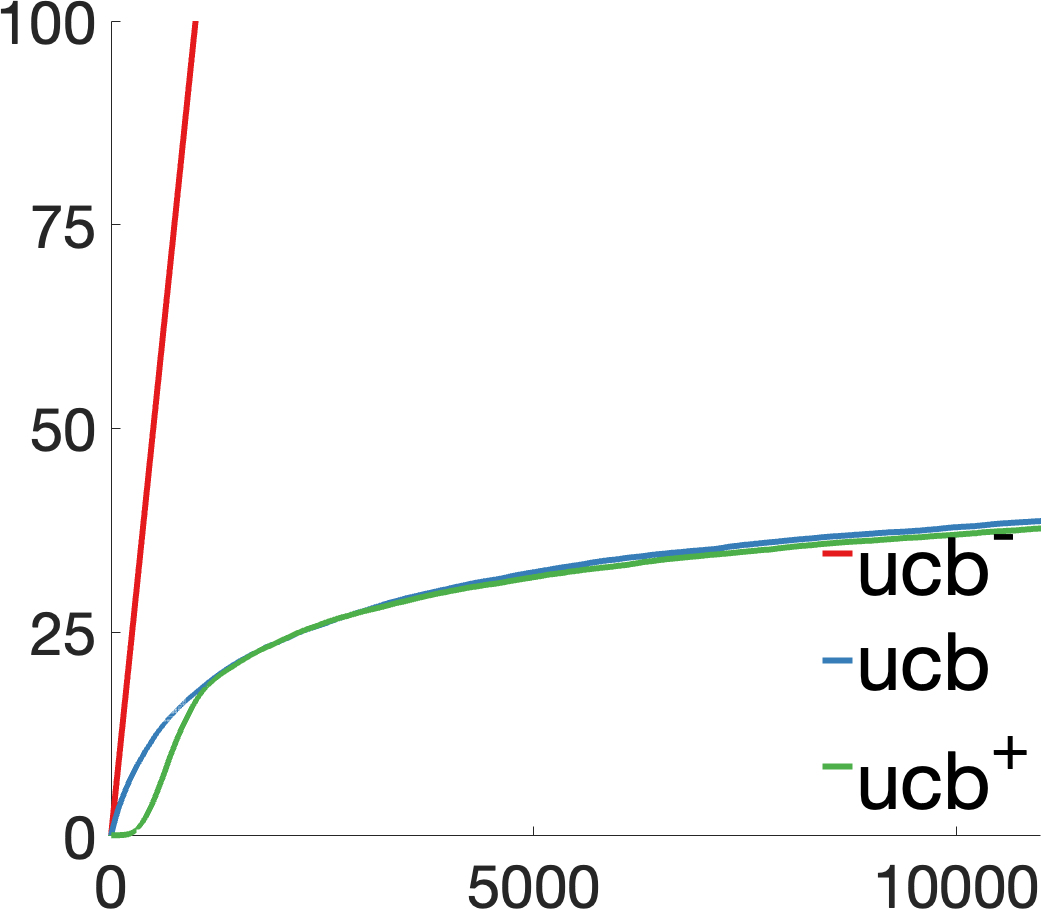}
		\subcaption{$\Delta = 0.1$}
		\label{fig:_5_1_mab1_b}
	\end{minipage}\hfill\null
	\caption{Simulation results comparing \texttt{UCB}\textsuperscript{+} learner augmented with causal bounds over the expected rewards, standard \texttt{UCB}, and \texttt{UCB}\textsuperscript{-} with direct transfer of observational data.}
	\label{fig:_5_1_mab1}
\end{figure}
\begin{experiment}\label{exp:_5_1_mab2}
	Fig.~\ref{fig:_5_1_mab1} shows the cumulative regret of \texttt{UCB}\textsuperscript{+} in the MAB environment $\1M^*$ described in Example~\ref{exp:_2_1_mab} with the suboptimal gap $\Delta = 0.3$ and $\Delta = 0.1$ respectively. It takes as input the natural bounds $\left[l_x, r_x \right]$ over the expected rewards of arm $x = 0, 1$, computed from taking as input $5,000$ observational samples drawn from the distribution $P(X, Y)$. As a baseline, we also include a vanilla \texttt{UCB} starting from scratch without utilizing any prior observations, and \texttt{UCB}\textsuperscript{-} with the confounded observational data directly transferred.

	One can see by inspection the significant disparity between the performance of \texttt{UCB}\textsuperscript{+} and \texttt{UCB} for $\Delta = 0.3$. In this case, the causal bound for the suboptimal arm $x = 1$ is $r_1 = 0.6 - \Delta = 0.3 < \mu_0$, and \texttt{UCB}\textsuperscript{+} converges to the optimal arm $x = 0$ almost immediately after the learning starts. On the other hand, when the suboptimal gap $\Delta = 0.1$ and the causal bound $r_1 = 0.6 - \Delta  = 0.5 > \mu_0$, the performance of \texttt{UCB}\textsuperscript{+} and \texttt{UCB} virtually coincides. The simulation results corroborate the theory that the transfer strategy of \texttt{UCB}\textsuperscript{+} enjoys no negative impact. As expected, the direct transfer \texttt{UCB}\textsuperscript{-} performs the worst among all strategies due to unobserved confounding. \hfill $\blacksquare$
\end{experiment}

\subsection{Online Learning in Sequential Decision-Making}\label{sec:_5_2}
The offline-to-online learning strategy described so far focuses on the MAB models with a single decision horizon $H = 1$.
The remainder of this section will extend this strategy to optimize a general CDM $\langle \1M^*, \Pi, \1R \rangle$ where the policy space $\Pi = \Braces{\Tuple{X_i, \*S_i}}_{i = 1}^H$ and the reward function $\1R: \D(\*Y) \mapsto \3R$ taking a set of signals $\*Y$ as input. We will first introduce a purely online learning algorithm to optimize a CDM $\langle \1M^*, \Pi, \1R \rangle$ without detailed parametrization of the underlying environment $\1M^*$. Compared to bandit algorithms previously described for MAB models, our proposed algorithm also interacts with the underlying SCM $\1M^*$ for repeated episodes $t = 1, \dots, T$. For each episode $t$, instead of selecting a single arm $X$, our algorithm will determine values of a sequence of actions $X_1, \dots, X_H$. More specifically, it will pick a policy $\pi^{(t)} = \Parens{\pi^{(t)}_1, \dots, \pi^{(t)}_H}$ at the beginning of episode $t$. For every step $i = 1, \dots, H$ of interventions, our algorithm observed state $\*S^{(t)}_i$, selects an action $X^{(t)}_i \sim \pi^{(t)}_i\Parens{X_i \mid \*S^{(t)}_i}$ following the decision rule $\pi^{(t)}_i$, and perform intervention $\doo\Parens{X_i \gets X^{(t)}_i}$. The cumulative regret for an online learning algorithm operating in a CDM $\langle \1M^*, \Pi, \1R \rangle$ after $T > 1$ episodes of interventions is defined as follows, compared to an idealized agent following the optimal policy $\pi^*$ for all episodes of interactions $t = 1, \dots, T$,
\begin{align}
	R(T, \M^*) = T \invE{\1R(\*Y) ; \M^*}{\pi^*} -  \sum_{t = 1}^{T} \invE{\1R\Parens{\*Y}}{\pi^{(t)}}. \label{eq:_5_2_cumulative_regret}
\end{align}
Similar to the bandit setting, a nice property for an online algorithm is to achieve a sublinear regret $R(T, \M^*) = o(T)$ so that it eventually converges to an optimal policy $\pi^*$.\footnote{See Sec.~\ref{sec:_4_2} for a detailed discussion of the online learning task and properties of cumulative regret.}

There exists an experimental design of randomized trials for optimizing policies over a finite sequence of actions $\*X$ in an unknown environment, called sequential multiple assignment randomized trials (for short, SMART \citep{murphy2005experimental}). It is an explore-then-commit algorithm. More specifically, fix a total number of trials $N \in \3N$. For the first $t \leq N$ episodes, the SMART algorithm explores by sampling values $X^{(t)}_i$ of every action $X_i$, for $i = 1, \dots, H$, from its domain $\D(X_i)$ uniformly at random. For episodes $t > N$, the algorithm commits to a policy $\pi^{(t)}$ maximizing the empirical reward estimates $\hat{\3E}_{\pi}\Brackets{\1R(\*Y)}$ computed from experimental data $\1D_{\text{exp}}^{(N)}$ collected during exploration. Details of the algorithm is summarized in Alg.~\ref{alg:_4_2_rct}.

When the total number of trials $N$ is sufficiently large, SMART is able to recover the effects of candidate policies and find an optimal policy from the experimental data \citep{murphy2001dtr,murphy2005generalization}. However, as previously discussed in Sec.~\ref{sec:_4_2}, explore-then-commit algorithms suffer from a linear regret during the exploration phase ($t \leq N$). Determining the optimal trial number $N$ is theoretically challenging, requiring prior parametric knowledge about the underlying environment and the total episodes of interactions $T$.

{We will next describe a novel online learning algorithm for optimizing a policy space over a sequence of actions $\*X = \{X_1, \dots, X_H\}$. It is able to achieve a sublinear regret $R(T, \M^*) = o(T)$ without parametric knowledge of the SCM $\1M^*$ and total episodes $T$. Our discussion begins with the decision with some necessary notations and technical tools. Recall that for every $i = 1, \dots, H$, $\bar{\*X}_i = \{X_1, \dots, X_i\}$ is a sequence of actions up to stage $i$ and $\bar{\*S}_i = \{\*S_1, \dots, \*S_i\}$ is the sequence of corresponding states. For any policy $\pi \in \Pi$, using basic probabilistic operations and the Bayes' rule, the expected reward $\invE{\mathcal{R}(\*Y)}{\pi}$ could be written as
\begin{align}
	\invE{\mathcal{R}(\*Y)}{\pi} & =\sum_{\bar{\*x}_H, \bar{\*s}_H} \underbrace{\invE{\1R(\*Y) \mid \bar{\*x}_H, \bar{\*s}_H}{\pi}}_{\text{reward}} \prod_{i = 0}^{H-1} \underbrace{\inv{ s_{i+1} \mid \bar{\*x}_i, \bar{\*s}_i}{\pi}}_{\text{transition probabilities}} \underbrace{\pi_{i+1}\left(x_{i+1} \mid \*s_{i+1}\right)}_{\text{policy}} \label{eq:_5_2_id1}
\end{align}
The above equation follows that in submodel $\1M^*_{\pi}$, values of every action $X_i$ are determined by the function $\pi_i$. Among the quantities above, probabilities of policy $\pi$ are known. It is sufficient to estimate transition distributions $\inv{ s_{i+1} \mid \bar{\*x}_i, \bar{\*s}_i}{\pi}$ and the conditional reward $\invE{\1R(\*Y) \mid \bar{\*x}_H, \bar{\*s}_H}{\pi}$.

The NUC condition holds in the post-interventional system $\langle \1M^*_{\pi}, \Pi, \1R \rangle$ (Lem.~\ref{lem:_4_2_nuc}). Therefore, for every $i = 0, \dots, H-1$, conditioning on past states $\bar{\*S}_i$ d-separates all backdoor paths between actions $\bar{\*X}_i$ and another variables in submodel $\1M^*_{\pi}$. Applying Rule 2 of do-calculus (Thm.~\ref{def:_4_3_do-calculus}),
\begin{align}
	 & \inv{ s_{i+1} \mid \bar{\*x}_i, \bar{\*s}_i}{\pi} = \inv{s_{i+1} \mid \bar{\*s}_i}{\bar{\*x}_i},   \label{eq:_5_2_transition}\\
	 & \invE{\1R(\*Y) \mid \bar{\*x}_H, \bar{\*s}_H}{\pi} = \invE{\1R(\*Y) \mid \bar{\*s}_H}{\bar{\*x}_H} \label{eq:_5_2_reward}
\end{align}
In words, the transition distribution $\inv{s_{i+1} \mid \bar{\*x}_i, \bar{\*s}_i}{\pi}$ and conditional reward $\invE{\1R(\*Y) \mid \bar{\*x}_H, \bar{\*s}_H}{\pi}$ remain invariant across policies $\pi \in \Pi$. Therefore, they could be consistently estimated by pooling interventional data collected by different candidate policies in $\Pi$. 

Throughout this section, we will consistently assume that the domains of the states $\*S$ and actions $\*X$ are finite; the reward function $\1R(\*Y)$ is bounded in a real interval $[0, 1]$. Fix a finite sequence $\pi^{(1)}, \dots, \pi^{(t)} \in \Pi$. Given finite samples $\Braces{\*V^{(1)}, \dots, \*V^{(t)}}$ drawn from distributions $\inv{\*V}{\pi^{(1)}} \dots, \inv{\*V}{\pi^{(t)}}$ respectively, empirical mean estimates for the transition distribution $\inv{\*S_i \mid \bar{\*s}_{i-1}}{\bar{\*x}_{i-1}}$, $i = 1, \dots, H-1$, and the conditional reward $\invE{Y\mid \bar{\*s}_H}{\bar{\*x}_H}$ are defined as:
\begin{align}
	\forall i = 1, \dots, H-1,\;\; & \hat{P}^{(t)}_{\bar{\*x}_{i}}\left (\*s_{i+1} \mid \bar{\*s}_{i} \right) =  \frac{\sum_{j = 1}^{t} \I\left \{\bar{\*X}^{(j)}_{i} = \bar{\*x}_{i}, \bar{\*S}^{(j)}_{i+1} = \bar{\*s}_{i+1} \right \}}{ N_{t}(\bar{\*x}_{i}, \bar{\*s}_{i}) }, \label{eq:_5_2_empirical1}              \\
	\text{and}\;\;                 & \hat{\E}^{(t)}_{\bar{\*x}_H} \left[\1R(\*Y) \mid \bar{\*s}_H \right] = \frac{\sum_{j = 1}^{t} \I\left \{\bar{\*X}^{(j)}_{H} = \bar{\*x}_{H}, \bar{\*S}^{(j)}_{H} = \bar{\*s}_{H} \right \} \1R\Parens{\*Y^{(j)}} }{ N_{t}(\bar{\*x}_{H}, \bar{\*s}_{H})}, \label{eq:_5_2_empirical2}
\end{align}
Among quantities in the above equations, for every $i = 1, \dots, H$, $N_t(\bar{\*x}_{i}, \bar{\*s}_{i}) $ is the event count for every state-action pair $(\bar{\*x}_i, \bar{\*s}_i) \in \D(\bar{\*X}_i\cup \bar{\*S}_i)$ defined as
\begin{align}
	\forall i = 1, \dots, H,\;\; N_t(\bar{\*x}_{i}, \bar{\*s}_{i}) = \max\left \{1, \sum_{j=1}^{t} \I\left \{\bar{\*X}^{(j)}_{i} = \bar{\*x}_{i}, \bar{\*S}^{(j)}_{i} = \bar{\*s}_{i} \right \} \right \}.
\end{align}

Alg.~\ref{alg:_5_2_ucb} shows details of \texttt{UCB} algorithm capable of optimizing an unknown CDM $\langle \1M^*, \Pi, \1R \rangle$. It works in phases of model construction, optimistic planning, and policy execution. In Step 2, \texttt{UCB} constructs a set $\3M_{t-1}(\delta)$ of plausible SCMs from interventional data $\Braces{\*V^{(1)}, \dots, \*V^{(t-1)}}$. For every SCM $\1M \in \3M_{t-1}(\delta)$, its transition distributions $\inv{\*S_{i+1} \mid \bar{\*S}_{i}}{\bar{\*x}_{i}}$, $i = 1, \dots, H-1$, and the conditional reward $\invE{\1R(\*Y) \mid \bar{\*s}_H}{\bar{\*x}_H}$ are contained in convex intervals centering around their corresponding empirical estimates computed from interventional data collected prior to episode $t$.
The error probability $\delta$ is set as a decreasing function of the episode number $t$ so that $\3M_{t-1}(\delta)$ contains the underlying SCM $\1M^*$ with high probability as the algorithm continues and more interventional data are collected. It then computes in Step 3 an optimal policy $\pi^{(t)}$ of the most optimistic SCM $\1M^{(t)} \in \3M_{t-1}(\delta)$ that induces the maximal expected reward. We will discuss the details of the model construction and the optimistic planning later. Finally, policy $\pi^{(t)}$ is executed throughout episode $t$ and new samples $\*V^{(t)} \sim \inv{\*V}{\pi^{(t)}}$ are collected (Step 4).

\begin{algorithm}[!t]
	\caption{Upper Confidence Bound (\texttt{UCB}) for CDMs}
	\label{alg:_5_2_ucb}
	\setlength{\textfloatsep}{0pt}
	\begin{algorithmic}[1]
		\Require a policy space $\Pi = \Braces{\Tuple{X_i, \*S_i}}_{i = 1}^H$, a reward function $\1R: \D(\*Y) \mapsto [0, 1]$
		\ForAll{episodes $t = 1, 2, \dots $}
		\State Construct a set $\3M_{t-1}(\delta)$ of all candidate SCMs $\1M$, with error probability $\delta = t^{-4}$, that are compatible with the interventional data $\Braces{\*V^{(1)}, \dots, \*V^{(t-1)}}$ collected up to episode $t$.
		\State Find the optimal policy $\pi^{(t)}$ of an optimistic SCM $\1M^{(t)} \in \3M_{t-1}(\delta)$ such that
		\begin{align}
			 & \E_{\pi^{(t)}}\left [\1R(\*Y) ; \1M^{(t)} \right]= \max_{\pi, \1M}\E_{\pi}\Big [\1R(\*Y) ; \1M \Big] \;\; \text{s.t.} \;\; \pi \in \Pi, \1M \in \3M_{t-1}\left( \delta \right). \label{eq:_5_2_optimize}
		\end{align}
		\State Perform $\doo\Parens{\pi^{(t)}}$ for episode $t$ and receive observations $\*V^{(t)}$.
		\EndFor
	\end{algorithmic}
\end{algorithm}

\paragraph{Model Construction}
Let $\Braces{Y^{(1)}, \dots, Y^{(n)}}$ be i.i.d. samples drawn from a discrete distribution $P(Y)$. Let $\hat{P}(y) = \frac{1}{n} \sum_{t = 1}^n \I\Braces{Y^{(i)} = y}$ be empirical estimates of probabilities $P(y)$. Generally, the $L_1$-deviation of the true distribution and the empirical distribution is bounded according to \citep{weissman2003inequalities}
\begin{align}
	 & P\left( \left \lVert \hat{P}\left( \cdot \right) - P \left ( \cdot  \right) \right \rVert_1 > \sqrt{ \frac{2 \left |\D(Y) \right | \log\left ( 2/\delta \right)}{n}} \right) \leq \delta \label{eq:_5_2_ucb_joint}
\end{align}
Fix $\delta \in (0, 1)$. Let $\3M_t(\delta)$ be the set of all SCMs $\1M$ with endogenous variables $\*V$, and with its transition distribution $\inv{\*S_{i+1} \mid \bar{\*s}_{i}}{\bar{\*x}_{i}}$ close to the empirical distribution $\hat{P}^{(t)}_{\bar{\*x}_{i}}\left (\*S_{i+1} \mid \bar{\*s}_{i} \right)$, $i = 0, \dots, H-1$, and the reward $\invE{\1R(\*Y)\mid \bar{\*s}_H}{\bar{\*x}_H}$ close to the empirical reward $\hat{\E}^{(t)}_{\bar{\*x}_H} \left[\1R(\*Y) \mid \bar{\*s}_H \right]$, i.e.,
\begin{align}
	\forall i = 0 \dots, H-1,\;\; & \left \lVert P_{\bar{\*x}_{i}}(\cdot | \bar{\*s}_{i}; \M) - \hat{P}^{(t)}_{\bar{\*x}_{i}}(\cdot | \bar{\*s}_{i}) \right \rVert_1 \leq \epsilon_i(\delta), \label{eq:_5_2_ci1} \\
	\text{and}\;\;                & \left | \E_{\bar{\*x}_H}[\1R(\*Y) | \bar{\*s}_H; \M] - \hat{\E}^{(t)}_{\bar{\*x}_H}[\1R(\*Y) | \bar{\*s}_H] \right | \leq \epsilon_H(\delta).\label{eq:_5_2_ci2}
\end{align}
where the confidence width $\epsilon_i(\delta)$ is a function given by,
\begin{align}
	\forall i= 0, \dots, H-1, \;\; & \epsilon_i(\delta) = \sqrt{\frac{2 \left |\D(\*S_{i+1}) \right | \log(4H \left |\D(\bar{\*S}_{i}\cup \bar{\*X}_{i}) \right | / \delta)}{ N_t(\bar{\*x}_{i}, \bar{\*s}_{i}) }} \label{eq:_3_2_ci3} \\
	\text{and } \;\;               & \epsilon_H(\delta) = \sqrt{\frac{\log(8H \left  |\D(\bar{\*S}_{H}\cup \bar{\*X}_{H}) \right | / \delta)}{ 2N_t(\bar{\*x}_H, \bar{\*s}_H)}} \label{eq:_5_2_ci4}
\end{align}
Applying a union bound over the concentration inequalities in Eq.~\ref{eq:_5_2_ucb_joint} and Hoeffding's inequality in Eq.~\ref{eq:_4_2_ucb1}, we  obtain the following error probability:
\begin{align}
	P\left (\1M^* \not \in \3M_t(\delta) \right) & \leq \sum_{i = 0}^{H - 1} \sum_{\bar{\*x}_{i}, \bar{\*s}_{i} } P\left( \left \lVert P_{\bar{\*x}_{i}}(\cdot | \bar{\*s}_{i}; \M^*) - \hat{P}^{(t)}_{\bar{\*x}_{i}}(\cdot | \bar{\*s}_{i}) \right \rVert_1 > \epsilon_i(\delta) \right) \\
	                                             & +  \sum_{\bar{\*x}_{H}, \bar{\*s}_{H}}P\left( \left | \E_{\bar{\*x}_H}[\1R(\*Y) | \bar{\*s}_H; \M] - \hat{\E}^{(t)}_{\bar{\*x}_H}[\1R(\*Y) | \bar{\*s}_H] \right | > \epsilon_H(\delta) \right)                                        \\
	                                             & =\sum_{i = 0}^{H - 1} \sum_{\bar{\*x}_{i}, \bar{\*s}_{i}} \frac{\delta}{2H \left |\D(\bar{\*S}_{i}\cup \bar{\*X}_{i}) \right |} + \sum_{\bar{\*x}_{H}, \bar{\*s}_{H}}\frac{\delta}{2H \left |\D(\bar{\*S}_{H}\cup \bar{\*X}_{H})\right|} \\
	                                             & < \delta \label{eq:_5_2_concentration}
\end{align}
That is, the underlying SCM $\1M^*$ is contained in SCM family $\3M_t(\delta)$ with probability at least $1 - \delta$.

\paragraph{Optimistic Planning} 
Step 7 of \texttt{UCB} tries to find an optimal policy $\pi^{(t)}$ for an optimistic SCM $\1M^{(t)}$. For a fixed SCM $\1M$, standard planning algorithms \citep{bellman57, koller2003multi} are applicable to allow one to find an optimal policy in space $\Pi$ that maximizes the expected reward. However, the optimization problem of Eq.~\ref{eq:_5_2_optimize} also requires the learner to find an SCM $\1M^{(t)}$ that defines the maximal optimal reward among all plausible SCMs in family $\3M_{t-1}(\delta)$.

Generally, we can formulate this problem as a polynomial program as follows. The decomposition in Eq.~\ref{eq:_5_2_id1}, together the invariances in Eqs.~\ref{eq:_5_2_transition} and \ref{eq:_5_2_reward}, allows one to write the expected reward $\invE{\mathcal{R}(\*Y)}{\pi}$ as follows
\begin{align}
	\invE{\mathcal{R}(\*Y)}{\pi} & =\sum_{\bar{\*x}_H, \bar{\*s}_H} \invE{\1R(\*Y) \mid\bar{\*s}_H}{\bar{\*x}_H} \prod_{i = 0}^{H-1}\inv{ s_{i+1} \mid \bar{\*s}_i}{\bar{\*x}_i} \pi_{i+1}\left(x_{i+1} \mid \*s_{i+1}\right) \label{eq:_5_2_id2}
\end{align} 
where every decision rule $\pi_i(X_i \mid \*S_i)$, $i = 1, \dots, H$, is a proper conditional distribution mapping from the domains of states $\*S_i$ to action $X_i$. Probabilities of transition distributions $\inv{S_{i+1} \mid \bar{\*s_i}}{\bar{\*x}_i}$ for $i = 0, \dots, H-1$ are contained in the convex set defined in Eq.~\ref{eq:_5_2_ci1}; values of the conditional reward $\invE{\1R(\*Y) \mid  \bar{\*s}_H}{\bar{\*x}_H}$ are contained in the convex polytope $\*{\1R}$ defined in Eq.~\ref{eq:_5_2_ci2}. 

Solving for a policy $\pi^{(t)}$ and an optimistic SCM $\1M^{(t)}$ in Eq.~\ref{eq:_5_2_optimize} is equivalent to solving a polynomial program with objective function $\invE{\1R(\*Y)}{\pi}$ defined in Eq.~\ref{eq:_5_2_id2} with transitional probabilities $\inv{S_{i+1} \mid \bar{\*s_i}}{\bar{\*x}_i} \in \*{\1P}_i$, $i = 0, \dots, H-1$, and reward mean $E_{\bar{\*x}_K}[\1R(\*Y) | \bar{\*s}_K] \in \*{\1R}$; and probabilistic constraints $\sum_{x_i} \pi_i(x_i \mid \*s_i) = 1$ and $\pi_i(x_i \mid \*s_i) \geq 0$, for every $i = 1, \dots, H$. There exists an efficient dynamic programming procedure to solve this polynomial program when the policy space $\Pi$ satisfies the \emph{perfect recall} condition \cite[Def.~23.5]{koller2009probabilistic}. In words, this conditions states that $\*S_i \cup \{X_i\} \subseteq \*S_j$ whenever $i < j$, which means that the agent does not forget the previous decision or information it once had. \footnote{In many real-world healthcare applications where the decision horizon $H$ is low, the perfect recall assumption is quite natural and automatically satisfied. 
On the other hand, in some practical settings where the horizon $H$ is high or even infinite, the policy space $\Pi$ satisfying the perfect recall is high-dimensional. Planning in $\Pi$ is computationally challenging even when parameters of the underlying SCM are fully known \citep{papadimitriou1987complexity}.} 
An optimistic policy $\pi^{(t)}$ is obtainable by solving following extended Bellman equations, for $\forall i= 1, \dots, H$,
\begin{align}
	                   & Q^*(\bar{\*s}_i, \bar{\*x}_{i-1}) = \max_{x_i} \left  \{ \max_{\inv{\cdot \mid \bar{\*s}_i}{\bar{\*x}_i} \in \*{\mathcal{P}}_i} \left  \{  \sum_{s_{i+1}} Q^*(\bar{\*s}_{i+1}, \bar{\*x}_i)P_{\bar{\*x}_i}(s_{i+1}| \bar{\*s}_i) \right  \} \right  \}, \notag \\
	\text{and}\;\;\;\; & Q^*(\bar{\*s}_H, \bar{\*x}_{H-1}) = \max_{x_H} \max_{\invE{\1R(\*Y) \mid \bar{\*s}_H}{\bar{\*x}_H} \in \*{\mathcal{R}}} \invE{\1R(\*Y) \mid \bar{\*s}_H}{\bar{\*x}_H}, \label{eq:_5_2_bellman}
\end{align}
The inner maximum in the above equation is a linear program (LP) over the convex polytope $\*{\mathcal{P}}_k$ (or $\*{\mathcal{R}}$), which is solvable using by an iterative algorithm introduced by \citep{strehl2008analysis}. For grounding purposes, we provide the complete algorithm in Appendix.~\ref{app:_extended_q}. 
\begin{restatable}{theorem}{thmucbdtr}\label{thm:_5_2_ucb}
	Let $\langle \1M^*, \Pi, \1R \rangle$ be a CDM where $\Pi = \left \{\langle X_i, \*S_i \rangle \right \}_{i=1}^H$ and $\1R: \D(\*Y) \mapsto [0, 1]$. For any $\epsilon \geq 0$, the regret of \texttt{UCB} in SCM $\1M^*$ after $T > 1$ episodes is bounded by
	\begin{align}
		R(T, \1M^*) \leq \max_{\pi \in  \Pi^{\mathrm{o}}: \Delta_{\pi} > \epsilon } \frac{17^2 H^2 \left | \D(\*X \cup \*S) \right| \log(T)}{\Delta_{\pi}} + \max_{\pi \in  \Pi^{\mathrm{o}}: \Delta_{\pi} \leq \epsilon} \Delta_{\pi} T + \frac{\pi^2}{6}. \label{eq:_5_2_regret_e}
	\end{align}
	where $\Pi^{\mathrm{o}} = \{\pi \in \Pi: \pi \text{ is deterministic} \}$ is the set of all deterministic policies in the policy space $\Pi$; and $\Delta_{\pi} = \invE{Y; \1M^*}{\pi^*} - \invE{Y; \1M^*}{\pi}$ is the gap from the optimal reward for any policy $\pi \in \Pi$. Moreover, fix $\epsilon = 0 $. The regret of \texttt{UCB} could be further written as
	\begin{align}
		R(T, \1M^*) \leq  \max_{\pi \in  \Pi^{\mathrm{o}}: \Delta_{\pi} > 0} \frac{17^2 H^2 \left | \D(\*X \cup \*S) \right| \log(T)}{ \Delta_{\pi}} +  \frac{\pi^2}{6}. \label{eq:_5_2_regret}
	\end{align} \hfill $\blacksquare$
\end{restatable}
Thm.~\ref{thm:_5_2_ucb} implies that Alg.~\ref{alg:_5_2_ucb} is able to achieve a sublinear regret $\mathcal{O} \left( H^2 \left | \D(\*X \cup \*S) \right| \log(T)/\Delta \right)$ where $H$ is the total number of actions (i.e., the decision horizon), $\left | \D(\*X \cup \*S) \right|$ is the cardinality of the state-action domain; $T$ is the total episodes of online interventions; and $\Delta$ is the gap in the expected reward between the second-best deterministic policy $\pi$ and the optimal policy $\pi$ evaluated in the underlying SCM $\1M^*$. This means that \texttt{UCB} is able to converge and eventually obtain an optimal policy $\pi^*$ as the total episodes of intervention $T$ increases. Moreover, suppose $\1M^*$ is an MAB model with $K$ candidate arms, i.e., $H = 1$ and $\left | \D(\*X \cup \*S) \right| = K$. The regret bound of Eq.~\ref{eq:_5_2_regret} is equal to $\mathcal{O} \left( K\log(T)/ \Delta\right)$, which matches the analytical result in Thm.~\ref{thm:_4_2_ucb}.

\begin{figure}[t]
	\centering
	\hfill
	\begin{minipage}[b]{0.33\linewidth}
		\centering
		\includegraphics[width=1.0\textwidth]{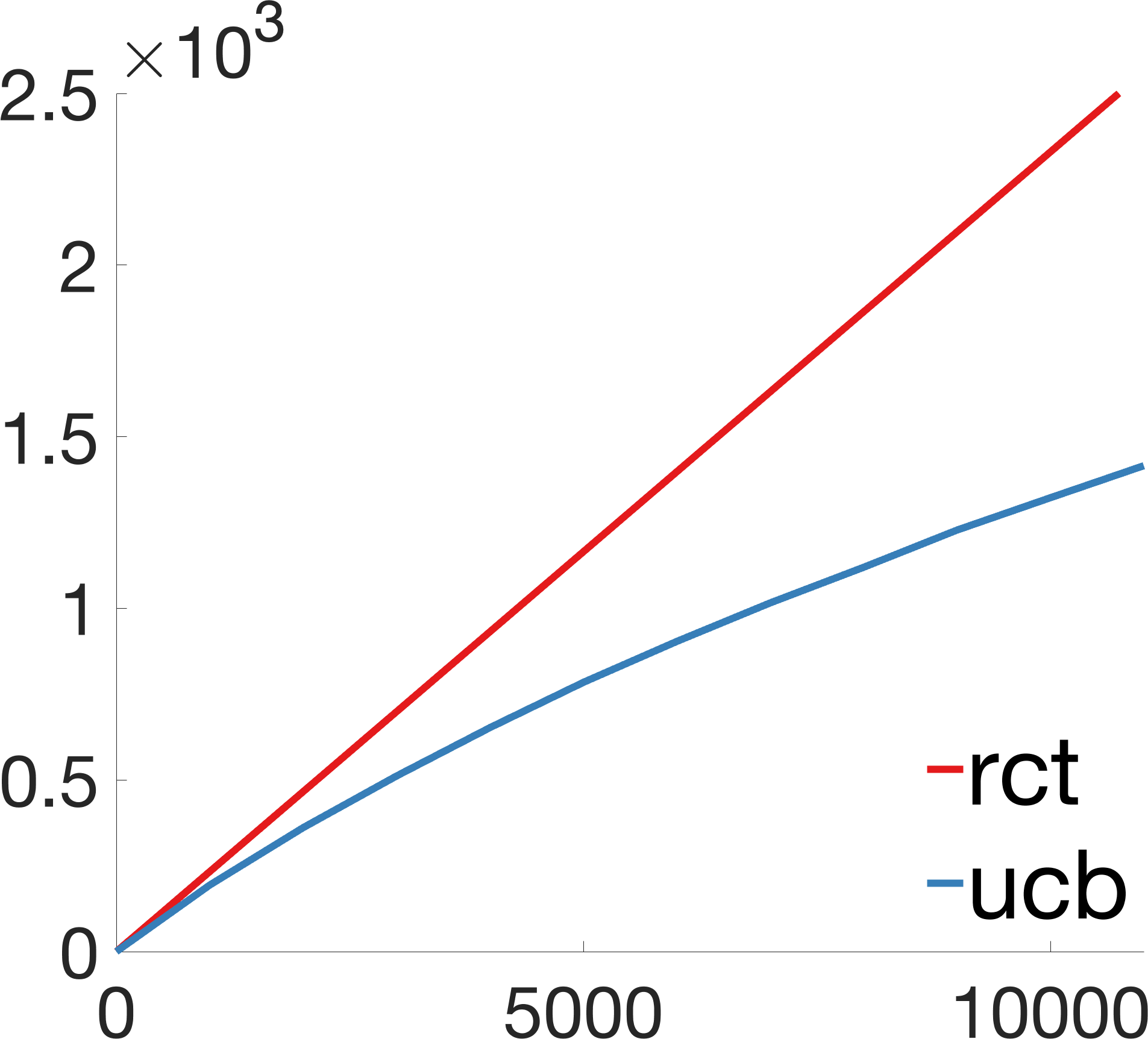}
		\subcaption{$\alpha_1 = 3, \alpha_2 = -3$}
		\label{fig:_5_2_dtr_a}
	\end{minipage}\hfill
	\begin{minipage}[b]{0.33\linewidth}
		\centering
		\includegraphics[width=1.0\textwidth]{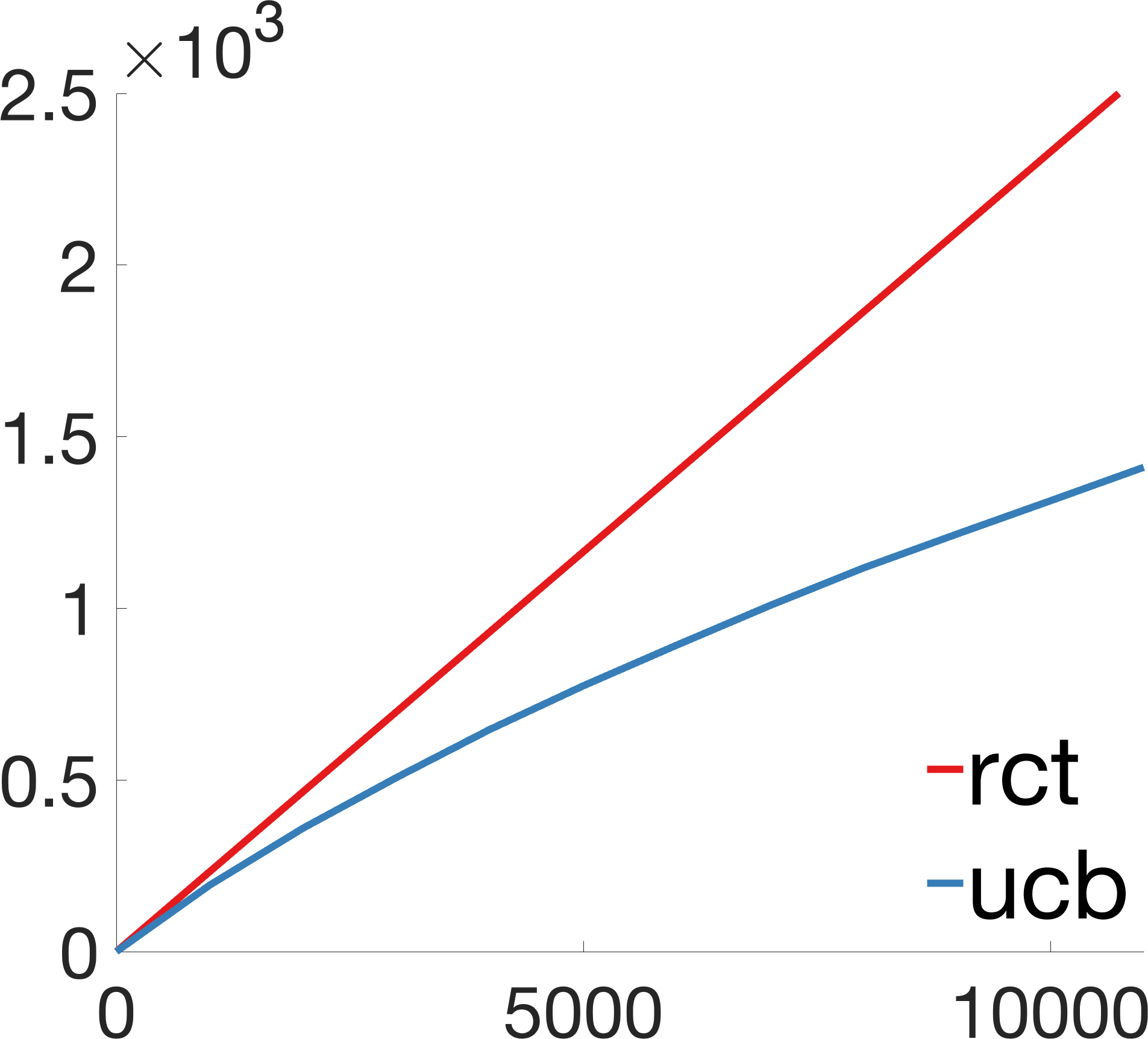}
		\subcaption{$\alpha_1 = -12, \alpha_2 = -3$}
		\label{fig:_5_2_dtr_b}
	\end{minipage}\hfill\null
	\caption{Simulation results comparing \texttt{UCB} learner optimizing a 2-stage DTR model and \texttt{RCT} determining values of actions $X_1, X_2$ uniformly at random.}
	\label{fig:_5_2_dtr}
\end{figure}

\begin{experiment}\label{exp:_5_2_dtr}
Fig.~\ref{fig:_5_2_dtr} shows the cumulative regret of \texttt{UCB}\textsuperscript{+} in the 2-stage DTR environment $\1M^*$ described in Example~\ref{exp:_3_1_dtr} with the coefficients $\Parens{\alpha_1, \alpha_2}$ set to $(3, -3)$ and $(-12, -3)$ respectively. 
It takes as input the causal bounds over $\inv{S_2\mid S_1}{X_1}$ and $\invE{Y\mid S_1, S_2}{X_1, X_2}$, computed from the observational distribution $P(S_1, X_1, S_2, X_2, Y)$. As a baseline, we include a randomized controlled trials (\texttt{RCT}) deciding treatments $X_1, X_2$ uniformly at random, as introduced by \citep{murphy2005experimental}, which extended the one-shot, classical treatment by \citep{fisher:35}. 

One can see by inspection \texttt{UCB} is able to achieve a sublinear regret in both DTR models. Simulations corroborate the analytical results \texttt{UCB} is able to eventually converge to an optimal policy $\pi^*$ as the total number of trials $T$ increases. As expected, the randomized strategy \texttt{RCT} performs the worst among all strategies due to the linear regret during exploration. \hfill $\blacksquare$
\end{experiment}

\subsection{Learning from Observational Data}\label{sec:_5_3}
Despite its performance guarantee, the online learning algorithm introduced in Alg.~\ref{alg:_5_2_ucb} does not make use of any knowledge in the observational distribution $P(\*V)$. 
When the NUC condition (Def.~\ref{def:_4_1_nuc}) holds in the underlying CDM $\langle \1M^*, \Pi, \1R \rangle$, the state and actions' history $\bar{\*X}_{i-1}, \bar{\*S}_i$ blocks all backdoor paths every action $X_i$ to any other variable in the causal diagram $\G$. 
One could thus estimate the transition distribution $\inv{\*S_{i+1} \mid \bar{\*s}_i}{\bar{\*x}_i}$, $i = 0, \dots, H-1$ and reward $\invE{\1R(\*Y)\mid \bar{\*s}_H}{\bar{\*x}_H}$ using the corresponding conditional distribution $P\left( \*S_{i+1} \mid \bar{\*x}_i, \bar{\*s}_i \right)$ and $\E\left[ \1R(\*Y) \mid \bar{\*x}_H, \bar{\*s}_H\right]$. 
The validity of the estimation procedure follows from Rule 2 of do-calculus (Def.~\ref{def:_4_3_do-calculus}). 
These estimations could then be directly transferred to ``warm-start'' \texttt{UCB} algorithm. 
However, issues of non-identifiability could arise in general settings where the NUC does not hold and no additional causal assumptions are provided.\footnote{The non-identifiability of transition distributions $\inv{\*S_{i+1} \mid \bar{\*s}_i}{\bar{\*x}_i}$, $i = 2, \dots, H-1$ and reward $\invE{ \1R(\*Y)\mid \bar{\*s}_H}{\bar{\*x}_H}$ has been acknowledged in \citep{lee2019gid,correa2019statistical}.}

Given such challenges, we then consider the \textit{partial identification} of the transition distributions and the expected reward from the observational distribution. Our first result bounds interventional transition probabilities $\inv{\*s_{i+1} \mid \bar{\*s}_i}{\bar{\*x}_i}$ from the observational distribution $P(\*V)$.
\begin{restatable}{theorem}{thmcboundt}\label{thm:_5_3_cbound1}
	Let $\langle \1M^*, \Pi, \1R \rangle$ be a CDM where the policy space $\Pi = \left \{\langle X_i, \*S_i \rangle \right \}_{i=1}^H$. For every $i = 1, \dots, H-1$, $\inv{\*s_{i+1} \mid \bar{\*s}_i}{\bar{\*x}_i}   \in \left [l_{\bar{\*x}_i}(\bar{\*s}_{i+1}), r_{\bar{\*x}_i}(\bar{\*s}_{i+1}) \right ]$  where
	\begin{align}
		 & l_{\bar{\*x}_i}(\bar{\*s}_{i+1}) = \frac{P(\bar{\*s}_{i+1}, \bar{\*x}_i)}{\Gamma(\bar{\*s}_{i}, \bar{\*x}_{i-1} )}, &  & r_{\bar{\*x}_i}(\bar{\*s}_{i+1}) = \frac{\Gamma(\bar{\*s}_{i+1}, \bar{\*x}_i ) }{ \Gamma(\bar{\*s}_{i}, \bar{\*x}_{i-1} )}. \label{eq:_5_3_cbound1}
	\end{align}
	and $\Gamma(\bar{\*s}_{i+1}, \bar{\*x}_i)$ is a function of the observational distribution $P(\*V)$ defined as:
	\begin{align}
		\Gamma(\bar{\*s}_{i+1}, \bar{\*x}_i) = \begin{cases}
			                                       P(\*s_1)                                                                                               & \mbox{if $i = 0$} \\
			                                       P(\bar{\*s}_{i+1}, \bar{\*x}_i) - P(\bar{\*s}_i, \bar{\*x}_i) + \Gamma(\bar{\*s}_{i}, \bar{\*x}_{i-1}) & \mbox{if $i = 1, \dots, H-1$}
		                                       \end{cases} \label{eq:_5_3_gamma}
	\end{align}  \hfill $\blacksquare$
\end{restatable}
The upper bound in Eq.~\ref{eq:_5_3_cbound1} could be written as:
\begin{align}
	{\Gamma(\bar{\*s}_{i+1}, \bar{\*x}_i ) \over \Gamma(\bar{\*s}_{i}, \bar{\*x}_{i-1} )} & = {P(\bar{\*s}_{i+1}, \bar{\*x}_i) - P(\bar{\*s}_i, \bar{\*x}_i) + \Gamma(\bar{\*s}_{i}, \bar{\*x}_{i-1}) \over \Gamma(\bar{\*s}_{i}, \bar{\*x}_{i-1} )} \\
	                                                                                      & = 1 - {P(\bar{\*s}_i, \bar{\*x}_i) - P(\bar{\*s}_{i+1}, \bar{\*x}_i)\over \Gamma(\bar{\*s}_{i}, \bar{\*x}_{i-1} )}
\end{align}
Considering the denominator, note that the gap $P(\bar{\*s}_i, \bar{\*x}_i) - P(\bar{\*s}_{i+1}, \bar{\*x}_i) > 0$ whenever observational probabilities $P(\*s, \*x) > 0$ are positive for all realizations of the state-action pair. This means that the causal bounds in Thm.~\ref{thm:_5_3_cbound1} are generally informative, i.e., strictly contained in the real interval $[0, 1]$. 
The bounds following Thm.~\ref{thm:_5_3_cbound1} can be seen as a generalization of the natural ones given in Thm.~\ref{thm:_5_1_nb} to the sequential settings with multiple actions $H > 1$. The following example illustrates this connection.
\begin{example}\label{exp:_5_3_dtr}
	Consider the $2$-stage DTR model $\1M^*$ described in Example~\ref{exp:_3_1_dtr}. We will bound the interventional distribution $\inv{S_2 \mid s_1}{x_1}$ from the observational distribution $P(S_1, X_1, S_2, X_2, Y)$. Applying Thm.~\ref{thm:_5_3_cbound1}, we obtain the lower bound
	\begin{align}
		\inv{s_2 \mid s_1}{x_1} & \geq \frac{P(s_1, s_2, x_1)}{\Gamma(s_1)} \\
		                        & \geq  P(s_2, x_1 \mid s_1)
	\end{align}
	The last step follows from $\Gamma(s_1) = P(s_1)$. Similarly, function $\Gamma(s_1, s_2, x_1)$ could be written as:
	\begin{align}
		\Gamma(s_1, s_2, x_1) & = P(s_1, s_2, x_1) - P(s_1, x_1) + \Gamma(s_1)               \\
		                      & =P(s_1, s_2, x_1) - P(s_1, x_1) + P(s_1) \label{eq:_5_3_dtr2}
	\end{align}
	Therefore, we could obtain the upper bound
	\begin{align}
		\inv{s_2 \mid s_1}{x_1} & \leq  \frac{\Gamma(s_1, s_2, x_1)}{\Gamma(s_1)}              \\
		                        & \leq  \frac{P(s_1, s_2, x_1) - P(s_1, x_1) + P(s_1)}{P(s_1)} \\
		                        & \leq  P(s_2, x_1\mid s_1) - P(x_1 \mid s_1) + 1
	\end{align}
	The above bounds could be seen as an application of natural bounds (Thm.~\ref{thm:_5_1_nb}) with action $X_1$ and outcome $S_2$ conditioning on the covariate $S_1$. We evaluate transition probabilities $\inv{s_2 \mid s_1}{x_1}$ in the underlying SCM $\1M^*$, compute their corresponding causal bounds in Table~\ref{tab:_5_dtr_a}. \hfill $\blacksquare$
\end{example}
Similarly, one could bound conditional rewards $\invE{\1R(\*Y) \mid \bar{\*s}_H}{\bar{\*x}_H}$ from the observational data.
\begin{restatable}{theorem}{thmcboundr}\label{thm:_5_3_cbound2}
	Let $\langle \1M^*, \Pi, \1R \rangle$ be a CDM where the policy space $\Pi = \left \{\langle X_i, \*S_i \rangle \right \}_{i=1}^H$ and the reward function $\1R: \D(\*Y) \mapsto [0, 1]$. $\invE{\1R(\*Y) \mid \bar{\*s}_H}{\bar{\*x}_H}  \in \left [l_{\bar{\*x}_H}(\bar{\*s}_H), r_{\bar{\*x}_H}(\bar{\*s}_H) \right ]$ where
	\begin{align}
		 & l_{\bar{\*x}_H}(\bar{\*s}_H) = {E \left [\1R(\*Y) \mid \bar{\*s}_{H}, \bar{\*x}_H \right]P\left (\bar{\*s}_{H}, \bar{\*x}_H \right ) \over \Gamma(\bar{\*s}_{H}, \bar{\*x}_{H-1} )} \notag                                           \\
		 & r_{\bar{\*x}_H}(\bar{\*s}_H) =  1 - { \left (1 - E\left [\1R(\*Y) \mid \bar{\*s}_{H}, \bar{\*x}_H \right ] \right )P\left (\bar{\*s}_{H}, \bar{\*x}_H \right ) \over \Gamma(\bar{\*s}_{H}, \bar{\*x}_{H-1} )} \label{eq:_5_3_cbound2}
	\end{align}
\end{restatable}
Since the conditional reward $E\left [\1R(\*Y) \mid \bar{\*s}_{H}, \bar{\*x}_H \right ] \in [0, 1]$, the bounds in Eq.~\ref{eq:_5_3_cbound2} must be strictly contained in $[0, 1]$ whenever the observational probability $P\left (\bar{\*s}_{H}, \bar{\*x}_H \right )$ is positive, and are thus informative. The bounds developed so far are functions of the observational distribution $P(\*V)$, which is identifiable by the sampling process, and so generally can be estimated consistently. We could estimate causal bounds in Thms.~\ref{thm:_5_3_cbound1} and \ref{thm:_5_3_cbound2} by the corresponding sample mean estimates. Standard concentration inequalities are applicable to control the uncertainties due to finite samples.
\begin{example}\label{exp:_5_3_dtr2}
	\begin{table}[t]
		\centering
		\hfill
		\begin{subtable}{\linewidth}
			\centering
			\renewcommand{\arraystretch}{1.25}
			\begin{tabular}{|ccc|c|cc||ccc|c|cc|}
				$S_1$ & $X_1$ & $S_2$ & $P$    & $l$    & $r$    & $S_1$ & $X_1$ & $S_2$ & $P$    & $l$    & $r$    \\
				\hline
				\hline
				0     & 0     & 0     & 0.4750 & 0.1021 & 0.8872 & 1     & 0     & 0     & 0.4502 & 0.0069 & 0.9915 \\
				0     & 0     & 1     & 0.5250 & 0.1128 & 0.8979 & 1     & 0     & 1     & 0.5498 & 0.0085 & 0.9931 \\
				0     & 1     & 0     & 0.4502 & 0.3534 & 0.5683 & 1     & 1     & 0     & 0.4256 & 0.4190 & 0.4344 \\
				0     & 1     & 1     & 0.5498 & 0.4317 & 0.6466 & 1     & 1     & 1     & 0.5744 & 0.5656 & 0.5810 \\
			\end{tabular}
			\caption{$\inv{S_2\mid S_1}{X_1}$}
			\label{tab:_5_dtr_a}
		\end{subtable}\hfill
		\vspace{0.1in}
		\begin{subtable}{\linewidth}
			\centering
			\renewcommand{\arraystretch}{1.25}
			\resizebox{\textwidth}{!}{
			\begin{tabular}{|cccc|c|cc|cccc|c|cc|}
				$S_1$ & $X_1$ & $S_2$ & $X_2$ & $\E$   & $l$    & $r$    & $S_1$ & $X_1$ & $S_2$ & $X_2$ & $\E$   & $l$    & $r$    \\
				\hline
				\hline
				0     & 0     & 0     & 0     & 0.7851 & 0.0587 & 0.9779 & 1     & 0     & 0     & 0     & 0.2149 & 0.0007 & 0.9959 \\
				0     & 0     & 0     & 1     & 0.9846 & 0.0333 & 0.9990 & 1     & 0     & 0     & 1     & 0.7851 & 0.0014 & 0.9992 \\
				0     & 0     & 1     & 0     & 0.7851 & 0.0138 & 0.9963 & 1     & 0     & 1     & 0     & 0.2149 & 0.0002 & 0.9991 \\
				0     & 0     & 1     & 1     & 0.7851 & 0.0744 & 0.9663 & 1     & 0     & 1     & 1     & 0.2149 & 0.0009 & 0.9934 \\
				0     & 1     & 0     & 0     & 0.2149 & 0.1279 & 0.6254 & 1     & 1     & 0     & 0     & 0.0008 & 0.0007 & 0.2417 \\
				0     & 1     & 0     & 1     & 0.9846 & 0.1170 & 0.9975 & 1     & 1     & 0     & 1     & 0.2149 & 0.0288 & 0.8232 \\
				0     & 1     & 1     & 0     & 0.2149 & 0.0490 & 0.8917 & 1     & 1     & 1     & 0     & 0.0008 & 0.0002 & 0.7897 \\
				0     & 1     & 1     & 1     & 0.7851 & 0.4021 & 0.8917 & 1     & 1     & 1     & 1     & 0.0154 & 0.0102 & 0.2472 \\
			\end{tabular}
			}
			\caption{$\invE{Y\mid S_1, S_2}{X_1, X_2}$}
			\label{tab:_5_dtr_b}
		\end{subtable}\hfill\null
		\caption{Interventional distributions $\inv{S_2\mid S_1}{X_1}$ and $\invE{Y\mid S_1, S_2}{X_1, X_2}$ and their causal bounds defined by a 2-stage DTR environment described in Example~\ref{exp:_3_1_dtr}.}
		\label{tab:_5_dtr}
	\end{table}
	Continue with the 2-stage DTR model in Example~\ref{exp:_5_3_dtr}. We also apply Thm.~\ref{thm:_5_3_cbound2} to bound the expected reward $\invE{Y\mid s_1, s_2}{x_1, x_2}$. Precisely,
	\begin{align}
		\invE{Y\mid s_1, s_2}{x_1, x_2} & \geq \frac{\E\left[Y \mid s_1, s_2, x_1, x_2\right] P\left(s_1, s_2, x_1, x_2 \right)}{\Gamma(s_1, s_2, x_1)}                                         \\
		                                & \geq  \frac{\E\left[Y \mid s_1, s_2, x_1, x_2\right] P\left(s_1, s_2, x_1, x_2 \right)}{P(s_1, s_2, x_1) - P(s_1, x_1) + P(s_1)}  \label{eq:_5_3_dtr3}
	\end{align}
	The last step follows from the evaluation of $\Gamma(s_1, s_2, x_1)$ in Eq.~\ref{eq:_5_3_dtr2}. Similarly,
	\begin{align}
		\invE{Y\mid s_1, s_2}{x_1, x_2} & \leq 1 -  \frac{ \left( 1 - \E\left[Y \mid s_1, s_2, x_1, x_2\right] \right) P\left(s_1, s_2, x_1, x_2 \right)}{\Gamma(s_1, s_2, x_1)}                                       \\
		                                & \leq 1 -  \frac{ \left( 1 - \E\left[Y \mid s_1, s_2, x_1, x_2\right] \right) P\left(s_1, s_2, x_1, x_2 \right)}{P(s_1, s_2, x_1) - P(s_1, x_1) + P(s_1)} \label{eq:_5_3_dtr4}
	\end{align}
	We evaluate the conditional reward $\invE{Y\mid s_1, s_2}{x_1, x_2}$ directly in the underlying SCM $\1M^*$, compute their corresponding causal bounds derived above and provide them in Table~\ref{tab:_5_dtr_b}. \hfill $\blacksquare$
\end{example}

\begin{algorithm}[t]
	\caption{Upper Confidence Bound with Causal Bounds (\texttt{UCB}\textsuperscript{+}) }
	\label{alg:_5_3_ucb+}
	\begin{algorithmic}[1]
		\Require a policy space $\Pi = \Braces{\Tuple{X_i, \*S_i}}_{i = 1}^H$, a reward function $\1R: \D(\*Y) \mapsto [0, 1]$, causal bounds $\left [l_{\bar{\*x}_i}(\bar{\*s}_{i+1}), r_{\bar{\*x}_i}(\bar{\*s}_{i+1}) \right ]$, $i = 1, \dots, H-1$, and $\left [l_{\bar{\*x}_H}(\bar{\*s}_H), r_{\bar{\*x}_H}(\bar{\*s}_H) \right ]$ for all state-action pairs $(\*x, \*s) \in \D(\*X\cup \*S)$.
		\State Let $\3M_{\text{c}}$ be the set of all SCMs $\1M$ with endogenous variables $\*V$, and with the transition distribution $\inv{\*S_{i+1} \mid \bar{\*s}_{i}; \1M}{\bar{\*x}_{i}}$ compatible with bounds $\left [l_{\bar{\*x}_i}(\bar{\*s}_{i+1}), r_{\bar{\*x}_i}(\bar{\*s}_{i+1}) \right ]$, $i = 0, \dots, H-1$, and rewards $\invE{Y\mid \bar{\*s}_H; \1M}{\bar{\*x}_H}$ compatible with  $\left [l_{\bar{\*x}_H}(\bar{\*s}_H), r_{\bar{\*x}_H}(\bar{\*s}_H) \right ]$, that is,
		\begin{align}
			\forall i = 0 \dots, H-1,\;\; & l_{\bar{\*x}_i}(\bar{\*s}_{i+1}) \leq P_{\bar{\*x}_{i}}\left (\*s_{i+1} \mid \bar{\*s}_{i}; \M \right) \leq r_{\bar{\*x}_i}(\bar{\*s}_{i+1}), \label{eq:_5_3_ci1} \\
			\text{and}\;\;                & l_{\bar{\*x}_H}(\bar{\*s}_H) \leq \E_{\bar{\*x}_n}\left [Y \mid  \bar{\*s}_n; \M \right] \leq  r_{\bar{\*x}_H}(\bar{\*s}_H).\label{eq:_5_3_ci2}
		\end{align}

		\ForAll{episodes $t = 1, 2, \dots $}
		\State Construct a set of plausible SCMs $\3M_{t-1}(\delta)$ with $\delta = t^{-4}$ following Steps 2-4 of Alg.~\ref{alg:_5_2_ucb}.
		\State Find the optimal policy $\pi^{(t)}$ of an optimistic SCM $\1M^{(t)} \in \3M_{t-1}(\delta) \cap \3M_{\text{c}}$ such that
		\begin{align}
			 & \E_{\pi^{(t)}}\left [Y ; \1M^{(t)} \right]= \max_{\pi, \1M}\E_{\pi}\Big [Y ; \1M \Big] \;\; \text{s.t.} \;\; \pi \in \Pi, \1M \in \3M_{t-1}(\delta) \cap \3M_{\text{c}}. \label{eq:_5_3_optimize}
		\end{align}
		\State Perform $\doo\Parens{\pi^{(t)}}$ for episode $t$ and receive observations $\*V^{(t)}$.
		\EndFor
	\end{algorithmic}
\end{algorithm}

We are ready to introduce a generalized \texttt{UCB}\textsuperscript{+} algorithm utilizing causal bounds in the sequential decision-making setting. Alg.~\ref{alg:_5_3_ucb+} summarizes the details of its implementation. It takes as input arguments a policy space $\Pi$, and bounds over transition probabilities $\left [l_{\bar{\*x}_i}(\bar{\*s}_{i+1}), r_{\bar{\*x}_i}(\bar{\*s}_{i+1}) \right ]$, $i = 1, \dots, H-1$, and rewards $\left [l_{\bar{\*x}_H}(\bar{\*s}_H), r_{\bar{\*x}_H}(\bar{\*s}_H) \right ]$ computed from the observational distribution $P(\*V)$, following the derivation in Thms.~\ref{thm:_5_3_cbound1} and \ref{thm:_5_3_cbound2}. More specifically, in Step 1, \texttt{UCB}\textsuperscript{+} constructs a family $\3M_{\text{c}}$ of plausible SCMs with transition distributions $\inv{\*s_{i+1}\mid \bar{\*s}_i}{\bar{\*x}_i} \in \left [l_{\bar{\*x}_i}(\bar{\*s}_{i+1}), r_{\bar{\*x}_i}(\bar{\*s}_{i+1}) \right ]$ and with rewards $\invE{Y\mid \bar{\*s}_H}{\bar{\*x}_H} \in \left [l_{\bar{\*x}_H}(\bar{\*s}_H), r_{\bar{\*x}_H}(\bar{\*s}_H) \right ]$ compatible with the provided causal bounds. Since bounds in Thms.~\ref{thm:_5_3_cbound1} and \ref{thm:_5_3_cbound2} are sound and the observational data are often abundant, this ensures that the underlying SCM $\1M^* \in \3M_{\text{c}}$ with high probability. For every episode $t$, it computes the optimal policy $\pi^{(t)}$ of an optimistic SCM $\M^{(t)}$ in set intersection $\3M_t \cap \3M_{\text{c}}$ (Step 4). The construction of the SCM family $\3M_t$ follows Steps 2-4 of \texttt{UCB} defined in Alg.~\ref{alg:_5_2_ucb}. Similar to the optimistic planning procedure described previously, $\pi^{(t)}$ could be obtained by solving a polynomial program with the objective Eq.~\ref{eq:_5_2_id1}, subject to interventional constraints of Eqs.~\ref{eq:_5_2_ci1} and \ref{eq:_5_2_ci2}, and additional constraints of Eqs.~\ref{eq:_5_3_ci1} and \ref{eq:_5_3_ci2} imposed by causal bounds.

The causal bounds in over transition distributions $\left [l_{\bar{\*x}_i}(\bar{\*s}_{i+1}), r_{\bar{\*x}_i}(\bar{\*s}_{i+1}) \right ]$, $i = 1, \dots, H-1$, and rewards $\left [l_{\bar{\*x}_H}(\bar{\*s}_H), r_{\bar{\*x}_H}(\bar{\*s}_H) \right ]$ also permits a partial identification strategy to bound the expected rewards of candidate policies $\pi \in \Pi$. Formally, the expected reward $\invE{\1R(\*Y)}{\pi} \in [l_{\pi}, r_{\pi}]$ such that
\begin{align}
	 & l_{\pi} = \min_{\1M \in \3M_{\text{c}}} \invE{\1R(\*Y); \1M}{\pi}, &  & r_{\pi} = \max_{\1M \in \3M_{\text{c}}} \invE{\1R(\*Y); \1M}{\pi}.
\end{align}
where $\3M_{\text{c}}$ is the set of all SCMs compatible with constraints imposed by causal bounds. Interestingly, these bounds $[l_{\pi}, r_{\pi}]$ also characterize conditions under which the confounded observational data accelerate the performance of online learning algorithms.
\begin{restatable}{theorem}{thmucbdtrplus}\label{thm:_5_3_ucb+}
	Let $\langle \1M^*, \Pi, \1R \rangle$ be a CDM where $\Pi = \left \{\langle X_i, \*S_i \rangle \right \}_{i=1}^H$ and $\1R: \D(\*Y) \mapsto [0, 1]$. The regret of \texttt{UCB}\textsuperscript{+} in SCM $\1M^*$ after $T > 1$ episodes is bounded by
	\begin{align}
		R(T, \1M^*) \leq  \max_{\pi \in  \Pi^{\mathrm{o}}: \substack{ \Delta_{\pi} > 0, \\ r_{\pi} \geq \mu^*}} {17^2 H^2 \left | \D(\*X \cup \*S) \right| \log(T) \over \Delta_{\pi}} + \frac{\pi^2}{6}. \label{eq:_5_3_regret}
	\end{align}
	where $\mu^* = \invE{\1R(\*Y); \1M^*}{\pi^*}$ is the expected reward of an optimal policy $\pi^* \in \Pi$. \hfill $\blacksquare$
\end{restatable}
Thm.~\ref{thm:_5_3_ucb+} implies that \texttt{UCB}\textsuperscript{+}, utilizing the observational data, consistently dominates \texttt{UCB} (Alg.~\ref{alg:_5_2_ucb}) in terms of the performance. Broadly, it enjoys the same asymptotic regret bound as \texttt{UCB}, provided in Thm.~\ref{thm:_5_2_ucb}. When the causal bounds are informative, i.e., there exist some suboptimal policies $\pi$ with the upper bound $r_{\pi} < \mu_{\pi^*}$ smaller than the expected reward of the optimal policy $\pi^*$, \texttt{UCB}\textsuperscript{+} is able to outperform \texttt{UCB} that learns from scratch, without using any prior observations. For instance, consider a multi-armed bandit model with action $|\D(X)| = K$ and states $\*S = \emptyset$. The regret bound of \texttt{UCB}\textsuperscript{+} is $\mathcal{O}(K \log(T) / \Delta_{x})$ where $\Delta_{x}$ is the smallest gap among sub-optimal arms $x$ with causal upper bound $r_{x} \geq \mu^*$. The improvement condition matches the analytical result in Thm.~\ref{thm:_5_1_ucb}, derived for the special case of MAB models.

\begin{figure}[t]
	\centering
	\hfill
	\begin{minipage}[b]{0.33\linewidth}
		\centering
		\includegraphics[width=1.0\textwidth]{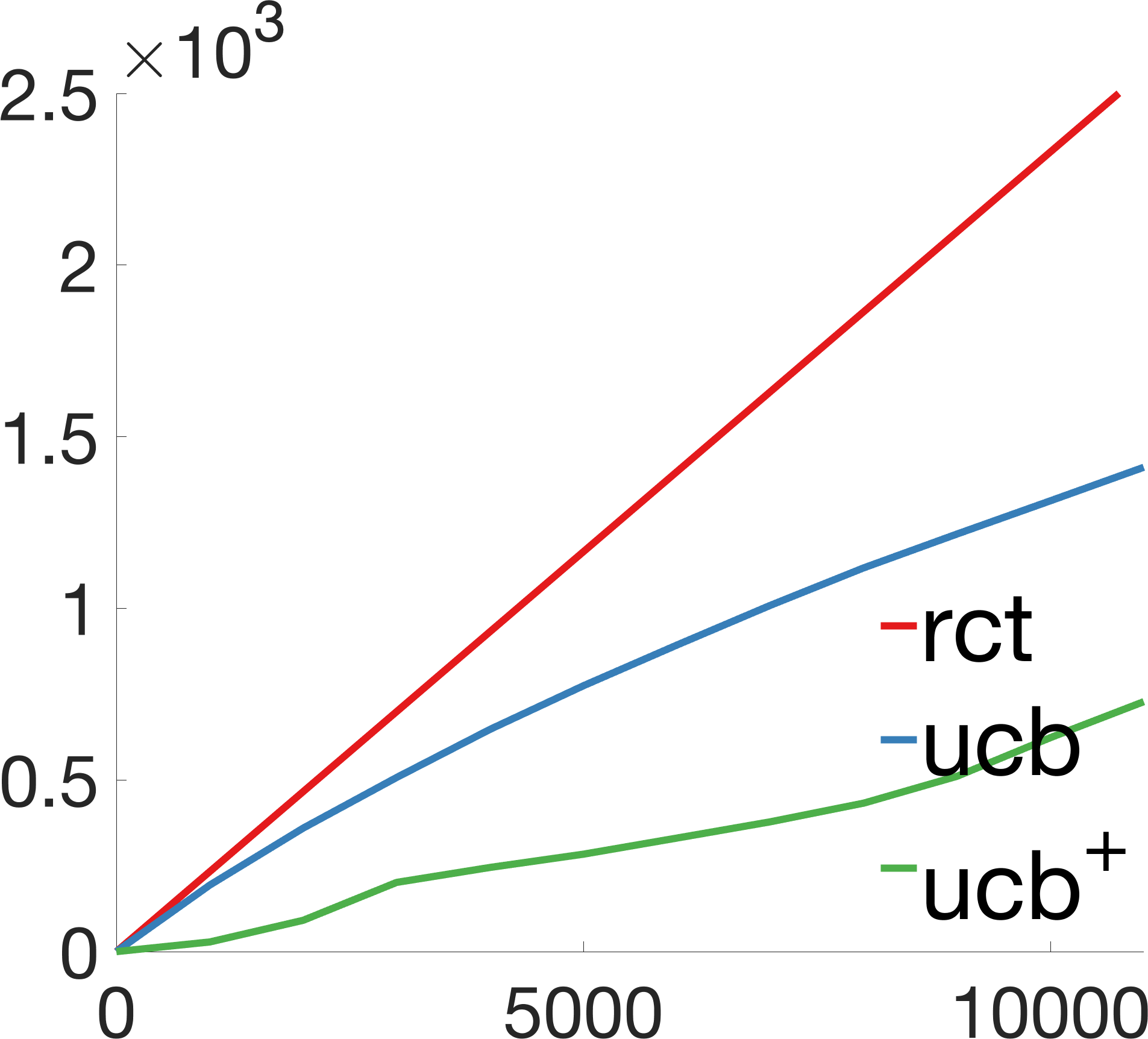}
		\subcaption{$\alpha_1 = 3, \alpha_2 = -3$}
		\label{fig:_5_3_dtr_a}
	\end{minipage}\hfill
	\begin{minipage}[b]{0.33\linewidth}
		\centering
		\includegraphics[width=1.0\textwidth]{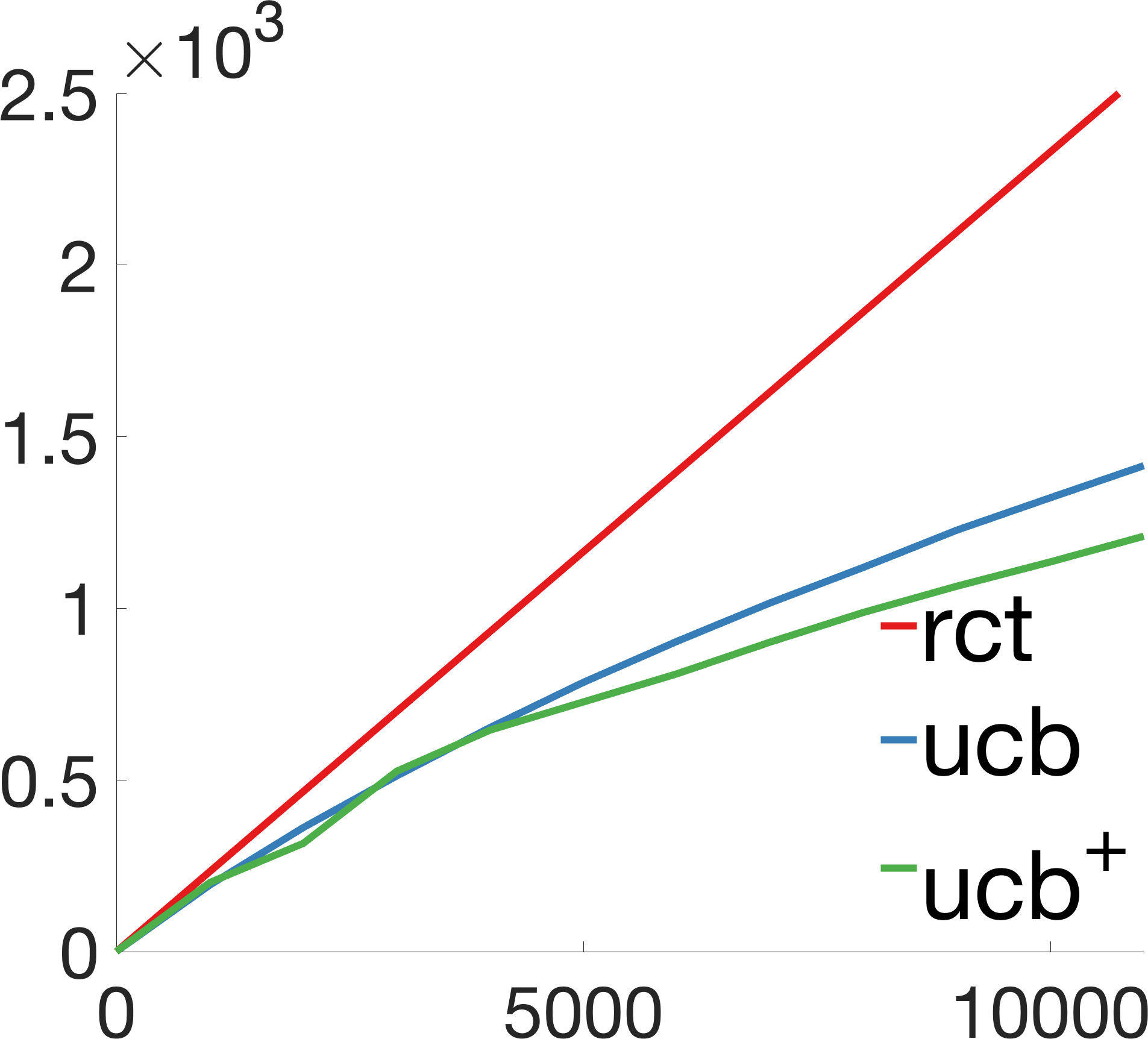}
		\subcaption{$\alpha_1 = -12, \alpha_2 = -3$}
		\label{fig:_5_3_dtr_b}
	\end{minipage}\hfill\null
	\caption{Simulation results comparing \texttt{UCB}\textsuperscript{+} learner augmented with causal bounds over the expected rewards, standard \texttt{UCB}, and \texttt{RCT} determining values of action uniformly at random.}
	\label{fig:_5_3_dtr}
\end{figure}

\begin{experiment}\label{exp:_5_3_dtr1}
	Fig.~\ref{fig:_5_3_dtr} shows the cumulative regret of \texttt{UCB}\textsuperscript{+} in the 2-stage DTR environment $\1M^*$ described in Example~\ref{exp:_3_1_dtr}. The setup is the same as in Experiment.~\ref{exp:_5_2_dtr}. It takes as input the causal bounds over $\inv{S_2\mid S_1}{X_1}$ and $\invE{Y\mid S_1, S_2}{X_1, X_2}$, computed from the observational distribution $P(S_1, X_1, S_2, X_2, Y)$. Simulation results show the significant disparity between the performance of \texttt{UCB}\textsuperscript{+} and \texttt{UCB} for $\Parens{\alpha_1, \alpha_2} = (3, -3)$. In this case, the causal bound for the expected reward of a suboptimal policy $\pi = (X_1 \gets 1, X_2 \gets 0)$ is $r = 0.6283$, which is smaller than the optimal expected reward $\mu_{\pi^*} = 0.6757$. On the other hand, when the coefficients $\Parens{\alpha_1, \alpha_2} = (-12, -3)$ and the causal bound $r = 0.9875 > \mu_{\pi^*}$, the performance of \texttt{UCB}\textsuperscript{+} and \texttt{UCB} coincides. These results corroborate the theory that the learning strategy of \texttt{UCB}\textsuperscript{+} enjoys no negative impact. \hfill $\blacksquare$
\end{experiment}

\begin{table}[!t]
	\setlength{\tabcolsep}{4.2pt}
	\centering
	\begin{tabular}{m{3cm}|m{2.5cm}|m{5.5cm}}
		\toprule
		                         \textbf{Decision Horizon}                                                          & \textbf{Algorithm} & \textbf{Regret Bound}                   \\ \midrule	                 \multirow{2}{1.5cm}[0pt]{$H = 1$}            & \texttt{UCB} (Alg.~\ref{alg:_4_2_ucb})                                                                                             & $\1O\Parens{|\D(X)| \log(T) / \Delta}$ \\[10pt] \cline{2-3}                                                                & \texttt{UCB}\textsuperscript{+}  (Alg.~\ref{alg:_5_1_ucb+})                                                                              & $\1O\Parens{|\D(X)| \log(T) / \Delta_*}$ \\[10pt] \midrule 
		 \multirow{2}{1.5cm}[0pt]{$H \geq 2$}                 & \texttt{UCB}  (Alg.~\ref{alg:_5_2_ucb})                                               &  $\1O\Parens{H^2 \left | \D(\*X \cup \*S) \right| \log(T) / \Delta}$\\[10pt] \cline{2-3}                                                                       & \texttt{UCB}\textsuperscript{+}  (Alg.~\ref{alg:_5_3_ucb+})                                         & $\1O\Parens{H^2 \left | \D(\*X \cup \*S) \right| \log(T) / \Delta_*}$                        \\[10pt] \bottomrule
	\end{tabular}
	\caption{Summary of \texttt{UCB} and \texttt{UCB}\textsuperscript{+} studied in  Secs.~\ref{sec:_4_1} and \ref{sec:_5_o2o}. The performance gap $\Delta \leq \Delta_*$.}
	\label{tab:_5_3_algs}
\end{table}

Table~\ref{tab:_5_3_algs} summarizes online learning and offline-to-online learning algorithms studied so far in Secs.~\ref{sec:_4_1} and \ref{sec:_5_o2o}. These algorithms could be categorized into two lines of learning strategies: \texttt{UCB} is an online algorithm that does not utilize any observational data, and \texttt{UCB}\textsuperscript{+} is an offline-to-online algorithm that leverages on observational data through causal bounds. More specifically,
	\begin{itemize}
		\item Consider first when the decision horizon $H = 1$ and the input covariates $\*S = \emptyset$. In this case, \texttt{UCB} achieves a regret bound $\1O\Parens{|\D(X)| \log(T) / \Delta}$ where $\Delta$ is the smallest performance gap between the optimal arm $x^*$ and a suboptimal arm $x$. Alg.~\ref{alg:_4_2_ucb} shows its implementation. On the other hand, \texttt{UCB}\textsuperscript{+} utilizes the causal bounds and achieves a regret bound $\1O\Parens{|\D(X)| \log(T) / \Delta_*}$ where $\Delta_*$ the smallest performance gap between the optimal arm $x^*$ and a suboptimal arm $x$ with a causal bound $r_x \geq \mu_{x^*}$. Alg.~\ref{alg:_5_1_ucb+} shows its implementation. Since by definition, the performance gap $\Delta \leq \Delta_*$,\texttt{UCB}\textsuperscript{+} performs at least as well as \texttt{UCB}. 
		\item We also studied settings where the decision horizon $H \geq 2$ and every actions $X_i$ is associated with a set of input covariates $\*S_i$ for $i = 1, \dots, H$. \texttt{UCB}, described in Alg.~\ref{alg:_5_2_ucb}, is able to achieve a regret bound $\1O\Parens{H^2 \left | \D(\*X \cup \*S) \right| \log(T) / \Delta}$ where $\Delta$ is the smallest performance gap between the optimal policy $\pi^*$ and a suboptimal deterministic policy $\pi$. Meanwhile, \texttt{UCB}\textsuperscript{+} (Alg.~\ref{alg:_5_3_ucb+}) exploits the causal bounds and enjoys a regret bound $\1O\Parens{H^2 \left | \D(\*X \cup \*S) \right| \log(T) / \Delta_*}$ where $\Delta_*$ is the smallest performance gap between the optimal policy $x^*$ and a suboptimal deterministic policy $x$ with a causal bound $r_{\pi} \geq \mu_{\pi^*}$. Again, since the performance gap $\Delta \leq \Delta_*$, \texttt{UCB}\textsuperscript{+} generally outperforms \texttt{UCB} in sequential settings. 
	\end{itemize}
After all, our analytical results reveal that causal bounds are robust to the confounding bias in the observational data, and could consistently improve the performance of online learners. 

\section{Mixed Policy Learning: Where to Intervene (CRL Task 2)}\label{sec:_6_wheredo}

Agents are deployed in complex and uncertain environments, where they are exposed and need to process high volumes of information while being expected to operate efficiently, surgically, and safely.
This requires the agent to identify an optimal policy to bring about a desirable state of affairs. 
A prevalent assumption in the literature, including the discussion in the previous sections, is that the action space is fixed. 
For example, it could be defined over variables $\*X$, where $|\*X| = k$. 
This implies that the agent will explore policies within the domain of $\*X$, for example, encompassing $2^k$  possible configurations in the binary case. 

In this section, we will relax the assumption that the policy scope is fixed and explore more flexible action spaces, including situations where the agent is not required to perform interventions. 
This is motivated by the observation that such a strategy, while viable in controlled, artificial environments  (e.g., actions executed in a simulator or gaming scenarios), where interventions are harmless, becomes less ideal and sometimes infeasible in real-world settings due to their potentially harmful side effects. 

Another complementary property, perhaps surprisingly, in non-Markovian causal systems is that controlling all intervenable variables, denoted as $do(\*X \gets \*x)$, does not necessarily lead to an optimal policy. 
In particular, we will show that in certain settings, a partial intervention, where $do(\*X' \gets \*x')$ with $\*X' \subset \*X$, can outperform the case where full control is exerted. 
In such systems, a larger search would be required to examine all possible subsets of $\*X$, including $3^k$ possible configurations, which will need to be evaluated in a systematic manner. 

In real-world causal systems, it's interesting to recognize that full controllability is not always necessary. 
For instance, natural mechanisms often govern the action variables $\*X$ in many scenarios, meaning that forcing the variable to take some value by external interventions might lead to undesirable effects. 
Robots inevitably obey the laws of physics, such as inertia and gravity, which makes their joints move naturally when their gears are disengaged; similarly, physicians treat patients based on their experience while adhering to varying rules and regulations by location. 
Therefore, in certain situations, it may be sufficient to intervene in only a subset of variables among $\*X$, allowing the remaining variables to vary according to their natural dynamics, as determined by the underlying mechanism $f_X$. 
Fig.~\ref{fig:_6_mpl} illustrates these dynamics, where the agent performs interventions over different sets of variables in each episode, and sometimes just acts naturally. 

\begin{figure}[t]
	\centering
	\centering
	\begin{tikzpicture}
		\def\outerr{3.5}
		\def\innerr{3.5}
		\def\dist{3.5}

		\draw[->, >={Latex}] (-1.5,0) -- (3*\dist+2,0) node[below] {Episode t};

		\node[vertex, minimum width=6mm] (X1) at (0, 2.2) {X\textsuperscript{(1)}};
		\node[vertex, minimum width=6mm] (Y1) at (1.25, 2.2) {Y\textsuperscript{(1)}};
		\node[vertex, minimum width=6mm] (Z1) at (-1.25, 2.2) {Z\textsuperscript{(1)}};
		\node[vertex, minimum width=6mm] (S1) at (0, 3.4) {S\textsuperscript{(1)}};
		\draw[bidir] (S1) to [bend right = 30] (X1);
		\draw[bidir] (S1) to [bend right = 30] (Z1);
		\draw[dir] (X1) -- (Y1);
		\draw[dir] (S1) -- (Y1);
		\draw[dir] (Z1) -- (X1);

		\node[vertex, minimum width=6mm] (X2) at (\dist, 2.2) {X\textsuperscript{(2)}};
		\node[vertex, minimum width=6mm] (Y2) at (\dist+1.25, 2.2) {Y\textsuperscript{(2)}};
		\node[vertex, minimum width=6mm] (Z2) at (\dist-1.25, 2.2) {Z\textsuperscript{(2)}};
		\node[vertex, minimum width=6mm] (S2) at (\dist, 3.4) {S\textsuperscript{(2)}};
		\draw[dir, betterblue] (S2) -- (X2);
		\draw[bidir] (S2) to [bend right = 30] (Z2);
		\draw[dir] (X2) -- (Y2);
		\draw[dir] (S2) -- (Y2);

		\node[vertex, minimum width=6mm] (X3) at (2*\dist, 2.2) {X\textsuperscript{(3)}};
		\node[vertex, minimum width=6mm] (Y3) at (2*\dist+1.25, 2.2) {Y\textsuperscript{(3)}};
		\node[vertex, minimum width=6mm] (Z3) at (2*\dist-1.25, 2.2) {Z\textsuperscript{(3)}};
		\node[vertex, minimum width=6mm] (S3) at (2*\dist, 3.4) {S\textsuperscript{(3)}};
		\draw[dir, betterblue] (S3) -- (Z3);
		\draw[bidir] (S3) to [bend right = 30] (X3);
		\draw[dir] (X3) -- (Y3);
		\draw[dir] (S3) -- (Y3);
		\draw[dir] (Z3) -- (X3);

		\node[vertex, minimum width=6mm] (X4) at (3*\dist, 2.2) {X\textsuperscript{(4)}};
		\node[vertex, minimum width=6mm] (Y4) at (3*\dist+1.25, 2.2) {Y\textsuperscript{(4)}};
		\node[vertex, minimum width=6mm] (Z4) at (3*\dist-1.25, 2.2) {Z\textsuperscript{(4)}};
		\node[vertex, minimum width=6mm] (S4) at (3*\dist, 3.4) {S\textsuperscript{(4)}};
		\draw[dir, betterblue] (S4) -- (Z4);
		\draw[dir] (X4) -- (Y4);
		\draw[dir] (S4) -- (Y4);
		\draw[dir, betterblue] (Z4) -- (X4);

		\node (d1) at (0, 1.4) {$\*V^{(1)} \sim P\left(\*V\right)$};
		\node (d2) at (\dist, 1.4) {$\*V^{(2)} \sim \inv{\*V}{\pi^{(2)}}$};
		\node (d3) at (2*\dist, 1.4) {$\*V^{(3)} \sim \inv{\*V}{\pi^{(3)}}$};
		\node (d4) at (3*\dist, 1.4) {$\*V^{(4)} \sim \inv{\*V}{\pi^{(4)}}$};

		\draw[very thick, betterblue, -] (0, 0) -- (0, 0.2);
		\draw[very thick, betterblue, -] (\dist, 0) -- (\dist, 0.2);
		\draw[very thick, bettergreen, -] (2*\dist, 0) -- (2*\dist, 0.2);
		\draw[very thick, bettergreen, -] (3*\dist, 0) -- (3*\dist, 0.2);

		\node [below] at (0, -0.05) { 0};
		\node [below] at (\dist, -0.05) { 1};
		\node [below] at (2*\dist, -0.05) { 2};
		\node [below] at (3*\dist, -0.05) { 3};

		\node [fill=betterblue!45] at (0, 0.6) {see};
		\node [fill=bettergreen!45] at (\dist, 0.6) {\small $\doo(\pi_X)$};
		\node [fill=bettergreen!45] at (2*\dist, 0.6) {\small $\doo(\pi_Z)$};
		\node [fill=bettergreen!45] at (3*\dist, 0.6) {\small $\doo(\pi_{X, Z})$};

		\begin{pgfonlayer}{back}
			\node[circle,fill=betterred!65,draw=none,minimum size=2*\innerr mm] at (Y1) {};
			\draw[fill=betterblue!25, draw = betterblue!45] \convexpath{S2, X2}{\outerr mm};
			\node[circle,fill=betterblue!65,draw=none,minimum size=2*\innerr mm] at (X2) {};
			\node[circle,fill=betterred!65,draw=none,minimum size=2*\innerr mm] at (Y2) {};
			\draw[fill=betterblue!25, draw = betterblue!45] \convexpath{S3, Z3}{\outerr mm};
			\node[circle,fill=betterblue!65,draw=none,minimum size=2*\innerr mm] at (Z3) {};
			\node[circle,fill=betterred!65,draw=none,minimum size=2*\innerr mm] at (Y3) {};
			\draw[fill=betterblue!25, draw = betterblue!45] \convexpath{Z4, S4}{1.1*\outerr mm};
			\draw[fill=betterblue!25, draw = betterblue!45] \convexpath{Z4, X4}{\outerr mm};
			\node[circle,fill=betterblue!65,draw=none,minimum size=2*\innerr mm] at (Z4) {};
			\node[circle,fill=betterblue!65,draw=none,minimum size=2*\innerr mm] at (X4) {};
			\node[circle,fill=betterred!65,draw=none,minimum size=2*\innerr mm] at (Y4) {};
		\end{pgfonlayer}
	\end{tikzpicture}
	\caption{Temporal diagram showing the dynamics of a mixed policy learning while the agent interacts with the environment with different policy scopes at each episode.}
	\label{fig:_6_mpl}
\end{figure}
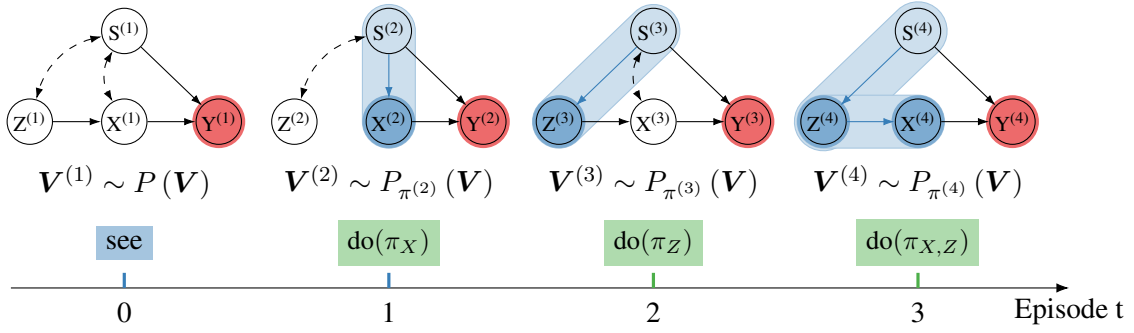

Towards formalizing this setting, we put these observations together and define a \textit{mixed policy space}, which is a collection of policy spaces in which each action and context are defined as subsets of intervenable variables $\*X^\star$ and context variables $\*S^\star$, respectively. That is,
\[\Pi_{\textsc{mix}} = \set{\set{\tuple{X, \*S_X}}_{X\in\*X'} : \*X'\subseteq \*X^\star, \*S_X\subseteq \*S^\star}.\] 
Every policy space in $\Pi_{\textsc{mix}}$ lies between two extremes --- observational and experimental policies.

\begin{definition}[Mixed Policy Space \& Policy]\label{def:mixed-policy}
Let $\G$ be a causal diagram, $Y\in \*V(\G)$ be a reward variable. 
	 $\*X^\star \subseteq \*V\setminus\set{Y}$ be a set of intervenable variables, 
	 and $\*S^\star\subseteq \*V\setminus \set{Y}$ be	a set of context variables. 
A mixed policy space $\Pi_{\textsc{mix}}$ is the collection of policy spaces where each policy space $\Pi \in \Pi_{\textsc{mix}}$ is defined with actions $\*X \subseteq \*X^\star$ and contexts $\*S \subseteq \*S^\star$ such that $\Pi = \set{\tuple{X_i, \*S_i}}_{i \mid X_i\in \*X}$, $\*S = \bigcup_{i\mid X_i\in\*X} \*S_i$, and $\G_{\Pi}$ is a DAG.	
 Given a mixed policy space $\Pi_{\textsc{mix}}$ with respect to $\tuple{\G, Y, \*X^\star, \*S^\star}$, 
	a \emph{mixed policy} $\pi \in \Pi \in \Pi_{\textsc{mix}}$ is a policy $\pi$ following the policy space $\Pi$. \hfill $\blacksquare$
\end{definition}
For simplicity, we may use $\pi\in \Pi_{\textsc{mix}}$, which is the shorthand notation for $\pi \in \Pi \in \Pi_{\textsc{mix}}$.
The following task signature characterizes this learning setting involved in a mixed policy space: 
\[
	\1T_{\textsc{mix}} = \Tuple{\1I = \text{do}, \1A= \G, \Pi_{\textsc{mix}} = \Braces{\Braces{\Tuple{ X_i, \*S_i } \mid \*S_i\subseteq \*S^\star }_{X_i \in \*X}}_{\*X\subseteq \*X^\star}, \1R = \D(Y)\mapsto \mathbb{R} }.
\]
Note that the policy space is not fixed but an element of a mixed policy space. This means that the agent will search for a policy $\pi^*$ such that
\begin{align}\label{eq:opt-crl-mixed}
	 & \pi^* = \argmax_{\textcolor{red}{\pi \in \Pi_{\textsc{mix}}}} \invEE{ \1R \left (\*Y \right )  \;\bigg\vert \;\G,  \;\mathcal{D}_{\text{exp}} \sim \inv{\*V}{\*x}  }{\pi}{\1M^*}, 
\end{align}
where the distinct feature of the task is the mixed policy scope. 

In Section~\ref{sec:mixed-policy-with-no-context}, we introduce the marginal case of mixed policy spaces, where no context variables are present. Even without considering contexts, the problem of deciding where the agent should intervene in the system is challenging and sets the ground for further explorations.
We then investigate in Section~\ref{sec:mixed-policy-with-context} mixed policy spaces with context variables, or where the agent should intervene (do) and look (see) to determine the optimal policy. 

\subsection{Mixed Policy with No Context}\label{sec:mixed-policy-with-no-context}
In this section, we investigate how an agent should behave to efficiently identify an optimal action given a mixed policy space and an underlying causal diagram. 
For simplicity, we focus on MAB settings where each policy $\pi \in \Pi_{\text{mix}}$ is an experimental policy $(\pi_1, \dots, \pi_{|{\*X}'|})$ over a subset of actions ${\*X}' \subseteq \*X^\star$ with each decision rule $\pi(x)$ involved in an empty context. Thus, $\pi(\*x') \in \Pi_{\text{mix}}$ for $\*x'\in \D(\*X')$ and $\*X'\subseteq \*X^\star$.
The following example illustrates the challenge of this task.

\begin{figure}[t]
    \begin{subfigure}{.29\textwidth}
        \begin{tikzpicture}[x=10mm, y=24mm]
            \def\outerr{3.2}
            \def\innerr{3}
            \node[] (XX) at (-1.5,+0.2) {$\*X$};
            \node[vertex] (X) {$X_2$};
            \node[vertex] (Z) at (-.6,1) {$X_1$};
            \node[vertex,dashed] (U) at (1,1) {$U$};
            \node[vertex] (Y) at (2,0) {$Y$};
            \draw[->] (Z) -- (X);
            \draw[->] (X) -- (Y);
            \draw[->,dashed] (U) -- (X);
            \draw[->,dashed] (U) -- (Y);
    		\path (Z) -- (X) coordinate[midway] (midpoint);
    		\draw[rotate around={33:(midpoint)}] ($(Z.north west)+(0.55,0.12)$) rectangle ($(X.south east)+(-0.56,-0.1)$);
        \end{tikzpicture}
        \caption{}
        \label{fig:mixed-iv-model-example-a}
    \end{subfigure}\hfill
    \begin{subfigure}{.2\textwidth}
        \begin{tikzpicture}[x=7mm, y=19mm]
            \def\outerr{3.2}
            \def\innerr{3}
            \node[vertex] (X)  at (-.6,-0.75) {$X_2$};
            \node[vertex] (Z) at (-.6,0.75) {$X_1$};
            \node[vertex,dashed] (U) at (1,1) {$U$};
            \node[vertex] (Y) at (2,0) {$Y$};
            \draw[->] (Z) -- (Y);
            \draw[->] (X) -- (Y);
            \draw[->,dashed] (U) -- (Y);
        \end{tikzpicture}
        \caption{}
        \label{fig:mixed-iv-model-example-b}
    \end{subfigure}\hfill
    \begin{subfigure}{.23\textwidth}
	\begin{tikzpicture}[
	    vertex/.style={draw, thick, rectangle, rounded corners, fill=gray!20, minimum size=3mm, align=center, text centered},
	    line/.style={thick},
	    set/.style={draw, rounded corners, dashed, inner sep=4pt}
	]

	\node[vertex] (E) {\scriptsize $\{\}$};
	\node[vertex] (X) [below left=1cm and 0.25cm of E] {\scriptsize $\{X_1\}$}; 
	\node[vertex] (Z) [below right=1cm and 0.25cm of E] {\scriptsize $\{X_2\}$}; 
	\node[vertex] (XZ) [below=2.8cm of E] {\scriptsize $\{X_1, X_2\}$}; 

	\draw[line] (Z) -- (XZ);
	\draw[line] (X) -- (XZ);
	\draw[line] (E) -- (X);
	\draw[line] (E) -- (Z);

    \begin{pgfonlayer}{back}
        \node[set, fit=(X) (Z)] {};
    \end{pgfonlayer}
    
	\end{tikzpicture}
    \caption{}
    \label{fig:mixed-iv-policy-space}
    \end{subfigure}\hfill
    \begin{subfigure}{.27\textwidth}\centering
    \includegraphics[width=\textwidth]{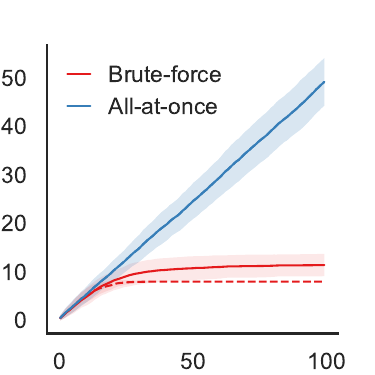}
    \caption{}
    \label{fig:mixed-policy-basic-plot}
    \end{subfigure}
    \caption{
(a)    True causal diagram $\G$ for environment $\1M^*$. (b) Hypothesized model in the agent's mind, after intervention. (c) Structure of the action space. (d) Two agents' cumulative regret with Thompson sampling (solid line) and UCB (dashed line) together with shaded areas representing 95\% confidence interval. Two lines for the All-at-once agent are overlapped.
    }
    \label{fig:mixed-iv-model-example}
\end{figure}
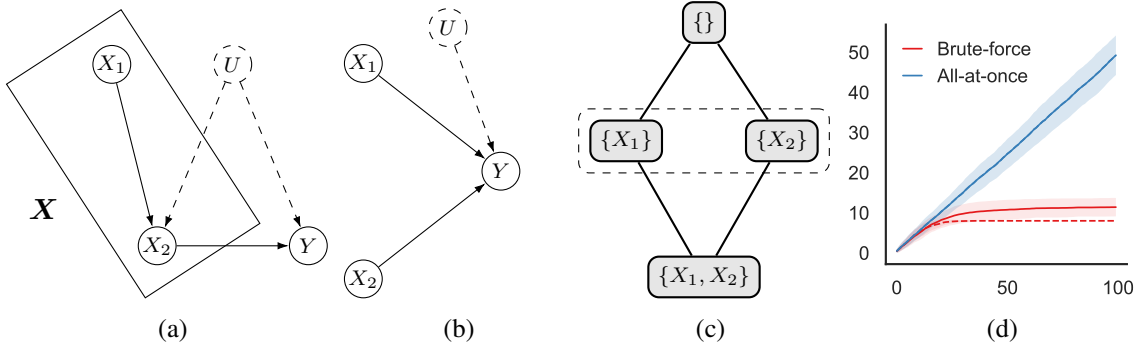

\begin{example}[Where to intervene]
	Consider a MAB environment described by an SCM
	\begin{align}
		\1M^* = \langle \*U = \{U_1, U_2\}, \*V =  \{X_1, X_2, Y\}, \2F, P(U_1, U_2)\rangle,
	\end{align}
	where the endogenous variables $\*V$ are all binary. 
	The causal mechanisms are the following:
\begin{align}
    \2F = \begin{cases}
            X_1     &\gets U_{1}, \\
            X_2     &\gets X_1\oplus U_{2}, \\     
            Y       &\gets X_2 \oplus U_{2}, 
       \end{cases}\label{eq:mixed-policy-basic-func-iv}
\end{align}
and the exogenous distribution: 
\begin{align}
 P(U_1) = P(U_2) = 1/2. 
\end{align}
The causal diagram associated with $\1M^*$ is shown in Fig.~\ref{fig:mixed-iv-model-example-a}. 

We now consider an agent deployed in $\1M^*$ with the goal of optimizing the outcome variable $Y$ while being capable of intervening on (or controlling) variables $X_1, X_2$. Specifically, the goal of the agent is to find 
\begin{align}\label{eq:opt-crl-mixed-ex}
	 & \pi^* = \argmax_{x_1, x_2} \invE{ Y; \1M^* }{X_1 \gets x_1, X_2 \gets x_2} 
\end{align}
The structure of the mixed policy space is shown in Fig.~\ref{fig:mixed-iv-policy-space}, which highlights the various policy scopes available. 
In particular, this space represents the power set of action variables $\*X^\star=\set{X_1,X_2}$, which entails $2^{|\*X^\star|}$ possible \textit{intervention sets}. 
This setting may be translated to a traditional MAB instance such that each arm corresponds to intervening on a subset of $\set{X_1,X_2}$ to a specific value, which results in 9 arms in this case. \footnote{$\{\doo(\emptyset), \doo(X_1 \gets 0), \doo(X_1 \gets 1), \doo(X_2 \gets 0), \doo(X_2 \gets 1), \doo(X_1 \gets 0,X_2 \gets 0), \doo(X_1 \gets 0,X_2 \gets 1), \doo(X_1 \gets 1,X_2 \gets 0),  \doo(X_1 \gets 1, X_2 \gets 1)\}$} $^,$ \footnote{The same observation applies for MDPs or any more complex model, as discussed later on in the section.  }

We start our analysis with an agent that is oblivious to the causal structure underlying the action space and adheres to a strict experimentalist approach. 
In other words, this means that it will abstract away the causal diagram and perform interventions on one action variable $\*X = \{X_1, X_2\}$. 
We call the agent following such a strategy ``all-at-once,'' since all variables are intervened together. 
\footnote{Formally, this can be thought of as a specific instance of a cluster causal diagram, an object that has been studied in the literature; for a more detailed discussion, refer to  \citep{anand2021cluster}.} 
Each intervention on $\doo(\*X)$ corresponds to interventions on the lower-level variables $\{\doo(X_1 \gets 0,X_2 \gets 0), \doo(X_1 \gets 0,X_2 \gets 1), \doo(X_1 \gets 1,X_2 \gets 0),  \doo(X_1 \gets 1, X_2 \gets 1)\}$. 
Under an interventional regime, the causal structure compatible with this strategy is shown in Fig.~\ref{fig:mixed-iv-model-example-b}, which is clearly different from the environment's causal diagram. 

Despite what is in the agent's mind, or optimization function, it will still be evaluated by the underlying SCM $\1M^*$.  
The natural question that arises here is whether it is okay to be oblivious to the pair $(\G, \M)$; would it be sufficient to perform more interventions to make up for the ignorance of the causal structure? 
In other words, more samples from the $\doo(X_1 \gets x_1, X_2 \gets x_2)$ distribution should be sufficient to eventually learn an optimal policy? 

This agent is deployed in the environment $\1M^*$ and after some interventions,  it is able to learn a policy and obtain the following reward: 
\begin{align}
     \invE{Y}{x_1, x_2}&= \E[x_2 \oplus U_{X_2,Y}] = 0.5x_2 + 0.5(1-x_2) = 0.5. 
\end{align}
This policy is, in fact, independent of the specific values of $X_1$ and $X_2$, and $Y$ reaches its highest value at most 0.5 of the time. 
At this point, the expectation by some is that there is an issue of sample complexity, but not of asymptotic convergence. In other words, the agent can be oblivious to the causal structure, $\G$, compensating by accumulating more samples, but it would eventually be able to learn the optimal policy. 

Now we examine an alternative policy within the mixed policy space in which the agent controls only one variable $X_1$. 
The evaluation of such a policy goes as follows: 
\begin{align}
     \invE{Y}{x_1} &= \E[(x_1 \oplus U_{X_2,Y}) \oplus U_{X_2,Y}] = x_1
\end{align}
This means that if the goal is to keep $Y$ as high as possible, the agent should perform an intervention $\doo(X_1 \gets 1)$, which would imply that $Y = 1$ in each subsequent round. 

The implication of such a result is that the strategy ``all-at-once'',  oblivious to the environment's structure, will never converge, no matter how many interactions are allowed to the agent. 
We compare empirically this with an alternative strategy called ``brute-force'', which searches over the entire policy space, including all possible subsets of $\set{X_1,X_2}$. 
The performance of both agents is shown in Figure~\ref{fig:mixed-policy-basic-plot}, where the y-axis represents a cumulative regret. 
In fact, the all-at-once agent does not converge while the brute-force approach is able to find the optimal policy, since $\doo(X_1 \gets 1)$ is inside the mixed policy space. 

The question here is whether we can do better by leveraging the underlying causal invariances of $\1M^*$, as represented in $\G$. 
We answer this question by examining the expected rewards over the entire mixed policy space. 
First, using Rule 3 of do-calculus, we note that $P(y\mid \doo(x_1,x_2)) =  P(y\mid \doo(x_2))$, since $X_1$ has no effect on $Y$ under intervention on $X_2$.
That is, the corresponding expected rewards are equivalent, $\mu_{x_1,x_2} = \mu_{x_2}$, for any $x_1$ and $x_2$. 
Hence, $\mu^*_{X_1,X_2} = \mu^*_{X_2}$. 
Since the all-at-once strategy can be discarded, we examine 5 arms based on 3 intervention sets as follows: 
\begin{align}
    \E[Y]             &= \E[(X_1 \oplus U_{X_2,Y}) \oplus U_{X_2,Y}] = \E[X_1]=0.5\\
    \invE{Y}{x_1} &= \E[(x_1 \oplus U_{X_2,Y}) \oplus U_{X_2,Y}] = x_1\\
    \invE{Y}{x_2} &= \E[x_2 \oplus U_{X_2,Y}] = 0.5x_2 + 0.5(1-x_2) = 0.5.
\end{align}
Therefore, the optimal action is $\doo(X_1 \gets 1)$ for the model with $\mu_{X_1 \gets 1}^*=1$. Again, intervening $\set{X_1,X_2}$ will incur regrets and cannot converge to the optimal solution.

To explain this further, we now investigate the relationships among interventional probabilities in the causal model in Fig.~\ref{fig:mixed-iv-model-example-a}.
The observational probability $P(y)=P(y\mid \doo(\emptyset))$ can be viewed as a convex combination of $\set{P(y\mid \doo(x_1))}_{x_1\in \D(X_1)}$,
\begin{align}
    P(y) = \sum_{x_1} P(y\mid x_1)P(x_1) = \sum_{x_1} \inv{y}{x_1}P(x_1).
\end{align}
That is, $\mu_\emptyset = \sum_{x_1} \mu_{x_1} P(x_1)$. By replacing $\mu_{x_1}$ to $\mu^*_{X_1} = \max_{x_1} \mu_{x_1}$, then,
\begin{align}
    \mu_\emptyset = \sum_{x_1} \mu_{x_1} P(x_1) \leq \sum_{x_1} \mu^*_{X_1} P(x_1) = \mu^*_{X_1}.
\end{align}
This equation holds for any model conforming to the causal diagram in Fig.~\ref{fig:mixed-iv-model-example-a}, namely, it affects whether the agent should play some arms since playing non-optimal arms will incur regrets. 
At this point, playing the arms over $X_2$ is preferred to the arms over both $X_1$ and $X_2$ (i.e., $\doo(X_1,X_2)$), since it minimizes the number of arms that need to be played to find the optimal, and $\doo(X_1)$ is preferred to $\doo()$ as $\mu_\emptyset$ cannot be strictly better than the best achievable expected reward obtainable by intervening on $X_1$ to ${x^*_1}$.

\begin{figure}
    \footnotesize
    \centering
    \begin{tikzpicture}[x=1.6cm,y=.6cm]
        \node[anchor=east] (exp) at (-1, 0) {\text{interventions}};
        \node (DOE) {$\doo(\emptyset)$};
        \node (DOX) at (4,0) {$\doo(x_2)$};
        \node (DOZ) at (2,0) {$\doo(x_1)$};
        \node (DOXZ) at (6,0) {$\doo(x_1,x_2)$};
        
        \node[anchor=east] (dist) at (-1, -2) {\text{distributions}};
        \node (PE) at (0, -2) {$P(y,x_1,x_2)$};
        \node (PX) at (4, -2) {$P_{x_2}(y,x_1)$};
        \node (PZ) at (2, -2) {$P_{x_1}(y,x_2)$};
        \node (PXZ) at (6, -2) {$P_{x_1,x_2}(y)$};
        
        \node[anchor=east] (er) at (-1, -4) {\text{expected rewards}};
        \node[rectangle,fill=gray!20] (E) at (0,-4) {$\mu_\emptyset$};
        \node[rectangle,fill=gray!20] (X) at (4,-4)  {$\mu_{x_2}$};
        \node[rectangle,fill=gray!20] (Z) at (2,-4)  {$\mu_{x_1}$};
        \node[rectangle,fill=gray!20] (XZ) at (6,-4)  {$\mu_{x_1,x_2}$};
        
        \node (eqeq) at (1.75,-6) {$\sum_{x_1} \mu_{x_1} P(x_1)$};
        
        \draw[<->] (X) -- (XZ);
        \draw[->] (PE) -- (E);
        \draw[->] (PX) -- (X);
        \draw[->] (PZ) -- (Z);
        \draw[->] (PXZ) -- (XZ);
        
        \draw[->] (DOE) -- (PE);
        \draw[->] (DOX) -- (PX);
        \draw[->] (DOZ) -- (PZ);
        \draw[->] (DOXZ) -- (PXZ);
        
        \draw[->] (PE) -- (PZ) node [midway,above] {$P(y,x_2\mid x_1)$};
        \draw[->] (PX) -- (PXZ) node [midway,above] {$P_{x_2}(y)$};
        \draw[->] (Z) -- (eqeq);
        \draw[->] (PE) -- (eqeq) node [midway,above,sloped] {$P(x_1)$};
        \draw[->] (PX) -- (eqeq) node [midway,below,sloped] {$P(x_1)=P_{x_2}(x_1)$};
        \draw[->] (eqeq) -- (E);
    \end{tikzpicture}
    \caption{Relationships among quantities such as probability distributions and expected rewards arising in the mixed policy relative to causal model in Fig.~\ref{fig:mixed-iv-model-example-a}. }
    \label{fig:iv-model-rels}
\end{figure}
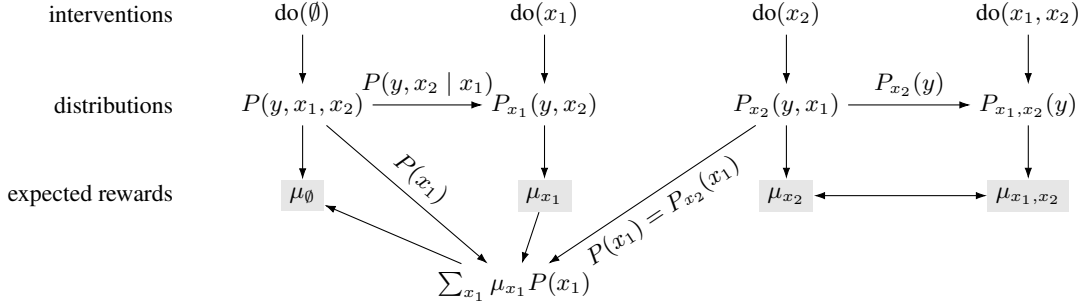

Regarding the superiority of intervening on a set of variables over other set of variables, a natural question is, then, whether the comparisons among $\mu_{x_2}$ and $\mu_{x_1}$ can be made as well, noting that $\mu^*_{X_2} < \mu^*_{X_1}$ in $\1M^*$. 
In fact, we can show that the inequality $\mu^*_{X_2} > \mu^*_{X_1}$ is also realizable.
To witness, consider an SCM $\1M'$ identical to the one defined previously (Eq.~\ref{eq:mixed-policy-basic-func-iv}) but for $Y$'s mechanism: 
\begin{align}
f_Y \leftarrow X_2 + U_{X_2,Y}.
\end{align}
Then, we can evaluate the expected rewards in $\1M'$ as follows: 
\begin{align*}
    \E[Y]             &= \E[(X_1 \oplus U_{X_2,Y}) + U_{X_2,Y}] = 1\\
    \invE{Y}{x_1} &= \E[(x_1 \oplus U_{X_2,Y}) + U_{X_2,Y}] = 0.5x_1 + 0.5(2-x_1) = 1  \\
    \invE{Y}{x_2} &= \E[x_2 + U_{X_2,Y}] = x_2 + 0.5
\end{align*}
Hence, the optimal action is $\doo(X_2 \gets 1)$ with $\mu^*_{X_2 \gets 1}=1.5$. 
This demonstrates the impossibility that arises in some cases of deciding a priori to prefer one interventional scope over the other solely based on the causal diagram, and this depends on the specific instantiation of the environment.  \hfill $\blacksquare$
\end{example}

Considering the example,  we note that ignoring the underlying causal structure, and the interplay between action space and reward, may result in a suboptimal performance due to playing such regret-incurring arms. 
If one is negligent to the influence of unobserved confounder and simply chooses to intervene on every variable, $\doo(x_1,x_2)$, or to intervene on the one closest to $Y$, $\doo(x_2)$, it is possible that the agent will never converge to the optimal arm, e.g., $\doo(x_1^*)$.

Exploring this example, we have shown the existence of equivalence classes among actions with respect to their expected rewards. 
Also, certain partial-orders emerge among subsets of action variables with respect to their optimal expected rewards. 
Also, the expected reward of an action is related to other actions, e.g., an observational probability written with probabilities from $\doo(x_1)$ and $\doo(x_2)$, $P(y)=\sum_{x_1} P_{x_1}(y)P(x_1)=\sum_{x_1} P_{x_1}(y)P_{x_2}(x_1)$.
Figure~\ref{fig:iv-model-rels} illustrates relationships among different distributions and their rewards.
Four different interventions are shown with their distributions where some distributions can yield other distributions, e.g., $P_{x_1}(y,x_2)$ from $P(y,x_1,x_2)$. Further, the expected reward for observation (mentioned above) can be represented as an expression made of probabilities from other arms.
We will later see that such a formula improves the performance of online learners. We now more formally investigate these phenomena as studied in \citep{LB:19a,LB:18a,lee2020characterizing}.

\subsubsection{Structural Properties in Mixed policy Learning in a MAB setting }
Here, we provide three structural properties emerging in a bandit setting with a mixed policy. These properties among  different actions arise due to the shared causal mechanisms and can be understood through do-calculus and related machinery.

\paragraph*{Property 1. Equivalence among actions}
Do-calculus provides rules to examine equivalence relationships in the space of conditional interventional distributions. Hence, it naturally partitions the space into equivalence classes.
In particular, we focus on Rule 3, which ascertains a graphical condition such that a set of interventions does not have an effect on the outcome variable, i.e., 
$P(y\mid  \doo(\*x, \*z), \*w)=P(y\mid  \doo(\*x), \*w)$. Since actions correspond to interventions (including the null intervention) and there is no contextual information, we consider examining $P(y\mid \doo(\*x,\*z))=P(y\mid \doo(\*x))$ through $\parens{Y\Perp \*Z \mid \*X}$ in $\G_{\overline{\*X\cup \*Z }}$, which implies that  $\mu_{\*x,\*z}=\mu_{\*x}$. 
If d-separation holds in the manipulated graph, this condition implies that it is sufficient to play only one action among actions in the equivalence class regarding finding an optimal arm efficiently. In an online learning setting where its objective is minimizing a cumulative regret, it is desired to play a smaller subset of arms given a set of arms as far as the subset contains the best arm. Against this background, we define a minimal intervention set.
\begin{definition}[Minimal Intervention Set (MIS)]
  \label{lb:def:MIS}A subset of action variables $\*X'\subseteq \*X^\star$
  is said to be a \textit{minimal intervention set} relative to $\G$, $\*X^\star$, and $Y$ if 
  there is no proper subset $\*X''\subset\*X'$
  such that $\mu_{\*x''}=\mu_{\*x'}$
  for every SCM conforming to $\G$ and $\*x''\in \D(\*X'')$ consistent with $\*x'$.  \hfill $\blacksquare$
\end{definition}
Whether a subset of action variables $\*X'\subseteq \*X^\star$ is an MIS can be examined through a rather simple procedure involving in an ancestral relationship.
\begin{proposition}[Minimality]
  \label{lb:prop:minimality}A set of variables $\*X'\subseteq\*X^\star$ is a minimal
  intervention set for $\G$ with respect to $Y$ if and only if $\*X'\subseteq an(Y)_{\G_{\overline{\*X'}}}$.  \hfill $\blacksquare$
\end{proposition}
This characterization demonstrates that one can consider only directed edges among variables, not the unobserved variables, in acquiring MISes. Intervening nothing or a single variable (an ancestor of $Y$ in $\G$) constitutes MISes. Further, intervening on $\*W = pa(\*Z)\setminus \*Z$ is also an MIS for any $\*Z\subseteq an(Y)_\G$ since each variable $W\in\*W$ has a directed path towards $Y$ without passing through the rest of $\*W$, i.e., $\*W\setminus\set{W}$.

\paragraph*{Property 2. Partial-orders among minimal intervention sets} 
We now explore the partial-orders among the subsets of $\*X^\star$ within the MISes. 
Given a causal diagram $\G$, it is possible that intervening on some variables is \emph{always} as good as intervening on another set of variables. Formally, there can be two different
sets of variables $\*W,\*Z\subseteq \*X^\star$
such that 
\[
    \max_{\*w\in \D(\*W)}\mu_{\*w}\leq\max_{\*z\in \D(\*Z)}\mu_{\*z}
\]
in every possible SCM conforming to $\G$. If that is the case, it would be unnecessary (and possibly harmful in terms of the sample efficiency) to play actions over $\D(\*W)$.
We define Possibly-Optimal MIS, which incorporates the partial-orderedness among MISes denoting the optimal value for $\*X'\subseteq\*X^\star$ given an SCM by $\*x'^{*}$.

\begin{definition}[Possibly-Optimal Minimal Intervention Set (POMIS)]
  \label{lb:def:POMIS}Given $\G$, $\*X^\star$ and $Y$, let $\*X' \subseteq \*X^\star$ be a MIS. If there exists an SCM 
  conforming to $\G$ such that $\mu_{\*x'^{*}}>\forall_{\*Z\in\mathbb{Z}{\setminus}\set{ \*X'} }\mu_{\*z^{*}}$, 
  where $\mathbb{Z}$ is the set of MISes with respect to $\G$, $\*X^\star$ and $Y$, 
  then $\*X'$ is a \textit{possibly-optimal minimal intervention
  set} with respect to $\G$, $\*X^\star$ and $Y$.  \hfill $\blacksquare$
\end{definition}
To determine whether intervening on a subset of $\*X^\star$ is a POMIS or not, one may list all possible partial-orders among MISes in a brute-force manner, and select those that are not dominated by any other MISes. However it is unclear whether we can compare two arbitrary MISes under what conditions and, further, whether such conditions are complete. As a starting point, we provide a way to obtain a single partial-order among two MISes. Consider intervening on an MIS $\*W$. By basic algebra, we can express the expected reward for $\doo(\*w)$ for some $\*Z\subseteq \*X^\star$:
\begin{align*}
    \invE{Y}{\*w}
            &=\sum_{\*z} \invE{Y \mid \*z }{\*w} \inv{\*z}{\*w}.
\intertext{If it is possible to exchange the observation $\*z$ to an intervention $\*z$ using Rule 2 of do-calculus, then the expression becomes}
           \invE{Y}{\*w} &= \sum_{\*z} \invE{Y}{\*z, \*w} \inv{\*z}{\*w}\\
            &\leq \sum_{\*z} \invE{Y}{(\*z, \*w)^*} \inv{\*z}{\*w}\\
            &= \invE{Y}{(\*z, \*w)^*}\\
            &= \invE{Y}{(\*z', \*w')^*}.
\end{align*}
where $\*Z'\cup\*W'$ is an MIS corresponding to the intervention set $\*Z\cup\*W$ and $^*$ over $(\*z,\*w)$ indicates the values for $(\*z,\*w)$ maximizing the expectation.

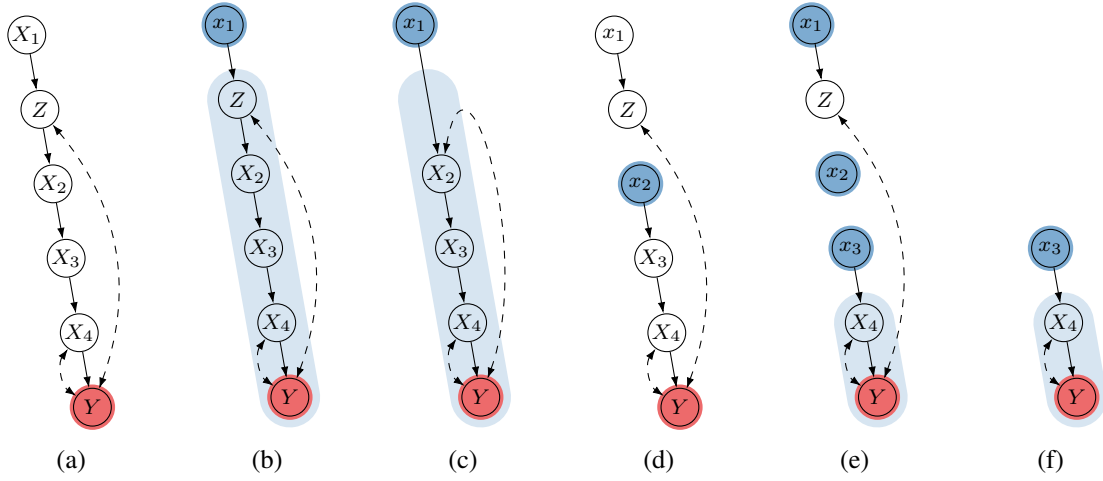
\begin{figure}
    \begin{subfigure}{.15\textwidth}\centering
        \begin{tikzpicture}
            \node[vertex] (A) at (100:5) {$X_1$};
            \node[vertex] (B) at (100:4) {$Z$};
            \node[vertex] (C) at (100:3) {$X_2$};
            \node[vertex] (D) at (100:2) {$X_3$};
            \node[vertex] (E) at (100:1) {$X_4$};
            \node[vertex] (Y) at (0,0) {$Y$};
            \draw[->] (A) -- (B);
            \draw[->] (B) -- (C);
            \draw[->] (C) -- (D);
            \draw[->] (D) -- (E);
            \draw[->] (E) -- (Y);
            \draw[<->,dashed] (B) to [bend left=30] (Y);
            \draw[<->,dashed] (E) to [bend right=45] (Y);
            \def\outerr{3.2}
            \def\innerr{3}
            \begin{pgfonlayer}{back}
                \node[circle,fill=betterred!65,draw=none,minimum size=2*\innerr mm] at (Y) {};
            \end{pgfonlayer}
        \end{tikzpicture}
        \caption{}
        \label{fig:pomis-example-simple-a}
    \end{subfigure}\hfill
    \begin{subfigure}{.15\textwidth}\centering
        \begin{tikzpicture}
            \node[vertex] (A) at (100:5) {$x_1$};
            \node[vertex] (B) at (100:4) {$Z$};
            \node[vertex] (C) at (100:3) {$X_2$};
            \node[vertex] (D) at (100:2) {$X_3$};
            \node[vertex] (E) at (100:1) {$X_4$};
            \node[vertex] (Y) at (0,0) {$Y$};
            \draw[->] (A) -- (B);
            \draw[->] (B) -- (C);
            \draw[->] (C) -- (D);
            \draw[->] (D) -- (E);
            \draw[->] (E) -- (Y);
            \draw[<->,dashed] (B) to [bend left=30] (Y);
            \draw[<->,dashed] (E) to [bend right=45] (Y);
            \begin{pgfonlayer}{back}
                \draw[fill=betterblue!20,draw=none] \convexpath{B,Y}{4mm};
            \end{pgfonlayer}
            \def\outerr{3.2}
            \def\innerr{3}
            \begin{pgfonlayer}{back}
                \node[circle,fill=betterblue!65,draw=none,minimum size=2*\innerr mm] at (A) {};
                \node[circle,fill=betterred!65,draw=none,minimum size=2*\innerr mm] at (Y) {};
            \end{pgfonlayer}
        \end{tikzpicture}
        \caption{}
        \label{fig:pomis-example-simple-b}
    \end{subfigure}\hfill
    \begin{subfigure}{.15\textwidth}\centering
        \begin{tikzpicture}
            \node[vertex] (A) at (100:5) {$x_1$};
            \node[vertex,opacity=0] (B) at (100:4) {$Z$};
            \node[vertex] (C) at (100:3) {$X_2$};
            \node[vertex] (D) at (100:2) {$X_3$};
            \node[vertex] (E) at (100:1) {$X_4$};
            \node[vertex] (Y) at (0,0) {$Y$};
            \draw[->] (A) -- (C);
            \draw[->] (C) -- (D);
            \draw[->] (D) -- (E);
            \draw[->] (E) -- (Y);
            \draw[<->,dashed] (C) to [out=80, in=70,looseness=1.5] (Y);
            \draw[<->,dashed] (E) to [bend right=45] (Y);
            
            \begin{pgfonlayer}{back}
                \draw[fill=betterblue!20,draw=none] \convexpath{B,Y}{4mm};
            \end{pgfonlayer}
            \def\outerr{3.2}
            \def\innerr{3}
            \begin{pgfonlayer}{back}
                \node[circle,fill=betterblue!65,draw=none,minimum size=2*\innerr mm] at (A) {};
                \node[circle,fill=betterred!65,draw=none,minimum size=2*\innerr mm] at (Y) {};
            \end{pgfonlayer}
    
        \end{tikzpicture}
        \caption{}
        \label{fig:pomis-example-simple-c}
    \end{subfigure}\hfill
    \begin{subfigure}{.15\textwidth}\centering
        \begin{tikzpicture}
            \node[vertex] (A) at (100:5) {$x_1$};
            \node[vertex] (B) at (100:4) {$Z$};
            \node[vertex] (C) at (100:3) {$x_2$};
            \node[vertex] (D) at (100:2) {$X_3$};
            \node[vertex] (E) at (100:1) {$X_4$};
            \node[vertex] (Y) at (0,0) {$Y$};
            \draw[->] (A) -- (B);
            \draw[->] (C) -- (D);
            \draw[->] (D) -- (E);
            \draw[->] (E) -- (Y);
            \draw[<->,dashed] (B) to [bend left=30] (Y);
            \draw[<->,dashed] (E) to [bend right=45] (Y);
    
            \def\outerr{3.2}
            \def\innerr{3}
            \begin{pgfonlayer}{back}
                \node[circle,fill=betterblue!65,draw=none,minimum size=2*\innerr mm] at (C) {};
                \node[circle,fill=betterred!65,draw=none,minimum size=2*\innerr mm] at (Y) {};
            \end{pgfonlayer}
        \end{tikzpicture}
        \caption{}
        \label{fig:pomis-example-simple-d}
    \end{subfigure}\hfill
    \begin{subfigure}{.15\textwidth}\centering
        \begin{tikzpicture}
            \node[vertex] (A) at (100:5) {$x_1$};
            \node[vertex] (B) at (100:4) {$Z$};
            \node[vertex] (C) at (100:3) {$x_2$};
            \node[vertex] (D) at (100:2) {$x_3$};
            \node[vertex] (E) at (100:1) {$X_4$};
            \node[vertex] (Y) at (0,0) {$Y$};
            \draw[->] (A) -- (B);
            \draw[->] (D) -- (E);
            \draw[->] (E) -- (Y);
            \draw[<->,dashed] (B) to [bend left=30] (Y);
            \draw[<->,dashed] (E) to [bend right=45] (Y);
            
            \begin{pgfonlayer}{back}
                \draw[fill=betterblue!20,draw=none] \convexpath{E,Y}{4mm};
            \end{pgfonlayer}
    
            \def\outerr{3.2}
            \def\innerr{3}
            \begin{pgfonlayer}{back}
                \node[circle,fill=betterblue!65,draw=none,minimum size=2*\innerr mm] at (A) {};
                \node[circle,fill=betterblue!65,draw=none,minimum size=2*\innerr mm] at (C) {};
                \node[circle,fill=betterblue!65,draw=none,minimum size=2*\innerr mm] at (D) {};
                \node[circle,fill=betterred!65,draw=none,minimum size=2*\innerr mm] at (Y) {};
            \end{pgfonlayer}
        \end{tikzpicture}
        \caption{}
        \label{fig:pomis-example-simple-e}
    \end{subfigure}\hfill
    \begin{subfigure}{.15\textwidth}\centering
        \begin{tikzpicture}
            \begin{scope}[opacity=0]
                \node[vertex] (A) at (100:5) {$x_1$};
                \node[vertex] (B) at (100:4) {$Z$};
                \node[vertex] (C) at (100:3) {$x_2$};
            \end{scope}
            \node[vertex] (D) at (100:2) {$x_3$};
            \node[vertex] (E) at (100:1) {$X_4$};
            \node[vertex] (Y) at (0,0) {$Y$};
            \draw[->] (D) -- (E);
            \draw[->] (E) -- (Y);
            \draw[<->,dashed] (E) to [bend right=45] (Y);
            
            \begin{pgfonlayer}{back}
                \draw[fill=betterblue!20,draw=none] \convexpath{E,Y}{4mm};
            \end{pgfonlayer}
            \def\outerr{3.2}
            \def\innerr{3}
            \begin{pgfonlayer}{back}
                \node[circle,fill=betterblue!65,draw=none,minimum size=2*\innerr mm] at (D) {};
                \node[circle,fill=betterred!65,draw=none,minimum size=2*\innerr mm] at (Y) {};
            \end{pgfonlayer}
        \end{tikzpicture}
        \caption{}
        \label{fig:pomis-example-simple-f}
    \end{subfigure}
    \caption{
        Illustrative examples demonstrating how partial-orders can be obtained from Figure~\ref{fig:pomis-example-simple-a}. (\subref{fig:pomis-example-simple-b} and \subref{fig:pomis-example-simple-c}) demonstrates no better intervention than $\doo(X_1)$ is obtained, (\subref{fig:pomis-example-simple-d} to \subref{fig:pomis-example-simple-f}) illustrates $\mu^*_{X_2} \leq \mu^*_{X_3}$. Light blue areas represent variables having a backdoor path to $Y$ under the intervention.
    }
    \label{fig:pomis-example-simple}
\end{figure}

To find such $\*Z\subseteq \*X^\star$ under the intervention on $\*W$, 
we examine Rule 2 of do-calculus: $\*Z$ must satisfy $(Y\Perp \*Z \mid \*W)$ in $\G_{\overline{\*W}\underline{\*Z}}$, which is equivalent to $(Y\Perp \*Z)$ in $(\G\setminus\*W)_{\underline{\*Z}}$. If no backdoor path from $\*Z$ to $Y$ exists in $\G\setminus \*W$, 
then $\mu^*_{\*W} \leq \mu^*_{\*W,\*Z} = \mu^*_{\*W',\*Z'}$. One may iteratively apply this idea to find a set which leads to a higher expected reward when intervened on until stuck at a graph where every non-intervened subset of $\*X^\star$ in the graph involves a backdoor path to $Y$.

A few examples are illustrated in Figure~\ref{fig:pomis-example-simple}. In the example with a causal diagram where none is intervened on (Figure~\ref{fig:pomis-example-simple-a}), every $X_i$ with $i>1$ has a backdoor path through $Z$ while $X_4$ is also directly confounded with $Y$. The light blue area in Figure~\ref{fig:pomis-example-simple-b} covers backdoor paths from a subset of $\*X$ to $Y$ in $\G_{\overline{X_1}}$. Thus, $\mu_\emptyset \leq \mu_{x_1}$ can be inferred. One may directly facilitate a graph obtained by projecting out variables neither $\*X$ nor $Y$, and still backdoor paths can be equally examined in the resulting graph (Figure~\ref{fig:pomis-example-simple-c}). For intervening on $X_2$ (Figure~\ref{fig:pomis-example-simple-d}), both $X_1$ and $X_3$ have no backdoor path to $Y$ (Figure~\ref{fig:pomis-example-simple-e}). Considering the minimality, we derive $\mu^*_{X_2} \leq \mu^*_{X_2,X_3,X_1} \leq  \mu^*_{X_3}$ (Figure~\ref{fig:pomis-example-simple-f}). We can avoid considering $X_1$ in the beginning by excluding variables ineffective to $Y$ under the intervention $\doo(x_2)$.
We define a set of variables having backdoor paths to $Y$ in a given graph:
\begin{definition}[Minimal Unobserved-Confounders' Territory]\label{lb:def:UCT}
    Given a diagram $\G$ and a node $Y$,
    let $\1H$ be $\G[An(Y)_{\G}]$. A set of variables
    $\*T\subseteq\*V(\1H)$ containing $Y$ is
    called a \textit{UC-territory} on $\G$ with respect to $Y$ if $\*T$ is closed under descendants and c-component, that is, $De(\*T)_{\1H}=\*T$ and there is no bidirected edge between $\*T$ and $\*V\setminus \*T$ in $\1H$. If there is no UC-territory $\*T'\subsetneq \*T$, then $\*T$ is a minimal UC-territory. \hfill $\blacksquare$
\end{definition}
The subgraph induced by Minimal Unobserved-Confounders' Territory (MUCT) represents how $Y$ is determined through the variables ruled by unobserved confounders under the intervention outside the MUCT. MUCT is tightly related to Rule 2 of do-calculus, and the procedure iteratively extends variables that have a backdoor path to $Y$ so that the rest of the variables can be exchangeable between condition and intervention.

This has a connection to the partial-orders we are seeking --- this demonstrates that $\mu_\emptyset \leq \mu^*_{\*V\setminus \*T}= \mu_{pa(\*T)\setminus \*T}$ if $\*X^\star$ is defined as $\*V\setminus\set{Y}$. We define $\operatorname{MUCT}(\G,Y)$ as the MUCT of $\G$ with respect to $Y$ and $\operatorname{IB}(\G,Y) = pa(\*T)_\G \setminus \*T$ as the Interventional Border (IB) of $\G$ with respect to $Y$ where $\*T=\operatorname{MUCT}(\G,Y)$.
For an arbitrary $\*X^\star \subseteq \*V\setminus \set{Y}$, one can obtain a MUCT and IB from, say $\1H$, the latent projection of $\G$ onto $\*X\cup\set{Y}$. Then, $\mu^*_{\*W} \leq \mu^*_{\operatorname{IB}({\1H}_{\overline{\*W}},Y)}$. 
A more involved example is shown in Figure~\ref{fig:pomis-abstract} where MUCT can be procedurally constructed by iteratively updating $\set{Y}$ by including the variables connected with it via bidirected edges and descendants of them. For example, MUCT under $\doo(B)$ in Figure~\ref{fig:pomis-abstract-c} can be obtained by first removing $A$ in the graph and expanding $\set{Y}$ with its confounded variables $C$, its descendants $E$, and, again, its confounded variable $F$ to ultimately arrive at $\set{Y,C,E,F}$.

\begin{figure}
    \begin{subfigure}{.24\textwidth}\centering
    \begin{tikzpicture}[x=9mm,y=9mm,rotate=20]
        \node[vertex] (Y) at (0:0) {$Y$};
        \node[vertex] (E) at (100:1) {$E$};
        \node[vertex] (D) at (100:2) {$D$};
        \node[vertex] (B) at (100:3) {$B$};
        \node[vertex] (A) [left=5mm of D] {$A$};
        \node[vertex] (C) at (60:1.6) {$C$};
        \node[vertex] (F) at (160:1.6) {$F$};
        
        \draw[->] (A) -- (B);
        \draw[->] (B) -- (D);
        \draw[->] (D) -- (E);
        \draw[->] (E) -- (Y);
        \draw[->] (F) -- (Y);
        \draw[->] (C) -- (E);
        \draw[<->,dashed] (A) to [bend right=40] (D);
        \draw[<->,dashed] (A) to [bend right] (F);
        \draw[<->,dashed] (F) to [bend left=10] (E);
        \draw[<->,dashed] (C) to [bend left=20] (Y);
        \draw[<->,dashed] (B) to [bend left] (C);

        \begin{pgfonlayer}{back}
            \begin{scope}[opacity=0]
                \draw[fill=betterblue!30,draw=none] \convexpath{C,Y}{4mm};
                \draw[fill=betterblue!30,draw=none] \convexpath{F,E,Y}{4mm};
            \end{scope}
        \end{pgfonlayer}
        \def\outerr{3.2}
        \def\innerr{3}
        \begin{pgfonlayer}{back}
            \node[circle,fill=betterred!65,draw=none,minimum size=2*\innerr mm] at (Y) {};
        \end{pgfonlayer}

    \end{tikzpicture}
    \caption{$\G$}
    \label{fig:pomis-abstract-a}
    \end{subfigure}\hfill
    \begin{subfigure}{.24\textwidth}\centering
    \begin{tikzpicture}[x=9mm,y=9mm,rotate=20]
        \node[vertex] (Y) at (0:0) {$Y$};
        \node[vertex] (E) at (100:1) {$E$};
        \node[vertex] (D) at (100:2) {$D$};
        \node[vertex] (B) at (100:3) {$b$};
        \node[vertex] (A) [left=5mm of D] {$A$};
        \node[vertex] (C) at (60:1.6) {$C$};
        \node[vertex] (F) at (160:1.6) {$F$};
        
        \draw[->] (B) -- (D);
        \draw[->] (D) -- (E);
        \draw[->] (E) -- (Y);
        \draw[->] (F) -- (Y);
        \draw[->] (C) -- (E);
        \draw[->] (E) -- (Y);
        \draw[<->,dashed] (A) to [bend right=40] (D);
        \draw[<->,dashed] (A) to [bend right] (F);
        \draw[<->,dashed] (F) to [bend left=10] (E);
        \draw[<->,dashed] (C) to [bend left=20] (Y);
        
        \begin{pgfonlayer}{back}
            \begin{scope}[opacity=0]
                \draw[fill=betterblue!30,draw=none] \convexpath{C,Y}{4mm};
                \draw[fill=betterblue!30,draw=none] \convexpath{F,E,Y}{4mm};
            \end{scope}
        \end{pgfonlayer}
        \def\outerr{3.2}
        \def\innerr{3}
        \begin{pgfonlayer}{back}
            \node[circle,fill=betterblue!65,draw=none,minimum size=2*\innerr mm] at (B) {};
            \node[circle,fill=betterred!65,draw=none,minimum size=2*\innerr mm] at (Y) {};
        \end{pgfonlayer}
    \end{tikzpicture}
    \caption{$\G_{\overline{B}}$}
    \label{fig:pomis-abstract-b}
    \end{subfigure}\hfill
    \begin{subfigure}{.24\textwidth}\centering
    \begin{tikzpicture}[x=9mm,y=9mm,rotate=20]
        \node[vertex] (Y) at (0:0) {$Y$};
        \node[vertex] (E) at (100:1) {$E$};
        \node[vertex] (D) at (100:2) {$D$};
        \node[vertex] (B) at (100:3) {$b$};
        \node[vertex,opacity=0] (A) [left=5mm of D] {$A$};
        \node[vertex] (C) at (60:1.6) {$C$};
        \node[vertex] (F) at (160:1.6) {$F$};
        
        \draw[->] (B) -- (D);
        \draw[->] (D) -- (E);
        \draw[->] (E) -- (Y);
        \draw[->] (F) -- (Y);
        \draw[->] (C) -- (E);
        \draw[<->,dashed] (F) to [bend left=10] (E);
        \draw[<->,dashed] (C) to [bend left=20] (Y);
        
        \begin{pgfonlayer}{back}
            \draw[fill=betterblue!30,draw=none] \convexpath{E,C,Y}{4mm};
            \draw[fill=betterblue!30,draw=none] \convexpath{F,E,Y}{4mm};
        \end{pgfonlayer}
        \def\outerr{3.2}
        \def\innerr{3}
        \begin{pgfonlayer}{back}
            \node[circle,fill=betterblue!65,draw=none,minimum size=2*\innerr mm] at (B) {};
            \node[circle,fill=betterred!65,draw=none,minimum size=2*\innerr mm] at (Y) {};
        \end{pgfonlayer}
    \end{tikzpicture}
    \caption{$\operatorname{MUCT}(\G_{\overline{B}},Y)$}
    \label{fig:pomis-abstract-c}
    \end{subfigure}\hfill
    \begin{subfigure}{.24\textwidth}\centering
    \begin{tikzpicture}[x=9mm,y=9mm,rotate=20]
        \node[vertex] (Y) at (0:0) {$Y$};
        \node[vertex] (E) at (100:1) {$E$};
        \node[vertex] (D) at (100:2) {$d$};
        \node[vertex,opacity=0] (B) at (100:3) {$b$};
        \node[vertex,opacity=0] (A) [left=5mm of D] {$A$};
        \node[vertex] (C) at (60:1.6) {$C$};
        \node[vertex] (F) at (160:1.6) {$F$};

        \draw[->] (D) -- (E);
        \draw[->] (E) -- (Y);
        \draw[->] (F) -- (Y);
        \draw[->] (C) -- (E);
        \draw[<->,dashed] (F) to [bend left=10] (E);
        \draw[<->,dashed] (C) to [bend left=20] (Y);
        
        \begin{pgfonlayer}{back}
            \draw[fill=betterblue!30,draw=none] \convexpath{E,C,Y}{4mm};
            \draw[fill=betterblue!30,draw=none] \convexpath{F,E,Y}{4mm};
        \end{pgfonlayer}
        \def\outerr{3.2}
        \def\innerr{3}
        \begin{pgfonlayer}{back}
            \node[circle,fill=bettergreen!65,draw=none,minimum size=2*\innerr mm] at (D) {};
            \node[circle,fill=betterred!65,draw=none,minimum size=2*\innerr mm] at (Y) {};
        \end{pgfonlayer}
    \end{tikzpicture}
    \caption{$\operatorname{MUCT}$ \& $\operatorname{IB}(\G_{\overline{B}},Y)$}
    \label{fig:pomis-abstract-d}
    \end{subfigure}
    \smallskip
    
    \begin{subfigure}{.24\textwidth}\centering
    \begin{tikzpicture}[x=9mm,y=9mm,rotate=20]
        \node[vertex] (Y) at (0:0) {$Y$};
        \node[vertex] (E) at (100:1) {$e$};
        \node[vertex] (D) at (100:2) {$D$};
        \node[vertex] (B) at (100:3) {$B$};
        \node[vertex] (A) [left=5mm of D] {$A$};
        \node[vertex] (C) at (60:1.6) {$C$};
        \node[vertex] (F) at (160:1.6) {$F$};
        
        \draw[->] (A) -- (B);
        \draw[->] (B) -- (D);
        \draw[->] (E) -- (Y);
        \draw[->] (F) -- (Y);
        \draw[->] (E) -- (Y);
        \draw[<->,dashed] (A) to [bend right=40] (D);
        \draw[<->,dashed] (A) to [bend right] (F);
        \draw[<->,dashed] (C) to [bend left=20] (Y);
        \draw[<->,dashed] (B) to [bend left] (C);

        \def\outerr{3.2}
        \def\innerr{3}
        \begin{pgfonlayer}{back}
            \node[circle,fill=betterblue!65,draw=none,minimum size=2*\innerr mm] at (E) {};
            \node[circle,fill=betterred!65,draw=none,minimum size=2*\innerr mm] at (Y) {};
        \end{pgfonlayer}
        \begin{pgfonlayer}{back}
            \begin{scope}[opacity=0]
                \draw[fill=betterblue!30,draw=none] \convexpath{C,Y}{4mm};
                \draw[fill=betterblue!30,draw=none] \convexpath{F,E,Y}{4mm};
            \end{scope}
        \end{pgfonlayer}
    \end{tikzpicture}
    \caption{$\G_{\overline{E}}$}
    \label{fig:pomis-abstract-e}
    \end{subfigure}\hfill
    \begin{subfigure}{.24\textwidth}\centering
    \begin{tikzpicture}[x=9mm,y=9mm,rotate=20]
        \node[vertex] (Y) at (0:0) {$Y$};
        \node[vertex] (E) at (100:1) {$e$};
        \node[vertex,opacity=0] (D) at (100:2) {$D$};
        \node[vertex,opacity=0] (B) at (100:3) {$B$};
        \node[vertex,opacity=0] (A) [left=5mm of D] {$A$};
        \node[vertex,opacity=0] (C) at (60:1.6) {$C$};
        \node[vertex] (F) at (160:1.6) {$f$};
        
        \draw[->] (E) -- (Y);
        \draw[->] (F) -- (Y);
        \draw[->] (E) -- (Y);

        \begin{pgfonlayer}{back}
            \node[circle,draw=none,fill=betterblue!30,minimum size=8mm] {};
        \end{pgfonlayer}
        \def\outerr{3.2}
        \def\innerr{3}
        \begin{pgfonlayer}{back}
            \node[circle,fill=bettergreen!65,draw=none,minimum size=2*\innerr mm] at (E) {};
            \node[circle,fill=bettergreen!65,draw=none,minimum size=2*\innerr mm] at (F) {};
            \node[circle,fill=betterred!65,draw=none,minimum size=2*\innerr mm] at (Y) {};
        \end{pgfonlayer}
    \end{tikzpicture}
    \caption{$\operatorname{MUCT}$ \& $\operatorname{IB}(\G_{\overline{E}},Y)$}
    \label{fig:pomis-abstract-f}
    \end{subfigure}\hfill
    \begin{subfigure}{.24\textwidth}\centering
    \begin{tikzpicture}[x=9mm,y=9mm,rotate=20]
        \node[vertex] (Y) at (0:0) {$Y$};
        \node[vertex] (E) at (100:1) {$E$};
        \node[vertex] (D) at (100:2) {$D$};
        \node[vertex] (B) at (100:3) {$B$};
        \node[vertex] (A) [left=5mm of D] {$A$};
        \node[vertex] (C) at (60:1.6) {$C$};
        \node[vertex] (F) at (160:1.6) {$f$};
        
        \draw[->] (A) -- (B);
        \draw[->] (B) -- (D);
        \draw[->] (D) -- (E);
        \draw[->] (E) -- (Y);
        \draw[->] (F) -- (Y);
        \draw[->] (C) -- (E);
        \draw[->] (E) -- (Y);
        \draw[<->,dashed] (A) to [bend right=40] (D);
        \draw[<->,dashed] (C) to [bend left=20] (Y);
        \draw[<->,dashed] (B) to [bend left] (C);
        \def\outerr{3.2}
        \def\innerr{3}
        \begin{pgfonlayer}{back}
            \node[circle,fill=betterblue!65,draw=none,minimum size=2*\innerr mm] at (F) {};
            \node[circle,fill=betterred!65,draw=none,minimum size=2*\innerr mm] at (Y) {};
        \end{pgfonlayer}
        \begin{pgfonlayer}{back}
            \begin{scope}[opacity=0]
                \draw[fill=betterblue!30,draw=none] \convexpath{C,Y}{4mm};
                \draw[fill=betterblue!30,draw=none] \convexpath{F,E,Y}{4mm};
            \end{scope}
        \end{pgfonlayer}
        
    \end{tikzpicture}
    \caption{$\G_{\overline{F}}$}
    \label{fig:pomis-abstract-g}
    \end{subfigure}\hfill
    \begin{subfigure}{.24\textwidth}\centering
    \begin{tikzpicture}[x=9mm,y=9mm,rotate=20]
        \node[vertex] (Y) at (0:0) {$Y$};
        \node[vertex] (E) at (100:1) {$E$};
        \node[vertex] (D) at (100:2) {$D$};
        \node[vertex] (B) at (100:3) {$B$};
        \node[vertex] (A) [left=5mm of D] {$A$};
        \node[vertex] (C) at (60:1.6) {$C$};
        \node[vertex] (F) at (160:1.6) {$f$};
        
        \draw[->] (A) -- (B);
        \draw[->] (B) -- (D);
        \draw[->] (D) -- (E);
        \draw[->] (E) -- (Y);
        \draw[->] (F) -- (Y);
        \draw[->] (C) -- (E);
        \draw[->] (E) -- (Y);
        \draw[<->,dashed] (A) to [bend right=40] (D);
        \draw[<->,dashed] (C) to [bend left=20] (Y);
        \draw[<->,dashed] (B) to [bend left] (C);

        \begin{pgfonlayer}{back}
            \draw[fill=betterblue!30,draw=none] \convexpath{B,Y}{4mm};
            \draw[fill=betterblue!30,draw=none] \convexpath{A,B}{4mm};
            \draw[fill=betterblue!30,draw=none] \convexpath{E,C,Y}{4mm};
        \end{pgfonlayer}
        \def\outerr{3.2}
        \def\innerr{3}
        \begin{pgfonlayer}{back}
            \node[circle,fill=bettergreen!65,draw=none,minimum size=2*\innerr mm] at (F) {};
            \node[circle,fill=betterred!65,draw=none,minimum size=2*\innerr mm] at (Y) {};
        \end{pgfonlayer}
    \end{tikzpicture}
    \caption{$\operatorname{MUCT}$ \& $\operatorname{IB}(\G_{\overline{F}},Y)$}
    \label{fig:pomis-abstract-h}
    \end{subfigure}
    \caption{Obtaining MUCTs (variables in light blue areas) and IBs (variable in green) under intervention (variables in blue) with $\*X^\star=\*V\setminus\set{Y}$ (or equivalently $\G$ maybe viewed as the latent projection of an original graph by retaining only $\*X^\star\cup \set{Y}$). Here, $\mu^*_{B} \leq \mu^*_{D}$ and $\mu^*_E\leq \mu^*_{E,F}$. }
    \label{fig:pomis-abstract}
\end{figure}
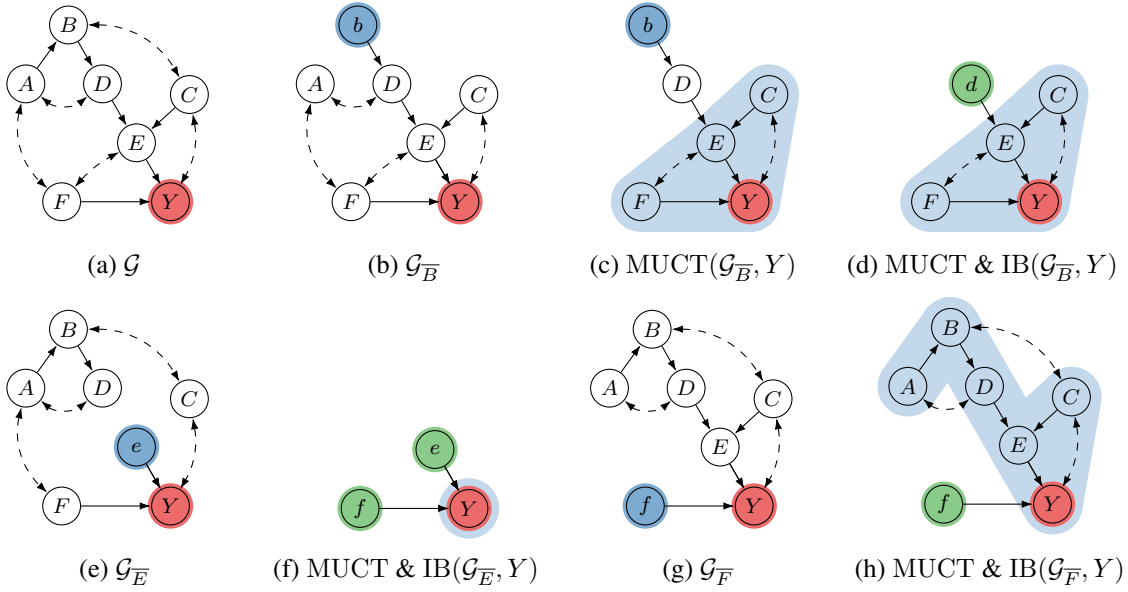

Equipped with MUCT and IB, one can check whether there exists a better MIS than a given MIS with respect to their maximum achievable expected rewards. However, the current procedure does not tell us whether there are other MISes better than the one obtained by an interventional border. The following theorem asserts that the interventional border approach provides a way to establishing a complete collection of POMISes given $\G$, $\*X^\star$, and $Y$:

\begin{theorem}\label{lb:lem:nece-suff}Given $\G$, $\*X^\star$, and $Y$, let $\*X'\subseteq \*X^\star$ be an MIS and let $\1H$ be the latent projection of $\G$ onto $\*X\cup\set{Y}$. Then, $\*X'$ is a POMIS if and only if $\operatorname{IB}(\1H_{\overline{\*X'}},Y)=\*X'$.  \hfill $\blacksquare$
\end{theorem}
This result \citep{LB:18a,LB:19a} can be best explained by that, under the distributions of the set of unobserved confounders in a MUCT, the mechanisms of the variables in the MUCT are orchestrated to yield the best result given a configuration (values set outside the MUCT). When any external force (interventions on any variables in the MUCT) is applied, the delicately orchestrated mechanism is disrupted and a subpar reward is obtained. In Figure~\ref{fig:pomis-example-simple}, arms intervening on $\set{X_1}$, $\set{X_3}$, and $\set{X_4}$ are POMISes.

\paragraph*{Property 3. Expressions among actions}
The two aforementioned structural properties help bandit agents to focus only on a set of minimal arms that can possibly be optimal without examining any collected data. We now consider the third property, which connects the causal effect of playing an arm and the distributions from data acquired through playing other arms.

Consider the causal diagram in Figure~\ref{fig:where-property-three} with $\*X^\star=\set{B,D,E}$ where POMISes are $\emptyset$, $\set{B}$, $\set{D}$, $\set{E}$, $\set{B, D}$, and $\set{D, E}$. Then, an online agent playing POMIS arms will obtain samples from $P(\*V)$, $P_b(\*V\setminus\set{B})$, $\dotsc$, and $P_{d,e}(\*V\setminus\set{D,E})$. In this example, one can rewrite the expected reward for, e.g., $\doo(\emptyset)$ according to c-factorization \citep{tian:pea02c}, as
\begin{align}
\E[Y]
    &= \sum_{a,b,c,d,e,y} y P_{d,e}(y,a,b) P_{a,c}(e) P_c(d) P_b(c) \\
    &= \sum_{a,b,c,d,e,y} y P_{d,e}(y\mid a,b)P_{d,e}(a,b) P_{a,c}(e) P_c(d) P_b(c) \label{eq:factrozied-prop3}
\end{align}
Other arms' expected rewards can be similarly factorized.
Here, each term can be replaced by other probabilities obtainable from different POMIS arms with the help of do-calculus (Table~\ref{tab:factrozied-prop3}), which is derived from subsequent applications of do-calculus.
\begin{table}
    \centering\footnotesize
    \begin{tabular}{@{}r|rrrrrr@{}}
                         & $\doo()$          & $\doo(b)$      & $\doo(d)$        & $\doo(e)$        & $\doo(b,d)$        & $\doo(d,e)$   \\ \midrule
        $P_{d,e}(y\mid a,b)$ & $P(y\mid  a,b,d,e)$ &              & $P_d(y\mid a,b,e)$ & $P_e(y\mid a,b,d)$ &                  & \\
        $P_{d,e}(a,b)$   & $P(a,b)$        &              & $P_d(a,b)$     & $P_e(a,b)$     &                  & \\
        $P_{a,c}(e)$     & $P(e\mid a,c)$      & $P_b(e\mid a,c)$ & $P_d(e\mid a,c)$   &                & $P_{b,d}(e\mid a,c)$ & \\
        $P_c(d)$         & $P(d\mid c)$        & $P_b(d\mid c)$   &                & $P_e(d\mid c)$     &                  & \\
        $P_b(c)$         & $P(c\mid b)$        &              & $P_d(c\mid b)$     & $P_e(c\mid b)$     & $P_{b,d}(c)$     & $P_{d,e}(c\mid b)$ \\ \bottomrule
    \end{tabular}
    \caption{For each term shown in Equation~\ref{eq:factrozied-prop3} (rows), its equal probability quantities that are obtainable from data sampled by playing POMIS arms (columns) are shown. Note, for example, that $P_e(d\mid c)=P_c(d)$ implies that this holds true for every $e\in \D(E)$.}
    \label{tab:factrozied-prop3}
\end{table}
These equalities not only imply that a term can be replaced by another but also suggest that each quantity can be estimated from the combination of them. 
For example, the term $P_c(d)$ can be estimated by a weighted combination of $P_c(d), P(d\mid c), P_e(d\mid c),P_b(d\mid c)$ as
\[
    \frac{N_c\hat{P}_c(d) + N_{\emptyset\mid c}\hat{P}(d\mid c) + \sum_e N_{e\mid c}\hat{P}_e(d\mid c) + \sum_b N_{b\mid c} \hat{P}_b(d\mid c)}{N_c+N_{\emptyset\mid c}+\sum_e N_{e\mid c}+\sum_b N_{b\mid c}},
\]
where $N_{\*w\mid \*z}$ is the number of samples with $\*Z=\*z$ in $\doo(\*W=\*w)$. This expression is nothing but estimating the probability of $D=d$ based on a maximum likelihood principle by aggregating compatible data instances together. Plugging in such estimator for each term in the expression for $\hat{\E}[Y]$ results in an estimator taking advantage of other arms' data.

Imagine an online learning scenario in which $\mu_\emptyset = \E[Y]$ is relatively smaller than the optimal arm. The agent in the scenario will likely play more on other arms with higher rewards. The agent would occasionally play less-played arms when the agent is not completely confident that such arms are not the best arm (e.g., UCB or Thompson sampling). With the expression provided and other arms' data utilized, the agent can improve confidence on what $\mu_\emptyset$ is by playing other arms only avoiding accumulating regrets.

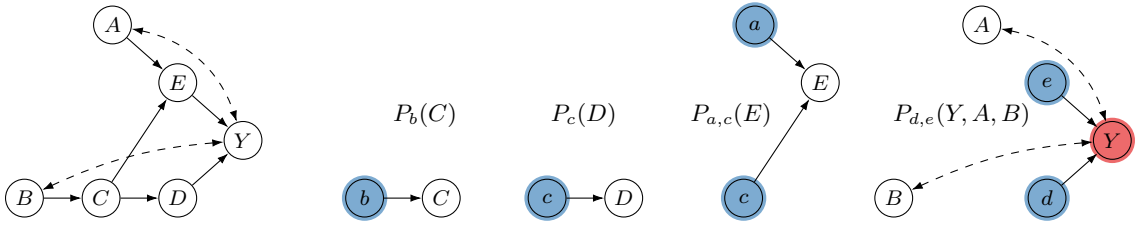
\begin{figure}
    \centering
    \begin{tikzpicture}[node distance=7mm and 5mm]
        \begin{scope}
            \node[vertex] (B) {$B$};
            \node[vertex] (C) [right=of B] {$C$};
            \node[vertex] (D) [right=of C] {$D$};
            \node[vertex] (Y) [above right=of D,yshift=-3mm] {$Y$};
            \node[vertex] (E) [above left=of Y,yshift=-3mm] {$E$};
            \node[vertex] (A) [above left=of E,yshift=-3mm] {$A$};
            
            \draw[->] (B) -- (C);
            \draw[->] (C) -- (D);
            \draw[->] (D) -- (Y);
            \draw[->] (A) -- (E);
            \draw[->] (E) -- (Y);
            \draw[->] (C) -- (E);
            
            \draw[<->,dashed] (B) to [bend left=10] (Y);
            \draw[<->,dashed] (A) to [bend left=30] (Y);
        \end{scope}
        \begin{scope}[xshift=4.5cm]
            \node[vertex] (B) {$b$};
            \node[vertex] (C) [right=of B] {$C$};
            \node[vertex,opacity=0] (D) [right=of C] {$D$};
            \node[vertex,opacity=0] (Y) [above right=of D,yshift=-3mm] {$Y$};
            \node[vertex,opacity=0] (E) [above left=of Y,yshift=-3mm] {$E$};
            \node[vertex,opacity=0] (A) [above left=of E,yshift=-3mm] {$A$};
            
            \draw[->] (B) -- (C);
            \def\outerr{3.2}
        \def\innerr{3}
        \begin{pgfonlayer}{back}
            \node[circle,fill=betterblue!65,draw=none,minimum size=2*\innerr mm] at (B) {};
        \end{pgfonlayer}
        \end{scope}
        \begin{scope}[xshift=5.9cm]
            \node[vertex,opacity=0] (B) {$B$};
            \node[vertex] (C) [right=of B] {$c$};
            \node[vertex] (D) [right=of C] {$D$};
            \node[vertex,opacity=0] (Y) [above right=of D,yshift=-3mm] {$Y$};
            \node[vertex,opacity=0] (E) [above left=of Y,yshift=-3mm] {$E$};
            \node[vertex,opacity=0] (A) [above left=of E,yshift=-3mm] {$A$};
            
            \draw[->] (C) -- (D);
            \def\outerr{3.2}
        \def\innerr{3}
        \begin{pgfonlayer}{back}
            \node[circle,fill=betterblue!65,draw=none,minimum size=2*\innerr mm] at (C) {};
        \end{pgfonlayer}
        \end{scope}
        \begin{scope}[xshift=8.5cm]
            \node[vertex,opacity=0] (B) {$B$};
            \node[vertex] (C) [right=of B] {$c$};
            \node[vertex,opacity=0] (D) [right=of C] {$D$};
            \node[vertex,opacity=0] (Y) [above right=of D,yshift=-3mm] {$Y$};
            \node[vertex] (E) [above left=of Y,yshift=-3mm] {$E$};
            \node[vertex] (A) [above left=of E,yshift=-3mm] {$a$};
            
            \draw[->] (A) -- (E);
            \draw[->] (C) -- (E);
            \def\outerr{3.2}
        \def\innerr{3}
        \begin{pgfonlayer}{back}
            \node[circle,fill=betterblue!65,draw=none,minimum size=2*\innerr mm] at (A) {};
            \node[circle,fill=betterblue!65,draw=none,minimum size=2*\innerr mm] at (C) {};
        \end{pgfonlayer}
        \end{scope}
        \begin{scope}[xshift=11.5cm]
            \node[vertex] (B) {$B$};
            \node[vertex,opacity=0] (C) [right=of B] {$C$};
            \node[vertex] (D) [right=of C] {$d$};
            \node[vertex] (Y) [above right=of D,yshift=-3mm] {$Y$};
            \node[vertex] (E) [above left=of Y,yshift=-3mm] {$e$};
            \node[vertex] (A) [above left=of E,yshift=-3mm] {$A$};
            
            \draw[->] (D) -- (Y);
            \draw[->] (E) -- (Y);

            \draw[<->,dashed] (B) to [bend left=10] (Y);
            \draw[<->,dashed] (A) to [bend left=30] (Y);
            \def\outerr{3.2}
        \def\innerr{3}
        \begin{pgfonlayer}{back}
            \node[circle,fill=betterblue!65,draw=none,minimum size=2*\innerr mm] at (D) {};
            \node[circle,fill=betterblue!65,draw=none,minimum size=2*\innerr mm] at (E) {};
            \node[circle,fill=betterred!65,draw=none,minimum size=2*\innerr mm] at (Y) {};
        \end{pgfonlayer}
        \end{scope}
        \node at (5.3,1.1) {\footnotesize $P_{b}(C)$};
        \node at (7.4,1.1) {\footnotesize $P_{c}(D)$};
        \node at (9.35,1.1) {\footnotesize $P_{a,c}(E)$};
        \node at (12.4,1.1) {\footnotesize $P_{d,e}(Y,A,B)$};
        
    \end{tikzpicture}
    \caption{A causal diagram $\G$ and the visualization of original c-factorization of $P(\*v)$.}
    \label{fig:where-property-three}
\end{figure}

In \citep{LB:19a}, the posterior of expected reward for each POMIS arm is approximated by bootstrapping samples from multiple data sources and integrated into Thompson sampling.
Similarly, based on the variance of expected reward from bootstraps, the effective number of arms played can be approximated and translated back to upper confidence bounds so as to be incorporated into a UCB algorithm.

\begin{figure}
\centering
\includegraphics[width=0.55\textwidth]{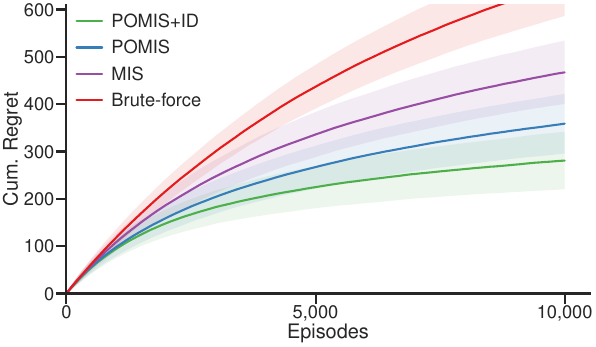} 
\caption{Cumulative regrets of different bandit agents based on Brute-force, MIS, POMIS, and POMIS with identification formula (POMIS+ID). Shaded areas represent standard deviation based on 2000 simulations.}
\label{fig:_6_1_6var}
\end{figure}

\begin{experiment}\label{exp:_6_1}
    Figure~\ref{fig:_6_1_6var} illustrates the cumulative regrets of \texttt{UCB} agents based on different arm candidates in an environment whose causal diagram is depicted in Figure~\ref{fig:where-property-three} (leftmost) and Table~\ref{tab:factrozied-prop3}. In this experiment, $A$ and $C$ are not intervenable, and the agent can intervene in the all possible combinations of $\set{B,D,E}$. An agent with brute-force strategy attempts to learn the expected reward for each combination while other agent with MIS or POMIS strategy respectively employs only intervening MISes or POMISes. POMIS+ID indicates an agent actively infers each arm's expected reward from other arms as well if possible.
    Simulation results illustrate clear gaps among the performances of different strategies. At the end (10,000th episodes), Brute-force, MIS, POMIS, and POMIS+ID respectively yields cumulative regret (mean and standard deviation) of 
    $655.96 \pm 69.68$,
    $467.19 \pm 66.85$,
    $358.86 \pm 63.11$, and
    $280.80 \pm 60.83$.
    These results demonstrate that the refinement of arms by considering causal structure improves the efficiency of agents in them interacting with the underlying domain. \hfill $\blacksquare$
\end{experiment}

In this section, we presented that an agent should be aware that choosing variables to intervene is not a trivial problem that can be simply answered like, i.e., intervene variables as many as it can, but a sophisticated problem that can be addressed with the knowledge of causal structure. In the next section, we further consider the case where agents can make use of states/contexts in their decision.

\subsection{Mixed Policy with Context}\label{sec:mixed-policy-with-context}

As seen in the previous section, a causal understanding of the underlying world enables us to recognize a wide range of policies across various policy spaces, allowing agents to choose their mode of interaction. This involves decisions about not only which variables to intervene in but also to observe as part of the context. In light of this, we explore the use of causal relationships in systematic decision-making over mixed policies with contexts involved. 
To illustrate the concept of mixed policy with context more clearly, let's consider an agent operating within an environment depicted as in Figure~\ref{fig:mixed-policy-intro-1}.

\begin{example}\label{ex:cpomis}
In this graph, we have 
intervenable variables $\*X^\star=\set{X_1,X_2}$ and contexts $\*S^\star = \set{C,X_1}$.
The primary objective of the agent is to maximize the reward, denoted as $\mu_{\pi}$, which is defined as the expected value of $Y$ when following a mixed policy $\pi$ chosen from a mixed policy space $\Pi_{\textsc{mix}}$.

\begin{figure}\centering
\def\dummytopA{1.25}
	\begin{subfigure}{.15\textwidth}
	\begin{tikzpicture}[x=1cm,y=.998cm]
		\node[opacity=0] (dummy) at (0,\dummytopA) {.};
		\node[vertex] (X1) at (-1.5,1) {$X_1$};
		\node[vertex] (X2) at (-1,0) {$X_2$};
		\node[vertex] (C) at (-.5,1) {$C$};
		\node[vertex] (Y) {$Y$};

		\draw[->] (X1) -- (X2);
		\draw[->] (X2) -- (Y);
		\draw[->] (C) -- (Y);
		\draw[->] (C) -- (X2);
		\draw[<->,dashed] (X1) to [bend left=30] (C);
		\draw[<->,dashed] (X2) to [bend right=30] (Y);
	\end{tikzpicture}
	\caption{$\G$}
	\label{fig:mixed-policy-intro-1}
	\end{subfigure}\hfill
	\begin{subfigure}{.17\textwidth}\centering
	\begin{tikzpicture}[x=1cm,y=.998cm]
		\node[opacity=0] (dummy) at (0,\dummytopA) {.};
		\node[vertex] (X1) at (-1.5,1) {$X_1$};
		\node[vertex] (X2) at (-1,0) {$X_2$};
		\node[vertex] (C) at (-.5,1) {$C$};
		\node[inner sep=0mm] (aux) at (-1.8,.0) {\scriptsize $\pi_{\*X}$};
		\node[vertex] (Y) {$Y$};
		
		\draw[->] (X2) -- (Y);
		\draw[->] (C) -- (Y);
		\draw[->] (C) -- (-1.15,.6);
		\draw[red,thin] \convexpath{X1,X2}{2.6mm};
		\draw[->] (aux) -- (-1.36,.45);
		\draw[<->,dashed,opacity=0] (X2) to [bend right=30] (Y);
        \def\outerr{3.2}
        \def\innerr{3}
        \begin{pgfonlayer}{back}
            \node[circle,fill=betterblue!65,draw=none,minimum size=2*\innerr mm] at (X1) {};
            \node[circle,fill=betterblue!65,draw=none,minimum size=2*\innerr mm] at (X2) {};
            \node[circle,fill=betterred!65,draw=none,minimum size=2*\innerr mm] at (Y) {};
        \end{pgfonlayer}
	\end{tikzpicture}
	\caption{CB (abstract)}
	\label{fig:new intro 2}
	\end{subfigure}\hfill
	\begin{subfigure}{.22\textwidth}\centering
	\begin{tikzpicture}[x=1cm,y=.998cm,node distance=3mm and 3mm]
		\node[opacity=0] (dummy) at (0,\dummytopA) {.};
		\node[vertex] (X1) at (-1.5,1) {$X_1$};
		\node[vertex] (X2) at (-1,0) {$X_2$};
		\node[vertex] (C) at (-.5,1) {$C$};
		\node[vertex] (Y) {$Y$};
		\node[inner sep=0mm,yshift=1mm] (aux1) [left=of X1] {\scriptsize$\pi_{X_1}$};
		\node[inner sep=0mm,yshift=1mm] (aux2) [left=of X2] {\scriptsize$\pi_{X_2}$};
		
		\draw[->] (C) -- (X1);
		\draw[->] (X1) -- (X2);
		\draw[->] (X2) -- (Y);
		\draw[->] (C) -- (Y);
		\draw[->] (C) -- (X2);
		\draw[<->,dashed,opacity=0] (X2) to [bend right=30] (Y);
		\draw[->] (aux1) -- (X1);
		\draw[->] (aux2) -- (X2);
         \def\outerr{3.2}
        \def\innerr{3}
        \begin{pgfonlayer}{back}
            \node[circle,fill=betterblue!65,draw=none,minimum size=2*\innerr mm] at (X1) {};
            \node[circle,fill=betterblue!65,draw=none,minimum size=2*\innerr mm] at (X2) {};
            \node[circle,fill=betterred!65,draw=none,minimum size=2*\innerr mm] at (Y) {};
        \end{pgfonlayer}
	\end{tikzpicture}
	\caption{$\G_{\Pi_\textsc{cb}}$}
	\label{fig:new intro 3}
	\end{subfigure}\ignorespaces
	\begin{subfigure}{.22\textwidth}\centering
	\begin{tikzpicture}[x=1cm,y=.998cm,node distance=3mm and 3mm]
		\node[opacity=0] (dummy) at (0,\dummytopA) {.};
		\node[vertex] (X1) at (-1.5,1) {$X_1$};
		\node[vertex] (X2) at (-1,0) {$X_2$};
		\node[vertex] (C) at (-.5,1) {$C$};
		\node[vertex] (Y) {$Y$};
		\node[inner sep=0mm,yshift=1mm] (aux1) [left=of X1] {\scriptsize$\pi_{X_1}$};
		
		\draw[->] (C) -- (X1);
		\draw[->] (X1) -- (X2);
		\draw[->] (X2) -- (Y);
		\draw[->] (C) -- (Y);
		\draw[->] (C) -- (X2);
		\draw[<->,dashed] (X2) to [bend right=30] (Y);
		\draw[->] (aux1) -- (X1);
        \def\outerr{3.2}
        \def\innerr{3}
        \begin{pgfonlayer}{back}
            \node[circle,fill=betterblue!65,draw=none,minimum size=2*\innerr mm] at (X1) {};
            \node[circle,fill=betterred!65,draw=none,minimum size=2*\innerr mm] at (Y) {};
        \end{pgfonlayer}
	\end{tikzpicture}
	\caption{$\G_{\Pi_d}$}
	\label{fig:new intro 4}
	\end{subfigure}\ignorespaces
	\begin{subfigure}{.2\textwidth}\centering
	\begin{tikzpicture}[x=1cm,y=.998cm,node distance=3mm and 3mm]
		\node[opacity=0] (dummy) at (0,\dummytopA) {.};
		\node[vertex] (X1) at (-1.5,1) {$X_1$};
        \node[vertex] (X2) at (-1,0) {$X_2$};
		\node[vertex] (C) at (-.5,1) {$C$};
		\node[vertex] (Y) {$Y$};
		\node[inner sep=0mm,yshift=1mm] (aux2) [left=of X2] {\scriptsize$\pi_{X_2}$};

		\draw[->] (X2) -- (Y);
		\draw[->] (C) -- (Y);
		\draw[->] (C) -- (X2);
        \draw[<->,dashed] (X1) to [bend left=30] (C);
		\draw[->] (aux2) -- (X2);

        \def\outerr{3.2}
        \def\innerr{3}
        \begin{pgfonlayer}{back}
            \node[circle,fill=betterblue!65,draw=none,minimum size=2*\innerr mm] at (X2) {};
            \node[circle,fill=betterred!65,draw=none,minimum size=2*\innerr mm] at (Y) {};
        \end{pgfonlayer}
	\end{tikzpicture}
	\caption{$\G_{\Pi_e}$}
	\label{fig:new intro 5}
	\end{subfigure}
	\caption{(a) a causal diagram, (b) abstract representation of a contextual bandit policy, and (c,d,e) policy-induced graphs. 
	$\pi$ nodes are intervention indicators, which will be left implicit throughout the section.
}
\end{figure}
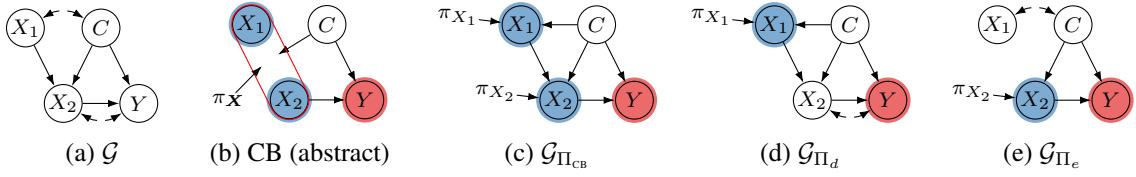 

In the realm of contextual bandit (CB) problems, the objective is to optimize a policy denoted as $\pi_{\textsc{cb}}$ (as depicted in Figure~\ref{fig:new intro 2}). This policy can be viewed as a stochastic mapping from contexts to actions. Alternatively, it can be represented as a pair of decision rules: $\pi_{\textsc{cb}}$ can be expressed as $\parens{\pi(X_1\mid C),\pi(X_2\mid X_1,C)}$ or more generally $\parens{\pi(X_1,X_2\mid C)}$ (illustrated in Figure~\ref{fig:new intro 3}).
Traditionally, this policy is optimized within a constrained space denoted as $\Pi_{\textsc{cb}}$, which consists of pairs, $\tuple{X_1,\set{C}}$ and $\tuple{X_2,\set{X_1,C}}$. However, an issue arises in that the optimal policy $\pi_{\textsc{cb}}^*$, determined as $\argmax_{\pi\in \Pi_{\textsc{cb}}} \mu_{\pi}$, may not necessarily be the best possible, i.e., $\mu^*_{\Pi_\textsc{cb}} \triangleq  \mu_{\pi_{\textsc{cb}}^*} < \mu^*$.

Consider a scenario where all variables are binary and $U_1$ and $U_2$ are unobserved confounders connected to $X_1$ and $X_2$,  respectively. Think of these as fair coin flips. Additionally, there's a noise $\epsilon$ associated with $X_1$, which follows a distribution with $P(\epsilon=1)=0.2$.
\begin{align}
    \2F = \begin{cases}
            X_1     &\gets U_1\oplus \epsilon, \\
            C       &\gets U_1, \\
            X_2     &\gets U_2\oplus X_1\oplus C, \\     
            Y       &\gets (1-(X_2\oplus U_2)) \vee C
       \end{cases}
       \label{eq:cpomis-example}
\end{align}
Given that the chosen policy renders $X_2$ independent of $U_2$ and that the context $C$ is also unrelated to $U_2$, it's deduced that the optimal value for $\mu^*_{\Pi_\textsc{cb}}$ equals 0.75.
In this setup, the most effective policy involves intervening solely on $X_1$ while considering the context $C$. This policy ensures that, when $X_1$ is set equal to $C$, the noise $\epsilon$ affecting $X_1$ is eliminated, resulting in $X_2$ becoming equivalent to $U_2$. Consequently, the policy attains an optimal expected reward of 1.0 in this environment. \hfill $\blacksquare$
\end{example}

\begin{figure}\centering
 \def\introgridh{4}
	\def\introgridw{7}
	\tikzset{srr/.style={draw=none,circle,inner sep=0mm, minimum size=1.8mm},font=\footnotesize,
	>={Latex[width=1.1mm,length=1.1mm]},
    	dagA/.pic={
    		\node[align=center, text width=21mm] (outer) {$\Pi_\textsc{cb}=$\\$\{\tuple{X_1,\set{C}},$\\$\tuple{X_2,\set{C,X_1}}\}$};},
    	dagB/.pic={
    		\node[align=center, text width=17mm] (outer)  {$\Pi_e=$\\$\set{\tuple{X_2,\set{C}}}$};},
    	dagC/.pic={
    		\node[align=center, text width=17mm] (outer) {$\Pi_d=$\\$\set{\tuple{X_1,\set{C}}}$};},
    	dagD/.pic={
    		\node[align=center, text width=9mm] (outer)  {$\Pi_a=$\\$\emptyset$};}
	}
	\begin{tikzpicture}[srr/.style={draw=none,circle,inner sep=0mm, minimum size=1.8mm,font=\footnotesize},>={Latex[width=1.1mm,length=1.1mm]},thin,x=.6cm,y=.533cm]
		\pic[shift={(0,\introgridh)}] (G1) {dagA};
		\pic[shift={(\introgridw,\introgridh)}] (G2) {dagB};
		\pic[shift={(0,0)}] (G3) {dagC};
		\pic[shift={(\introgridw,0)}] (G4) {dagD};
        \draw[-] (G1outer) -- node[pos=.5,above=0mm] {$\supset,=_\mu$} (G2outer);
		\draw[-] (G3outer) -- node[midway,above=0mm,sloped] {$\subset$} (G1outer);
		\draw[-] (G3outer) -- node[pos=.4,above=0mm,sloped] {$\supset,\geq_\mu$} (G4outer);
		\draw[-] (G1outer) -- node[pos=.5,above=0mm,sloped] {$\supset$} (G4outer);
		\draw[-] (G2outer) -- node[midway,above=0mm,sloped] {$\supset$} (G4outer);
    \end{tikzpicture}
	\caption{Relationships among the policy spaces based on two aspects.}
	\label{fig:intro motivational}
\end{figure}
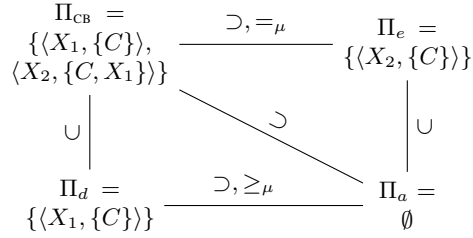

\paragraph{Desiderata for Optimal Mixed Policies}
From the mixed policy space associated with Figure~\ref{fig:mixed-policy-intro-1}, we can elicit 15 policy spaces from the mixed policy space.
These different modes of interaction can be categorized based on two desiderata: \textit{minimality} and \textit{optimality}. 
We explain these desiderata through an illustration (Figure~\ref{fig:intro motivational}) of the four policy spaces
$\Pi_{a}=\set{}$, $\Pi_\textsc{cb}$, $\Pi_d=\set{\tuple{X_1,\set{C}}}$, and $\Pi_e=\set{\tuple{X_2,\set{C}}}$ where  each subscript represents the label of figure.
We say $\Pi$ subsumes $\Pi'$, denoted by $\Pi'\subseteq\Pi$, if
$\*X(\Pi')\subseteq \*X(\Pi)$ and $\*S'_X \subseteq \*S_X$, for every $\tuple{X,\*S'_X}\in \Pi'$ where $\*X(\cdot)$ is a set of intervened variables in the policy space.
We use $\geq_{\mu}$ (or $=_{\mu}$) to indicate whether one's optimal reward is as good as or better than the other's in every scenario compatible with a causal diagram. This establishes equivalence classes among policy spaces based on their optimal rewards.

In simpler terms (to be formalized later on), \textit{minimality} means that removing any actions or contexts from a policy space can worsen its performance. In other words, given two policy spaces $\Pi$ and $\Pi'$, if $\Pi\supsetneq \Pi'$ and $\Pi=_{\mu}\Pi'$, then $\Pi$ is said to be \textit{redundant}.
For instance, since $\Pi_\textsc{cb} \supset \Pi_e$ while $\Pi_\textsc{cb} =_{\mu} \Pi_e$, the CB policy (Figure~\ref{fig:new intro 3}) is redundant and the CB agent wastes its resources not only for intervening on $X_1$ (a redundant action) but also for taking $X_1$ into account for $X_2$ (a redundant context).

Furthermore, \textit{optimality} of a policy space $\Pi$ represents that there exists no other policy space $\Pi'$ (not in the equivalence class of $\Pi$) such that $\Pi'\geq_\mu \Pi$.
For example, $\Pi_d$, when optimized, is at least as good as $\Pi_a$ (i.e., $\mu^*_{\Pi_d} \geq \mu^*_{\Pi_a}$) in every environment, and can outperform it in some environments (i.e., $\mu^*_{\Pi_d} > \mu^*_{\Pi_a}$), which demonstrates that $\Pi_a$ does not meet the optimality criterion. 
Not all policy spaces can be directly comparable: $\Pi_e$ is not comparable to $\Pi_a$ nor $\Pi_d$.  After careful examination, we find that policy spaces $\Pi_\textsc{cb}$, $\Pi_d$, $\Pi_e$ meet the optimality criterion.
Both minimality and optimality are satisfied only by $\Pi_d$ and $\Pi_e$ among all 15 policy spaces. This example illustrates that a smart agent should selectively intervene in variables with relevant contexts to achieve optimal rewards. Against this background, we will delve into the evaluation of mixed policies in terms of their expected rewards.

\subsubsection{Contextual Minimality in Optimal Mixed Policies}

Optimizing a mixed policy involves assessments of the effectiveness of its policy space so that 
an agent can avoid intervening or observing unnecessary actions or contexts.
It is well-known that an action on $X$ is worthy if it can affect $Y$ through the change of its mechanism $\pi_X$ and each context in $S\in \*S_X$ is relevant to its associated action $X$ if the context provides information relative to other contexts $\*S_X\setminus \set{S}$ \citep{lauritzen2001representing,zhang2020designing}. This simple characterization of (non-)minimality of an individual action and an individual context of policy space is, unfortunately, insufficient to fully grasp, e.g., whether a subset of contexts over multiple actions would be still relevant, especially when $\pi \in \Pi$ is optimized. We explain this insufficiency through an example.

\begin{example}In Figure~\ref{fig:nr-opt marginal}, 
both $X_1$ and $X_2$ utilize $C_3$ as their contexts where  $\mu_{\pi}=\E_{C_3}[\E_{\pi}[Y\mid C_3]]$. 
Since there exists $c^*_3 = \argmax_{c_3\in \D({C_3})} \E_{\pi}[Y\mid c_3]$, we can derive that $\mu_{\pi}\leq \E_{C_3}[\E_{\pi}[Y\mid c^*_3]] = \E_{\pi}[Y\mid c^*_3]$.
Given that $c^*_3$ is merely a constant, 
new decision rules \[\pi'(x_i\mid c_1)\triangleq P_{\pi}(x_i\mid c_i,c^*_3)=\pi(x_i\mid c_i,c^*_3)\] for $i\in\set{1,2}$ yield the same optimal reward. That is, $X_1$ and $X_2$ (as if they are two agents) agree to assume that $C_3$ is fixed to some value. \hfill $\blacksquare$
\end{example}
This example first demonstrates the idea of \textit{fixing} where it is viable to treat a variable in the context as a fixed value without sacrificing optimal performance. Once fixed, decision rules in the mixed policy can free the variable from being a context, resulting in a simpler policy space.
A more sophisticated example is shown in Figure~\ref{fig:nr-opt conditional} where a redundant context can be \textit{fixed} \textit{conditionally} on the remaining contexts.

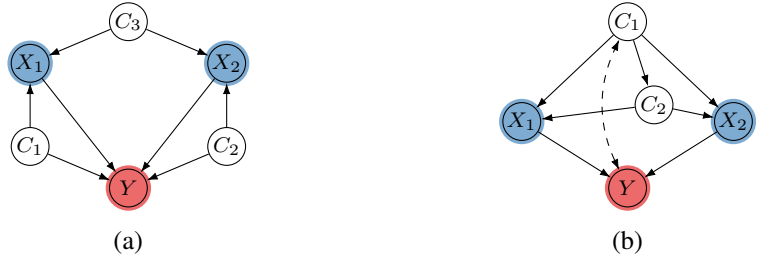
\begin{figure}
    \hfill
	\begin{subfigure}{.3\textwidth}\centering
		\begin{tikzpicture}[x=13mm, y=11mm]
            \node[vertex] (Y) {$Y$};
			\node[vertex] (C1) at (-1,0.5) {$C_1$};
			\node[vertex] (C2) at (1,0.5) {$C_2$};
			\node[vertex] (X1) at (-1,1.5) {$X_1$};
			\node[vertex] (X2) at (1,1.5) {$X_2$};
			\node[vertex] (C3) at (0,2) {$C_3$};
			\draw[->] (C3) -- (X1);
			\draw[->] (C3) -- (X2);
			\draw[->] (X1) -- (Y);
			\draw[->] (X2) -- (Y);
			\draw[->] (C1) -- (X1);
			\draw[->] (C1) -- (Y);
			\draw[->] (C2) -- (X2);
			\draw[->] (C2) -- (Y);

            \def\outerr{3.2}
            \def\innerr{3}
            \begin{pgfonlayer}{back}
                \node[circle,fill=betterblue!65,draw=none,minimum size=2*\innerr mm] at (X1) {};
                \node[circle,fill=betterblue!65,draw=none,minimum size=2*\innerr mm] at (X2) {};
                \node[circle,fill=betterred!65,draw=none,minimum size=2*\innerr mm] at (Y) {};
            \end{pgfonlayer}
		\end{tikzpicture}
		\caption{}
		\label{fig:nr-opt marginal}
	\end{subfigure}\hfill
	\begin{subfigure}{.3\textwidth}\centering
		\begin{tikzpicture}[x=14mm, y=11mm]
			\node[vertex]      (Y)  at (0,0) {$Y$};
			\node[vertex]  (X1) at (-1,1-0.2) {$X_1$};
			\node[vertex] (C2) at (0.25,1) {$C_2$};
			\node[vertex]  (X2) at (1,1-0.2) {$X_2$};
			\node[vertex] (C1) at (0,2) {$C_1$};
			\draw[<->,dashed] (Y) to [bend left=25] (C1);
			\draw[->] (C1) -- (X1);
			\draw[->] (C1) -- (X2);
			\draw[->] (C1) -- (C2);
			\draw[->] (C2) -- (X1);
			\draw[->] (C2) -- (X2);
			\draw[->] (X1) -- (Y);
			\draw[->] (X2) -- (Y);

            \def\outerr{3.2}
            \def\innerr{3}
            \begin{pgfonlayer}{back}
                \node[circle,fill=betterblue!65,draw=none,minimum size=2*\innerr mm] at (X1) {};
                \node[circle,fill=betterblue!65,draw=none,minimum size=2*\innerr mm] at (X2) {};
                \node[circle,fill=betterred!65,draw=none,minimum size=2*\innerr mm] at (Y) {};
            \end{pgfonlayer}
		\end{tikzpicture}
		\caption{}
		\label{fig:nr-opt conditional}
	\end{subfigure}\hfill\null
\caption{Causal diagrams where the relevance of some contexts can be further eliminated under the optimality of policy.}
\label{fig:movedmovedmoved}
\end{figure}

\begin{example}
In the policy illustrated in Figure~\ref{fig:nr-opt conditional}, both intervened variables are relying on the same contexts $C_1$ and $C_2$. The expected reward is expressed as
\begin{align}
    \mu_{\pi} 
        & = \sum_{y,\*x,\*s} y P_{\*x}(y,c_1,c_2) \pi(x_1\mid c_1,c_2)\pi(x_2\mid c_1,c_2). \notag
\intertext{Given that $P_{\*x}(y,c_1,c_2)=P_{\*x}(c_2\mid y,c_1)P_{\*x}(y,c_1)=P(c_2\mid c_1)P_{\*x}(y,c_1)$ based on basic do-calculus,}
        &= \sum_{c_1,c_2} P(c_2\mid c_1)  \underbrace{\sum_{y,\*x} yP_{\*x}(y,c_1)\pi(x_1\mid c_1,c_2)\pi(x_2\mid c_1,c_2)}_{\textrm{define } \mu_{\pi}(c_1,c_2)} \label{eq:mixed-characteristic}\\
\intertext{This expression can be viewed as the sum of weighted rewards for each $C_1$ value. Let $c^*_2$ be a function taking $c_1$ such that $c^*_2(c_1)=\argmax_{c_2}\mu_{\pi}(c_1,c_2)$ for $c_1\in \D(C_1)$. Then,}
        & \leq \sum_{c_1,c_2} P(c_2\mid c_1)   \mu_{\pi}(c_1,c^*_2(c_1)) \notag\\
        & = \sum_{c_1} \mu_{\pi}(c_1,c^*_2(c_1)) \notag\\
        &= \sum_{y,c_1,\*x} yP_{\*x}(y,c_1)\pi(x_1\mid c_1,c^*_2(c_1))\pi(x_2\mid c_1,c^*_2(c_1)). \notag
\intertext{By incorporating $c^*_2$ into $\pi$, we can devise a smaller mixed policy $\pi'$ such that 
 $\pi'(x_1\mid c_1)=\pi(x_1\mid c_1,c^*_2(c_1))$ and $\pi'(x_2\mid c_1)=\pi(x_2\mid c_1,c^*_2(c_1))$.}
        &= \sum_{y,c_1,\*x} yP_{\*x}(y,c_1)\pi'(x_1\mid c_1)\pi'(x_2\mid c_1) = \mu_{\pi'}. \notag
\end{align}  \hfill $\blacksquare$
\end{example}
The key idea demonstrated in this derivation is finding contexts that can be fixed to some desirable values relative to other contexts (or none) so that decision rules can be equivalently performed without relying on the contexts fixable via remaining contexts. Relationships between the two types of contexts best captured in Equation~\ref{eq:mixed-characteristic}.
More generally, the values to be fixed does not have to be contexts to yield an expression that is `greater than equal to' the previous expression. The restriction is (to meet our purpose to rewrite the expression to the expected reward of a smaller policy space) that any fixed context should be inferred from the rest of context (e.g., $c_2$ relative to $c_1$). Further, this process may involve the use of intervened variables that are fixed (determined) by their contexts under optimality (to show later).

Against this background, we will define and characterize the minimality of policy space under optimality, which has practical implications to an agent learning an optimal policy.
\begin{definition}[Minimality under Optimality]\label{def:NR-MPS-OPT}
	Given $\tuple{\G,Y, \*X^\star,\*S^\star}$, 
	a policy space $\Pi$ is said to be \emph{minimal under optimality}
	if there exists an SCM $\M$ compatible with $\G$ such that $\mu^*_{\Pi}>\mu^*_{\Pi'}$ for every strictly subsumed policy space $\Pi'\subsetneq\Pi$, that is, 
    \[\exists {\M\sim \G} \,\forall {\Pi'\subsetneq \Pi}\, (\mu^*_{\Pi}> \mu^*_{\Pi'}).\]  \hfill $\blacksquare$
\end{definition}
%
We will develop a sufficient condition for non-minimality under optimality by generalizing the idea presented earlier. The condition is made of two parts. The first part is obtaining a specific form of an intermediate expression for expected reward given a set of variables to fix. In the next part, based on the intermediate expression, it checks whether the variables can indeed be fixed to yield a new simpler policy. Before we proceed to investigate conditions for finding a simpler policy, let us briefly discuss what we mean by \textit{fixing}. Consider the following form of expression,
\[
    \sum_{a,\*b,\*d} P(a\mid \*b, \*c) f(a,\*b,\*d) \leq  \sum_{\*b,\*d} f(a^*(\*b),\*b,\*d) = \sum_{\*b,\*d} f'(\*b,\*d).
\]
With $\*C$ fixed to a constant, we say $a$ is \textit{fixed conditional} on $\*B$. For the purpose of eliciting a simpler policy, decision rules (implicit in $f$) also drop $a$ from its argument by inferring it from $\*b$ and $\*c$.

\paragraph{Step 1: obtaining an intermediate expression from an expected reward}
Given a policy space $\Pi$ satisfying the basic minimality, 
let $\*X'\subseteq \*X(\Pi)$ be actions of interest (of which we would like to change its decision rules), 
$\*S'\subsetneq \*S_{\*X'}\setminus \*X'$ non-action contexts of interest (that is, contexts to keep among the contexts of $\*X'$). 
Given (i) a subset of exogenous variables $\*U'$ in $\G_\Pi$\footnote{Formally speaking, we are selecting unobserved variables associated with a clique formed by bidirected edges in $\G_\Pi$.}, (ii) a subset of endogenous variables $\*Z$ in $\G_\Pi$ that disjoints with $\*S'\cup\*X'$ and subsumes unselected contexts $\*S_{\*X'}\setminus(\*S'\cup\*X')$, and (iii) an order $\prec$ over $\*V'\triangleq \*S'\cup\*X'\cup\*Z$, if the triple $\tuple{\*U'$, $\*Z$, $\prec}$ satisfies certain conditions \citep[Lemma~1]{lee2020characterizing}, then we can write $\mu_\pi$ as 
\begin{align}
    \mu_{\pi}=
    \overbrace{\sum_{\*u'}P_{\pi}(\*u')}^{\textrm{marginally fixable}}
    {\sum_{y,\*s',\*x'}} y\underbrace{{Q'}_{\*x'}(y,\*s')\vphantom{\prod_{\*Z}}}_{\textrm{irrelevant to }\*Z}
    \overbrace{{\sum_{\*z}}\underbrace{{\prod_{{Z\in\*Z}}} P_{\pi}(z\mid \*v'_{\prec Z},\*u')}_{\textrm{defines dependency}}}^{\textrm{to fix conditionally}}
    {\prod_{{X\in\*X'}}} \underbrace{\pi(x\mid \overbrace{\*s_X\setminus\*z\vphantom{\prod_{\*Z}}}^{\textrm{given}},\overbrace{\*s_X\vphantom\setminus \cap\*z\vphantom{\prod_{\*Z}}}^{\textrm{to infer}})}_{\textrm{to become $\pi'(x\mid \*s_X\setminus \*z)$}},
\end{align}
where $Q' = P_{\pi\setminus \*X'}$ is a distribution under $\pi$ except $\*X'$. The conditions are mainly designed to elicit the term $Q'_{\*x'}(y,\*s')$ without $\*Z$ so that fixing does not affect the term, and later we can transform the expression into the expected reward for a simpler policy relying only on context $\*S'$.

We explain the intermediate expression. To begin with we replace $P_{\pi}(Z\mid \*V'_{\prec Z},\*U')$ to $P_{\pi}(Z\mid \*V_Z)$ where  $\*V_Z$ is a minimal subset of $\*V'_{\prec Z}\cup \*U'$ dependent to $Z$.
The purpose of intermediate expression is to be transformed to $\mu_{\pi'}$ such that all decision rules $\parens{\pi(X\mid \*S_X)}$ for $X\in \*X'$ become $\parens{\pi(X\mid \*S_X\setminus \*Z)}$. We achieve this by fixing all the contexts containing $\*Z$ as dictated in $P_{\pi}(z \mid \*v_Z)$ similar to $P(c_2 \mid c_1)$ in the earlier example. 
What we have found is that fixing variables other than unnecessary contexts can ultimately help to fix and remove those unnecessary contexts. This explains why some of the exogenous variables and variables other than unnecessary contexts are involved. Finally, the order explicitly decides how probability terms are factorized following a chain rule and, thus, how variables are fixed relative to other variables.

\paragraph{Step 2: transforming intermediate expression into the expected reward for a simpler policy}
Once we attain the intermediate expression for some $\tuple{\*U',\*Z, \prec}$, 
we examine whether the expression can be converted to $\mu_{\pi'}$ where $\Pi'\triangleq (\Pi\setminus\*X') \cup \set{\tuple{X,\*S_X\setminus\*Z}}_{X\in\*X'}$.
Algorithmically speaking, we can first fix $\*u'$ unconditionally. Then, check whether any currently unfixed $Z\in\*Z$ can be fixed since $\*V'_{\prec Z}$ are all fixed, which are made of contexts, actions, and other endogenous variables. Other than fixing value of $Z$ to the best possible value to drop the term $P(Z\mid \cdot)$, action variables can be fixed in a similar way.

\begin{figure}
\hfill\begin{subfigure}{.3\textwidth}\centering
	\begin{tikzpicture}[node distance=5mm and 8mm]
		\node[vertex] (Y) {$Y$};
		\node[vertex] (X1) [above=of Y] {$X_1$};
		\node[vertex,yshift=1mm] (C2) [above=of X1] {$C_2$};
		\node[vertex,yshift=-2mm] (C1) [left=of C2] {$C_1$};
		\node[vertex] (X2) [below right=of C2] {$X_2$};
		\node[vertex,yshift=-5mm] (C3) [above right=of C2] {$C_3$};
        \draw[->] (C1) -- (X1);
		\draw[->] (C2) -- (X1);
		\draw[->] (C3) -- (X2);
		\draw[->] (C1) to [bend right=20] (Y);
		\draw[->] (X1) -- (Y);
		\draw[->] (X2) to [bend left=20] (Y);
		\draw[->] (X2) -- (C2);
		\draw[->] (C3) -- (C2);
		\draw[<->,dashed] (C1) to [bend left] (C2);
        \def\outerr{3.2}
        \def\innerr{3}
        \begin{pgfonlayer}{back}
            \node[circle,fill=betterblue!65,draw=none,minimum size=2*\innerr mm] at (X1) {};
            \node[circle,fill=betterblue!65,draw=none,minimum size=2*\innerr mm] at (X2) {};
            \node[circle,fill=betterred!65,draw=none,minimum size=2*\innerr mm] at (Y) {};
        \end{pgfonlayer}
        
	\end{tikzpicture}
 \caption{}
 \end{subfigure}\hfill
 \begin{subfigure}{.3\textwidth}\centering
 \begin{tikzpicture}[x=1cm, y=1cm]
        \node[vertex] (C3) at (-1.25, 0.5) {$C_3$};		
        \node[vertex] (X2) at (0,0.25) {$X_2$};
        \node[vertex] (C1) at (1,0) {$C_1$};
		\node[vertex] (C2) at (0,-.95) {$C_2$};		
  
        \draw[->] (C1) -- (C2);
        \draw[->] (C3) -- (X2);
        \draw[->] (C3) -- (C2);
        \draw[->] (X2) -- (C2);
        \def\outerr{3.2}
        \def\innerr{3}
        \begin{pgfonlayer}{back}
            \node[circle,fill=bettergreen!65,draw=none,minimum size=2*\innerr mm] at (C1) {};
        \end{pgfonlayer}
 \end{tikzpicture}
 \caption{}
 \label{fig:nro-3-dependency}
 \end{subfigure}\hfill\null
	\caption{(a) A minimal policy space and (b) its dependency graph derived in Example~\ref{ex:nro-3} where $C_1$ is given.}
	\label{fig:nr-example}
\end{figure}
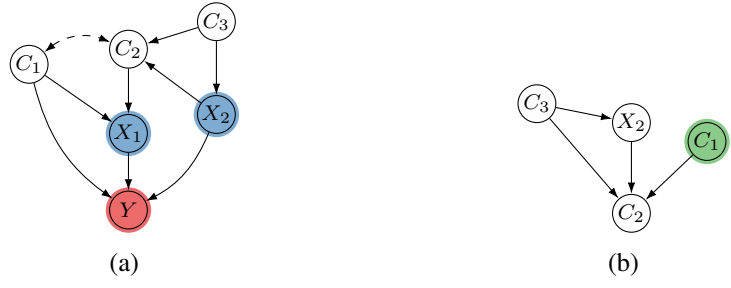

\begin{example}\label{ex:nro-3} 
    Through Figure~\ref{fig:nr-example}, we will show that, $C_2$ and $C_3$ are \emph{redundant contexts under optimality}. Given $\*S'=\set{C_1}$ and $\*X'=\set{X_1,X_2}$, consider $\*Z=\set{C_2,C_3}$, $\*U'=\emptyset$, 
    and order $\prec=\tuple{C_3,C_1,X_2,C_2,X_1}$. We can derive the following expression for the expected reward (with subscripts concatenated), 
    \begingroup
    \allowdisplaybreaks
    \begin{align}
        \mu^*_{\Pi} & = \sum_{y,\*x,c_1}yQ'_{\*x}(y\mid c_1)\sum_{c_{23}}P_{\pi}(c_{123},\*x)  \notag \\
                     & = \sum_{y,\*x,c_1}yQ'_{\*x}(y\mid c_1)\sum_{c_{23}}P_{\pi}(c_3)P_{\pi}(c_1\mid c_3)P_{\pi}(x_2\mid c_{13})P_{\pi}(c_2\mid c_{13},x_2)P_{\pi}(x_1\mid c_{123},x_2)  \notag  \\
                     & = \sum_{c_{3}}P_{\pi}(c_3)\sum_{y,\*x,c_1}yQ'_{\*x}(y,c_1)\sum_{c_{2}}P_{\pi}(c_2\mid c_{13},x_2)\pi(x_2\mid c_3)\pi(x_1\mid c_{12}). \label{eq:nro-main-text-example-first-part}
        \intertext{$C_3$ can be fixed to a constant $c^*_3$ so that,}
            & \leq \sum_{y,\*x,c_1}yQ'_{\*x}(y,c_1)\sum_{c_{2}}P_{\pi}(c_2\mid c_1,c^*_3,x_2)\pi(x_2\mid c^*_3)\pi(x_1\mid c_1,c_2). \hphantom{testtesttestte.} \notag 
        \intertext{This expression can be rearranged so that $\sum_{x_2} \pi(x_2\mid c_3^*)$ starts the expression where there exists $x^*_2\in\D(X_2)$, which allows us to substitute $\pi(x_2\mid c_3^*)$ with $\pi'(x_2)$ where $\pi'(x^*_2)=1$.}
            & \leq \sum_{y,\*x,c_1}yQ'_{\*x}(y,c_1)\sum_{c_{2}}P_{\pi}(c_2\mid c_1,c^*_3,x^*_2)\pi'(x_2)\pi(x_1\mid c_1,c_2). \notag
        \intertext{
        Here, although we can drop both $x_2$ from summation and $\pi'(x_2)$ from the expression, we keep them to connect to the expected reward of the resulting simpler policy.  
        Next, the optimal $c_2$ is determined with respect to $c_1$, i.e., $P_{\pi}(c_2\mid c_1,c^*_3,x^*_2)$, where we can replace $\pi(x_1\mid c_1,c^*_2(c_1))$ by $\pi'(x_1\mid c_1)$. We start by reordering terms for readability.}
        & = \sum_{c_1, c_{2}}P_{\pi}(c_2\mid c_1,c^*_3,x^*_2) \sum_{y,\*x}yQ'_{\*x}(y,c_1)\pi'(x_2)\pi(x_1\mid c_1,c_2) \notag \\
        & \leq \sum_{y,\*x,c_1}yQ'_{\*x}(y,c_1)\pi'(x_2)\pi(x_1\mid c_1,c^*_2(c_1)) \notag  \\
        &= \sum_{y,\*x,c_1}yQ'_{\*x}(y,c_1)\pi'(x_1\mid c_1)\pi'(x_2) = \mu^*_{\Pi'}. \label{eq:nropt-deriv-x1}
    \end{align}
    \endgroup
    Since $\mu^*_{\Pi'}\leq\mu^*_{\Pi}$ by the existence of $\pi\in \Pi$ that can emulate $\pi'\in\Pi'$, and $\mu^*_{\Pi'}\geq\mu^*_{\Pi}$ by the derivation (Equation~\ref{eq:nropt-deriv-x1}), we can conclude that $\mu^*_{\Pi'}=\mu^*_{\Pi}$. As a consequence, policy space $\Pi$ is  \textit{not minimal under optimality} due to the ineffective contexts $\set{C_2,C_3}$ with respect to $\set{X_1,X_2}$.  \hfill $\blacksquare$
\end{example}

The procedure leading to a simpler policy from the intermediate expression can be described as constructing and examining a dependency graph as follows.
Based on the distributions over $\*Z$ and $\*U'$, we can construct a dependency graph in the form of DAG.
Initially, vertices are $\*U'$, $\*Z$, $\*Z$'s parents, and the parents of intervened variables. Directed edges are added so that the parents of each node is the set of conditions in its distribution, that is, $P_\pi(Z \mid \*V_Z)$ yields $Z$ having $\*V_Z$ as its parents in the graph. This similarly applies to the decision rules $\pi(x \mid \*s_x)$. The dependency graph for the Example~\ref{ex:nro-3} is shown in Figure~\ref{fig:nro-3-dependency} with $\*Z=\set{C_2, C_3}$ and their parents determined through factorization in Equation~\ref{eq:nro-main-text-example-first-part}.

Once a dependency graph is constructed, we can then figure out whether all $\*Z$'s can be fixed to yield a simpler policy. Contexts to keep $\*S'$ do not have parents in the dependency graph and are marked to indicate they are available to fix other variables. Here, $C_1$ is available to fix, e.g., $C_2$, for its values. Nodes are marked if its parents are all marked. For example, $C_3$ can be marked unconditionally. $X_2$ is marked given that $C_3$ is marked. With $C_3$ and $X_2$ marked, altogether with $C_1$, $C_2$ is marked, resulting in fixing to $c_2^*(c^*_3,x^*_2(c^*_3), c_1)=c_2^*(c_1)$.
Finally, if (i) all $\*Z$'s in the dependency graph are marked, and (ii) the ancestors of each intervened variable $X$ do not include any of context variables that will not be available in the simpler policy, that is, $\*S'\setminus (\*S_X \setminus \*Z)$, then we can yield a simpler policy. More detailed results are presented in Theorem~2 \citep{lee2020characterizing}.

\begin{figure}
    \centering\includegraphics[width=.45\textwidth]{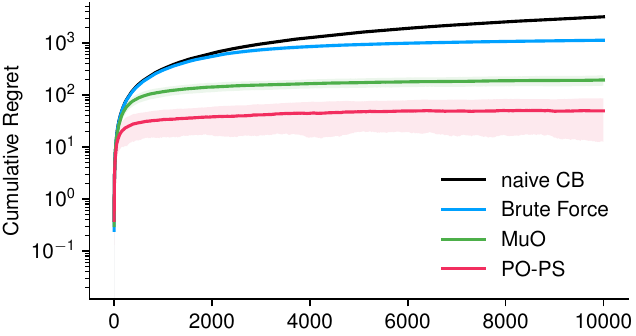}\hfill
    \includegraphics[width=.45\textwidth]{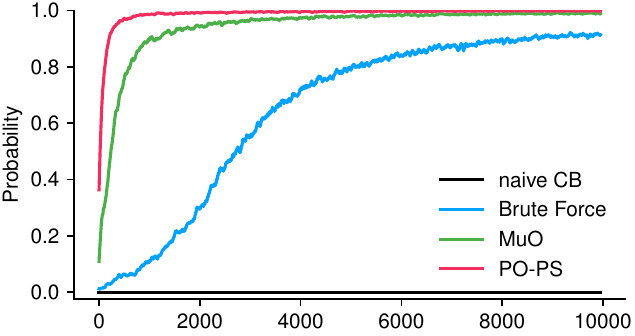}
    \caption{(left) Cumulative regrets (in a log scale) of different arm strategies based on all possible mixed policy spaces (brute-force, BF), naive contextual bandit (CB), Minimality under Optimality (MuO), and possibly optimal policy spaces (PO-PS). Shaded areas represent standard deviation based on 100 simulations. (right) Probability each agent selecting the optimal arm.}
    \label{fig:exp:_6_2}
\end{figure}

\begin{experiment}\label{exp:_6_2}
	Figure~\ref{fig:exp:_6_2} depicts the cumulative regrets of \texttt{UCB} agents based on different arm selection strategies in the environment described in Example~\ref{ex:cpomis}. There are four strategies where each strategy considers a subset of policy spaces in the mixed policy space. Brute-force makes use of all policy spaces. A naive contextual bandit (CB) implements intervening both given $C$ (Figure~\ref{fig:new intro 3}), which can be modeled as intervening on $X_1$ given $\set{C}$ and $X_2$ given $\set{C,X_1}$. The minimality under optimality (MuO) agent plays only the arms without redundant interventions or contexts. Finally, possibly optimal policy spaces (PO-PS) discards policy spaces among the minimal policy spaces that are not strictly better than other policy spaces (see \cite{lee2020characterizing} for details). 
    With a smaller number of policy spaces to consider, PO-PS converges faster than other strategies. MuO is better than the brute-force agent, which finds out the optimal policy among all possible options. Although brute-force agent is slow, it converges since the optimal policy space is included. However, the naive CB agent is unable to converge since, in this example, intervening on $X_2$ leads to suboptimal policy due to the mechanism involving the unobserved confounder between $X_2$ and $Y$. Average cumulative regret of PO-PS, MuO, BF, and naive CB respectively was 49.60, 195.03, 1134.84, and 3198.42.  \hfill $\blacksquare$
\end{experiment}

\paragraph{Discussion}
Mixed policy learning provides a flexible framework when variables to be intervened associated with natural mechanisms for how their values to be determined. We have showed that naively intervening on the underlying system while ignoring natural mechanisms can result in a suboptimal policy, incurring regrets indefinitely.
We have characterized mixed policies with and without contexts. In the case without contexts, complete characterizations for the minimality and possible-optimality of a policy space within a given mixed policy space are provided. With context, we provide a procedure to detect non-minimality. These characterizations do not require samples through interacting the given environment. \cite{lee2020characterizing} provide accounts on the possible-optimality of policy space with context. A complete characterization of both minimality and possible-optimality (partial-order) for the case of mixed policy space with contexts is an open problem.

\graphicspath{{ch_decisions_img/}}
\makeatletter
\def\input@path{{ch_decisions_img/}}
\makeatother

\section{Counterfactual Decision-Making (CRL Task 3)}\label{sec:_7_ctf-rand}
In this section, we investigate a novel type of interaction between the agent and the environment via layer 3 of the PCH.
For the tasks described so far, the agent interacts with the underlying environment through passive observations  (seeing),  active experimentation (doing), or a combination of both, which in the context of the PCH evokes layers 1 and 2-interactions. 
The new interactive modality will allow the agent to search over the large space of counterfactual policies, $\Pi_{\textsc{ctf}}$. 
More specifically, an online learning task with counterfactual randomization is characterized by a signature defined as follows:
\begin{align}
	\1T_{\text{ctf-rand}} = \Tuple{\1I= \text{ctf}, \1A= \emptyset, \Pi = \Pi_{\textsc{ctf}}, \1R = \D(\*Y) \mapsto \3R}.
\end{align}
This means that the agent will try to find a policy $\pi^*$ such that
\begin{align}\label{eq:opt-crl-ctf}
	 & \pi^* = \argmax_{\pi \in \Pi_{\textsc{ctf}}} \invEE{ \1R \left (\*Y \right )  \;\bigg\vert  \;\textcolor{red}{\mathcal{D}_{\text{ctf}} \sim P \Parens{\*V_{\*x} \mid \*x'}}  }{\pi}{\1M^*}, 
\end{align}
where the distinct feature of the task is the counterfactual type of interactions. To make this argument more precise, recall that for an experimental policy space $\Pi_{\textsc{exp}}$ (Def.~\ref{def:_3_space}), performing an intervention $\doo(\pi)$ following a policy $\pi \in \Pi_{\textsc{exp}}$ induces an interventional distribution $\inv{\*V}{\pi}$ (Def.~\ref{def:_2_2_inv_dist}), which follows from Fisherian randomization. On the other hand, interaction following a counterfactual policy $\pi \in \Pi_{\textsc{ctf}}$ allows the agent to access a specific type of counterfactual distribution, the $x$-specific effect of the decision on the outcome \cite[Sec.~4.1.1]{plecko:bar2022}.
\footnote{This quantity is also known in the literature as the effect of treated on the treated (ETT). For further discussion and historical context, refer to \citep{heckman1992randomization} and \citep[Ch.~8.2.]{pearl:2k}.} 

As it will become clear throughout this section, 
we will introduce a new type of counterfactual randomization (for short, \emph{ctf-rand}), which will allow the agent to behave optimally in challenging decision-making settings.
The following example illustrates one of such scenario using an instance of MAB environment, graphically described in Fig.~\ref{fig:_3_1_mab}.

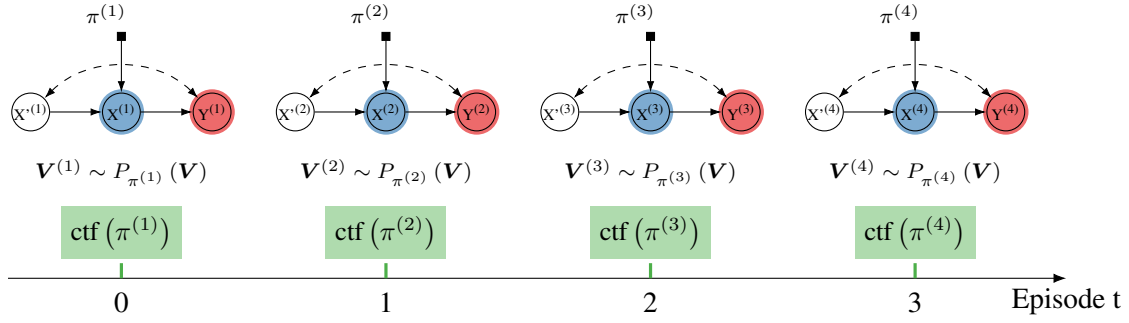
\begin{figure}[t]
	\centering
	\begin{tikzpicture}
		\def\outerr{3}
		\def\innerr{3}
		\def\dist{3.5}

		\draw[->, >={Latex}] (-1.5,0) -- (3*\dist+2,0) node[below] {Episode t};

		\node[vertex] (Xp1) at (-1.2, 2.2) {\tiny X'\textsuperscript{(1)}};
		\node[vertex] (X1) at (0, 2.2) {\tiny X\textsuperscript{(1)}};
		\node[vertex] (Y1) at (1.2, 2.2) {\tiny Y\textsuperscript{(1)}};
		\node[regime] (p1) at (0, 3.2) {};
		\node[draw=none] (text) at (-0.2, 3.5) {\scriptsize $\pi^{(1)}$};
		\draw[bidir] (Xp1) to [bend left = 45] (Y1);
		\draw[dir] (X1) -- (Y1);
		\draw[dir] (Xp1) -- (X1);
		\draw[dir] (p1) -- (X1);

		\node[vertex] (Xp2) at (\dist -1.2, 2.2) {\tiny X'\textsuperscript{(2)}};
		\node[vertex] (X2) at (\dist, 2.2) {\tiny X\textsuperscript{(2)}};
		\node[vertex] (Y2) at (\dist+1.2, 2.2) {\tiny Y\textsuperscript{(2)}};
		\node[regime] (p2) at (\dist, 3.2) {};
		\node[draw=none] (text) at (\dist-0.2, 3.5) {\scriptsize $\pi^{(2)}$};
		\draw[bidir] (Xp2) to [bend left = 45] (Y2);
		\draw[dir] (X2) -- (Y2);
		\draw[dir] (Xp2) -- (X2);
		\draw[dir] (p2) -- (X2);

		\node[vertex] (Xp3) at (2*\dist-1.2, 2.2) {\tiny X'\textsuperscript{(3)}};
		\node[vertex] (X3) at (2*\dist, 2.2) {\tiny X\textsuperscript{(3)}};
		\node[vertex] (Y3) at (2*\dist+1.2, 2.2) {\tiny Y\textsuperscript{(3)}};
		\node[regime] (p3) at (2*\dist, 3.2) {};
		\node[draw=none] (text) at (2*\dist-.2, 3.5) {\scriptsize $\pi^{(3)}$};
		\draw[bidir] (Xp3) to [bend left = 45] (Y3);
		\draw[dir] (X3) -- (Y3);
		\draw[dir] (Xp3) -- (X3);
		\draw[dir] (p3) -- (X3);

		\node[vertex] (Xp4) at (3*\dist-1.2, 2.2) {\tiny X'\textsuperscript{(4)}};
		\node[vertex] (X4) at (3*\dist, 2.2) {\tiny X\textsuperscript{(4)}};
		\node[vertex] (Y4) at (3*\dist+1.2, 2.2) {\tiny Y\textsuperscript{(4)}};
		\node[regime] (p4) at (3*\dist, 3.2) {};
		\node[draw=none] (text) at (3*\dist-0.2, 3.5) {\scriptsize $\pi^{(4)}$};
		\draw[bidir] (Xp4) to [bend left = 45] (Y4);
		\draw[dir] (X4) -- (Y4);
		\draw[dir] (Xp4) -- (X4);
		\draw[dir] (p4) -- (X4);

		\node (d1) at (0, 1.4) {\scriptsize $\*V^{(1)} \sim \inv{\*V}{\pi^{(1)}}$};
		\node (d2) at (\dist, 1.4) {\scriptsize $\*V^{(2)} \sim \inv{\*V}{\pi^{(2)}}$};
		\node (d3) at (2*\dist, 1.4) {\scriptsize $\*V^{(3)} \sim \inv{\*V}{\pi^{(3)}}$};
		\node (d4) at (3*\dist, 1.4) {\scriptsize $\*V^{(4)} \sim \inv{\*V}{\pi^{(4)}}$};

		\draw[very thick, bettergreen, -] (0, 0) -- (0, 0.2);
		\draw[very thick, bettergreen, -] (\dist, 0) -- (\dist, 0.2);
		\draw[very thick, bettergreen, -] (2*\dist, 0) -- (2*\dist, 0.2);
		\draw[very thick, bettergreen, -] (3*\dist, 0) -- (3*\dist, 0.2);

		\node [below] at (0, -0.05) { 0};
		\node [below] at (\dist, -0.05) { 1};
		\node [below] at (2*\dist, -0.05) { 2};
		\node [below] at (3*\dist, -0.05) { 3};

		\node [fill=bettergreen!45] at (0, 0.6) {\small $\ctf \Parens{\pi^{(1)}}$};
		\node [fill=bettergreen!45] at (\dist, 0.6) {\small $\ctf \Parens{\pi^{(2)}}$};
		\node [fill=bettergreen!45] at (2*\dist, 0.6) {\small $\ctf \Parens{\pi^{(3)}}$};
		\node [fill=bettergreen!45] at (3*\dist, 0.6) {\small $\ctf \Parens{\pi^{(4)}}$};

		\begin{pgfonlayer}{back}
			\node[circle,fill=betterblue!65,draw=none,minimum size=2*\innerr mm] at (X1) {};
			\node[circle,fill=betterred!65,draw=none,minimum size=2*\innerr mm] at (Y1) {};
			\node[circle,fill=betterblue!65,draw=none,minimum size=2*\innerr mm] at (X2) {};
			\node[circle,fill=betterred!65,draw=none,minimum size=2*\innerr mm] at (Y2) {};
			\node[circle,fill=betterblue!65,draw=none,minimum size=2*\innerr mm] at (X3) {};
			\node[circle,fill=betterred!65,draw=none,minimum size=2*\innerr mm] at (Y3) {};
			\node[circle,fill=betterblue!65,draw=none,minimum size=2*\innerr mm] at (X4) {};
			\node[circle,fill=betterred!65,draw=none,minimum size=2*\innerr mm] at (Y4) {};
		\end{pgfonlayer}
	\end{tikzpicture}
	\caption{Temporal diagram showing an agent interacting with the environment for repeated episodes through counterfactual policies.}
	\label{fig:_7_ctf}
\end{figure}

\begin{example}[Greedy Casino]\label{exp:_7_casino}
A group of investors decide to develop a new casino in Las Vegas and wants to make their machines as lucrative as possible at all costs, which we will call the Greedy casino incorporated (GCI). 
GCI's owners are determined about their mission and divide their efforts in three phases: research, setup, and operations. 

\paragraph{Phase 1. Research}
GCI's executives hire a team of cognitive scientists, psychologists, and cognitive scientists to investigate human's behaviors in the casinos' floors currently in operation around town. 
The team conducts a battery of studies and discovers that two features, out of hundreds examined, accurately predict the gambling behavior of players on a casino floor: each player's inebriation and the machine's conspicuousness (e.g., whether a machine is blinking and making noise).
Coding these traits as binary variables, we let $U_B \in \{0, 1\}$ denote whether or not a machine is blinking, and $U_D \in \{0, 1\}$ denote whether or not the gambler is drunk. 

As another outcome of the team's comprehensive study, they discover that the gambling population tends to prefer attracting less attention and naturally tends to be shy. However, their behavior changes when they become intoxicated, and they are more drawn towards the more effusive machines, such as those blinking and making noise. Formally, a gambler's 'natural' choice is described by the following mechanism (starting at 0):
\begin{align} 
		X \gets f_X(U_B, U_D) = \lnot (U_D \oplus U_B),  \label{eq:_7_policy}
\end{align}  
where $X = 1$ represents staying in the current machine, and $X=0$ represents switching to the neighbor machine. 
For instance, the gambler will stay in the slot machine they are currently in (``$X = 1$'') whenever the current machine is blinking ($U_B = 1$) and he is drunk ($U_D = 1$), or this machine is not blinking ($U_B = 0$) and he is sober ($U_D=0$). 
Alternatively, the gambler will get uncomfortable and switch machines (``$X = 0$'') whenever        the machine is blinking ($U_B = 1$) but he is not drunk ($U_D = 0$), or the machine is not blinking ($U_B = 0$) and he is drunk ($U_D = 1$). 
The table in Fig. \ref{tab:_7_mab_b} summarizes this behavior. 
\begin{figure}[t]
		\centering
		\hfill
		\begin{subfigure}[b]{0.5\textwidth}	\centering	\includegraphics[width=0.75\textwidth]{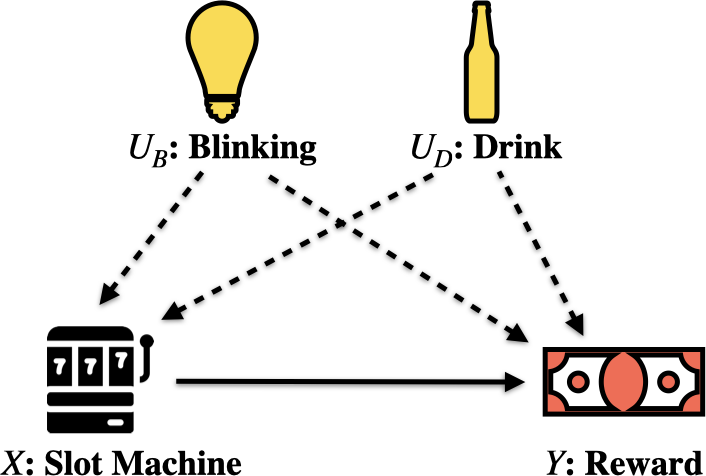}
		\caption{}
		\label{fig:_7_1_mab_a}
		\end{subfigure}\hfill
		\begin{subfigure}[b]{0.5\textwidth}\centering
    \begin{tabular}{|l|l|l|}
    \hline
        $U_B$ & $U_D$ & $X$ \\ \hline
        0 & 0 & 1 \\ \hline
        0 & 1 & 0 \\ \hline
        1 & 0 & 0 \\ \hline
        1 & 1 & 1 \\ \hline
    \end{tabular}			\caption{}
			\label{tab:_7_1_mab_b}
		\end{subfigure}
		\hfill \null
		\caption{(a) Illustration of the machines configurations. (b) Table summarizing gamblers natural predispositions.}
		\label{fig:_7_1_mab}
\end{figure}

\paragraph{Phase 2. Setup}
The GC owners take the information gathered during the research stage and devise a new plan to leverage it in order to maximize profitability.
Specifically, they purchase brand new machines with several important capabilities, including:
\begin{enumerate}[topsep=0pt,itemsep=-1ex,partopsep=1ex,parsep=1ex]
\item High-definition cameras capable of recording the gamblers' faces and body language.
\item New lighting and sound systems that enable the machines to blink and make noise.
\item The latest deep learning software that allows the machines to analyze the gamblers' behavior.
\end{enumerate}

The machines can be configured to operate in a network, so that they can share capabilities to make the look and feel of the casino's floor more pleasant. 
In practice, GC's executives decide to have the machines operate in pairs, meaning they share the same source of randomness. 
This configuration is illustrated in Fig.~\ref{fig:_7_1_mab_a}, where a blinking/noisy machine is paired with a non-blinking/silent one. 

The gambler arrives at one of the machines and can immediately play it or switch to the alternative one. 
(Importantly, the gamblers are already drunk or not just before arriving, and the machines are already blinking or not regardless of the gambler, those are two independent events.)   
We will call the former by machine 1 ($X=1$) and the latter by 0 ($X=0$). 
Also, the outcome of each gamble is represented by a variable $Y$, where $0$ means that the patron lost, and $1$ otherwise. 

Moreover, to keep the system in an equilibrium, and under the radar, the casino distributes free drinks and sets up the machines such that every gambler has an equal chance of being intoxicated and each machine has an equal chance of blinking its lights at a given time. 
In formal notation, this means that $P(U_D = 0) = P(U_D = 1) = 0.5$ and $P(U_B = 0) = P(U_B = 1) = 0.5$.

The owners are also cognizant of current's state gambling regulations that require casinos to maintain a minimum attainable payout rate for slots of 30\%. 
While still wanting to maximize profits, GCI executives decide to take advantage of the players' propensities by leveraging the machines' sensing capabilities.
They then set up the payout rates $f_Y$ of the machines as depicted in Table~\ref{tab:_7_mab_a}.
\begin{table}[t]
	\centering
	\hfill
	\begin{subfigure}[b]{0.5\textwidth}\centering
		\begin{tabular}{@{}c|rr|rr@{}} \toprule
					& \multicolumn{2}{c|}{$U_D=0$} & \multicolumn{2}{c}{$U_D=1$}                     \\ \cmidrule(r){2-3} \cmidrule(l){4-5}
					& $U_B$ = 0                    & $U_B$ = 1                   & $U_B$ = 0 & $U_B$ = 1 \\ \midrule
			$X = 0$ & 0.50                      & *0.10                      & *0.20    & 0.40   \\
			$X = 1$ & *0.10                       & 0.50                     & 0.40   & *0.20    \\ \bottomrule
		\end{tabular}
		\caption{}
		\label{tab:_7_mab_a}
	\end{subfigure}\hfill
	\begin{subfigure}[b]{0.5\textwidth}\centering
		\begin{tabular}{@{}c|rr@{}} \toprule
					& $\E[Y | X]$ & $\invE{Y}{x}$ \\ \midrule
			$X = 0$ & 0.15        & 0.3           \\
			$X = 1$ & 0.15        & 0.3           \\ \bottomrule
		\end{tabular}
		\caption{}
		\label{tab:_7_mab_b}
	\end{subfigure}
	\hfill \null
	\caption{(a) Payout rates $P(Y = 1 \mid X, U_D, U_B)$ decided by reactive slot machines as a function of arm choice, sobriety, and machine conspicuousness. Players' natural arm choices under $U_D, U_B$ are indicated by asterisks. (b) Payout rates according to the observational, $P(Y=1|X)$, and interventional $\inv{Y=1}{x}$, distributions, where $Y=1$ represents winning (shown in the table).}
	\label{tab:_7_mab}
\end{table}
In words, if $U_D=0,U_B=0$, the player will naturally select action $X=1$, following Eq.~\ref{eq:_7_policy}, which will lead to a positive outcome, $Y=1$, only $10\%$ of the time (the same with $U_D=1, U_B=1$). 
Also, if $U_D=0,U_B=1$ (or $U_D=1, U_B=0$), the player will decide for action $X=0$, to switch, which will lead to a positive outcome $20\%$ of the time. 
These configurations are marked with an $*$ in the table.

\paragraph{Phase 3. Operational}
The GC debuts and is a big hit; many new patrons enjoy their evenings playing in the new machines. 
Interestingly, they are not aware (conscious) that their behavior is influenced by their inebriation and whether the machine is looking conspicuous. The variables $U_B, U_D$ are exogenous and remain unobserved, following the causal language introduced earlier. 

Still, some patrons know about GC owners' reputations and are suspicious of the casino's ethical standards. 
These patrons decide to collect some data on the other gamblers' behavior, through random sampling, which leads to the distribution shown in Table~\ref{tab:_7_mab_b}. 
In other words, it seems that the casino is paying ordinary gamblers only 15\% of the time. Also, no matter whether they play machine $X=0$ or $X=1$, the average payout is the same. 

The state is called to investigate the issue and, being blind to the GC's payout strategy, claims that this data is observational (non-causal), and, therefore, is ``invalid''.
They then decide to conduct a randomized study to verify whether the win rates in the floor meet the legal standards. 
The government's inspectors follow the RCT procedure discussed in Sec.~\ref{sec:_4_2}. 
First, they recruit random players from the casino floor, pay them to play a random slot, and then observe the outcome. 
The experiment yields a favorable outcome for the casino, with win rates precisely meeting the 30\% cutoff -- no more, no less than. 
The data looks like Table~\ref{tab:_7_mab_b}, and is again insensitive to the machine's choice.

As RL enthusiasts, we decide to run a series of experiments using more refined and sample-efficient adaptive strategies (e.g., $\epsilon$-greedy, Thompson Sampling, UCB1, EXP3) to test the new slot machines on the casino's floor. 
We obtain data encoded in Fig.~\ref{fig:_7_regret}. The first plot shows that the probability of choosing the correct action is no better than a random coin flip even after a considerable number of steps. 
We note, somewhat surprised, that the cumulative regret continues without abating, indicating our inability to learn a superior arm. 
We also realize that the results obtained by the standard algorithms align with the randomized study (orange line).

After all, the casino seems to be, at the same time, (1) exploiting gamblers' natural predilections as a function of their intoxication and the machine's blinking behavior (based on Eq.~\ref{eq:_7_policy}), (2) paying, on average, less than the legally allowed (15\% instead of 30\%), and (3) fooling state's inspectors since the randomized trial payout meets the legal requirement. \hfill $\blacksquare$
\end{example}

Some observations are worth noting after this example. Firstly, the situation described in the greedy casino is far from contrived. There is a growing body of literature in the cognitive sciences that recognizes a significant aspect of human decision-making occurring at a subconscious level, with individuals often unaware of the reasons behind their actions \citep{kouider2010rich}.
\footnote{Interestingly, the work of psychology Professor Daniel Kahneman, a Nobel Prize laureate, revolves around recognizing and studying various biases and mechanisms in human decision-making. For more insights, refer to \citep{tversky1974judgment,bargh1999unbearable,dijksterhuis2006theory} for a survey on these results.}"

\begin{figure}[t]
	\centering
	\hfill
	\begin{minipage}[b]{0.45\linewidth}
		\centering
		\includegraphics[width=1.0\textwidth]{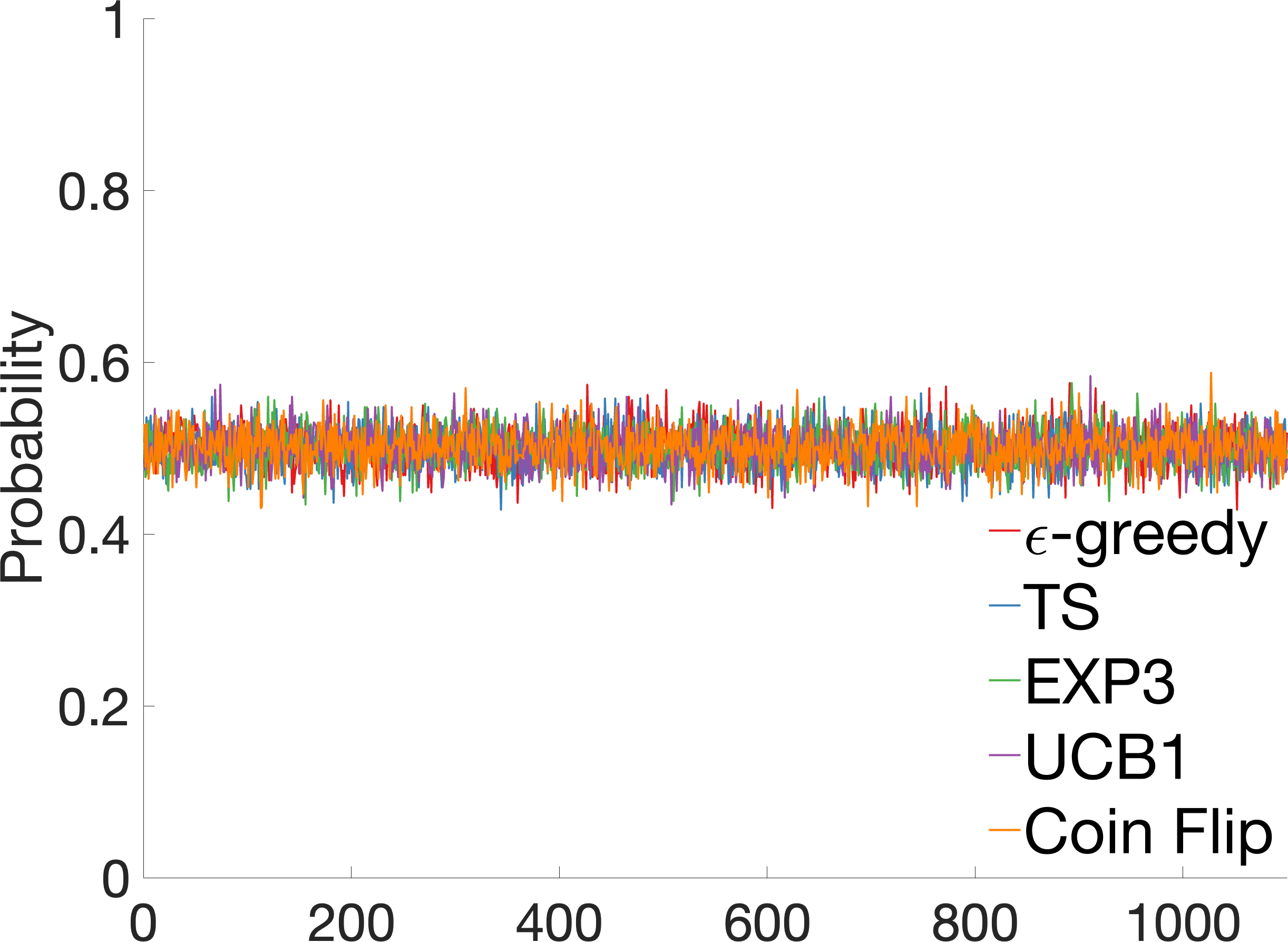}
		\subcaption{}
		\label{fig:_7_regret_a}
	\end{minipage}\hfill
	\begin{minipage}[b]{0.45\linewidth}
		\centering
		\includegraphics[width=1.0\textwidth]{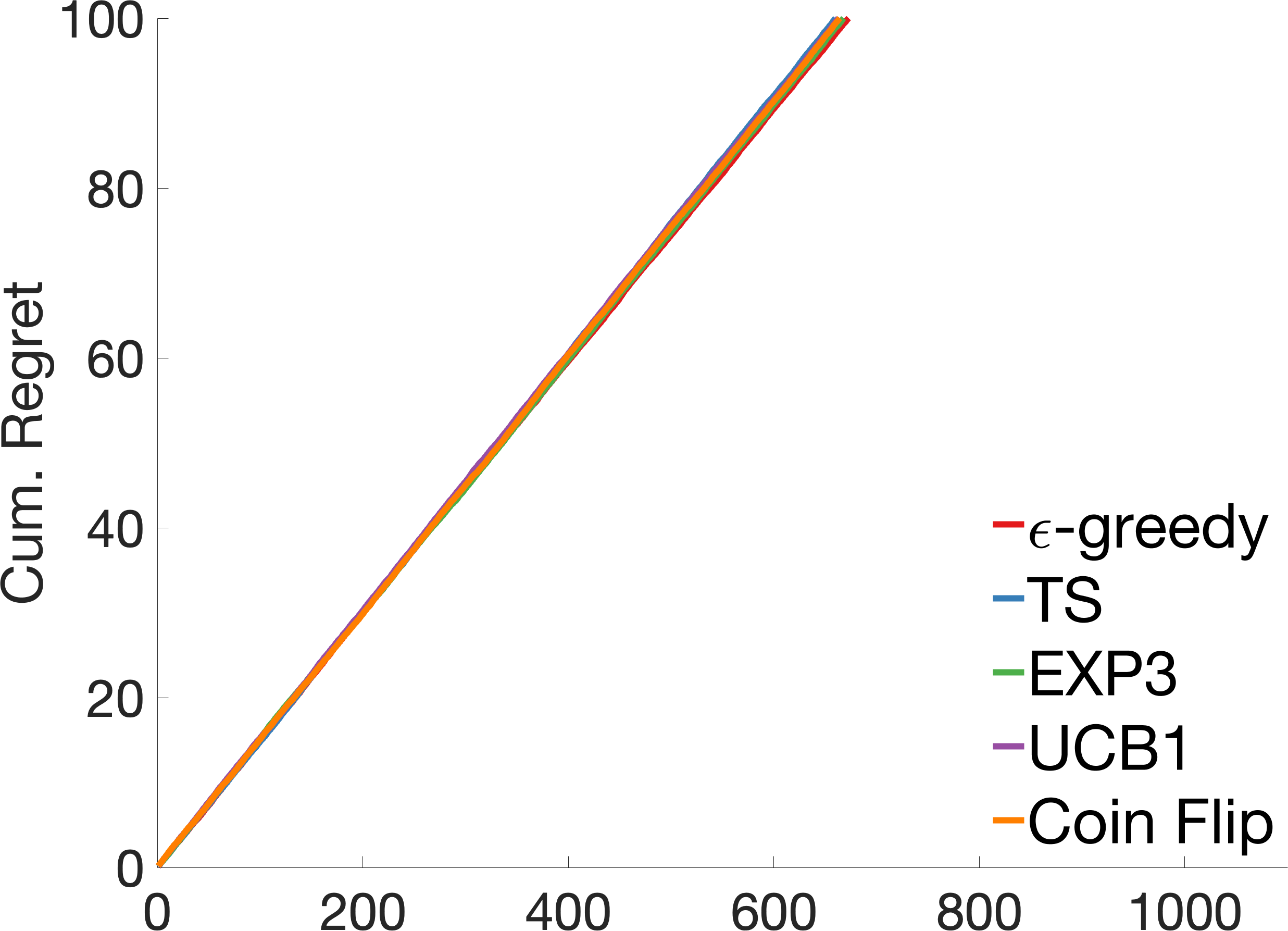}
		\subcaption{}
		\label{fig:_7_regret_b}
	\end{minipage}\hfill\null
	\caption{Performance of different bandit strategies in the greedy casino example; x-axis represents the total episodes of interactions. (\subref{fig:_7_regret_a}) No algorithm is able to perform better than random guessing.  (\subref{fig:_7_regret_b}) Regret grows without bounds.} \label{fig:_7_regret}
\end{figure}

Second, under the presence of unobserved confounders,  such as in the GC example, the interventional quantity $\invE{Y}{x}$ used throughout the RL literature does not seem to capture critical information required to maximize payout, but rather the average payout akin to choosing arms by a coin flip (as shown in the plot earlier). 
Specifically, the payout given by coin flipping is the same for both machines, 
\begin{align}
\invE{Y}{X \gets 0} = \invE{Y}{X \gets 1} = 0.3,
\end{align}
which means that the arms are statistically indistinguishable in the limit of a large sample size. 
Further, if we consider using the observational data from watching gamblers on the casino floor (based on their natural predilections), the average payoff is also independent of the machine choice,
\begin{align}
\E \Brackets{Y = 1 \mid X = 0} = \E \Brackets{Y = 1 \mid X = 1} = 0.15, 	
\end{align}
albeit with an even lower payout.
Based on these observations, we can see why no arm choice is better than the other under either distribution alone, which explains the reason any algorithm based on these distributions will fail to learn an optimal policy. 

Third, and more fundamentally, one may be puzzled by the discrepancy between observational and interventional distributions.
After all, even though the interventional distribution refers to the causal effect, the typical player receives payouts that come from the observational data; how can this be reconciled? 
Furthermore, could this difference reveal insights about the unobserved confounders, offering clues on how to differentiate the arms?
Lastly, from a more practical standpoint, acknowledging this phenomenon and considering the data in Table~\ref{tab:_7_mab_b}, what would be the optimal way to play at the GC? Is it possible to devise a strategy that yields a higher payout than the two methods previously discussed? 

In this section, our goal is to further understand and answer these questions. 
To achieve this, we will introduce novel causal machinery designed to exploit PCH's layer 3 distributions, thereby enabling the agent to achieve higher performance in challenging decision-making scenarios, such as the greedy casino discussed in Example~\ref{exp:_7_casino}. 
We aim to formalize the concept of counterfactual randomization, including canonical environments such as MABs (Fig.~\ref{fig:_3_1_mab}) and MDPs (Fig.~\ref{fig:_3_1_mdp}). 
We will provide a systematic augmentation procedure that empowers existing online learning agents to optimize counterfactual policies in these environments. 
The contributions of these sections are summarized as follows: 
\begin{itemize}
	\item Sec.~\ref{sec:_7_1} proposes a novel \emph{counterfactual decision criterion} in bandit models. This strategy determines the values of actions based on a specific type of counterfactual quantity, namely, the effect of the treatment on the treated. This approach leads to a new family of counterfactual policies that take advantage of the agent's intended actions (i.e., intuitions). Sec.~\ref{sec:_7_1_1} extends this counterfactual criterion to a canonical family of sequential decision-making environments, specifically Markov decision processes with unobserved confounders (MDP).

	\item Sec.~\ref{sec:_7_2} develops a novel \emph{counterfactual randomization} to enable the realization of counterfactual decision-making in the underlying environment. Sec.~\ref{sec:_7_2_1} extends this novel randomization procedure to enhance existing online algorithms in MDP environments. We demonstrate this procedure by augmentating a state-of-art MDP algorithm \texttt{UCBVI} \citep{azar2017minimax}. The analysis suggests that the new counterfactual randomization strategy is statistically efficient and consistently outperforms standard online algorithms that lack counterfactual reasoning.
	
	\item Sec.~\ref{sec:_7_3} connects the idea of counterfactual policies and the notion of autonomy in decision-making. In particular, we introduce a novel trade-off between \emph{autonomy and optimality}: while a fully autonomous system is preferable, discarding the human input could harm the optimal performance of the decision system, leading to suboptimal strategies. An effective planning algorithm is developed to balance this autonomy-optimality trade-off, enabling an agent to learn an optimal counterfactual policy in an MDP environment under a budget constraint on the frequency of using human input.
\end{itemize}

\subsection{Counterfactual Decision Criterion}\label{sec:_7_1}
We begin the discussion by describing the causal mechanisms that encode the agent's interaction regimes with the MAB environment. 
First, when the agent passively observes events unfolding in the underlying causal model, the underlying causal mechanisms remain unchanged. Fig.~\ref{fig:_7_1_see} illustrates the causal diagram of a canonical MAB environment. 
Note that $f_X$ represents the system's behavioral policy generating the observational data. It takes as input the unobserved factors $U$ affecting the reward signal $Y$ and determines an observed arm choice $X$; the presence of $U$ is represented by the bidirected arrow $X \bidirectarrow Y$ (Def.~\ref{def:_2_3_diagram}). 
One way to interpret this arrow is through the concept of players' natural predilections. For instance, in the greedy casino (Example~\ref{exp:_7_casino}), the predilection could correspond to choices made by gamblers when allowed to play freely on the casino's floor (e.g., intoxicated players favoring blinking machines), or doctors prescribing drugs based on their ``gut feeling'' (e.g., physicians prescribing the more expensive drug to wealthier patients). 
The rewards associated with these predilections are encoded in the observational quantity $\E\Brackets{Y \mid x}$.

On the other hand, the interventional quantity $\invE{Y}{x}$ encodes the reward induced by the process in which the natural predilections are overridden or ceased by external and deliberate policies. 
In the casino's example, this reward arises when the government's inspectors flip a coin and direct gamblers to machines based on the coin's outcome via intervention $\doo(x)$, regardless of their natural predilections. Fig.~\ref{fig:_7_1_do} depicts the causal diagram of the post-interventional MAB environment. Here, the bidirected arrow between $X $ and $Y$ is removed, as the unobserved confounder $U$ does not influence the intervention $\doo(x)$ and how the value of $X$ attains its value. The exogenous distribution $P(U)$ and the reward function $f_Y(X, U)$ encompass the expected reward parameters for each arm, which are typically the focus of analysis in RL literature. 
\footnote{The standard MAB literature focuses on the unconfounded case depicted in Fig.~\ref{fig:_7_1_do}), where $\invE{Y}{x} = \E[Y \mid x]$ (Def.~\ref{def:_4_1_nuc}). Throughout this section, we focus on the general setting where the NUC assumption does not hold and the unobserved confounder $U$ cannot be ruled out a priori.}

Among the two interaction regimes described above, the observational agent (``see'') does not deliberately search for a better alternative other than its natural predilections. 
 Meanwhile, the interventional agent (``do'') explores the environment by randomly playing arms, thus disregarding its natural predilections. 
 Remarkably, it is possible to leverage the information embedded in these distinct interaction regimes (and their corresponding distributions) to understand and improve the agents' natural predilections in these MAB instances.

To witness, assume the agent is now more introspective and starts operating through the following protocol on the casino's floor. 
The gambler observes itself and intercepts its decision-flow when is just about to pull the arm of machine ``$0$''. He then contemplates whether following his natural predilection (``$0$'') or going against it (playing ``$1$'') would lead to a better outcome. 
The drunk gambler, for example, who is a clever machine learning student and familiar with Figs.~\ref{fig:_7_1_see} and \ref{fig:_7_1_do}, says that such evaluation cannot be computed a priori. 
He affirms that, despite spending hours on the casino estimating the expected payoff based on players' natural predilections (namely, $\E[Y \mid x]$), it is not feasible to relate this natural predilection with the hypothetical construction what would have happened had he decided to play differently. 
He further acknowledges that the interventional quantity $\invE{Y}{x}$, devoid of the gamblers' predilections, does not support any clear comparison against his personal strategy. The oracle says this type of reasoning is possible, but first one needs to define the concept of counterfactual distributions: 
\begin{definition}[Counterfactual Distribution \citep{bareinboim2020pearl}]\label{def:_7_1_ctf}
\sloppy	An SCM $\M= \tuple{\*U, \*V,$ $\2F, P}$ induces a family of joint distributions over counterfactual events $\*Y_{\*x}, \ldots, \*Z_{\*w}$, for any $\*Y, \*Z, \dots, \*X, \*W \subseteq \*V$:
	\begin{align}
		P\Parens{\*y_{\*x}, \dots, \*z_{\*w}} \equiv \sum_{\*u} \I\left \{ \*Y_{\*x}(\*u) = \*y, \dots, \*Z_{\*w}(\*u) = \*w\right\}P(\*u), \label{eq:_7_1_ctf}
	\end{align} \hfill $\blacksquare$
\end{definition}
Note that the l.h.s. of Eq.~\ref{eq:_7_1_ctf} contains variables with different subscripts, which, syntactically, encode different counterfactual worlds. The evaluation implied by this equation can be described as the following process:
\begin{enumerate}
	\item For each set of subscripts relative to each set of variables (e.g., $\doo(\*x), \dots, \doo(\*w)$ for $\*Y, \dots$, $\*Z$, respectively), replace the corresponding mechanisms with the appropriate constants and generate  $\2F_{\*x}, \dots, \2F_{\*w}$ (Def.~\ref{def:_2_2_submodel}), creating submodels $\M_{\*x},\dots, \M_{\*w}$ (of $\M$);
	\item For each situation $\*U = \*u$, the environment evaluates the modified causal mechanisms (e.g., $\2F_{\*x}, \dots, \2F_{\*w}$) following a valid order (i.e., any variable in the l.h.s. is evaluated after the ones in the r.h.s.) to obtain the potential responses of the observables, and
	\item  The probability mass $P(\*U = \*u)$ is then accumulated for each instantiation $\*U = \*u$ that is consistent with the events over the counterfactual variables -- for instance, ${\*Y}_{\*x}={\*y}\dots, {\*Z}_{\*w}={\*z}$, i.e., ${\*Y} = {\*y}, \dots, {\*Z} = {\*z}$ in the submodels $\M_{\*x}, \dots, \M_{\*w}$, respectively.
\end{enumerate}

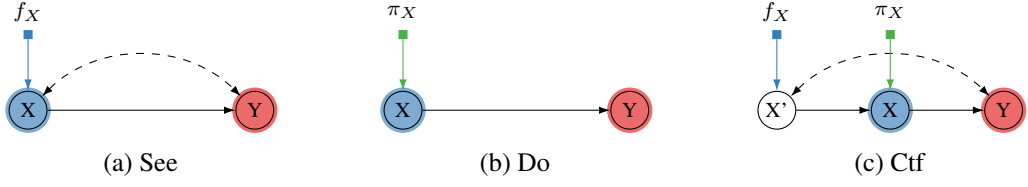
\begin{figure}[t]
	\hfill%
	\begin{subfigure}{0.3\linewidth}\centering
		\begin{tikzpicture}
			\def\outerr{3.2}
			\def\innerr{3}

			\node[vertex, opacity=0] (U) at (1, 1.7) {};
			\node[vertex] (X) at (0, 0) {X};
			\node[vertex] (Y) at (3, 0) {Y};
			\draw[bidir] (X) to [bend left = 45] (Y);

			\node[regime, betterblue] (p) at (0, 1) {};
			\node[draw=none] (text) at (0, 1.3) {\scriptsize $f_X$};
			\draw[dir, betterblue] (p) -- (X);

			\draw[dir] (X) -- (Y);

			\begin{pgfonlayer}{back}
				\node[circle,fill=betterblue!65,draw=none,minimum size=2*\innerr mm] at (X) {};
				\node[circle,fill=betterred!65,draw=none,minimum size=2*\innerr mm] at (Y) {};
			\end{pgfonlayer}
		\end{tikzpicture}
		\caption{See}
		\label{fig:_7_1_see}
	\end{subfigure}\hfill
	\begin{subfigure}{0.3\linewidth}\centering
		\begin{tikzpicture}
			\def\outerr{3.2}
			\def\innerr{3}

			\node[vertex, opacity=0] (U) at (1, 1.7) {};
			\node[vertex] (X) at (0, 0) {X};
			\node[vertex] (Y) at (3, 0) {Y};
			\draw[bidir, opacity = 0] (X) to [bend left = 45] (Y);

			\node[regime, bettergreen] (p) at (0, 1) {};
			\node[draw=none] (text) at (0, 1.3) {\scriptsize $\pi_X$};
			\draw[dir, bettergreen] (p) -- (X);

			\draw[dir] (X) -- (Y);

			\begin{pgfonlayer}{back}
				\node[circle,fill=betterblue!65,draw=none,minimum size=2*\innerr mm] at (X) {};
				\node[circle,fill=betterred!65,draw=none,minimum size=2*\innerr mm] at (Y) {};
			\end{pgfonlayer}
		\end{tikzpicture}
		\caption{Do}
		\label{fig:_7_1_do}
	\end{subfigure}\hfill
	\begin{subfigure}{0.3\linewidth}\centering
		\begin{tikzpicture}
			\def\outerr{3.2}
			\def\innerr{3}

			\node[vertex, opacity=0] (U) at (1, 1.7) {};
			\node[vertex] (X) at (0, 0) {X'};
			\node[vertex] (Xp) at (1.5, 0) {X};
			\node[vertex] (Y) at (3, 0) {Y};
			\draw[bidir] (X) to [bend left = 45] (Y);

			\node[regime, betterblue] (p1) at (0, 1) {};
			\node[draw=none] (text) at (0, 1.3) {\scriptsize $f_X$};
			\draw[dir, betterblue] (p1) -- (X);

			\node[regime, bettergreen] (p2) at (1.5, 1) {};
			\node[draw=none] (text) at (1.5, 1.3) {\scriptsize $\pi_X$};
			\draw[dir, bettergreen] (p2) -- (Xp);

			\draw[dir] (Xp) -- (Y);
			\draw[dir] (X) -- (Xp);

			\begin{pgfonlayer}{back}
				\node[circle,fill=betterblue!65,draw=none,minimum size=2*\innerr mm] at (Xp) {};
				\node[circle,fill=betterred!65,draw=none,minimum size=2*\innerr mm] at (Y) {};
			\end{pgfonlayer}
		\end{tikzpicture}
		\caption{Ctf}
		\label{fig:_7_1_ctf}
	\end{subfigure}\hfill\null
	\caption{Causal diagrams for different interaction regimes in the MAB environment.}
	\label{fig:_7_1_interact}
\end{figure}
Among quantities in Def.~\ref{def:_7_1_ctf}, the counterfactual event $\*Y_{\*x} (\*u) = \*y$ could be read as a sentence: ``$\*Y$ would be $\*y$ (in situation $\*U = \*u$), had $\*X$ been $\*x$.'' This definition of counterfactuals naturally leads to the judgment suggested by the oracle.
\begin{example}[Counterfactual reasoning through SCMs (GC continued)]\label{exp:_7_1_casino}
\sloppy Consider again the MAB model $\1M^*$ describing the greedy casino environment in Example~\ref{exp:_7_casino}. 
We are interested in evaluating whether the new statement, ``Would I (the agent) win ($Y = 1$) had I played $X = 0$?'' 
can be formally written as a counterfactual event. 
Assuming that the agent's natural predilection is to play machine $X = 1$, the agent now engages in introspection to compare the odds of winning following his ``gut feeling'' or going against his intuition. 
This statement can be written in counterfactual language, formally,	as 
\begin{align}\label{eq:xspecific1} 
\E \Brackets{Y_{X \gets 0} \mid X = 1}, 
\end{align}	
which reads as ``the expected value of winning ($Y = 1$) had I played $X = 0$ given that I am about to play $X=1$''. 
This statement contrasts with the alternative quantity
\begin{align}\label{eq:xspecific2}
\E \Brackets{Y_{X \gets 1} \mid X = 1}
\end{align}
which reads as ``the expected value of winning ($Y = 1$) had I played $X = 1$ given that I am about to play $X=1$''.
This quantity is also called the $X$-specific effect of $X$ on $Y$ \cite[Sec.~4.1.1]{plecko:bar2022}.. 

More specifically, for quantities in the form of Eq.~\ref{eq:xspecific2}, the composition axiom \citep[Ch.~7.3]{pearl:2k} implies that
\begin{align}
\E \Brackets{Y_{X \gets 1} \mid X = 1} = \E \Brackets{Y \mid X = 1},
\end{align}
where the l.h.s. is computable from the observational distribution $P(X, Y)$. 
Using the previous discussion in Example~\ref{exp:_7_casino}, computing the above equation gives 
\begin{align}
\E \Brackets{Y_{X \gets 1} \mid X = 1} = 0.15
\end{align}

The counterfactual quantity in the form of Eq.~\ref{eq:xspecific1} could be written as
	\begin{align}
		\E \Brackets{Y_{X \gets 0} \mid X = 1} & = P(Y_{X\gets 0}=1 \mid X=1)                                     \\
			& = \frac{P(Y_{X\gets 0}=1, X=1)}{P(X=1)} \label{eq:_7_1_casino_1}
	\end{align}
	where the denominator is trivially obtainable since it only involves observational probabilities. 
	Following the behavioral policy in Eq.~\ref{eq:_7_policy}, we obtain
	\begin{align}
		P(X = 1) & = P(U_D  = 0, U_B = 1) + P(U_D  = 1, U_B = 0) \\
		         & = 0.5
	\end{align}
On the other hand, the numerator, $P(Y_{X \gets 1}=1, X=0)$ is more interesting and refers to two different worlds and cannot be written in the languages of observational and interventional distributions since they do not allow for probability expressions involving more than one subscript (each encoding a different version of the environment).  Using the procedure dictated in Eq.~\ref{eq:_7_1_ctf}, we obtain
	\begin{align}
		P(Y_{X \gets 0}=1, X=1) & = \sum_{u_B,u_D} \I  \{Y_{X \gets 0}(u_B, u_D)=1, X(u_B, u_D)=1\} P(u_B, u_D)
	\end{align}
Noting that $U_D=1, U_B=1$ and $U_D=0, U_B=0$ are not compatible with the event $X=0$, through Eq.~\ref{eq:_7_policy}, we can write:	
\begin{align}\label{eq:ctf-eval1}
P(Y_{X \gets 0}=1, X=1) = 0.25 \times 
\Big ( & \{Y_{X \gets 0}(1, 0)=1, X(1,0)=1\}         \\
& +  \{Y_{X \gets 0}(0, 1)=1, X(0,1)=1\} \Big )
\end{align}
Considering the first factor and going back to $f_Y$, as shown in Table 12a, we note that $Y=1$ when $D=0, B=1, X=0$ with probability $0.5$, and when $D=1, B=0, X=0$ with probability $0.4$. 
Putting this back into Eq.~\ref{eq:_7_1_casino_1} leads to: 
\begin{align}
P(Y_{X \gets 0}=1, X=1) = \frac{0.25}{0.5} * 0.9  = 0.45 
\end{align}

For completeness, we also compute the other values for the $x$-specific effect, $\E[Y_{x} \mid x']$, for all $x, x' = 0, 1$, as shown in Table~\ref{tab:_7_mab_ett}. 
The conclusion following this analysis is clear -- the payout rate would have tripled had the agent played machine $X = 0$ in situations where its natural predilections suggest $X = 1$, and machine $X = 1$ in situations where its natural predilections suggest $X = 0$. 
	\begin{table}[t]
		\centering
		\begin{tabular}{@{}c|r|r@{}} \toprule
			        & $\E[Y_x \mid X = 0]$ & $\E[Y_x \mid X = 1]$ \\ \midrule
			$X = 0$ & 0.15                 & *0.45                \\
			$X = 1$ & *0.45                & 0.15                 \\ \bottomrule
		\end{tabular}
		\caption{Payout rates according to the counterfactual distribution $P\Parens{Y_{X \gets x} \mid X = x'}$.}
		\label{tab:_7_mab_ett}
	\end{table} \hfill $\blacksquare$
\end{example}
The counterfactual analysis in the previous example suggests a novel decision-making criterion in the MAB environment.
Instead of using a decision rule comparing the average payouts associated with arms, namely (for action $x$),
\begin{align}
	x^* = \argmax_{x} \invE{Y}{x},
\end{align}
we should consider the rule using the comparison between the average payouts obtained by players for choosing in favor or against their intuition, respectively,
\begin{align}
	\pi^*(x) = \argmax_{x} \E\Brackets{ \underbrace{Y_{X \gets x}}_{\text{realized action}} \mid \underbrace{X = x'}_{\text{intended action}}}, \;\; \forall x \in \{0, 1\} \label{eq:_7_1_ctf_opt}
\end{align}
where $x'$ is the player's natural predilection, and $x$ is their final decision. 

We call this procedure \textit{counterfactual decision criterion} (CDC) to emphasize the counterfactual nature of this reasoning step and the idea of either following or disobeying the agent's intuition. 
Remarkably, CDC takes into account the agent's individuality and the fact that their natural inclination provides valuable information about the confounders that also affect the payout, importantly, even when unknown to the very agent. 
In the binary case, for example, assuming that $X = 1$ is the player's natural choice at some time step, if 
\begin{align}
\E \Brackets{Y_{X \gets 0} \mid X = 1} \geq \E \Brackets{Y_{X \gets 1} \mid X = 1}, 	
\end{align}
this would suggest that the player should act against their intuition, i.e., refraining of playing machine $X = 1$ in favor of playing machine $X = 0$.
Conversely, if 
\begin{align}
\E \Brackets{Y_{X \gets 0} \mid X = 1} \leq \E \Brackets{Y_{X \gets 1} \mid X = 1}, 	
\end{align}
it would imply that the player should follow their intuition, in this case, playing machine $X = 1$. 
Performing CDC leads to a novel type of counterfactual policies in MAB environment. 
\begin{definition}[Counterfactual Policy - MAB]\label{def:_7_1_ctf_space}
Let $\1M$ be an MAB environment graphically described in Fig.~\ref{fig:_7_1_see}. 
A counterfactual policy space $\Pi_{\textsc{ctf}}$ is a collection of policies $\pi(X \mid X')$ mapping from the domain of an intended action $X'$ to the space of probability distribution over the domain of a realized action $X$. 
Henceforth, we will consistently denote such a policy space by $\Pi_{\textsc{ctf}} = \Braces{\Tuple{X, \{X'\}}}$. 
\end{definition}
At first glance, a counterfactual policy $\pi(X \mid X')$ might be counter-intuitive since the input and output of the policy $\pi$ appear to be the same, over the domain of action variable $X$. 
This suggests that deploying a counterfactual policy introduces a self-reference in the underlying environment. 
Semantically speaking, this is a well-defined counterfactual quantity and computable for any SCM $\1M$. 
While its input $X'$ and output $X$ share the same domain, they represent action variables in two different worlds --  the input $X'$ is the agent's natural predilection, similar to the obtained in the observational distribution in layer 1 (Fig.~\ref{fig:_7_1_see}), while the output $X$ is the agent's realized action generating the interventional distribution in layer 2 (Fig.~\ref{fig:_7_1_do}). 
Determining the realized intervention $\doo(X \gets x)$ based on the agent's natural predilection $X = x'$ leads to the expected reward of a counterfactual policy $\pi(X \mid X')$. 
\begin{definition}[Counterfactual Submodel, MAB]\label{def:_7_2_submodel}
	Let $\1M$ be an MAB environment graphically described in Fig.~\ref{fig:_7_1_see} such that 
\begin{align}
\M = \Tuple{\*U = \{U\}, \*V = \{X, Y\}, \2F = \{f_X, f_Y\}, P = P(U)}	
\end{align}	
Let $\pi(X \mid X')$ be a counterfactual policy over action $X$. A submodel $\1M_{\pi}$ of $\1M$ is a modified SCM
	\begin{align}
		\1M_{\pi} = \Tuple{\*U = \{U\}, \*V_{\pi} = \{X', X, Y\}, \2F_{\pi},  P = P(U)},
	\end{align}
	where the structural functions $\2F_{\pi}$ are defined as
	\begin{align}
		\2F_{\pi} = \begin{cases}
			            X' \gets f_X(U)         \\
			            X \sim \pi(X \mid X') \\
			            Y \gets f_Y(X', U)
		            \end{cases} \label{eq:_7_2_submodel}
	\end{align} 	\hfill $\blacksquare$
\end{definition}
Fig.~\ref{fig:_7_1_ctf} provides a graphical representation of the submodel $\1M_{\pi}$ induced by a counterfactual policy $\pi(X \mid X')$. 
Note that a new, virtual variable $X'$ is added to represent the intended action, which is used as an input to determine the realized action $X$ affecting the reward signal $Y$. Formally, we define the expected reward $\invE{Y; \1M}{\pi}$ of a counterfactual policy $\pi(X \mid X')$ evaluated in an MAB environment as the expected reward $\E \Brackets{Y ; \1M_{\pi}}$ evaluated in the submodel $\1M_{\pi}$. That is,
\begin{align}
	\invE{Y; \1M}{\pi} & = \sum_{x', x} \E \Brackets{Y \mid x, x'; \1M_{\pi}} P\left (x \mid x'; \1M_{\pi} \right) P\left( x'; \1M_{\pi} \right)               \\
	                            & =\sum_{x', x} \E \Brackets{Y \mid x, x'; \1M_{\pi}} \pi \left (x \mid x' \right) P\left(x' \right)                                  
\end{align}
The last step holds since in the counterfactual submodel $\1M_{\pi}$ defined by Eq.~\ref{eq:_7_2_submodel}, the values of the intended action $X'$ are decided by the behavioral policy $f_X$ determining action in the original SCM $\1M$. 
Since the counterfactual policy $\pi$ does not take the unobserved confounder $U$ affecting other variables in the system, conditioning on the intended action $X'$ ``blocks'' the backdoor paths from the realized action $X$ to the subsequent reward $Y$. Computing the expected reward of $Y$ conditioning on both the intended and realized actions $X', X$ in submodel thus recovers the counterfactual ETT quantity. The above equation can be further written as 
\begin{align}
	\invE{Y; \1M}{\pi} & =\sum_{x, x'} \E \Brackets{Y_{X \gets x} \mid X = x'} \pi \left (x \mid x' \right) P\left(x' \right)           \label{eq:_7_1_reward}
\end{align}
The following example demonstrates counterfactual policies in the Greedy Casino environment described previously in Example~\ref{exp:_7_casino}.
\begin{example}\label{exp:_7_2_mab1}
	Consider the MAB environment $\1M^*$ described in Example~\ref{exp:_7_casino}. 
	Performing intervention $\ctf(\pi)$ following a counterfactual policy $\pi \triangleq X \gets \neg X'$ defines a submodel $\1M^*_{\pi}$ described by the following tuple
	\begin{align}
		\1M^*_{\pi} =\Tuple{\*U = \{U_D, U_B\}, \*V_{\pi} = \{X', X, Y\}, \2F_{\pi},  P = P(U_D, U_B)}
	\end{align}
	The structural functions in $\2F_{\pi}$ are defined as
	\begin{align}
		\2F_{\pi} = \begin{cases}
			            X' \gets U_D \oplus U_B \\
			            X \gets \neg X'    \\
			            Y \gets f_Y(X, U_D, U_B)  \label{eq:_7_2_mab1}
		            \end{cases}
	\end{align}
	\begin{table}[t]
		\centering
		\begin{tabular}{@{}c|rr|rr@{}} \toprule
			        & \multicolumn{2}{c|}{$X'=0$} & \multicolumn{2}{c}{$X'=1$}                       \\ \cmidrule(r){2-3} \cmidrule(l){4-5}
			        & $X$ = 0                   & $X$ = 1                  & $X$ = 0 & $X$ = 1 \\ \midrule
			$Y = 0$ & 0                          & 0.275                     & 0.275    & 0        \\
			$Y = 1$ & 0                          & 0.225                     & 0.225    & 0        \\ \bottomrule
		\end{tabular}
		\caption{The joint distribution $P(X', X, Y)$ evaluated in the submodel $\M^*_{\pi}$ induced by a counterfactual policy $\pi \triangleq X \gets \neg X'$.}
		\label{tab:_7_2_ett}
	\end{table}
	Table~\ref{tab:_7_2_ett} shows the detailed parametrization of the joint distribution $P(X, X', Y)$ evaluated in the counterfactual submodel $\M^*_{\pi}$ described in Example~\ref{exp:_7_2_mab1}. The expected reward of $Y$ in submodel $\M^*_{\pi}$ is given by
	\begin{align}
		\invE{Y}{\pi} & = P\Parens{X' = 0, X = 0, Y = 1} + P\Parens{X' = 1, X = 0, Y = 1} \\
		                               & + P\Parens{X' = 0, X = 1, Y = 1} + P\Parens{X' = 1, X = 1, Y = 1}
	\end{align}
	Computing the above equation gives $\invE{Y}{\pi} = 0.45$,  which outperforms the expected reward of atomic intervention $\invE{Y}{x} = 0.15$ for every arm $x = 0, 1$. \hfill $\blacksquare$
\end{example}
More generally, note that the counterfactual policy space $\Pi_{\textsc{ctf}}$ contains the experimental policy $\Pi_{\textsc{exp}} = \Braces{\Tuple{X, \emptyset }}$ in MAB environments. For every experimental policy $\pi(X)$, one could simulate it using a counterfactual $\pi(X \mid X)$ by selecting realized action $\doo(X \gets x)$ regardless of natural predilection $X' = x'$, i.e., $\pi(x) = \pi(x \mid x')$, $\forall x, x'$. 
In this case, the expected reward in Eq.~\ref{eq:_7_1_reward} could be further written as:
\begin{align}
	\invE{Y}{\pi} & = \sum_{x} \pi(x)  \sum_{x'} \E\Brackets{Y_{X \gets x} \mid X = x'} P(x') \\
	              & =\sum_{x} \pi(x) \E[Y_{x}]
\end{align}
By the definition of potential outcomes (Def.~\ref{def:_2_2_po}) and interventional distributions (Def.~\ref{def:_2_2_inv_dist}), the counterfactual quantity $\E[Y_x] = \invE{Y}{x}$. 
The above equation thus coincides with the expected reward of an experimental policy $\pi(x)$. 
As shown next, an optimal counterfactual policy consistently dominates the best possible experimental policy in terms of performance.
\begin{theorem}[Counterfactual dominates Interventional Policies (MAB)]\label{thm:_7_1_ctf}
	For an MAB environment $\1M^*$, let policy spaces $\Pi_{\textsc{ctf}} = \Braces{\Tuple{X, \{X\}}}$ and $\Pi_{\textsc{exp}} = \Braces{\Tuple{X, \emptyset }}$. Then, an optimal counterfactual policy is never worse than an optimal interventional policy, namely, 
	\begin{align}
\argmax_{\pi \in \Pi_{\textsc{ctf}}} \invE{Y}{\pi} \geq  \argmax_{\pi \in \Pi_{\textsc{exp}}} \invE{Y}{\pi}
	\end{align} \hfill $\blacksquare$
\end{theorem}
A natural question at this point is when the equality in the equation holds, and the standard interventional agent is able to achieve the optimal performance of an counterfactual agent. When the NUC condition (Def.~\ref{def:_4_1_nuc}) holds in the underlying MAB environment, there is no unobserved confounder affecting the action $X$ and the reward $Y$, simultaneously. 
This implies that the agent's intended action $X$ is independent of the potential outcome $Y_x$ induced by intervention $\doo(x)$, namely, \footnote{This independence relationship is also referred to as \emph{ignorability} in the literature \citep{rosenbaum1983central}.}
\begin{align}
	(X \ci Y_x)
\end{align}
When the above independence relationship holds, Eq.~\ref{eq:_7_1_reward} could be further written as:
\begin{align}
		\E \Brackets{Y ; \1M_{\pi}} & =\sum_{x} \E \Brackets{Y_{X \gets x}} \sum_{x'} \pi \left (x \mid x' \right) P\left(x' \right)\\
		& =\sum_{x} \E \Brackets{Y_{X \gets x}} P_{\pi}(x)
\end{align}
In the last step, probabilities $P_{\pi}(x) = \sum_{x} \pi \left (x \mid x' \right) P\left(x' \right)$ are obtained by marginalizing over the domain of the intended action $X$. An agent could thus simulate the performance of the counterfactual policy $\pi(x'|x)$ using an experimental policy $\pi(x) = P_{\pi}(x)$. 
In other words, the optimal performance of counterfactual and experimental policies coincide whenever the NUC holds.

\subsubsection{Markov Decision Process with Unobserved Confounders}\label{sec:_7_1_1}
The remainder of this section expands on the concept of counterfactual policies to a more general sequential decision-making setting where the agent must decide on the values of a sequence of actions. 
Our discussion will focus on a canonical family of environments that extend Markov decision processes \citep{puterman1994markov} through the language of structural causality. 
\begin{definition}[MDP Environment]\label{def:_7_3_mdp}
Consider an SCM describing an MDP environment
\begin{align}
\1M = \Tuple{\*U, \*V, \2F, P(\*U)}, 
\end{align}
where for a decision horizon $H \in \3N^+$,\footnote{The decision horizon $H$ could be finitely large, i.e. $H = \infty$. The model $\1M$ is called an infinite-horizon MDP.}
	\begin{itemize}
		\item $\*U = \{U_1, \dots, U_H \}$ is a sequence of exogenous variables $U_i$;
		\item $\*V = \{\*S, \*X, \*Y\}$ is a set of endogenous variables consisting of a sequence of states $\*S = \{S_1, \dots, S_H \}$, actions $\*X = \{X_1, \dots, X_H \}$, and rewards $\*Y = \{Y_1, \dots, Y_H\}$;
		\item $\2F$ is a set of functions determining values of $\*S, \*X, \*Y$ such that for every $i = 1, \dots, H$, \footnote{With a slight abuse of notation, we denote by $S_{i - 1} = \emptyset, X_{i - 1} = \emptyset$ if $i = 1$, i.e., the initial state $S_1 \gets f_S(U_1)$.}
		      \begin{align}
			      \2F = \begin{cases}
				            S_i \gets f_S(S_{i-1}, X_{i-1}, U_i) \\
				            X_i \gets f_X(S_{i}, U_i)            \\
				            Y_i \gets f_Y(S_i, X_i, U_i)
			            \end{cases}
		      \end{align}
		\item $P$ is a joint distribution over $\*U$ such that $P(\*U) = \prod_{i = 1}^{H} P(U_i)$ and $U_1, \dots, U_H$ are i.i.d. variables drawn over a domain $\D(U)$, i.e., $P(U_1) = \dots = P(U_H)$.\footnote{Compared with dynamic treatment regimes \citep{murphy2001dtr}, MDPs explicitly encode the \emph{locality} in both the underlying causal mechanisms and exogenous noises affecting states, actions and rewards at every time step. This structural constraint manifests in the Markov property in system dynamics, as discussed in Sec.~\ref{sec:_3_3}.}
	\end{itemize} \hfill $\blacksquare$
\end{definition}
Def.~\ref{def:_7_3_mdp} describes a generalized family of environments similar to MDPs, where the NUC condition does not hold and unobserved confounders are not excluded a priori. 
Consequently, this family of environments is referred to in the literature as MDP with unobserved confounders (MDPUCs) \citep{zhang2020loop}. 
In an MDP environment, the Markov property holds for both the observational distribution $P(\*S, \*X, \*Y)$ and interventional distribution $\inv{\*S, \*X, \*Y}{\pi}$ induced by a policy $\pi = \Parens{\pi_1(X_1 \mid S_1), \dots, \pi_H(X_H \mid S_H)}$. For every time step $i = 1, \dots, H$, all the future state $\bar{\*S}_{i+1:H}$, actions $\bar{\*X}_{i:H}$ and rewards $\bar{\*Y}_{i:H}$ are independent of the history $\bar{\*S}_{1:i-1}, \bar{\*X}_{1:i-1}, \bar{\*Y}_{1:i-1}$ given the current state $S_i$, namely: 
\begin{align}
	\Parens{\bar{\*S}_{i+1:H}, \bar{\*X}_{i:H}, \bar{\*Y}_{i:H} \ci \bar{\*S}_{1:i-1}, \bar{\*X}_{1:i-1}, \bar{\*Y}_{1:i-1} \mid S_i}
\end{align}
The above independence relationship could be read from the causal diagram of the MDP environment, shown in Fig.~\ref{fig:_7_1_1_mdp_a}, following the $d$-separation rules (Def.~\ref{def:_2_3_dsep}). However, due to the presence of unobserved confounders, the transition probabilities and conditional reward in the observational and interventional distributions do not necessarily coincide, i.e. for every step $i = 1, \dots, H$,
\begin{align}
	&P \Parens{s_{i+1} \mid s_i, x_i} \neq \inv{s_{i+1} \mid s_i}{x_i}\\
	&\E \Brackets{Y_i \mid s_i, x_i} \neq \invE{Y_i \mid s_i}{x_i}
\end{align}
Recall the more detailed discussion of the Markov property and the NUC assumption in Sec.~\ref{sec:_3_3}. For instance, Example~\ref{exp:_2_1_mdp} shows an MDP environment concerning the inventory management of a retail store; potential unobserved confounders include human errors of the store manager, uncertainties in the customers' demands, and monetary values of the goods. Both its observational and interventional distributions can be compactly represented using finite-state automata. However, detailed parameters of these automata differ significantly due to the presence of unobserved confounders; detailed computation is provided in Examples~\ref{exp:_2_2_obs-mdp} and \ref{exp:_2_2_inv-mdp}.
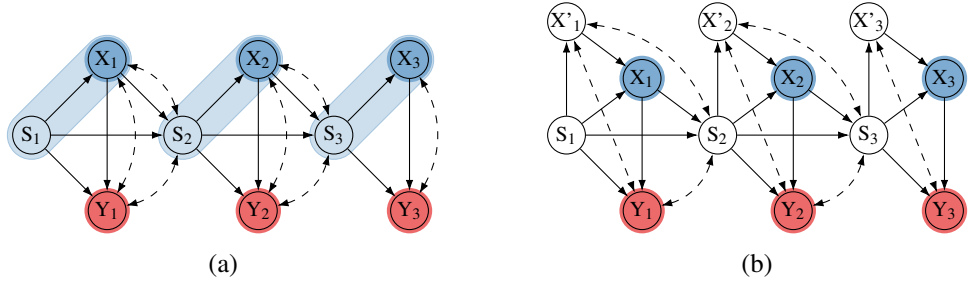
\begin{figure}[t]
	\hfill
	\begin{subfigure}{0.38\linewidth}\centering
		\begin{tikzpicture}
			\def\outerr{3.2}
			\def\innerr{3}
			\node[vertex] (S1) at (-1, -1) {S\textsubscript{1}};
			\node[vertex, opacity=0] (U) at (2, 0.5) {};
			\node[vertex] (X1) at (0, 0) {X\textsubscript{1}};
			\node[vertex] (Y1) at (0, -2) {Y\textsubscript{1}};
			\node[vertex] (S2) at (1, -1) {S\textsubscript{2}};
			\node[vertex] (X2) at (2, 0) {X\textsubscript{2}};
			\node[vertex] (Y2) at (2, -2) {Y\textsubscript{2}};
			\node[vertex] (S3) at (3, -1) {S\textsubscript{3}};
			\node[vertex] (X3) at (4, 0) {X\textsubscript{3}};
			\node[vertex] (Y3) at (4, -2) {Y\textsubscript{3}};

			\draw[dir] (S1) to (S2);
			\draw[dir] (S1) to (X1);
			\draw[dir] (S1) to (Y1);
			\draw[dir] (X1) to (Y1);
			\draw[dir] (X1) to (S2);

			\draw[bidir] (X1) to [bend left = 30] (S2);
			\draw[bidir] (X1) to [bend left = 30] (Y1);
			\draw[bidir] (Y1) to [bend right = 30] (S2);

			\draw[dir] (S2) to (S3);
			\draw[dir] (S2) to (X2);
			\draw[dir] (S2) to (Y2);
			\draw[dir] (X2) to (Y2);
			\draw[dir] (X2) to (S3);

			\draw[bidir] (X2) to [bend left = 30] (S3);
			\draw[bidir] (X2) to  [bend left = 30] (Y2);
			\draw[bidir] (Y2) to [bend right = 30] (S3);

			\draw[dir] (S3) to (X3);
			\draw[dir] (S3) to (Y3);
			\draw[dir] (X3) to (Y3);

			\draw[bidir] (X3) to [bend left = 30] (Y3);

			\begin{pgfonlayer}{back}
				\draw[fill=betterblue!25, draw = betterblue!45] \convexpath{X1, S1}{\outerr mm};
				\draw[fill=betterblue!25, draw = betterblue!45] \convexpath{X2, S2}{\outerr mm};
				\draw[fill=betterblue!25, draw = betterblue!45] \convexpath{X3, S3}{\outerr mm};
				\node[circle,fill=betterblue!65,draw=none,minimum size=2*\innerr mm] at (X1) {};
				\node[circle,fill=betterblue!65,draw=none,minimum size=2*\innerr mm] at (X2) {};
				\node[circle,fill=betterblue!65,draw=none,minimum size=2*\innerr mm] at (X3) {};
				\node[circle,fill=betterred!65,draw=none,minimum size=2*\innerr mm] at (Y1) {};
				\node[circle,fill=betterred!65,draw=none,minimum size=2*\innerr mm] at (Y2) {};
				\node[circle,fill=betterred!65,draw=none,minimum size=2*\innerr mm] at (Y3) {};
			\end{pgfonlayer}
		\end{tikzpicture}
		\caption{}
		\label{fig:_7_1_1_mdp_a}
	\end{subfigure}\hfill
	\begin{subfigure}{0.4\linewidth}\centering
		\begin{tikzpicture}
			\def\outerr{3.2}
			\def\innerr{3}
			\node[vertex] (S1) at (-1, -1) {S\textsubscript{1}};
			\node[vertex] (X1) at (-1, 0.5) {X'\textsubscript{1}};
			\node[vertex] (Xp1) at (0, -0.25) {X\textsubscript{1}};
			\node[vertex] (Y1) at (0, -2) {Y\textsubscript{1}};
			\node[vertex] (S2) at (1, -1) {S\textsubscript{2}};
			\node[vertex] (X2) at (1, 0.5) {X'\textsubscript{2}};
			\node[vertex] (Xp2) at (2, -0.25) {X\textsubscript{2}};
			\node[vertex] (Y2) at (2, -2) {Y\textsubscript{2}};
			\node[vertex] (S3) at (3, -1) {S\textsubscript{3}};
			\node[vertex] (X3) at (3, 0.5) {X'\textsubscript{3}};
			\node[vertex] (Xp3) at (4, -0.25) {X\textsubscript{3}};
			\node[vertex] (Y3) at (4, -2) {Y\textsubscript{3}};

			\draw[dir] (S1) to (S2);
			\draw[dir] (S1) to (X1);
			\draw[dir] (S1) to (Y1);
			\draw[dir] (S1) to (Xp1);
			\draw[dir] (X1) to (Xp1);
			\draw[dir] (Xp1) to (Y1);
			\draw[dir] (Xp1) to (S2);

			\draw[bidir] (X1) to [bend left = 30] (S2);
			\draw[bidir] (X1) to (Y1);
			\draw[bidir] (Y1) to [bend right = 30] (S2);

			\draw[dir] (S2) to (S3);
			\draw[dir] (S2) to (X2);
			\draw[dir] (S2) to (Y2);
			\draw[dir] (S2) to (Xp2);
			\draw[dir] (X2) to (Xp2);
			\draw[dir] (Xp2) to (Y2);
			\draw[dir] (Xp2) to (S3);

			\draw[bidir] (X2) to [bend left = 30] (S3);
			\draw[bidir] (X2) to (Y2);
			\draw[bidir] (Y2) to [bend right = 30] (S3);

			\draw[dir] (S3) to (X3);
			\draw[dir] (S3) to (Y3);
			\draw[dir] (S3) to (Xp3);
			\draw[dir] (X3) to (Xp3);
			\draw[dir] (Xp3) to (Y3);

			\draw[bidir] (X3) to (Y3);

			\begin{pgfonlayer}{back}
				\node[circle,fill=betterblue!65,draw=none,minimum size=2*\innerr mm] at (Xp1) {};
				\node[circle,fill=betterblue!65,draw=none,minimum size=2*\innerr mm] at (Xp2) {};
				\node[circle,fill=betterblue!65,draw=none,minimum size=2*\innerr mm] at (Xp3) {};
				\node[circle,fill=betterred!65,draw=none,minimum size=2*\innerr mm] at (Y1) {};
				\node[circle,fill=betterred!65,draw=none,minimum size=2*\innerr mm] at (Y2) {};
				\node[circle,fill=betterred!65,draw=none,minimum size=2*\innerr mm] at (Y3) {};
			\end{pgfonlayer}
		\end{tikzpicture}
		\caption{}
		\label{fig:_7_1_1_mdp_b}
	\end{subfigure}\hfill\null
	\caption{Causal diagrams for the MDP environment and its submodel induced by a counterfactual policy $\pi = \Parens{\pi_1(X'_1 \mid S_1, X_1), \dots, \pi_H(X'_H \mid S_H, X_H)}$.}
	\label{fig:_7_1_1_mdp}
\end{figure}

We next formalize the concept of counterfactual policies in MDP environments. 
Similar to MABs, an agent following a counterfactual policy can be thought of as selecting the values of every action $X_i \in \*X$ based on its original intended action. 
However, unlike in the previous MAB settings, the agent will also consider observed values of the current $S_i$ before taking action $X_i$. 
\begin{definition}[Counterfactual Policy - MDP]\label{def:_7_3_ctf_space}
	For an MDP environment $\1M^*$, a counterfactual policy space $\Pi_{\textsc{ctf}}$ is a collection of policies
\begin{align}
\pi = \Parens{\pi_1(X_1 \mid S_1, X'_1),  \dots, \pi_H(X_H \mid S_H, X'_H)},		
\end{align}
where every decision rule $\pi_i(X_i \mid S_i, X'_i)$ is a function mapping from the domain of state $S_i$ and intended action $X'_i$ to the space of probability distribution over the domain of realized action $X_i$. Henceforth, we will consistently denote such a policy space by $\Pi_{\textsc{ctf}} = \Braces{\Tuple{X_i, \{S_i, X'_i\}}}_{i = 1}^{H}$. \hfill $\blacksquare$
\end{definition}
Similar to MAB environments, an agent interacting an MDP environment $\1M$ following a counterfactual policy $\pi$ leads to a submodel $\1M_{\pi}$ with additional intended actions $X_i'$ mediating between every realized action $X_i$ and its direct parents, including the current state $S_{i}$ and the unobserved confounder $U_i$. Formally,
\begin{definition}[Counterfactual Submodel, MDP]\label{def:_7_3_submodel}
	Let $\1M = \Tuple{\*U, \*V, \2F, P(\*U)}$ be an MDP environment, and $\pi = \Parens{\pi_1(X_1 \mid S_1, X'_1), \dots, \pi_H(X_H \mid S_H, X'_H)}$ be a counterfactual policy over actions $X_1, X_2, \dots$. 
	A submodel $\1M_{\pi}$ of $\1M$ is an SCM
	\begin{align}
		\1M_{\pi} = \Tuple{\*U, \*V_{\pi} = \Braces{\*S, \*X', \*X, \*Y}, \2F_{\pi},  P = P(\*U)}, 
	\end{align}
	where $\*X = \{X_1, \dots, X_H\}$ is a sequence of realized actions; $\2F_{\pi}$ is a set of structural functions defined as
	\begin{align}
		\2F_{\pi} = \begin{cases}
			            S_i \gets f_S(S_{i-1}, X_{i-1}, U_i) \\
			            X'_i \gets f_X(S_{i}, U_i)            \\
			            X_i \sim \pi_i(X_i \mid S_i, X'_i)    \\
			            Y_i \gets f_Y(S_i, X_i, U_i)
		            \end{cases} \label{eq:_7_3_submodel}
	\end{align} \hfill $\blacksquare$
\end{definition}
The causal diagram  in Fig.~\ref{fig:_7_1_1_mdp_b} is associated with the submodel $\1M_{\pi}$ induced by an MDP environment and a counterfactual policy $\pi(X_i, \mid S_i, X'_i)$. 
Formally, we define $\inv{\*S, \*X', \*X, \*Y}{\pi}$ of a counterfactual policy $\pi(X_i, \mid S_i, X'_i)$ as the joint distribution over endogenous variables $\*S, \*X', \*X, \*Y$ in submodel $\1M_{\pi}$. 
One could see by inspection that the data-generating mechanisms in Fig.~\ref{fig:_7_1_1_mdp_b} define a Markov chain \citep{puterman1994markov}. 
For every stage of intervention $i = 1, 2, \dots$, the state $S_i$ and intended action $X'_i$ satisfy the Markov property with regard to past state and actions' history. More specifically, the following independent relationships hold in the counterfactual submodel $\1M_{\pi}$,
\begin{align}
	P\Parens{S_{i+1}, X'_{i+1} \mid \bar{\*{S}}_{1:i}, \bar{\*{X}'}_{1:i}, \bar{\*{X}}_{1:i} ; \1M_{\pi}} & = P\Parens{S_{i+1}, X'_{i+1} \mid S_{i}, X'_{i}, X_i; \1M_{\pi}} \label{eq:_7_3_markov_s}\\
	\E \Brackets{Y_i \mid \bar{\*{S}}_{1:i}, \bar{\*{X}'}_{1:i}, \bar{\*{X}}_{1:i}; \1M_{\pi}}           & = \E \Brackets{Y_i \mid S_{i}, X'_{i}, X_i; \1M_{\pi}} \label{eq:_7_3_markov_y}
\end{align}
The following example demonstrates the Markov property in a counterfactual MDP submodel.
\begin{table}[t]
			\centering
			\renewcommand{\arraystretch}{1.25}
			\begin{tabular}{|ccccc|c|ccccc|c|}
				$S_{i+1}$ & $X_{i+1}$ & $S_i$ & $X'_i$ & $X_i$ & $P$ &$S_{i+1}$ & $X_{i+1}$ & $S_i$ & $X'_i$ & $X_i$ & $P$ \\
				\hline
				\hline
				0     & 0     & 0 & 0     &0     & 0.09 &1     & 0     & 0 & 0     &0     & 0.09 \\
				0     & 0     & 0 & 0     &1      &0.01 &1     & 0     & 0 & 0     &1      & 0.81           \\
				0     & 0     & 0 & 1     &0     &0.01 &1     & 0     & 0 & 1     &0     &0.81 \\
				0     & 0     & 0 & 1     &1      &0.09 &1     & 0     & 0 & 1     &1      & 0.09          \\
				0     & 0     & 1 & 0     &0     & 0.09 &1     & 0     & 1 & 0     &0     &0.09 \\
				0     &0     & 1 & 0     &1      & 0.09 &1     & 0     & 1 & 0     &1      &0.09          \\
				0     & 0     & 1 & 1     &0     & 0.01 &1     & 0     & 1 & 1     &0     &0.81 \\
				0     & 0     & 1 & 1     &1      & 0.01 &1     & 0     & 1 & 1     &1      &0.81          \\
				0     & 1     & 0 & 0     &0     & 0.09 &1     & 1     & 0 & 0     &0     &0.01 \\
				0     & 1     & 0 & 0     &1      & 0.01 &1     & 1     & 0 & 0     &1      & 0.09         \\
				0     & 1     & 0 & 1     &0     &0.01 &1     & 1     & 0 & 1     &0     &0.09 \\
				0     & 1     & 0 & 1     &1      &0.09 &1     & 1     & 0 & 1     &1      & 0.01          \\
				0     & 1     & 1 & 0     &0     &0.09 &1     & 1     & 1 & 0     &0     &0.01 \\
				0     & 1     & 1 & 0     &1      &0.09 &1     & 1     & 1 & 0     &1      & 0.01          \\
				0     & 1     & 1 & 1     &0     &0.01 &1     & 1     & 1 & 1     &0     &0.09 \\
				0     & 1     & 1 & 1     &1      &0.01 &1    & 1     & 1 & 1     &1      &0.09        \\
			\end{tabular}
		\caption{Evaluation of the counterfactual transition distribution $P(S_{i+1_{x_i}}, X_{i+1_{x_i} }\mid S_i, X_i = x'_i)$ evaluated in the MDP environment of Example~\ref{exp:_2_1_mdp}.}
		\label{tab:_7_3_mdp_s}
\end{table}
The following example demonstrates counterfactual policies in MDP environments.
\begin{example}[Autonomous vs. Semi-autonomous Systems]\label{exp:_7_1_1_mdp1}
	Consider the MDP environment $\1M^*$ described in Eq.~\ref{eq:_2_1_mdp}, where the decision horizon $H = \infty$. Recall that the experimental policy space $\Pi_{\textsc{exp}} = \Braces{\Tuple{X_i, \{S_i\}}}_{i = 1}^{\infty}$ contains a collection of policies $\pi = \Parens{\pi_1(X_1 \mid S_1), \pi_2(X_2 \mid S_2), \dots}$. Every decision rule $\pi_i(X_i \mid S_i)$ is a probability distribution mapping from state $S_i$ to action $X_i$. Operationally, the experimental decision model $\Tuple{\1M^*, \Pi_{\textsc{exp}}, \1R}$ defines an autonomous inventory management system that determines whether to refill $X_i$ based on the current inventory size $S_i$. Note that the manager's intended decision $X_i$ is not accounted in the system's decision-making process, and can thus be discarded.
	
	We now consider the counterfactual policy space $\Pi_{\textsc{ctf}} = \Braces{\Tuple{X_i, \{S_i, X_i\}}}_{i = 1}^{\infty}$. Every counterfactual policy $\pi \in \Pi_{\textsc{ctf}}$ is a sequence of decision rules $\Parens{\pi_1(X'_1 \mid S_1, X_1), \pi_2(X'_2 \mid S_2, X_2), \dots}$. Operationally, the counterfactual decision model $\Tuple{\1M^*, \Pi_{\textsc{ctf}}, \1R}$ defines an inventory management system that repeatedly calibrates the manager's intended action $X_i$ based on the observed state $S_i$. Note that this decision-making system is semi-autonomous since it proactively accounts for the manager's decision. Compared with the autonomous system described above, the counterfactual intervention does not entirely replace the human behavioral policy operating in the environment. \footnote{The connection between the counterfactual intervention and semi-autonomous systems was first formalized and explored in \citep{zhang2020loop}.} 
	
 More specifically, let a counterfactual policy $\pi = \Parens{\pi_1(X_1 \mid X_1, S_1), \pi_2(X_2 \mid X_2, S_2), \dots}$ such that for every $i = 1, 2, \dots$, the decision rule $\pi_i$ is defined as, 
	\begin{align}
		\pi_i \triangleq X_i \gets S_i \oplus X'_i \oplus U_{i, 4}
	\end{align}
	where $U_{i, 4}$ is a new independent noise uniformly drawn over $\{0, 1\}$. Performing counterfactual intervention $\ctf(\pi)$ leads to a submodel $\1M^*_{\pi}$ described as a tuple
	\begin{align}
	\M^*_{\pi} = \left \langle \*U = \{U_{i,1}, \dots, U_{i,4}\}, \*V_{\pi} =  \{X'_i, Y_i, S_i, X_i\}, \2F_{\pi}, P(\*U) \right \rangle_{i = 1, 2, \dots}, 
	\end{align}
	The structural functions $\2F_{\pi}$ are defined as, for every $i = 1, 2, \dots$,
 \begin{align}
    \2F_i = \begin{cases}
      S_i \gets \left (S_{i-1} \vee X'_{i-1} \right) \oplus U_{i-1, 1} \oplus U_{i-1,2},\\
      X'_i \gets S_i \oplus U_{i, 1} \\
      X_i \gets S_i \oplus X'_i \oplus U_{i, 4}\\
      Y_i \gets S_i \oplus X_i \oplus U_{i, 1} \oplus U_{i, 3}
    \end{cases} \label{eq:_7_1_mdp}
  \end{align}
  Note that the structural function $X_i \gets S_i \oplus U_{i, 1}$. Given values of observed state $S_i = s_i$ and action $X_i = x_i$ in submodel $\1M^*_{\pi}$, one could infer values of the unobserved confounder $U_{i, 1}$ as
		\begin{align}
		U_{i, 1} = x_i \oplus s_i \label{eq:_7_3_mdp2_1}
	\end{align}
	Since the next state $S_{i+1} \gets (S_i \vee X_i) \oplus U_{i, 1} \oplus U_{i, 3}$, given the current state $S_i = s_i$, intended action $X_i = x_i$, and realized action $X'_i = x'_i$, event $S_{i+1} = s_{i+1}$ implies the following 
	\begin{align}
		U_{i, 3} &= s_{i+1} \oplus (s_i \vee x'_i) \oplus U_{i, 1} \\
		&= s_{i+1} \oplus (s_i \vee x'_i) \oplus x_i \oplus s_i \label{eq:_7_3_mdp2_2}
	\end{align}
	The last step follows from Eq.~\ref{eq:_7_3_mdp2_1}. Evaluating the transition distribution on the next state $S_{i+1}$ and next intended action $X_{i+1}$ given the current state $S_i$, intended action $X_i$, and realized action $X'_i$ in submodel $\1M^*_{\pi}$ gives
	\begin{align}
		&P\Parens{S_{i+1} = s_{i+1}, X_{i+1} = x_{i+1} \mid S_i = s_i, X_i = x_i, X'_i =  x'_i} \\
		&= P\Parens{U_{i, 3} =  s_{i+1} \oplus (s_i \vee x'_i) \oplus x_i \oplus s_i, X_{i+1} = x_{i+1} \mid S_i = s_i, X_i = x_i, X'_i =  x'_i}\\
		&= P\Parens{U_{i, 3} =  s_{i+1} \oplus (s_i \vee x'_i) \oplus x_i \oplus s_i,  U_{i+1, 1} = x_{i+1} \oplus s_{i+1} } \label{eq:_7_3_mdp2_s}
	\end{align} 
	The first step follows from Eq.~\ref{eq:_7_3_mdp2_2}; the second step follows from the equation $X'_{i+1} \gets S_{i+1} \oplus U_{i+1, 1}$. Moreover, given values of the current $S_{i}, X'_i, X_i$, the past history $S_1, \dots, S_{i-1}$, $X'_1, \dots, X'_{i-1}$, and $X_1, \dots, X_{i-1}$ are independent from the exogenous variables $U_{i, 3}, U_{i+1, 1}$, i.e., the Markov property holds. 
	We compute the detailed parametrization of the conditional transition distribution  $P\Parens{S_{i+1}, X'_{i+1} \mid S_i, X'_i, X_i}$ and provide them in Table~\ref{tab:_7_3_mdp_s}. 
	
\begin{table}[t]
			\centering
			\renewcommand{\arraystretch}{1.25}
			\begin{tabular}{|ccc|c|ccc|c|}
				 $S_i$ & $X_i$ & $X'_i$ & $\E$ & $S_i$ & $X_i$ & $X'_i$ & $\E$ \\
				\hline
				\hline
				0 & 0     &0     &0.1 &1 & 0     &0     &0.1 \\
				0 & 0     &1     &0.9 &1 & 0     &1     &0.9 \\
				0 & 1     &0     &0.9 &1 & 1     &0     &0.9 \\
				0 & 1     &1     &0.1 &1 & 1     &1     &0.1 \\
			\end{tabular}
		\caption{Evaluation of the counterfactual expected reward $\E[Y_{i_{x_i}}\mid S_i, X_i = x'_i]$ evaluated in the MDP environment of Example~\ref{exp:_2_1_mdp}.}
		\label{tab:_7_3_mdp_y}
\end{table}
	 Similarly, note that values of the reward signal $Y_i \gets S_i \oplus X_i \oplus U_{i, 1} \oplus U_{i, 2}$. Given state $S_i = s_i$, intended action $X'_i = x'_i$ and realized action $X_i = x_i$, event $Y_{i} = y_i$ implies the following
	\begin{align}
		U_{i, 2} &= y_i \oplus s_i \oplus x_i \oplus U_{i, 1}\\
		&= y_i \oplus s_i \oplus x_i \oplus s_i \oplus x'_i\\
		&= y_i \oplus x_i \oplus x'_i
	\end{align}
	The second step follows from Eq.~\ref{eq:_7_3_mdp2_1}. Evaluating the expected reward $Y_i$ conditioning on the state $S_i$, intended action $X'_i$ and realized action $X_i$ in submodel $\1M^*_{\pi}$ gives
	\begin{align}
		&\E \Brackets{Y_{i} \mid S_i = s_{i}, X'_{i} = x'_i, X_i = x_i} \\
		&= P\Parens{Y_{i} = 1 \oplus x_i \oplus x'_i\mid S_i = s_{i}, X'_{i} = x'_i, X_i = x_i}\\
		&= P \Parens{U_{i, 2} = 1 \oplus x_i \oplus x'_i}  \label{eq:_7_3_mdp2_y}
	\end{align}
	Detailed parametrization of $\E \Brackets{Y_{i} \mid s_{i}, x'_i, x_i} $ are computed and provided in Table~\ref{tab:_7_3_mdp_y}. \hfill $\blacksquare$
\end{example}
More importantly, it is possible to show that the above conditional distributions in submodel coincide with ETTs of action $X_i$ on the reward signal $Y_i$ and next state $S_{i+1}$ and action $X_{i+1}$, provided with the current state $S_i$. Such ETTs remain invariant across every stage $i = 1, \dots, H$, 
\begin{lemma}\label{lem:_7_3_ctf}
	Let $\1M$ be an MDP environment, $\pi(X_i \mid S_i, X'_i)$ be a counterfactual policy over actions $X_1, \dots, X_H$, and $\1M_{\pi}$ be an induced counterfactual submodel of $\1M$. Then, for every $i = 1, \dots, H$, the transition distribution over $S_{i+1}$ and the expected reward over $Y_i$ conditioning on $S_i, X'_i, X_i$ in submodel $\1M_{\pi}$ is equal to
	\begin{align}
		P\Parens{S_{i+1}, X_{i+1} \mid s_i, x'_i, x_i; \1M_{\pi}} & = P\Parens{S_{i+1_{X_i \gets x_i}}, X_{i+1_{X_i \gets x_i}} \mid s_i, X'_i = x'_i; \1M } \label{eq:_7_3_ctf_1}\\
		\E \Brackets{Y_i \mid s_{i}, x'_{i}, x_i; \1M_{\pi}}      & = \E \Brackets{Y_{i_{X_i \gets x_i}} \mid s_{i}, X'_{i} = x'_i; \1M} \label{eq:_7_3_ctf_2}
	\end{align}
	Moreover, the above quantities remain invariant across stage $i = 1, \dots, H$, i.e.,
	\begin{align}
		P\Parens{S_{i+1_{X_i \gets x_i}}, X_{i+1_{X_i \gets x_i}} \mid s_i, X'_i = x'_i} & = \cdots = P\Parens{S_{2_{X_1 \gets x_1}}, X_{2_{X_1 \gets x_1}} \mid s_1, X'_1 = x'_1 } \label{eq:_7_3_ctf_3}\\
		\E \Brackets{Y_{i_{X_i \gets x_i}} \mid s_{i}, X'_{i} = x'_i}                     & = \cdots = \E \Brackets{Y_{1_{X_1 \gets x_1}} \mid s_{1}, X'_{1} = x'_1} \label{eq:_7_3_ctf_4}
	\end{align}\hfill $\blacksquare$
\end{lemma}
Among the above equations, Eqs.~\ref{eq:_7_3_ctf_1} and \ref{eq:_7_3_ctf_2} follow from the definition of counterfactual intervention. Eqs.~\ref{eq:_7_3_ctf_3} and \ref{eq:_7_3_ctf_4} hold since structural functions $f_X, f_Y, f_S$ and the exogenous distribution $P(U_i)$ remain invariant across all stages of interventions $i = 1, \dots, H$. 
\begin{example}\label{exp:_7_3_mdp3}
	Consider the MDP environment $\1M^*$ described in Eq.~\ref{eq:_2_1_mdp}. Given values of state $S_i = s_i, X'_i = x'_i$, one could infer values of the unobserved confounder $U_{i, 1}$ as
	\begin{align}
		U_{i, 1} = x'_i \oplus s_i \label{eq:_7_3_mdp3_1}
	\end{align}
	Given current state and action $S_i = s_i, X'_i = x'_i$, the counterfactual event $S_{i+1_{X_i \gets x_i}} = s_{i+1}$ implies
	\begin{align}
		U_{i, 3} &= s_{i+1} \oplus (S_i \vee X'_i)  \oplus U_{i, 1} \\
		&= s_{i+1} \oplus (s_i \vee x_i) \oplus x'_i \oplus s_i \label{eq:_7_3_mdp3_2}
	\end{align}
	The last step follows from Eq.~\ref{eq:_7_3_mdp3_1}. 
	Evaluating the ETT of action $X_i$ on the next state $S_{i+1}$ and action $X_{i+1}$ conditioning on the current state $S_i$ in the underlying MDP environment $\1M^*$ gives
	\begin{align}
		&P\Parens{S_{i+1_{X_i \gets x_i}} = s_{i+1}, X_{i+1_{X_i \gets x_i}} = x_{i+1} \mid S_i = s_i, X'_i = x'_i} \\
		&= P\Parens{U_{i, 3} =  s_{i+1} \oplus (s_i \vee x_i) \oplus x'_i \oplus s_i, X_{i+1_{X_i \gets x_i}} = x'_{i+1}\mid S_i = s_i, X'_i = x'_i}\\
		&= P\Parens{U_{i, 3} =  s_{i+1} \oplus (s_i \vee x_i) \oplus x'_i \oplus s_i,  U_{i+1, 1} = x'_{i+1} \oplus s_{i+1} }
	\end{align} 
	The above equation coincides with Eq.~\ref{eq:_7_3_mdp2_s}. 
	This means that the counterfactual ETT distribution $P\Parens{S_{i+1_{X_i \gets x_i}}, X_{i+1_{X_i \gets x_i}} \mid s_i, x'_i}$ evaluated in the MDP environment $\1M^*$ is equal to the conditional transition distribution $P\Parens{S_{i+1}, X'_{i+1} \mid s_i, x'_i, x_i}$ evaluated in submodel $\1M^*_{\pi}$. 
	Its detailed parametrizations are provided in Table~\ref{tab:_7_3_mdp_s}. Moreover, this counterfactual distribution remains the same for all decision horizon $i = 1, 2, \dots$ since structural functions $\2F$ and exogenous distribution $P(U_{i, 1}, U_{i, 2}, U_{i, 3})$ are invariant with regard to the horizon $i$.

	Similarly, given state and action $S_i = s_i, X'_i = x'_i$, the potential outcome $Y_{i_{X_i \gets x_i}} = y_i$ implies
	\begin{align}
		U_{i, 2} &= y_i \oplus s_i \oplus x_i \oplus U_{i, 1}\\
		&= y_i \oplus s_i \oplus x_i \oplus s_i \oplus x'_i\\
		&= y_i \oplus x_i \oplus x'_i
	\end{align}
	The second step follows from Eq.~\ref{eq:_7_3_mdp3_1}. Evaluating the ETT of action $X_i$ on the reward signal $Y_i$ conditioning on the current state $S_i$ in the underlying MDP environment $\1M^*$ gives
	\begin{align}
		\E \Brackets{Y_{i_{X_i \gets x_i}} \mid S_i = s_{i}, X'_{i} = x'_i} &= P\Parens{Y_{i_{X_i \gets x_i}} = 1 \oplus x_i \oplus x'_i\mid S_i = s_{i}, X'_{i} = x'_i}\\
		&= P \Parens{U_{i, 2} = 1 \oplus x_i \oplus x'_i}
	\end{align}
	The last step coincides with Eq.~\ref{eq:_7_3_mdp2_y}. This means that the conditional $x$-specific causal effects $\E \Brackets{Y_{i_{X_i \gets x_i}} \mid s_{i}, x'_i}$ evaluated in the MDP environment $\1M^*$ equates to the conditional expected reward $\E[Y_i \mid s_i, x'_i, x_i]$ evaluated in submodel $\1M^*_{\pi}$. 
	Its detailed parametrizations remain invariant for every decision horizon $i = 1, 2, \dots$, and are provided in Table~\ref{tab:_7_3_mdp_y}. \hfill $\blacksquare$
\end{example}
\sloppy Lem.~\ref{lem:_7_3_ctf} permits us to represent the distribution $\inv{\*S, \*X', \*X, \*Y}{\pi}$ induced by a counterfactual policy $\pi = \Parens{\pi_1(X_1 \mid S_1, X'_1), \dots, \pi_H(X_H \mid S_H, X'_H)}$ using a standard MDP (Def.~\ref{def:_3_3_mdp})
\begin{align}
\Tuple{\D(S)\times \D(X), \D(X), \1T_{\text{ctf}}, \1R_{\text{ctf}}}
\end{align}
Here, $\D(S)$ and $\D(X)$ are, respectively, the domain of state $S_i$ and action $X_i$ for every stage $i = 1, \dots, H$. The transitional distribution $\1T_{\text{ctf}}$ and the reward function $\1R_{\text{ctf}}$ are conditional ETTs  evaluated in the underlying MDP environment $\1M^*$ given by, for any $s, s' \in \D(S)$ and any $x, x', x'' \in \D(X)$,
\begin{align}
	 \1T_{\text{ctf}}(s, x, x', s', x'') &= P\Parens{S_{i+1_{X_i \gets x'}} = s', X_{i+1_{X_i \gets x'}} = x'' \mid S_i = s, X_i = x} \label{eq:_7_3_mdp_t} \\
	 \1R_{\text{ctf}}(s, x, x') &= \E \Brackets{Y_{i_{X_i \gets x'}} \mid S_{i} = s, X_{i} = x}  \label{eq:_7_3_mdp_r}
\end{align}
Let $\1R(\*Y) \in \3R$ be a reward function taking reward signal $\*Y = \{Y_1, \dots, Y_H\}$ as input. Our goal is to obtain an optimal counterfactual policy $\pi^* \in \Pi_{\textsc{ctf}}$ maximizing the expected reward over $\invE{\1R(\*Y)}{\pi}$ evaluated in the underlying MDP environment $\1M^*$. The reduction to a standard MDP model $\Tuple{\D(S)\times \D(X), \D(X), \1T_{\text{ctf}}, \1R_{\text{ctf}}}$ allows us to solve for an optimal counterfactual policy $\pi^*$ using standard dynamic programming algorithms \citep{bellman1966dynamic}, provided that the detailed parametrization of the underlying MDP environment $\1M^*$ is available. 

To make this argument more precise, we will consider a discounted cumulative reward function $\1R(\*Y) = \sum_{i = 1}^{\infty} \gamma^{i-1} Y_{i}$ over an infinite horizon $H = \infty$, where the discount rate $\gamma \in (0, 1)$. 
The classic result in planning literature \citep{puterman1994markov} implies that it is sufficient to consider a class of stationary counterfactual policies $\pi = \Parens{\pi_1(X_1 \mid S_1, X_1), \pi_2(X_2 \mid S_2, X_2), \dots}$ such that the decision rule $\pi_i$ remains invariant across the decision horizon $i = 1, 2, \dots$, i.e., $\pi_1 = \pi_2 = \dots$.

The state-action value function $Q_{\pi}: \D(S) \times \D(X) \times \D(X) \to \3R$ for a counterfactual policy $\pi$ evaluated in the MDP environment $\1M^*$ is defined as the expected cumulative reward following policy $\pi$, given the starting state $s$, intended action $x$, and realized action $x'$. Formally,
\begin{align}
	Q_{\pi}(s, x, x') = \E \Brackets{ \sum_{j = 0}^{\infty} \gamma^{j} Y_{i+j} \mid S_i = s, X_i = x, X'_i = x'; \1M^*_{\pi}}
\end{align}
By exploring the Markov property in submodel $\1M^*_{\pi}$ (Eqs.~\ref{eq:_7_3_markov_s} and \ref{eq:_7_3_markov_y}, the above Q-function could be further written as an augmented \emph{Bellman Equation} using the intended action $X$ as an additional side information:
\begin{align}
	&Q_{\pi}(s, x, x') \\
	& = \E \Brackets{  Y_i + \gamma Y_{i+1} + \gamma^2 Y_{i+2} + \cdots \mid S_i = s, X_i = x, X'_i = x'; \1M^*_{\pi}} \\
	& = \E \Brackets{  Y_i + \gamma Q_{\pi}(S_{i+1}, X_{i+1}, X'_{i+1})  \mid S_i = s, X_i = x, X'_i = x'; \1M^*_{\pi}}                                      \\
	& = \1R_{\text{ctf}}(s, x, x') + \gamma \sum_{s', x''} \1T_{\text{ctf}}(s,  x, x', s', x'') \sum_{x'''}\pi_{i+1}(x'''\mid s', x'') Q_{\pi}(s', x'', x''')
\end{align}
The last step follows from the equality relationships in Lem.~\ref{lem:_7_3_ctf}. An optimal counterfactual policy $\pi^*$ is obtainable by recursively computing the Q-function and optimizing the realized action $x'$ for every state $s$ and intended action $x$. The following example demonstrates such a procedure. 
\begin{example}\label{exp:_7_3_mdp}
	\begin{table}[t]
			\centering
			\renewcommand{\arraystretch}{1.25}
			\begin{tabular}{|ccc|c|ccc|c|}
				 $S_i$ & $X_i$ & $X'_i$ & $Q_*$ & $S_i$ & $X_i$ & $X'_i$ & $Q_*$ \\
				\hline
				\hline
				0 & 0     &0     &1 &1 & 0     &0     &1 \\
				0 & 0     &1     &\textbf{9} &1 & 0     &1     &\textbf{9} \\
				0 & 1     &0     &\textbf{9} &1 & 1     &0     &\textbf{9} \\
				0 & 1     &1     &1 &1 & 1     &1     &1 \\
			\end{tabular}
		\caption{Optimal augmented Q-function $Q_*(s, x, x')$ evaluated in the MDP model of Example~\ref{exp:_2_1_mdp}.}
		\label{tab:_7_3_mdp_q}
	\end{table}

	Consider the MDP environment $\1M^*$ defined in Eq.~\ref{eq:_2_1_mdp}. Note that an optimal counterfactual policy $\pi^*$ is such that its induced value function $Q_{\pi^*}(s, x, x') \geq Q_{\pi}(s, x, x')$ for all state $s$, intended action $x$, and realized action $x'$. Optimizing Q-function leads to a counterfactual augmented (ctf-augmented) \emph{Bellman optimality equation} using the intended action $X$ as an additional context, i.e,
	\begin{align}
		Q_{*}(s, x, x') = \1R_{\text{ctf}}(s, x, x') + \gamma \sum_{s', x''} \1T_{\text{ctf}}(s, x, x', s', x'') \max_{x'''} Q_{*}(s', x'', x''') \label{eq:_7_3_mdp_optimal_bellman}
	\end{align}
	The optimal policy $\pi^*$ is given by, for every stage $i = 1,2, \dots$, for any state $s \in \D(S)$,
	\begin{align}
		\pi^*(S_i = s, X_i = x) = \argmax_{x'} Q_*(s, x, x')
	\end{align}
	We compute the optimal augmented Q-function evaluated in the MDP environment $\1M^*$ using the value iteration algorithm \citep{sutton1998reinforcement}. Detailed parametrizations are provided in Table~\ref{tab:_7_3_mdp_q}. The optimal policy $\pi^* = \Parens{\pi^*_i(X_i \mid S_i, X_i)}_{i = 1}^{\infty}$ is given by $\pi^*_i \triangleq X_i \gets \neg X_i$, for every $i = 1, 2, \dots$. 
	
After all, this means that, in this case, the agent should always go against its intended action. 
Evaluating its expected return gives $\invE{\sum_{i=1}^{\infty} \gamma^{i-1}Y_i}{\pi^*} = 9$, which outperforms the best possible experimental policy $\pi_i \triangleq X_i \gets \neg S_i$. Derivations of the best experimental policy are provided in Example~\ref{exp:_3_1_mdp}. \hfill $\blacksquare$
\end{example}
More generally, experimental policies in $\Pi_{\textsc{exp}} = \Braces{\Tuple{X_i, \{S_i\} }}_{i = 1}^{H}$ are contained in the counterfactual policy space $\Pi_{\textsc{ctf}} = \Braces{\Tuple{X_i, \{S_i, X_i\} }}_{i = 1}^{H}$. This means that one could simulate the expected reward of any experimental policy $\pi' = \Parens{\pi'_i(X_i \mid S_i)}_{i = 1}^{H}$ using a counterfactual policy $\pi = \Parens{\pi_i(X_i \mid S_i, X_i)}_{i = 1}^{H}$ such that $\pi'_i(x' \mid s) = \pi_i(x' \mid s, x)$ for all $s, x, x'$. In this case, intended actions $\*X$ do not affect values of reward signals $\*Y$ in submodel $\1M^*_{\pi}$ induced by counterfactual intervention $\ctf(\pi)$. Marginalizing intended actions $\*X$ reduces $\1M_{\pi}$ to the experimental submodel $\1M^*_{\pi'}$ induced by intervention $\doo(\pi')$. The performance of the counterfactual policy $\pi$ and the experimental policy $\pi'$ thus coincides, i.e., $\invE{\1R(\*Y)}{\pi} = \invE{\1R(\*Y)}{\pi'}$. This observation implies that an agent optimizing the environment following counterfactual interventions $\ctf(\pi)$ must perform at least as well as its counterpart following experimental intervention $\doo(\pi)$. Formally,
\begin{theorem}[Counterfactual dominates Interventional policies]\label{thm:_7_3_ctf}
	For an MDP environment $\1M^*$, let $\1R$ be a reward function over reward signals $\*Y = \{Y_1,\dots, Y_H\}$. Let policy spaces $\Pi_{\textsc{ctf}} = \Braces{\Tuple{X_i, \{S_i, X_i\}}}_{i = 1}^{H}$ and $\Pi_{\textsc{exp}} = \Braces{\Tuple{X, \{S_i\} }}_{i = 1}^{H}$. Then, 
\begin{align}
\argmax_{\pi \in \Pi_{\textsc{ctf}}} \invE{\1R(\*Y)}{\pi} \geq  \argmax_{\pi \in \Pi_{\textsc{exp}}} \invE{\1R(\*Y)}{\pi}
\end{align}	 \hfill $\blacksquare$
\end{theorem}
Thm.~\ref{thm:_7_3_ctf} implies that it is preferable to learn an optimal counterfactual policy in the MDP environment when determining values for a sequence of actions, which consistently dominates the best possible experimental policy that does not account for the agent's intended actions $X_i$ for every stage of decision $i = 1, \dots, H$. On the other hand, when the NUC holds in $\1M^*$ (Def.~\ref{def:_4_1_nuc}), conditioning on current state $S_i$ and realized action $X_i$ d-separates the intended action $X_i$ from all the future rewards $Y_i, Y_{i+1}, \dots$ and states $S_{i+1}, S_{i+2}, \dots$. We thus have the following, for any $x_i, s_i, x'_i$,
	\begin{align}
		P\Parens{S_{i+1_{X_i \gets x'_i}}, X_{i+1_{X_i \gets x'_i}} \mid S_i = s_i, X_i = x_i} & =  \inv{S_{i+1} \mid S_i = s_i}{X_i \gets x'_i}\\
		\E \Brackets{Y_{i_{X_i \gets x'_i}} \mid s_{i}, X_{i} = x_i}&= \invE{Y_i \mid S_i = s_i}{X_i \gets x'_i}
	\end{align}
In words, the counterfactual transition probabilities $\1T_{\text{ctf}}$ and reward function $\1R_{\text{ctf}}$ coincide with their experimental counterparts $\1T_{\text{exp}}$ and $\1R_{\text{exp}}$. An agent can thus simulate the performance of an optimal counterfactual policy using an experimental one; observing the agent's intended action provides no value of information to the learning task and could be ignored.

\subsection{Counterfactual Randomization}\label{sec:_7_2}
So far, we have described effective planning algorithms to obtain an optimal counterfactual policies that account for the agent's intended actions in canonical decision environments, such as MABs and MDPs. 
However, how can an optimal counterfactual policy be learned by computing the counterfactual quantities entailed by the underlying, unknown environment? 
To illustrate, consider the MAB environment as an example. 
When the intended action (coming from $f_x$) and the executed action match (i.e., $x' = x$), the counterfactual quantity $\E[Y_{x} \mid x]$ coincides with the observational reward $\E [Y \mid X = x]$, following the composition axiom \citep[Ch.~7.3]{pearl:2k}. 
For the general case when intended and executed actions do not match ($x' \neq x$), the $x$-specific effect $\E[Y_{x} \mid x']$ is not computable from any combination of passive observations and controlled experiments, without the detailed parametrization or additional assumptions of the underlying causal model. 
\footnote{One exception is the binary case, as elaborated in \cite[Sec.~8.2]{pearl:2k}.} 

In settings where the $x$-specific effect is identifiable, a more challenging question arises: How can an agent implement the counterfactual policy in the environment? 
This question stems from the observation that agents may consider various alternatives during the deliberation process and change their opinion about the best course of action. 
Consequently, only the final choice matters, actually representing the agent's natural predilections. 
To illustrate this concept, the diagram in Fig.~\ref{fig:multi-natural-decision} depicts an example of an agent's deliberation process. 
Initially, the agent intends to play $X'=x_1$ but reconsiders, thinking it might be sub-optimal, and decides to switch to $X'=x_2$, where $x_1 \neq x_2$. 
As time passes, the agent may realize that  $X'=x_{t-1}$ was not ideal and switch to an alternative, $X'=x_t$. 
Ultimately, the final decision defines the agent's individuality, irrespective of the path taken to reach it.%
\footnote{Note that whenever the agent pursues an interventional (layer 2) strategy by leveraging Fisherian randomization, the entire deliberation process is bypassed. In this case, the randomization itself determines which action should be executed. This is illustrated in Fig.~\ref{fig:multi-natural-decision}. } 

This challenge calls for novel counterfactual machinery to allow for the counterfactual interaction following layer 3 as discussed in the previous section, in theory. 
Here, we introduce a novel type of randomization for intention-specific groups, namely, interrupt any reasoning agent before they execute their choice, treat this choice as their intention, deliberate, and then act. 
The $x$-specific effect will then be computed in an alternative fashion, based on the idea of intention-specific randomization. This section discusses the algorithmic implementation of this randomization.
	
\usetikzlibrary{fit}
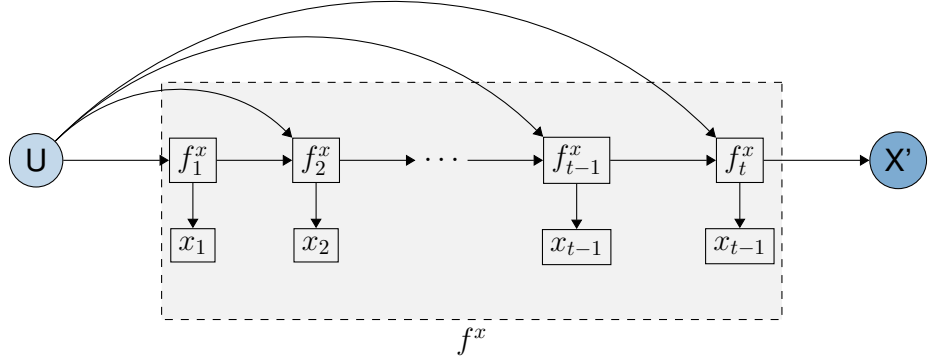
\begin{figure}\centering
\begin{tikzpicture}[
    font=\sffamily\Large,
    >=Triangle,
    node distance=1cm,
    every node/.style={scale=0.8},
    auto
]

  \node[draw, fill=betterblue!30, circle] (U) {U};
 
  \node[draw, rectangle, right=of U, xshift=0.5cm] (f1) {\(f^x_1\)};
  \node[draw, rectangle, below=of f1, yshift=0.5cm] (f1x) {\(x_1\)};
  \node[draw, rectangle, right=of f1] (f2) {\(f^x_2\)};
  \node[draw, rectangle, below=of f2, yshift=0.5cm] (f2x) {\(x_2\)};
  \node[right=of f2] (dots) {\(\cdots\)};
  \node[draw, rectangle, right=of dots] (ft1) {\(f^x_{t-1}\)};
  \node[draw, rectangle, below=of ft1, yshift=0.5cm] (ft1x) {\(x_{t-1}\)};
  \node[draw, rectangle, right=of ft1, xshift=0.5cm] (ft) {\(f^x_t\)};
  \node[draw, rectangle, below=of ft, yshift=0.5cm] (ftx) {\(x_{t-1}\)};
  \node[draw, fill=betterblue!65,  circle, right=of ft, xshift=0.5cm] (X) {X'};

  \begin{pgfonlayer}{back}
    \node[draw, dashed, rectangle, fill=gray!10, minimum height=1cm,  inner sep=32pt, fit=(f1) (ft) (f1x) (ftx), label=below:\(f^x\)] (container) {};
  \end{pgfonlayer}

  \draw[->] (U) -- (f1);
  \draw[->] (f1) -- (f1x);
  \draw[->] (U.45) to [out=45,in=135, looseness=1] (f2.north west);
  \draw[->] (U.45) to [out=45,in=135, looseness=1] (ft1.north west);
  \draw[->] (U.45) to [out=45,in=135, looseness=1] (ft.north west);
  \draw[->] (f1) -- (f2);
  \draw[->] (f2) -- (f2x);
  \draw[->] (f2) -- (dots);
  \draw[->] (dots) -- (ft1);
  \draw[->] (ft1) -- (ft);
  \draw[->] (ft1) -- (ft1x);
  \draw[->] (ft) -- (X);
  \draw[->] (ft) -- (ftx);

\end{tikzpicture}
\caption{Illustration of decision flow, $f_X$, where $U$ is taken as input and the natural predilections $X'$ is returned as output. The process is refined through multiple stages. } \label{fig:multi-natural-decision}
\end{figure}

Alg.~\ref{alg:_7_2_rct} shows the detailed design of randomized controlled trials augmented with counterfactual interventions, which we name \texttt{Ctf-RCT}. More specifically, it takes as input the domain of action $X$ and an integer $N$ indicating the total number of trials. 
For every episode $t$, the agent first perceives its intended action $X^{(t)}$, decided by the behavioral policy $f_X$. 
During the exploration phase (i.e., episode $t \leq N$), the algorithm selects a realized action $X^{(t)}$ from the action space $\1D(X)$ uniformly at random. 
During the exploitation phase (episode $t > N$), it selects a realized action maximizing the empirical estimates of the $x$-specific effect provided with the intended action $X = X^{(t)}$, computed from samples collected from the first $N$ episodes of interventions. Formally, the empirical estimates of the counterfactual quantity $\E \Brackets{Y_{X \gets x} \mid X = x'}$ computed from samples up to episode $t$ are defined as:
\begin{align}
	\hat{\E}^{(t)}\Brackets{Y_{X \gets x} \mid X = x'} = \frac{\sum_{i = 1}^{t} Y^{(i)} \I\Braces{X'^{(i)} = x', X^{(i)} = x}}{N_t\Parens{x', x}} \label{eq:_7_2_empirical}
\end{align}
where $N_t(x, x') = \sum_{i = 1}^t \I\Braces{X^{(i)} = x, X'^{(i)} = x}$ is the total number of occurrence of intercepting intended actions $X^{(i)} = x$ and selecting realized actions $X'^{(i)} = x'$. It follows that Eq.~\ref{eq:_7_2_empirical} is a consistent estimate of the $x$-specific effect of $X$ on $Y$. 
Finally, \texttt{Ctf-RCT} performs an intervention $\doo(X'^{(t)})$ following the selected action throughout episode $t$ and receives subsequent reward $Y^{(t)}$.

\begin{algorithm}[t]
	\caption{Counterfactual Randomized Controlled Trials (\texttt{Ctf-RCT}) in MAB}
	\label{alg:_7_2_rct}
	\setlength{\textfloatsep}{0pt}
	\begin{algorithmic}[1]
		\Require the domain of action $\D(X)$, the total number of trials $N \in \3N$.
		\For{episodes $t = 1, 2, \dots $}
		\State Perceive an intended action $X^{(t)}$ and store it.
		\State Choose a realize action $X^{'(t)}$ as follows.
		\begin{align}
			X'^{(t)} = \begin{dcases}
				           \texttt{Unif}(\D(X))                                                & \mbox{if } t \leq N \\
				           \argmax_{x} \hat{\E}^{(N)}\Brackets{Y_{X \gets x} \mid X = X^{(t)}} & \mbox{if } t > N
			           \end{dcases}
		\end{align}
		\State Perform $\doo(X'^{(t)})$ for episode $t$ and receive reward $Y^{(t)}$.
		\EndFor
	\end{algorithmic}
\end{algorithm}

One could also apply the same principle of counterfactual interventions to adaptive online randomization algorithms such as \texttt{UCB} (Alg.~\ref{alg:_4_2_ucb}) to obtain an optimal counterfactual policy with sublinear regret. Precisely, by applying Hoeffiding's inequalities for every intended action $X = x'$, we define the upper confidence bound over ETT $\E \Brackets{Y_{X \gets x} \mid X = x'}$ computed from samples collected up to episode $t$ as follows, for every pair $x', x \in \D(X)$,
\begin{align}
	\text{UCB}_{t}\left (x, x', \delta \right) = \hat{\E}^{(t)}\Brackets{Y_{X \gets x'} \mid X = x} + \sqrt{\frac{\log(1/\delta)}{2N_t(x, x')}},
\end{align}
where the error probability $\delta \in (0, 1)$ is an arbitrary real value. For every episode $t = 1, 2, \dots$, the algorithm incorporating counterfactual interventions first perceives and intercepts the agent's intended action $X^{(t)}$. It then computes the confidence bounds $\text{UCB}_{t-1}\left (x, X^{(t)}, \delta \right)$ for every arm $x'$ from samples collected up to episode $t - 1$, and picks a realized action $X^{(t)}$ with the highest UCB estimates provided with the intended action $X'^{(t)}$. Finally, the algorithm performs an intervention $\doo(X^{(t)})$ following the selected action throughout episode $t$ and receives subsequent reward $Y^{(t)}$.

The detailed implementation of the counterfactual \texttt{UCB} algorithm, named \texttt{Ctf-UCB}, is summarized in Alg.~\ref{alg:_7_2_ucb}. For every episode $t = 1, 2, \dots$, the error probability $\delta$ is set as a non-increasing function of the total occurrences $N_t\Parens{X^{(t)}} = \sum_{j = 1}^t \I \Braces{X^{(j)} = X^{(t)}}$ of the agent's intended action perceived at episode $t$ from past samples collected so far. In words, the algorithm uses a separate instance of \texttt{UCB} for every intended action $X = x$. Let $K = | \D(X)|$ denote the total number of candidate arms. The cumulative regret of \texttt{Ctf-UCB} after $T$ episodes of interventions is bounded by summing the regrets of each \texttt{UCB} instance for every intended action $X = x$. Specifically, summing regrets in Eq.~\ref{eq:_4_2_ucb4} gives
\begin{align}
	R(T, \1M^*) \leq \sum_{x}  C \sqrt{K \sum_{t = 1}^T \I \Braces{X^{(t)} = x} \log(T)},
\end{align}
where $C$ is a universal constant. Since the square root is a concave function, applying Jensen's inequality allows us to further bound the regret as follows.
\begin{theorem}[Regrets of \texttt{Ctf-UCB} in MABs]\label{thm:_7_2_ucb}
	For an MAB $\langle \1M^*, \Pi, Y \rangle$, let $\Pi$ be a counterfactual policy space $\Set{\Tuple{X, \{X\}}}$, $Y$ be the reward variable with support on $[0, 1]$, and let the domain of action $X$ be $\D(X) = \{1, \dots, K\}$. The regret of \texttt{Ctf-UCB} in SCM $\1M^*$ after $T > 1$ episodes is bounded by
	\begin{align}
		R(T, \1M^*) \leq C K\sqrt{T\log(T)}\label{eq:_7_2_ucb_regret}
	\end{align}
	where $C$ is a universal constant. \hfill $\blacksquare$
\end{theorem}

\begin{algorithm}[t]
	\caption{Counterfactual Upper Confidence Bound (\texttt{Ctf-UCB}) in MAB}
	\label{alg:_7_2_ucb}
	\setlength{\textfloatsep}{0pt}
	\begin{algorithmic}[1]
		\Require the domain of action $\D(X)$.
		\For{episodes $t = 1, 2, \dots $}
		\State Receive an intended action $X^{(t)}$.
		\State Choose an arm $X'^{(t)} = \argmax_{x} \text{UCB}_{t-1}\left (x, X^{(t)}, \delta \right)$ where $\delta = N_t\Parens{X^{(t)}}^{-4}$.
		\State Perform $\doo(X'^{(t)})$ for episode $t$ and receive reward $Y^{(t)}$.
		\EndFor
	\end{algorithmic}
\end{algorithm}

Thm.~\ref{thm:_7_2_ucb} implies that \texttt{Ctf-UCB} is able to eventually learn an optimal counterfactual policy $\pi^*(X \mid X')$ that consistently improves the agent's intended action. 
On the other hand, without perceiving the agent's intended action, the standard \texttt{UCB} algorithm only performs atomic interventions $\doo(x)$. 
Note that an optimal counterfactual policy consistently dominates the best possible interventional one (Thm.~\ref{thm:_7_1_ctf}). 
One important observation is that the standard \texttt{UCB} generally experiences linear regret when compared with an optimal counterfactual agent.
\begin{corollary}
	Let $\Pi$ be an experimental policy space $\Set{\Tuple{X, \emptyset }}$, $Y$ be the reward variable with support on $[0, 1]$, and let the domain of action $X$ be $\D(X) = \{1, \dots, K\}$. There exists an MAB environment $\1M^*$ such that for any algorithm (e.g., \texttt{UCB}) optimizing over space $\Pi$ after $T > 1$ episodes is lower bounded by
	\begin{align}
		R(T, \1M^*) \geq 0.5T
	\end{align} \hfill $\blacksquare$
\end{corollary}
In words, there is an MAB environment such that for any online algorithm employing Fisherian randomization, it must incur at least $0.5$ regret on average per episode of interaction. It is thus unable to achieve an optimal counterfactual policy accounting for the agent's intuition.\footnote{See Example~\ref{exp:_7_casino} for details of the construction of this MAB environment.}

The proposed augmentation procedure may appear to be an immediate extension of \texttt{UCB} in contextual bandits using the agent's natural predilection as an extra context. However, the augmented \texttt{Ctf-UCB} differs from contextual \texttt{UCB} in the following. \footnote{This augmentation procedure is applicable to empower other bandit algorithms, including Thompson sampling \citep{bareinboim2015bandits,forney2017counterfactual,forney2019counterfactual}, with the capability of counterfactual randomization and obtain an optimal counterfactual policy.}
\begin{enumerate}
	\item The agent's intended action $X$ is semantically different from a context $S$. The former is a variable only existing under the observational regime (see), as shown in Fig.~\ref{fig:_7_1_see}. It is replaced by the agent's realized action and does not appear under the interventional regime (do), as shown in Fig.~\ref{fig:_7_1_do}. On the other hand, the context variable $S$ is not affected by the agent's interaction regime with the environment.
	\item The realization of using the agent's intended action $X$ as an additional context is a consequence of counterfactual intervention (ctf-do). On the other hand, online RL algorithms interacts with the environment by repeatedly performing randomized interventions (do), discarding the agent's natural predilection.
\end{enumerate}
\begin{figure}[t]
	\centering
	\hfill
	\begin{minipage}[b]{0.45\linewidth}
		\centering
		\includegraphics[width=1.0\textwidth]{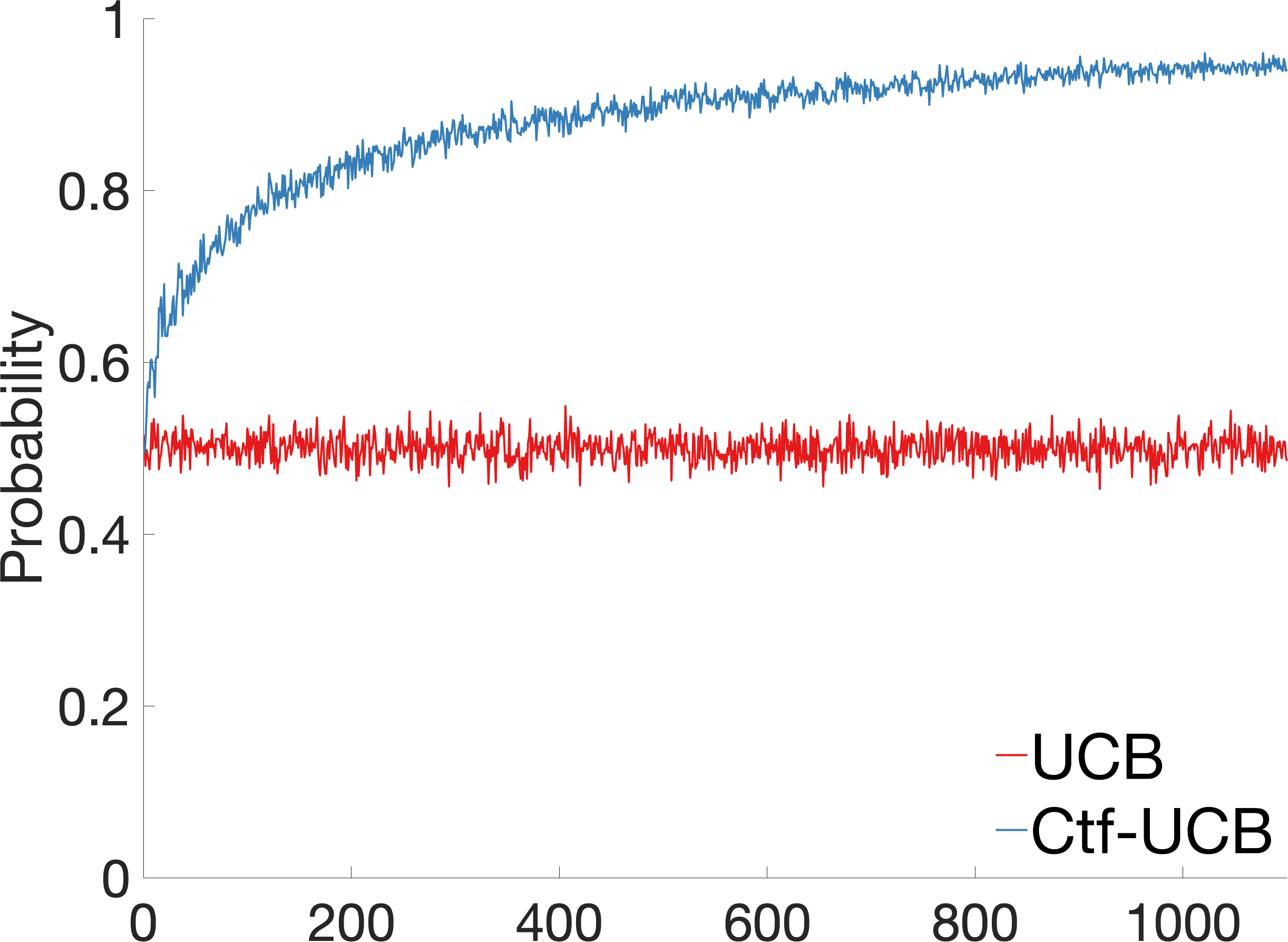}
		\subcaption{}
		\label{fig:_7_2_regret_a}
	\end{minipage}\hfill
	\begin{minipage}[b]{0.45\linewidth}
		\centering
		\includegraphics[width=1.0\textwidth]{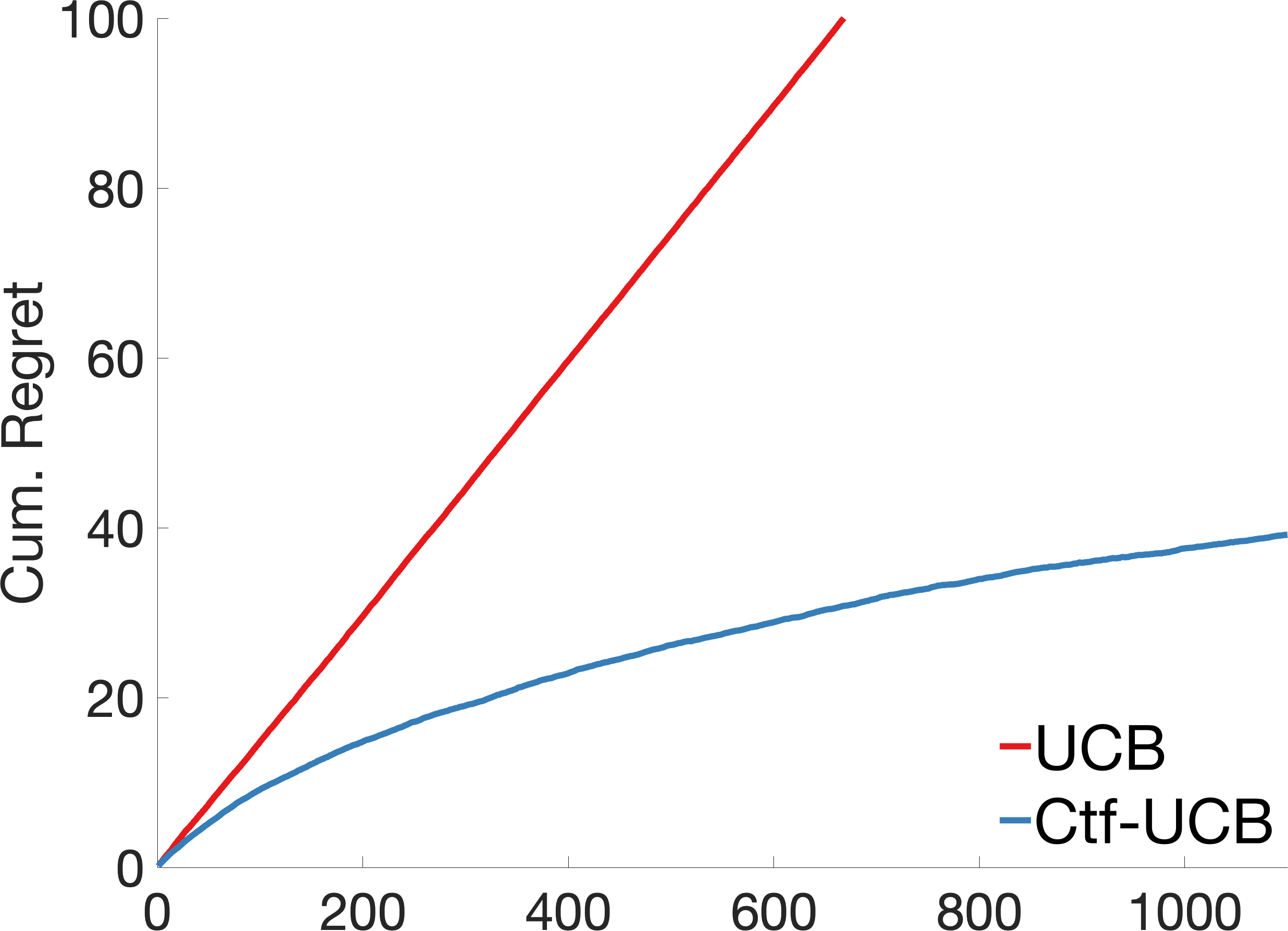}
		\subcaption{}
		\label{fig:_7_2_regret_b}
	\end{minipage}\hfill\null
	\caption{Performance of standard \texttt{UCB} performing atomic interventions and the augmented \texttt{Ctf-UCB} using counterfactual interventions; x-axis represents the total episodes of interactions. The x-axis represents, respectively, the probability of picking an optimal action and the cumulative regret in (\subref{fig:_7_2_regret_a}) and (\subref{fig:_7_2_regret_b}); the y-axis represents the number of episodes in both (\subref{fig:_7_2_regret_a}) and (\subref{fig:_7_2_regret_b}).} \label{fig:_7_2_regret}
\end{figure}

\begin{experiment}
	We evaluate the standard \texttt{UCB} algorithm that attempts to maximize rewards based on $\invE{Y}{x}$, ignoring the agent's intended arm choice, and \texttt{Ctf-UCB} described in Alg.~\ref{alg:_7_2_ucb}, which maximizes the rewards based on ETT $\E \Brackets{Y_{X \gets x'} \mid X = x}$ via counterfactual interventions. 
	All reported simulations are partitioned into rounds of $T = 1,000$ trials averaged over $N = 1,000$ repetitions.

	The Greedy Casino parameterization (specified in Table~\ref{tab:_7_mab}) illustrates the scenario where each arm's payout appears to be equivalent under the observational and experimental distributions alone. Only when we concert the two distributions and condition on a player's predilection the optimal policy can be obtained. 
Fig.~\ref{fig:_7_2_regret} shows the cumulative regret and probability of selecting optimal arms for evaluated algorithms. Simulations support the efficacy of counterfactual interventions. Analyses revealed a significant difference in the regret experienced by \texttt{Ctf-UCB} compared to standard \texttt{UCB}, which, predictably, is not a competitor experiencing linear regret. \hfill $\blacksquare$
\end{experiment}
So far we have introduced a novel type of interaction between the agent and the environment, i.e., the counterfactual randomization, in single-stage decision-making settings such as MABs. We showed online learning agents using counterfactual randomizations consistently outperform their experimental counterparts that do not actively consider the agent's intended action. Novel adaptive counterfactual randomization procedures were proposed to optimize an unknown MAB environment. 

\subsubsection{Counterfactual Randomization for MDPs}\label{sec:_7_2_1}
In this section, we generalize the counterfactual randomization to the more generalized sequential decision-making setting, e.g., MDPs. 
We will endow online algorithms in an unknown MDP with the ability of counterfactual reasoning so that they can learn optimal counterfactual policies, accounting for the agent's intended action and natural predilections. 

To make the argument more precise, we will focus on the episodic learning setting in an unknown MDP environment $\1M^*$ with a finite horizon $H$. The algorithm will interact with $\1M^*$ for repeated episodes $t = 1, 2, \dots, T$. 
For every episode $t$, the algorithm picks a counterfactual policy $\pi^{(t)} = \Parens{\pi^{(t)}_{1}(X_1 \mid S_1, X'_1), \dots, \pi^{(t)}_{H}(X_H \mid S_H, X'_H)}$ in the counterfactual space $\Pi_{\textsc{ctf}}$, performs intervention $\ctf\Parens{\pi^{(t)}}$, and receives subsequent reward signals $\*Y^{(t)} = \Braces{Y^{(t)}_1, \dots, Y^{(t)}_H}$. We are interested in maximizing the undiscounted cumulative reward $\1R(\*Y) = \sum_{i=1}^{H} Y_i$. \footnote{If the decision horizon $H$ is sufficiently large, the undiscounted cumulative reward provides an approximation to the discounted cumulative reward with an infinite horizon \citep{kearns1999approximate}.} For the convenience of the analysis, we will assume that parameters of the counterfactual reward function $\1R_{\text{ctf}}(s, x, x') = \E \Brackets{Y_{i_{X_i \gets x'}} \mid S_{i} = s, X_{i} = x}$ are known. However, our analysis generalizes immediately to settings where the reward function is not accessible.

\begin{algorithm}[!t]
	\caption{Counterfactual \texttt{UCBVI} in MDP (\texttt{Ctf-UCBVI})}
	\label{alg:_7_4_ucb}
	\setlength{\textfloatsep}{0pt}
	\begin{algorithmic}[1]
		\Require a policy space $\Pi = \Braces{\Tuple{X_i, \{S_i, X_i\}}}_{i = 1}^H$, a reward function $\1R(\*Y) = \sum_{i = 1}^{H} Y_i$.
		\State Initialize data $\*H^{(0)} = \emptyset$.
		\For{episodes $t = 1, 2, \dots $}
		\State $\hat{Q}_{t-1} = \texttt{UCB-Q-values}(\*H^{(t-1)})$.
		\For{step $i = 1, \dots, H$}
		\State Observe current state $S_i = S^{(t)}_i$.
		\State Intercept the agent's intended action $X_i = X^{(t)}_i$.
		\State Pick a new action $X'^{(t)}_i = \argmax_{x} \hat{Q}^{(i)}_{t-1}\Parens{S^{(t)}_i, X^{(t)}_i, x}$.
		\State Perform $\doo(X_t \gets X'^{(t)}_i)$ and receive reward $Y^{(t)}_i$.
		\State Update data $\*H^{(t)} = \*H^{(t-1)} \cup \Braces{S^{(t)}_i, X^{(t)}_i, X'^{(t)}_i}$.
		\EndFor
		\EndFor
	\end{algorithmic}
\end{algorithm}

\begin{algorithm}[!t]
	\caption{\texttt{UCB-Q-values}}
	\label{alg:_7_4_ucb_q}
	\setlength{\textfloatsep}{0pt}
	\begin{algorithmic}[1]
		\Require Data $\*H^{(t)} = \Braces{\*S^{(i)}, \*X^{(i)}, \*X'^{(i)}, \*Y^{(i)}}_{i = 1}^{t}$

		\State For all $s, s' \in \D(S)$ and all $x, x', x'' \in \D(X)$, compute from data $\*H^{(t)}$
		\begin{align}
			N_{t}(s, x, x') &= \sum_{k = 1}^{t} \sum_{i = 1}^H \I \Braces{S^{(k)}_i = s, X^{(k)}_i = x, X'^{(k)}_i = x'} \\
			N'_{t}(s, x, x') &= \sum_{k = 1}^{t} \sum_{i = 1}^H Y^{(k)}_i \I \Braces{S^{(k)}_i = s, X^{(k)}_i = x, X'^{(k)}_i = x'} \\
			N_{t}(s, x, x', s', x'') &= \sum_{k = 1}^{t} \sum_{i  = 1}^H \I \Braces{S^{(k)}_i = s, X^{(k)}_i = x, X'^{(k)}_i = x', S^{(k)}_{i+1} = s, X'^{(k)}_{i+1} = x''}
		\end{align}
		\State Let $\1K = \Braces{(s, x, x') \in \D(S) \times \D(X) \times \D(X) \mid N_{t}(s, x, x') > 0}$.
		\State For $(s, x, x') \in \1K$, compute estimates
		\begin{align}
			 &\hat{\1T}_t(s, x, x', s', x'') = \frac{N_{t}(s, x, x', s', x'')}{N_{t}(s, x, x')}, &\hat{\1R}_t(s, x, x') = \frac{N'_{t}(s, x, x')}{N_{t}(s, x, x')}
		\end{align}
		\State Initialize $V^{(H+1)}_{t}(s, x) = 0$, $Q^{(H)}_t(s, x, x') = H$ for all $(s, x, x') \in \D(S) \times \D(X) \times \D(X)$.
		\For{$i = H, H-1, \dots, 1$}
		\State For all $(s, x, x') \in \1K$, compute function $Q^{(h)}_t$ as
		\begin{align}
			\!\! Q^{(h)}_t(s, x, x') = 
			\min \Braces{Q^{(h)}_{t-1}(s, x, x'), H, \hat{\1R}_t(s, x, x') + \Parens{\hat{\1T}_t  V^{(h+1)}_t}(s, x, x') + b^{(h)}_t(s, x, x')} \notag 
		\end{align}
		where the bonus function $b^{(h)}_t$ is defined as
		\begin{align}
			b^{(h)}_t(s, x, x') = 7H\sqrt{ \ln \Parens{5|\D(S) \times \D(X)\times \D(X)| T /\delta
			 } \over N_{t}(s, x, x')}
		\end{align}
		\State Let $V^{(h)}_t(s, x) = \max_{x'} Q^{(h)}_t(s, x, x')$.
		\EndFor
	\end{algorithmic}
\end{algorithm}
We will utilize \texttt{UCBVI} \citep{azar2017minimax}, an online reinforcement learning algorithm that can learn the best possible experimental policy $\pi \in \Pi_{\textsc{exp}}$ in a finite-horizon MDP environment. 
Alg.~\ref{alg:_7_4_ucb} shows an augmented procedure that incorporates counterfactual randomization which we call \texttt{Ctf-UCBVI}. 
More specifically, for every episode $t$, it computes a policy $\pi^{(t)}$ based on the data $\*H^{(t-1)}$ collected prior to episode $t$. At Step 3, it calls \texttt{UCB-Q-values} (Alg.~\ref{alg:_7_4_ucb_q}),  which returns upper confidence bounds on the optimal Q-values $Q_*(s, x, x')$. This is computed using an empirical Bellman operator with an additional confidence bonus, estimated based on Chernoff-Hoeffding's concentration inequality. The empirical estimates of the counterfactual transition distributions $\1T_{\text{ctf}}$ and the conditional reward $\1R_{\text{ctf}}$ are consistent following Lem.~\ref{lem:_7_3_ctf}. At Step 6 in \texttt{UCB-Q-values}, the linear operator $ \Parens{\hat{\1T}_t \cdot V^{(h+1)}_t}(s, x, x')$ is defined as,
\begin{align}
	\Parens{\hat{\1T}_t \cdot V^{(h+1)}_t}(s, x, x') = \sum_{s', x''} \hat{\1T}_t(s, x, x', s', x'') V^{(h+1)}_t(s', x'')
\end{align}
At Steps 5 -- 9, \texttt{Ctf-UCBVI} sequentially performs counterfactual intervention $\ctf$ on every action $X_1, \dots, X_H$. For every decision horizon $i = 1, \dots, H$, it observes the current state $S_i = S^{(t)}_i$, and intercepts the agent's intended action $X'_i = X^{(t)}_i$. 
The algorithm then computes an alternative action $X^{(t)}_i$ by maximizing the empirical Q-values $\hat{Q}_{t-1}$ computed from data $\*H^{(t)}$. Finally, it performs the selected action $\doo(X_t \gets X'^{(t)}_i)$ and receives a subsequent reward $Y^{(t)}_i$. 

Following the derivation in \citep{azar2017minimax}, it is possible to show that \texttt{Ctf-UCBVI}, empowered with counterfactual randomization, is able to obtain an optimal counterfactual policy in $\Pi_{\textsc{ctf}}$ while achieving a sublinear regret. Formally,
\begin{theorem}[Regrets of \texttt{Ctf-UCBVI} in MDPs]\label{thm:_7_4_ucb}
	For an MDP $\langle \1M^*, \Pi, \1R \rangle$ with horizon $H \in \3N$, let $\Pi$ be a counterfactual policy space $\Set{\Tuple{X_i, \{S_i, X_i\}}}_{i=1}^{N}$, $\1R(\*Y) = \sum_{i = 1}^{H} Y_i$ be a cumulative reward function over bounded reward signals $Y_i \in [0, 1]$. It holds the regret of \texttt{Ctf-UCBVI} in SCM $\1M^*$ after $T > 1$ episodes is bounded by
	\begin{align}
		R(T, \1M^*) \leq  CH^{3/2}\sqrt{|\D(S) \times \D(X)| T\log(T)} \label{eq:_7_4_ucb_regret}
	\end{align}
	where $C$ is a universal constant; $\D(S)$ and $\D(X)$ are domains of every state $S_i$ and action $X_i$, $i = 1, 2, \dots$, respectively. \hfill $\blacksquare$

\end{theorem}
On the other hand, without considering the agent's intended action, online algorithms performing randomized experiments may never be able to converge to an optimal counterfactual policy. The linear regret could occur when the intended action $X_i$ reveals valuable information about the unobserved confounder $U_i$, as highlighted by the next proposition. 
\begin{corollary}
	Let $\Pi$ be an experimental policy space $\Set{\Tuple{X_i, \{S_i, X_i\}}}_{i=1}^{H}$, $\1R(\*Y) = \sum_{i = 1}^{H} Y_i$ be a reward function over bounded reward signal $Y_i \in [0, 1]$. There exists an MDP environment $\1M^*$ such that for any algorithm (e.g., \texttt{UCBVI}) optimizing over space $\Pi$ after $T > 1$ episodes is lower bounded by
	\begin{align}
		R(T, \1M^*) \geq 0.08 HT
	\end{align} \hfill $\blacksquare$
\end{corollary}
Similarly to the MAB setting, the above proposition implies that there is an MDP environment such that for any online algorithm following Fisherian randomization, it suffers at least a constant regret on average per every step of the interaction. 
Therefore, these existing algorithms are generally incapable of obtaining an optimal counterfactual policy in MDPs while achieving a sublinear regret. Fortunately, one could augment these RL algorithms with counterfactual reasoning by replacing standard interventions with counterfactual interventions. 
The proposed \texttt{Ctf-UCBVI} (Alg.~\ref{alg:_7_4_ucb}) demonstrates this augmentation procedure in the \texttt{UCBVI} algorithm. The following simulation demonstrates the performance of \texttt{Ctf-UCBVI} in a simple MDP environment.

\begin{figure}[t]
	\centering
	\hfill
	\begin{minipage}[b]{0.45\linewidth}
		\centering
		\includegraphics[width=1.0\textwidth]{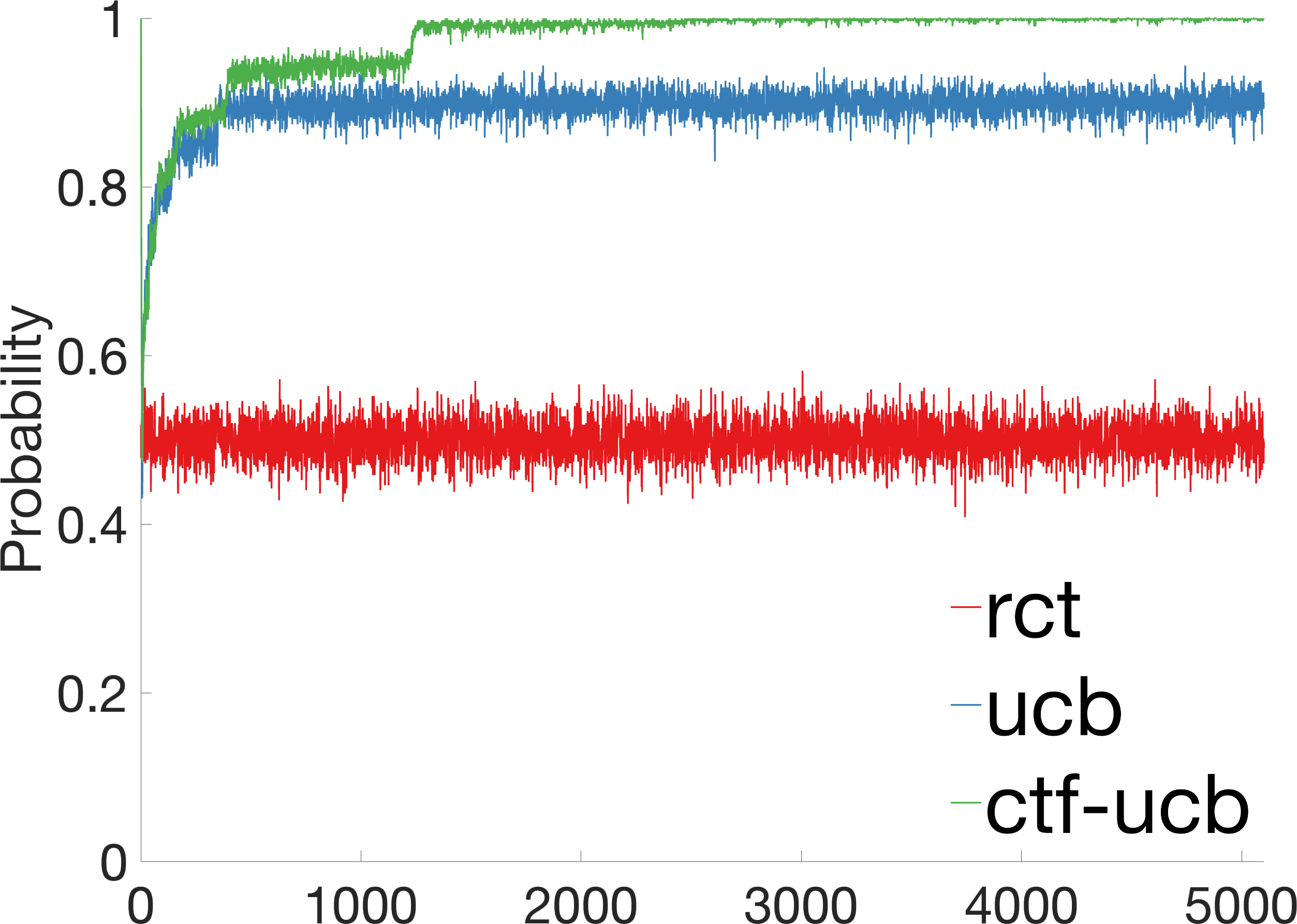}
		\subcaption{}
		\label{fig:_7_4_regret_a}
	\end{minipage}\hfill
	\begin{minipage}[b]{0.45\linewidth}
		\centering
		\includegraphics[width=1.0\textwidth]{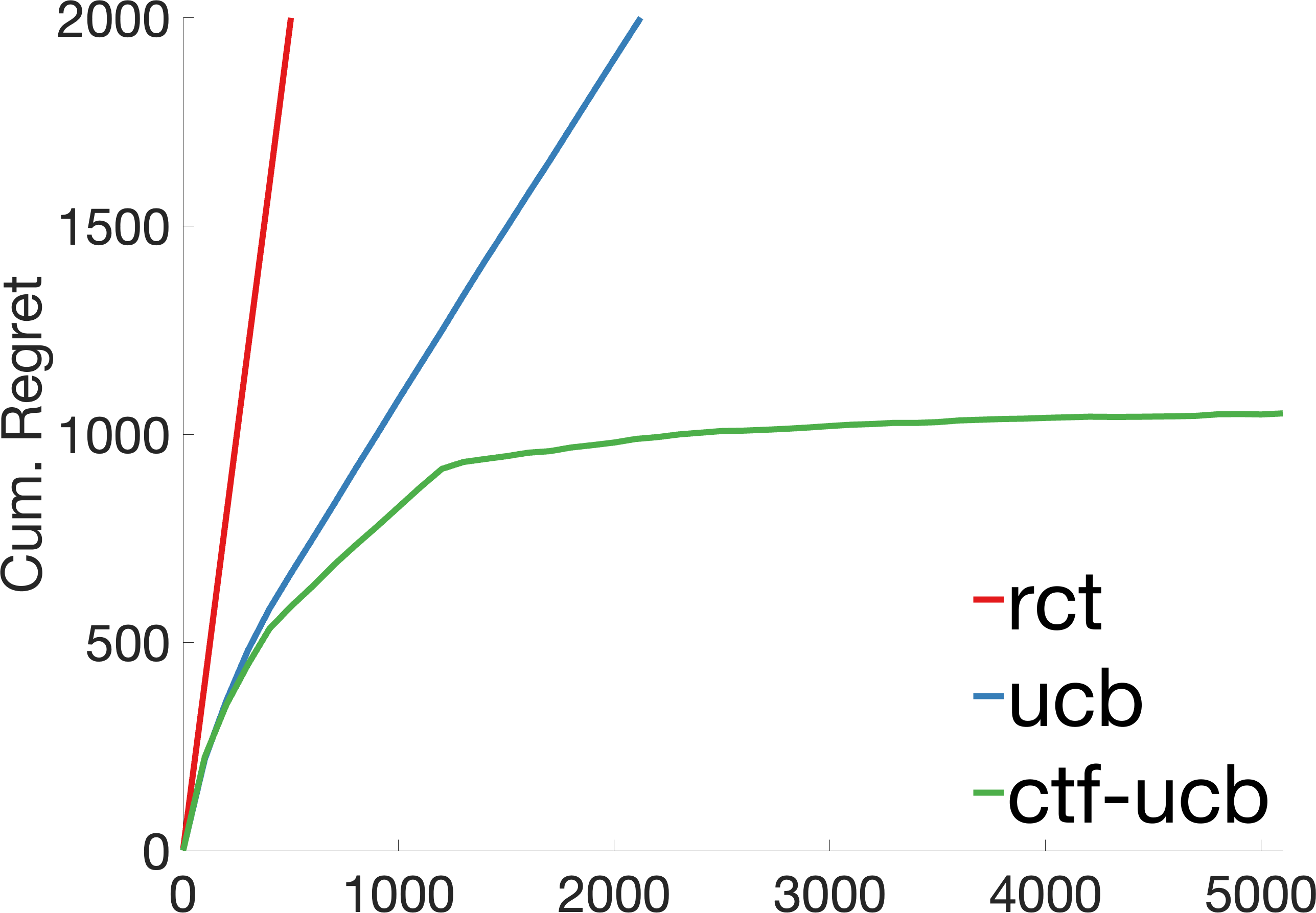}
		\subcaption{}
		\label{fig:_7_4_regret_b}
	\end{minipage}\hfill\null
	\caption{Performance of standard \texttt{UCBVI} performing atomic interventions and the augmented \texttt{Ctf-UCBVI} using counterfactual interventions.} \label{fig:_7_4_regret}
\end{figure}
\begin{experiment}\label{exp:_7_4_mdp}
	We evaluate standard \texttt{UCBVI} that attempts to maximize rewards based on interventional transitional probabilities $\1T_{\text{exp}}$ and reward function $\1R_{\text{exp}}$, while it ignores the agent's intended actions. 
	We also evaluate the augmented \texttt{Ctf-UCBVI} algorithm described in Alg.~\ref{alg:_7_4_ucb} which attempts to maximize cumulative reward based on counterfactual transitional probabilities $\1T_{\text{ctf}}$ and reward function $\1R_{\text{ctf}}$. It actively accounts for the agent's intended actions by performing counterfactual interventions. All reported simulations are partitioned into rounds of $T = 5000$ trials averaged over $N = 1000$ repetitions.

	The detailed parameterization of the MDP environment with unobserved confounders is provided in Eq.~\ref{eq:_2_1_mdp} where the decision horizon $H = 10$. Fig.~\ref{fig:_7_4_regret} shows the cumulative regret and probability of selecting optimal arms for evaluated algorithms. Simulation results support our proposed online RL algorithms using counterfactual interventions. Analyses revealed that standard \texttt{UCBVI} suffered from a linear regret. Meanwhile, the augmented \texttt{Ctf-UCBVI} achieved a sublinear regret, showing that it is able to obtain an optimal counterfactual policy actively accounting for the agent's intended actions in the decision-making process. \hfill $\blacksquare$
\end{experiment}

\subsection{The Tradeoff between Autonomy and Optimality} \label{sec:_7_3}

\begin{table}[t]	
\centering
  \setlength{\tabcolsep}{10pt}
   \renewcommand{\arraystretch}{1.25}
        \begin{tabular}{|c | c | c | c | c | c | c|}
        \toprule
        \multirow{2}{*}{Environment} &\multirow{2}{*}{Structural Assumptions} &\multicolumn{2}{c|}{Optimality} &\multirow{2}{*}{Autonomy}\\ \cline{3-4}
        & & $\Pi_{\textsc{exp}}$ & $\Pi_{\textsc{ctf}}$ & \\ \hline
        \multirow{2}{*}{MAB}& NUC &\cmark & \cmark &\cmark  \\ \cline{2-5}
        & - &\xmark & \cmark &\xmark  \\ \hline
        \multirow{2}{*}{MDP}& NUC &\cmark & \cmark &\cmark  \\ \cline{2-5}
        & - &\xmark &\cmark  & \xmark \\ \bottomrule
    \end{tabular}
    \makeatletter\def\@captype{table}\makeatother
    \caption{The performance of counterfactual policies $\Pi_{\textsc{exp}}$ and experimental policies $\Pi_{\textsc{ctf}}$ in canonical environments including MABs and MDPs.}
    \label{tab:_7_3}
\end{table}

We have established that an agent following a counterfactual policy generally outperforms an interventional one, as summarized in Table~\ref{tab:_7_3}. 
The ``Environment'' column lists canonical decision-making environments covered in this section, such as MABs and MDPs. 
The ``Structural Assumptions'' column details additional structural assumptions applied to the environment, including no unmeasured confounding (NUC, Def.~\ref{def:_4_1_nuc}). 
The ``Optimality'' column indicates whether optimizing within the corresponding policy space could lead to an optimal decision strategy, using an optimal counterfactual policy in $\Pi_{\textsc{ctf}}$ as the baseline. Here, a check (cross) mark in $\Pi_{\textsc{exp}}$ under ``Optimality'' denotes whether experimental policies are capable (or not) of achieving optimal performance in the respective environment, compared to their counterfactual counterparts. 
In environments where unobserved confounders generally exist, full autonomy (as indicated in the ``Autonomy'' column) is attainable only when it does not compromise optimality. This occurs when the performance of experimental policies $\Pi_{\textsc{exp}}$ and counterfactual policies $\Pi_{\textsc{ctf}}$ coincide. 

This suggests a fundamental tradeoff between optimality and autonomy in the design of RL systems. While full autonomy is preferable, the agent could potentially achieve superior performance by leveraging a human's capabilities through counterfactual reasoning. 
For instance, as demonstrated in Example~\ref{exp:_7_1_1_mdp1}, a counterfactual policy in $\Pi_{\textsc{ctf}}$ characterizes interactions in a semi-autonomous system that incorporates the human operator's intended action as input. 
On the other hand, deploying an interventional policy in $\Pi_{\textsc{exp}}$ eliminates the need for a human operator in the underlying environment, resulting in a fully autonomous system. Consequently, Thm.~\ref{thm:_7_3_ctf} implies that while full autonomy is preferable, the agent could potentially achieve better performance by leveraging the human's intuition through a counterfactual decision criterion. 

We will model this autonomy-optimality tradeoff as a constrained transfer of control (TOC) problem such that a decision system tries to maximize its rewards while repeatedly switching between experimental and counterfactual policies, subject to a budget constraint over the total time of using the agent's intended action, i.e., no more than $\delta \in (0, 1)$ ratio of the total running time. For instance, an autonomous vehicle could ask for the human driver's input when necessary, but no more than $\delta = 10\%$ of the total expected driving time. 
Formally, we first define hybrid policies, which is a restricted family of counterfactual policies $\Pi_{\textsc{ctf}}$ that contains experimental policies $\Pi_{\textsc{exp}}$. 
\begin{definition}[Hybrid Policies]\label{def:_7_3_hybrid_space}
	For an MDP environment $\1M^*$, a hybrid policy space $\Pi_{\textsc{hyb}}$ is a subset of counterfactual policies $\Pi_{\textsc{ctf}}$. For any hybrid policy $\pi = \Parens{\pi_1, \dots, \pi_H}$ in $\Pi_{\textsc{hyb}}$, its decision rule $\pi_i = (f_i \circ g_i)$ is a composition of functions $f_i, g_i$ such that, for $i = 1, \dots, H$,
	\begin{itemize}
	\item $g_i$ is a probability distribution mapping from the current state $S_i$ to an extended action $A_i \in \{0, 1\}$ where ``$0$'' stands for \emph{experimental decision criterion} and ``$1$'' for \emph{counterfactual decision criterion}. That is, 
	\begin{align}
		A_i \sim g_i\Parens{A_i \mid S_i}
	\end{align}
	\item $f_i$ is a probability distribution mapping from the current state $S_i$, the intended action $X_i$, and the extended acton $A_i$ to the realized action $X'_i$. Moreover, the extended action $A_i$ decides the realized action $X_i$ as follows:
	\begin{align}
		&f_i\Parens{X_i \mid S_i, X'_i, A_i} = \begin{cases}
                f_i\left(X_i \mid S_i\right), &\text{if }A_i = 0\\
                f_i\left(X_i \mid S_i, X'_i\right), &\text{if }A_i = 1
                \end{cases}
	\end{align}
	\end{itemize} \hfill $\blacksquare$
\end{definition}
In words, the decision-making process of a hybrid policy consists of two phases. First, it determines an extended action $A_i \in \{0, 1\}$ based on the current state $S_i$. Here, $A_i$ is a switch variable indicating whether to utilize counterfactual reasoning ($A_i = 1$), or stay in the standard experimental decision criterion ($A_i = 0$). The human's intuition $X_i$ is included as an evidence to decide the realized decision $X'_i$ if $A_i = 1$; otherwise, it is ignored. Fig.~\ref{fig:_7_3_mdp} shows the augmented causal diagram of an MDP environment induced by a hybrid policy where extended actions $A_i$ are added to control the modes of decision-making criterion.  Fix a budget $\delta \in [0, 1]$. For a hybrid policy $\pi$ that does not apply counterfactual decision criterion with a probability higher than $\delta$ as the decision stage $i$ grows, it must satisfy the following constraint:
\begin{align}
	\lim_{i \to \infty} P_{\pi}\Parens{A_i = 1} \leq \delta. \label{eq:_7_3_budget}
\end{align}
If the above equation holds, this means that an agent following policy $\pi$ will not utilize the human's intuition more than $\delta \times 100 \%$ of the total running time in long-term. 

\pgfsetlayers{back,main}

\begin{figure}[t]
\centering
		\begin{tikzpicture}
			\def\outerr{3.2}
			\def\innerr{3}
			\node[vertex] (S1) at (-2.5, -1) {S\textsubscript{1}};
			\node[vertex] (X1) at (-1.5, 0) {X\textsubscript{1}};
			\node[vertex] (Xp1) at (0, 0) {X'\textsubscript{1}};
			\node[vertex] (A1) at (-1.5, -2) {A\textsubscript{1}};
			\node[vertex] (Y1) at (0, -2) {Y\textsubscript{1}};
			\node[vertex] (S2) at (1, -1) {S\textsubscript{2}};
			\node[vertex] (X2) at (2, 0) {X\textsubscript{2}};
			\node[vertex] (Xp2) at (3.5, 0) {X'\textsubscript{2}};
			\node[vertex] (A2) at (2, -2) {A\textsubscript{2}};
			\node[vertex] (Y2) at (3.5, -2) {Y\textsubscript{2}};
			\node[vertex] (S3) at (4.5, -1) {S\textsubscript{3}};
			\node[vertex] (X3) at (5.5, 0) {X\textsubscript{3}};
			\node[vertex] (Xp3) at (7, 0) {X'\textsubscript{3}};
			\node[vertex] (A3) at (5.5, -2) {A\textsubscript{3}};
			\node[vertex] (Y3) at (7, -2) {Y\textsubscript{3}};

			\draw[dir] (S1) to (S2);
			\draw[dir] (S1) to (X1);
			\draw[dir] (S1) to (Y1);
			\draw[dir] (S1) to (Xp1);
			\draw[dir] (X1) to (Xp1);
			\draw[dir] (Xp1) to (Y1);
			\draw[dir] (Xp1) to (S2);

			\draw[bidir] (X1) to  (S2);
			\draw[bidir] (X1) to (Y1);
			\draw[bidir] (Y1) to [bend right = 30] (S2);

			\draw[dir] (S2) to (S3);
			\draw[dir] (S2) to (X2);
			\draw[dir] (S2) to (Y2);
			\draw[dir] (S2) to (Xp2);
			\draw[dir] (X2) to (Xp2);
			\draw[dir] (Xp2) to (Y2);
			\draw[dir] (Xp2) to (S3);

			\draw[bidir] (X2) to (S3);
			\draw[bidir] (X2) to (Y2);
			\draw[bidir] (Y2) to [bend right = 30] (S3);

			\draw[dir] (S3) to (X3);
			\draw[dir] (S3) to (Y3);
			\draw[dir] (S3) to (Xp3);
			\draw[dir] (X3) to (Xp3);
			\draw[dir] (Xp3) to (Y3);
			
			\draw[dir] (S1) to (A1);
			\draw[dir] (S2) to (A2);
			\draw[dir] (S3) to (A3);
			\draw[dir] (A1) to (Xp1);
			\draw[dir] (A2) to (Xp2);
			\draw[dir] (A3) to (Xp3);
			
			\draw[bidir] (X3) to (Y3);

			\begin{pgfonlayer}{back}
				\node[circle,fill=betterblue!65,draw=none,minimum size=2*\innerr mm] at (X1) {};
				\node[circle,fill=betterblue!65,draw=none,minimum size=2*\innerr mm] at (X2) {};
				\node[circle,fill=betterblue!65,draw=none,minimum size=2*\innerr mm] at (X3) {};
				\node[circle,fill=betterred!65,draw=none,minimum size=2*\innerr mm] at (Y1) {};
				\node[circle,fill=betterred!65,draw=none,minimum size=2*\innerr mm] at (Y2) {};
				\node[circle,fill=betterred!65,draw=none,minimum size=2*\innerr mm] at (Y3) {};
			\end{pgfonlayer}
		\end{tikzpicture}

	\caption{Causal diagram for the MDP environment induced by a hybrid policy where extended actions $A_i$ are added to control the modes of experimental and counterfactual decision criteria.}
	\label{fig:_7_3_mdp}
\end{figure}
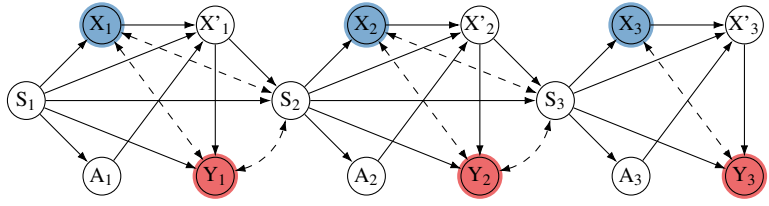

\begin{example}\label{exp:_7_3_mdp1}
	Recall the MDP environment $\1M^*$ described in Eq.~\ref{eq:_2_1_mdp}, where the decision horizon $H = \infty$. Let a hybrid policy $\pi = \Parens{\pi_1(X_1 \mid X_1, S_1), \pi_2(X_2 \mid X_2, S_2), \dots}$ such that for every $i = 1, 2, \dots$, the decision rule $\pi_i$ is defined as, 
	\begin{align}
		\pi_i \triangleq X_i \gets \neg X'_i \cdot A_i +  \neg S_i \cdot (1-A_i)
	\end{align}
	where the extended action $A_i$ is drawn over the binary domain $\{0, 1\}$ such that $P(A_i = 1) = 0.1$. Evidently, an agent following such a hybrid policy $\pi$ must satisfy $P_{\pi}\Parens{A_i = 1} = 0.1$ for every $i = 1, 2, \dots$. That is, the agent will not utilize the intended action for more than $0.1 \times 100\%$ of total running time in long term. \hfill $\blacksquare$
\end{example}
We next introduce a planning algorithm to solve for an optimal hybrid policy in an MDP environment subject to the budget constraint in Eq.~\ref{eq:_7_3_budget}. Our previous discussion provided dynamic programming approaches (e.g., value iteration and policy iteration) for optimizing the discounted expected cumulative rewards over counterfactual policies. Here, we first describe an alternative planning strategy using linear programming. Formally, optimizing discounted rewards over counterfactual policies $\Pi_{\textsc{ctf}}$ in an MDP environment $\1M^*$ can be reduced to solving the following equivalent linear program (LP) \citep{d1963probabilistic,kallenberg1983linear}, 
  \begin{equation}
  \begin{aligned}
    \text{max}\;\; & \sum_{s, x, x'} \1R_{\text{ctf}}(s, x, x')\phi(s, x, x') \\
    \text{subject to}\;\; & \forall s, x \in \1S \times \1X, \;\; \phi(s, x, x')  \geq 0 \\
    & \sum_{x'} \phi(s, x, x')  = \alpha\Parens{s, x} + \gamma \sum_{s', x'', x'}\phi\Parens{s', x'', x'}\1T_{\text{ctf}}(s, x, x', s', x'')
    \end{aligned} \label{eq:_7_3_lp_ctf}
  \end{equation} 
where $\alpha\Parens{s, x'} = P(S^{(1)} = s, X^{(1)} = x')$ specifies the observational distribution over the initial state and action. The optimization variables $\phi(s, x, x')$ are called the \textit{occupation measure} of a policy, where $\phi(s, x, x')$ is the total discounted number of times action $X_i = x$ is realized in the observed state $S_i = s$, provided with the intended action $X_t = x$. An optimal counterfactual policy in $\Pi_{\textsc{ctf}}$ is stationary and can be computed from a solution to the above LP as, for $i = 1, 2, \dots$,
\begin{align}
	\pi^*_{i}(x'|s, x) = \frac{\phi(s, x, x')}{\sum_{x'}\phi(s, x, x')}.
\end{align}
Next we extend the LP formulation in Eq.~\ref{eq:_7_3_lp_ctf} to solve for an optimal hybrid policy under a budget constraint. Let optimization variables $\phi(s, x, x', a)$ denote the \textit{occupation measure} of a hybrid policy over the realized action $X'_i = x'$, observed state $S_i = s$, the intended action $X_t = x$, and the extended action $A_i = a$. Since the extended action $A_i$ does not directly affect the reward signal $Y_i$ and next state $S_{i+1}$, the transition probabilities $\1T_{\text{ctf}}$ and reward function $\1R_{\text{ctf}}$ induced by hybrid policies remain the same as those induced by   counterfactual policies. An unconstrained hybrid policy optimizing the MDP environment is thus obtainable by solving the following LP, 
  \begin{equation}
  \begin{aligned}
    \text{max}\;\; & \sum_{s, x, x', a} \1R_{\text{ctf}}(s, x, x')\phi(s, x, x', a) \\
    \text{subject to}\;\; & \forall s, x \in \1S \times \1X, \forall a \in \{0, 1\}, \;\; \phi(s, x, x', a)  \geq 0 \\
    & \sum_{x', a} \phi(s, x, x', a)  = \alpha\Parens{s, x} + \gamma \sum_{s', x'', x', a}\phi\Parens{s', x'', x', a}\1T_{\text{ctf}}(s, x, x', s', x'')
    \end{aligned} \label{eq:_7_3_lp_hyb}
  \end{equation} 
Meanwhile, the budget constraint over the intended action in Eq.~\ref{eq:_7_3_budget} could be written as:
\begin{align}
        \sum_{s, x, x'}\phi \Parens{s, x, x', 1} & \leq \delta \sum_{s, x, x', a}\phi \Parens{s, x, x', a} \label{eq:_7_3_budget1}\\
        \frac{\phi \Parens{s, x, x', 0}}{\sum_{x'}\phi \Parens{s, x, x', 0}} &= \frac{\sum_{x} \phi \Parens{s, x, x', 0}}{\sum_{x, x'}\phi \Parens{s, x, x', 0}} \label{eq:_7_3_budget2}\\
        \frac{ \sum_{x'} \phi \Parens{s, x, x', a}}{ \sum_{x', a} \phi \Parens{s, x, x', a}} &= \frac{ \sum_{x, x'} \phi \Parens{s, x, x', a}}{ \sum_{x, x', a} \phi \Parens{s, x, x', a}} \label{eq:_7_3_budget3}
\end{align}
Among the above equations, Eq.~\ref{eq:_7_3_budget1} ensures that for an agent operating in the MDP environment, its total time steps applying counterfactual decision criterion ($A_i = 1$) is no more than $\delta \times 100 \%$ of the total time steps (discounted so that future visits count less than present ones). Eq.~\ref{eq:_7_3_budget2} ensures that when an agent applies the experimental decision criterion ($A_i = 0$), the policy $\pi$ does not take the intended action $X'_i = x'$ as an input. 
Finally, Eq.~\ref{eq:_7_3_budget3} reflects the functional constraint that the extended action $A_i$ only depends on the current state $S_i$. An optimal hybrid policy in $\Pi_{\textsc{hyb}}$ satisfying the $\delta$-budget constraint is thus obtainable by solving the LP specified in Eq.~\ref{eq:_7_3_lp_hyb} subject to additional constraints in Eqs.~\ref{eq:_7_3_budget1} - \ref{eq:_7_3_budget3}. This mathematical program forms a polynomial optimization problem \citep{tuy1998convex}, which is neither linear nor convex. Despite its difficulty, several efficient methods of polynomial optimization can be used in this case, for example, the RLT method \citep{sherali2013reformulation}, and a SDP relaxation method \citep{lasserre2001global}.

\begin{figure}[t]
	\centering
	\hfill
	\begin{minipage}[b]{0.45\linewidth}
		\centering
		\includegraphics[width=1.0\textwidth]{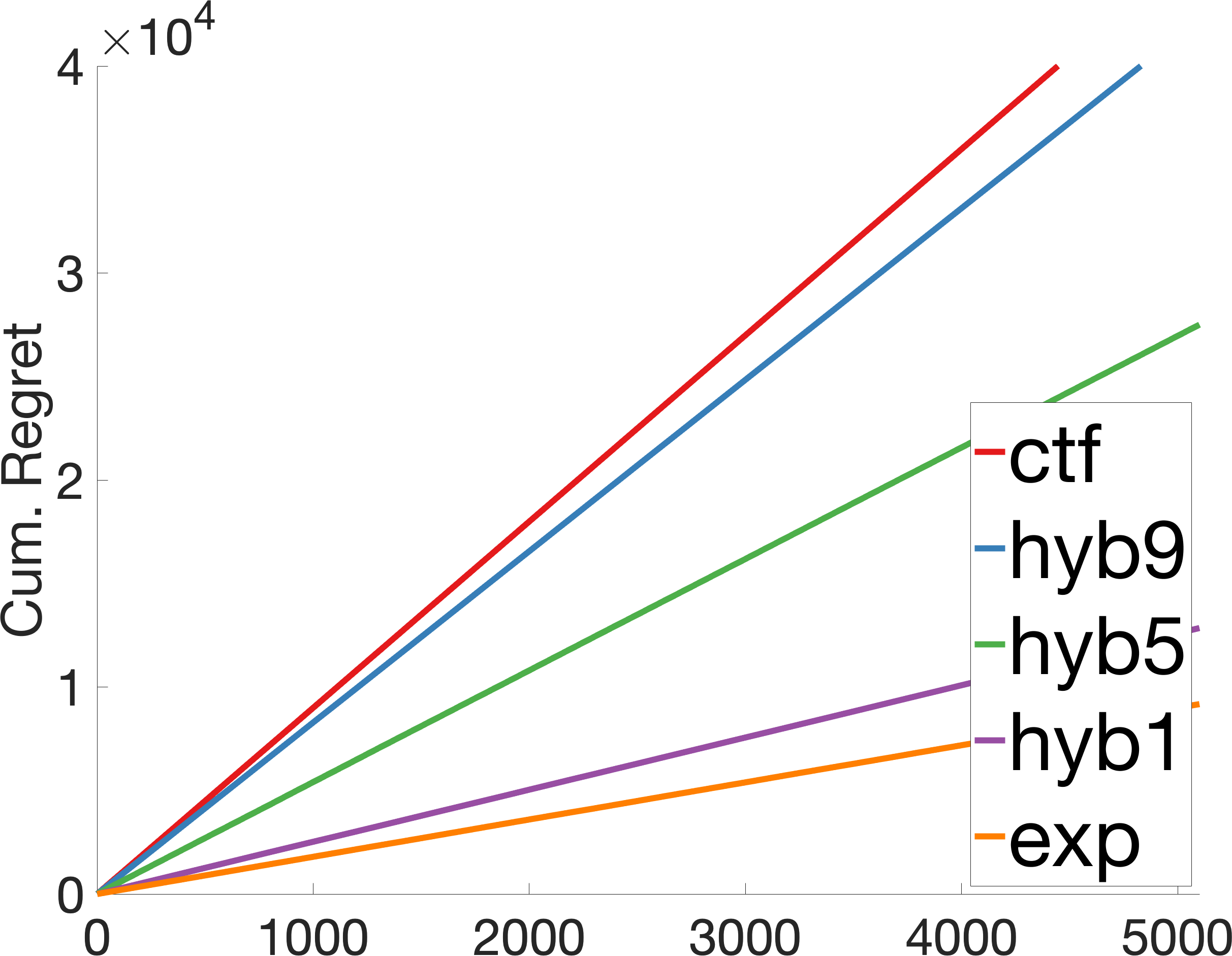}
		\subcaption{}
		\label{fig:_7_3_reward_a}
	\end{minipage}\hfill
	\begin{minipage}[b]{0.45\linewidth}
		\centering
		\includegraphics[width=1.0\textwidth]{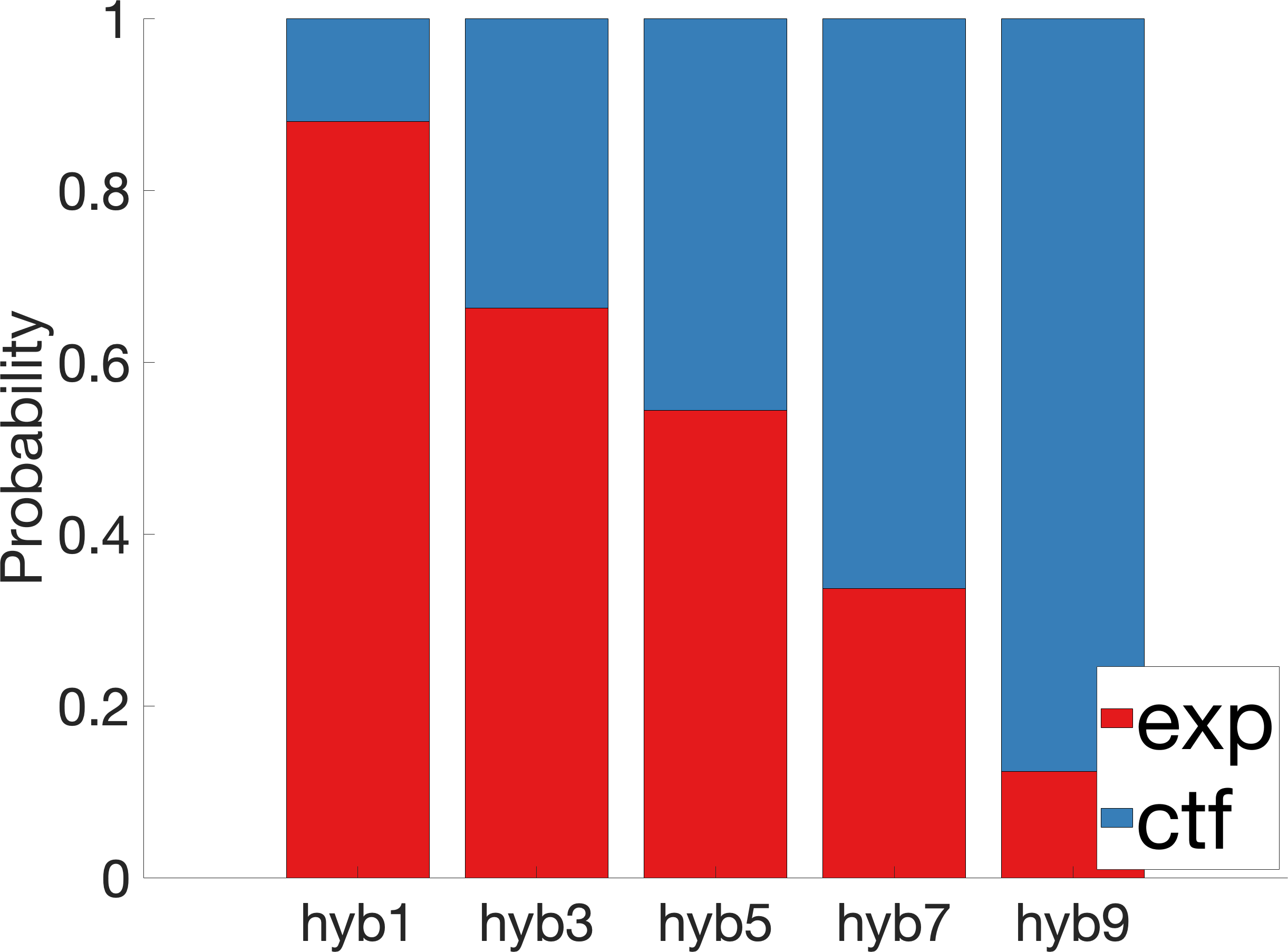}
		\subcaption{}
		\label{fig:_7_3_reward_b}
	\end{minipage}\hfill\null
	\caption{Simulations comparing the performance (\subref{fig:_7_3_reward_a}) and occupancy composition (\subref{fig:_7_3_reward_b}) of \emph{exp} and \emph{ctf} decision criteria; $y$-axis in (\subref{fig:_7_3_reward_b}) represents the ratio of total time performing the experimental or counterfactual decision criterion.} \label{fig:_7_3_reward}
\end{figure}
\begin{experiment}\label{exp:_7_3_mdp2}
We evaluate the performance of optimal hybrid policies in an MDP environment subject to different budget constraints. Recall that $\delta$ is a constraint over the ratio between the total time of performing the counterfactual decision criterion (using the intended action) and the total running time of the system. We compute policies for three hybrid agents with the ratio constraint $\delta$ set to $0.1, 0.5, 0.9$, labeled as \emph{hyb1}, \emph{hyb5} and \emph{hyb9}, respectively. We also include the best experimental and counterfactual policies in $\Pi_{\textsc{exp}}$ and $\Pi_{\textsc{ctf}}$ as the baseline, labeled as \emph{exp} and \emph{ctf} respectively. We use value iteration for MDP planning. As for hybrid policy planning, we employ the SDP relaxation method for polynomial optimization. SDPs are constructed using SparsePOP \citep{waki2008algorithm} and solved with SeDuMi \citep{sturm1999using}.

Detailed parameterization of the MDP environment with unobserved confounders is provided in Eq.~\ref{eq:_2_1_mdp}. Fig.~\ref{fig:_7_3_reward_a} shows the discounted cumulative reward for all algorithms. Simulation results reveal that the performance of hybrid policies converges to the best possible counterfactual policy as $\delta \rightarrow 1$. In particular, \emph{hyb1} ($\delta = 0.1$) shows limited performance improvement over the experimental policy \emph{exp}, while \emph{hyb9} ($\delta = 0.9$) experiences higher cumulative reward, which is comparable to the best counterfactual policy \emph{ctf}. Predictably, the performance of \emph{hyb5} ($\delta = 0.5$) lies in between \emph{hyb1} and \emph{hyb9}. We also show the composition graph of experimental and counterfactual decision criteria in \ref{fig:_7_3_reward_b}. Two hybrid policies with $\delta= 0.3, 0.7$ are included. The simulations support that the polynomial optimization reduction worked as expected, where \emph{hyb1} and \emph{hyb3} tend to stay in the autonomous mode. In contrast, \emph{hyb7} and \emph{hyb9} are more semi-autonomous and actively account for the human's intended action. Unsurprisingly, \emph{hyb5} kept neutral. \hfill $\blacksquare$
\end{experiment}

This section investigated a novel interaction regime, called counterfactual randomization, that allows the agent to actively account for human intuition during decision-making using counterfactual reasoning. Our analysis revealed that in almost all cases, a standard agent employing Fisherian randomization is constrained to sub-optimal behaviors; while a counterfactual agent is able to consistently achieve better performance. More generally, our results implied that human intuition should be kept ``in the loop'' as long as it has access to information about the tasks at hand, even after the agent completes its learning and builds a model of the environment. To resolve the tension between the autonomy and optimality of the system, we proposed a novel RL task subject to a budget constraint. Automated decision-making systems are playing an increasingly prominent role in society, and we hope this work constitutes a step towards a better understanding of the principles underlying human-machine interactions.

\section{Causal Imitation Learning (CRL Task 4)}\label{sec:_8_imitation}
Reinforcement Learning (RL) has been deployed and shown to perform exceptionally well in highly complex environments in the past decades \citep{sutton1998reinforcement,mnih2013playing,silver2016mastering,berner2019dota,kumar2022should}. One critical assumption behind many of the classical RL algorithms is that the reward function could be well-specified. In many real-world applications, however, it might be impractical to design a suitable reward function that evaluates each and every scenario \citep{randlov1998learning,ng1999policy}. For example, in the context of human driving, it is challenging to design a precise reward function, and experimenting in the environment could be ill-advised; still, watching expert drivers operate is usually feasible.

In the context of reinforcement learning, the \emph{imitation learning} (IL) paradigm investigates the problem of how an agent should behave and learn in an environment with an unknown reward function by observing demonstrations from a human expert \citep{argall2009survey,billard2008survey,hussein2017imitation,osa2018algorithmic}. 
Formally, a causal imitation learning task is characterized by the following signature.
\begin{align}
	\1T_{\text{imitate}} = \Tuple{\1I= \text{see}, \1A= \G, \Pi = \Braces{\Tuple{X_i, \*S_i}}_{i=1}^H, \1R = \emptyset}.
\end{align}
\begin{wrapfigure}[9]{r}{0.3\textwidth}
    \includegraphics[width=0.3\textwidth]{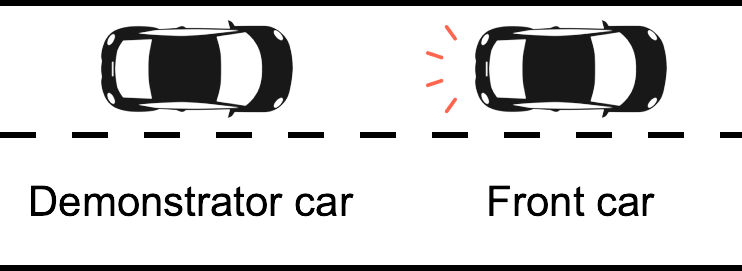}
    \caption{The tail light of the front car is unobserved in highway (aerial) drone data.}
    \label{fig:_8_car}
\end{wrapfigure}
This means that the agent will try to find a policy $\pi^*$ such that
\begin{align}\label{eq:opt-crl-cil}
	 & \pi^* = \argmax_{\pi \in \Pi_{\textsc{exp}}} \invEE{ \1R \left (\*Y \right ) \bigg \vert \G, \1D_{\text{obs}} \sim P(\*V), \textcolor{red}{\1R = \emptyset} }{\pi}{\1M^*}, 
\end{align}
the distinct feature of the task is that the reward function measuring the system's performance is not fully specified and is unknown from the learner's perspective.

An imitation learning agent attempts to learn a policy in space $\Pi$ from the demonstration data generated by a  demonstrator (e.g., a driver, a physician), following a different behavioral policy. For every episode $t = 1, \dots, T$, the agent observes the demonstrator operating in the underlying environment $\1M^*$, and receives trajectories $\*V^{(t)} \sim P(\*V)$ drawn from the observational distribution. It could also access certain structural assumptions of the environment, e.g., a causal diagram $\1A = \G$. 
Compared with off-policy learning and causal identification tasks, the departing point of imitation learning is that the reward function is not revealed to the learning agent and is not well-specified ($\1R = \emptyset$), posing a significant learning challenge.

\begin{corollary}\label{corol:_8_nonid}
	Let endogenous variables $\*X, \*Y \subseteq \*V$. If detailed parametrization of the reward function $\1R: \D(\*Y) \mapsto \3R$ is unknown, the expected reward $\invE{\1R(\*Y)}{\pi}$ for any policy $\pi$ over actions $\*X$ is not identifiable from a causal diagram $\G$.
\end{corollary}
The following example demonstrates non-identifiability posed by an unknown reward function.
\begin{example}\label{exp:_8_mab1}
For concreteness, consider a learning scenario depicted in Fig.~\ref{fig:_8_car}, describing trajectories of human-driven cars collected by drones flying over highways \citep{highDdataset,etesami2020causal}. Using such data, we want to learn a driving policy $\pi(x)$ deciding on the acceleration (action) $X$ of the following car to optimize the distance $Y$ between the following car and the front car. In reality, the human demonstrator's behaviors are affected by an unobserved noise $U$ representing the operational error. The driver's performance is evaluated by an unknown polynomial reward function $\1R(Y)$ over the car distance $Y$.

More specifically, consider an MAB environment $\1M^*$ described by the tuple
	\begin{align}
		\1M^* = \langle \*U = \{U\}, \*V = \{X, Y\}, \2F, P(\*U) \rangle \end{align}
where the structural functions $\2F$ are given by:
	\begin{align}\label{eq:_8_mab}
		\2F = \begin{cases}
			      X \gets U, \\
			      Y \gets X
		      \end{cases}
	\end{align}
$U \in \{0, 1\}$ is a binary variable drawn from the exogenous distribution $P(U = 1) = 0.9$. Due to the uncertainty of the reward function $\1R(Y)$, the expected reward $\invE{\1R(Y)}{\pi}$ for any driving policy $\pi(X)$ is not identifiable from the observational distribution $P(X, Y)$.

To make this argument more precise, let $\1R(Y) = \alpha Y$ be a linear function with an unknown real coefficient $\alpha \in \3R$. For any policy $\pi(X)$, the expected reward $\invE{\1R(Y)}{\pi}$ is given by
	\begin{align}
		\invE{\1R(Y)}{\pi} &= \alpha \invE{Y}{X\gets 0}\pi(X = 0)  + \alpha \invE{Y}{X\gets 1} \pi(X = 1)                 \\
		& = \alpha\pi(X = 1) \label{eq:_8_mab1}
	\end{align}
The last step holds since values of $Y$ are determined by $Y \gets X$. Note that every coefficient $\alpha \in \3R$ defines a unique expected reward $\invE{\1R(Y)}{\pi}$. Since $\alpha$ is not a parameter of the SCM $\1M^*$, changing values of $\alpha$ does not affect the evaluation of the observational distribution $P(X, Y)$. This means the expected reward $\invE{\1R(Y)}{\pi}$ is not uniquely discernible from the observational distribution $P(X, Y)$ in MAB models, i.e., $\invE{\1R(Y)}{\pi}$ is not identifiable if the reward function $\1R$ is unknown. \hfill $\blacksquare$
\end{example}
Corol.~\ref{corol:_8_nonid} implies that when the reward function $\1R$ is unknown, it is infeasible to uniquely determine the expected reward $\invE{\1R(\*Y)}{\pi}$ from the observational distribution $P(\*V)$ in the causal diagram $\G$. This precludes direct applications of causal identification approaches described in Sec.~\ref{sec:_4_3}, including do-calculus learning \citep{pearl:2k}, \texttt{Identify} algorithm \citep{tian:02}, and soft-do-calculus learning \citep{correa2020calculus}. To circumvent issues of non-identifiability, a common approach is to assume that the observed trajectories are generated by an ``expert'' demonstrator with satisfactory performance $\E\left[ \1R(\*Y)\right]$, e.g., no less than a certain threshold ($\E\left[ \1R(\*Y)\right] \geq \tau$). If we could find a policy $\pi$ that performs at least as well as the expert's policy, the agent's performance is also guaranteed to be satisfactory.
\begin{definition}\label{def:_8_ip}
	For a CDM $\Tuple{\1M^*, \Pi, \1R}$, an \emph{imitating policy} $\pi^*$ is a policy such that its expected reward is lower bounded by the expert's reward, i.e.,
	\begin{align}
		\underbrace{\invE{\1R(\*Y); \1M^*}{\pi^*}}_{\text{Agent's Performance}} \geq \underbrace{\E \left [\1R(\*Y); \1M^*\right]}_{\text{Expert's performance}}
	\end{align} \hfill $\blacksquare$
\end{definition}
In words, the right-hand side represents the expert's performance that the agent wants to achieve, while the left-hand side represents the real expected reward experienced by the agent evaluated in the underlying environment $\1M^*$.
We will call this the \textit{fundamental equation of the imitation learning problem}, which the agent aims to solve toward finding a policy $\pi^*$ that could perform at least as well as the demonstrating expert.  

The literature can be partitioned into two major learning modalities that realize imitation :
\begin{itemize}
	\item \emph{ behavioral cloning} (BC) \citep{widrow1964pattern,pomerleau1989alvinn,muller2006off,mulling2013learning,mahler2017learning}, and 
	\item \emph{inverse reinforcement learning} (IRL) \citep{ng2000algorithms,ziebart2008maximum,ho2016generative,fu2017learning}.
\end{itemize}
Specifically, BC methods attempt to directly mimic the expert's behavior policy by learning a mapping from the observed states to the expert's action via supervised learning. 
On the other hand, IRL methods first learn a surrogate reward function under which the expert's behavior policy is optimal. 
The imitator then obtains a policy using standard off-policy learning methods (see Sec.~\ref{sec:_4_1}) to maximize the learned reward function. 
Under some common assumptions, both BC and IRL can obtain policies that achieve the expert's performance \citep{ng2000algorithms,abbeel2004apprenticeship}. 
When additional parametric knowledge about the reward function is provided, IRL may produce a policy that outperforms the expert's in the underlying environment \citep{syed2008game, li2017infogail, yu2020intrinsic}.

Despite the performance guarantees provided by existing  BC and IRL methods, these are contingent on the assumption that the expert's input observations match those available to the imitator. 
On the other hand, when some expert's observed states remain latent to the imitator, unobserved confounders (UCs) are generally present in the demonstration data, violating the NUC assumption (Def.~\ref{def:_4_1_nuc}). Perhaps surprisingly, we will show later in this section, when the NUC does not hold, naively applying BC or IRL methods does not necessarily lead to satisfactory performance, even though the expert itself behaves optimally. After all, it is unclear how to perform imitation learning with unobserved confounding in the expert's demonstrations. This section answers this question and, more broadly, investigates the problem of imitation learning through causal lenses. We will provide novel algorithms capable of learning an imitating policy that performs at least as well as the expert from the demonstration data while allowing the presence of UCs. In particular, our contributions are summarized as follows.
\begin{itemize}
	\item \textbf{Confounding Robust BC. }Sec.~\ref{sec:_8_1} introduces a sufficient and necessary graphical criterion for determining the feasibility of BC-type learning procedure from demonstration data and qualitative knowledge about the data-generating process represented as a causal diagram. When such a condition holds, an imitating policy is obtainable using standard BC algorithms to achieve the expert's performance.
	\item \textbf{Confounding Robust IRL. }Sec.~\ref{sec:_8_2} derives a new graphical condition for deciding whether an imitating policy can be computed from the available data and knowledge, which provides a robust generalization of current IRL algorithms to general settings where the NUC assumption does not hold. These algorithms include GAIL \citep{ho2016generative}, and MWAL \citep{syed2008game}.
\end{itemize}

\subsection{Causal Behavioral Cloning} \label{sec:_8_1}
We first consider the behavioral cloning approach, where the agent attempts to learn an imitating policy by mimicking conditional observational distributions $P(X_i \mid \*Z_i)$ over domains of every action $\*X_i$ given a subset of input states $\*Z_i \subseteq \*S_i$. Formally
\begin{definition}[Behavioral Cloning Policy]\label{def:_8_1_bc}
	Let $\Pi = \Braces{\Tuple{X_i, \*S_i}}_{i = 1}^H$ be a policy space, and $P(\*V)$ be an observational distribution. A \emph{behavioral cloning policy} $\pi \in \Pi$ is an expression in terms of $P(\*V)$ such that for every $i = 1, \dots, H$, $\pi_i(X_i \mid \*Z_i) = P(X_i \mid \*Z_i)$ for some $\*Z_i \subseteq \*S_i$. \footnote{There could exist multiple behavioral policies in a policy space $\Pi$ simulating the same conditional distributions $P(X_i \mid \*Z_i)$. For instance, for $\Pi = \Braces{\Tuple{X, \emptyset}}$ and $P(X = 1) = 0.9$, behavioral policies $\pi, \pi' \in \Pi$ are given by $\pi(X = 1) = \I\{U \leq 0.9\}$ and $\pi'(X = 1) = \I\{U \geq 0.1\}$ where $U$ is an uniform distribution over $[0, 1]$.} \hfill $\blacksquare$
\end{definition}
There exist algorithms in imitation learning literature to perform behavioral cloning from observed demonstration data \citep{widrow1964pattern,pomerleau1989alvinn,muller2006off,mahler2017learning}. The following example illustrates BC learning in an MAB environment.
\begin{example}[Behavioral Cloning in MAB]\label{exp:_8_1_mab}
	Consider again the MAB environment $\1M^*$ described in Eq.~\ref{eq:_8_mab} concerning with learning a driving policy $\pi(X)$ following the front car. Since $\pi(X)$ belongs to a policy space $\Pi = \Braces{\Tuple{X, \emptyset}}$, a behavioral cloning policy $\pi_{\textsc{bc}}(X)$ is given by
	\begin{align}
		\pi_{\textsc{bc}}(X = 1) &= P(X = 1) \\
		&=P(U = 1)
	\end{align}
	Computing the above equation gives $\pi_{\textsc{bc}}(X = 1) = 0.9$. Recall the reward function $\1R(Y) \gets \alpha Y$ is a linear function with an unknown coefficient $\alpha$. Evaluating the expected reward $\1R(Y)$ in submodel $\1M^*_{\pi_{\textsc{bc}}}$ gives, following the decomposition in Eq.~\ref{eq:_8_mab1}:
	\begin{align}
		\invE{\1R(Y)}{\pi_{\textsc{bc}}} &= \alpha \pi_{\textsc{bc}}(X = 1)\\
		&= 0.9\alpha
	\end{align}
	We will next show that the BC policy $\pi_{\textsc{bc}}$ achieves the demonstrator's performance. By evaluating the expected reward $\1R(Y)$ in SCM $\1M^*$, we obtain
	\begin{align}
		\E\Brackets{\1R(Y)} &= \alpha \E\Brackets{Y}\\
		&=\alpha P(U = 1)\label{eq:_8_1_mab1}
	\end{align}
	Computing the above equation gives the evaluation of the demonstrator's performance $\E\Brackets{\1R(Y)} = 0.9\alpha$, which matches the performance of BC policy $\pi_{\textsc{bc}}$. \hfill $\blacksquare$
\end{example}
More generally, behavioral cloning is able to imitate the expert's performance when the NUC assumption holds (Def.~\ref{def:_4_1_nuc}) and the input states $\*S_i$ for every action $X_i$ are sufficiently large (to be defined). More precisely, provided with the NUC, for any policy $\pi \in \Pi$, the expected reward $\invE{\1R(\*Y)}{\pi}$ could be decomposed as, following the IPW identification formula (Thm.~\ref{thm:_4_1_ipw}),
\begin{align}
			\invE{\1R(\*Y); \1M^*}{\pi} = \sum_{\bar{\*x}_H, \bar{\*s}_H} E\left[\1R(\*Y) \mid \bar{\*x}_H, \bar{\*s}_H \right] P\left (\bar{\*x}_H, \bar{\*s}_H \right) \prod_{i = 1}^H \frac{\pi_{i} \left(x_i \mid \*s_i\right)}{P\left( x_i \mid \bar{\*x}_{i-1}, \bar{\*s}_i \right)}.
\end{align}
Let $\pi$ be a behavioral cloning policy in $\Pi$ such that $\pi_i(X_i \mid \*S_i) = P(X_i \mid \*S_i)$ for action $X_i$. The above equation could be written as
\begin{align}
			\invE{\1R(\*Y); \1M^*}{\pi} = \sum_{\bar{\*x}_H, \bar{\*s}_H} E\left[\1R(\*Y) \mid \bar{\*x}_H, \bar{\*s}_H \right] P\left (\bar{\*x}_H, \bar{\*s}_H \right) \prod_{i = 1}^H \frac{P \left(x_i \mid \*s_i\right)}{P\left( x_i \mid \bar{\*x}_{i-1}, \bar{\*s}_i \right)}. \label{eq:_8_1_ipw1}
\end{align}
Let the input states $\*S_i$ for every action $X_i$ be sufficiently large such that the following independence relationships hold in the observational distribution $P(\*V)$,
\begin{align}
	\Parens{X_i \ci \bar{\*X}_{i-1}, \bar{\*S}_{i-1}  \mid \*S_i} \;\forall i = 1, \dots, H. \label{eq:_8_1_markov}
\end{align}
One example for the above condition to hold is when states $\*S_i$ contain all observed parents $\*\PA_i$ for action $X_i$. Since the NUC holds, values of every $X_i$ are determined by independent noise $\*U_i$ given input states $\*S_i$. Eq.~\ref{eq:_8_1_ipw1} could be further written as
\begin{align}
	\invE{\1R(\*Y); \1M^*}{\pi} &= \sum_{\bar{\*x}_H, \bar{\*s}_H} E\left[\1R(\*Y) \mid \bar{\*x}_H, \bar{\*s}_H \right] P\left (\bar{\*x}_H, \bar{\*s}_H \right) \prod_{i = 1}^H \frac{P \left(x_i \mid \*s_i\right)}{P\left( x_i \mid \*s_i \right)}\\
	&=\sum_{\bar{\*x}_H, \bar{\*s}_H} E\left[\1R(\*Y) \mid \bar{\*x}_H, \bar{\*s}_H \right] P\left (\bar{\*x}_H, \bar{\*s}_H \right)\\
	&= \E \Brackets{\1R(\*Y)}
\end{align}
The last step follows by summing over states and actions $\bar{\*S}_H, \bar{\*X}_H$. In words, the behavior policy $\pi \in \Pi$ achieves the expert's performance. Formally, 
\begin{restatable}[Behavioral Cloning from NUC]{theorem}{thmbcnuc}\label{thm:_8_1_nuc}
	Let $\langle \1M^*, \Pi, \1R \rangle$ be a CDM where $\Pi = \left \{\langle X_i, \*S_i \rangle \right \}_{i = 1}^H$ and $\1R: \Omega(\*Y) \mapsto \3R$. Consider the following conditions:
	\begin{enumerate}
		\item The NUC condition (Def.~\ref{def:_4_1_nuc}) holds for the policy space $\Pi$ in SCM $\1M^*$;
		\item For every action $X_i$, $i = 1, \dots, H$, its endogenous parents $\*\PA_i \subseteq \*S_i$.
	\end{enumerate}
	Then, there is a behavioral cloning policy $\pi \in \Pi$ (Def.~\ref{def:_8_1_bc}), where the expected reward $\E_{\pi}\left[ \1R(\*Y) \right]$ matches the expert's performance evaluated in $\1M^*$,
	\begin{align}
		\invE{\1R(\*Y); \1M^*}{\pi} = \E \Brackets{\1R(\*Y); \1M^*}. 
	\end{align}
	Moreover, such a policy $\pi$ is given by $\pi_i(X_1 \mid \*S_i) = P(X_i\mid \*S_i)$ for every $i = 1, \dots, H$. \hfill $\blacksquare$
\end{restatable}
Example~\ref{exp:_8_1_mab} shows that behavioral cloning is able to achieve the expert's performance in an MAB environment. Our next example demonstrates behavioral cloning in the sequential setting.
\begin{example}[Behavioral Cloning in DTR]\label{exp:_8_1_dtr}
Consider the 2-stage DTR $\Tuple{\1M^*, \Pi, Y}$ where SCM $\1M^*$ is described by Eq.~\ref{eq:_3_1_dtr1} with coefficients $\alpha_1 = \alpha_2 = 0$; and the policy space $\Pi = \Braces{\Tuple{X_1, \braces{S_1}}, \Tuple{X_2, \braces{S_1, X_1, S_2}}}$. Evidently, Condition 1 of Thm.~\ref{thm:_8_1_nuc} is satisfied since NUC holds in this model (please revisit Example.~\ref{exp:_4_1_nuc1}). Also, Condition 2 of Thm.~\ref{thm:_8_1_nuc} holds since the endogenous parents of the actions $X_1, X_2$ are $\*{\PA}_1 = \{S_1\}$ and $\*{\PA}_2 = \{S_2\}$, respectively, which are both contained in the input states. Applying Thm.~\ref{thm:_8_1_nuc} implies the learner could achieve the expert's performance with a behavioral cloning policy $\pi = (\pi_1, \pi_2)$ given by
	\begin{align}
		 &\pi_1(X_1 \mid S_1) = P\Parens{X_1 \mid S_1}, &&\pi_2(X_2 \mid S_1, X_1, S_2) = P\Parens{X_2 \mid S_1, X_1, S_2}
	\end{align}
	Evaluating the above equation gives decision rules
	\begin{align}
		 & \pi_i: X_i \gets \I\{3S_i + U_i > 0\}, \;\; \forall i = 1, 2
	\end{align}
	where $U_i$, $i = 1, 2$, are independent variables drawn from a logistic distribution $\texttt{Logistic}(0, 1)$. It follows from Eq.~\ref{eq:_3_1_dtr1} submodel $\1M^*_{\pi}$ coincides with SCM $\1M^*$. As a consequence, we must have $\invE{Y}{\pi} = \E[Y]$, i.e., the learned policy achieves the expert's performance. \hfill $\blacksquare$
\end{example}
However, the behavioral cloning strategy does not always achieve the expert's performance in all environments, especially when the input variables of the imitator's and the expert's policy mismatch, and unobserved confounders generally exist in the demonstration data. Our next example illustrates the challenges of unobserved confounding.
\begin{example}[Behavioral Cloning fails without NUC]\label{exp:_8_1_mab2}
	Consider again the driving scenario described in Example~\ref{exp:_8_mab1}. Suppose now that the distance $Y$ between the demonstrator and font car is affected by the deceleration $U$ of the front car. Also, the human driver perceives the deceleration $U$ of the front car through its tail light and determines the action $X$. However, since the tail light is not recorded in the drone footage, variable $U$ is thus unobserved from the imitator's perspective. 
	
	More specifically, this environment is described by an SCM $\1M$ defined as 
	\begin{align}
		\1M = \langle \*U = \{U\}, \*V = \{X, Y\}, \2F, P(\*U) \rangle
	\end{align}
	where structural functions $\2F$ are given by
	\begin{align}\label{eq:_8_1_mab}
		\2F = \begin{cases}
			      X \gets \neg U, \\
			      Y \gets X \oplus U
		      \end{cases}
	\end{align}
	$U \in \{0, 1\}$ is a binary variable drawn from the exogenous distribution $P(U = 1) = 0.5$. Since $U$ is now an observed confounder affecting both action $X$ and outcome $Y$, the NUC condition does not hold in this environment $\1M$.
	
	Let the driver's performance be measured by a reward function $\1R(Y) = \alpha Y$ with an unknown coefficient $\alpha \in \3R$. Evaluating the expected reward in $\1M$ gives:
	\begin{align}
		\E \Brackets{\1R(Y)} &= \alpha \E\Brackets{Y}\\
		&= \alpha \E\Brackets{X \oplus U}\\
		&= \alpha \E\Brackets{\neg U \oplus U}
	\end{align}
The last step holds since $X \gets \neg U$ in SCM $\1M^*$. Computing the above equation gives the expert's performance $\E \Brackets{\1R(Y)} = \alpha$. 

We now apply the behavioral cloning strategy in Thm.~\ref{thm:_8_1_nuc} and see if it imitates the expert's performance even when the NUC does not hold. Mimicking the marginal distribution $P(X)$ results in a behavioral policy $\pi_{\textsc{bc}}(X)$ such that
\begin{align}
	\pi_{\textsc{bc}}(X = 1) &= P(X = 1)\\
	&=P(U = 1)
\end{align}
Computing the above equation gives $\pi_{\textsc{bc}}(X = 1) = 0.5$, i.e., the imitator randomly accelerates the demonstrator car. Evaluating the expected reward $\1R(Y)$ in submodel $\1M_{\pi_{\textsc{bc}}}$ implies
\begin{align}
	\invE{\1R(Y)}{\pi_{\textsc{bc}}} &= \sum_{x} \alpha \invE{Y}{x} \pi_{\textsc{bc}}(x)\\
	&= 0.5\alpha \Parens{\invE{Y}{X \gets 0}  + \invE{Y}{X\gets 1} }
\end{align}
Since values of $Y$ are given by $Y \gets X \oplus U$, we further have
\begin{align}
	\invE{\1R(Y)}{\pi_{\textsc{bc}}} &= 0.5\alpha \E[0 \oplus U + 1 \oplus U]\\
	&=0.5\alpha
\end{align}
Suppose the actual coefficient $\alpha > 0$ is positive. The imitator's performance $\invE{\1R(Y)}{\pi_{\textsc{bc}}} = 0.5\alpha$ is far from the expert's performance $\E \Brackets{\1R(Y)} = \alpha$. \hfill $\blacksquare$
\end{example}
In words, behavioral cloning does not guarantee to achieve the expert's performance when the NUC condition does not hold, which calls for alternative cloning strategies. We will next study a more generalized imitation setting from the expert's demonstration, provided with a causal diagram encoding the underlying qualitative knowledge about the environment.

\subsubsection{Backdoor Criterion for Imitation}
Our discussion starts with a variant of the sequential backdoor criterion (Def.~\ref{def:_4_3_sbc}) that allows the learner to imitate the expert's performance \citep{zhang2020causal,kumor2021causal}. Recall that for any policy $\pi \in \Pi$, $\G_{\pi_{i+1}, \dots, \pi_H}$, $i = 0, \dots, H-1$, is a manipulated graph obtained from the causal diagram $\G$ by replacing incoming arrows of every action node $X_j \in \left \{X_{i+1}, \dots, X_H \right \}$, with arrows from input states in $\*S_j$ to $X_j$. 
In the context of imitation learning, $\G_{\pi_{i+1}, \dots, \pi_H}$ can be seen as $\1G$ with all future actions after the $i$-th stage of the intervention is already encoded in the graph. Formally, the imitation backdoor criterion is defined as follows: 
\begin{definition}[Imitation Backdoor Condition]\label{def:_8_1_ibc}
	Let $\G$ be a causal diagram and $\*X, \*Y \in \*V$ be subsets of variables. A policy space $\Pi = \Braces{\Tuple{X_i, \*S_i}}_{i = 1}^H$ is said to satisfy the \emph{imitation backdoor condition} w.r.t. $\*Y$ in $\G$ (for short, $\Pi$ is imitation admissible) if for every policy $\pi \in \Pi$, every action $X_i \in \*X$, one of the following conditions hold:
	\begin{enumerate}
		\item $X_i$ is not an ancestor of $\*Y$ in $\G_{\pi_{i+1}, \dots, \pi_H}$, i.e., $X \not \in \An(\*Y)_{\G_{\pi_{i+1}, \dots, \pi_H}}$;
		\item $\*S_i$ $d$-separates all backdoor path from node $X_i$ to nodes in $\*Y$ in $\G_{\pi_{i+1}, \dots, \pi_H}$, i.e., $(\*Y \ci X_i | \*S_i)$ in $\G_{\underline{X_i},\pi_{i+1}, \dots, \pi_H}$.  \hfill $\blacksquare$
	\end{enumerate}
\end{definition}
The first condition in Def.~\ref{def:_8_1_ibc} corresponds to the case where an action at $X_i$ does not affect the value of $\*Y$ once future actions are taken. 
Since $\G_{\pi_{i+1}, \dots, \pi_H}$ has modified parents for future actions $\bar{\*X}_{i+1:H}$, the value of $X_i$ might no longer be relevant at all to $Y$, i.e. $Y$ would get the same input distribution no matter what policy is chosen for $X_i$. 
This allows $X_i$ to fail Condition (2), meaning that it is not clonable by itself, but still be part of a clonable set $\*X$, because future actions can shield $Y$ from errors made at $X_i$. 
The second condition is similar to the backdoor criterion where $\*Z_i$ is a set of variables that effectively encodes all information relevant to imitating $X_i$ with respect to $Y$. 
In other words, if the joint distribution $P(\*Z_i, X_i)$ over the observed states $\*Z_i$ and action $X_i$ matches when both expert and imitator are acting, then an adversarial reward function $Y$ cannot distinguish between the two and imitation could be successfully realized.
\begin{figure}
	\begin{subfigure}{0.32\linewidth}\centering
		\begin{tikzpicture}
			\def\outerr{3.5}
			\def\innerr{3}
			\node[vertex] (Z1) at (0.5, 1) {Z};
			\node[vertex] (X1) at (0, 0) {X\textsubscript{1}};
			\node[vertex] (Z2) at (1, 0) {W};
			\node[vertex] (X2) at (2, 0) {X\textsubscript{2}};
			\node[vertex] (Y) at (3, 0) {Y};

			\draw[dir] (Z1) to (Y);

			\draw[dir] (X1) to (Z2);

			\draw[dir] (Z2) to (X2);

			\draw[bidir] (Z2) to [bend left = 30] (X2);

			\draw[dir] (X2) to (Y);

			\draw[dir] (Z1) to (Z2);
			\draw[bidir]  (Z1) to [bend right = 30] (X1);

			\begin{pgfonlayer}{back}
				\node[circle,fill=betterblue!65,draw=none,minimum size=2*\innerr mm] at (X1) {};
				\node[circle,fill=betterblue!65,draw=none,minimum size=2*\innerr mm] at (X2) {};
				\node[circle,fill=betterred!65,draw=none,minimum size=2*\innerr mm] at (Y) {};
			\end{pgfonlayer}
		\end{tikzpicture}
		\caption{}
		\label{fig:_8_1_ibc1a}
	\end{subfigure}\hfill
	\begin{subfigure}{0.32\linewidth}\centering
		\begin{tikzpicture}
			\def\outerr{3.5}
			\def\innerr{3}
			\node[vertex] (Z1) at (0.5, 1) {Z};
			\node[vertex] (X1) at (0, 0) {X\textsubscript{1}};
			\node[vertex] (Z2) at (1, 0) {W};
			\node[vertex] (X2) at (2, 0) {X\textsubscript{2}};
			\node[vertex] (Y) at (3, 0) {Y};

			\draw[dir] (Z1) to (Y);

			\draw[dir] (X1) to (Z2);

			\draw[dir, betterblue] (Z1) to (X2);

			\draw[dir] (X2) to (Y);

			\draw[dir] (Z1) to (Z2);
			\draw[bidir]  (Z1) to [bend right = 30] (X1);
			\begin{pgfonlayer}{back}
				\node[circle,fill=betterblue!65,draw=none,minimum size=2*\innerr mm] at (X1) {};
				\node[circle,fill=betterblue!65,draw=none,minimum size=2*\innerr mm] at (X2) {};
				\node[circle,fill=betterred!65,draw=none,minimum size=2*\innerr mm] at (Y) {};
			\end{pgfonlayer}
		\end{tikzpicture}
		\caption{}
		\label{fig:_8_1_ibc1b}
	\end{subfigure}\hfill
	\begin{subfigure}{0.32\linewidth}\centering
		\begin{tikzpicture}
			\def\outerr{3.5}
			\def\innerr{3}
			\node[vertex] (Z1) at (0.5, 1) {Z};
			\node[vertex] (X1) at (0, 0) {X\textsubscript{1}};
			\node[vertex] (Z2) at (1, 0) {W};
			\node[vertex] (X2) at (2, 0) {X\textsubscript{2}};
			\node[vertex] (Y) at (3, 0) {Y};

			\draw[dir] (Z1) to (Y);

			\draw[dir] (X1) to (Z2);

			\draw[dir, betterblue] (Z2) to (X2);

			\draw[dir] (X2) to (Y);

			\draw[dir] (Z1) to (Z2);
			\draw[bidir]  (Z1) to [bend right = 30] (X1);
			\begin{pgfonlayer}{back}
				\node[circle,fill=betterblue!65,draw=none,minimum size=2*\innerr mm] at (X1) {};
				\node[circle,fill=betterblue!65,draw=none,minimum size=2*\innerr mm] at (X2) {};
				\node[circle,fill=betterred!65,draw=none,minimum size=2*\innerr mm] at (Y) {};
			\end{pgfonlayer}
		\end{tikzpicture}
		\caption{}
		\label{fig:_8_1_ibc1c}
	\end{subfigure}\hfill\null
	\caption{A causal diagram and its manipulated subgraphs.}
	\label{fig:_8_1_ibc1}
\end{figure}
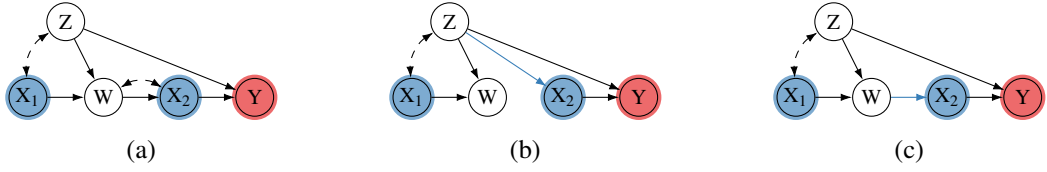
\begin{example}[Imitation Backdoor $\not \Rightarrow$ Sequential Backdoor]\label{exp:_8_1_ibc1}
We will illustrate the distinction between Conditions (1) and (2) in the causal diagram $\G$ of Fig.~\ref{fig:_8_1_ibc1a}. Consider a policy space 
\begin{align}
\Pi_1= \left \{ \langle X_1, \emptyset \rangle, \langle X_2, \{Z\} \rangle \right\}. 
\end{align}
For every policy $(\pi_1, \pi_2) \in \Pi_1$, the manipulated diagram $\G_{\pi_2}$ is shown in Fig.~\ref{fig:_8_1_ibc1b}. As for action $X_1$, since its input states $\*S_1 = \emptyset$, there is no valid adjustment set that can $d$-separate $X_1$ from $Y$. However, since the policy for action $X_2$ uses $Z$ as input instead of $W$ or $X_1$ (i.e. $\pi_{2}(X_2 \mid Z)$), $X_1$ will no longer be an ancestor of $Y$ in $\G_{\pi_2}$ and Condition (1) holds. 
In effect, the action made at $X_2$ ignores the mistakes made at $X_1$ due to not having access to unobserved confounders when taking action. Conditioning on covariate node $Z$ $d$-separates all backdoor paths between $X_2$ and $Y$ in the subgraph $\G$, satisfying Condition (2). Therefore, the policy space $\Pi_1$ is imitation admissible w.r.t. the reward signal $Y$ in the causal diagram $\G$. \hfill $\blacksquare$
\end{example}
An interesting observation follows from the above example. While $\Pi_1= \left \{ \langle X_1, \emptyset \rangle, \langle X_2, \{Z\} \rangle \right\}$ is imitation admissible in Fig.~\ref{fig:_8_1_ibc1a}, $\Pi_1$ does not satisfy the sequential backdoor condition of Def.~\ref{def:_4_3_sbc}. This is the case since the input state $\*S_1 = \emptyset$ fails to block the backdoor path between $X_1$ and $Y$ via covariate $Z$. On the other hand, in some settings, there exist policy spaces satisfying the sequential backdoor condition (Def.~\ref{def:_4_3_sbc}) but are not imitation admissible.
\begin{example}[Sequential Backdoor $\not \Rightarrow$ Imitation Backdoor]\label{exp:_8_1_ibc2}
	Consider the causal diagram $\G$ described in Fig.~\ref{fig:_8_1_ibc1a} and a policy space $\Pi_2: \left \{ \langle X_1, \{Z\} \rangle, \langle X_2, \{W\} \rangle \right\}$. For every policy $\Parens{\pi_1, \pi_2} \in \Pi_2$, the manipulated diagram $\G_{\pi_2}$ is shown in Fig.~\ref{fig:_8_1_ibc1c}. As for action $X_1$, Condition (2) holds since policy $\pi_1(X_1\mid Z)$ for $X_1$ takes $Z$ as input, and conditioning on $Z$ $d$-separates all backdoor paths from $X_1$ to reward $Y$ in $\G_{\pi_2}$. As for action $X_2$, Condition (1) does not hold since $X_2$ is a direct parent of $Y$. Condition (2) fails to apply since input variable $W$ is a collider and conditioning on $W$ opens the backdoor path between $X$ and $Y$ in $\G$, e.g., $X_2 \bidirectarrow W \leftarrow Z \rightarrow Y$. 
Still, conditioning on all past actions and states' history $X_1, Z, W$ $ d$-separates backdoor paths from $X_2$ to $Y$ in $\G$. This means $\Pi_2$ satisfies the sequential backdoor condition of Def.~\ref{def:_4_3_sbc} and the expected reward of policy $\pi \in \Pi_2$ is identifiable from $P(X_1, Z, W, X_2, Y)$. \hfill $\blacksquare$
\end{example}
The imitation backdoor condition provides an effective algorithm for deciding whether a policy compatible with a policy space in a causal diagram is imitable or not. Next, we describe some necessary notations.
\begin{definition}[Policy Subspace]\label{def:_8_1_subspace}
	For a policy space $\Pi = \left \{ \langle X_1, \*S_i \rangle  \right\}_{i = 1}^H$, a policy subspace $\Pi'$ of $\Pi$, denoted by $\Pi' \subseteq \Pi$, is a sequence $\left \{ \langle X_i, \*Z_i \rangle  \right\}_{i = 1}^H$ where $\*Z_i \subseteq \*S_i$ for every action $X_i \in \*X$. \hfill $\blacksquare$
\end{definition}

In words, every policy space $\Pi$ is also a subspace of itself. A subspace $\Pi'$ contained in $\Pi$ is \emph{proper} if $\Pi' \neq \Pi$ is not equal to $\Pi$, which we denote by $\Pi' \subset \Pi$. Note that for every policy $\pi' \in \Pi'$ compatible with a subspace $\Pi'$, one could always simulate it using a policy $\pi \in \Pi$ such that $\pi_i(x_i \mid \*s_i) = \pi'_i(x_i \mid \*z_i)$, $i = 1, \dots, H$, for all realizations $x_i, \*s_i, \*z_i$. That is, input states in the set difference $\*S_i \setminus \*Z_i$ do not affect values of action $X_i$. It follows that policy $\pi' \in \Pi$ is also compatible with space $\Pi$ if it is compatible with a subspace $\Pi' \subseteq \Pi$.
\begin{restatable}[Behavioral Cloning from Imitation Backdoor]{theorem}{thmibc}\label{thm:_8_1_ibc}
	Let $\G$ be a causal diagram, $\Pi$ be a policy space over actions $\*X$, and $\*Y \subseteq \*V$ be a subset of variables. If there exists a subspace $\Pi' = \left \{ \langle X_i, \*Z_i \rangle  \right\}_{i = 1}^H$ contained in $\Pi$ such that $\Pi'$ is imitation admissible w.r.t. $\*Y$ in $\1G$, then there is a behavior cloning policy $\pi \in \Pi'$ such that for any SCM $\1M^*$ compatible with diagram $\G$,
	\begin{align}
		\invE{\1R(\*Y); \1M^*}{\pi} = \E \Brackets{\1R(\*Y); \1M^*}. 
	\end{align}
	 Moreover, such a policy $\pi \in \Pi'$ is given by $\pi_i\left(X_i | \*Z_i \right) = P\left(X_i | \*Z_i \right)$ for every $i = 1, \dots, H$. \hfill $\blacksquare$
\end{restatable}
Thm.~\ref{thm:_8_1_ibc} implies that whenever an imitation admissible subspace $\Pi' \subseteq \Pi$ is found, the expert's performance is achievable using behavioral cloning, i.e., mimicking the conditional distribution $P(X_i|\*Z_i)$ for every action $X_i \in \*X$. Moreover, it has been shown that the imitation backdoor criterion is also necessary for determining the feasibility of behavioral cloning for a general class of policy spaces such that for every action $X_i$, the input states $\*S_i$ contain all variables preceding $X_i$ following a temporal ordering in the diagram $\G$ \citep{zhang2020causal,kumor2021causal}. \footnote{Indeed, this is the largest possible policy space defined over action $\*X$ in an SCM.} That is, if there is no imitation admissible subspace in $\Pi$, then for any behavioral cloning policy $\pi \in \Pi$, one could always construct an SCM $\1M$ compatible with the causal diagram $\G$ such the BC policy $\pi$ fails to achieve the expert's performance.  
\begin{example}\label{exp:_8_1_ibc3}
	Consider again the causal diagram $\G$ described in  Fig.~\ref{fig:_8_1_ibc1a} and a policy space $\Pi = \left \{ \langle X_1, \{Z\} \rangle, \langle X_2, \{Z, W\}\rangle \right\}$. Note that $\Pi_1= \left \{ \langle X_1, \emptyset \rangle, \langle X_2, \{Z\} \rangle \right\}$ is a subspace contained $\Pi$ and, as discussed previously, is imitation admissible in diagram $\G$. It follows from Thm.~\ref{thm:_8_1_ibc} that the expert's performance $\E\Brackets{\1R(Y)}$ is achievable from observational data using a behavioral cloning policy $\pi = \left ( \pi_1, \pi_2 \right)$ given by $\pi_1(X_1) = P(X_1)$ and $\pi_2(X_2 \mid Z) = P(X_2 \mid Z)$. \hfill $\blacksquare$
\end{example}
For every action $X_i \in \*X$, the imitation backdoor criterion requires that the covariates $\*Z_{i}$ is a back-door adjustment set in the manipulated diagram $\G_{\pi_{i+1}, \dots, \pi_H}$. There exist efficient methods for finding adjustment sets in the literature \citep{van2020finding}.
The learner could run these algorithms on each action $X_i$ iteratively to find each backdoor admissible set $\*Z_i$ following a reverse topological ordering $X_H \succ X_{H-1}  \succ  \dots  \succ  X_1$, which will lead to an imitation admissible subspace $\Pi' = \left \{ \langle X_i, \*Z_i \rangle \right\}_{X_i \in \*X}$ in the end. When the state variables $\*S_i$, $i = 1, \dots, H$, contain all variables preceding every action $X_i$ following a topological ordering in the diagram $\G$, \citep{kumor2021causal} provides a polynomial-time algorithm to find an imitation admissible subspace $\Pi'$. This means that a policy subspace $\Pi'$ satisfying the imitation backdoor condition in Def.~\ref{def:_8_1_ibc} is generally easier to obtain if it exists. 
If that is the case, an imitating policy could be obtained from demonstrations by ``cloning'' the expert's nominal policy, following standard behavioral cloning algorithms \citep{widrow1964pattern,pomerleau1989alvinn}.
\begin{table}[t]
	\centering
	\begin{tabular}[t]{c @{\hspace{.5\tabcolsep}} c | c | c | c}
		\toprule
		\#     & Causal Diagram   & Causal BC                        & BC -- Observed Parents           & BC -- All Observed \\
		\midrule
		$\G_1$ &
		\begin{minipage}{2.8cm}
			\centering
			\begin{tikzpicture}
				\def\outerr{3.5}
				\def\innerr{3}
				\node[vertex] (x1) at (0,0) {$X_1$};
				\node[vertex] (x2) at (1,0) {$X_2$};
				\node[vertex] (y) at (2,0) {$Y$};

				\node[vertex] (z) at (1,1) {$Z$};

				\draw[dir] (x1) to (x2);
				\draw[dir] (x2) to (y);

				\draw[bidir] (z) to [bend left=30] (y);
				\draw[bidir] (z) to [bend right=30] (x1);
				\draw[bidir] (z) to [bend right=50] (x2);

				\begin{pgfonlayer}{back}
					\draw[fill=betterblue!25, draw = betterblue!45] \convexpath{x2, x1, z}{1.2*\outerr mm};

					\draw[fill=betterblue!25, draw = betterblue!45] \convexpath{x1, z}{\outerr mm};
					\node[circle,fill=betterblue!65,draw=none,minimum size=2*\innerr mm] at (x1) {};
					\node[circle,fill=betterblue!65,draw=none,minimum size=2*\innerr mm] at (x2) {};
					\node[circle,fill=betterred!65,draw=none,minimum size=2*\innerr mm] at (y) {};
				\end{pgfonlayer}
			\end{tikzpicture}
		\end{minipage}
		\vspace{0.1cm}
		       & $0.04\pm0.04\%$  & $0.05\pm0.04\%$                  & {\color{red}$\*{0.13\pm0.18\%}$}                      \\         \midrule
		$\G_2$ &
		\begin{minipage}{2.8cm}
			\centering
			\begin{tikzpicture}
				\def\outerr{3.5}
				\def\innerr{3}
				\node[vertex] (x1) at (0,0) {$X_1$};
				\node[vertex] (x2) at (1,0) {$X_2$};
				\node[vertex] (y) at (2,0) {$Y$};

				\node[vertex] (z) at (1,1) {$Z$};

				\draw[dir] (x1) to (x2);
				\draw[dir] (x2) to (y);

				\draw[dir] (z) to [bend left=30] (y);
				\draw[bidir] (z) to [bend right=30] (x1);

				\begin{pgfonlayer}{back}
					\draw[fill=betterblue!25, draw = betterblue!45] \convexpath{x2, x1, z}{1.2*\outerr mm};

					\draw[fill=betterblue!25, draw = betterblue!45] \convexpath{x1, z}{\outerr mm};
					\node[circle,fill=betterblue!65,draw=none,minimum size=2*\innerr mm] at (x1) {};
					\node[circle,fill=betterblue!65,draw=none,minimum size=2*\innerr mm] at (x2) {};
					\node[circle,fill=betterred!65,draw=none,minimum size=2*\innerr mm] at (y) {};
				\end{pgfonlayer}
			\end{tikzpicture}
		\end{minipage}
		\vspace{0.1cm}
		       & $0.05\pm 0.03\%$ & {\color{red}$\*{0.20\pm0.25\%}$} & $0.05\pm 0.03\%$                                      \\         \midrule
		$\G_3$ &
		\begin{minipage}{2.8cm}
			\centering
			\begin{tikzpicture}
				\def\outerr{3.5}
				\def\innerr{3}
				\node[vertex] (x1) at (0,0) {$X_1$};
				\node[vertex] (x2) at (1,0) {$X_2$};
				\node[vertex] (y) at (2,0) {$Y$};

				\node[vertex] (z) at (1,1) {$Z$};

				\draw[dir] (x1) to (x2);
				\draw[dir] (x2) to (y);

				\draw[dir] (z) to [bend left=30] (y);
				\draw[bidir] (z) to [bend right=30] (x1);

				\begin{pgfonlayer}{back}
					\draw[fill=betterblue!25, draw = betterblue!45] \convexpath{x2, x1, z}{1.2*\outerr mm};
					\node[circle,fill=betterblue!25,draw=betterblue!45,minimum size=2*\outerr mm] at (X1) {};

					\node[circle,fill=betterblue!65,draw=none,minimum size=2*\innerr mm] at (x1) {};
					\node[circle,fill=betterblue!65,draw=none,minimum size=2*\innerr mm] at (x2) {};
					\node[circle,fill=betterred!65,draw=none,minimum size=2*\innerr mm] at (y) {};
				\end{pgfonlayer}
			\end{tikzpicture}
		\end{minipage}
		\vspace{0.1cm}
		       & $0.04\pm 0.03\%$ & {\color{red}$\*{0.27\pm0.40\%}$} & {\color{red}$\*{0.26\pm0.39\%}$}                      \\         \midrule
		$\G_4$ &
		\begin{minipage}{2.8cm}
			\centering
			\begin{tikzpicture}
				\def\outerr{3.5}
				\def\innerr{3}
				\node[vertex] (x1) at (0,0) {$X_1$};
				\node[vertex] (x2) at (1,0) {$X_2$};
				\node[vertex] (y) at (2,0) {$Y$};

				\node[vertex] (z) at (1,1) {$Z$};

				\draw[dir] (x1) to [bend right = 30] (y);
				\draw[dir] (x2) to (y);
				\draw[dir] (z) to (x2);

				\draw[bidir] (z) to [bend left=30] (y);
				\draw[bidir] (z) to [bend right=30] (x1);

				\begin{pgfonlayer}{back}
					\draw[fill=betterblue!25, draw = betterblue!45] \convexpath{x2, x1, z}{1.2*\outerr mm};

					\node[circle,fill=betterblue!25,draw=betterblue!45,minimum size=2*\outerr mm] at (X1) {};
					\node[circle,fill=betterblue!65,draw=none,minimum size=2*\innerr mm] at (x1) {};
					\node[circle,fill=betterblue!65,draw=none,minimum size=2*\innerr mm] at (x2) {};
					\node[circle,fill=betterred!65,draw=none,minimum size=2*\innerr mm] at (y) {};
				\end{pgfonlayer}
			\end{tikzpicture}
		\end{minipage}
		\vspace{0.1cm}
		       & Not Imitable     & {\color{red}$\*{0.19\pm0.29\%}$} & {\color{red}$\*{0.19\pm0.29\%}$}                      \\ 
		\bottomrule
		\\
	\end{tabular}

	\caption{Performance gap $|\invE{Y}{\pi}- \E[Y]|$ from behavioral cloning using different input states in randomly sampled SCMs consistent with each causal diagram.}
	\label{tab:_7_1_bc}
\end{table}
\begin{experiment}\label{exp:_8_1_bc}
	We evaluate BC algorithms in randomly sampled SCMs consistent with various causal diagrams. These algorithms use different criteria to select input states/features for every action, which we summarize as follows. (1) Our proposed causal BC selects input states using the imitation backdoor condition, following the procedure described in Thm.~\ref{thm:_8_1_ibc}. (2) Standard BC algorithm mimics the expert's nominal policy, taking observed direct parents for every action as input. (3) Standard BC algorithm considers all state variables available to the imitator at the time of each action, described by the policy space.

	For each causal diagram, $10,000$ random discrete causal models are sampled, the expert's performance is measured, and then the expert's policy is replaced with imitating policies $\pi_i(X_i \mid \*Z_i) =P(X_i|\*Z_i)$ for every action $X_i \in \*X$, with input covariates $\*Z_i$ determined by the tested BC algorithm described above. The performance of algorithms is evaluated using the gap between the expert's performance $\E[Y]$ and the expected reward of the policy $\invE{Y}{\pi}$ obtained by the imitator. Simulation results are shown in Table \ref{tab:_7_1_bc}, with causal diagrams and policy spaces described in the first column, followed by the performance gap between the expert and the imitator.

	For the diagram $\G_1$, including $Z$ when developing a policy for $X_1, X_2$ leads to a biased answer, which makes the average error of using all observed covariates (red) larger than just the sampling fluctuations present in the other columns.
	Similarly, $Z$ needs to be considered in $\G_2$, but it is not explicitly used by $X_2$, so a method relying only on observed parents leads to bias here. In $\G_3$, $Z$ is not observed at the time of determining action $X_1$, making standard BC algorithms fail to achieve the expert's performance. Our method recognizes that $X_2$'s policy can fix the imitation error made at $X_1$, and is the only method that leads to an unbiased result. Finally, in $\G_4$, the non-causal approaches cannot determine non-clonability, and return biased results in all such cases. \hfill $\blacksquare$
\end{experiment}
\subsection{Causal Inverse RL} \label{sec:_8_2}
This section studies an alternative imitation learning strategy, called \emph{inverse reinforcement learning} (IRL, \cite{ng2000algorithms,ziebart2008maximum,ho2016generative,fu2017learning}), in a CDM $\Tuple{\1M^*, \Pi, \1R}$. Similar to behavioral cloning, detailed parametrizations of the underlying environment $\1M^*$ and the reward function $\1R$ are not fully unknown. However, the imitator now has access to a parametric family $\2R \subseteq \left \{\forall \1R: \D(\*Y) \mapsto \3R \right\}$ containing the actual reward function $\1R$. We will start the discussion by describing an IRL strategy when the NUC assumption (Def.~\ref{def:_4_1_nuc}) holds. We will next relax the NUC and study inverse RL in a more general class of causal models, provided with a causal diagram $\G$ encoding its qualitative knowledge.

Following the game-theoretic approach introduced in \citep{syed2008game}, we formulate imitating learning (Def.~\ref{def:_8_ip}) as learning to play a two-player zero-sum game in which the agent chooses a policy, and the adversarial (e.g., Nature) chooses a worst-case reward function from the parametric reward family $\2R$. Now consider the optimization problem defined as follows.
\begin{align}
	\nu^* = \min_{\pi \in \Pi} \max_{\1R \in \2R} \E\left[ \1R(\*Y); \1M^* \right ] - \invE{\1R(\*Y);\1M^*}{\pi}. \label{eq:_8_2_opt}
\end{align}
The inner maximization in the above equation can be viewed as a \emph{causal IRL} step where we attempt to ``guess'' a worst-case reward function $\hat{\1R} \in \2R$ that prioritizes the expert's policy. That is, the gap in the performance between the expert's and the imitator's policies is maximized. Meanwhile, note that the expert's reward $\E\left[ \1R(\*Y); \1M^* \right ]$ is not affected by the imitating's policy $\pi$. The outer minimization is equivalent to a planning step that finds a policy $\pi^*$ optimizing a CDM $\Tuple{\1M^*, \Pi, \hat{\1R}}$ with the worst-case reward $\hat{\1R}$. Obviously, the solution $\pi^*$ is an imitating policy if the performance gap $\nu^* = 0$. 
In cases where the expert is sub-optimal, we may have $\nu^* < 0$, i.e., 
\begin{align}
\E\left[ \hat{\1R}(\*Y); \1M^* \right ]< \invE{\hat{\1R}(\*Y); \1M^*}{\pi}, \; \exists \pi \in \Pi
\end{align} 
In words, the solution $\pi^*$ will dominate the expert's policy $f_{\*X}$ in the worst-case scenario, regardless of the detailed form of the reward function $\1R$. To some extent, the imitating policy $\pi^*$ ignores the sub-optimal expert and instead exploits prior knowledge about the unknown reward function. When the prior knowledge is informative, solving the optimization program in Eq.~\ref{eq:_8_2_opt} could produce a policy that could significantly outperform the expert in the underlying environment with respect to the unknown reward function, while at the same time guaranteed to be no worse.
\begin{example}[Inverse RL in MAB]\label{exp:_8_2_mab}
Consider again the MAB environment $\1M^*$ described in Eq.~\ref{eq:_8_mab} concerning learning a driving policy from highway footage. Suppose that the reward function $\1R(Y) = \alpha Y$ is linear with a positive coefficient $\alpha > 0$. The minimax program in Eq.~\ref{eq:_8_2_opt} could be written as
\begin{align}
	\nu^* &= \min_{\pi(X)} \max_{\alpha > 0} \alpha \E\left[ Y \right ] - \alpha  \invE{Y}{\pi} \label{eq:_8_2_mab1}
\end{align}
Evaluating the distance from the front car $Y$ in the environment $\1M^*$ gives:
\begin{align}
	\E[Y] &= P(X = 1) = 0.9 \label{eq:_8_2_mab2}
\end{align}
Note that the NUC condition holds in this environment. The interventional quantity $\invE{Y}{\pi}$ is a function of the observational distribution $P(X, Y)$ and policy $\pi(x)$ is given by, following the DP formula in Thm.~\ref{thm:_4_1_dp} (or the IPW formula in Thm.~\ref{thm:_4_1_ipw}),
\begin{align}
	\invE{Y}{\pi} &=  \E[Y \mid X = 0] \pi(X = 0) + \E[Y \mid X = 1] \pi(X = 1)\\
	&= \pi(X = 1) \label{eq:_8_2_mab3}
\end{align}
The last step holds since values of $Y$ are determined by $Y \gets X$. By substituting Eqs.~\ref{eq:_8_2_mab2} and \ref{eq:_8_2_mab3} into Eq.~\ref{eq:_8_2_mab1}, we can further write the performance gap $\nu^*$ as:
\begin{align}
	\nu^* &= \min_{\pi(X)} \max_{\alpha > 0} \alpha \Parens{0.9 - \pi(X = 1)}
\end{align} 
For any coefficient $\alpha >0$, the above program is minimized with a solution $\pi(X = 1) = 1$. Solving the above equation thus leads to an IRL policy $\pi_{\textsc{irl}}: X \gets 1$. In this case, the performance gap is equal to $\nu^* = -0.1\alpha < 0$, which means that the IRL imitator outperforms the expert.

To verify this intuition, we evaluate the expected reward $\1R(Y)$ in submodel $\1M^*_{\pi_{\textsc{irl}}}$, following the evaluation formula in Eq.~\ref{eq:_8_mab1},
\begin{align}
	\invE{\1R(Y)}{\pi_{\textsc{irl}}} = \alpha \pi_{\textsc{irl}}(X = 1) 
\end{align}
This means that the IRL policy achieves the expected reward $\invE{\1R(Y)}{\pi_{\textsc{irl}}} = \alpha$, which outperforms both the expert and BC's policies $\E\Brackets{\1R(Y)} =  \invE{\1R(Y)}{\pi_{\textsc{bc}}} = 0.9\alpha$ (see Example~\ref{exp:_8_1_mab} for detailed computations). \hfill $\blacksquare$
\end{example}
Despite its clear semantics, solving the optimization problem in Eq.~\ref{eq:_8_2_opt} requires the detailed parametrization of the underlying SCM $\1M^*$, which is not accessible to the agent in most real-world settings. It is then important to study conditions under which the solution of Eq.~\ref{eq:_8_2_opt} is identifiable and could be formulated from the observational distribution $P(\*V)$. Fix a reward function $\1R \in \2R$. First, the expect's performance $\E\left[ \1R(\*Y); \1M^* \right ]$ is obtainable from $P(\*V)$ by computing the arithmetic mean of $\1R(\*Y)$ weighted by the marginal distribution $P(\*Y)$. If the NUC assumption (Def.~\ref{def:_4_1_nuc}) holds, the imitator's performance $\invE{\1R(\*Y);\1M^*}{\pi}$ is computable from the observational distribution $P(\*V)$, following off-policy learning algorithms including IPW (Thm.~\ref{thm:_4_1_ipw}) and DP (Thm.~\ref{thm:_4_1_dp}). The imitator could then formulate the minimax program in Eq.~\ref{eq:_8_2_opt} from the observational distribution $P(\*V)$, the hypothesis reward class $\2R$, and the policy space $\Pi$. Solving this optimization program leads to an imitating policy. We demonstrate in Example~\ref{exp:_8_2_mab} the IRL strategy under NUC in an MAB environment.

On the other hand, however, when the input variables of the expert and the imitator's policies mismatch, and unobserved confounders generally exist, performing IRL with the standard off-policy evaluation does not necessarily lead to an imitating policy achieving the expert's performance. The following example illustrates the challenges of unobserved confounders for IRL methods. 
\begin{example}[Inverse RL fails without NUC]\label{exp:_8_2_mab2}
	Consider the alternative MAB environment $\1M$ described in Eq.~\ref{eq:_8_1_mab} where the front car's deceleration $U$ is an unobserved confounder affecting both the human demonstrator's action $X$ and the distance $Y$ between the demonstrator and the front car; so the NUC does not hold in this environment. 
	
	Evaluating the expected value of $Y$ in this MAB environment $\1M$ gives
	\begin{align}
		\E[Y] &= \E[X \oplus U]\\
		&=\E[\neg U \oplus U] 
	\end{align}
	Computing the above equation gives the evaluation $\E[Y] = 1$. Applying the DP formula in Thm.~\ref{thm:_4_1_dp} (or the IPW formula in Thm.~\ref{thm:_4_1_ipw}) gives the following evaluation, for any policy $\pi(X)$, 
	\begin{align}
		\invE{Y}{\pi} &= \E[Y \mid X = 0] \pi(X = 0) +\E[Y \mid X = 1] \pi(X = 1) \label{eq:_8_2_mab4}
	\end{align}
	Among the above quantities, the conditional mean $\E[Y\mid X]$ is given by, for any $x$,
	\begin{align}
		\E[Y \mid X = x] &= \E[X \oplus U \mid X = x]\\
		&= \E[x \oplus \neg x \mid X = x]
	\end{align}
	The last step holds since values of $X$ are given by $X \gets \neg U$ in the MAB environment $\1M$. Computing the above equation gives $\E[Y \mid X = x] = 1$ for $x = 0, 1$. Eq.~\ref{eq:_8_2_mab4} could be further written as
	\begin{align}
		\invE{Y}{\pi} &= \pi(X = 0) +\pi(X = 1) = 1
	\end{align}
	Again, let $\1R(Y) = \alpha Y$ be a linear reward function with a positive coefficient $\alpha > 0$. By substituting evaluations $\E[Y] = 1$ and $\invE{Y}{\pi} = 1$ into Eq.~\ref{eq:_8_2_mab1}, we obtain the following minimax program:
\begin{align}
	\nu^* &= \min_{\pi(X)} \max_{\alpha > 0} \alpha \E\left[ Y \right ] - \alpha  \invE{Y}{\pi}\\
	&= \min_{\pi(X)} \max_{\alpha > 0} \alpha - \alpha\\
	&= 0
\end{align}
This means that any fixed coefficient $\alpha > 0$, the imitator is able to achieve the expert's performance using any policy $\pi(x)$. To verify this conclusion, let an IRL policy $\pi_{\textsc{irl}}: X \gets 1$. Evaluating the expected reward $\1R(Y)$ in submodel $\1M_{\pi_{\textsc{irl}}}$ implies
\begin{align}
	\invE{\1R(Y)}{\pi_{\textsc{irl}}} &= \alpha \invE{Y}{X \gets 1}\\
	&=\alpha \E[1 \oplus U]\\
	&=0.5 \alpha
\end{align}
The last step holds since $U$ is uniformly drawn over the binary domain $\{0, 1\}$. This means that the IRL policy ($\invE{\1R(Y)}{\pi_{\textsc{irl}}} = 0.5\alpha$) fails to achieve the expert's performance $\E[\1R(Y)] = \alpha$. \hfill $\blacksquare$
\end{example}

\subsubsection{Minimal Imitation Backdoor}\label{sec:_8_2_1}
We will next study causal IRL in more general settings where the NUC assumption does not hold, and there exist unobserved confounders in the demonstration data affecting both actions and other variables in the environment. Our algorithm relies on a refinement of the imitation backdoor condition (Def.~\ref{def:_8_1_ibc}), based on the concept of minimal $d$-separating sets.
\begin{definition}[Minimal Imitation Backdoor]\label{def:_8_2_minimal_ibc}
	Let $\G$ be a causal diagram and $\*X, \*Y \in \*V$ be subsets of variables. An imitation admissible space $\Pi$ over $\*X$ is said to be \emph{minimal} if there exists no proper subspace $\Pi' \subset \Pi$ satisfying the imitation backdoor w.r.t. $\*Y$ in $\G$. \hfill $\blacksquare$
\end{definition}
In words, an imitation admissible space $\Pi = \left \{ \langle X_i, \*S_i \rangle \right\}_{X_i \in \*X}$ is minimal if for every action $X_i \in \*X$, $\*S_i$ is a minimal $d$-separating set between action $X_i$ and reward signals $\*Y$ is the manipulated diagram $\1G_{\underline{X_i}, \pi_{i+1}, \dots, \pi_H}$; or states $\*S_i = \emptyset$ whenever $X_i$ is not an ancestor of $\*Y$ in diagram $\1G_{\pi_{i+1}, \dots, \pi_H}$.

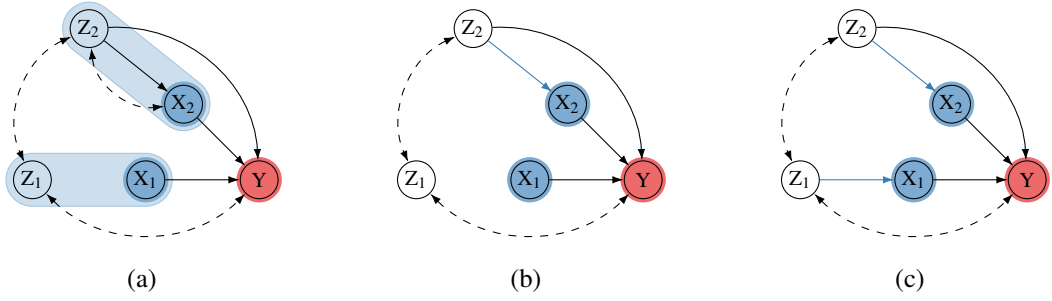
\begin{figure}[t]
	\hfill
	\begin{subfigure}{0.25\linewidth}\centering
		\begin{tikzpicture}
			\def\outerr{3.5}
			\def\innerr{3}

			\node[vertex] (Z2) at (0, 0) {Z\textsubscript{1}};
			\node[vertex] (X2) at (1.5, 0) {X\textsubscript{1}};
			\node[vertex] (Z1) at (0.75, 2) {Z\textsubscript{2}};
			\node[vertex] (X1) at (2, 1) {X\textsubscript{2}};
			\node[vertex] (Y) at (3, 0) {Y};

			\draw[dir] (Z1) -- (X1);
			\draw[dir] (X1) -- (Y);
			\draw[dir] (X2) -- (Y);

			\draw[bidir] (Z1) to [bend right=45] (Z2);
			\draw[dir] (Z1) to [bend left=45] (Y);
			\draw[bidir] (Z1) to [bend right=45] (X1);

			\draw[bidir] (Z2) to [bend right=45] (Y);

			\begin{pgfonlayer}{back}
				\draw[fill=betterblue!25,draw=betterblue!45] \convexpath{Z1, X1}{\outerr mm};
				\draw[fill=betterblue!25,draw=betterblue!45] \convexpath{Z2, X2}{\outerr mm};
				\node[circle,fill=betterblue!65,draw=none,minimum size=2*\innerr mm] at (X1) {};
				\node[circle,fill=betterblue!65,draw=none,minimum size=2*\innerr mm] at (X2) {};
				\node[circle,fill=betterred!65,draw=none,minimum size=2*\innerr mm] at (Y) {};
			\end{pgfonlayer}
		\end{tikzpicture}
		\caption{}
		\label{fig:_8_2_1a}
	\end{subfigure}\hfill\hfill
	\begin{subfigure}{0.25\linewidth}\centering
		\begin{tikzpicture}
			\def\outerr{3.5}
			\def\innerr{3}

			\node[vertex] (Z2) at (0, 0) {Z\textsubscript{1}};
			\node[vertex] (X2) at (1.5, 0) {X\textsubscript{1}};
			\node[vertex] (Z1) at (0.75, 2) {Z\textsubscript{2}};
			\node[vertex] (X1) at (2, 1) {X\textsubscript{2}};
			\node[vertex] (Y) at (3, 0) {Y};

			\draw[dir,betterblue] (Z1) -- (X1);
			\draw[dir] (X1) -- (Y);
			\draw[dir] (X2) -- (Y);

			\draw[bidir] (Z1) to [bend right=45] (Z2);
			\draw[dir] (Z1) to [bend left=45] (Y);

			\draw[bidir] (Z2) to [bend right=45] (Y);

			\begin{pgfonlayer}{back}
				\node[circle,fill=betterblue!65,draw=none,minimum size=2*\innerr mm] at (X1) {};
				\node[circle,fill=betterblue!65,draw=none,minimum size=2*\innerr mm] at (X2) {};
				\node[circle,fill=betterred!65,draw=none,minimum size=2*\innerr mm] at (Y) {};
			\end{pgfonlayer}
		\end{tikzpicture}
		\caption{}
		\label{fig:_8_2_1b}
	\end{subfigure}\hfill\hfill
	\begin{subfigure}{0.25\linewidth}\centering
		\begin{tikzpicture}
			\def\outerr{3.5}
			\def\innerr{3}

			\node[vertex] (Z2) at (0, 0) {Z\textsubscript{1}};
			\node[vertex] (X2) at (1.5, 0) {X\textsubscript{1}};
			\node[vertex] (Z1) at (0.75, 2) {Z\textsubscript{2}};
			\node[vertex] (X1) at (2, 1) {X\textsubscript{2}};
			\node[vertex] (Y) at (3, 0) {Y};

			\draw[dir,betterblue] (Z1) -- (X1);
			\draw[dir] (X1) -- (Y);
			\draw[dir] (X2) -- (Y);

			\draw[bidir] (Z1) to [bend right=45] (Z2);
			\draw[dir] (Z1) to [bend left=45] (Y);
			\draw[dir,betterblue] (Z2) -- (X2);

			\draw[bidir] (Z2) to [bend right=45] (Y);

			\begin{pgfonlayer}{back}
				\node[circle,fill=betterblue!65,draw=none,minimum size=2*\innerr mm] at (X1) {};
				\node[circle,fill=betterblue!65,draw=none,minimum size=2*\innerr mm] at (X2) {};
				\node[circle,fill=betterred!65,draw=none,minimum size=2*\innerr mm] at (Y) {};
			\end{pgfonlayer}
		\end{tikzpicture}
		\caption{}
		\label{fig:_8_2_1c}
	\end{subfigure}\hfill\null
	\caption{Causal diagrams where $X$ represents an action (shaded blue) and $Y$ represents a latent reward (shaded red). Input covariates of the policy space $\Pi$ are shaded in light blue.}
	\label{fig:_8_2_1}
\end{figure}
\begin{example}\label{exp:_8_2_minimal_ibc}
	Consider the causal diagram $\G$ described in Fig.~\ref{fig:_8_2_1a} and a policy space $\Pi_1 = \left \{\langle X_1, \{Z_1\}\rangle, \langle X_2, \{Z_2\}\rangle\right \}$. For a policy $\Parens{\pi_1, \pi_2} \in \Pi$, the manipulated diagram $\G_{\pi_2}$ is shown in Fig.~\ref{fig:_8_2_1b}. It is verifiable that $\Pi_1$ satisfies the imitation backdoor condition w.r.t. the outcome $Y$ in $\G$ since the following independence relationships hold: $(X_1 \ci Y \mid Z_1)$ in $\G_{\underline{X_1}, \pi_2}$ and $(X_2 \ci Y \mid Z_2)$ in $\G_{\underline{X_2}}$,  respectively. However, the same space $\Pi_1$ is not minimal since $\{Z_1\}$ is not a minimal $d$-separating set and $(X_1 \ci Y)$ holds in $\G^{(1)}_{\underline{X_1}}$. On the other hand, $\Pi_2 = \left \{\langle X_1, \emptyset \rangle, \langle X_2, \{Z_2\}\rangle\right \}$ is minimal imitation admissible since conditioning on the covariate set $\{Z_2\}$ $d$-separates the backdoor path $X_2 \leftarrow Z_2 \rightarrow Y$ in diagram $\G_{\underline{X_2}}$; removing node $Z_2$ opens the backdoor path. \hfill $\blacksquare$
\end{example}
A key property of a minimal imitation admissible space $\Pi$ is that for every policy $\pi \sim \Pi$, the interventional distribution $\inv{\*Y}{\pi}$ is identifiable from the observational distribution $P(\*V)$, provided with the structural assumptions encoded in the causal diagram $\G$.
\begin{restatable}{theorem}{thmminimalibc}\label{thm:_8_2_minimal_ibc}
	Let $\G$ be a causal diagram, $\Pi$ be a policy space over actions $\*X$, and $\*Y \subseteq \*V$ be a subset of variables. If there exists a subspace $\Pi' = \left \{ \langle X_i, \*Z_i \rangle  \right\}_{i = 1}^H$ contained in $\Pi$ such that $\Pi'$ is minimal imitation admissible w.r.t. $\*Y$ in $\1G$, then for every policy $\pi \in \Pi'$, the interventional distribution $\inv{\*Y}{\pi}$ is computable from $P(\*V)$ and given by 
	\begin{align}
		\inv{\*y}{\pi} = \sum_{\bar{\*x}_H, \bar{\*z}_H} P\left(\*y\mid \bar{\*x}_H, \bar{\*z}_H \right) \prod_{i = 1}^H  P\left ( \*z_i  \mid \bar{\*x}_{i-1}, \bar{\*z}_{i-1}\right)  \pi_{i} \left(x_i \mid \*z_i\right).  \label{eq:_8_2_minimal_ibc}
	\end{align}
	where $\bar{\*X}_{i} = \Braces{X_1, \dots, X_i}$ and $\bar{\*Z}_i = \Braces{\*Z_1, \dots, \*Z_i}$ are sequences of actions and input covariates up to the decision horizon $i = 1, \dots, H$. \hfill $\blacksquare$
\end{restatable}
However, the same identifiability result does not generally hold for policies in a non-minimal imitation admissible space. The following example demonstrates such an instance.
\begin{example}
Consider the causal diagram $\G$ of Fig.~\ref{fig:_8_2_1a} again. Let us focus on the minimal imitation admissible space
\begin{align}
\Pi_2 = \left \{\langle X_1, \emptyset \rangle, \langle X_2, \{Z_2\}\rangle\right \}	
\end{align}
For ever policy $\pi \in \Pi_2$, the post-interventional diagram $\G_{\pi}$ is shown in Fig.~\ref{fig:_8_2_1b}. By applying Thm.~\ref{thm:_8_2_minimal_ibc} we obtain
	\begin{align}
		\inv{y}{\pi} =  \sum_{x_1, x_2, z_2} P\left(y\mid x_1, x_2, z_2\right) P(z_2 \mid x_1) \pi_2(x_2 \mid z_2)\pi_1(x_1)
	\end{align}
On the other hand, the same identification result in Thm.~\ref{thm:_8_2_minimal_ibc} does not necessarily hold for a non-minimal imitation admissible space. 
	
More specifically, consider a policy space 
\begin{align}
\Pi_1 = \left \{\langle X_1, \{Z_1\} \rangle, \langle X_2, \{Z_2\}\rangle\right \}	
\end{align}
For ever policy $\pi \in \Pi_1$, the post-interventional diagram $\G_{\pi}$ is shown in Fig.~\ref{fig:_8_2_1c}. As discussed previously (Example~\ref{exp:_8_2_minimal_ibc}), $\Pi_1$ is not minimal. This means that, for any policy $\pi \in \Pi_1$, the interventional distribution $\inv{Y}{\pi}$ is not computable from the identification formula in Eq.~\ref{eq:_8_2_minimal_ibc}. More generally, $\inv{Y}{\pi}$ for policies  $\pi \in \Pi_1$ is not identifiable from the observational distribution $P(\*V)$ in diagram $\G$. Following the decomposition in Eq.~\ref{eq:_4_3_pid1}, $\inv{Y}{\pi}$ can be written as, 
\begin{align}
	\inv{y}{\pi} = \sum_{x_1, x_2, z_1, z_2} \inv{y, z_1, z_2}{x_1, x_2} \pi_1(x_1 \mid z_1)\pi_2(x_2 \mid z_2)
\end{align}
Prop.~\ref{lem:_4_3_pid} implies that $\inv{Y}{\pi}$ is identifiable if and only if the interventional distribution $\inv{Y, Z_1, Z_2}{x_1, x_2}$ is identifiable in the causal diagram $\G$. However, such quantity $\inv{Y, Z_1, Z_2}{x_1, x_2}$ is not identifiable due to the presence of the bi-directed path $X_2 \bidirectarrow Z_2 \bidirectarrow Z_1 \bidirectarrow Y$ \citep[Thm.~16]{tian:02}. Indeed, the non-identifiability of the effects of policies $\pi \in \Pi_1$ in the causal diagram $\G$ described in Fig.~\ref{fig:_8_2_1a} has been shown in \citep{tian2008dsp,correa2019statistical}. \hfill $\blacksquare$
\end{example}
The concept of minimal imitation backdoor in Def.~\ref{def:_8_2_minimal_ibc} and the identification result in Thm.~\ref{thm:_8_2_minimal_ibc} provide a natural algorithm for performing causal IRL when unobserved confounders generally exist. Instead of searching for imitating policies in the policy space $\Pi$, the agent will focus on a minimal imitation admissible subspace $\Pi' \subseteq \Pi$.
Specifically, as discussed previously in Sec.~\ref{sec:_8_1}, there exist efficient algorithms finding admissible policy subspaces satisfying the imitation backdoor. Once such an admissible subspace is found, one could obtain a minimal imitation admissible subspace by iteratively removing input state variables from $\*S_i$ for every action $X_i$ until the imitation backdoor does not hold. This procedure could be done in polynomial steps with regard to the total number of states $\*S$ and actions $\*X$ variables.

\subsubsection{Imitation via Inverse RL}\label{sec:_8_2_2}
Once a minimal imitation admissible subspace $\Pi' \subseteq \Pi$ is obtained, one could obtain an imitating policy by solving the minimax program in Eq.~\ref{eq:_8_2_opt} with the policy space $\Pi$ substituted with $\Pi'$. By expanding values of $\*Y$, this optimization program could be written as,
\begin{align}
	\nu^* = \min_{\pi \in \Pi'} \max_{\1R \in \2R} \sum_{\*y} \1R(\*y) (\underbrace{P(\*y)}_{\text{expert's occupancy}}- \underbrace{\inv{\*y}{\pi}}_{\text{imitator's occupancy}} )\label{eq:_8_2_canonical}
\end{align}
Among quantities in the above equation, the first term is the expert's occupancy measures over domains of signals $\*Y$, which is a marginal observational distribution $P(\*Y)$. The second term is the expert's occupancy measures over domains of $\*Y$, which is an interventional distribution $\inv{\*Y}{\pi}$. Since $\Pi'$ is a minimal subspace satisfying the imitation backdoor criterion, applying Thm.~\ref{thm:_8_2_minimal_ibc} permits one to compute $\inv{\*Y}{\pi}$ from the observational distribution $P(\*V)$ and the policy $\pi$. \footnote{More generally, the imitator could search over all policies $\pi \in \Pi$ such that the imitator's occupancy measure $\inv{\*Y}{\pi}$ induced by $\doo(\pi)$ is identifiable in diagram $\G$. This imitation approach has been studied in \citep{ruan2023causal}.}

Provided with some common choices of the hypothesis class $\2R$, the minimax program in Eq.~\ref{eq:_8_2_canonical} is solvable using some state-of-art IRL algorithms. Due to this reason, we consistently refer to Eq.~\ref{eq:_8_2_canonical} as the \emph{canonical IRL program}. To make this argument more precise, we will demonstrate this reduction procedure with the multiplicative-weights algorithm (MWAL) \citep{syed2008game} and the generative adversarial imitation learning (GAIL) \citep{ho2016generative}.

\paragraph{Causal MWAL}\citep{abbeel2004apprenticeship,syed2008game} study IRL in Markov decision processes where the reward function $\1R(\*y)$ is a linear combination of $k$-length \emph{feature expectations} vectors $\*\phi(\*y)$. Particularly, let $\1R(\*y) = \*w \cdot \*\phi(\*y)$ for a coefficient vector $\*w$ in a convex set
\begin{align}
	\3P^k = \left \{ \*w \in \3R^k \mid \lVert \*w \rVert_1 = 1 \text{ and } \*w \succeq \*0\right\}.
\end{align}
Let $\*\phi^{(i)}$ be the $i$-th component of feature vector $\*\phi$ and let deterministic policies with space $\Pi$ be ordered by $\pi^{(1)}, \dots, \pi^{(n)}$. The canonical program in Eq.~\ref{eq:_8_2_canonical} is reducible to a two-person zero-sum matrix game under linearity.
\begin{restatable}{proposition}{propmwal}\label{prop:_8_2_mwal}
	For a hypothesis class $\2R = \{\1R =\*w \cdot \*\phi \mid \*w \in \3P^k \}$, the solution $\nu^*$ of the canonical program in Eq.~\ref{eq:_8_2_canonical} is obtainable by solving the following minimax problem
	\begin{align}
		\nu^* = \min_{\pi \in \Pi'} \max_{\*w \in \3P^k} \*w^\top \*G \pi, \label{eq:_8_2_mwal}
	\end{align}
	where $\*G$ is a $k \times n$ matrix given by $\*G(i, j) = \sum_{\*y} \*\phi^{(i)} (\*y) \left (P(\*y) - P_{\pi^{(j)}}(\*y) \right)$. \hfill $\blacksquare$
\end{restatable}
There exist effective multiplicative weights algorithms for solving the matrix game in Eq.~\ref{eq:_8_2_mwal}, including MW \citep{freund1999adaptive} and MWAL \citep{syed2008game}.

\paragraph{Causal GAIL}\citep{ho2016generative} introduces the GAIL algorithm for learning an imitating policy in Markov decision processes with a general family of non-linear reward functions. In particular, $\1R(\*y)$ takes values in the real space $\3R$, i.e., $\1R \in \3R^{\*Y}$ where $\3R^{\*Y} = \{r : \D(\*Y) \mapsto \3R\}$. The complexity of reward function $\1R$ is penalized by a convex regularization function $\psi\left(\1R\right)$, i.e.,
\begin{align}
	\nu^* = \min_{\pi \in \Pi'} \max_{\1R \in \3R^{\*Y} } \sum_{\*y} \1R(\*y) \left (P(\*y) -P_{\pi}(\*y) \right) - \psi(\1R) \label{eq:_8_2_gail}
\end{align}
Henceforth, we will consistently refer to Eq.~\ref{eq:_8_2_gail} as the \emph{penalized canonical program} of causal IRL. It is often preferable to solve its conjugate form. Formally,
\begin{restatable}{proposition}{propgail}\label{prop:_8_2_gail}
	For a hypothesis class $\2R = \{\1R :\D(\*Y) \mapsto \3R\}$ regularized by $\psi$, the solution $\nu^*$ of the penalized canonical program in Eq.~\ref{eq:_8_2_gail} is obtainable by solving the following problem
	\begin{align}
		\nu^* = \min_{\pi \in \Pi'} \psi^* \left (P - P_{\pi} \right) \label{eq:_8_2_gail2}
	\end{align}
	where $\psi^*$ be a conjugate function of $\psi$ and is given by $\psi^* = \max_{\1R \in \3R^{\*Y}} a^\top \1R - \psi(\1R)$. \hfill $\blacksquare$
\end{restatable}
Eq.~\ref{eq:_8_2_gail2} seeks a policy $\pi$ which minimizes the divergence of joint probabilities over reward signals $\*Y$ between the imitator and the expert, as measured by the function $\psi^*$. 
When we utilize a regularizer $\psi(r)$ similar to \cite[Eq.~13]{ho2016generative}, the convex conjugate function $\psi^*$ in Eq.~\ref{eq:_8_2_gail2} is further written as:
\begin{align}
	\min_{\pi \in \Pi'} \psi^* \left (P - P_{\pi}  \right) = \min_{\pi \in \Pi'} \max_{D \in (0,1)^{\*Y}} E \left[ \log(D(\*Y)) \right] + \invE{\log(1 - D(\*Y))}{\pi}, \label{eq:_8_2_gail3}
\end{align}
where function $D \in \D(\*Y) \mapsto (0, 1)$ is a discriminator classifier (e.g, a neural network). The above equation draws the connection between causal imitation learning and the computational framework of generative adversarial networks \citep{goodfellow2014generative}, which could be viewed as two neural networks competing against each other in a zero-sum game. When the discriminator $D$ cannot distinguish the occupancy measure generated by the policy $\pi$ from the expert, then $\pi$ has successfully matched the expert's performance. Solving the minimax program of Eq.~\ref{eq:_8_2_gail3} requires finding a saddle point $(\pi, D)$. This could be done by iteratively optimizing policy parameters $\pi$ and discriminator $D$ following the implementation procedure of GAIL algorithm \citep{ho2016generative}.

\begin{experiment}\label{exp:_8_2_bd}
	We demonstrate our causal imitation framework on an SCM $\1M^*$ compatible with the causal diagram in Fig.~\ref{fig:_8_2_1a}. Particularly,
	\begin{align}
		\1M^* = \langle \*U = \{U_1, U_2, U_3, U_4\}, \*L = \emptyset, \*V = \{X_1, X_2, Z_1, X_2, Y\}, \2F, P(\*U) \rangle \label{eq:_8_2_bd1}
	\end{align}
	where structural functions $\2F$ is defined as
	\begin{align}
		\2F = \begin{cases}
			      Z_1  \gets U_1 \oplus U_3,            \\
			      X_1 \sim \texttt{Bern}(0.68)          \\
			      Z_2  \gets U_1 \oplus U_2 \oplus U_4, \\
			      X_2 \gets U_2 \oplus Z_2              \\
			      Y \gets \Parens{X_1, X_2, Z_1, Z_2, U_3}
		      \end{cases}
	\end{align}
Among quantities in the above equation, reward signal $Y = \Parens{Y_1, \dots Y_5}$ is a feature vector containing $5$ elements; the exogenous distribution $P(U_1, U_2, U_3, U_4)$ is defined such that $U_i$, $i = 1, \dots, 4$ are independent variables given by
	\begin{align}
		 & U_1 \sim \texttt{Bern}(0.8), &  & U_2 \sim \texttt{Bern}(0.8), &  & U_3 \sim \texttt{Bern}(0.2) & U_4 \sim \texttt{Bern}(0.1)
	\end{align}
	The agent's goal is to optimize a CDM $\Tuple{\1M^*, \Pi, \1R}$ where environment $\1M^*$ is defined in Eq.~\ref{eq:_8_2_bd1}; the policy space $\Pi = \Braces{\Tuple{X_1, \{Z_1\}}, \Tuple{X_2, \{Z_1, X_1, Z_2\}}}$; and the reward function $\1R(Y) = \oplus_{i = 1}^5 Y_i$.

	\begin{figure}[!t]
		\centering
		\includegraphics[width=0.5\linewidth]{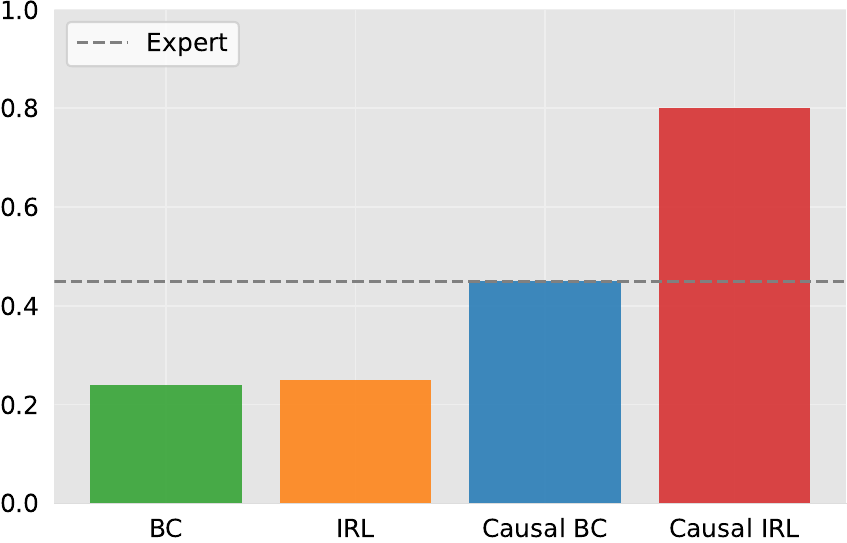}
		\caption{Simulation results evaluating causal IRL when imitation backdoor condition holds.}\label{fig:_8_2_2}
	\end{figure}

We will then apply different imitation strategies to learn an imitating policy in space $\Pi$ without detailed parametrization of the reward function $\1R(Y)$. These imitation algorithms are
	\begin{itemize}
		\item Standard BC algorithm utilizes all observed states $\*S_i$ for every action $X_i \in \*X$.

		\item Standard IRL algorithm utilizes all observed states $\*S_i$ for every action $X_i \in \*X$. We will apply GAIL algorithm \citep{syed2008game} when $\1R(Y)$ is non-linear;

		\item Causal-BC algorithm, described in Thm.~\ref{thm:_8_1_ibc}, selects a set of covariates $\*Z_i$ for every action $X_i \in \*X$ following the imitation backdoor criterion (Def.~\ref{def:_8_1_ibc}). It then learns an imitating policy $\pi$ with subspace $\Pi' = \{\langle X_i, \*Z_i \rangle\}_{X_i \in \*X}$ using standard BC algorithms.

		\item Our proposed Causal-IRL algorithm first finds a minimal imitation admissible subspace $\Pi'$ (Thm.~\ref{thm:_8_2_minimal_ibc}) and then obtains an imitating policy by solving the canonical program in Eq.~\ref{eq:_8_2_canonical}. We will use Causal GAIL algorithm since $\1R(Y)$ is non-linear. Reward augmentation (RA) is performed to incorporate the parametric knowledge that $\1R(Y)$ is a monotone function concerning values of $Y_1, Y_2$ \citep{li2017infogail}. This is done by adding an additional regularization function in Eq.~\ref{eq:_8_2_gail3} to encourage assigning higher values of features $Y_1, Y_2$.
	\end{itemize}
Simulation results are shown in Fig.~\ref{fig:_8_2_2}. The analysis reveals that Causal-IRL consistently outperforms the expert's policy and other imitation strategies by exploiting additional parametric knowledge about the reward function; Causal-BC obtains a policy that mimics the expert's performance. As expected, BC and IRL failed to obtain a policy that matches the expert's performance. \hfill $\blacksquare$
\end{experiment}
This section investigates imitation learning in the semantics of structural causal models. The goal is to find an imitating policy that can perform at least as well as the expert behaviors from combinations of demonstration data and qualitative knowledge about the data-generating process represented as a causal diagram. First, we provided a novel graphical criterion that is sufficient for determining the feasibility of learning an imitating policy that mimics the expert's performance. When such a condition holds, one could obtain an effective imitating learning using standard behavioral cloning. We also investigate imitation learning via inverse reinforcement learning (IRL), provided with additional quantitative knowledge about the reward function. We provide a graphical criterion based on the sequential backdoor, which allows one to obtain an imitating policy by solving a canonical optimization equation of causal IRL. Such a canonical formulation addresses the challenge of the presence of unobserved confounders (UCs) and is solvable by leveraging standard IRL algorithms. 


\section{Conclusions}
\label{sec:_9_conclusions}
The current generation of AI agents capable of optimal decision-making builds on the theoretical framework of reinforcement learning (RL). Most of these RL systems do not explicitly represent the underlying causal models or engage in causal reasoning. On the other hand, there is a growing recognition across many fields and sciences that effective decision-making relies on an understanding of the causal mechanisms in the environment. For example, an intelligent robot needs to grasp the cause-and-effect relationships within its surroundings to plan its actions effectively; a physician must understand the effects of available medications to devise a suitable treatment strategy for her patients; an economist, too, needs to envision the relationship between skill sets and the future job market in order to create an effective educational policy. These scenarios illustrate how decision-making across various sectors of society depends on understanding complex, dynamic, and often unobserved causal mechanisms. Although there have been some attempts to integrate causal knowledge into RL tasks, a systematic approach and a cohesive foundation are still lacking.

To address this challenge, we combine the capabilities of RL agents with Pearl's Structural Causal Models (SCMs) theory to encode causal knowledge and perform counterfactual reasoning. This marriage leads to an algorithmic and theoretical framework for robust decision-making under uncertainties, which is part of an emerging branch of research called \textit{Causal Reinforcement Learning} (CRL). Building on this framework, we are able to bring improvement to RL algorithms in some key aspects. First, The CRL framework enables us to relax certain key assumptions regarding the causal mechanisms that generate the observed data. We have developed innovative algorithms that are resilient to unobserved confounding bias in offline settings, which include off-policy learning and imitation learning. Additionally, we proposed more efficient online learning algorithms that effectively identify optimal policies while achieving near-optimal regret. These advancements leverage causal conclusions drawn from biased offline data.

The final important distinction presented in this manuscript is the difference between the regimes in which AI agents operate to interact with their environment. Specifically, supervised learning agents identify patterns from data gathered through passive observation, while reinforcement learning (RL) agents actively engage with the system and modify their policies based on the responses they receive. By generalizing these interaction regimes, we open up new learning possibilities that have not been explored in the existing literature. The problem of where to intervene allows us to design agents to achieve better performance by combining both passive observation and active intervention. Counterfactual randomization generalizes the classic Fisherian randomized experience, enriching agents with capabilities of counterfactual reasoning, which we believe is critical to design the next generation of AI systems.

\section*{Acknowledgements}
This research was supported in part by the NSF, ONR, AFOSR, DARPA, DoE, Amazon, JP Morgan,
and The Alfred P. Sloan Foundation.

\newpage

\bibliography{book}

\newpage

\appendix
\section{Comparison with Partially Observed MDPs}\label{sec:_3_4}
In this section, we will extend the Causal Hierachy Theorem (CHT, \citep[Thm.~1]{bareinboim2020pearl}) to a general family of stochastic processes where the Markov property does not hold for states and actions across time steps. Specifically, we focus on the partially observed Markov decision processes (POMDP, \cite{sondik1971optimal}) that model the dependence between observations over a long sequence of time steps through latent states and dynamics. Formally,
\begin{definition}[Standard POMDP \citep{sondik1971optimal}]\label{def:_3_4_mdp}
\sloppy	A partially observable Markov decision process is tuple $\Tuple{\D(S), \D(X), \D(O), \1T, \1R, \1O}$ where 
	\begin{enumerate}
            \item $\D(S), \D(X), \1T$ and $\1R$ describe a Markov decision process;
		\item $\D(O)$ is a finite set of observations the agent can perceive of its world, called the observation space;
		\item $\1O(x, s, o)$ is the observation function, which gives, for each action $X_i = x$ and resulting state $S_{i+1} = s$, a probability distribution over possible observations $O_{i+1} = o$.
	\end{enumerate}
\end{definition}
A policy $\pi$ in a standard POMDP is a sequence of stochastic decision rules $\{\pi_1, \pi_2, \dots \}$; each decision rule $\pi_i$ is a function mapping from the observations and actions history $\bar{\*O}_{1:i}, \bar{\*X}_{1:i-1}$ to a probability distribution over the action space $\1D(X)$. Given a policy $\pi$ and a distribution over the initial state and observation $P(S_1, O_1)$, every standard PMDP model defines a joint distribution over observations $\bar{\*O}_{1:H}$, actions $\bar{\*X}_{1:H}$, and rewards $\bar{\*Y}_{1:H}$ up to decision horizon $H$, i.e.,
\begin{equation}
    \begin{split}
        &P_{\pi}(\bar{\*o}_{1:H}, \bar{\*x}_{1:H}, \bar{\*y}_{1:H}) \\
        &= \sum_{\bar{\*s}_{1:H}} P(s_1, o_1) \prod_{i = 1}^H \pi(x_i\mid s_i) \1T(s_i, x_i, s_{i+1}) \1O(x_i, s_{i+1}, o_{i+1}) \I\{\1R(s_i, x_i) = y_i\}
    \end{split}
\end{equation}
Due to the presence of latent states, the Markov property no longer holds with regard to the perceived observations. This argument is corroborated with the network structure in the causal diagram $\1G_{\textsc{pomdp}}$ of Fig.~\ref{fig:_3_1_pomdp}. For every stage $i = 1, 2, \dots$, conditioning on the observation $O_i$ and action $X_i$ fails to block all paths from observed history $O_j, X_j, Y_j$ for $j < i$ to any future observation $O_{k}$, action $X_{k}$, and reward $Y_k$ for $k > i$ (Def.~\ref{def:_2_3_dsep}). Specifically, such long-sequence dependency is generated from the latent states $S_i$, e.g., the open causal path $O_1 \leftarrow S_1 \rightarrow S_2 \rightarrow O_2$, which could be represented using a standard POMDP.
\begin{example}[POMDP, Observational]\label{exp:_3_4_pomdp_obs}
\sloppy Consider the following SCM environment $\M^*$ adapted from Eq.~\ref{eq:_2_1_mdp}, unrolling over stages $i = 1, 2, \dots$
\begin{align}
\M^{*} =
 \Tuple{\*U = \{U_{i, 1}, U_{i,2}, U_{i, 3}\}, \*V =  \{X_i, Y_i, S_i, O_i\}, \2F = \left \{\2F^{*}_i \right \}, P^{*}(\*U)}_{i = 1, 2, \dots}. 
\end{align}
The above SCM is identical to the model $\M^{*}$ defined in Eq.~\ref{eq:_2_1_mdp}, except that the underlying state $S_i$ is now latent to the learner; and the endogenous variables include an observation $O_i$ fixed at a constant $O_i \gets 0$. This means that the learning agent, interacting with the environment, accesses samples drawn from marginal observational $P(\bar{\*o}_{1:H}, \bar{\*x}_{1:H}, \bar{\*y}_{1:H})$ or interventional distribution $P_{\pi}(\bar{\*o}_{1:H}, \bar{\*x}_{1:H}, \bar{\*y}_{1:H})$, depending on the regimes of interactions. 

We compute the observational distributions $P\Parens{S_{i+1} \mid \bar{\*O}_{1:i}, \bar{\*X}_{1:i}}$ and $\3E \Brackets{Y_i \mid \bar{\*O}_{1:i}, \bar{\*X}_{1:i}}$ and summarize them using a standard POMDP $\Tuple{\D(S), \D(X), \D(O), \1T_{\text{obs}}, \1R_{\text{obs}}, \1O}$ where $\D(S), \D(X), \1T_{\text{obs}}, \1R_{\text{obs}}$ form a standard MDP described in Example~\ref{exp:_3_3_mdp_obs}; the observation function $\1O(x, s, o) = 1$ for observation $o=0$ given any action $X_i = x$ and subsequent state $S_{i+1} = s$. Fig.~\ref{fig:_3_4_cht_pomdp} (a) shows a finite automaton that describes its detailed system dynamics. The shaded ellipse around the states $S = 0$ and $S = 1$ indicates that both states yield the same observation. \hfill $\blacksquare$
\end{example}
Following a similar argument, we could show that any interventional distribution evaluated in a SCM $\1M^*$ graphically described in Fig.~\ref{fig:_3_1_pomdp} violates the Markov property, leading to an alternative standard POMDP representation. 
\begin{example}[POMDP, Interventional]\label{exp:_3_4_pomdp_inv}
\sloppy Consider again the SCM $\M^*$ described in Example~\ref{exp:_3_4_pomdp_obs}. Its interventional distributions $\inv{S_{i+1} \mid \bar{\*O}_{1:i}}{\bar{\*X}_{1:i}}$ and $\invE{Y\mid \bar{\*O}_{1:i}}{\bar{\*X}_{1:i}}$ a standard POMDP $\Tuple{\D(S), \D(X), \D(O), \1T_{\text{exp}}, \1R_{\text{exp}}, \1O}$ where 
$\D(S), \D(X), \1T_{\text{exp}}, \1R_{\text{exp}}$ are described in the standard MDP of Example~\ref{exp:_3_3_mdp_inv}; the observation function $\1O(x, s, o = 0) = 1$ given any action $X_i = x$ and state $S_{i+1} = s$. The system dynamics of this POMDP are described in the finite automaton of Fig.~\ref{fig:_3_4_cht_pomdp} (b). The ellipse around the states indicates that both states yield the same observation. \hfill $\blacksquare$
\end{example}
In both examples above, the observational and interventional distributions evaluated in the SCM $\1M^*$ could be represented using standard POMDPs. However, the detailed system dynamics in these standard POMDPs differ, as illustrated in the finite automata shown in Fig.~\ref{fig:_3_4_cht_pomdp}. One may wonder if it is possible to recover interventional quantities $\1T_{\text{exp}}$ and $\1R_{\text{exp}}$ from the observational data in POMDP environments. Our next result shows this is not the case.
\begin{figure}[t]
\centering
  \resizebox{\linewidth}{!}{
  \begin{tikzpicture}
      \node (env) [environment, fill=gray!10, dashed, minimum height=1cm, text width=1cm, label=above:{Eq.~(5)}] at (6.25, 5.5) {$\M^*$};
      \node (env1) [environment, fill=gray!10, dashed, minimum height=1cm, text width=1cm, label=above:{Eq.~(119)}] at (2.25, 5) {$\M^{(1)}$};
      \node (env2) [environment, fill=gray!10, dashed, minimum height=1cm, text width=1cm, label=above:{Eq.~(121)}] at (10.25, 5) {$\M^{(2)}$};

      \draw[dashed, fill=gray!10] (2.25,0) ellipse (3cm and 1cm);
      \node[vertex, minimum width=6mm] (S0) at (0, 0) {S=0};
      \node[vertex, minimum width=6mm] (S1) at (4.5, 0) {S=1};
      \node[action, label={[shift={(0,0)}]\scriptsize X=0}] (X00) at (1, 2) {};
      \node[action, label={[shift={(0,-0.5)}]\scriptsize X=1}] (X10) at (1, -2) {};
      \node[action, label={[shift={(0,0)}]\scriptsize X=0}] (X01) at (3.5, 2) {};
      \node[action, label={[shift={(0,-0.5)}]\scriptsize X=1}] (X11) at (3.5, -2) {};
      
      \draw[dir] (S0) to (X00);
      \draw[dir] (S0) to (X10);
      \draw[dir] (S1) to (X01);
      \draw[dir] (S1) to (X11);
      
      \node[text width = 1.3cm, align = right, left] at (0.2, 1.5) {\scriptsize 0.9,Y=0.1};
      \node[text width = 1.3cm, align = right, left] at (0.2, -1.5) {\scriptsize 0.9,Y=0.1};
      
      \node[text width = 1.3cm, align = right] at (1.3, 0.4) {\scriptsize 0.1,Y=0.1};
      \node[text width = 1.3cm, align = right] at (1.3, -0.4) {\scriptsize 0.1,Y=0.1};
      
      \node[text width = 1.3cm, align = left] at (3.2, 0.4) {\scriptsize 0.1,Y=0.1};
      \node[text width = 1.3cm, align = left] at (3.2, -0.4) {\scriptsize 0.1,Y=0.1};
      
      \node[text width = 1.3cm, right] at (4.3, 1.5) {\scriptsize 0.9,Y=0.1};
      \node[text width = 1.3cm, right] at (4.3, -1.5) {\scriptsize 0.9,Y=0.1};

      \draw[dir] (X00) to [bend right = 45] (S0);
      \draw[dir] (X10) to [bend left = 45] (S0);
      \draw[dir] (X00) to [bend left = 15] (S1);
      \draw[dir] (X10) to [bend right = 15] (S1);
      
      \draw[dir] (X01) to [bend left = 45] (S1);
      \draw[dir] (X11) to [bend right = 45] (S1);
      \draw[dir] (X01) to [bend right = 15] (S0);
      \draw[dir] (X11) to [bend left = 15] (S0);
      
      \node (mdpa) at (2.25, 0) [draw, gray, ultra thin, minimum width=7.5cm,minimum height=5.5cm] {};

      \draw[dashed, fill=gray!10] (10.25,0) ellipse (3cm and 1cm);
      \node[vertex, minimum width=6mm] (S0i) at (8, 0) {S=0};
      \node[vertex, minimum width=6mm] (S1i) at (12.5, 0) {S=1};
      \node[action, label={[shift={(0,0)}]\scriptsize X=0}] (X00i) at (9, 2) {};
      \node[action, label={[shift={(0,-0.5)}]\scriptsize X=1}] (X10i) at (9, -2) {};
      \node[action, label={[shift={(0,0)}]\scriptsize X=0}] (X01i) at (11.5, 2) {};
      \node[action, label={[shift={(0,-0.5)}]\scriptsize X=1}] (X11i) at (11.5, -2) {};
      
      \draw[dir] (S0i) to (X00i);
      \draw[dir] (S0i) to (X10i);
      \draw[dir] (S1i) to (X01i);
      \draw[dir] (S1i) to (X11i);
      
      \node[text width = 1.4cm, align = right, left] at (8.2, 1.5) {\scriptsize 0.18,Y=0.82};
      \node[text width = 1.4cm, align = right, left] at (8.2, -1.5) {\scriptsize 0.82,Y=0.18};
      
      \node[text width = 1.4cm, align = right] at (9.3, 0.4) {\scriptsize 0.82,Y=0.18};
      \node[text width = 1.4cm, align = right] at (9.3, -0.4) {\scriptsize 0.82,Y=0.82};
      
      \node[text width = 1.4cm, align = left] at (11.2, 0.4) {\scriptsize 0.82,Y=0.82};
      \node[text width = 1.4cm, align = left] at (11.2, -0.4) {\scriptsize 0.18,Y=0.18};
      
      \node[text width = 1.4cm, right] at (12.3, 1.5) {\scriptsize 0.18,Y=0.18};
      \node[text width = 1.4cm, right] at (12.3, -1.5) {\scriptsize 0.18,Y=0.82};
      
      \draw[dir] (X00i) to [bend right = 45] (S0i);
      \draw[dir] (X10i) to [bend left = 45] (S0i);
      \draw[dir] (X00i) to [bend left = 15] (S1i);
      \draw[dir] (X10i) to [bend right = 15] (S1i);
      
      \draw[dir] (X01i) to [bend left = 45] (S1i);
      \draw[dir] (X11i) to [bend right = 45] (S1i);
      \draw[dir] (X01i) to [bend right = 15] (S0i);
      \draw[dir] (X11i) to [bend left = 15] (S0i);
      
       \node (mdpb) at (10.25, 0) [draw, gray, ultra thin, minimum width=7.5cm,minimum height=5.5cm] {};
       
      \node (a) at (2.25, -2.5) {(a)};
      \node (b) at (10.25, -2.5) {(b)};
      
      \path [-Latex] (env) edge node[anchor=west, text width=1.9cm] {\hphantom{a}Obs. ($\1L_1$)} (mdpa.80);
      \path [-Latex] (env) edge node[anchor=east, text width=1.7cm] {Inv. ($\1L_2$)} (mdpb.100);
      
      \path [-Latex] (env1) edge node[anchor=east, text width=1.7cm, align = right] {Obs. ($\1L_1$)\\ Inv. ($\1L_2$)} (mdpa.north);
      \path [-Latex] (env2) edge node[anchor=west, text width=1.7cm] {Obs. ($\1L_1$)\\ Inv. ($\1L_2$)} (mdpb.north);
      
  \end{tikzpicture}
  }
\caption{Causal Hierarchy Theorem (CHT) in POMDP environments.}
\label{fig:_3_4_cht_pomdp}
\end{figure}
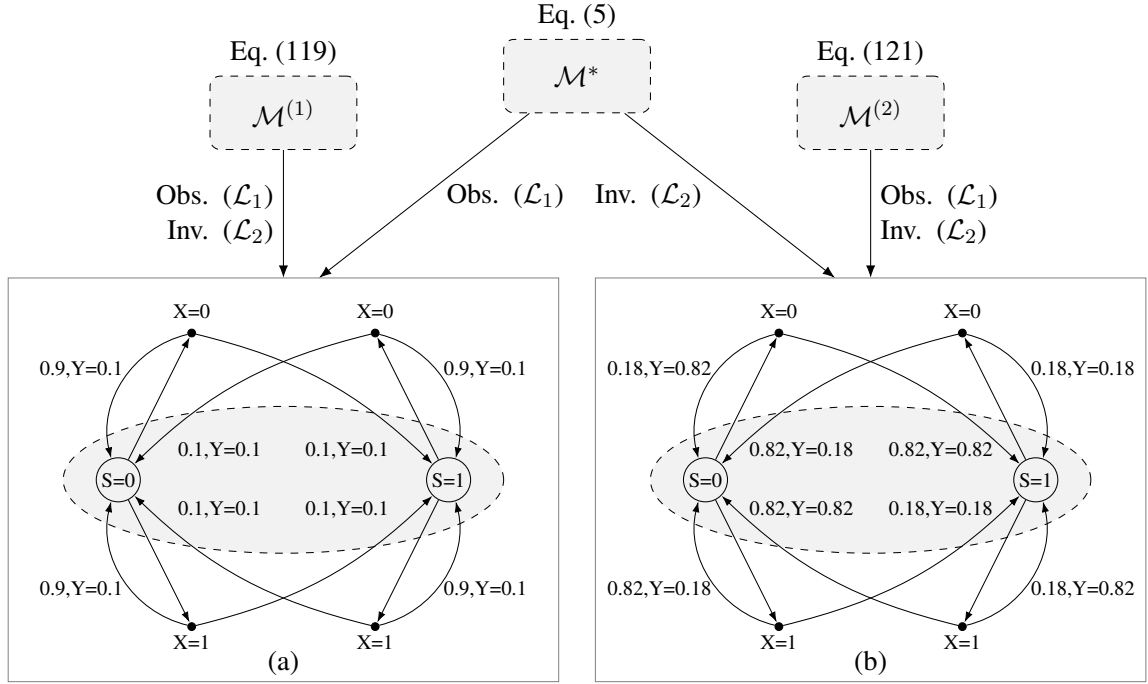%

\begin{proposition}\label{prop:_3_4_pomdp_obs}
	For any SCM $\1M^*$ compatible with the causal diagram $\G_{\textsc{pomdp}}$ of Fig.~\ref{fig:_3_1_pomdp}, there is an SCM $\1M^{(1)}$ compatible with $\G_{\textsc{pomdp}}$ such that for every stage $i = 1, 2, \dots$,
	\begin{align}
		&P^{(1)}\left ( o_{i+1} \mid \bar{\*o}_{1:i}, \bar{\*x}_{1:i} \right ) = P^{*}\left ( o_{i+1} \mid \bar{\*o}_{1:i}, \bar{\*x}_{1:i}\right ), &&\E^{(1)}\left[ Y_i \mid \bar{\*o}_{1:i}, \bar{\*x}_{1:i} \right] = \E^{*}\left[ Y_i \mid \bar{\*o}_{1:i}, \bar{\*x}_{1:i} \right] \label{eq:_3_3_mdp_obs1}
	\end{align}
	while 
	\begin{align}
		&P^{(1)}_{\bar{\*x}_{1:i}}\left ( o_{i+1} \mid \bar{\*o}_{1:i} \right ) \neq P^{*}_{\bar{\*x}_{1:i}}\left ( o_{i+1} \mid \bar{\*o}_{1:i} \right ), &&\E^{(1)}_{\bar{\*x}_{1:i}}\left[ Y_i \mid \bar{\*o}_{1:i}\right] \neq \E^{*}_{\bar{\*x}_{1:i}}\left[ Y_i \mid \bar{\*o}_{1:i}\right] \label{eq:_3_3_mdp_obs2}
	\end{align} \hfill $\blacksquare$
\end{proposition}
The following example constructs an alternative SCM $\1M^{(1)}$ that generates the observational distribution as the underlying environment $\1M^*$, but differs significantly in interventional distributions. 
\begin{example}[POMDP, Observational $\not \Rightarrow$ Interventional]\label{exp:_3_4_obs-pomdp}
\sloppy We will construct an alternative SCM $\M^{(1)}$ where its system dynamics collapse to the associational layer ($\1L_1$) with respect to the SCM $\M^{*}$ described in Example~\ref{exp:_3_4_pomdp_obs}. More specifically,
\begin{align}
\M^{(1)} =
 \Tuple{\*U = \{U_{i, 1}, U_{i,2}, U_{i, 3}\}, \*V =  \{X_i, Y_i, S_i, O_i\}, \2F = \left \{\2F^{(1)}_i \right \}, P^{(1)}(\*U)}_{i = 1, 2, \dots}. 
\end{align}
Similarly to the construction of $\M^{*}$, the above SCM is identical to the model $\M^{(1)}$ defined in Example~\ref{exp:_2_2_obs-mdp} except that for every stage $i = 1, 2, \dots$, the learner does not observe the state $S_i$, but only receives new endogenous variable $O_i \gets 0$. We compute the observational distributions $P\Parens{S_{i+1} \mid \bar{\*O}_{1:i}, \bar{\*X}_{1:i}}$ and $\3E \Brackets{Y_i \mid \bar{\*O}_{1:i}, \bar{\*X}_{1:i}}$ and the interventional distributions $\inv{S_{i+1} \mid \bar{\*O}_{1:i}}{\bar{\*X}_{1:i}}$ and $\invE{Y\mid \bar{\*O}_{1:i}}{\bar{\*X}_{1:i}}$ evaluated in $\1M^{(1)}$, following the discussion in Example~\ref{exp:_2_2_obs-mdp}. The analysis suggests that the observational and interventional distributions collapse in the model $\1M^{(1)}$, which could be described using the same finite-state automaton shown in Fig.~\ref{fig:_3_4_cht_pomdp}(a). 
 
Compared with system dynamics in $\M^*$, the model $\M^{(1)}$ coincides with $\M^*$ in the observational distributions ($\1L_1$), but deviates significantly in the interventional distributions ($\1L_2$). More specifically, given any observations and actions' history $\bar{\*o}_{1:i}, \bar{\*x}_{1:i}$, the induced reward function in $\M^{(1)}$ is given by,
\begin{align}
    \E^{(1)}_{\bar{\*x}_{1:i}}\left[ Y_i \mid \bar{\*o}_{1:i}\right] &= \E^{(1)}\left[ Y_i \mid \bar{\*o}_{1:i}, \bar{\*x}_{1:i} \right]\\
    &= 0.1
\end{align}
On the other hand, suppose the state occupancy rate $P(s_i) = 0.5$ for any state $s_i = 0, 1$ in the model $\M^*$. Evaluating the transition distribution in $\M^*$ gives, for any action $x_i = 0, 1$ and any state $s_{i+1} = 0, 1$,
\begin{align}
    P_{x_i}\Parens{s_{i+1}} = 0.5.
\end{align}
For instance, let $x_i = 0$ and $s_{i+1} = 1$. Expanding on the current state $S_i$ gives,
\begin{align}
    &P_{X_i \gets 0}\Parens{S_{i+1} = 1}\\
    &=  P_{X_i \gets 0}\Parens{S_{i+1} = 1 \mid S_i = 0} P(S_i = 0) +  P_{X_i \gets 0}\Parens{S_{i+1} = 1 \mid S_i = 1} P(S_i = 1)\\
    &= 0.82 \times 0.5 + 0.18 \times 0.5\\
    &=0.5
\end{align}
The above equations imply the following intermediate reward evaluated in the model $\M^*$
\begin{align}
    \E^{*}_{\bar{\*x}_{1:i}}\left[ Y_i \mid \bar{\*o}_{1:i}\right] &= \sum_{s_i} \E^{*}_{x_i}\left[ Y_i \mid s_i \right] P^*_{\bar{\*x}_{1:i-1}}(s_i \mid \bar{\*o}_{1:i})\\
    &=0.5,
\end{align}
which deviates from the corresponding query in $\M^{(1)}$. For instance, let $x_i = 0$. Expanding on the current state $S_i$ gives,
\begin{align}
    &\E^{*}_{\bar{\*x}_{1:i}}\left[ Y_i \mid \bar{\*o}_{1:i}\right] \\
    &= \E^{*}_{X_i \gets 0}\left[ Y_i \mid S_i = 0 \right] P^*_{\bar{\*x}_{1:i-1}}(S_i = 0 \mid \bar{\*o}_{1:i}) + \E^{*}_{X_i \gets 0}\left[ Y_i \mid S_i = 1 \right] P^*_{\bar{\*x}_{1:i-1}}(S_i = 1 \mid \bar{\*o}_{1:i})\\
    &= 0.82 P^*_{\bar{\*x}_{1:i-1}}(S_i = 0 \mid \bar{\*o}_{1:i}) + 0.18 P^*_{\bar{\*x}_{1:i-1}}(S_i = 1 \mid \bar{\*o}_{1:i})\\
    &=0.5
\end{align}
The last step holds since the occupancy rate $P^*_{\bar{\*x}_{1:i-1}}(s_i \mid \bar{\*o}_{1:i}) = 0.5$ for any state $s_i = 0, 1$. This example corroborates Prop.~\ref{prop:_3_4_pomdp_obs} and illustrates that observational queries are generally under-determined by randomized experiments in POMDP environments. \hfill $\blacksquare$
\end{example} 
The above example shows that interventional distributions in an unknown POMDP environment are generally not fully determined from the observational distribution. Conversely, we also show that it is generally infeasible to recover observational quantities from randomized experiments in non-Markov processes.
\begin{proposition}\label{prop:_3_4_mdp_inv}
	For any SCM $\1M^*$ compatible with the causal diagram $\G_{\textsc{pomdp}}$ of Fig.~\ref{fig:_3_1_pomdp}, there is an SCM $\1M^{(2)}$ compatible with $\G_{\textsc{pomdp}}$ such that for every stage $i = 1, 2, \dots$,
	\begin{align}
		&P^{(2)}_{\bar{\*x}_{1:i}}\left ( o_{i+1} \mid \bar{\*o}_{1:i} \right ) = P^{*}_{\bar{\*x}_{1:i}}\left ( o_{i+1} \mid \bar{\*o}_{1:i} \right ), &&\E^{(2)}_{\bar{\*x}_{1:i}}\left[ Y_i \mid \bar{\*o}_{1:i}\right] = \E^{*}_{\bar{\*x}_{1:i}}\left[ Y_i \mid \bar{\*o}_{1:i}\right]\label{eq:_3_3_mdp_inv1}
	\end{align}
	while 
	\begin{align}
		&P^{(1)}\left ( o_{i+1} \mid \bar{\*o}_{1:i}, \bar{\*x}_{1:i} \right ) \neq P^{*}\left ( o_{i+1} \mid \bar{\*o}_{1:i}, \bar{\*x}_{1:i}\right ), &&\E^{(1)}\left[ Y_i \mid \bar{\*o}_{1:i}, \bar{\*x}_{1:i} \right] \neq \E^{*}\left[ Y_i \mid \bar{\*o}_{1:i}, \bar{\*x}_{1:i} \right]  \label{eq:_3_3_mdp_inv2}
	\end{align} \hfill $\blacksquare$
\end{proposition}
The following example corroborates the proposition mentioned above by constructing an alternative SCM $\1M^{(2)}$ compatible with the causal diagram of Fig.~\ref{fig:_3_1_pomdp} that induces the same interventional distribution in the underlying environment but generates different observations. 
\begin{example}[POMDP, Interventional $\not \Rightarrow$ Observational]\label{exp:_3_4_inv-pomdp}
\sloppy Consider the following SCM environment rolling over stages $i = 1, 2, \dots$,
\begin{align}
\M^{(2)} =
 \Tuple{\*U = \{U_{i, 1}, U_{i,2}, U_{i, 3}\}, \*V =  \{X_i, Y_i, S_i, O_i\}, \2F = \left \{\2F^{(2)}_i \right \}, P^{(2)}(\*U)}_{i = 1, 2, \dots}. 
\end{align}
Similar to the previous example, the above SCM is identical to the model $\M^{(2)}$ defined in Example~\ref{exp:_2_2_inv-mdp} except that for every stage $i = 1, 2, \dots$, the learner does not observe the state $S_i$, but only receives new endogenous variable $O_i \gets 0$. We compute the observational distributions $P\Parens{S_{i+1} \mid \bar{\*O}_{1:i}, \bar{\*X}_{1:i}}$ and $\3E \Brackets{Y_i \mid \bar{\*O}_{1:i}, \bar{\*X}_{1:i}}$ and the interventional distributions $\inv{S_{i+1} \mid \bar{\*O}_{1:i}}{\bar{\*X}_{1:i}}$ and $\invE{Y\mid \bar{\*O}_{1:i}}{\bar{\*X}_{1:i}}$ evaluated in $\1M^{(2)}$, following the discussion in Example~\ref{exp:_2_2_inv-mdp}. The analysis suggests that the observational and interventional distributions coincide in the model $\1M^{(2)}$, which could be described using the same finite automaton of Fig.~\ref{fig:_3_4_cht_pomdp}(b). 
 
Comparing with system dynamics in the SCM $\M^*$ described in Example~\ref{exp:_3_4_obs-pomdp}, we find that model $\M^{(2)}$ coincides with $\M^*$ in the interventional distributions ($\1L_2$), but disagree in the observational distributions ($\1L_1$). More specifically, given any observations and actions' history $\bar{\*o}_{1:i}, \bar{\*x}_{1:i}$, the observed intermediate reward in $\M^{(2)}$ is given by,
\begin{align}
    \E^{(2)}\left[ Y_i \mid \bar{\*o}_{1:i}, \bar{\*x}_{1:i} \right] &= \E^{(2)}_{\bar{\*x}_{1:i}}\left[ Y_i \mid \bar{\*o}_{1:i}\right]\\
    &= 0.5
\end{align}
On the other hand, evaluating the observed intermediate reward given the history $\bar{\*o}_{1:i}, \bar{\*x}_{1:i}$ in the model $\M^*$ gives
\begin{align}
    \E^{*}\left[ Y_i \mid \bar{\*o}_{1:i}, \bar{\*x}_{1:i} \right] = 0.1,
\end{align}
which differs from the corresponding query in $\M^{(2)}$. This complements previous examples and illustrates that interventional queries are generally non-identifiable from the observational data in POMDP environments. \hfill $\blacksquare$
\end{example}
We organize the examples and results discussed in this section and summarize them in Fig.~\ref{fig:_3_4_cht_pomdp}. The agent interacts with the ground-truth SCM $\M^*$ (Example~\ref{exp:_3_4_obs-pomdp}) in the middle, through passive observation or active intervention, and generates the observational and interventional distributions. The system dynamics of these distributions do not satisfy the Markov property and can be represented using the latent states in standard POMDPs. The finite automata in Fig.~\ref{fig:_3_4_cht_pomdp} (a, b) describes these latent dynamics, respectively.

Assuming only observational data ($\1L_1$) is available, one can construct an alternative SCM $\M^{(1)}$ (left side) that generates the same the observational data but have different interventional distributions (i.e., $\1L^{*}_1 = \1L^{(1)}_1$, $\1L^{*}_2 \neq \1L^{(1)}_2$). This implies that, in practice, natural trajectories of other behavioral agents collected from passive observations are generally insufficient to make claims about the learning agent's actions and performance. Conversely, whenever the interventional data ($\1L_2$) is available, one can construct an alternative SCM $\M^{(2)}$ (right side) that generates the same interventional distribution but has a different observational one (i.e., $\1L^{*}_1 \neq \1L^{(2)}_1$, $\1L^{*}_2 = \1L^{(2)}_2$). This might seem counterintuitive, as interventions are generally thought to be more informative than simply observing a system as it evolves over time. However, in practice, this approach does not enable the learning agent to predict how other agents will behave within the same environment.

\newpage
\section{Detailed Computation of CDM Planning}  \label{app:_compute}

\section{Offline-to-Online Learning in Thompson Sampling}  \label{app:_ts}

\section{Extended Dynamic Programming}  \label{app:_extended_q}

\section{Appendix x. } 

Model-free versus Model-based RL orthogonal to causal-model-based 

\section{Appendix x. }  \label{app:mdp-mab}
 \label{appendix-faq}
\section{Counterfactual Distributions and Axioms}\label{app:_ctf}
This section will review the formal definition of counterfactual distributions in the language of structural causality and related axioms for inference. Specifically, counterfactual reasoning refers to thinking about alternative ways the world could be, including ways that might conflict with how the world, in fact, currently is. For instance, if the patient took the aspirin and the headache was cured, would the headache still be gone had they not taken the drug? Counterfactual queries can be directly determined from a fully specified structural causal model, as described below.
\begin{definition}[Counterfactual Distribution]\label{def:_app_3_ctf}
	An SCM $\M= \tuple{\*U, \*V, \2F, P}$ induces a family of joint distributions over counterfactual events $\*Y_{\pi_{\*X}}, \ldots, \*Z_{\pi_{\*W}}$, for any $\*Y, \*Z, \dots, \*X, \*W \subseteq \*V$:
	\begin{align}
		P\Parens{\*y_{\pi_{\*X}}, \dots, \*z_{\pi_{\*W}}} \equiv \sum_{\*u} \I\left \{ \*Y_{\pi_{\*X}}(\*u) = \*y, \dots, \*Z_{\pi_{\*W}}(\*u) = \*w\right\}P(\*u), \label{eq:_app_3_ctf}
	\end{align}
\end{definition}
Note that the l.h.s. of Eq.~\ref{eq:_app_3_ctf} contains variables with different subscripts, which, syntactically, encode different counterfactual ``worlds.'' The evaluation implied by this equation can be described as the following process:
\begin{enumerate}
	\item For each set of subscripts relative to each set of variables (e.g., $\pi_{\*X}, \dots, \pi_{\*W}$ for $\*Y, \dots, \*Z$, respectively), replace the corresponding mechanisms with the appropriate constants and generate  $\2F_{\pi_{\*X}}, \dots, \2F_{\pi_{\*W}}$ (Eq.~\ref{def:_2_2_submodel}), creating submodels $\M_{\pi_{\*X}},\dots, \M_{\pi_{\*W}}$ (of $\M$);
	\item For each unit $\*U = \*u$, Nature evaluates the modified mechanisms (e.g., $\2F_{\pi_{\*X}}, \dots, \2F_{\pi_{\*W}}$) following a valid order (i.e., any variable in the l.h.s. is evaluated after the ones in the r.h.s.) to obtain the potential responses of the observables, and
	\item  The probability mass $P(\*U = \*u)$ is then accumulated for each instantiation $U = u$ that is consistent with the events over the counterfactual variables -- for instance, ${\*Y}_{\pi_{\*X}}={\*y}\dots, {\*Z}_{\pi_{\*W}}={\*z}$, i.e., ${\*Y} = {\*y}, \dots, {\*Z} = {\*z}$ in the submodels $\M_{\pi_{\*X}}, \dots, \M_{\pi_{\*W}}$, respectively.
\end{enumerate}
As a special case, we denote by $P \left(\*Y_{\*x}, \dots, \*Z_{\*w}\right)$ the counterfactual distributions entailed by atomic interventions $\doo(\*x), \dots, \doo(\*w)$, which is a joint distribution over variables $\*Y, \dots, \*Z$ in submodels $\M_{\*x}, \dots, \1M_{\*w}$ respectively.
\begin{example}\label{exp_app_3_mab}
	Consider the MAB model $\M^*$ described in Eq.~\ref{eq:_2_1_mab}. Since there is a group of patients who did not receive the treatment and died ($X=0, Y=0$), one may wonder whether these patients would have been alive ($Y=1$) had they been given the treatment ($X=1$). This counterfactual question is written as $P(Y_{X\gets 1}=1 \mid X=0,Y=0)$. Note that
	\begin{align}
		P(Y_{X\gets 1}=1 \mid X=0,Y=0) = \frac{P(Y_{X\gets 1}=1, X=0,Y=0)}{P(X=0,Y=0)}
	\end{align}
	where the denominator is trivially obtainable since it only involves observational probabilities. Following the calculation in Eq.~\ref{exp:_2_3_mab} we obtain
	\begin{align}
		P(X = 0, Y= 0) = P(U \geq 0.2) = 0.2
	\end{align}
	On the other hand, the numerator, $P(Y_{X \gets 1}=1, X=0,Y=0)$, refers to two different worlds and cannot be written in the languages of observational and interventional distributions since they do not allow for probability expressions involving more than one subscript (each encoding a different world).  Using the procedure dictated in Eq.~\ref{eq:_app_3_ctf}, we obtain
	\begin{align}
		P(Y_{X \gets 1}=1, X=0,Y=0) & = \int_{0}^1 \I\{u < 0.4 - \Delta, u \geq 0.8, u \geq 0.4\} du \\
		                            & = P(U < 0.4 - \Delta, U \geq 0.8) \label{eq:_app_3_mab1}
	\end{align}
	Since $\Delta \in (0, 0.4)$, compute Eq.~\ref{eq:_app_3_mab1} gives, $P(Y_{X \gets 1}=1, X=0,Y=0) = 0$. This implies
	\begin{align}
		P(Y_{X\gets 1}=1 \mid X=0,Y=0) = \frac{P(Y_{X\gets 1}=1, X=0,Y=0)}{P(X=0,Y=0)} = 0
	\end{align}
	The conclusion following this counterfactual analysis is clear: even if we had given the treatment to everyone who did not survive, none of them would have survived. In other words, the treatment would not have prevented their death.
\end{example}
Counterfactual variables follow three properties of composition, effectiveness, and reversibility, which hold in all structural causal models.
\begin{theorem}[Counterfactual Axioms]\label{thm:_app_3_ctf}
	Let $\M= \tuple{\*U, \*L, \*V, \2F, P}$ be an SCM. For any three sets of variables $\*X, \*Y, \*W \subseteq \*V$, the following properties hold
	\begin{enumerate}
		\item \textbf{Composition.} $\*W_{\*x}(\*u) = \*w \Rightarrow \*Y_{\*x,\*w}(\*u) = \*Y_{\*x}(\*u)$;
		\item \textbf{Effectiveness.} $\*X_{\*x,\*w}(\*u) = \*x$;
		\item \textbf{Reversibility} $\Parens{\*Y_{\*x, \*w}(\*u) = \*y} \& \Parens{\*W_{\*x, \*y}(\*u) = \*w} \Rightarrow \*Y_{\*x}(\*u) = \*y$.
	\end{enumerate}
\end{theorem}

\section{Proofs for Chapter 4}\label{app:_sec_4}
\thmipw*
\begin{proof}
	Note that the interventional distribution $\inv{\*v}{\pi}$ is the joint distribution over $\*Y$ evaluated in submodel $\1M^*_{\pi}$. By factorizing over variables in submodel $\1M^*_{\pi}$ we have
	\begin{align}
		\inv{\*v}{\pi} & = \sum_{\*u} \prod_{V \in \*V \setminus \*X} P(v \mid \*\pa_V, \*u_V) \prod_{i = 1}^H \pi_i(x_i \mid \*s_i)P(\*u)  \\
		               & = \prod_{i = 1}^H \pi_i(x_i \mid \*s_i) \sum_{\*u} \prod_{V \in \*V \setminus \*X} P(v \mid \*\pa_V, \*u_V) P(\*u)
	\end{align}
	The last step holds since actions and states $\*X \cup \*S \subseteq \*V$ are all observed. Summing over unobserved variables $\*U$ gives
	\begin{align}
		\inv{\*v}{\pi} = \prod_{i = 1}^H \pi_i(x_i \mid \*s_i) \inv{\*v \setminus \*x}{\*x} \label{eq:_app_4_ipw2}
	\end{align}
	Similarly, factorizing variables in SCM $\1M^*$ gives
	\begin{align}
		P(\*v) = \sum_{\*u} \prod_{V \in \*V \setminus \*X} P(v \mid \*\pa_V, \*u_V) \prod_{i = 1}^H P(x_i \mid \*\pa_i, \*u_i) P(\*u)
	\end{align}
	Since the NUC holds, Condition (1) of Def.~\ref{def:_4_1_nuc} says that for every action $X_i$, endogenous parents $\*\PA_i \subseteq \bar{\*X}_{i-1}\cup \bar{\*S}_i$. The above equation can be further written as
	\begin{align}
		P(\*v) = \sum_{\*u} \prod_{V \in \*V \setminus \*X} P(v \mid \*\pa_V, \*u_V) \prod_{i = 1}^H P(x_i \mid \bar{\*x}_{i-1}, \bar{\*s}_i, \*u_i) P(\*u)
	\end{align}
	Condition (2) of Def.~\ref{def:_4_1_nuc} says that for every action $X_i$, exogenous parents $\*U_i$ are indpendent noises affecting only $X_i$. We must have
	\begin{align}
		P(\*v) & = \prod_{i = 1}^H \sum_{\*u_i} P(x_i \mid \bar{\*x}_{i-1}, \bar{\*s}_i, \*u_i)P(\*u_i) \sum_{\*u} \prod_{V \in \*V \setminus \*X} P(v \mid \*\pa_V, \*u_V)  P(\*u) \\
		       & = \prod_{i = 1}^H P(x_i \mid \bar{\*x}_{i-1}, \bar{\*s}_i) \inv{\*v \setminus \*x}{\*x} \label{eq:_app_4_ipw1}
	\end{align}
	Eq.~\ref{eq:_app_4_ipw1} implies the interventional distribution $\inv{\*V \setminus \*X}{\*x}$ is identifiable from the observational distribution $P(\*V)$ and is given by
	\begin{align}
		\inv{\*v \setminus \*x}{\*x} = P(\*v)\prod_{i = 1}^H \frac{1}{P(x_i \mid \bar{\*x}_{i-1}, \bar{\*s}_i)} \label{eq:_app_4_ipw3}
	\end{align}
	Eqs.~\ref{eq:_app_4_ipw2} and \ref{eq:_app_4_ipw3} give
	\begin{align}
		\inv{\*v}{\pi} = P(\*v)\prod_{i = 1}^H \frac{\pi_i(x_i \mid \*s_i)}{P(x_i \mid \bar{\*x}_{i-1}, \bar{\*s}_i)}
	\end{align}
	The expected reward $\invE{\1R(\*Y)}{\pi}$ can thus be written as
	\begin{align}
		\invE{\1R(\*Y)}{\pi} & = \sum_{\*v \setminus \*x, \*s} \1R(\*y) \inv{\*v}{\pi}                                                                                \\
		                     & =  \sum_{\*v \setminus \*x, \*s} \1R(\*y) P(\*v)\prod_{i = 1}^H \frac{\pi_i(x_i \mid \*s_i)}{P(x_i \mid \bar{\*x}_{i-1}, \bar{\*s}_i)}
	\end{align}
	Summing over domains of variables $\*V \setminus (\*X \cup \*S)$ gives the IPW adjustment formula.
\end{proof}

\thmdp*
\begin{proof}
	Let function $Q_{\pi}^{(H)} (\bar{\*x}_{H}, \bar{\*s}_H) = \E\left[\1R(\*Y) \mid \bar{\*x}_H, \bar{\*s}_H\right]$. The expected reward $\invE{\1R(\*Y)}{\pi}$  can thus be written as, following the decomposition in Eq.~\ref{eq:_4_1_dp_derivation},
	\begin{align}
		\invE{\1R(\*Y)}{\pi} & = \sum_{\bar{\*x}_H, \bar{\*s}_H} Q_{\pi}^{(H)} (\bar{\*x}_{H}, \bar{\*s}_H) \prod_{i = 1}^H  P\left ( s_i  \mid \bar{\*x}_{i-1}, \bar{\*s}_{i-1}\right)  \pi_{i} \left(x_i \mid \*s_i\right) \\
		                     & =\sum_{\bar{\*x}_{H-1}, \bar{\*s}_{H-1}} \left( \sum_{x_H, \*s_H}  Q_{\pi}^{(H)} (\bar{\*x}_{H}, \bar{\*s}_H) \pi_H(x_H \mid \*S_H) P(\*s_i \mid \bar{\*x}_{H-1}, \bar{\*s}_{H-1}) \right)           \\
		                     &\cdot \prod_{i = 1}^{H-1}  P\left ( s_i  \mid \bar{\*x}_{i-1}, \bar{\*s}_{i-1}\right)  \pi_{i} \left(x_i \mid \*s_i\right)
	\end{align}
	Summing over domains of states $\*S_H$ and $\*X_H$ gives
	\begin{align}
		\invE{\1R(\*Y)}{\pi} = \sum_{\bar{\*x}_{H-1}, \bar{\*s}_{H-1}} Q_{\pi}^{(H-1)} (\bar{\*x}_{H-1}, \bar{\*s}_{H-1}) \prod_{i = 1}^{H-1}  P\left ( s_i  \mid \bar{\*x}_{i-1}, \bar{\*s}_{i-1}\right)  \pi_{i} \left(x_i \mid \*s_i\right), \label{eq:_app_4_dp1}
	\end{align}
	where function $Q_{\pi}^{(H-1)} (\bar{\*x}_{H-1}, \bar{\*s}_{H-1})$ is defined as
	\begin{align}
		Q_{\pi}^{(H-1)} (\bar{\*x}_{H-1}, \bar{\*s}_{H-1}) & = \sum_{x_H, \*s_H} Q_{\pi}^{(H)} (\bar{\*x}_{H}, \bar{\*s}_H) \pi_H(x_H \mid \*S_H) P(\*s_i \mid \bar{\*x}_{H-1}, \bar{\*s}_{H-1})                                      \\
		                                                   & = E \left[\sum_{x_{H}}  Q_{\pi}^{(H)} (\bar{\*x}_{H}, \bar{\*s}_{H-1}, \*S_{H}) \pi_{H}\left(x_{H} \mid \*S_{H} \right) \mmid \bar{\*x}_{H-1}, \bar{\*s}_{H-1}  \right ]
	\end{align}
	By repeatedly applying the above marginalization procedure, we obtain
	\begin{align}
		\invE{\1R(\*Y)}{\pi} & = \sum_{x_1, \*s_1}  Q_{\pi}^{(1)}(x_1, \*s_1) \pi_1(x_1\mid \*s_1) P(\*s_1)  \\
		                     & = \E \left[ \sum_{x_1} Q_{\pi}^{(1)}(x_1, \*S_1) \pi_1(x_1\mid \*S_1)\right],
	\end{align}
	where function $Q_{\pi}^{(i)} (\bar{\*x}_{i}, \bar{\*s}_i)$ is given by, for $i = 1, \dots, H-1$,
	\begin{align}
		Q_{\pi}^{(i)} (\bar{\*x}_{i}, \bar{\*s}_i) = \E\left [\sum_{x_{i+1}}  Q_{\pi}^{(i+1)} (\bar{\*x}_{i+1}, \bar{\*s}_i, \*S_{i+1}) \pi_{i+1}\left(x_{i+1} \mid \*S_{i+1} \right) \mmid \bar{\*x}_{i}, \bar{\*s}_i \right].
	\end{align}
	This proves the DP identification formula.
\end{proof}

\lemonlinenuc*
\begin{proof}
	We will examine the NUC condition for every action $X_i \in \*X$ and show that it holds in the decision model $\Tuple{\1M^*_{\pi}, \Pi, \1R}$ induced by intervention $\doo(\pi)$. Recall that intervention $\doo(\pi)$ replace the function $f_{X_i}$ associated with every action $X_i \in \*X$ in the SCM $\1M^*$ with the decision rule $\pi(X_i \mid \*S_i)$. For every $i = 1, \dots, H$, input states $\*S_i$ are observed parents of action $X_i$ in the submodel $\1M^*_{\pi}$; Condition (1) of Def.~\ref{def:_4_1_nuc} holds. Also, in the submodel $\1M^*_{\pi}$, the exogenous parent $U_i$ for every $i = 1, \dots, H$, is an independent noise only affecting action $X_i$; Condition (1) of Def.~\ref{def:_4_1_nuc} holds.
\end{proof}

\thmsbc*
\begin{proof}
	By summing over domains of action $X_1$ and states $\*S_1$, the interventional distribution $\inv{\*Y}{\pi}$ for every policy $\pi \in \Pi$ can be written as
	\begin{align}
		\inv{\*y}{\pi} & = \sum_{x_1, \*s_1} \inv{\*y \mid x_1, \*s_1}{\pi} \inv{x_1 \mid \*s_1}{\pi} \inv{\*s_1}{\pi}             \\
		               & = \sum_{x_1, \*s_1} \inv{\*y \mid \*s_1}{x_1, \pi_2, \dots, \pi_H} \pi_1(x_1 \mid \*s_1) \inv{\*s_1}{\pi}
	\end{align}
	The last step holds since in submodel $\1M^*_{\pi}$, values of $X_1$ are determined by the decision rule $\pi_1(X_1 \mid \*S_1)$. Note that states $\*S_1$ are non-descendants of actions $X_1 \dots, X_H$ in the manipulated diagram $\G_{\pi}$. The above equation can be further written as
	\begin{align}
		\inv{\*y}{\pi} & = \sum_{x_1, \*s_1} \inv{\*y \mid \*s_1}{x_1, \pi_2, \dots, \pi_H} \pi_1(x_1 \mid \*s_1) P(\*s_1) \label{eq:_app_4_sbc1}
	\end{align}
	Since the policy space $\Pi$ is backdoor-admissible, this means that conditioning on nodes $\*S_1$ blocks all backdoor path from $X_1$ to reward singals $\*Y$ in the manipulated diagram $\G_{\pi}$, i.e.,
	\begin{align}
		\left(X_1 \ci \*Y \mid \*S_1 \right)_{\G_{\underline{X_1},\pi_{2}, \dots, \pi_H}}
	\end{align}
	Applying Rule 2 of do-calculus in Def.~\ref{def:_4_3_do-calculus} implies the following
	\begin{align}
		\inv{\*y \mid \*s_1}{x_1, \pi_{2}, \dots, \pi_H} = \inv{\*y \mid x_1, \*s_1}{\pi_{2}, \dots, \pi_H} \label{eq:_app_4_sbc2}
	\end{align}
	Eqs.~\ref{eq:_app_4_sbc1} and \ref{eq:_app_4_sbc2} imply
	\begin{align}
		\inv{\*y}{\pi} & = \sum_{x_1, \*s_1} \inv{\*y \mid x_1, \*s_1}{\pi_{2}, \dots, \pi_H} \pi_1(x_1 \mid \*s_1) P(\*s_1) \label{eq:_app_4_sbc3}
	\end{align}
	By repeatedly applying procedures for actions $X_2, \dots, X_H$ we obtain
	\begin{align}
		\inv{\*y}{\pi} = \sum_{\bar{\*x}_H, \bar{\*s}_H} P(\*y \mid \bar{\*x}_H, \bar{\*s}_H) \prod_{i = 1}^H  P\left ( s_i  \mid \bar{\*x}_{i-1}, \bar{\*s}_{i-1}\right)  \pi_{i} \left(x_i \mid \*s_i\right)
	\end{align}
	The expected reward $\invE{\1R(\*Y)}{\pi}$ of every policy $\pi \in \Pi$ can thus be identified from the observational distribution $P(\*V)$ and is given by
	\begin{align}
		\E_{\pi}\left [\1R(\*Y)\right] = \sum_{\bar{\*x}_H, \bar{\*s}_H} \E\left[\1R(\*Y) \mid \bar{\*x}_H, \bar{\*s}_H \right]\prod_{i = 1}^H  P\left ( s_i  \mid \bar{\*x}_{i-1}, \bar{\*s}_{i-1}\right)  \pi_{i} \left(x_i \mid \*s_i\right)
	\end{align}
	The above identification formula is equivalent to Thms.~\ref{thm:_4_1_ipw} and \ref{thm:_4_1_dp}. Details of the derivations follow the discussion in Sec.~\ref{sec:_4_1_1}.
\end{proof}
\section{Proofs for Chapter 5}\label{app:_sec_5}

\thmucbandit*
\begin{proof}
	Recall that $N_t(x)$ denotes the number of times an arm $x$ was chosen up to episode $t$. The cumulative regret $\1R(T, \1M^*)$ could be decomposed as
	\begin{align}
		\1R(T, \1M^*) = \sum_{x \in \D(X)} \Delta_x \E\left[ N_T(x)\right]
	\end{align}
	For any arm $x$, we upper bound $N_T(x)$ on any sequence of interventions. We will discuss cases based on the causal upper bound $\invE{Y}{x} \leq r_x$.

	\paragraph{If $r_x < \mu^*$.} In this case, the indicator function of $X^{(t)} = x$ is bounded as follows
	\begin{align}
		N_T(x) & = \sum_{t = 1}^{T} \I\Braces{X^{(t)} = x}                                                                                                                             \\
		       & \leq \sum_{t = 1}^{T} \I\Braces{\overline{\text{UCB}}_t(x^*, \delta) < \mu^*} + \I\Braces{\overline{\text{UCB}}_t(x^*, \delta) \geq \mu^*, X^{(t)} = x}               \\
		       & \leq \sum_{t = 1}^{T} \I\Braces{\overline{\text{UCB}}_t(x^*, \delta) < \mu^*} + \I\Braces{\overline{\text{UCB}}_t(x, \delta) \geq \mu^*} \label{eq:_app_5_ucb_bandit1}
	\end{align}
	The last step holds since an arm $x$ is played at time step $t$ if and only if $\overline{\text{UCB}}_t(x, \delta)$ is maximal. Recall that the clipped confidence bound is compatible with the causal bound $[l_x, r_x]$, i.e.,
	\begin{align}
		l_x \leq \overline{\text{UCB}}_t(x, \delta) \leq r_x
	\end{align}
	Since $r_x < \mu^*$, we must have $\overline{\text{UCB}}_t(x, \delta) < \mu^*$. Eq.~\ref{eq:_app_5_ucb_bandit1} could thus be written as
	\begin{align}
		N_T(x) & \leq \sum_{t = 1}^{T} \I\Braces{\overline{\text{UCB}}_{t-1}(x^*, \delta) < \mu^*} \\
		       & \leq \sum_{t = 1}^{T} \I\Braces{\text{UCB}_{t-1}(x^*, \delta) < \mu^*}
	\end{align}
	So we get
	\begin{align}
		\E\Brackets{N_T(x)} & \leq \sum_{t = 1}^T P \Parens{\text{UCB}_{{t-1}}(x^*, \delta) < \mu^*}                                                                   \\
		                    & \leq  \sum_{t = 1}^T P\Parens{\hat{\E}^{(t-1)}_{x^*}[Y] + \sqrt{\frac{\log(1/\delta)}{2N_{t-1}(x^*)}} < \mu^*}                           \\
		                    & \leq \sum_{t = 1}^T \sum_{2N_{t-1}(x^*) = 1}^t P\Parens{\hat{\E}^{(t-1)}_{x^*}[Y] + \sqrt{\frac{\log(1/\delta)}{2N_{t-1}(x^*)}} < \mu^*} \\
		                    & \leq  \sum_{t = 1}^T \sum_{2N_{t-1}(x^*) = 1}^t \delta \label{eq:_app_5_ucb_bandit2}
	\end{align}
	The last step follows Hoeffding’s inequalities (Eq.~\ref{eq:_4_2_ucb1}). Since $\delta = t^{-4}$, we must have
	\begin{align}
		\E\Brackets{N_T(x)} & \leq \sum_{t = 1}^T \sum_{n}^{t-1} \frac{1}{t^4}  \\
		                    & \leq \sum_{t = 1}^T \frac{1}{t^3}                 \\
		                    & \leq \frac{\pi^2}{6} \label{eq:_app_5_ucb_bandit3}
	\end{align}

	\paragraph{If $r_x \geq \mu^*$.} Let $l$ be an arbitrary positive integer. The indicator function of $X^{(t)} = x$ is bounded as
	\begin{align}
		N_T(x) & = \sum_{t = 1}^{T} \I\Braces{X^{(t)} = x}                                                                                                    \\
		       & \leq l + \sum_{t = 1}^{T} \I\Braces{\overline{\text{UCB}}_{t-1}(x^*, \delta) \leq \overline{\text{UCB}}_{t-1}(x, \delta), N_{t-1}(x) \geq l}
	\end{align}
	Now observe that $\overline{\text{UCB}}_{t-1}(x^*, \delta) \leq \overline{\text{UCB}}_{t-1}(x, \delta)$ implies that at least one of the following events must hold true
	\begin{align}
		 & \hat{\E}_{x^*}^{(t-1)} + \sqrt{\frac{\log(1/\delta)}{2N_{t-1}(x^*)}} \leq \invE{Y}{x^*} \\
		 & \hat{\E}_{x}^{(t-1)} - \sqrt{\frac{\log(1/\delta)}{2N_{t-1}(x)}} \geq \invE{Y}{x}       \\
		 & \invE{Y}{x^*} < \invE{Y}{x} + 2\sqrt{\frac{\log(1/\delta)}{2N_{t-1}(x)}}
	\end{align}
	It follows from \cite[Theorem 1]{Auer2002} that
	\begin{align}
		\E\Brackets{N_T(x)} & \leq \frac{8\log(T)}{\Delta_x^2} + 1 + \frac{\pi^2}{3} \label{eq:_app_5_ucb_bandit4}
	\end{align}
	By applying Eqs.~\ref{eq:_app_5_ucb_bandit2} and \ref{eq:_app_5_ucb_bandit3}, we bound the cumulative regret as
	\begin{align}
		\1R(T, \1M^*) & = \sum_{x: r_x < \mu^*} \Delta_x \E\left[ N_T(x)\right] + \sum_{x: r_x \geq \mu^*} \Delta_x \E\left[ N_T(x)\right]                                 \\
		              & \leq \sum_{x: r_x < \mu^*} \frac{\pi^2}{6}\Delta_x + \sum_{x: r_x \geq \mu^*} \frac{8\log(T)}{\Delta_x} + \left(1 + \frac{\pi^2}{3}\right)\Delta_x
	\end{align}
	Simplifying the above equation completes the proof.
\end{proof}

\thmucbdtr*
\begin{proof}
	The cumulative regret could be written as
	\begin{align}
		R(T, \1M^*) & = \sum_{t = 1}^T \invE{\1R{(\*Y)}; \1M^*}{\pi^*} - \E \Brackets{\1R\Parens{\*Y^{(t)} }} \\
		            & = \E \Brackets{\sum_{t = 1}^T \Delta_{\pi^{(t)}} }
	\end{align}
	Let $\epsilon \geq 0$ be an arbitrary real value. The sum of performance gaps decomposes as follows
	\begin{align}
		\sum_{t = 1}^T \Delta_{\pi^{(t)}} & \leq  \sum_{t = 1}^T\I\Braces{\M^* \not \in \3M_{t-1}(\delta)} + \sum_{t = 1}^T \Delta_{\pi^{(t)}} \I\Braces{\M^* \in \3M_{t-1}(\delta) }                        \\
		                                  & \leq \underbrace{\sum_{t = 1}^T\I\Braces{\M^* \not \in \3M_{t-1}(\delta)}}_{\text{Term 1}}                                                                       \\
		                                  & + \underbrace{\sum_{t = 1}^T \Delta_{\pi^{(t)}} \I\Braces{\Delta_{\pi^{(t)}} \leq \epsilon, \M^* \in \3M_{t-1}(\delta)}}_{\text{Term 2}}                         \\
		                                  & +  \underbrace{\sum_{t = 1}^T \Delta_{\pi^{(t)}} \I\Braces{\Delta_{\pi^{(t)}} > \epsilon, \M^* \in \3M_{t-1}(\delta)}}_{\text{Term 3}} \label{eq:_app_5_ucb_dtr8}
	\end{align}
	We will next bound the above terms separately.

	\paragraph{Bounding Term 1.} By applying a union bound over all possible values of state-action counts $N_t(\bar{\*x}_i, \bar{\*s}_i)$ for every $i = 1, \dots, H$, we obtain from Eq.~\ref{eq:_5_2_concentration}
	\begin{align}
		\sum_{t = 1}^T P\Parens{\M^* \not \in \3M_{t-1}(\delta)} & \leq \sum_{t = 1}^T \sum_{n = 1}^{t-1} P\Parens{\M^* \not \in \3M_{t-1}(\delta)} \\
		                                                         & \leq \sum_{t = 1}^T \sum_{n = 1}^{t-1} \delta
	\end{align}
	Since $\delta = t^{-4}$, the above equation could be further written as
	\begin{align}
		\sum_{t = 1}^T P\Parens{\M^* \not \in \3M_{t-1}(\delta)} \leq \sum_{t = 1}^T \sum_{n = 1}^{t-1} \frac{1}{t^4} \leq \frac{\pi^2}{6} \label{eq:_app_5_ucb_dtr_term1}
	\end{align}

	\paragraph{Bounding Term 2.} Observe that the algorithm only selects deterministic policies in $\Pi^{\mathrm{o}}$. This implies that for any policy $\pi^{(t)}$ such that $\Delta_{\pi^{(t)}} \leq \epsilon$, its performance gap is bounded by that of the worst deterministic policy $\pi \in \Pi^{\mathrm{o}}$ with $\Delta_{\pi} \leq \epsilon$. We thus have
	\begin{align}
		\sum_{t = 1}^T \Delta_{\pi^{(t)}} P\Parens{\Delta_{\pi^{(t)}} \leq \epsilon, \M^* \in \3M_{t-1}(\delta)} \leq \max_{\pi \in \Pi^{\mathrm{o}}: \Delta_{\pi} \leq \epsilon }\Delta{\pi} T \label{eq:_app_5_ucb_dtr_term2}
	\end{align}

	\paragraph{Bounding Term 3.} Let $\Delta(\epsilon, T)$ denote Term 3. Since $\1M^* \in \3M_{t-1}(\delta)$, the optimistic planning in Eq.~\ref{eq:_5_3_optimize} implies
	\begin{align}
		\Delta(\epsilon, T) & \leq \sum_{t = 1}^T \Parens{\invE{\1R(\*Y); \1M^*}{\pi^*} - \invE{\1R(\*Y); \1M^*}{\pi^{(t)}}} \I\Braces{\M^* \in \3M_{t-1}(\delta)} \\
		                    & \leq \sum_{t = 1}^T \invE{\1R(\*Y); \1M^{(t)}}{\pi^{(t)}} - \invE{\1R(\*Y); \1M^*}{\pi^{(t)}} \label{eq:_app_5_ucb_dtr1}
	\end{align}
	For convenience, we will use the superscript $P^{(t)}$ to denote a distribution evaluated in SCM $\1M^{(t)}$ and $P$ for a distribution evaluated in the true SCM $\1M^*$. Fix a policy $\pi \in \Pi$. For every $i = 0, \dots, H$, define $V^{(i)}_{\pi}(\bar{\*x}_{i}, \bar{\*s}_{i})$ as a function given by
	\begin{align}
		V^{(i)}_{\pi} = \sum_{\bar{\*x}_H, \bar{\*s}_H} \E^{(t)}_{\bar{\*x}_H}\Brackets{\1R(\*Y) \mid \bar{\*s}_H} & \prod_{j = i}^{H - 1} P^{(t)}_{\bar{\*x}_j}\Brackets{s_{j+1} \mid \bar{\*s}_j} \pi_{j+1}\left(x_{j+1} \mid \*s_{j+1}\right) \\
		                                                                                                           & \prod_{j = 0}^{i - 1} \inv{s_{j+1} \mid \bar{\*s}_j}{\bar{\*x}_j} \pi_{j+1}\left(x_{j+1} \mid \*s_{j+1}\right)
	\end{align}
	and define $V^{(H+1)}_{\pi} = \invE{\1R(\*Y); \1M^*}{\pi}$. Observe that $V^{(0)}_{\pi} = \invE{\1R(\*Y); \1M^{(t)}}{\pi}$. Eq.~\ref{eq:_app_5_ucb_dtr1} can thus be written as a telescoping sum
	\begin{align}
		\Delta(\epsilon, T) & \leq \sum_{t = 1}^T  \sum_{i = 0}^{H} V^{(i)}_{\pi^{(t)}} - \bar{V}^{(i+1)}_{\pi^{(t)}} \label{eq:_app_5_ucb_dtr4}
	\end{align}
	For every $i = 0, \dots, H-1$,
	\begin{align}
		V^{(i)}_{\pi^{(t)}} - V^{(i+1)}_{\pi^{(t)}} & \leq \sum_{\bar{\*x}_i, \bar{\*s}_i} \left \lVert \inv{\cdot \mid \bar{\*s}_i; \1M^{(t)} }{\bar{\*x}_i} - \inv{\cdot \mid \bar{\*s}_i; \1M^*}{\bar{\*x}_i} \right \rVert_1                                                             \\
		                                            & \leq 2\sqrt{ 2 \left |\D(\*S_{i+1}) \right | \log(4H\left |\D(\bar{\*S}_{i}\cup \bar{\*X}_{i}) \right | / \delta)} \sum_{\bar{\*x}_i, \bar{\*s}_i } \frac{1}{ \sqrt{N_t\Parens{\bar{\*x}_i, \bar{\*s}_i}  }} \label{eq:_app_5_ucb_dtr2}
	\end{align}
	and similarly,
	\begin{align}
		V^{(H)}_{\pi^{(t)}}- V^{(H+1)}_{\pi^{(t)}} & \leq \sum_{\bar{\*x}_H, \bar{\*s}_H }  \left \lvert \invE{\1R(\*Y) \mid \bar{\*s}_H; \1M^{(t)}}{\bar{\*x}_H} - \invE{\1R(\*Y) \mid \bar{\*s}_H; \1M^*}{\bar{\*x}_H}  \right \rvert                     \\
		                                           & \leq 2\sqrt{\log(8H \left  |\D(\bar{\*S}_{H}\cup \bar{\*X}_{H}) \right | / \delta)} \sum_{\bar{\*x}_H, \bar{\*s}_H }  \frac{1}{\sqrt{N_t\Parens{\bar{\*x}_H, \bar{\*s}_H}}}  \label{eq:_app_5_ucb_dtr3}
	\end{align}
	Eqs.~\ref{eq:_app_5_ucb_dtr2} and \ref{eq:_app_5_ucb_dtr3} imply
	\begingroup\allowdisplaybreaks \begin{align}
		 & \sum_{t = 1}^T  \sum_{i = 0}^{H} V^{(i)}_{\pi^{(t)}} - \bar{V}^{(i+1)}_{\pi^{(t)}}                                                                                                                                                           \\
		 & \leq \sum_{i = 0}^{H-1} 2\sqrt{2 \left |\D(\*S_{i+1}) \right | \log(4H\left |\D(\bar{\*S}_{i}\cup \bar{\*X}_{i}) \right | / \delta)}\sum_{\bar{\*x}_i, \bar{\*s}_i } \sum_{t = 1}^T \frac{1}{ \sqrt{N_t\Parens{\bar{\*x}_i, \bar{\*s}_i}  }} \\
		 & +\sqrt{2\log(8H \left  |\D(\bar{\*S}_{H}\cup \bar{\*X}_{H}) \right | / \delta)} \sum_{\bar{\*x}_H, \bar{\*s}_H } \sum_{t = 1}^T  \frac{1}{\sqrt{2N_t\Parens{\bar{\*x}_H, \bar{\*s}_H}}}                                                      \\
		 & \leq \sum_{i = 0}^{H-1} 4\sqrt{2 \left |\D(\*S_{i+1}) \right | \log(4H\left |\D(\bar{\*S}_{i}\cup \bar{\*X}_{i}) \right | / \delta)}\sum_{\bar{\*x}_i, \bar{\*s}_i }  \sqrt{N_T\Parens{\bar{\*x}_i, \bar{\*s}_i} }                           \\
		 & +2\sqrt{2\log(8H \left  |\D(\bar{\*S}_{H}\cup \bar{\*X}_{H}) \right | / \delta)} \sum_{\bar{\*x}_H, \bar{\*s}_H }  \sqrt{N_T\Parens{\bar{\*x}_H, \bar{\*s}_H}}
	\end{align}\endgroup
	The last step follows from $\sum_{n = 1}^{\infty} \frac{1}{\sqrt{n}} < 2\sqrt{n} - 1$. Eq.~\ref{eq:_app_5_ucb_dtr4} could thus be written as
	\begin{align}
		\Delta(\epsilon, T) & \leq \sum_{i = 0}^{H-1} 4\sqrt{2 \left |\D(\*S_{i+1}) \right | \log(4H\left |\D(\bar{\*S}_{i}\cup \bar{\*X}_{i}) \right | / \delta)}  \sum_{\bar{\*x}_i, \bar{\*s}_i} \sqrt{N_T\Parens{\bar{\*x}_i, \bar{\*s}_i}} \\
		                    & +2\sqrt{2\log(8H \left  |\D(\bar{\*S}_{H}\cup \bar{\*X}_{H}) \right | / \delta)}  \sum_{\bar{\*x}_H, \bar{\*s}_H} \sqrt{N_T\Parens{\bar{\*x}_H, \bar{\*s}_H}}
	\end{align}
	Let $T_{\epsilon} = \sum_{t = 1}^T  \I\Braces{\Delta_{\pi^{(t)}} > \epsilon}$. Applying Jensen's inequality gives
	\begin{align}
		\Delta(\epsilon, T) & \leq \sum_{i = 0}^{H-1} 4\sqrt{2 \left |\D(\*S_{i+1}) \right | \log(4H\left |\D(\bar{\*S}_{i}\cup \bar{\*X}_{i}) \right | / \delta)}\sqrt{ \left |\D(\bar{\*S}_{i}\cup \bar{\*X}_{i}) \right |T_{\epsilon}} \\
		                    & +2\sqrt{2 \log(8H \left  |\D(\bar{\*S}_{H}\cup \bar{\*X}_{H}) \right | / \delta)} \sqrt{\left  |\D(\bar{\*S}_{H}\cup \bar{\*X}_{H}) \right |T_{\epsilon}}
	\end{align}
	Let $\delta = t^{-4}$. A few simplifications give
	\begin{align}
		\Delta(\epsilon, T) \leq 17 H\sqrt{ T_{\epsilon} \left|\D(\bar{\*S}_{H}\cup \bar{\*X}_{H}) \right |  \log(T) } \label{eq:_app_5_ucb_dtr5}
	\end{align}
	Observe that the algorithm only selects deterministic policies in $\Pi^{\mathrm{o}}$. This implies that for any policy $\pi^{(t)}$ such that $\Delta_{\pi^{(t)}} > \epsilon$, its performance gap is lower bounded by that of the best possible deterministic policy $\pi \in \Pi^{\mathrm{o}}$ with $\Delta_{\pi} > \epsilon$. This gives
	\begin{align}
		\min_{\pi \in \Pi^{\mathrm{o}}: \Delta_{\pi} > \epsilon} \Delta_{\pi} T_{\epsilon} \leq \Delta(\epsilon, T)\label{eq:_app_5_ucb_dtr6}
	\end{align}
	Eqs.~\ref{eq:_app_5_ucb_dtr5} and \ref{eq:_app_5_ucb_dtr6} imply
	\begin{align}
		T_{\epsilon} \leq \max_{\pi \in \Pi^{\mathrm{o}}: \Delta_{\pi} > \epsilon} \frac{17^2 H^2 \left|\D(\bar{\*S}_{H}\cup \bar{\*X}_{H}) \right |  \log(T)}{\Delta^2_{\pi}} \label{eq:_app_5_ucb_dtr7}
	\end{align}
	Note that $\bar{\*S}_H = \*S$ and $\bar{\*X}_H = \*X$. Eqs.~\ref{eq:_app_5_ucb_dtr5} and \ref{eq:_app_5_ucb_dtr7} imply
	\begin{align}
		\Delta(\epsilon, T) \leq \max_{\pi \in \Pi^{\mathrm{o}}: \Delta_{\pi} > \epsilon} \frac{17^2 H^2 \left|\D(\*S \cup \*X )\right |  \log(T)}{\Delta_{\pi}} \label{eq:_app_5_ucb_dtr_term3}
	\end{align}
	By Eqs.~\ref{eq:_app_5_ucb_dtr_term1}, \ref{eq:_app_5_ucb_dtr_term2} and \ref{eq:_app_5_ucb_dtr_term3}, we have
	\begin{align}
		R(T, \1M^*) \leq \max_{\pi \in \Pi^{\mathrm{o}}: \Delta_{\pi} > \epsilon} \frac{17^2 H^2 \left|\D(\*S \cup \*X )\right |  \log(T)}{\Delta_{\pi}} + \max_{\pi \in \Pi^{\mathrm{o}}: \Delta_{\pi} \leq \epsilon }\Delta{\pi} T + \frac{\pi^2}{6}
	\end{align}
	This completes the proof.
\end{proof}

\begin{lemma}\label{lem:_app_5_cbound1}
	Let $\langle \1M^*, \Pi, \1R \rangle$ be a CDM where the policy space $\Pi = \left \{\langle X_i, \*S_i \rangle \right \}_{X_i \in \*X}$. For every $i = 1, \dots, H-1$,
	\begin{align}
		 & \inv{\bar{\*s}_{i+1}}{{\bar{\*x}_i}} - \inv{\bar{\*s}_i}{\bar{\*x}_i} \leq P(\bar{\*s}_{i+1}, \bar{\*x}_i) - P(\bar{\*s}_i, \bar{\*x}_i).
	\end{align}
\end{lemma}
\begin{proof}
	For any set of variables $\*V$, let $\neg \*v$ denote an event $\*V \neq \*v$. Note that $\inv{\bar{\*s}_{i+1}}{\bar{\*x}_i}$ can be written as the counterfactual distribution $P(\bar{\*s}_{i+1_{\bar{\*x}_i}})$. We thus have
	\begin{align}
		\inv{\bar{\*s}_{i+1}}{\bar{\*x}_i} = P(\bar{\*s}_{i+1_{\bar{\*x}_i}}, \bar{\*x}_i) + P(\bar{\*s}_{i+1_{\bar{\*x}_i}}, \neg x_i, \bar{\*x}_{i-1}) + P(\bar{\*s}_{i+1_{\bar{\*x}_i}}, \neg \bar{\*x}_{i-1}),
	\end{align}
	By the composition axiom (Thm.~\ref{thm:_app_3_ctf}),  $\bar{\*S}_{i+1_{\bar{\*x}_i}}(\*u) = \bar{\*s}_{i+1}$ if $\bar{\*X}_i(\*u) = \bar{\*x}_i$. So,
	\begingroup\allowdisplaybreaks\begin{align*}
		\inv{\bar{\*s}_{i+1}}{\bar{\*x}_i} & = P(\bar{\*s}_{i+1}, \bar{\*x}_i) + P(\bar{\*s}_{i+1_{\bar{\*x}_i}}, \neg x_i, \bar{\*x}_{i-1}) + P(\bar{\*s}_{i+1_{\bar{\*x}_i}}, \neg \bar{\*x}_{i-1})                                                            \\
		                                   & \leq P(\bar{\*s}_{i+1}, \bar{\*x}_i) + P(\bar{\*s}_{i_{\bar{\*x}_i}}, \neg x_i, \bar{\*x}_{i-1}) + P(\bar{\*s}_{i_{\bar{\*x}_i}}, \neg \bar{\*x}_{i-1})                                                             \\
		                                   & =P(\bar{\*s}_{i+1}, \bar{\*x}_i) + P(\bar{\*s}_{i_{\bar{\*x}_i}}, \bar{\*x}_{i-1}) - P(\bar{\*s}_{i_{\bar{\*x}_i}}, \bar{\*x}_i) + P(\bar{\*s}_{i_{\bar{\*x}_i}}) - P(\bar{\*s}_{i_{\bar{\*x}_i}}, \bar{\*x}_{i-1}) \\
		                                   & = P(\bar{\*s}_{i_{\bar{\*x}_i}}) + P(\bar{\*s}_{i+1}, \bar{\*x}_i)- P(\bar{\*s}_{i_{\bar{\*x}_i}}, \bar{\*x}_i)                                                                                                     \\
		                                   & = P_{\bar{\*x}_i}(\bar{\*s}_i) + P(\bar{\*s}_{i+1}, \bar{\*x}_i)- P(\bar{\*s}_{i_{\bar{\*x}_i}}, \bar{\*x}_i).
	\end{align*}\endgroup
	The last step holds sine $P(\bar{\*s}_{i_{\bar{\*x}_i}})=P_{\bar{\*x}_i}(\bar{\*s}_i)$. Since $\bar{\*S}_{i_{\bar{\*x}_i}}(\*u) = \bar{\*S}_i$ if $\bar{\*X}_i(\*u) = \bar{\*x}_i$, applying the composition axiom  (Thm.~\ref{thm:_app_3_ctf}) again gives
	\begin{align}
		\inv{\bar{\*s}_{i+1}}{\bar{\*x}_i} \leq P_{\bar{\*x}_i}(\bar{\*s}_i) + P(\bar{\*s}_{i+1}, \bar{\*x}_i)- P(\bar{\*s}_i, \bar{\*x}_i)
	\end{align}
	Rearranging the above equation completes the proof.
\end{proof}

\begin{lemma}\label{lem:_app_5_cbound2}
	Let $\langle \1M^*, \Pi, \1R \rangle$ be a CDM where the policy space $\Pi = \left \{\langle X_i, \*S_i \rangle \right \}_{X_i \in \*X}$. For every $i = 1, \dots, H-1$, $\inv{\bar{\*s}_{i+1}}{\bar{\*x}_i} \leq \Gamma(\bar{\*s}_{i+1}, \bar{\*x}_i)$ where $\Gamma(\bar{\*s}_{i+1}, \bar{\*x}_i)$ is a function defined in Eq.~\ref{eq:_5_3_gamma}.
\end{lemma}
\begin{proof}
	We will prove this statement by induction.
	\paragraph{Base Case: $i = 0$}By definition, $\Gamma(s_1) = P(s_1)$. We thus have $P(s_1) \leq \Gamma(s_1)$.
	\paragraph{Induction Step}Assume that the statement holds for $k$, i.e., $P_{\bar{\*x}_i}(\bar{\*s}_{i+1}) \leq \Gamma(\bar{\*s}_{i+1}, \bar{\*x}_i)$. We will next show that the statement holds for $i + 1$, i.e.,
	\begin{align}
		\inv{\bar{\*s}_{i+2}}{\bar{\*x}_{i+1}} \leq \Gamma(\bar{\*s}_{i+2}, \bar{\*x}_{i+1})
	\end{align}
	To begin with,
	\begin{align}
		P_{\bar{\*x}_{i+1}}(\bar{\*s}_{i+2}) = P_{\bar{\*x}_{i+1}}(\bar{\*s}_{i+2}) - P_{\bar{\*x}_{i+1}}(\bar{\*s}_{i+1}) + P_{\bar{\*x}_{i+1}}(\bar{\*s}_{i+1}).
	\end{align}
	Applying Lem.~\ref{lem:_app_5_cbound1} gives
	\begin{align}
		P_{\bar{\*x}_{i+1}}(\bar{\*s}_{i+2}) & \leq P(\bar{\*s}_{i+2}, \bar{\*x}_{i+1}) - P(\bar{\*s}_{i+1}, \bar{\*x}_{i+1}) + P_{\bar{\*x}_{i+1}}(\bar{\*s}_{i+1}) \\
		                                     & \leq P(\bar{\*s}_{i+2}, \bar{\*x}_{i+1}) - P(\bar{\*s}_{i+1}, \bar{\*x}_{i+1}) + P_{\bar{\*x}_i}(\bar{\*s}_{i+1})
	\end{align}
	The last step holds since $\bar{\*S}_{i+1}$ are non-descendants of $X_{i+1}$. By the induction assumption, $P_{\bar{\*x}_i}(\bar{\*s}_{i+1}) \leq \Gamma(\bar{\*s}_{i+1}, \bar{\*x}_i)$. This implies
	\begin{align}
		P_{\bar{\*x}_{i+1}}(\bar{\*s}_{i+2}) & \leq P(\bar{\*s}_{i+2}, \bar{\*x}_{i+1}) - P(\bar{\*s}_{i+1}, \bar{\*x}_{i+1}) + \Gamma(\bar{\*s}_{i+1}, \bar{\*x}_i) \\
		                                     & \leq \Gamma(\bar{\*s}_{i+2}, \bar{\*x}_{i+1}).
	\end{align}
\end{proof}

\thmcboundt*
\begin{proof}
	By the definition of the conditional distribution,
	\begin{align}
		P_{\bar{\*x}_i}(s_{i+1} | \bar{\*s}_i) = {P_{\bar{\*x}_i}(\bar{\*s}_{i+1}) \over P_{\bar{\*x}_i}(\bar{\*s}_i) }.
	\end{align}
	Applying Lem.~\ref{lem:_app_5_cbound1} gives:
	\begin{align}
		P_{\bar{\*x}_i}(s_{k+1} | \bar{\*s}_i) & \leq 1 + {P(\bar{\*s}_{i+1}, \bar{\*x}_i) - P(\bar{\*s}_i, \bar{\*x}_i) \over P_{\bar{\*x}_i}(\bar{\*s}_i) }     \\
		                                       & \leq 1 + {P(\bar{\*s}_{i+1}, \bar{\*x}_i) - P(\bar{\*s}_i, \bar{\*x}_i) \over P_{\bar{\*x}_{i-1}}(\bar{\*s}_i) }
	\end{align}
	The last step holds since $\bar{\*S}_i$ are non-descendants of action $X_{i}$.
	Among the above quantities, the numerator $P(\bar{\*s}_{i+1}, \bar{\*x}_i) - P(\bar{\*s}_i, \bar{\*x}_i) \leq 0$, since the marginal probability is not larger than its joint. Therefore, the above upper bound is maximized when the denominator $P_{\bar{\*x}_i}(\bar{\*s}_i)$ is the maximal. By applying Lem.~\ref{lem:_app_5_cbound2}, we obtain the upper bound
	\begin{align}
		P_{\bar{\*x}_i}(s_{k+1} | \bar{\*s}_i) & \leq 1 + {P(\bar{\*s}_{i+1}, \bar{\*x}_i) - P(\bar{\*s}_i, \bar{\*x}_i) \over \Gamma(\bar{\*s}_i, \bar{\*x}_{i-1}) } \\
		                                       & \leq {\Gamma(\bar{\*s}_{i+1}, \bar{\*x}_i ) \over \Gamma(\bar{\*s}_i, \bar{\*x}_{i-1} )}.
	\end{align}

	We will now derive the lower bound. By the definition of counterfactual distributions (Def.~\ref{def:_app_3_ctf}),
	\begin{align}
		P_{\bar{\*x}_i}(\bar{\*s}_{i+1}) = P(\bar{\*s}_{{i+1}_{\bar{\*x}_i}})
	\end{align}
	By expanding over the domain of observed actions $\bar{\*X}_i$,
	\begin{align}
		P_{\bar{\*x}_i}(s_{k+1} | \bar{\*s}_i) & = {P(\bar{\*s}_{{i+1}_{\bar{\*x}_i}}, \bar{\*x}_i) + P(\bar{\*s}_{{i+1}_{\bar{\*x}_i}}, \neg \bar{\*x}_i) \over P_{\bar{\*x}_i}(\bar{\*s}_i) } \\
		                                       & \geq {P(\bar{\*s}_{{i+1}_{\bar{\*x}_i}}, \bar{\*x}_i) \over P_{\bar{\*x}_i}(\bar{\*s}_i) }.
	\end{align}
	Since $\bar{\*S}_{i+1_{\bar{\*x}_i}}(\*u) = \bar{\*S}_{i+1}$ if $\bar{\*X}_i(\*u) = \bar{\*x}_i$, the composition axiom (Thm.~\ref{thm:_app_3_ctf}) implies
	\begin{align}
		P_{\bar{\*x}_i}(s_{k+1} | \bar{\*s}_i) \geq {P(\bar{\*s}_{i+1}, \bar{\*x}_i) \over P_{\bar{\*x}_i}(\bar{\*s}_i) }
	\end{align}
	By applying Lem.~\ref{lem:_app_5_cbound2} again, we obtain the lower bound
	\begin{align}
		P_{\bar{\*x}_i}(s_{k+1} | \bar{\*s}_i) \geq  {P(\bar{\*s}_{i+1}, \bar{\*x}_i) \over \Gamma(\bar{\*s}_i, \bar{\*x}_{i-1} )}.
	\end{align}
	This completes the proof.
\end{proof}

\thmcboundr*
\begin{proof}
	By basic probabilistic operations,
	\begin{align}
		E_{\bar{\*x}_H}[Y \mid \bar{\*s}_H] = {E_{\bar{\*x}_H}[\1R(\*Y) \mid \bar{\*s}_H]P_{\bar{\*x}_H}(\bar{\*s}_H) \over P_{\bar{\*x}_H}(\bar{\*s}_H) }  \label{eq:_app_5_cboundr2}
	\end{align}
	Note that the reward function $\1R(\*y) \in [0, 1]$. By following a similar argument as Lem.~\ref{lem:_app_5_cbound1}, we could show that
	\begin{align}
		E_{\bar{\*x}_H}\left[\1R(\*Y) \mid \bar{\*s}_H \right]P_{\bar{\*x}_H}(\bar{\*s}_H) - P_{\bar{\*x}_H}(\bar{\*s}_H) \leq E \left[\1R(\*Y) \mid  \bar{\*s}_H, \bar{\*x}_H \right] P(\bar{\*s}_H, \bar{\*x}_H) - P(\bar{\*s}_H, \bar{\*x}_H). \label{eq:_app_5_cboundr1}
	\end{align}
	Rearranging Eq.~\ref{eq:_app_5_cboundr1} gives
	\begin{align}
		E_{\bar{\*x}_H}\left[\1R(\*Y) \mid  \bar{\*s}_H \right] & \leq 1 - {\left (1 - E\left [\1R(\*Y) \mid \bar{\*s}_H, \bar{\*x}_H \right] \right)P(\bar{\*s}_H, \bar{\*x}_H) \over P_{\bar{\*x}_H}(\bar{\*s}_H)}     \\
		                                                        & \leq 1 - {\left (1 - E\left [\1R(\*Y) \mid \bar{\*s}_H, \bar{\*x}_H \right] \right)P(\bar{\*s}_H, \bar{\*x}_H) \over P_{\bar{\*x}_{H-1}}(\bar{\*s}_H)}
	\end{align}
	The last step holds since $\bar{\*S}_H$ are non-descendants of $X_H$. Since $E \left[\1R(\*Y) \mid \bar{\*s}_H, \bar{\*x}_H \right] \leq 1$, the above equation is maximized when $P_{\bar{\*x}_H}(\bar{\*s}_H)$ is the maximal. Applying Lem.~\ref{lem:_app_5_cbound2} gives the upper bound
	\begin{align}
		E_{\bar{\*x}_H}\left [\1R(\*Y) \mid  \bar{\*s}_H \right]  \leq 1 - {\left (1 - E\left[\1R(\*Y) \mid \bar{\*s}_H, \bar{\*x}_H \right] \right) P(\bar{\*s}_H, \bar{\*x}_H) \over \Gamma(\bar{\*s}_k, \bar{\*x}_{k-1} )}.
	\end{align}

	We will next derive the lower bound. By the definition of counterfactual distributions (Def.~\ref{def:_app_3_ctf}),
	\begin{align}
		P_{\bar{\*x}_H}(\*y, \bar{\*s}_H) = P(\*y_{\bar{\*x}_H}, \bar{\*s}_{K_{\bar{\*x}_{H-1}}}) = P(\*y_{\bar{\*x}_H}, \bar{\*s}_{K_{\bar{\*x}_{H}}})
	\end{align}
	The last step holds since $\bar{\*S}_H$ are non-descendants of $X_H$. By basic probabilistic operations,
	\begin{align}
		E_{\bar{\*x}_H}\left [\1R(\*Y) \mid  \bar{\*s}_H \right] & = \frac{\sum_{\*y} \1R(\*y) P \left(\*y_{\bar{\*x}_H}, \bar{\*s}_{H_{\bar{\*x}_{H-1}}} \right) }{P_{\bar{\*x}_{H-1}}(\bar{\*s}_H)}                \\
		                                                         & \geq \frac{\sum_{\*y} \1R(\*y) P \left(\*y_{\bar{\*x}_H}, \bar{\*s}_{H_{\bar{\*x}_{H-1}}}, \bar{\*x}_H \right)}{P_{\bar{\*x}_{H-1}}(\bar{\*s}_H)} \\
		                                                         & \geq \frac{ \sum_{\*y} \1R(\*y) P \left(\*y, \bar{\*s}, \bar{\*x}_H \right) }{P_{\bar{\*x}_{H-1}}(\bar{\*s}_H)}
	\end{align}
	The last follows from the composition axiom (Thm.~\ref{thm:_app_3_ctf}), $\*Y_{\bar{\*x}_H}(\*u) = \*Y(\*u), \bar{\*S}_{H_{\bar{\*x}_{H-1}}}(\*u)= \bar{\*S}_{H-1}$ if $\bar{\*X}_H(\*u) = \bar{\*x}_H$. Applying Lem.~\ref{lem:_app_5_cbound2} gives the lower bound
	\begin{align}
		E_{\bar{\*x}_H}\left [\1R(\*Y) \mid  \bar{\*s}_H \right] & \geq \frac{ \sum_{\*y} \1R(\*y) P \left(\*y, \bar{\*s}, \bar{\*x}_H \right) }{ \Gamma(\bar{\*s}_H, \bar{\*x}_{H-1} ) } \\
		                                                         & = {E[\1R(\*Y) \mid \bar{\*s}_H, \bar{\*x}_H]P(\bar{\*s}_H, \bar{\*x}_H) \over \Gamma(\bar{\*s}_H, \bar{\*x}_{H-1} )}
	\end{align}
	This completes the proof.
\end{proof}

\thmucbdtrplus*
\begin{proof}
	Let $\epsilon \geq 0$ be an arbitrary real value. The sum of performance gaps is bounded as follows.
	\begin{align}
		\sum_{t = 1}^T \Delta_{\pi^{(t)}} & \leq  \underbrace{\sum_{t = 1}^T\I\Braces{\M^* \not \in \3M_{t-1}(\delta) \cap \3M_c}}_{\text{Term 1}} + \underbrace{\sum_{t = 1}^T \Delta_{\pi^{(t)}} \I\Braces{\M^* \in \3M_{t-1}(\delta) \cap \3M_c }}_{\text{Term 2}}
	\end{align}
	Among the quantities in the above equation, Term 1 is bounded following Eq.~\ref{eq:_app_5_ucb_dtr_term1},
	\begin{align}
		\sum_{t = 1}^T P\Parens{\M^* \not \in \3M_{t-1}(\delta) \cap \3M_c} \leq \sum_{t = 1}^T P\Parens{\M^* \not \in \3M_{t-1}(\delta)} \leq \frac{\pi^2}{6} \label{eq:_app_5_ucb_dtr_plus_term1}
	\end{align}
	Let $T_{0} = \sum_{t = 1}^T \I\{\Delta_{\pi^{(t)}} > 0\}$ be the total number of episodes where a suboptimal policy is picked. Let $\Delta(T)$ denote Term 2, which could be bounded following Eq.~\ref{eq:_app_5_ucb_dtr5}
	\begin{align}
		\Delta(T) \leq 17 H \sqrt{ T_0 \left|\D(\bar{\*S}_{H}\cup \bar{\*X}_{H}) \right |  \log(T) } \label{eq:_app_5_ucb_dtr_plus_1}
	\end{align}
	Observe that when $\M^* \in \3M_{t-1}(\delta) \cap \3M_c$, it follows from the optimistic planning in Eq.~\ref{eq:_5_3_optimize} that a suboptimal policy $\pi$ with $r_{\pi} < \mu^*$ is never selected. This gives
	\begin{align}
		\Delta(T) \geq \min_{\pi \in \Pi^{\mathrm{o}}: \Delta_{\pi} > 0, r_{\pi} \geq \mu^*} \Delta_{\pi} T_0 \label{eq:_app_5_ucb_dtr_plus_2}
	\end{align}
	Eqs.~\ref{eq:_app_5_ucb_dtr_plus_1} and \ref{eq:_app_5_ucb_dtr_plus_2}
	\begin{align}
		T_0 \leq \max_{\pi \in \Pi^{\mathrm{o}}: \Delta_{\pi} > \epsilon, r_{\pi} \geq \mu^*} \frac{17^2 H^2 \left|\D(\*S \cup \*X )\right |  \log(T)}{\Delta^2_{\pi}} \label{eq:_app_5_ucb_dtr_plus_3}
	\end{align}
	Eqs.~\ref{eq:_app_5_ucb_dtr_plus_1} and \ref{eq:_app_5_ucb_dtr_plus_3} imply
	\begin{align}
		\Delta(T) \leq \max_{\pi \in \Pi^{\mathrm{o}}: \Delta_{\pi} > \epsilon, r_{\pi} \geq \mu^*} \frac{17^2 H^2 \left|\D(\*S \cup \*X )\right |  \log(T)}{\Delta_{\pi}} \label{eq:_app_5_ucb_dtr_plus_term2}
	\end{align}
	By Eqs.~\ref{eq:_app_5_ucb_dtr_plus_term1} and \ref{eq:_app_5_ucb_dtr_plus_term2}, we have
	\begin{align}
		R(T, \1M^*) = \E \Brackets{\sum_{t = 1}^T \Delta_{\pi^{(t)}}} \leq \max_{\pi \in \Pi^{\mathrm{o}}: \Delta_{\pi} > \epsilon, r_{\pi} \geq \mu^*} \frac{17^2 H^2 \left|\D(\*S \cup \*X )\right |  \log(T)}{\Delta_{\pi}} + \frac{\pi^2}{6}
	\end{align}
	This completes the proof.
\end{proof}
\section{Proofs for Chapter 8}\label{app:_sec_8}
\thmbcnuc*
\begin{proof}
	Since the NUC condition holds, applying the IPW formula in Thm.~\ref{thm:_4_1_ipw} implies,  for any policy $\pi \in \Pi$,
	\begin{align}
		\E_{\pi}\left[\1R(\*Y)\right] = \sum_{\bar{\*x}_H, \bar{\*s}_H} E\left[\1R(\*Y) \mid \bar{\*x}_H, \bar{\*s}_H \right] P\left (\bar{\*x}_H, \bar{\*s}_H \right)\prod_{i = 1}^H \frac{\pi_{i} \left(x_i \mid \*s_i\right)}{P\left( x_i \mid \bar{\*x}_{i-1}, \bar{\*s}_i \right)} \label{eq:_app_8_nuc1}
	\end{align}
	Since the local Markov property holds, we must have
	\begin{align}
		P\left( x_i \mid \bar{\*x}_{i-1}, \bar{\*s}_i \right) = P\left( x_i \mid \*s_i \right) \label{eq:_app_8_nuc2}
	\end{align}
	Eqs.~\ref{eq:_app_8_nuc1} and \ref{eq:_app_8_nuc2} imply
	\begin{align}
		\E_{\pi}\left [\1R(\*Y)\right] = \sum_{\bar{\*x}_H, \bar{\*s}_H} E\left[\1R(\*Y) \mid \bar{\*x}_H, \bar{\*s}_H \right] P\left (\bar{\*x}_H, \bar{\*s}_H \right)\prod_{i = 1}^H \frac{\pi_{i} \left(x_i \mid \*s_i\right)}{P\left( x_i \mid \*s_i \right)}
	\end{align}
	Let policy $\pi$ be such that $\pi_i(X_i \mid \*S_i) = P(X_i \mid \*S_i)$ for every $i =1, \dots, H$. The expected reward of $\pi$ could be further written as
	\begin{align}
		\E_{\pi}\left [\1R(\*Y)\right] & = \sum_{\bar{\*x}_H, \bar{\*s}_H} E\left[\1R(\*Y) \mid \bar{\*x}_H, \bar{\*s}_H \right] P\left (\bar{\*x}_H, \bar{\*s}_H \right)\prod_{i = 1}^H \frac{P\left( x_i \mid \*s_i \right)}{P\left( x_i \mid \*s_i \right)} \\
		                               & = \E\left[\1R(\*Y)\right]
	\end{align}
	That is, behavioral cloning perfectly imitates the expert's performance.
\end{proof}

\thmibc*
\begin{proof}
	We first show the sufficiency of the imitation backdoor criterion (Def.~\ref{def:_8_1_ibc}). Consider the following claim, for every $i = 1, \dots, H$,
	\begin{align}
		\inv{\*Y}{\pi_i, \dots, \pi_H} = \inv{\*Y}{\pi_{i+1}, \dots, \pi_H} \label{eq:_app_8_ibc1}
	\end{align}
	If $X_i \not \in \An(\*Y)$ is not an ancestor of nodes in $\*Y$ in the manipulated diagram $\G_{\pi_i, \dots, \pi_H}$, Eq.~\ref{eq:_app_8_ibc1} trivially holds. We now consider the second condition, $X_i \in \An(\*Y)$ in $\G_{\pi_i, \dots, \pi_H}$. By conditioning on action $X_i$ and covariates $\*Z_i$,
	\begin{align}
		\inv{\*y}{\pi_i, \dots, \pi_H} & = \sum_{x_i, \*z_i} \inv{\*y, \*z_i}{x_i, \pi_{i+1}, \dots, \pi_H} \pi_i(x_i \mid \*z_i)            \\
		                               & =\sum_{x_i, \*z_i} \inv{\*y\mid \*z_i}{x_i, \pi_{i+1}, \dots, \pi_H} \pi_i(x_i \mid \*z_i) P(\*z_i)
	\end{align}
	The last step holds since $\*Z_i$ are non-descendants of actions $X_i, \dots, X_H$ in the manipulated diagram $\G_{\pi_{i}, \dots, \pi_H}$. Since $\Pi'$ is imitation admissible, covariates $\*Z_i$ block all backdoor paths from node $X_i$ to $\*Y$ in $\G_{\pi_{i}, \dots, \pi_H}$, i.e.,
	\begin{align}
		\Parens{\*Y \ci X_i \mid \*Z_i}_{\G_{\underline{X_i}, \pi_{i+1}, \dots, \pi_H}}
	\end{align}
	Applying Rule 2 of do-calculus (Def.~\ref{def:_4_3_do-calculus}) implies
	\begin{align}
		\inv{\*y\mid \*z_i}{x_i, \pi_{i+1}, \dots, \pi_H} = \inv{\*y\mid x_i, \*z_i}{\pi_{i+1}, \dots, \pi_H} \label{eq:_app_8_ibc2}
	\end{align}
	By Eqs.~\ref{eq:_app_8_ibc1} and \ref{eq:_app_8_ibc2} we have
	\begin{align}
		\inv{\*y}{\pi_i, \dots, \pi_H} & =\sum_{x_i, \*z_i} \inv{\*y\mid x_i, \*z_i}{\pi_{i+1}, \dots, \pi_H} \pi_i(x_i \mid \*z_i) P(\*z_i)
	\end{align}
	Let the decision rule $\pi_i(X_i \mid \*Z_i) = P(X_i \mid \*Z_i)$. The above equation could be written as
	\begin{align}
		\inv{\*y}{\pi_i, \dots, \pi_H} & =\sum_{x_i, \*z_i} \inv{\*y\mid x_i, \*z_i}{\pi_{i+1}, \dots, \pi_H} P(x_i \mid \*z_i) P(\*z_i) \\
		                               & = \inv{\*y}{\pi_{i+1}, \dots, \pi_H}
	\end{align}
	Now, repeatedly applying Eq.~\ref{eq:_app_8_ibc1} for every $i = 1, \dots, H$ gives
	\begin{align}
		\inv{\*y}{\pi} = \inv{\*y}{\pi_1, \dots, \pi_H} = \inv{\*y}{\pi_2, \dots, \pi_H} = \cdots = P(\*y)
	\end{align}
	That is, policy $\pi$ perfectly imitates the expert's performance.
\end{proof}

\propmwal*
\begin{proof}
	Note that $\pi$ is a \emph{mixed policy} which is a probability distribution over deterministic policies compatible with space $\Pi'$. With a slight abuse of notation, we will treat $\pi$ as a vector where $\pi(i)$ is the probability assigned to the $i$-th deterministic policy $\pi^{(i)}$. The claimed statement follows immediately from the definition of the game matrix $\*G$.
\end{proof}

\propgail*
\begin{proof}
	Let the vector $\alpha = \rho - \rho_{\*\pi}$. The claimed statement follows from the definition of the conjugate function $\psi^*$.
\end{proof}

\thmminimalibc*
\begin{proof}
	For any $i = 1, \dots, N$, let $\pi_{i:H}$ be a sequence of decision rules $\{\pi_{i}, \pi_{i+1}, \dots, \pi_{H}\}$. It is sufficient to show that for any $i = 1, \dots, n$,
	\begin{align}
		\inv{\*y, \bar{\*x}_{i-1}, \bar{\*z}_{i-1}}{\pi_{i:H}}
		= \sum_{x_i, \*z_i} \inv{\*y, \bar{\*x}_i, \bar{\*z}_i}{\pi_{i+1:H}} \frac{\pi_i(x_i \mid \*z_i)}{P(x_i \mid \bar{\*x}_{i-1}, \bar{\*z}_{i})}
	\end{align}
	By basic probabilistic operations,
	\begin{align}
		\inv{\*y, \bar{\*x}_{i-1}, \bar{\*z}_{i-1}}{\pi_{i:H}} & = \sum_{x_i, \*z_i}  \inv{\*y, \bar{\*x}_{i}, \bar{\*z}_{i}}{\pi_{i:H}}                                                                                                             \\
		                                                       & =\sum_{x_i, \*z_i} \inv{\*y \mid \bar{\*x}_{i}, \bar{\*z}_{i}}{\pi_{i:H}}  \inv{x_i \mid \bar{\*x}_{i-1}, \bar{\*z}_{i}}{\pi_{i:H}} \inv{\bar{\*x}_{i-1}, \bar{\*z}_{i}}{\pi_{i:H}} \\
		                                                       & =\sum_{x_i, \*z_i} \inv{\*y \mid \bar{\*x}_{i-1}, \bar{\*z}_{i}}{x_i, \pi_{i+1:H}} \pi_i(x_i \mid \*z_{i}) P(\bar{\*x}_{i-1}, \bar{\*z}_{i})
	\end{align}
	In the last step, $\inv{\bar{\*x}_{i-1}, \bar{\*z}_{i}}{\pi_{i:H}} = P(\bar{\*x}_{i-1}, \bar{\*z}_{i})$ since every causal diagram $\G_{\pi_{i:H}}$ is acyclic, and $\bar{\*X}_{i-1}, \bar{\*Z}_{i}$ are non-descendants of actions $X_{i}, \dots, X_{H}$. Next, we will show that
	\begin{align}
		\inv{\*y \mid \bar{\*x}_{i-1}, \bar{\*z}_{i}}{x_i, \pi_{i+1:H}} = \inv{\*y \mid \bar{\*x}_{i}, \bar{\*z}_{i}}{\pi_{i+1:H}} \label{eq:_app_8_min_ibc1}
	\end{align}
	This is equivalent to showing that
	\begin{align}
		(\*Y \ci X_i | \bar{\*X}_{i-1}, \bar{\*Z}_{i}) \text{ in } \G_{\underline{X_i}, \pi_{i+1}, \dots, \pi_H} \label{eq:_app_8_min_ibc2}
	\end{align}
	First, by definition of imitation backdoor,
	\begin{align}
		(\*Y \ci X_i | \*Z_i) \text{ in } \G_{\underline{X_i}, \pi_{i+1}, \dots, \pi_H} \label{eq:_app_8_min_ibc3}
	\end{align}
	Suppose Eq.~\ref{eq:_app_8_min_ibc2} does not hold. We must have a node $W \in \bar{\*Z}_{i-1} \cup \{X_i\}$ opening a collider path from $X_i$ to $\*Y$. Assume that there exists a directed path from $W$ to $\*Y$ in the manipulated diagram $\G_{\underline{X_i}, \pi_{i+1}, \dots, \pi_H}$. Then one could construct a backdoor path from node $X_i$ to a node in $\*Y$ via $W$, which contradicts the independence relationship in Eq.~\ref{eq:_app_8_min_ibc3}.

	What remains is to show that there must exist a directed path from $W$ to $\*Y$ in the manipulated diagram $\G_{\underline{X_i}, \pi_{i+1}, \dots, \pi_H}$. Since $\Pi'$ is a minimal imitation admissible, it is verifiable from \cite[Lem.~3.4]{van2014constructing} that $W \in \An(\*Y)$ in $\G_{\pi_{i+1}, \dots, \pi_H}$. Consequently, the only case that $W \not \in \An(\*Y)$ in $\G_{\underline{X_i}, \pi_{i+1}, \dots, \pi_H}$ is when $W \in \An(X_i)$ in $\G_{\pi_{i+1}, \dots, \pi_H}$. Again, this allows us to construct a backdoor path between node $X_i$ and nodes in $\*Y$ via node $W$ in the manipulated diagram $\G_{\pi_{i+1}, \dots, \pi_H}$, which contradicts Eq.~\ref{eq:_app_8_min_ibc3}.
\end{proof}

\end{document}